\PassOptionsToPackage{dvipsnames}{xcolor}
\documentclass[%
    paper=A4,               
    twoside=true,           
    openright,              
    parskip=half,           
    chapterprefix=true,     
    10pt,                   
    headings=normal,        
    bibliography=totoc,     
    listof=totoc,           
    titlepage=on,           
    captions=tableabove,    
    chapterprefix=false,    
    appendixprefix=false,   
    draft=false,            
]{scrreprt}%

\PassOptionsToPackage{utf8}{inputenc}
\usepackage{inputenc}

\usepackage[author={Auteur1}]{pdfcomment}

\newcommand{\thesisTitle}{Physics-informed machine learning: A mathematical framework with applications to time series forecasting}

\newcommand{\thesisName}{Nathan Doumèche}
\newcommand{\thesisSubject}{Distributed Systems}

\newcommand{\thesisDate}{7 juillet 2025}

\usepackage{morewrites}
\usepackage[pass]{geometry}
\usepackage{scrhack}
\usepackage{listings}
\usepackage{tikz}
\usetikzlibrary{shapes.misc, shapes.geometric, positioning, arrows.meta, calc, arrows}
\usepackage{import}
\usepackage[most]{tcolorbox}
\tcbuselibrary{listings}
\usepackage{amssymb}
\usepackage{caption}
\usepackage{subcaption}
\usepackage{adjustbox}
\usepackage{colortbl}
\usepackage{placeins}[section, below]
\usepackage{hyperref}
\usepackage{macros}
\usepackage[defblank]{paralist}  
\usepackage[inline,shortlabels]{enumitem} 
\usepackage{wrapfig}

\usepackage[english]{babel} 
\PassOptionsToPackage{
    figuresep=colon,%
    hangfigurecaption=false,%
    hangsection=true,%
    hangsubsection=true,%
    sansserif=false,%
    configurelistings=true,%
    colorize=full,%
    colortheme=bluemagenta,%
    configurebiblatex=true,%
    bibsys=bibtex,%
    bibfile=main,%
    bibstyle=alphabetic,%
    bibsorting=nty,%
}{cleanthesis}
\usepackage{cleanthesis}

\hypersetup{
    pdftitle={\thesisTitle},    
    pdfsubject={\thesisSubject},
    pdfauthor={\thesisName},    
    plainpages=false,           
    colorlinks=false,           
    pdfborder={0 0 0},          
    breaklinks=true,            
    bookmarksnumbered=true,     %
    bookmarksopen=true          %
}

\definecolor{red}{HTML}{9B0000}
\definecolor{lightred}{HTML}{FF5131}
\definecolor{green}{HTML}{1FAA00}
\definecolor{lightgreen}{HTML}{9CFF57}
\definecolor{purple}{HTML}{7200CA}
\definecolor{blue}{HTML}{0064B7}
\definecolor{verylightgrey}{HTML}{F1F1F1}
\colorlet{ctcolorfootertitlelight}{ctcolorfootertitle!20}

\lstdefinelanguage{diff}{
  morecomment=[f][\color{blue}]{@},
  morecomment=[f][\color{green}]{+},
  morecomment=[f][\color{red}]{-},
  keepspaces=true,
  identifierstyle=\color{black},
}

\lstdefinelanguage{spdiff}{
  morecomment=[s][\color{blue}]{@@}{@@},
  morecomment=[s][\color{blue}]{@rule1@}{@@},
  morecomment=[s][\color{blue}]{@rule2@}{@@},
  morecomment=[f][\color{green}]{+},
  morecomment=[f][\color{red}]{-},
  morekeywords={expression},
  morekeywords={identifier},
  keepspaces=true,
  identifierstyle=\color{black},
}

\newtcolorbox{notice}[1][]{enhanced,
  before skip=3mm,after skip=7mm,
  boxrule=0pt,left=7mm,right=2mm,top=1mm,bottom=1mm,
  colback=ctcolorfootertitlelight,
  sharp corners,
  frame hidden,
  underlay={%
    \path[fill=ctcolorfootertitle,draw=none] (interior.south west) rectangle node[white]{\huge\bfseries !} ([xshift=6mm]interior.north west);
    },
  #1}

\newtcblisting{namedlisting}[2][]{%
  enhanced,
  sharp corners,
  frame hidden,
  boxsep=0pt,
  title={\hspace*{3.5mm}#2},
  titlerule=0pt,
  colbacktitle=ctcolormain!20,
  coltitle=black,
  toptitle=1mm,
  bottomtitle=1mm,
  listing only,
  left=0pt,
  top=0pt,
  bottom=0pt,
  right=0pt,
  toprule=0pt,
  leftrule=0pt,
  bottomrule=0pt,
  rightrule=0pt,
  colframe=ctcolormain,
  listing options={aboveskip=0pt, belowskip=0pt, #1},
}

\tikzset{%
  >=latex,
  edge/.style={
    ->,
    rounded corners=3pt
  },
  node/.style={%
    rectangle,
    rounded corners=1pt,
    draw,
    thick,
    fill=ctcolorgraylighter,
    minimum height=1.5em,
  },
  empty/.style={%
    minimum height=0,
  },
  removed/.style={
    text=red,
    draw=red,
  },
  added/.style={
    text=green,
    draw=green,
  },
  fake/.style={
    densely dotted,
    text=ctcolorgray,
  },
  special/.style={
    trapezium,
    trapezium left angle=60,
    trapezium right angle=120,
    minimum width=2cm,
  },
  entry port/.style={
    alias=this,
    append after command={%
      \pgfextra
        \draw[-(, thick] (this) -- ($(this.north) + (0, 0.25)$);
      \endpgfextra
    },
  },
  exit port/.style={
    alias=this,
    append after command={%
      \pgfextra
        \draw[-o, thick] (this) -- ($(this.south) + (0, -0.35)$);
      \endpgfextra
    },
  },
}

\usepackage{amsthm,amsmath,amsfonts,amssymb}
\usepackage{mathrsfs}
\usepackage{nicefrac}

\newtheorem{thm}{Theorem}[section]
\newtheorem{lem}[thm]{Lemma}
\newtheorem{prop}[thm]{Proposition}
\newtheorem{cor}[thm]{Corollary}

\newtheorem{ex}[thm]{Example}
\newtheorem{remark}[thm]{Remark}
\newtheorem{defi}[thm]{Definition}
\newtheorem{assumption}[thm]{Assumption}

\DeclareOldFontCommand{\bf}{\normalfont\bfseries}{\mathbf}
\DeclareOldFontCommand{\it}{}{\textit}

\newcommand\bx{{\bf x}}
\newcommand\by{{\bf y}}
\newcommand\bX{{\bf X}}

\DeclareMathOperator*{\argmin}{arg\,min}
\DeclareMathOperator*{\oequivalent}{o}
\DeclareMathOperator*{\Oequivalent}{\mathcal{O}}

\newcommand{\appendixlink}{the 
 Appendix}
\newcommand{\supplementary}
{the Supplementary Material \citep{supplement2023}}
\newcommand{\supplementSecTwo}{\citet[][Supplementary Material, Section 2]{supplement2023}}
\newcommand{\supplementSecFive}{\citep[see][Supplementary Material, Section 5]{supplement2023}}

\usepackage{soul}
\usepackage{wrapfig}
\usepackage{booktabs} 
\usepackage[bbgreekl]{mathbbol}

\usepackage{tikz}
\usetikzlibrary{shapes,arrows,positioning}

\tikzstyle{block} = [draw, rectangle, minimum height=3em, minimum width=3em]
\tikzstyle{virtual} = [coordinate]

\newcommand{\lenghtfig}{12}
\begin{document}



\renewcaptionname{english}{\figurename}{Fig.}
\renewcaptionname{english}{\tablename}{Tab.}

\renewcommand*{\lstlistlistingname}{List of Listings}

\pagenumbering{roman}			

%

\newcommand{\jurymember}[5]{\hspace{1.5em}\textbf{{#1} \textsc{{#2}}}, {#3}, {#4} \hfill \textit{{#5}}\\}

\newgeometry{left=2.8cm, right=2.5cm, top=2cm, bottom=2cm}
\begin{titlepage}
	\pdfbookmark[0]{Cover}{Cover}
	\begin{flushleft}
      \includegraphics[height=1.7cm]{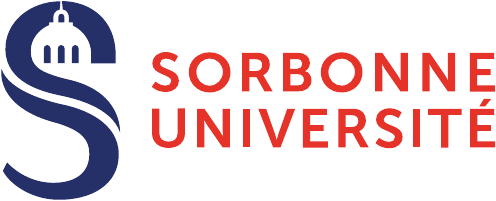}
      \hspace{5em}
      \includegraphics[height=1.7cm, trim={0 3.4cm 0 2cm},clip]{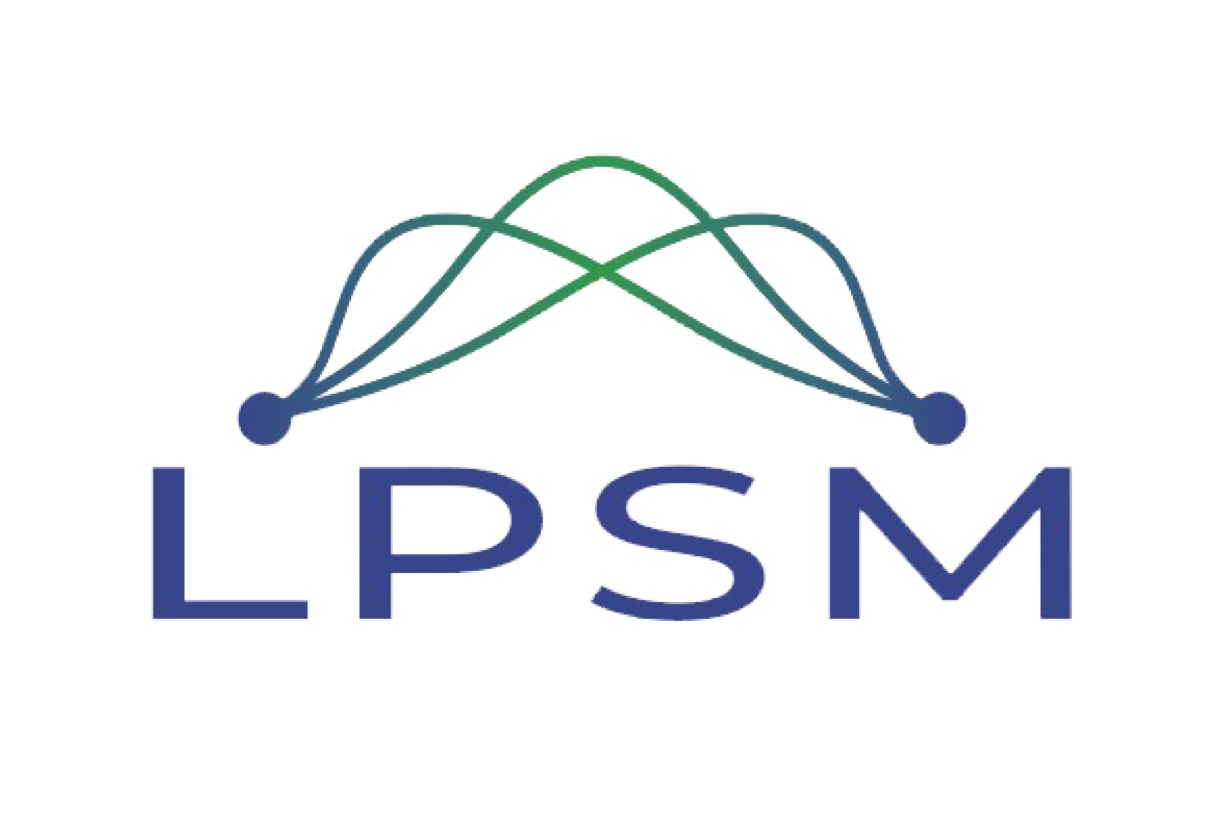}
      \hfill
      \includegraphics[height=1.7cm]{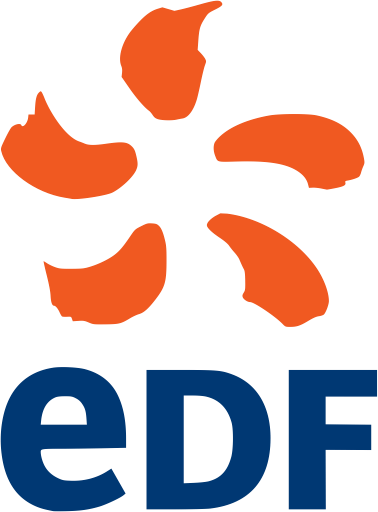}
	\end{flushleft}
	\vfill
	\begin{center}
        \vspace{1em}
        {\large Thèse présentée pour l'obtention du grade de}\\
        \vspace{1em}
        {\LARGE DOCTEUR de SORBONNE UNIVERSITÉ}\\
        \vspace{2em}
		{\large Discipline / Spécialité}\\
		\vspace{0.5em}
		{\Large Mathématiques appliquées / Statistique}\\
		\vspace{2em}
        {\large École doctorale}\\
		\vspace{0.5em}
		{\Large Sciences Mathématiques de Paris Centre (ED 386)}\\
    \end{center}
	\vfill
	\hfill
	\vfill
	{\centering
    {\LARGE\thesisTitle \par}}
	\rule[5pt]{\textwidth}{.4pt} \par
	\begin{flushright}
	{\Large\thesisName}
	\vfill
	\end{flushright}
        
	Soutenue publiquement le \textit{\large\thesisDate}
	\begin{flushleft}
		Devant un jury composé de :\\
    \jurymember{Gilles}{Blanchard}{Professeur}{Université Paris-Saclay}{Rapporteur}
    \jurymember{Richard}{Nickl}{Professeur}{Université de Cambridge}{Rapporteur}
    \jurymember{Gabriel}{Peyré}{Directeur de recherche}{CNRS}{Président du jury}
    \jurymember{Jalal}{Fadili}{Professeur}{ENSICAEN}{Examinateur}
    \jurymember{Mathilde}{Mougeot}{Professeure}{ENSIIE}{Examinatrice}
    \jurymember{Francis}{Bach}{Directeur de recherche}{CNRS}{Membre invité}
    \jurymember{Stéphane}{Tanguy}{Directeur de recherche}{CIO \& CTO à EDF Labs}{Membre invité}
        \jurymember{Gérard}{Biau}{Professeur}{Sorbonne Université}{Directeur de thèse}
        \jurymember{Claire}{Boyer}{Professeure}{Université Paris-Saclay}{Co-directrice de thèse}
        \jurymember{Yannig}{Goude}{Professeur associé}{Université Paris-Saclay}{Encadrant industriel de thèse}
		
	\end{flushleft}
\end{titlepage}
\restoregeometry
\clearpage

\vfill
\begin{flushright}
    \vfill
	\hfill
	\vfill
\end{flushright}

\pagestyle{plain}				
%
%
\pdfbookmark[0]{Remerciements}{Remerciements}
\addchap*{Remerciements}

Trois ans sont donc passés...
D'aucuns supplièrent jadis les heures de suspendre leur cours. Sans succès.
Quant à moi, pour endiguer le vertige qui me submerge à la clôture de ce chapitre de ma vie, je n'ai d'autre artifice que de recourir au rite consacré.
Aussi, j'ai le bonheur teinté de nostalgie d'entamer cette thèse en remerciant ceux qui l'ont rendue possible.

Au commencement était Gérard, et de Gérard naquit cette thèse. 
Voilà cinq ans maintenant que tu me formes avec exigence et me guides avec bienveillance.
Véritable amoureux de la statistique mathématique, tu partages avec générosité ton entrain scientifique et ta vision pour une rédaction scientifique réellement pédagogique.
Travailleur acharné, insatisfait de tes rôles de chercheur chevronné et d'enseignant apprécié, tu prends aussi de ton temps pour façonner la statistique de demain et représenter la Sorbonne.
J'ose donc affirmer que, fidèle au poète qui t'est cher, tu incarnes bien \textit{l'homme des utopies, les pieds ici, les yeux ailleurs.}
Pour tout cela, je suis heureux et fier d'avoir été ton élève.

Deuxième pilier de cette thèse, il me faut remercier Claire. 
Tu harmonises rigueur et curiosité scientifique par un plaisir solaire à apprendre, découvrir et écrire.
Aventurière des statistiques, tu m'as suivi sur tous les chemins, des réseaux de neurones à la mesurabilité des estimateurs, aux noyaux et aux EDP. 
C'est sans relâche que tu entretiens ton goût du verbe et de la bonne formule.
Ton empathie, ton sang-froid et ton humour m'ont aidé à faire face aux moments les plus durs de cette thèse.
Merci aussi d'avoir toujours eu à c\oe ur de mettre en valeur mon travail, et de m'encourager à participer à des conférences, séminaires et écoles d'été.

Ultime panneau de ce triptyque doctoral, je veux témoigner de toute ma gratitude envers Yannig. 
Avatar de l'esprit sportif, tu appliques avec méthode et ardeur à la prévision énergétique  tous les algorithmes prometteurs qui te sont présentés. 
En compétition, il faut faire feu de tout bois !
De tous les statisticiens que j'ai rencontrés, tu es sûrement celui dont je partage le plus la vision des mathématiques appliquées.
En entraîneur attentionné, tu as sans cesse veillé à ma bonne intégration au sein de l'équipe à EDF, et à mon épanouissement tant sur le plan pratique que théorique.
Merci pour ta patience, ton expérience, et tes encouragements.

Plus généralement, je tiens à exprimer ma reconnaissance aux membres du jury, qui me consacrent une journée de leur temps que je sais fort précieux. 
Merci à Gilles et à Richard pour l'enthousiasme dont ils ont fait preuve à la relecture de ce manuscrit de thèse. 
Bien que le sérieux de l'évaluation repose sur le fait que nous ne nous connaissons pas personnellement, j'ai lu avec gourmandise les travaux de Gilles sur les noyaux; de Jalal, Mathilde et Richard sur les méthodes informées par la physique; et de Gabriel sur les fondements mathématiques de l'IA.
C'est un grand honneur pour moi que de vous voir siéger à mon jury.

Ma thèse n'aurait pas été possible sans la joyeuse farandole des statisticiens qui m'ont accompagné, au détour d'un article ou d'un stage, et ont enrichi mon paysage informatique et mathématique. 
Merci à Stéfania pour ta connaissance des indices téléphoniques, à Yann pour m'avoir épaulé sur ma toute première base de données, et à Yvenn pour tes talents d'organisateur de challenge. 
Un merci tout particulier à Francis qui, par l'exercice d'un prosélytisme dont il ne se cache pas, m'a totalement converti aux méthodes à noyaux et m'a conduit à la béatitude exaltée que confère la dimension effective. 
Merci et bravo à mes stagiaires, \'Eloi et Guillhem, pour le travail, la confiance, et  la joie qu'ils m'ont apportés.
J'ai été très fier de vous voir tant progresser.
Merci en retour à Adeline et Pierre, membres éminents de la dynastie des Gérardiens et dont j'ai été le stagiaire dans le temps, pour m'avoir donné envie de faire des statistiques.

Ces trois ans n'auraient pas eu la même saveur sans l'ambiance chaleureuse des équipes d'EDF et de la Sorbonne. Merci à Caroline, \'Eloi, Ferdinand, Guillaume Lambert, Guillaume Principato, Julie et Stanislas pour avoir tant égayé les bureaux que nous avons partagés.
Chers collègues d'EDF, merci
à Amaury pour ta musique punchy, 
à Bachir pour ton goût pour la bonne chère, 
à Christian pour nos discussions de régression linéaire et de randonnée, 
à \'Elaine pour notre dévotion commune à Arte et Élisabeth Quin,
à \'Elise pour ta gentillesse sans égal,
à Félicie pour ta joie de vivre,
à Hugo pour ton humour pince-sans-rire,
à Gilles, Sandra et Véronique pour le maintien de la tradition du café et de la conversation matinale,   
à Joseph pour m'avoir montré la voie,  
à Manel pour ton affection presque maternelle, 
à Margaux pour la force de tes convictions,
à Virgile pour ton regard émerveillé sur le monde.
Honorables collègues du LPSM, merci à mes anciens professeurs de master Anna, Antoine, Arnaud, Erwan, Ismaël et Lorenzo pour leur savoir encyclopédique, à Charlotte pour m'avoir accueilli les bras ouverts comme chargé de travaux dirigés, et à Alice, Miguel, et Paul pour leurs conseils d'enseignement.
Merci également à Hugues, Natalie, Nisrine, Nora et Xavier, sans qui les rouages administratifs m'auraient sans doute avalé depuis longtemps.

Au-delà de ma thèse, je sais ce que je dois à ceux qui me soutiennent au quotidien et agrémentent ma vie de leur présence loufoque. 
Merci à mes amis d'enfance, qui portent en eux la chaleur et la tranquillité du Sud. À Gilliane, avec qui j'ai coévolué au point d'entendre 
mentalement sa voix. Merci pour ton rire communicatif, le théâtre, nos vacances, et notre amour incompris du Top D17. 
À mes amis de lycée, Gabrielle et Léo en particulier, pour nos innombrables soirées plage. 
Merci à mes amis des Mines, fièrement rassemblés sous la bannière de la Piche et régulièrement convoqués par notre roi élu.
À Agathe pour avoir partagé ma détresse sur l'Île-Molène, à Amandine pour notre danse sur \textit{I Like To Move It} à Barcelone, à Antoine pour tes extravagances mégalomanes, à Charlène et Jean pour votre sens du chic et de la fête, à Denis pour tes traits d'esprit caustiques, à Félix pour m'avoir transmis ta passion du Japon, à Victor pour avoir créé un si beau \textit{Donjon} où tu nous accueilles toujours avec l'hospitalité médiévale de circonstance. 
Merci à mes amis de l'ENS, notamment aux autoproclamés \textit{malins}, qui réapparaissent périodiquement pour me professer leur sagesse douteuse. Paraît-il qu'un bol n'est jamais plus utile que quand il est vide...

Une place toute spéciale est prise en mon c\oe ur par ce quatuor étrange et, il faut bien l'avouer, un peu disparate, qui toutefois s'accorde merveilleusement à l'harmonie de mon existence\footnote{Les férus de solfège reconnaissent ici une septième diminuée.}.
À Baptiste, avec qui j'ai souvent festoyé jusqu'à l'évanouissement, co-inventeur des fameuses pâtes au gras. 
À toi, que l'honneur mal placé exhorte régulièrement à des aventures qui forcent l'admiration, tant en sport qu'en sciences, et qui n'a pourtant pas encore pleinement conscience de sa force.
À \'Eric, qui envisage l'existence comme un jeu, et avec qui je prends un plaisir non dissimulé (certes, parfois après-coup) à pousser aux extrêmes limites nos capacités physiques et mentales.
À toi, cher voisin du C2 en mes temps de pape, toi qui m'a fait survivre sur une île bordée de phoques, toi qui m'a traîné, mourant, sur la muraille de Chine, bref, toi en qui j'ai une confiance irraisonnable.
À Nataniel, dragon majestueux et sautillant, un peu bruyant par moments, mon Doppelgänger flamboyant.
À toi qui me suit, chaque année, selon le rituel, arpenter les cimes du Mercantour. 
À ton amour débordant pour la vie, les amis, les animaux mignons (et les limaces !?), les champignons, la littérature, la musique, la danse...  
À Alexis, qui partage courageusement ma vie.
À tes passions hétéroclites pour les chats, le matcha, le karaoké, la pop, les voyages, le clubbing, les jeux vidéo...
À toi qui préfères acheter la whey que pousser à la salle, et à tous ces autres traits qui te rendent si attachant.  

Il est des dons que l'on ne peut rendre. Je tâche ici au moins d'en rendre compte.
Merci à ma mère, pour ton soutien inconditionnel, indéfectible.
Pour ta foi en l'école et dans le savoir, qui m'a porté jusque-là. 
Pour ta force devant la maladie, qui t'a fait soigner les autres.
À mon père, pour m'avoir transmis ton amour des sciences et de la nature.
Pour ces journées passées à pêcher, cueillir les plantes sauvages, tailler des silex, observer les animaux, et ergoter sur les espèces d'arbres.
À Anne qui prend soin au quotidien de cet homme préhistorique en puissance.
À ma grande s\oe ur, Andréa, qui m'a toujours servi de modèle et qui a initié en moi un intérêt regrettable pour la télé-réalité et les séries.
À mon adorable neveu, Arthur, qui fait preuve d'une patience rare, mêlée d'un intérêt sincère, à écouter mes histoires.
À mes oncles, tantes, cousins et cousines, \'Elie, Fabien, Irène, Jokyo, Lucie, Magali, Pierre, Prune, et Rodrigue.
À ma marraine Jocelyne, pour sa quête du spirituel et sa tendresse pour la montagne.
Aux amis de mes parents, Alain et Thavy, Christian, Christophe, Georges et Janine, qui m'ont tant appris sur les vérités cachées de la vie.
J'ai ici une pensée pour ceux dont le feu s'est éteint.
À papé Jean Doumèche, médaillé de la Résistance en 47, pour les exploits dont les récits ont bercé ma jeunesse, et qui m'a donné son nom.
À pépé Marcel, incorrigible musicien et blagueur, qui m'a légué ses partitions.
Le monde en soit témoin, vous êtes partis comme vous avez vécu: dignes.

%
%
\pdfbookmark[0]{Abstract}{Abstract}
\addchap*{Abstract}
\label{chap:abstract}

\vspace*{0mm}

Physics-informed machine learning (PIML) is an emerging framework that integrates physical knowledge into machine learning models. 
This physical prior often takes the form of a partial differential equation (PDE) system that the regression function must satisfy.
In the first part of this dissertation, we analyze the statistical properties of PIML methods.
In particular, we study the properties of physics-informed neural networks (PINNs) in terms of approximation, consistency, overfitting, and convergence.
We then show how PIML problems can be framed as kernel methods, making it possible to apply the tools of kernel ridge regression to better understand their behavior.
In addition, we use this kernel formulation to develop novel physics-informed algorithms and implement them efficiently on GPUs.
The second part explores industrial applications in forecasting energy signals during atypical periods. We present results from the Smarter Mobility challenge on electric vehicle charging occupancy and examine the impact of mobility on electricity demand. Finally, we introduce a physics-constrained framework for designing and enforcing constraints in time series, applying it to load forecasting and tourism forecasting in various countries.

\textbf{Keywords:} 
Physics-informed machine learning, neural networks, kernel methods, load forecasting, time series

\vspace*{3em}

{\huge Résumé}

\vspace*{8mm}

L'apprentissage automatique informé par la physique est un domaine récent qui consiste à intégrer des connaissances physiques dans des modèles d'apprentissage automatique. 
L'information physique prend souvent la forme d'un système d'équations aux dérivées partielles (EDPs) que la fonction de régression doit satisfaire.
Dans la première partie de cette thèse, nous analysons les propriétés statistiques des méthodes d'apprentissage automatique informé par la physique.
En particulier, nous étudions les propriétés des réseaux de neurones informés par la physique, en termes d'approximation, de consistance, de surapprentissage et de convergence. 
En outre, nous montrons comment l'apprentissage statistique pénalisé par des systèmes d'EDPs linéaires peut se réécrire comme une méthode à noyaux. 
En s'appuyant sur cette reformulation, nous développons de nouveaux algorithmes informés par la physique, que nous implémentons ensuite efficacement sur carte graphique.
La deuxième partie se concentre sur des applications industrielles en prévision de signaux énergétiques en périodes atypiques. 
Nous présentons les résultats du Smarter Mobility Data Challenge sur l'occupation de la charge des véhicules électriques, et examinons l'impact de la mobilité sur la demande d'électricité. 
Enfin, nous développons un cadre pour la conception et l'application de contraintes dans les séries temporelles, en l'appliquant à la prévision de la consommation électrique et à la prévision du tourisme dans différents pays.

\textbf{Keywords:} 
Apprentissage statistique informé par la physique, réseaux de neurones, méthodes à noyau, prévision de consommation électrique, séries temporelles
\clearpage
%
\addchap*{Présentation de la thèse}
\label{chap:resume_fr}

La thèse de doctorat présentée ici est le fruit d'une collaboration entre l'entreprise EDF, spécialisée dans la production et la vente d'électricité, et Sorbonne Université. 
Afin d'améliorer la performance et l'explicabilité des modèles de prévision énergétique, EDF s'intéresse à l'incorporation de connaissances humaines (souvent regroupées sous le nom de l'expertise métier) dans ses méthodes statistiques d'aide à la décision.  
À cette fin, le domaine de l'apprentissage automatique informé par la physique (PIML en anglais) permet d'intégrer des contraintes dans des modèles d'apprentissage automatique. 
Il a été introduit en 2019 par l'invention des réseaux neuronaux informés par la physique. 
Dans cette thèse, nous nous focalisons donc sur l'étude des algorithmes d'apprentissage automatique informés par la physique et sur la prévision énergétique en période atypique. 
Par conséquent, nous abordons à la fois des aspects théoriques, tels l'impact des contraintes physiques sur les propriétés statistiques des estimateurs, ainsi que des applications industrielles sur données réelles.

\section*{Intégration de contraintes physiques dans les méthodes statistiques et applications industrielles}
\subsection*{Apprentissage automatique informé par la physique}

\paragraph{Interprétabilité des modèles.} Les algorithmes d'apprentissage automatique affichent des performances remarquables sur de nombreuses tâches complexes en analyse et de génération de données \citep{wang2023scientific}.
Cependant, malgré d'impressionnants résultats en reconnaissance d'images, en traitement du langage et en interaction avec des environnements adversoriaux (apprentissage par renforcement), les techniques modernes d'apprentissage profond peinent encore à accomplir certaines tâches simples mais cruciales pour de nombreuses applications pratiques.
La prévision de séries temporelles en est peut-être l'exemple le plus saillant \citep{MAKRIDAKIS20221346, NEURIPS2023_f06d5ebd, zeng2023transformers, Tayal2024}. 
De surcroît, même lorsqu'ils sont performants, les algorithmes d'apprentissage profond ne présentent pas les mêmes garanties théoriques que les méthodes statistiques plus standard.
Il est donc risqué de s'appuyer sur de tels algorithmes, souvent qualifiés de "boîte noire", pour des applications industrielles sensibles ou à fort enjeu \citep{vollert2021interpretable}.
En complément d'efforts de recherche pour mieux comprendre les algorithmes d'apprentissage profond, de nombreux travaux visent à développer de nouveaux algorithmes avec de meilleures garanties théoriques, sous le nom d'apprentissage automatique interprétable ou explicable \citep{lim2021time, lisboa2023the}.
Une piste prometteuse pour améliorer l'interprétabilité des algorithmes consiste à intégrer des connaissances issue de la modélisation dans les modèles statistiques.
Ces informations préalables sur les caractéristiques du signal à prévoir peuvent prendre la forme d'un modèle physique.

\paragraph{Physique et statistique.} L'idée de coupler une modélisation physique à des modèles statistiques n'est pas nouvelle.
D'une part, la déduction empirique de lois fondamentales de la nature est dans l'essence même de la physique en tant que science expérimentale.
Par exemple, les célèbres lois de Kepler de 1609 furent découvertes empiriquement, par l'ajustement de courbes sur des observations du mouvement des planètes.
D'autre part, la modélisation mathématique de l'intégration d'équations aux dérivées partielles (EDPs) issues de la physique dans des modèles statistiques était déjà formulée dans les travaux de \citet{wahba1990spline} en 1990.
Cependant, jusqu'à peu, deux problèmes principaux étaient restés en suspens.
Premièrement, aucun algorithme n'était capable d'incorporer efficacement et de façon systématique des connaissances physiques au sein de méthodes de régression statistique.
Cela signifie que les praticiens devaient fournir un travail conséquent afin d'incorporer leurs modélisations physiques dans des méthodes statistiques.
Deuxièmement, l'impact mathématique  de l'ajout de contraintes physique sur la performance des méthodes statistiques était inconnu.
Bien qu'il soit intuitif que les modèles avec plus d'informations devraient être plus puissants, en pratique, l'ajout de connaissances physiques augmente souvent la complexité informatique des modèles et en détériore l'optimisation.

\paragraph{L'apprentissage automatique informé de la physique.} 
Au cours des dernières années, de nouveaux moyens furent découverts pour adapter des algorithmes bien connus afin qu'ils emploient efficacement des a priori physiques.
En effet, l'incorporation d'EDPs dans les algorithmes statistiques a été illustrée par \citet{raissi2019PINN} sur les réseaux neuronaux et \citet{nickl2023bayesian} sur les méthodes de Monte-Carlo par chaînes de Markov (MCMC en anglais), tandis que l'utilisation des techniques d'EDP pour des tâches statistiques a été illustrée par \citet{arnone2022spatialRegression} sur la méthode des éléments finis (FEM).
Dans cette thèse, nous montrons comment faire de même avec des méthodes à noyaux \citep{doumeche2024physicsinformed}.
En particulier, le concept d'incorporer de la physique à des algorithmes d'apprentissage automatique classiques est maintenant connu sous le nom d'apprentissage automatique informé par la physique (PIML en anglais) \citet{karniadakis2021piml}.
S'appuyer sur des algorithmes bien connus pour l'intégration de contraintes physiques est très avantageux d'un point de vue informatique.
Par exemple, les réseaux neuronaux informés par la physique (PINNs en anglais) de \citet{raissi2019PINN} exploitent les écosystèmes publics \texttt{Pytorch} et \texttt{Tensorflow}, développés et maintenus par la communauté de l'apprentissage automatique et soutenus par de riches entreprises d'intelligence artificielles, comme Google et Meta.
L'inscription dans ces écosystèmes permet de directement tirer parti des puissantes accélérations de calcul et d'une gestion optimisée de la mémoire issues d'années de recherche en apprentissage automatique.
En outre, cela permet de facilement implémenter les algorithmes informés par la physique sur du matériel informatique très efficace.
Tyiquement, il s'agit d'exécuter les algorithmes sur des cartes graphiques (GPU en anglais), qui sont des unités de calcul ultra-optimisées pour effectuer des opérations mathématiques  spécifiques d'algèbre linéaire, comme le produit matrice-vecteur.
Dans cette thèse, nous montrons comment mettre en oeuvre nos méthodes à noyaux sur GPU. 
Toutes ces accélérations, ainsi que la bonne gestion de la mémoire, sont des améliorations cruciales pour le PIML, car l'ajout de physique à un algorithme statistique rend généralement son apprentissage plus coûteux en calculs.

\paragraph{Vers des contraintes plus faibles.} 
La plupart des algorithmes informés par la physique furent développés pour incorporer des a priori physiques prenant la forme de systèmes d'EDPs. 
Bien que ce cadre soit approprié pour la description de systèmes physiques dont les lois sont connues, les connaissances humaines ne prennent pas toujours une forme aussi rigide.
Par exemple, c'est le cas de nombreux signaux macroéconomiques, tels que la demande d'électricité.
Il n'en demeure pas moins que certaines contraintes peuvent être incorporées dans les modèles de prévision de tels signaux. 
Par exemple, les lois de la macroéconomie stipulent que, toutes choses égales par ailleurs, la demande d'électricité diminuera lorsque le prix augmentera.
De telles formes plus faibles de physique (où le terme « physique » est entendu ici au sens large de toute information issue d'une modélisation) ont été intégrées avec succès dans les PINNs \citep[voir, par exemple, ][]{daw2022lake}.
Nous expliquons au chapitre~7 comment appliquer des contraintes faibles aux séries temporelles.

\subsection*{Défis mathématiques de l'apprentissage informé de la physique}
\paragraph{Cadre mathématique.}
Le PIML est généralement divisé en quatre tâches \citep{raissi2019PINN,karniadakis2021piml, cuomo2022scientific}. Soient $d\in \mathbb N^\star$ la dimension du problème, $\Omega \subseteq \mathbb R^d$ le domaine d'intérêt, et $\mathscr{D}$ l'opérateur différentiel correspondant à la physique du problème. 
L'exemple typique est $d=2$, $\Omega = [0,1]^2$, et $\mathscr{D} = \partial^2_{1,1}+\partial^2_{2,2}$ est l'opérateur laplacien.

La première tâche est la résolution d'EDP. Étant donné un ensemble de conditions limites, l'objectif est de trouver une fonction satisfaisant à la fois le système d'EDPs et les conditions limites.
Par exemple, la condition limite de Dirichlet $h:\partial\Omega \to \mathbb R$ se traduit par la recherche d'une fonction $f_h$ telle que pour tout $ x\in \Omega, \;\mathscr{D}(f_h)(x) = 0$, et $f_h|_{\partial\Omega} = h$.
Dans l'exemple précédent, cela revient à résoudre l'EDP $\partial^2_{1,1}f_h+\partial^2_{1,1}f_h=0$ compte tenu de la condition aux limites $h$. 
Dans ce contexte, l'utilisateur spécifie la condition limite $h$ et entraîne ensuite un algorithme d'apprentissage (par exemple, un PINN) à apprendre la solution de l'EDP.
Bien que les méthodes d'EDP telles que la méthode des éléments finis (FEM) soient déjà très efficaces dans la résolution des EDP, elles peuvent être coûteuses en termes de calcul. 
Ici, les méthodes d'apprentissage automatique sont utiles pour trouver rapidement des approximations pour $f_h$, appelées modèles de substitution.
Ceci est particulièrement intéressant lorsque l'EDP doit être résolue de manière répétée, comme c'est le cas des prévisions météorologiques.

La deuxième tâche est la modélisation hybride.
\'Etant données $n\in \mathbb N^\star$ observations $(X_1, Y_1)$, $\dots$, $(X_n, Y_n)$ indépendantes et identiquement distribuées (i.i.d.) selon la loi de la variable aléatoire $(X,Y)$, où $Y = f^\star(X)+\varepsilon$ et $\varepsilon$ est un bruit, l'objectif est d'estimer la fonction $f^\star$. 
La particularité de ce cadre d'apprentissage supervisé est l'a priori que $f^\star$ est solution de l'EDP $\mathscr{D}(f^\star) = 0$, toutefois avec une éventuelle erreur de modélisation. 
Ce cadre est particulièrement pertinent lorsque l'a priori physique est incomplet. 
Par exemple, les conditions aux limites peuvent ne pas être entièrement spécifiées, ou l'ensemble des équations différentielles peut admettre un nombre infini de solutions.
Dans ce cas, les techniques traditionnelles d'EDP ne peuvent pas être appliquées directement et les données sont nécessaires pour résoudre le problème.

La troisième tâche est l'apprentissage d'EDP. 
\'Etant données $n\in \mathbb N^\star$ observations $(X_1, Y_1)$, $\dots$, $(X_n, Y_n)$ indépendantes et identiquement distribuées (i.i.d.) selon la loi de la variable aléatoire $(X,Y)$, où $Y = f^\star(X)+\varepsilon$ et $\varepsilon$ est un bruit, l'objectif est d'estimer la fonction $f^\star$ ainsi que la loi physique à laquelle $f^\star$ obéit. 
Dans ce contexte d'apprentissage supervisé, la seule connaissance préalable est le fait que la fonction $f^\star$ est solution d'un système d'EDPs à coefficients inconnus. Un exemple de tel a priori physique pourrait être que $f^\star$ satisfait l'EDP $\Delta f^\star = \lambda^\star f^\star$, où $\lambda^\star \in \mathbb R$ est un paramètre inconnu devant être estimé grâce aux données.
Les PINNs \citet{raissi2019PINN}, la régression LASSO \citep{kaheman2020sindy} et les MCMCs \citep{nickl2023bayesian} sont des algorithmes efficaces d'apprentissage d'EDP. 

La quatrième tâche consiste à directement construire un solveur d'EDP. 
Formellement, il s'agit d'apprendre l'opérateur $\phi : h\in L^2(\partial\Omega) \mapsto f_h\in L^2(\Omega) $ qui, à une condition limite $h$, associe l'unique solution $f_h$ telle que $f_h|_{\partial\Omega} = h$ et $\forall x\in \Omega, \;\mathscr{D}(f_h)(x) = 0$. 
Ce contexte est appelé apprentissage d'opérateur.
L'objectif ici est de fournir des modèles de substitution plus rapides, en apprenant un modèle global $\phi$, plutôt que d'entraîner un algorithme spécifique pour chaque nouvelle condition limite $h$.
DeepONet \citep{lu2021learning} et les opérateurs neuronaux de Fourier \citep{li2021fourier} sont des algorithmes d'apprentissage d'opérateur efficaces.

Dans ce document, nous nous concentrons principalement sur la résolution d'EDP et la modélisation hybride. Les deux autres tâches seront l'objet de de travaux ultérieurs. En effet, l'apprentissage d'EDP et l'apprentissage d'opérateur sont des opérations plus complexes, et leurs propriétés statistiques sont encores inconnues.

\paragraph{Comment créer des algorithmes intégrant de l'information physique ?}
L'un des principaux défis du PIML est de concevoir des algorithmes efficaces pour traiter les tâches susmentionnées.
En pratique, la plupart des implémentations reposent sur la minimisation d'un risque empirique constitué d'une partie d'accroche aux données et d'une pénalité physique. 
En modélisation hybride, l'EDP est intégrée comme une pénalité $L^2$, et le risque empirique devient
\[\mathscr{R}_n(f) = \sum_{j=1}^n \|f(X_j)-Y_j\|_2^2 + \lambda\int_\Omega |\mathscr{D}(f)(x)|^2dx,\]
où $\lambda>0$ est un hyperparamètre fixé par l'utilisateur.
Cette minimisation est effectuée sur une classe de fonctions, à savoir les réseaux neuronaux pour les PINNs \citep{raissi2019PINN}, et une base de Fourier pour nos méthodes à noyaux \citep{doumèche2024physicsinformedkernellearning}.
Cependant, la pénalité EDP rend particulièrement difficile  la recherche d'un minimiseur global de $\mathscr{R}_n$ sur une classe de fonctions donnée. 
Il faut pour cela pouvoir être capable de dériver une fonction de la classe de fonctions considérée, ce qui n'est pas toujours possible (par exemple, les forêts aléatoires ne sont pas dérivables).
D'ailleurs, bien que les PINNs soient la technique qui a reçu le plus d'attention récemment, l'algorithme proposé par \citet{raissi2019PINN} est susceptible de surapprendre \citep{wang2022when, doumeche2023convergence}, tandis que son optimisation est fortement dégradée lorsque l'opérateur différentielle $\mathscr{D}$ est non linéaire \citep{bonfanti2024challengesnonlinearregimephysicsinformed}.
En outre, la plupart des algorithmes proposés dans la littérature sont très gourmands en ressources informatiques.
En particulier, les PINNs nécessitent des milliers d'étapes de descente de gradient pour converger, alors qu'ils ne sont pas toujours significativement plus performants que les solveurs d'EDP traditionnels, tant en termes de vitesse de calcul que de précision, comme le révèle la méta-analyse de \citet{mcgreivy2024weakbaselinesreportingbiases}. 

\paragraph{Comment quantifier les gains de la physique ?}
D'un point de vue théorique, mesurer l'impact de la physique sur la performance de l'algorithme est 
une question qui n'a toujours pas trouvé de réponse complète.
Un avantage de l'apprentissage informé de la physique est que, en s'appuyant sur des algorithmes connus, il devient possible d'adapter les outils spécifiques à l'algorithme initial pour comprendre les propriétés théoriques des versions physiquement pénalisées.
Par exemple, l'analyse du noyau tangent neuronal (NTK) des PINNs caractérise la convergence de leur descente de gradient \citep{bonfanti2024challengesnonlinearregimephysicsinformed}, tandis que les outils de l'inférence bayésienne non paramétrique permettent d'étudier le taux de convergence des MCMCs physiquement informées \citep{nickl2019convergence}. 
Dans cette thèse, nous nous appuyons sur l'analyse de la dimension effective de \citet{caponnetto2007optimal} et \citet{blanchard2008Statistical} pour caractériser le taux de convergence de nos méthodes à noyaux \citep{doumeche2024physicsinformed}.
La plupart de ces résultats théoriques confirment l'intuition selon laquelle les algorithmes PIML sont plus difficiles à optimiser, mais offrent de meilleures performances statistiques lorsqu'ils sont entraînés dans de bonnes conditions.

\subsection*{Contexte industriel}
Au-delà des aspects théoriques liés aux propriétés statistiques et à la complexité des algorithmes, l'apprentissage automatique informé de la physique a démontré son efficacité dans des applications industrielles. Cette thèse, menée en collaboration avec l'entreprise EDF, se concentre sur les applications aux séries temporelles, en particulier dans le domaine de la prévision énergétique.

\paragraph{Séries temporelles.} 
Les séries temporelles sont omniprésentes dans les applications industrielles \citep{petropoulos2022forecasting}, englobant la prévision des signaux macroéconomiques (offre, demande, prix, mobilité humaine...), l'estimation de la propagation des maladies et du trafic hospitalier, le suivi des processus industriels en temps réel (processus de fabrication, réactions chimiques, maintenance préventive...) et l'anticipation des événements environnementaux (vagues de chaleur, incendies, précipitations...).
Cependant, les séries temporelles sont particulièrement difficiles à traiter d'un point de vue statistique.
Tout d'abord, les observations sont corrélées, ce qui signifie que la loi des grands nombres et le théorème central limite ne peuvent pas être directement appliqués pour créer des estimateurs.
Ensuite, il est courant que la distribution des séries temporelles évolue au cours du temps, que cela soit en raison d'une tendance, d'une saisonnalité ou de ruptures.
En outre, les séries temporelles comportent souvent des valeurs manquantes, en conséquence de fréquences d'échantillonnage potentiellement différentes dans les données, ou encore de défaillances de capteurs.
Tous ces phénomènes limitent la quantité de données pertinentes disponibles pour l'apprentissage et rendent difficile la prévision des séries temporelles.
L'ajout de contraintes physiques aux modèles de prévision apparaît comme un moyen prometteur pour en améliorer les performances.
Les techniques de PIML ont été appliquées avec succès à des séries temporelles du monde réel présentant des dépendances physiques bien connues, telles que les prévisions météorologiques \citep{kashinath2021physics}, la prévision de production d'énergie renouvelable \citep{lagomarsino2023physics}, ou le contrôle en temps réel de réactions chimiques industrielles \citep{nguyen2022physics}.

\paragraph{Prévision de signaux énergétiques.} 
Dans cette thèse, nous nous concentrons principalement sur la prévision de signaux énergétiques, comme l'occupation de bornes de recharge de véhicules électriques et la demande d'électricité. 
Ces deux tâches de prévision sont difficiles, mais utiles pour l'industrie de l'énergie.
En effet, le marché des véhicules électriques est émergent et en forte croissance, et les fournisseurs doivent adapter le réseau électrique à ses demandes de haute intensité électrique. 
Cependant, les bases de données et les modèles de machine learning appliquées aux véhicules électriques sont encore rares \citep{amara-ouali2021a}.
En ce qui concerne la prévision de demande d'électricité (également appelée prévision de charge), le stockage de l'électricité est coûteux et limité, tandis que l'offre doit correspondre à la demande à tout moment pour éviter les pannes.
La prévision de la demande est donc nécessaire pour ajuster la production et agir efficacement sur les marchés de l'électricité \citep{hammad2020methods}.
Cependant, aucun de ces signaux n'est régi par un ensemble connu d'EDPs. 
De fait, l'ensemble complet des variables explicatives responsables de leur variations est inconnu ou non mesuré.
Des modèles statistiques, potentiellement guidée par notre connaissance du comportement de ces signaux, sont donc nécessaires pour combler ces défauts de modélisation.

\subsection*{Organisation du document}

Ce document est composé d'une introduction, d'une partie théorique (Partie I), d'une partie appliquée (Partie II) et d'une conclusion.
Chaque partie est divisée en plusieurs chapitres, chacun correspondant à une contribution autonome.
Nous donnons ci-dessous un bref aperçu de chaque chapitre. 
Chacun a donné ou donnera lieu à une publication.\\

\textbf{\large Partie I : Quelques résultats mathématiques sur l'apprentissage automatique informé par la physique}

\textit{On the convergence of PINNs}, Nathan Doumèche, Gérard Biau (Sorbonne Université), et Claire Boyer (Université Paris-Saclay). Publié à Bernoulli.

\fbox{%
\begin{minipage}{\textwidth}
   Le chapitre~\ref{ch:my-domain} est consacré à l'analyse des propriétés statistiques des réseaux neuronaux informés par la physique (PINNs) pour la résolution d'EDPs et la modélisation hybride. Nous nous concentrons sur l'approximation, la consistance du risque et la cohérence physique des PINNs. 
   Au travers d'exemples, nous illustrons comment les réseaux neuronaux informés par la physique sont sujet à un surapprentissage systémique.
   Nous montrons également que les techniques usuelles de régularisation sont efficaces pour s'assurer de leur consistance.
\end{minipage}
}\vspace{1em}

\textit{Physics-informed machine learning as a kernel method}, Nathan Doumèche, Francis Bach (INRIA Paris), Gérard Biau (Sorbonne Université), et Claire Boyer (Université Paris-Saclay). Publié aux Proceedings of Thirty Seventh Conference on Learning Theory (COLT 2024).

\fbox{%
\begin{minipage}{\textwidth}
   Dans le chapitre~\ref{ch:requirements}, nous prouvons que, pour les EDPs linéaires, la résolution d'EDPs et la modélisation hybride sont des méthodes à noyaux. 
   En s'appuyant sur la théorie des méthodes à noyaux, nous montrons que l'estimateur informé par la physique converge au moins au taux minimax de Sobolev. 
   Des taux plus rapides peuvent être atteints, mettant alors en évidence les bénéfices de l'a priori physique.
\end{minipage}
}\vspace{1em}

\textit{Physics-informed kernel learning}, Nathan Doumèche, Francis Bach (INRIA Paris), Gérard Biau (Sorbonne Université), et Claire Boyer (Université Paris-Saclay). Accepté avec révisions mineures au Journal of Machine Learning Research (JMLR).

\fbox{%
\begin{minipage}{\textwidth}
   Le chapitre~\ref{ch:related-work} est dédié à l'emploi de séries de Fourier pour approximer le noyau susmentionné.
   Nous y proposons un estimateur implémentable minimisant le risque empirique informée par la physique. 
   Nous illustrons la performance de l'estimateur à noyaux par des expériences numériques, tant pour la modélisation hybride que pour la résolution d'EDP.
\end{minipage}
}\vspace{3em}

\textbf{\large Partie II: Prévision de séries temporelles en périodes atypiques}

\textit{Forecasting Electric Vehicle Charging Station Occupancy: Smarter Mobility Data Challenge}, Yvenn Amara-Ouali (Université Paris-Saclay), Yannig Goude (Université Paris-Saclay), Nathan Doumèche (Sorbonne Université), Pascal Veyret (EDF R\&D), et al. Publié au Journal of Data-centric Machine Learning Research (DMLR).

\fbox{%
\begin{minipage}{\textwidth}
   Le chapitre~\ref{ch:contrib-1} est un chapitre spécial décrivant les résultats des trois équipes gagnantes du Smarter Mobility Data Challenge. 
   L'objectif de ce défi était de prédire l'occupation des bornes de recharge de véhicules électriques à Paris en 2021.
   Notre équipe s'est classée 3ème de ce challenge.
\end{minipage}
}\vspace{1em}

\textit{Human spatial dynamics for electricity demand forecasting}, Nathan Doumèche, Yannig Goude (Université Paris-Saclay), Stefania Rubrichi (Orange Innovation), et Yann Allioux (EDF R\&D). En évaluation par les pairs.

\fbox{%
\begin{minipage}{\textwidth}
  Dans le chapitre~\ref{ch:contrib-2}, nous explorons l'impact des données liées au travail sur la prévision de la demande d'électricité. 
  Nous démontrons que les indices de mobilité dérivés des données des réseaux mobiles améliorent de manière significative la performance des modèles de l'état de l'art, en particulier pendant la période de sobriété énergétique de la France de l'hiver 2022-2023.
\end{minipage}
}\vspace{1em}

\textit{Forecasting time series with constraints}, Nathan Doumèche, Francis Bach (INRIA Paris), Eloi Bedek (EDF R\&D), Gérard Biau (Sorbonne Université), Claire Boyer (Université Paris-Saclay), et Yannig Goude (Université Paris-Saclay). En évaluation par les pairs.

\fbox{%
\begin{minipage}{\textwidth}
   Le chapitre~\ref{ch:contrib-3} se concentre sur l'extension du cadre de Fourier développé au chapitre~3 à des contraintes spécifiques aux séries temporelles. 
   Les séries temporelles macroéconomiques ne satisfaisant que rarement un jeu d'EDPs connu, nous nous concentrons sur des contraintes plus faibles, comme les modèles additifs, l'adaptation en ligne aux ruptures, la prévision hiérarchique et l'apprentissage par transfert. Nous démontrons que les méthodes à noyaux qui en résultent atteignent des performances de pointe en prévision de charge et de tourisme.
\end{minipage}
}\vspace{2em}	
\cleardoublepage
\currentpdfbookmark{\contentsname}{toc}
\setcounter{tocdepth}{2}		
\tableofcontents				
\cleardoublepage

\pagestyle{empty}				
\pagenumbering{arabic}			
\pagestyle{scrheadings}			

%
\chapter{Introduction}
\label{sec:intro}

The work presented in this manuscript is the result of a collaboration between EDF, a company specializing in the production and sale of electricity, and Sorbonne University.
To improve the performance and explainability of its forecasting models, EDF is particularly interested in incorporating human knowledge into its statistical methods.
To this end, physics-informed machine learning (PIML) is a new framework designed to integrate physical constraints into established machine learning models.
It was introduced in 2019 with physics-informed neural networks (PINNs).
In this thesis, we investigate the mathematical properties of PIML, as well as applications to energy forecasting during atypical periods. 
Consequently, this thesis addresses both theoretical aspects, such as the impact of physical constraints on the statistical properties of the physics-informed estimators, and real-world applications.

\section{Integrating physical prior into machine learning for industrial applications}
\label{sec:integrating}
\subsection*{Physics-informed machine learning}

\paragraph{Towards interpretable machine learning models.} Machine learning techniques have achieved remarkable performance in many complex tasks of data analysis and generation \citep{wang2023scientific}.
However, despite impressive results in image recognition, language processing, and interaction with adversarial environments (reinforcement learning), modern deep learning techniques still struggle with some simpler tasks which are crucial for practical applications.
Time series forecasting is perhaps the most striking of such examples \citep{MAKRIDAKIS20221346, NEURIPS2023_f06d5ebd, zeng2023transformers, Tayal2024}. 
Moreover, even when they perform well, deep learning algorithms do not have the same theoretical guarantees as standard statistical techniques.
This makes it risky to rely on such black-box algorithms for sensitive or high-stakes industrial applications \citep{vollert2021interpretable}.
In addition to studying the mathematical properties of efficient deep learning algorithms to better understand their behavior, new algorithms with theoretical guarantees are being developed under the name of interpretable or explainable machine learning \citep{lim2021time, lisboa2023the}.
A promising way to achieve efficient and explainable machine learning is to develop algorithms that are able to integrate expert knowledge.
Such prior knowledge can take the form of a physical model of the phenomenon at hand.

\paragraph{Physics and Statistics.} 
Mixing physical modeling and statistical models is nothing new.
On the one hand, using data to infer physical laws is the essence of physics as an experimental science.
For example, the famous Keplerian laws of 1609 were derived by fitting curves from observations of planetary motion.
On the other hand, the integration of partial differential equations (PDEs) from physics into statistical models was already formally mathematically modeled by \citet{wahba1990spline} in 1990.
However, up until recently, two main problems remained unsolved.
First, no algorithm was able to efficiently and systematically incorporate physical priors into statistical regression problems.
This meant that practitioners had to work a lot to incorporate their physical knowledge into statistical methods.
Second, the mathematical impact of adding physics to statistical methods in terms of performance was unknown.
In fact, although common sense suggests that models with more information should be more powerful, doing so adds complexity to the statistical models and make their training more challenging.

\paragraph{Physics-informed machine learning.} 
What has changed in recent years is the discovery of new ways to efficiently incorporate physical priors into statistical problems using well-known algorithms.
Indeed, the incorporation of physics into statistical algorithms has been exemplified by \citet{raissi2019PINN} with neural networks and \citet{nickl2023bayesian} with Monte Carlo Markov Chains (MCMCs), while the use of PDE techniques for statistical tasks has been illustrated by \citet{arnone2022spatialRegression} with the finite element method (FEM).
In this dissertation, we show how physics can be incorporated into kernel methods \citep{doumeche2024physicsinformed}.
In particular, this idea of adding physics into well-known machine learning algorithms is now called physics-informed machine learning (PIML), as theorized by \citet{karniadakis2021piml}.
Relying on well-known algorithms to incorporate physical prior is extremely advantageous from a computational point of view.
Indeed, the physics-informed neural networks (PINNs) of \citet{raissi2019PINN} can leverage the open source \texttt{Pytorch} and \texttt{Tensorflow} ecosystems, developed and maintained by the machine learning community and supported by wealthy AI companies like Google and Meta.
This allows the many powerful computational speedups and efficient memory allocations from years of machine learning research to be implemented directly in PIML.
In addition, it allows PIML algorithms to run on powerful hardware such as graphics processing units (GPUs), the ultra-optimized computing units that have been improved by AMD, INTEL, and NVIDIA for many years to perform specific mathematical operations (such as matrix-vector products).
In this dissertation, we will show how to implement our kernel methods on GPUs \citep{doumèche2024physicsinformedkernellearning}. 
All of these speedups and memory implementations are crucial for PIML, because adding physics to a statistical algorithm generally makes its training more computationally expensive.

\paragraph{Towards weaker constraints.} 
Most PIML algorithms have been developed to incorporate physical priors taking the form of PDE systems. 
Although this setting is appropriate for well-studied physical systems, expert knowledge does not always take such a rigid form.
For example, many macroeconomic signals, such as electricity demand, do not satisfy a known set of PDEs.
Nevertheless, there are some physical insights that can be incorporated into forecasting models. 
For example, the laws of macroeconomics say that demand for electricity will fall as the price rises.
Such weaker forms of physics ---where "physics" is understood here as modeling information--- have been successfully integrated into PINNs \citep[see, e.g., ][]{daw2022lake}.
We will discuss how to apply weak constraints to time series in Chapter~7.

\subsection*{Mathematical challenges of PIML}
\paragraph{Mathematical framings of PIML.}
PIML is usually divided into four different tasks \citep{raissi2019PINN,karniadakis2021piml, cuomo2022scientific}. Let $d\in \mathbb N^\star$ be the dimension of the problem, $\Omega \subseteq \mathbb R^d$ be the domain of interest, and $\mathscr{D}$ be a differential operator. 
The typical example is $d=2$, $\Omega = [0,1]^2$, and $\mathscr{D} = \partial^2_{1,1}+\partial^2_{2,2}$ is the Laplacian operator.

The first PIML task is PDE solving. Given a set of boundary conditions, the goal is to find a function that satisfies both the PDE system and the boundary conditions.
For example, the Dirichlet boundary condition $h:\partial\Omega \to \mathbb R$ translates into finding a function $f_h$ such that for all $ x\in \Omega, \;\mathscr{D}(f_h)(x) = 0$, and  $f_h|_{\partial\Omega} = h$.
In the previous example, this amounts to solving the PDE  $\partial^2_{1,1}f_h+\partial^2_{1,1}f_h=0$ given the boundary condition $h$. 
In this context, the user specifies the boundary condition $h$ and then trains a learning algorithm (e.g., a PINN) to learn the solution to the PDE.
Altough PDE methods like the finite element method (FEM) are already very effective in PDE solving, they can be computationally expensive. 
Here, methods from machine learning are helpful to find computationally efficient approximations for $f_h$, called surrogate models.
This is especially interesting when the PDE has to be solved quickly and many times, as in the case of daily weather forecasts \citep{cheng2022data}.

The second task is hybrid modeling. 
Given $n\in \mathbb N^\star$ i.i.d. observations $(X_1, Y_1)$, $\dots$, $(X_n, Y_n)$ distributed as the random variable $(X,Y)$ such that $Y = f^\star(X)+\varepsilon$, where $\varepsilon$ is a random noise, the goal is to estimate $f^\star$.
What makes this supervised learning setting special is that we know that $f^\star$ follows the PDE $\mathscr{D}(f^\star) = 0$, up to a possible modeling error. 
This setting is particularly relevant when the physical prior is incomplete, in the sense that it is ill-posed. 
For example, the boundary condition may not be fully specified, or the set of differential equations may admit an infinite number of solutions.
Therefore, traditional PDE techniques cannot be applied directly, and the data is needed to solve the problem.

The third PIML task is PDE learning. 
Given $n\in \mathbb N^\star$ i.i.d. observations $(X_1, Y_1)$, $\dots$, $(X_n, Y_n)$ distributed as the random variable $(X,Y)$ such that $Y = f^\star(X)+\varepsilon$, where $\varepsilon$ is a random noise, the goal is to estimate $f^\star$ as well as the physical law it obeys. 
In this supervised learning setting, the only prior knowledge is that $f^\star$ is the solution to a PDE system with unknown coefficients. For instance, the physical prior could be that we know that $f^\star$ satisfies the PDE $\Delta f^\star = \lambda^\star f^\star$, where $\lambda^\star \in \mathbb R$ is an unknown parameter which must be inferred from the data.
Efficient implementation of PDE learning algorithms include PINNs \citep{raissi2019PINN}, LASSO regression \citep{kaheman2020sindy}, and MCMCs \citep{nickl2023bayesian}.

The fourth PIML task is to learn the PDE solver directly. 
Formally, the goal is to learn the operator $\phi: h\in L^2(\partial\Omega) \mapsto f_h\in L^2(\Omega) $ associating to a boundary condition $h$ the unique solution $f_h$ such that  $f_h|_{\partial\Omega} = h$ and $\forall x\in \Omega, \;\mathscr{D}(f_h)(x) = 0$. 
This context is called operator learning.
The goal here is to provide faster surrogate models by training a large model to learn the operator $\phi$, instead of training a new algorithm for each new boundary condition $h$.
Efficient operator learning algorithms include DeepONet \citep{lu2021learning} and Fourier neural operators \citep{li2021fourier}.

In this manuscript, we will mainly focus on PDE solving and hybrid modeling. The two other tasks are left for future works and will be discussed in the conclusion section. Indeed, operator learning and PDE learning are more complex problems, which statistical properties are still to be uncovered.

\paragraph{How to create efficient PIML algorithms?}
One of the main challenges in PIML is to design effective algorithms to handle the aforementioned tasks.
In practice, many implementations rely on minimizing an empirical loss with a data-driven part and a physical penalty. For instance, most hybrid modeling frameworks use the PDE as a soft penalty, and intend to minimize the empirical risk 
\[\mathscr{R}_n(f) = \sum_{j=1}^n \|f(X_j)-Y_j\|_2^2 + \lambda\int_\Omega |\mathscr{D}(f)(x)|^2dx,\]
where $\lambda>0$ is an hyperparameter to be scaled by the user.
This minimization is performed over a class of functions, such as neural networks for PINNs \citep{raissi2019PINN} or low frequency Fourier modes in our kernel methods \citep{doumèche2024physicsinformedkernellearning}.
However, finding a global minimizer of $\mathscr{R}_n$ over a given class of function is made particularly difficult because of the PDE penalty. 
This requires being able to differentiate a function from the function class of interest, which is not always possible (e.g., random forests are not differentiable).
In fact, although PINNs is the technique that has received the most attention recently, the algorithm proposed by \citet{raissi2019PINN} has been shown to be prone to overfitting \citep{wang2022when, doumeche2023convergence}, while its optimization is highly degraded when $\mathscr{D}$ is nonlinear \citep{bonfanti2024challengesnonlinearregimephysicsinformed}.
Moreover, many of the algorithms proposed in the literature are computationally intensive.
In particular, PINNs require thousands of gradient descent steps to converge, and do not clearly outperform the much faster traditional PDE solvers on PDE solving tasks, as revealed by the meta-analysis of \citet{mcgreivy2024weakbaselinesreportingbiases}. 

\paragraph{How to quantify the gains from the physics?}
From a theoretical point of view, measuring the impact of the physics on the performance of the algorithm is challenging.
An advantage of PIML is that, by relying on known algorithms, it becomes possible to adapt some well-established tools to understand the theoretical properties of the physically penalized versions of the algorithms.
For example, the neural tangent kernel (NTK) analysis of PINNs better characterizes the convergence of their gradient descent \citep{bonfanti2024challengesnonlinearregimephysicsinformed}, whereas tools from nonparametric Bayesian analysis makes it possible to study the convergence rate of physics-informed MCMCs \citep{nickl2019convergence}. 
In this dissertation, we rely on the effective dimension analysis of \citet{caponnetto2007optimal} and \citet{blanchard2008Statistical} to characterize the convergence rate of our kernel techniques \citep{doumeche2024physicsinformed}.
Most of these theoretical results confirm the intuition that PIML algorithms are harder to optimize, but yield better statistical performance when done right.

\subsection*{Industrial context}
Beyond the theoretical aspects related to the statistical properties and complexity of the algorithms, PIML has demonstrated its efficiency in industrial applications. This PhD, conducted in collaboration with the EDF energy company, focuses on time series applications, particularly in energy forecasting.

\paragraph{Time series.} 
Time series are ubiquitous in real-world applications \citep{petropoulos2022forecasting}, including forecasting  macroeconomic signals (supply, demand, prices, human mobility...), estimating the spread of disease and hospital traffic, monitoring industrial processes in real time (manufacturing, chemical reactions, preventive maintenance...), and anticipating environmental events (heat waves, wildfires, rainfall...).
However, time series are particularly difficult to handle from a statistical point of view.
First, the observations are correlated, which means that the law of large numbers and the central limit theorem cannot be directly applied to create estimators.
Then, the distribution of the target time series often changes over time, either due to trend, seasonality, or breaks.
Moreover, time series often have missing values, due to different sampling frequencies in the data or sensor failures.
All of these phenomena limit the amount of relevant data available and make time series forecasting difficult.
Adding physical constraints to the models appears like a promising way to improve the performance of the forecasts.
PIML techniques have been successfully applied to real-world time series with well-known physical dependencies, such as weather forecasting \citep{kashinath2021physics}, renewable energy production forecasting \citep{lagomarsino2023physics}, or real-time control of industrial chemical reactions  \citep{nguyen2022physics}.

\paragraph{Energy forecasting.} In this dissertation, we will mainly focus on forecasting energy signals, like electric vehicle charging station occupancy and electricity demand. 
Both of these forecasting tasks are challenging but valuable to the energy industry.
Indeed, the electric vehicle market is emerging and fast-growing, and providers need to adapt electricity grids to its high-intensity demands. 
However, open data sets and models are still rare \citep{amara-ouali2021a}.
As for electricity demand forecasting (also called load forecasting), electricity is expensive to store, while supply must match demand at all times to avoid blackouts.
Forecasting demand is thus necessary to adjust production, and to act on the electricity markets \citep{hammad2020methods}.
Neither of these signals is governed by a known set of PDEs. 
In fact, even the set of explanatory variables responsible for their fluctuation is unknown or unmeasured.
Therefore, statistical models are needed to fill this modeling gap.

\subsection*{Organization of the manuscript}

\label{chap:contribution}
This thesis consists of an introduction, a theoretical part (Part I), an applied part (Part II), and a conclusion.
Each part is separated in several chapters, each corresponding to a standalone contribution.
Thus, each chapter has led to or should lead to a publication.



The following sections introduce the basic concepts of each chapter and present the main contributions of this thesis. 
Note that the notation in these sections is unified and may differ slightly from the notation in the chapters, which corresponds to the notation in the corresponding papers.

\section{Some mathematical insights on physics-informed machine learning}
\subsection*{Chapter~\ref{ch:my-domain}: On the convergence of PINNs}

\textit{On the convergence of PINNs}, Nathan Doumèche, Gérard Biau (Sorbonne Université), and Claire Boyer (Université Paris-Saclay). Published in Bernoulli.

\fbox{%
\begin{minipage}{\textwidth}
   In Chapter~\ref{ch:my-domain}, we analyze the statistical properties of physics-informed neural networks (PINNs) for PDE solving and hybrid modeling, focusing on approximation, risk consistency, and physical inconsistency. 
   Through specific examples, we show that PINNs are prone to overfitting and demonstrate that standard regularization techniques are effective in ensuring their consistency.
\end{minipage}
}

\paragraph{Theoretical risk in hybrid modeling.} 
PINNs intend to tackle hybrid modeling tasks by minimizing a physically penalized theoretical risk over a class of neural networks. In this supervised learning setting, the goal is to learn the function $f^\star$ such that $Y=f^\star(X)+\varepsilon$, where $\varepsilon$ is a random noise, given i.i.d. observation $(X_1, Y_1)$, $\dots$, $(X_n, Y_n)$ of the process $(X,Y)\in \Omega\times \mathbb R^{d_2}$, where $\Omega$ is a bounded Lipschitz domain of $\mathbb R^{d_1}$. 
Lispchitz domains are a generalization of $C^1$-manifolds, which encompasses the square $[0,1]^{d_1}$ for example.
What makes this regression setting special is the prior physical knowledge that $f^\star$ satisfies the boundary conditions $\forall \bx \in \partial \Omega$, $f^\star(x) = h(x)$, and the PDE system, $\forall 1\leq k \leq M$, $\forall \bx \in \Omega$, $\mathscr{D}_k(f)(x) = 0$.
The theoretical risk takes the form
\[\mathscr{R}_n(f) = \frac{\lambda_d}{n} \sum_{i=1}^n \|f(X_i) - Y_i\|_2^2 + \lambda_e \mathbb{E}\|f(X^{(e)})-h(X^{(e)})\|_2^2 + \frac{1}{|\Omega|}\sum_{k=1}^M \int_\Omega \mathscr{D}_k(f)(x)^2 dx,\]
where $f$ is a neural network, $\lambda_d>0$ and $\lambda_e>0$ are hyperparameters, $\Omega$ is the domain of interest, $\mathscr{D}_j$ are differential operators, $\bX^{(e)}$ is a random variable sampled on $\partial \Omega$.
Though we assume that the practitioner has no control on the data $(X_i,Y_i)$, the distribution of the random variable $X^{(e)}$ can be chosen freely to incorporate the boundary condition. 
Usually, $X^{(e)}$ is taken as the uniform distribution on $\partial\Omega$.
The interest of using neural networks is that it is easy to compute their derivatives by backpropagation, and that neural networks are universal approximators, meaning that there is no bias intrinsic to the neural network class (given the neural networks are big enough). 

\paragraph{The neural network class.}
To be able to evaluate the operators $\mathscr{D}_k(f)$, the neural network $f$ must be differentiable. 
Thus, instead of considering the usual $\mathrm{relu}$ activation function, the PINNs community relies on the hyperbolic tangent function $\tanh$.
The class of interest is thus the class of fully-connected feedforward neural networks with $H\in\mathbb{N}^\star$ hidden layers of sizes $(L_1, \hdots, L_H) :=(D,\hdots ,  D) \in (\mathbb{N}^\star)^H$ and activation $\tanh$.
This corresponds to the space of functions $f_\theta$ from $\mathbb R^{d_1}$ to $\mathbb R^{d_2}$, defined by
\begin{equation*}
    f_{\theta} = \mathcal{A}_{H+1}\circ (\tanh \circ \mathcal{A}_H) \circ \cdots \circ (\tanh \circ \mathcal{A}_1),
\end{equation*}
where $\tanh$ is applied element-wise.  Each $\mathcal{A}_k : \mathbb{R}^{L_{k-1}} \rightarrow \mathbb{R}^{L_{k}}$ is an affine function of the form $\mathcal{A}_k(\bx) = W_k \bx + b_k$, with $W_k$ a ($L_{k-1} \times L_k$)-matrix,  $b_k \in \mathbb{R}^{L_k}$ a vector, 
$L_0 = d_1$, and $L_{H+1} = d_2$.
The neural network $f_{\theta}$ is thus parameterized by $\theta = (W_1, b_1, \hdots, W_{H+1}, b_{H+1}) \in \Theta_{H,D}$, where $\Theta_{H,D}=\mathbb{R}^{\sum_{i=0}^H (L_i+1) \times L_{i+1}}$.

\paragraph{The discretized version of the risk.} The idea behind PINNs is that, though minimizing $\mathscr{R}_n$ is difficult, it is possible to minimized the following discretized version by gradient descent 
\begin{align*}
R_{n, n_e, n_r}(f_{\theta}) &= \frac{\lambda_d}{n}\sum_{i=1}^{n} \|f_\theta(X_i)-Y_i\|_2^2 + \frac{\lambda_e}{n_e}\sum_{j=1}^{n_e} \|f_\theta(X^{(e)}_j)-h(X^{(e)}_j)\|_2^2 \\
& \quad + \frac{1}{n_r}\sum_{k=1}^M \sum_{\ell=1}^{n_r}  \mathscr{D}_k(f_\theta)(X^{(r)}_\ell)^2,
\end{align*}
where $n_e$ and $n_r$ are chosen by the practitioner, the $n_e$ points $X^{(e)}_j$ are sampled according to the distribution of $X^{(e)}$, and the $n_r$ collocation points $X^{(r)}_\ell$ are sampled according to the uniform distribution on $\Omega$.
Indeed, the gradient of $R_{n, n_e, n_r}$ with respect to $\theta$ can be efficiently computed by backpropagation.
However, $R_{n, n_e, n_r}(f_{\theta})$ is not convex in $\theta$, meaning that the gradient descent is not guaranteed to converge towards a global minimum.
In what follows, to simplify the analysis, we assume to have at hand a minimizing sequence $(\hat \theta(p, n_e, n_r, D))_{p\in \mathbb{N}} \in \Theta_{H,D}^\mathbb{N}$, i.e.,
\[ 
\lim_{p \to \infty}R_{n, n_e, n_r}(f_{\hat \theta(p, n_e, n_r, D)}) = \inf_{\theta \in \Theta_{H,D}}\,R_{n, n_e, n_r}(f_\theta).
\]
The hope is that, in the limit $n_e, n_r \to \infty$, minimizing the discretized risk $R_{n, n_e, n_r}$ is similar to minimizing the theoretical risk $\mathscr{R}_n$.
When both these minimizations are equivalent, i.e., $\lim_{n_e,n_r \to \infty}\lim_{p \to \infty}
\mathscr{R}_n(f_{\hat{\theta}(p, n_e,n_r, D)}) = \inf_{f \in \text{NN}_H(D)} \mathscr{R}_n(f)$, we say that PINNs are risk-consistent. 
Otherwise, we say that they overfit.

\paragraph{Contributions.} In this chapter, we prove the following theoretical results on the convergence of PINNs.
\begin{itemize}
    \item[(i)] [Proposition~\ref{prop:densite}] The class of neural networks is indeed able to approximate simultaneously any function and its derivatives. Formally, for all differentiation order $k \in \mathbb N$, this class is dense in the space $C^\infty(\bar \Omega, \mathbb R^{d_2})$ with respect to the $\|\cdot\|_{C^k(\Omega)}$ norm. This generalizes the result of \citet{ryck2021approximation}.
    \item[(ii)] [Propositions~\ref{prop:friction} and~\ref{prop:wave}] We have exhibited general cases where PINNs overfit.
    These results were later complemented by the NTK analysis of \citet{bonfanti2024challengesnonlinearregimephysicsinformed}.
    \item[(iii)] [Theorem~\ref{thm:generalization_error}] When adding a tailored ridge penalty $\|\theta\|_2^2$ to the discretized risk, PINNs become risk-consistent. This result is very general, as it covers systems of linear and nonlinear PDEs.
    \item[(iv)] [Examples~\ref{ex:dataIncorpo} and~\ref{ex:degeneratePINN}] Because of the challenging topological properties induced by the PDE penalty in the theoretical risk $\mathscr{R}_n$, risk-consistency is not enough to recover a physically-coherent neural network.
    \item[(v)] [Theorem~\ref{cor:sPINNsConsistency}] Adding a Sobolev penalty to the empirical risk ensures that, in the limit $p, n_e, n_r, D\to \infty$, the PINN $f_{\hat{\theta}(p, n_e,n_r, D)}$ converge to $f^\star$ at least at a $n^{-1/2}$ rate, and that $f_{\hat{\theta}(p, n_e,n_r, D)}$ respect the physical prior.
    This results is only proven for linear PDE systems.
    \item[(vi)] [Theorem~\ref{prop:pdeSolverFunctional}] We show how PDE solving can be seen as a particular instance of hybrid modeling without data, i.e., $n=0$. We show the convergence of the PINN to the unique solution of a PDE system, when the PDE is well-posed. This result complements those of \citet{shin2020convergence}, \citet{shin2023error}, \citet{mishra2022generalization}, \citet{ryck2022kolmogorov},   \citet{wu2022convergence}, and \citet{qian2023error} who focused on intractable modifications of PINNs.
    \item[(vii)] [Figure~\ref{fig:linReg}] We carry out numerical experiments, confirming empirically our results on the convergence rate of PINNs.
\end{itemize}

\paragraph{Remark on Sobolev spaces.} 
The notion of Sobolev space is central throughout all this manuscript. 
For instance, here, the Sobolev penalty used in PINNs is nothing but a squared Sobolev norm. 
Let $s\in \mathbb N$. 
The Sobolev space $H^s(\Omega, \mathbb R^{d_2})$ is a generalization of the Hölder space $C^s(\Omega, \mathbb R^{d_2})$ to functions that are not differentiable under the usual definition (called strong differentiability), but to a weaker sense (involving so-called weak derivatives). 
Formally, $H^s(\Omega, \mathbb R^{d_2})$ is the topological closure of $C^s(\Omega, \mathbb R^{d_2})$ with respect to the Sobolev norm $\|f\|_{H^s(\Omega)}^2 = \sum_{\alpha \in \mathbb N^{d_1}, \; \|\alpha\|_1\leq s} \|\partial^\alpha f\|_{L^2(\Omega)}^2$, where $\partial^{(\alpha_1, \hdots, \alpha_d)}f = \partial_1^{\alpha_1}\hdots \partial^{\alpha_d}_d f$.
This norm corresponds to the sum of the $L^2$ norms of the derivatives up to the order $s$.
For example, though the function $f:x\mapsto |x|$ is not differentiable at $x=0$ and thus does not belongs to $C^1([-1,1], \mathbb R)$, it belongs to $H^1([-1,1], \mathbb R)$.
This $L^2$ framework is particularly well-suited to PIML, where PDEs are penalized in the risk $\mathscr{R}_n$ by an $L^2$ penalty.
Sections~\ref{sec:notation} and~\ref{sec:rem_functional_ana} offer a more detailed introduction to weak derivatives, Sobolev spaces, and Lipschitz domains.

\subsection*{Chapter~\ref{ch:requirements}: PIML as a kernel method}
\textit{Physics-informed machine learning as a kernel method}, Nathan Doumèche, Francis Bach (INRIA Paris), Gérard Biau (Sorbonne Université), and Claire Boyer (Université Paris-Saclay). Published in the Proceedings of Thirty Seventh Conference on Learning Theory (COLT 2024).

\fbox{%
\begin{minipage}{\textwidth}
   In Chapter~\ref{ch:requirements}, we prove that for linear PDEs, PDE solving and hybrid modeling are kernel regression tasks. 
   By leveraging the theory of kernel methods, we show that the physics-informed estimator converges at least at the minimax Sobolev rate. 
   Faster rates can be achieved, highlighting the benefits of the physical prior.
\end{minipage}
}

\paragraph{Minimax convergence rate.} In Chapter~\ref{ch:my-domain}, we showed that PINNs with extra regularization terms are risk-consistent, and that they converge to $f^\star$ at a $n^{-1/2}$ rate if the PDE system is linear.
This result is satisfying because of its generality.
Indeed, it encompasses large classes of PDEs, holds for any Lispchitz domain $\Omega$, and only requires the initial condition $h$ to be Lipschitz.
An interesting results to compare with is the Sobolev minimax rate
\citep[see, e.g., Theorem 2.11,][]{tsybakov2009introduction}.
It states that no algorithm can learn an unknown function of the Sobolev ball $\{f, \|f\|_{H^s(\Omega, \mathbb{R})}\leq 1\}$ more quickly than the rate $n^{-2s/(2s+d_1)}$.
Note how the curse of the dimension appears in this rate because of its exponential dependency in $d_1$.
This rate is attained by many algorithms, like Sobolev kernel methods.
In particular, if the target function $f^\star$ is very smooth and belongs to $C^\infty(\bar\Omega, \mathbb R^{d_2})$, then it can be learnt at the parametric rate $n^{-1}$.
Thus, the rate of $n^{-1/2}$ that we computed for PINNs convergence is not optimal.

\paragraph{From linear regression to kernels.}  Our objective in this chapter is to rely on tools from kernel regression theory to better characterize the mathematical properties of PIML estimators.
Informally, a supervised learning task is said to be a kernel method if it can be cast as a linear regression with respect to some transformation $\phi$ of the features $X$.
For example, if $d_1 = d_2 =1$, the polynomial model $Y = \theta_1^\star + \theta_2^\star X + \theta^\star_3 X^2 + \hdots + \theta^\star_d X^{d-1} + \varepsilon$ is a linear model on the transformed feature $\phi(X) = (1, X, X^2, \hdots, X^{d-1})$.
In this setting, note that $f^\star(x) =  \langle \theta^\star, \phi(x)\rangle$.
The parameter $\theta^\star = (\theta^\star_1, \hdots, \theta^\star_{d})\in \mathbb R^d$ can thus be estimated from an i.i.d. sample $(X_1, Y_1)$, $\hdots$, $(X_n, Y_n)$ by finding the minimizer $\hat \theta$ over $\theta\in \mathbb R^d$ of the empirical risk $n^{-1}\sum_{j=1}^n|Y_j-\langle \theta, \phi(X_j)\rangle|^2 + \lambda \|\theta\|_2^2$, where $\lambda >0$ is a hyperparameter and $\|\theta\|_2^2$ is a ridge penalty.
This ridge regression admits the closed-form solution $\hat \theta = (\mathbb{\Phi}^\top \mathbb \Phi + \lambda \mathrm{Id})^{-1} \mathbb{\Phi}^\top \mathbb Y$, where $\mathbb{\Phi} = (\phi(X_1) \mid\hdots\mid \phi(X_n))^\top$ is the $n\times d$ feature matrix, $\mathbb Y = (Y_1, \hdots, Y_n)^\top \in \mathbb R^n$, and $\mathrm{Id}$ is the identity matrix.
Note that, in this case, $\mathbb{\Phi}^\top \mathbb \Phi$ is a $d \times d$ matrix, and so storing the matrix $\mathbb{\Phi}^\top \mathbb \Phi$ becomes computationally expensive as $d \to \infty$.  
Interestingly, the so-called "kernel trick" states that the estimated function $\hat f(x) = \langle \hat \theta, \phi(x)\rangle$ is also given by the formula $\hat f(x)(x) = \langle \hat \theta, \phi(x)\rangle = (K(x, X_1), \hdots, K(x, X_n)) (\mathbb K + \lambda n \mathrm{Id})^{-1} \mathbb Y$, where the function $K: (x,y)\in \Omega^2 \mapsto \langle \phi(x), \phi(y)\rangle\in \mathbb R$ is called the kernel function, and the kernel matrix $\mathbb K$ is the $n\times n$ Gram matrix $\mathbb K_{i,j} = \langle \phi(X_i), \phi(X_j)\rangle$. 

This formula allows to generalize this technique to infinite-dimensional maps $\phi$ (i.e., $d=\infty$) whenever the kernel function $K$ is well-defined.
Note that the Cauchy-Schwarz inequality states that $|K(x,y)|^2\leq K(x,x) K(y,y)$, and thus, one only needs to check that the diagonal terms $K(x,x)$ are well-defined for all $x\in \Omega$.
For instance, the infinite polynomial kernel of feature map $\phi(x) = (x^\ell)_{\ell \in \mathbb N}$ is well-defined on $\Omega = [-1/2,1/2]$, since $K(x,x) = \sum_{\ell \in \mathbb N} x^{2\ell} = \frac{1}{1-x^2}$ is bounded by $4/3$.
The kernel can then be recovered using the polarization identity: $K(x,y) = (K(x+y, x+y) - K(x-y, x-y))/4$.
Since any continuous function on $[-1/2, 1/2]$ can be approximated with arbitrary precision with polynomials (as a consequence of the Weierstrass theorem), the kernel $K(x,y) = (\frac{1}{1-(x+y)^2}-\frac{1}{1-(x-y)^2})/4$ can be used to perform nonparametric regression.

\paragraph{Convergence rate of kernels.} 
Building upon last example, we say that an estimator $\hat f$ is a kernel method if there exist a separable Hilbert space $(\mathcal{H},\; \langle\cdot, \cdot\rangle_\mathcal H)$, an hyperparameter $\lambda >0$, and a function $\phi: \Omega\to \mathcal{H}$ called the feature map, such that 
\begin{itemize}
    \item[(i)] the kernel function $K(x,y):=\langle \phi(x), \phi(y)\rangle$ is well-defined on $\Omega^2$,
    \item[(ii)] $\hat f(x) = (K(x, X_1), \hdots, K(x, X_n)) (\mathbb K + \lambda n \mathrm{Id})^{-1} \mathbb Y$.
\end{itemize}
Under the following assumptions on the distribution of $(X,Y)$ and on the regularity of the kernel, 
\begin{itemize}
    \item[(i)] the observations $(X_i, Y_i)$ are independent and identically distributed (i.i.d.),
    \item[(ii)] $\varepsilon$ is a noise (i.e., $\mathbb E(\varepsilon\mid X) = 0$) of bounded conditional variance (i.e., there is a constant $\sigma > 0$ such that $\mathbb E(\varepsilon^2\mid X) \leq \sigma^2$),
    \item[(iii)] the target function $f^\star$ is given by $f^\star(x) = \langle \theta^\star, \phi(x)\rangle$, for some $\theta^\star \in \mathcal H$, and
    \item[(iv)] the kernel $K$ is bounded on $\Omega^2$,
\end{itemize}
 \citet[see, e.g.,][Proposition~7.6]{bach2024learning} states that $\hat f$ converges to $f^\star$ at the following speed:
\begin{equation}
    \label{eq:noyau_bach}
    \mathbb E_{(X, Y)^{\otimes n}}\Big( \int |f^\star(x)-\hat f(x)|^2d\mathbb P_X(x)\Big) = O_{n\to \infty}(\lambda \|\theta^\star\|_{\mathcal H}^2  + \sigma^2\mathscr{N}(\lambda)n^{-1}),
\end{equation}
where the function $\mathscr{N}$ is effective dimension \citep{caponnetto2007optimal}.
This bound is minimax \citep[see, e.g.,][]{blanchard2020kernel}.
Note that the expectancy is taken with respect to the data set $(X_1, Y_1)$, $\dots$, $(X_n,Y_n) \sim (X,Y)^{\otimes n}$, the function $\hat f$ being then estimated from the observations.
Moreover, the distance between $f^\star$ and $\hat f$ is measured with respect to the measure $\mathbb P_X$ induced by the random variable $X$.
This is consistent with the fact the high-density regions for $\mathbb P_X$ are where we will sample the most data points $(X_i, Y_i)$, and therefore where we will learn $f^\star$ with more accuracy.
Here, the effective dimension is defined as the trace of the operator $(L_K+\lambda \mathrm{Id})^{-1}L_K$, where the integral kernel operator $L_K:  L^2(\Omega, \mathbb P_X) \to L^2(\Omega, \mathbb P_X) $ is defined by $L_K(g)(x) = \int_\Omega K(x,y) g(y)d\mathbb P_X(y)$. 
The only remaining difficulty consists in bounding $\mathscr{N}(\lambda)$.
There are general techniques to obtain bounds of the form $\mathscr{N}(\lambda) = O(\lambda^{\alpha})$, for some $\alpha >0$.
Taking $\lambda = n^{1/(1-\alpha)}$ minimizes the right-hand term in \eqref{eq:noyau_bach}, resulting in \[\mathbb E_{(X, Y)^{\otimes n}}\Big( \int |f^\star(x)-\hat f(x)|^2d\mathbb P_X(x)\Big) = O_{n\to \infty}(n^{1/(1-\alpha)}).\]


\paragraph{Contributions.} In this chapter, with have proven the following theoretical results on hybrid modeling and PDE solving. 
To simplify the setting, we assume here that the PIML problem is set with a unique PDE instead of a system (i.e., $M=1$), that $d_2=1$, and that we do not enforce boundary conditions (i.e., $\lambda_e = 0$). The case where $M >1$, $d_2 > 1$, and $\lambda_e > 0$ is an easy extension of the framework that will follow. In this context, $\hat f$ is a minimizer of the theoretical risk
\[\mathscr{R}_n(f) = \sum_{j=1}^n \|f(X_j)-Y_j\|_2^2 + \lambda_n \|f\|_{H^s(\Omega)}^2 + \mu_n\|\mathscr{D}(f)\|^2_{L^2(\Omega)}.\]
\begin{itemize}
    \item[(i)] [Theorem~\ref{thm:PDE_kernel}] Under the assumption that the PDE $\mathscr{D}$ is linear, both hybrid modeling and PDE solving are kernel methods.
    The minimizer $\hat f$ of $\mathscr{R}_n$ is therefore unique.
    \item[(ii)] [Proposition~\ref{prop:kernel_characterization}] The associated kernel $K$ is the unique solution to a weak PDE involving $\mathscr{D}$.
    \item[(iii)] [Proposition~\ref{prop:minSpeed}] When penalized by both the PDE penalty and the squared Sobolev norm, the PIML estimator converges at least at the Sobolev minimax rate $n^{-2s/(2s+d_1)}$, where the smoothness coefficient $s>d_1/2$ is such that $f^\star\in H^s(\Omega)$.
    \item[(iv)] [Theorem~\ref{prop:eigenfunction}] The eigenfunctions of $L_K$ are solutions of a PDE system.
    \item[(v)] [Theorem~\ref{prop:kernel_speed_up}] In the simple case where $d_1 = 1$, with $\Omega = [-L,L]$, $s=1$, $f^\star \in H^1(\Omega)$ and $\mathscr{D} = \frac{d}{dx}$, the kernel $K$ has an analytical expression and the eigenvalues of $L_K$ can be precisely bounded. We show in this case that 
    \begin{align*}
        \mathbb{E}\int_{[-L,L]} |\hat f-f^\star|^2 d{\mathbb P}_X &= \;\mathcal{O}_{n\to\infty} \big(  \|\mathscr{D}(f^\star)\|_{L^2(\Omega)} n^{-2/3}\log^3(n)+ \|f^\star\|_{H^1(\Omega)}^2 n^{-1} \log^3(n)\big) .
    \end{align*}
    If the ODE is exactly satisfies (i.e., $f^\star$ is constant), then the estimator converges at the parametric rate $n^{-1}$.
    Otherwise, we recover the $H^1(\Omega)$ minimax rate of $n^{-2/3}$.
    The modeling error $\|\mathscr{D}(f^\star)\|_{L^2(\Omega)}$ scales both convergence rates.
    \item[(vi)] [Figure~\ref{fig:numerical_experiment}] We carry out numerical experiment which confirm empirically our results on the convergence rate.
\end{itemize}

\paragraph{Kernel and estimators.} Since the unique minimizer of the theoretical risk $\mathscr{R}_n$ is $\hat f$, any algorithm minimizing $\mathscr{R}_n$ (be it PINNs, MCMCs...) will in fact be strictly equivalent to the kernel method of kernel $K$.
Thus, the theoretical properties established for the kernel method directly apply to all other methods minimizing $\mathscr{R}_n$.
Moreover, upon computing $K$, the kernel method associated to the kernel $K$ can also be seen as another algorithm for PIML tasks.

\subsection*{Chapter~\ref{ch:related-work}: Physics-informed kernel learning}
\textit{Physics-informed kernel learning}, Nathan Doumèche, Francis Bach (INRIA Paris), Gérard Biau (Sorbonne Université), and Claire Boyer (Université Paris-Saclay). In review: accepted with minor revisions in the Journal of Machine Learning Research (JMLR).

\fbox{%
\begin{minipage}{\textwidth}
   In Chapter~\ref{ch:related-work}, we rely on Fourier series to approximate the aforementioned kernel and propose an implementable estimator that minimizes the physics-informed risk function. We illustrate the estimator's performance through numerical experiments, both in the context of hybrid modeling and partial differential equation (PDE) solving.
\end{minipage}
}

\paragraph{Kernel approximation.} 
To implement the kernel method developed in Chapter~\ref{ch:requirements}, we need to compute or approximate the kernel $K$ given the PDE that it satisfies. 
The PDE literature has developed many methods to approximate the solutions of such PDEs, the most famous of these methods being the finite element method (FEM).
However, FEM requires to approximate the functions $K(X_i, \cdot)$ for each of the data point $X_i$, leading to a complexity $O(nh^3)$, where $h$ is the number of elements. 
Then, it requires to invert the kernel matrix $\mathbb K$, leading to a total complexity of $O(nh^3 + n^3)$.

\paragraph{Fourier expansion.}
In this chapter, we propose an algorithm with a better complexity, and that can be efficiently run on GPU. 
Instead of approximating the kernel $K$ directly, we discretize the associated Hilbert space.
This discretization consists in a low-frequency decomposition.  
Indeed, let $L$ be the length such that $\Omega \subseteq [-L,L]^{d_1}$.
For $k\in \mathbb Z^d$, we call $\phi_k$ the Fourier basis function $\phi_k(x)=(4L)^{-d/2} e^{\frac{i \pi}{2L}\langle k, x\rangle}$, and we approximate the Sobolev space $H^s(\Omega)$ by $H_m = \mathrm{Span}(\phi_k)_{\|k\|_\infty \leq m}$, where $m\geq 0$. 
The number of Fourier modes in $H_m$ is therefore $(2m+1)^d$.
Note that $f\in H_m$ if and only if there is a Fourier vector $z\in \mathbb C^{(2m+1)^d}$, such that $\forall x\in [-2L,2L]^{d_1}, \;f(x) = \sum_{k_1=-m}^m\hdots \sum_{k_{d_1}=-m}^m z_{k_1, \hdots, k_{d_1}}\phi_{k_1, \hdots, k_{d_1}}(x)$.
It is a non-trivial result that any function in $H^s(\Omega)$ can be expended into this Fourier basis (it is proven in Proposition~\ref{prop:dec_four_lip}).
Instead of penalizing the Sobolev norm $\|\cdot\|_{H^s(\Omega)}$, we consider the equivalent norm $\|\cdot\|_{H^s([-2L,2L]^{d_1})}$ of the Fourier decomposition on the extended domain $[-2L,2L]^{d_1}$.
Indeed, for all $f\in H_m$, the Sobolev norm $\|f\|_{H^s([-2L,2L]^{d_1})}^2 = \sum_{\|k\|_\infty \leq m} |z_k|_2^2 (1+ \|k\|_2^{2s})$ is easy to compute.
To stress that this norm is applied to Fourier series, which are by nature periodic, we denote $\|\cdot\|_{H^s_{\mathrm{per}}([-2L,2L]^{d_1})}$ instead of $\|\cdot\|_{H^s([-2L,2L]^{d_1})}$.

In this context, the kernel estimator is taken as a minimizer of the empirical risk
\[\bar{\mathscr{R}}_n(f)  = \sum_{j=1}^n \|f(X_j)-Y_j\|_2^2 + \lambda_n \|f\|_{H^s_{\mathrm{per}}([-2L,2L]^{d_1})}^2 + \mu_n\|\mathscr{D}(f)\|^2_{L^2(\Omega)}.\]
We call the resulting estimator the physics-informed kernel learner (PIKL).

\paragraph{Contributions.} In this chapter, with have proven the following theoretical results on the PIKL.
\begin{itemize}
    \item[(i)] [Section~\ref{sec:finite_approx}] The PIKL can be computed in $O(nh^2+h^3)$, where $h=(2m+1)^d$ is the number of Fourier modes. 
    Thus, it has a better complexity than the finite element method.
    Moreover, it can be efficiently implemented on GPU.
    \item[(ii)] [Theorem~\ref{thm:convergence_eff_dim}] The effective dimension of the PIKL problem converges as $m\to \infty$ to the effective dimension of the setting with $m=\infty$. 
    This corresponds to the same setting as the kernel method developped in Chapter~\ref{ch:requirements}.
\end{itemize}
We have also carried out the following experiments.
\begin{itemize}
    \item[(iii)] [Section~\ref{sec:finite_approx}] Direct approximation of the kernel $K$ from Chapter~\ref{ch:requirements} by the FEM. 
    \item[(iv)] [Section~\ref{sec:hybrid_mod}] Implementation of PIKL for hybrid modeling tasks and comparisons with other kernel methods.
    \item[(v)] [Section~\ref{sec:eff_dim}] Experimental evaluation of the effective dimension of several hybrid modeling tasks in dimensions $d_1=1$ and $d_1 = 2$.
    \item[(vi)] [Section~\ref{sec:PDE_solving}] Comparison of the PIKL, PINNs, and classical schemes in PDE solving.
    The PIKL significantly outperforms the PINNs and give similar results than the classical schemes.
    When the boundary conditions are noised, the PIKL outperforms all other methods.
\end{itemize}

\section{Time series forecasting in atypical periods}
The next three chapters are devoted to industrial applications in energy forecasting. Chapters~\ref{ch:contrib-1} and \ref{ch:contrib-2} are unrelated to physics-informed machine learning, while Chapter~\ref{ch:contrib-3} links both topics.

\subsection*{Chapter~\ref{ch:contrib-1}: Smarter Mobility Data Challenge }

\textit{Forecasting Electric Vehicle Charging Station Occupancy: Smarter Mobility Data Challenge}, Yvenn Amara-Ouali (Université Paris-Saclay), Yannig Goude (Université Paris-Saclay), Nathan Doumèche (Sorbonne Université), Pascal Veyret (EDF R\&D), et al. Published in the Journal of Data-centric Machine Learning Research (DMLR).

\fbox{%
\begin{minipage}{\textwidth}
   Chapter~\ref{ch:contrib-1} is a special chapter that describes the results of the three winning teams of the Smarter Mobility Data Challenge. 
   The goal of this challenge was to predict the occupancy of electric vehicle charging stations in Paris in 2021.
   Our team ranked 3rd in this challenge.
\end{minipage}
}

\paragraph{Smarter Mobility Data Challenge.}
The Smarter Mobility Data Challenge was organized in 2022 by the \textit{Manifeste IA} network, regrouping 16 French industrial (including EDF), and the European project TAILOR, which aims at promoting trustworthy artificial intelligence.
It gathered 169 participants and was open to all students from the European Union.
In this data challenge, the team of Nathan Doumèche and Alexis Thomas ranked 3rd.
Following the challenge, the data set was made public, and the results were published in the paper \citep{amara-ouali2024forecasting}.

\paragraph{Electric mobility forecasting.} 
The electric car market is emerging and evolves quickly \citep{IEA_EV_2022}.
As a result, energy providers must adapt the electricity network so that it can support the high-intensity needs related to charging an electric vehicles.
New solutions to accommodate these needs are being studied, such that pricing strategies, smart charging, and coupling with renewable production  \citep{dallinger2012grid, wang2016smart, alizadeh2017optimal, Moghaddam2018smart, crozier2020a, hafeez2023utilization}.  
However, their implementation requires a precise understanding of charging behaviors, and better EV charging models are necessary to grasp the impact of EVs on the grid \citep{Gopalakrishnan2016DemandPA, kaya2022electric, ciociola2023data, Andrenacci2023}. 
In particular, forecasting the occupancy of a charging station is a critical need for utilities to optimise their production units according to charging demand \citep{ZHANG2023}. 
On the user side, knowing when and where a charging station will be available is helful to find a parking place.
Nevertheless, large-scale datasets on EVs are rare \citep{calearo2021a, amara-ouali2021a}, which motivated this challenge.

\paragraph{Overview of the challenge.}
The goal of the challenge is to forecast the state of 91 charging stations for electric vehicle in Paris, each charging station being in either of the four states: \textit{available}, \textit{charging}, \textit{other}, or \textit{passive}.
The test period ranges from 19 February 2021 to 10 March 2021, while the data available to train the models ranges from 3 July 2020 to 18 February 2021.
The data set presents missing data, which limits the models which can be implemented.
The performance is measured according to a hierarchical structure of the problem.

\paragraph{Contributions.} 
In this chapter, we have provided the following contributions on forecasting the occupancy of electric vehicles.
\begin{itemize}
    \item[(i)] An open dataset on electric vehicle behaviors gathering both spatial and hierarchical features. Datasets with such features are rare and valuable for electric network management.
    \item[(ii)] An in-depth descriptive analysis of this dataset revealing meaningful user behaviors, such as work behaviors, daily and weekly patterns, and the impact of the pricing strategy.
    \item[(iii)] A detailed and reproducible benchmark for forecasting the EV charging station occupancy. This benchmark compares the winning solutions, as well as state-of-the-art forecasting models. One take-away is that neural network models did not perform well as compared to gradient boosting techniques.
    The best model, which is in fact the online aggregation of the three winning models, has an error about $40\%$ lower than the baseline model (consisting of forecasting the time series by their median).  
\end{itemize}
Apart from designing their model, the specific contributions of the team of Nathan Doumèche and Alexis Thomas are the following ones.
\begin{itemize}
    \item[(iv)] [Figure~\ref{fig:Other2}] We have evidenced that the data were not missing at random. Indeed, they tend to follow the state \textit{other}, corresponding to maintenance, and to be correlated between stations. 
    \item[(v)] [Table~\ref{fig:perf}] Our analysis showed that the data distribution was not stationary. Therefore, we trained our model in such a way that it gave more weight to recent observations. 
\end{itemize}

\subsection*{Chapter~\ref{ch:contrib-2}: Human spatial dynamics for electricity demand forecasting}

\textit{Human spatial dynamics for electricity demand forecasting}, Nathan Doumèche, Yannig Goude (Université Paris-Saclay), Stefania Rubrichi (Orange Innovation), and Yann Allioux (EDF R\&D). In review.

\fbox{%
\begin{minipage}{\textwidth}
  In Chapter~\ref{ch:contrib-2}, we explore the impact of work-related data on electricity demand forecasting. We demonstrate that mobility indices derived from mobile network data significantly enhance the performance of state-of-the-art models, particularly during France's energy sobriety period in the winter of 2022–2023.
\end{minipage}
}

\paragraph{Load forecasting and mobility data.} 
 Recently, machine learning techniques have been applied to load forecasting to ensure the electricity grid remains balanced \cite{pinheiro2023short} and to reduce electricity wastage. 
As France's electricity storage capacity is limited and expensive to run, electricity supply must match demand at all times. 
As a result, electricity load forecasting at different forecast horizons has attracted increasing interest over the last few years \cite{hong2020energy}. 
Here, we focus on the 24-hour ahead load forecasting, which is particularly relevant for operational usage in industry and the electricity market  \cite{nti2020review, hammad2020methods}. 
Most state-of-the-art models rely on historical electricity load data, seasonal data such as holidays or the position of the day in the week, and meteorological data such as temperature and humidity \cite{nti2020review}. 
However, such data cannot accurately account for complex human behaviours. 
As a result, traditional models struggle to account for unexpected large-scale societal events such as the COVID-19 lockdowns or energy savings following economic, geopolitical, and environmental crises \cite{obst2021adaptative}.
New datasets capturing social behaviors are therefore needed to better model electricity demand. 
Over recent decades, datasets generated from mobile networks, location-based services, and remote sensors in general, have been used to study human behavior \cite{blondel_understanding_2015}.
In terms of day-ahead load forecasting, mobility data from SafeGraph, Google, and Apple mobility reports were strongly correlated with electricity load drops in the US during the COVID-19 outbreaks \cite{chen2020using, ruan2020cross}, as well as in Ireland \cite{zarbakhsh2022human} and in France \cite{antoniadis2021hierarchical}. 
These works show that social behaviors like lockdowns and remote working significantly affect the electricity demand, and that these changes can be predicted by using mobility data.

\paragraph{Contributions.}
In this chapter, we establish the following experimental insights on load forecasting with mobility data.
\begin{itemize}
    \item[(i)] The mobility data from the Orange mobile network is correlated with other well-known socio-economic indices. Thus, it manages to quantify spatial dynamics related to mobility.
    \item[(ii)] We show that models using mobility data outperform the state-of-the-art in electricity demand forecasting by 10\% with respect to usual metrics.
    \item[(iii)] To better understand this result, we characterise electricity savings during the sobriety period in France. 
    \item[(iv)] Finally, we show that our {\it work} index has a distinctive effect on electricity demand, and is able to explain observed drops in electricity demand during holidays. 
Other human spatial dynamics indices such as tourism at the national level did not prove to have a significant effect on national electricity demand.
\end{itemize}

\subsection*{Chapter~\ref{ch:contrib-3}: Forecasting time series with constraints}

\textit{Forecasting time series with constraints}, Nathan Doumèche, Francis Bach (INRIA Paris), Eloi Bedek (EDF R\&D), Gérard Biau (Sorbonne Université), Claire Boyer (Université Paris-Saclay), and Yannig Goude (Université Paris-Saclay). In review.

\fbox{%
\begin{minipage}{\textwidth}
   In Chapter~\ref{ch:contrib-3}, we extend the Fourier framework from Chapter~\ref{ch:related-work} to incorporate constraints in time series analysis. Since macroeconomic time series rarely satisfy known PDEs, we focus on weak constraints, including additive models, online adaptation to structural breaks, hierarchical forecasting, and transfer learning. We demonstrate that the resulting kernel methods achieve state-of-the-art performance in load and tourism forecasting.
\end{minipage}
}

\paragraph{Weak constraints in time series.}
Forecasting time series presents unique challenges due to inherent data characteristics such as observation correlations, non-stationarity, irregular sampling intervals, and missing values. These challenges limit the availability of relevant data and make it difficult for complex black-box or overparameterized learning architectures to perform effectively, even with rich historical data \citep{lim2021time}. 
In this context, many modern frameworks incorporate constraints to improve the performance and interpretability of forecasting models. The strongest form of such constraints are typically derived from fundamental physical properties of the time series data and are represented by systems of differential equations. For example, weather forecasting often relies on solutions to the Navier-Stokes equations \citep[][]{schultz2021can}.
However, time series rarely satisfy strict differential constraints, often adhering instead to more relaxed forms of constraints \citep[][]{coletta2023on}.
Perhaps the most successful example of such weak constraints are the generalized additive models \citep[GAMs,][]{hastie1986generalized}, which have been applied to time series forecasting in epidemiology \citep{wood2017generalized}, earth sciences \citep{augusting2009modeling}, and energy forecasting \citep{fasiolo2021fast}. 
GAMs model the target time series (or some parameters of its distribution) as a sum of nonlinear effects of the features, thereby constraining the shape of the regression function.
Another example of weak constraint appears in the context of spatiotemporal time series with hierarchical forecasting. Here, the goal is to combine regional forecasts into a global forecast by enforcing that the global forecast must be equal to the sum of the regional forecasts \citep{Wickramasuriya2019optimal}.
Although this may seem like a simple constraint, hierarchical forecasting is challenging because of a trade-off: using more granular regional data increases the available information, but also introduces more noise as compared to the aggregated total. Another common and powerful constraint in time series forecasting arises when combining multiple forecasts \citep{gaillard2014second}. 
This is done by creating a final forecast by weighting each of the initial forecasts, with the constraint that the sum of the weights must equal one.

\paragraph{Contributions.} In this chapter, we have proven the following theoretical results on time series forecasting.
\begin{itemize}
    \item[(i)] We have developed a unified framework to integrate well-established constraints in time series: additive models, online adaption after a break, forecast combinations, transfer learning, hierarchical forecasting, and forecasting under differential constraints.
    \item[(ii)] We explicit a kernel which encodes each of the constraint. All the constraints can be effortlessly combined and efficiently implemented on GPU.
    \item[(iii)] [Proposition~\ref{prop:prop_lin}] We formally prove that adding linear constraints on $f^\star$ systematically improves the statistical performance. 
\end{itemize}
We also complement these theoretical results with the following experiments.
\begin{itemize}
    \item[(iv)] [Table~\ref{tab:ieee}] We implement the kernel method associated with the constraint of online adaption after a break to the IEEE DataPort Competition on Day-Ahead Electricity Load Forecasting. The resulting algorithm outperforms the winning team by about $10\%$.
    \item[(v)] [Table~\ref{table_score_target_agg2bid}] We implement the kernel method associated with the constraint of online adaption after a break to forecast the French load. The resulting algorithm outperforms the state-of-the-art by about $10\%$.
    \item[(vi)] [Table~\ref{table_australia}] We implement the kernel method associated with the hierarchical forecasting constraint to forecast the Australian domestic tourism. 
    The resulting algorithm outperforms the state-of-the-art reconciliation techniques by $7\%$.
\end{itemize}   

\part{Some mathematical insights on physics-informed machine learning}
\label{part:part1}
%

\chapter{On the convergence of PINNs}
\label{ch:my-domain}

This chapter corresponds to the following publication: \citet{doumeche2023convergence}.

\section{Introduction}

\noindent\textbf{Physics-informed machine learning}
Advances in machine learning and deep learning have led to significant breakthroughs in almost all areas of science and technology. However, despite remarkable achievements, modern machine learning models are difficult to interpret and do not necessarily obey the fundamental governing laws of physical systems \citep{linardatos2021explainability}. Moreover, they often fail to extrapolate scenarios beyond those on which they were trained \citep{xu2021extrapolation}. 
On the contrary, numerical or pure physical methods struggle to capture nonlinear relationships in complex and high-dimensional systems, while lacking flexibility and being prone to computational problems.  
This state of affairs has led to a growing consensus that data-driven machine learning methods need to be coupled with prior scientific knowledge based on physics. 
This emerging field, often called {physics-informed machine learning} \citep{raissi2019PINN}, seeks to combine the predictive power of machine learning techniques with the interpretability and robustness of physical modeling. 
The literature in this field is still disorganized, with a somewhat unstable nomenclature. In particular, the terms physics-informed, physics-based, physics-guided, and theory-guided are used interchangeably. 
For a comprehensive account, we refer to the reviews by \citet{rai2020review}, \citet{karniadakis2021piml}, \citet{cuomo2022scientific}, and \citet{Hao2022review}, which survey some of the prevailing trends in embedding physical knowledge in machine learning, present some of the current challenges, and discuss various applications.

\noindent\textbf{Vocabulary and use cases} Depending on the nature of the interaction between machine learning and physics, physics-informed machine learning is usually achieved by preprocessing the features \citep{rai2020review}, by designing innovative network architectures that incorporate the physics of the problem \citep{karniadakis2021piml}, or by forcing physics infusion into the loss function \citep{cuomo2022scientific}. 
It is this latter approach, which is most often referred to as physics regularization \citep{rai2020review}, to which our article is devoted.  Note that other names are possible, including physics consistency penalty \citep{wang2020superResolution}, knowledge-based loss term \citep{vonRueden2021informed}, and physics-guided neural networks \citep{cunha2022review}. In the following, we will focus more specifically on neural networks incorporating a physical regularization, called PINNs (for physics-informed neural networks, \citealt{raissi2019PINN}).
Such models have been successfully applied to $(i)$ model hybrid learning tasks, where the data-driven loss is regularized to satisfy a  physical prior, and $(ii)$ design efficient solvers of partial differential equations (PDEs).
A significant advantage of PINNs is that they are easy to implement compared to other PDE solvers, and that they rely on the backpropagation algorithm, resulting in reasonable computational cost.
Although $(i)$ and $(ii)$ are different facets of the same mathematical problem, they differ in their geometry and the nature of the data on which they are based, as we will see later.

\noindent\textbf{Related work and contributions} Despite a rapidly growing literature highlighting the capabilities of PINNs in various real-world applications, there are still few theoretical guarantees regarding {the} overfitting, consistency, and error analysis of the approach. Most existing theoretical work focuses either on intractable modifications of PINNs \citep{cuomo2022scientific} or on negative results, such as in \citet{krishnapriyan2021characterizing} and \citet{wang2022when}. 

Our goal in the present article is to provide a comprehensive theoretical analysis of the mathematical forces driving PINNs, in both the hybrid modeling and PDE solver settings, 
with the constant concern to provide approaches that can be implemented in practice. Our results complement those of \citet{shin2020convergence}, \citet{shin2023error}, \citet{mishra2022generalization}, \citet{ryck2022kolmogorov},   \citet{wu2022convergence}, and \citet{qian2023error} for the PDE solver problem. \citet{shin2020convergence} and \citet{wu2022convergence} focus on modifications of PINNs using the Hölder norm of the neural network in the loss function, which is unfortunately intractable in practice. In the context of linear PDEs, \citet{shin2023error} analyze the expected generalization error of PINNs using the Rademacher complexity of the image of the neural network class by a differential operator. However, this Rademacher complexity does not obviously vanish with increasing sample size. 
Similarly, \citet{mishra2022generalization} bound the generalization error by a quadrature rule depending on the Hölder norm of the neural network, which does not necessarily tend to zero as the number of training points tends to infinity. \citet{ryck2022kolmogorov} derive bounds on the expectation of the $L^2$ error, provided that the weights of the neural networks are bounded. 
In contrast to this series of works, we consider models and assumptions that can be practically verified or implemented. Moreover, our approach includes hybrid modeling, for which, as pointed out by \citet{karniadakis2021piml}, no theoretical guarantees have been given so far. Preliminary interesting results on the statistical consistency of a regression function penalized by a PDE are reported in \citet{arnone2022spatialRegression}. 
The original point of our approach lies in the use of a mix of statistical and functional analysis arguments \citep{evans2010partial} to characterize the PINN problem.

\noindent\textbf{Overview} After correctly defining the PINN problem in Section \ref{TPF}, we show in Section \ref{POF} that an additional regularization term is needed in the loss, otherwise PINNs can overfit. This first important result is consistent with the approach of \citet{shin2020convergence}, which penalizes PINNs by Hölder norms to ensure their convergence, and with the experiments of \citet{nabian2019engineering}, which improve performance by adding an extra-regularization term. In Section \ref{sec:consistency}, we establish the consistency of ridge PINNs by proving in Theorem \ref{thm:generalization_error} that a slowly va\-ni\-shing ridge penalty is sufficient to prevent overfitting.
Finally, in Section \ref{sec:functional}, we show that an additional level of regularization is sufficient in order to guarantee the strong convergence of PINNs (Theorem \ref{thm:functionalCv}).
We also prove that an adapted tuning of the hyperparameters allows to reconstruct the solution in the PDE solver setting (Theorem \ref{prop:pdeSolverFunctional}), as well as to ensure both statistical and physics consistency in the hybrid modeling setting (Theorem \ref{cor:sPINNsConsistency}). All proofs are postponed to \supplementary.
The code of all the numerical experiments can be found at \citet{supplement2023code} or at \url{https://github.com/NathanDoumeche/Convergence_and_error_analysis_of_PINNs}.

\section{The PINN framework}
\label{TPF}

In its most general formulation, the PINN method can be described as an empirical risk minimization problem, penalized by a PDE system.

\noindent\textbf{Notation} Throughout this article, the symbol $\mathbb E$ denotes expectation and $\|\cdot\|_2$ (resp.,  $\langle \cdot, \cdot\rangle$) denotes the Euclidean norm (resp., scalar product) in $\mathbb R^d$, where $d$ may vary depending on the context. 
Let $\Omega \subset \mathbb R^{d_1}$ be a bounded Lipschitz domain with boundary $\partial \Omega$ and closure $\bar \Omega$, and let $(\bX,Y) \in \Omega \times \mathbb{R}^{d_2}$ be a pair of random variables. Recall that Lipschitz domains are a general category of open sets that includes bounded convex domains (such as $]0,1[^{d_1})$ and usual manifolds with $C^1$ boundaries (see \appendixlink). 
This level of generality with respect to the domain $\Omega$ is necessary to encompass most of the physical problems, such as those presented in \citet{arzani2021uncovering}, which use non-trivial (but Lipschitz) geometries. 
For $K \in \mathbb{N}$, the space of functions from $\Omega$ to $\mathbb{R}^{d_2}$ that are $K$ times continuously differentiable is denoted by $C^K(\Omega, \mathbb{R}^{d_2})$. 

Let $C^\infty(\Omega, \mathbb{R}^{d_2}) = \cap_{K \geqslant 0}C^K(\Omega, \mathbb{R}^{d_2})$ be the space of infinitely differentiable functions. 
The space $C^K(\Omega, \mathbb{R}^{d_2})$ is endowed with the Hölder norm $\|\cdot\|_{C^K(\Omega)}$, 
defined for any $u$ by $\|u\|_{C^K(\Omega)} = \max_{|\alpha|\leqslant K} \|\partial^\alpha u\|_{\infty, \Omega}$. 
The space $C^\infty(\bar{\Omega}, \mathbb{R}^{d_2})$ of smooth functions is defined as the subspace of continuous functions $u:\bar{\Omega} \to \mathbb{R}^{d_2}$ satisfying $u|_\Omega \in C^\infty(\Omega, \mathbb{R}^{d_2})$ and, for all $K\in \mathbb{N}$, $\|u\|_{C^K(\Omega)} < \infty$. 
 A differential operator $\mathscr{F} : C^\infty(\Omega, \mathbb{R}^{d_2}) \times \Omega \to \mathbb{R}$ is said to be of order $K$ if it can be expressed as a function over the partial derivatives of order less than or equal to $K$. For example, the operator $\mathscr{F}(u, \bx) = \partial_1 u(\bx)  \partial^2_{1,2} u(\bx) + u(\bx)\sin(\bx) $ has order 2. 
 A summary of the mathematical notation used in this paper is to be found in \appendixlink.

\noindent\textbf{Hybrid modeling} As in classical regression analysis, we are interested in estimating the unknown regression function $u^\star$ such that $Y = u^\star({\bf X}) +\varepsilon$, for some random noise $\varepsilon$ that satisfies $\mathbb E(\varepsilon|\bX)=0$. 
What makes the problem original is that the function $u^{\star}$ is assumed to satisfy (at least approximately) a collection of $M \geqslant 1$ PDE-type constraints of order at most $K$, denoted in a standard form by $\mathscr{F}_k(u^\star,\bx)\simeq 0$ for $1 \leqslant k \leqslant M$. 
It is therefore assumed that $u^\star$ can be derived $K$ times. 
Moreover, there exists some subset $E \subseteq \partial \Omega$ and an boundary/initial condition function $h:E\to \mathbb R^{d_2}$ such that, for all $\bx\in E$, $u^\star(\bx) \simeq h(\bx)$. 
We stress that $E$ can be strictly included in $\Omega$, as shown in 
Example \ref{ex:spatioTemp} for a spatio-temporal domain $\Omega$. 
The specific case $E = \partial \Omega$ corresponds to Dirichlet boundary conditions. 

These constraints model some a priori physical information about $u^{\star}$. However, this knowledge may be incomplete (e.g., the PDE system may be ill-posed and have no or multiple solutions) and/or imperfect (i.e., there is some modeling error, that is, $\mathscr{F}_k(u^{\star},\bx)\neq 0$ and $u^\star|_E \neq h$).
This again emphasizes that $u^\star$ is not necessarily a solution of the system of differential equations. 
\begin{ex}[Maxwell equations]
Let $\bx = (x, y, z, t) \in \mathbb{R}^3 \times \mathbb{R}_+$, and consider  Maxwell equations describing the evolution of an electro-magnetic field $u^\star = (E^\star, B^\star)$ in vacuum, defined by
\[
\left\{\begin{array}{rcl}
      \mathscr{F}_1(u^\star, \bx) &=& \mathrm{div} E^\star(\bx)\\
      \mathscr{F}_2(u^\star, \bx) &=& \mathrm{div} B^\star(\bx)\\
      (\mathscr{F}_{3},\ \mathscr{F}_{4},\ \mathscr{F}_{5}) (u^\star, \bx) &=& \partial_t E^\star(\bx)  - \mathrm{curl} B^\star(\bx)\\
      (\mathscr{F}_{6},\ \mathscr{F}_{7},\ \mathscr{F}_{8})(u^\star, \bx) &=& \partial_t B^\star(\bx) + \mathrm{curl} E^\star(\bx),\\
\end{array}\right.
\]
where $E^\star\in C^1(\mathbb{R}^4, \mathbb{R}^3)$ is the electric field,  $B^\star \in C^1(\mathbb{R}^4, \mathbb{R}^3)$ the magnetic field, and the $\mathrm{div}$ and $\mathrm{curl}$ operators are respectively defined for $F = (F_x, F_y, F_z) \in C^1(\mathbb{R}^4, \mathbb{R}^3)$ by
\[ 
\mathrm{div} F = \partial_{x} F_x + \partial_{y} F_y + \partial_{z} F_z \quad \mbox{and} \quad
\mathrm{curl} F =  (\partial_y F_z - \partial_z F_y,\ \partial_z F_x - \partial_x F_z,\ \partial_x F_y - \partial_y F_x).
\]
In this case, $d_1=4$, $d_2 = 6$, and $M = 8$. 
\end{ex}

\begin{ex}[Spatio-temporal condition function]
\label{ex:spatioTemp}
    Assume that the domain $\Omega \subseteq \mathbb{R}^{d_1}$ is of the form $\Omega=\Omega_1 \times ]0,T[$, 
where $\Omega_1 \subseteq \mathbb{R}^{d_1-1}$ is a bounded Lipschitz domain and $T\geqslant 0$ is a finite time horizon. The spatio-temporal PDE system admits (spatial) boundary conditions specified by a function $f:\partial \Omega_1 \to \mathbb R^{d_2}$, i.e.,
\[
\forall x \in \partial \Omega_1, \ \forall t \in [0, T], \quad u^\star(x, t) = f(x),
\]
and a (temporal) initial condition specified by a function $g: \Omega_1 \to \mathbb R^{d_2}$, that is
\[
\forall x \in \Omega_1, \quad u^\star(x, 0) = g(x).
\]
The set on which the boundary and initial conditions are defined is $E = (\Omega_1\times \{0\}) \cup (\partial \Omega_1 \times [0,T]) $, and the associated condition function $h : E \to \mathbb{R}^{d_2}$ is
\[
h(\bx) = \left\{\begin{array}{lll}
       f(x) &\text{if} &\bx= (x,t) \in \partial \Omega_1 \times [0,T]\\
       g(x) &\text{if} &\bx=(x,t) \in \Omega_1\times \{0\}.
\end{array}\right.
\]
Notice that $E\subsetneq \partial \Omega$. 
\end{ex}
\hfill \\
In order to estimate $u^\star$, we assume to have at hand three sets of data:
\begin{itemize}
\item[$(i)$] A collection of i.i.d.~random variables $(\bX_1,Y_1), \hdots, (\bX_n,Y_n)$ distributed as $(\bX,Y) \in \Omega \times \mathbb R^{d_2}$, the distribution of which is \textit{ unknown};
\item[$(ii)$] A collection of i.i.d.~random variables $\bX^{(e)}_1, \hdots, \bX^{(e)}_{n_e}$ distributed according to some \textit{known} distribution $\mu_E$ on $E$;
\item[$(iii)$] A sample of i.i.d.~random variables $\bX^{(r)}_1, \hdots, \bX^{(r)}_{n_r}$ \textit{ uniformly distributed} on $\Omega$.
\end{itemize}
The function $u^\star$ is then estimated by minimizing the empirical risk function
\begin{align}
R_{n, n_e, n_r}(u_{\theta}) &= \frac{\lambda_d}{n}\sum_{i=1}^{n} \|u_\theta(\bX_i)-Y_i\|_2^2 + \frac{\lambda_e}{n_e}\sum_{j=1}^{n_e} \|u_\theta(\bX^{(e)}_j)-h(\bX^{(e)}_j)\|_2^2 \nonumber\\
& \quad + \frac{1}{n_r}\sum_{k=1}^M \sum_{\ell=1}^{n_r}  \mathscr{F}_k(u_\theta, \bX^{(r)}_\ell)^2\label{lossgenerale}
\end{align}
over the class $\text{NN}_H(D):=\{u_\theta, \theta \in \Theta_{H,D}\}$ 
of feedforward neural networks with $H$ hidden layers of common width $D$ (see below for a precise definition), where $(\lambda_d, \lambda_e) \in \mathbb{R}_+^2\backslash (0,0)$ are hyperparameters that establish a tradeoff between the three terms. 
In practice, one often encounters the case where $\lambda_e=0$ (data + PDEs).
Another situation of interest is when $\lambda_d=0$ (PDEs + boundary/initial conditions), which corresponds to the special case of a PDE solver. 
Setting \eqref{lossgenerale} is more general as it includes all the combinations data + PDEs + boundary/initial conditions. 
Since a minimizer of the empirical risk function \eqref{lossgenerale} does not necessarily exist, we denote by $(\hat \theta(p, n_e, n_r, D))_{p\in \mathbb{N}} \in \Theta_{H,D}^\mathbb{N}$ any minimizing sequence, i.e.,
\[ 
\lim_{p \to \infty}R_{n, n_e, n_r}(u_{\hat \theta(p, n_e, n_r, D)}) = \inf_{\theta \in \Theta_{H,D}}\,R_{n, n_e, n_r}(u_\theta).
\]
In practice, such a sequence is usually obtained by implementing some optimization procedure, the exact description of which is not important for our purpose.

On the practical side, simulations using hybrid modeling have been successfully applied to model image denoising \citep{wang2020superResolution},
turbulence \citep{wang2020turbulence},
blood streams \citep{arzani2021uncovering},
wave propagation \citep{davini2021using},
and ocean streams \citep{wolff2021ocean}. Experiments with real data have been performed to assess 
the sea temperature \citep{bezenac2017processes},
subsurface transport \citep{he2020subsurface},
fused filament fabrication \citep{kapusuzoglu2020manufacturing},
seismic response \citep{zhang2020seismic},
glacier dynamic \citep{riel2021glacier}, 
lake temperature \citep{daw2022lake},
thermal modeling of buildings \citep{gokhale2022thermal},  
 blasts \citep{pannell2022blast}, and
 heat transfers \citep{ramezankhani2022multifidelity}.
 The generality and flexibility of the empirical risk function \eqref{lossgenerale} allows it to encompass most PINN-like problems. For example, the case $M \geqslant 2$ is considered in \citet{bezenac2017processes} and \citet{riel2021glacier}, while \citet{zhang2020seismic} and \citet{wang2020turbulence} assume that $d_1 = d_2 = 3$. 
Importantly, the situation where $\lambda_d > 0$ {\it and} $\lambda_e>0$ (data + boundary conditions + PDEs) is also interesting from a physical point of view. This is, for example, the approach advocated by \citet{arzani2021uncovering}, which uses both data and boundary conditions (see also \citealp{cuomo2022scientific}, and \citealp{Hao2022review}).

\noindent\textbf{The PDE solver case} The particular case $\lambda_d=0$ deserves a special comment. 
In this setting, without physical measures $(\bX_i, Y_i)$, the function $u^{\star}$ is viewed as the unknown solution of the system of PDEs $\mathscr{F}_1, \hdots, \mathscr{F}_M$ with boundary/initial conditions $h$. 
The goal is to estimate the solution $u^{\star}$ of the PDE problem
\[
    \left\{\begin{array}{lrcl}
      \forall k, \, \forall \bx \in \Omega, & \mathscr{F}_k(u^{\star},\bx) &=& 0 \\
      \forall \bx \in E, & u^{\star}(\bx) &=& h(\bx),
\end{array}\right.
\]
with neural networks from $\mathrm{NN}_H(D)$.
In this case, the empirical risk function \eqref{lossgenerale} becomes 
\[
R_{n_e, n_r}(u_\theta) = \frac{\lambda_e}{n_e}\sum_{j=1}^{n_e} \|u_\theta(\bX^{(e)}_j)-h(\bX^{(e)}_j)\|_2^2  + \frac{1}{n_r}{\sum_{k=1}^M}\sum_{\ell=1}^{n_r} \mathscr{F}_k(u_\theta, \bX^{(r)}_\ell)^2,
\]
where the boundary and initial conditions $(\bX^{(e)}_1,h(\bX^{(e)}_1)), \hdots, (\bX^{(e)}_{n_e},h(\bX^{(e)}_{n_e}))$ are sampled on $E\times \mathbb{R}^{d_2}$ according to some known distribution $\mu_E$, and $(\bX^{(r)}_1, \hdots, \bX^{(r)}_{n_r})$ are uniformly distributed on $\Omega$.
Note that, for simplicity, we write $R_{n_e,n_r}(u_{\theta})$ instead of $R_{n, n_e,n_r}(u_{\theta})$ because no $\bX_i$ is involved in this context. Since no confusion is possible, the same convention is used for all subsequent risk functions throughout the paper. 
The first term of $R_{n_e,n_r}(u_{\theta})$ measures the gap between the network $u_\theta$ and the condition function $h$ on $E$, while the second term forces $u_{\theta}$ to obey the PDE in a discretized way. 
Since both the condition function $h$ and the distribution $\mu_E$ are known, it is reasonable to think of $n_e$ and $n_r$ as large (up to the computational resources).
In this scientific computing perspective, PINNs have been successfully applied to solve a wide variety of linear and nonlinear problems, including motion, advection, heat, Euler, high-frequency Helmholtz, Schr\"odinger, Blasius, Burgers, and Navier-Stokes equations, covering various fields ranging from classical (mechanics, fluid dynamics, thermodynamics, and electromagnetism) to quantum physics \citep[e.g.,][]{cuomo2022scientific, li2023aphysics}.

\noindent\textbf{The class of neural networks}  
A fully-connected feedforward neural network with $H\in\mathbb{N}^\star$ hidden layers of sizes $(L_1, \hdots, L_H) :=(D,\hdots ,  D) \in (\mathbb{N}^\star)^H$ and activation $\tanh$, is a function from $\mathbb R^{d_1}$ to $\mathbb R^{d_2}$, defined by
\begin{equation*}
    u_{\theta} = \mathcal{A}_{H+1}\circ (\tanh \circ \mathcal{A}_H) \circ \cdots \circ (\tanh \circ \mathcal{A}_1),
\end{equation*}
where the hyperbolic tangent function $\tanh$ is applied element-wise.  Each $\mathcal{A}_k : \mathbb{R}^{L_{k-1}} \rightarrow \mathbb{R}^{L_{k}}$ is an affine function of the form $\mathcal{A}_k(\bx) = W_k \bx + b_k$, with $W_k$ a ($L_{k-1} \times L_k$)-matrix,  $b_k \in \mathbb{R}^{L_k}$ a vector, 
$L_0 = d_1$, and $L_{H+1} = d_2$.
The neural network $u_{\theta}$ is parameterized by $\theta = (W_1, b_1, \hdots, W_{H+1}, b_{H+1}) \in \Theta_{H,D}$, where $\Theta_{H,D}=\mathbb{R}^{\sum_{i=0}^H (L_i+1) \times L_{i+1}}$.
Throughout, we let $\text{NN}_H(D)=\{u_{\theta}, \, \, \theta \in \Theta_{H,D}\}$.
We emphasize that the $\tanh$ function is the most common activation in PINNs \citep[see, e.g.,][]{cuomo2022scientific}.
It is preferable to the classical $\text{ReLU }(x)=\max(x,0)$ activation. In fact, since ReLU neural networks are a subset of piecewise linear functions, their high derivatives vanish and therefore cannot be captured by the penalty term $\frac{1}{n_r}\sum_{k=1}^M \sum_{\ell=1}^{n_r} \mathscr{F}_k(u_\theta, \bX^{(r)}_\ell)^2$.

The parameter space $\text{NN}_H(D)$ must be chosen large enough to approximate both the solutions of the PDEs and their derivatives. This property is encapsulated in Proposition \ref{prop:densite}, which shows that for any number $H \geqslant 2$ of hidden layers, the set $\text{NN}_H:=\cup_{D}\text{NN}_H(D)$ is dense in the space $(C^\infty(\bar{\Omega}, \mathbb{R}^{d_2}), \|\cdot\|_{C^K(\Omega)})$. This generalizes
Theorem 5.1 in \citet{ryck2021approximation} which states that $\text{NN}_2$ is dense in $(C^\infty([0,1]^{d_1}, \mathbb{R}), \|\cdot\|_{C^K(]0,1[^{d_1})})$ for all $d_1 \geqslant 1$ and $K \in \mathbb{N}$. 
\begin{prop}[Density of neural networks in Hölder spaces]
\label{prop:densite}
Let $K \in \mathbb{N}$, $H\geqslant 2$, and $\Omega \subseteq \mathbb{R}^{d_1}$ be a bounded Lipschitz domain. Then $\mathrm{NN}_H:=\cup_{D}\mathrm{NN}_H(D)$ is dense in $(C^\infty(\bar{\Omega}, \mathbb{R}^{d_2}), \|\cdot\|_{C^K(\Omega)})$, i.e., for any function $u\in C^\infty(\bar{\Omega}, \mathbb{R}^{d_2})$, there exists a sequence $(u_{p})_{p\in \mathbb{N}}\in \mathrm{NN}_H^{\mathbb{N}}$ such that $\lim_{p \to \infty} \|u-u_p\|_{C^K(\Omega)} = 0$. 
\end{prop}
In the remainder of the article, the number $H$ of hidden layers is considered to be fixed. 
\citet{krishnapriyan2021characterizing} use $\text{NN}_4(50)$, \citet{xu2021extrapolation} take $\text{NN}_5(100)$, whereas \citet{arzani2021uncovering} employ $\text{NN}_{10}(100)$. 
It is worth noting that in this series of papers the width $D$ is much larger than $H$, as in Proposition \ref{prop:densite}.

\section{PINNs can overfit}
\label{POF}
Our goal in this section is to show through two examples how learning with standard PINNs can lead to severe overfitting problems. This weakness has already been noted in \citet{costabal2020physics},  \citet{nabian2019engineering},  \citet{chandrajit2023recipes}, and \citet{esfahani2023adatadriven}, which propose to improve the performance of their models by resorting to an additional regularization strategy. The pathological cases that we highlight both rely on neural networks with exploding derivatives. 

The theoretical risk function is defined by 
\begin{equation}
\mathscr{R}_n(u) = \frac{\lambda_d}{n} \sum_{i=1}^n \|u(\bX_i) - Y_i\|_2^2 + \lambda_e \mathbb{E}\|u(\bX^{(e)})-h(\bX^{(e)})\|_2^2 + \frac{1}{|\Omega|}\sum_{k=1}^M \int_\Omega \mathscr{F}_k(u, \bx)^2 d\bx.\label{lossTheorical}
\end{equation}
Observe that in $\mathscr{R}_n(u)$ we take expectation with respect to $\mu_E$ (for the boundary/initial condition part) and integrate with respect to the uniform measure on $\Omega$ (for the PDE part), but keep the term $\sum_{i=1}^n \|u_{{\theta}}(\bX_i) - Y_i\|_2^2$ intact. 
This regime corresponds to the limit of the em\-pi\-ri\-cal risk function \eqref{lossgenerale}, holding $n$ fixed and letting  $n_e, n_r \to \infty$. The rationale is that while the random samples $(\bX_i,Y_i)$ may be limited in number (e.g., because their acquisition is more delicate and require physical measurements), this is not the case for $\bX^{(e)}_j$ or $\bX^{(r)}_j$, which can be freely sampled (up to computational resources).
Note however that in the PDE solver setting, the first term is not included. 

Given any minimizing sequence $(\hat{\theta}(p, n_e,n_r, D))_{p \in \mathbb{N}}$ of the empirical risk, satisfying
\[\lim_{p\to \infty} R_{n, n_e,n_r}(u_{\hat{\theta}(p, n_e,n_r, D)}) = \inf_{\theta \in \Theta_{H,D}}R_{n, n_e,n_r}(u_\theta),
\] 
a natural requirement, called risk-consistency, is that 
\[
\lim_{n_e,n_r \to \infty}\lim_{p \to \infty}
\mathscr{R}_n(u_{\hat{\theta}(p, n_e,n_r, D)}) = \inf_{u \in \text{NN}_H(D)} \mathscr{R}_n(u).
\]
We show below that standard PINNs can dramatically fail to be risk-consistent, through two counterexamples, one in the hybrid modeling context and one in the specific PDE solver setting.

\noindent\textbf{The case of dynamics with friction} 
Consider the following ordinary differential constraint, defined on the domain $\Omega = ]0, T[$ (with closure ${\bar{\Omega}} = [0,T]$) by
\begin{align}
\label{eq:friction_constraint}
\forall u\in C^2({\bar \Omega},\mathbb{R}), \ \forall \bx \in \Omega, \quad \mathscr{F}(u, \bx) &= mu''(\bx) + \gamma u'(\bx).
\end{align} 
This models the dynamics of an object of mass $m > 0$, subjected to a fluid force of friction coefficient $\gamma >0$. 
The goal is to reconstruct the real trajectory $u^\star$ by taking advantage of the model $\mathscr{F}$ and the noisy observations $Y_i$ at the $\bX_i$. This is an example where the modeling is perfect, i.e., $\mathscr{F}(u^\star, \cdot) = 0$, but the challenge is that the physical model is incomplete because the boundary conditions are unknown.
Following the hybrid modeling framework, the trajectory $u^{\star}$ is estimated by minimizing over the space $\text{NN}_H(D)$ the empirical risk function
\[
R_{n, n_r}(u_\theta) = \frac{\lambda_d}{n}\sum_{i=1}^{n} |u_\theta(\bX_i)-Y_i|^2 + \frac{1}{n_r}\sum_{\ell=1}^{n_r} \mathscr{F}(u_\theta, \bX^{(r)}_\ell)^2.
\] 
\begin{prop}[Overfitting] 
\label{prop:friction}
    Consider the dynamics with friction model \eqref{eq:friction_constraint}, and assume that there are two observations such that $Y_i \neq Y_j$. Then, whenever $D \geqslant n-1$, for any integer $n_r$, for all $\bX^{(r)}_1, \hdots, \bX^{(r)}_{n_r}$, there exists a minimizing sequence $(u_{\hat{\theta}(p, n_r, D)})_{p\in \mathbb{N}} \in \mathrm{NN}_H(D)^{\mathbb{N}}$ such that 
    $\lim_{p \to \infty} R_{n,n_r}(u_{\hat{\theta}(p, n_r, D)}) = 0$ but $ \lim_{p \to \infty} \mathscr{R}_n(u_{\hat{\theta}(p, n_r, D)}) = \infty$. So, this PINN estimator is not consistent.  
\end{prop}
\begin{figure}
    \centering
    \includegraphics[width = 0.7\textwidth]{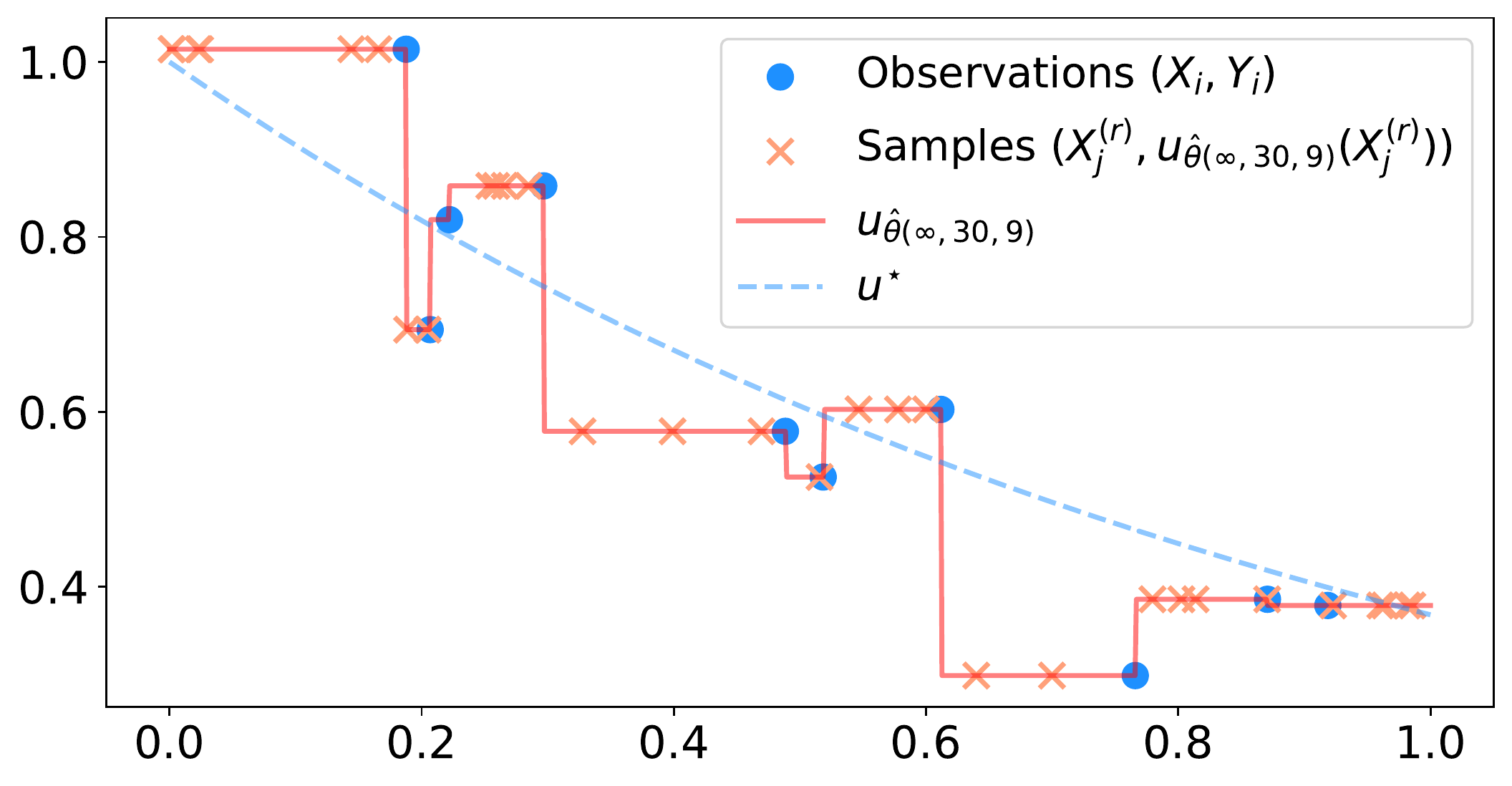}
    \caption{An inconsistent PINN estimator in hybrid modeling with $m = \gamma = 1$, $\varepsilon \sim \mathcal{N}(0, 10^{-2})$, and $n=10$. }
    \label{fig:failure_hybrid}
\end{figure}
Proposition \ref{prop:friction} illustrates how fitting a PINN by minimizing the empirical risk alone can lead to a catastrophic situation, where the empirical risk of the minimizing sequence is (close to) zero, while its theoretical risk is infinite. This phenomenon is explained by the existence of piecewise constant functions interpolating the observations $\bX_1,\hdots, \bX_n$, whose derivatives are null at the points $\bX^{(r)}_1, \hdots, \bX^{(r)}_{n_r}$, but diverge between these points (see Figure \ref{fig:failure_hybrid}). These functions correspond to neural networks $u_\theta$ such that $\|\theta\|_2 \to \infty$.

\begin{figure}
    \includegraphics[width = \textwidth]{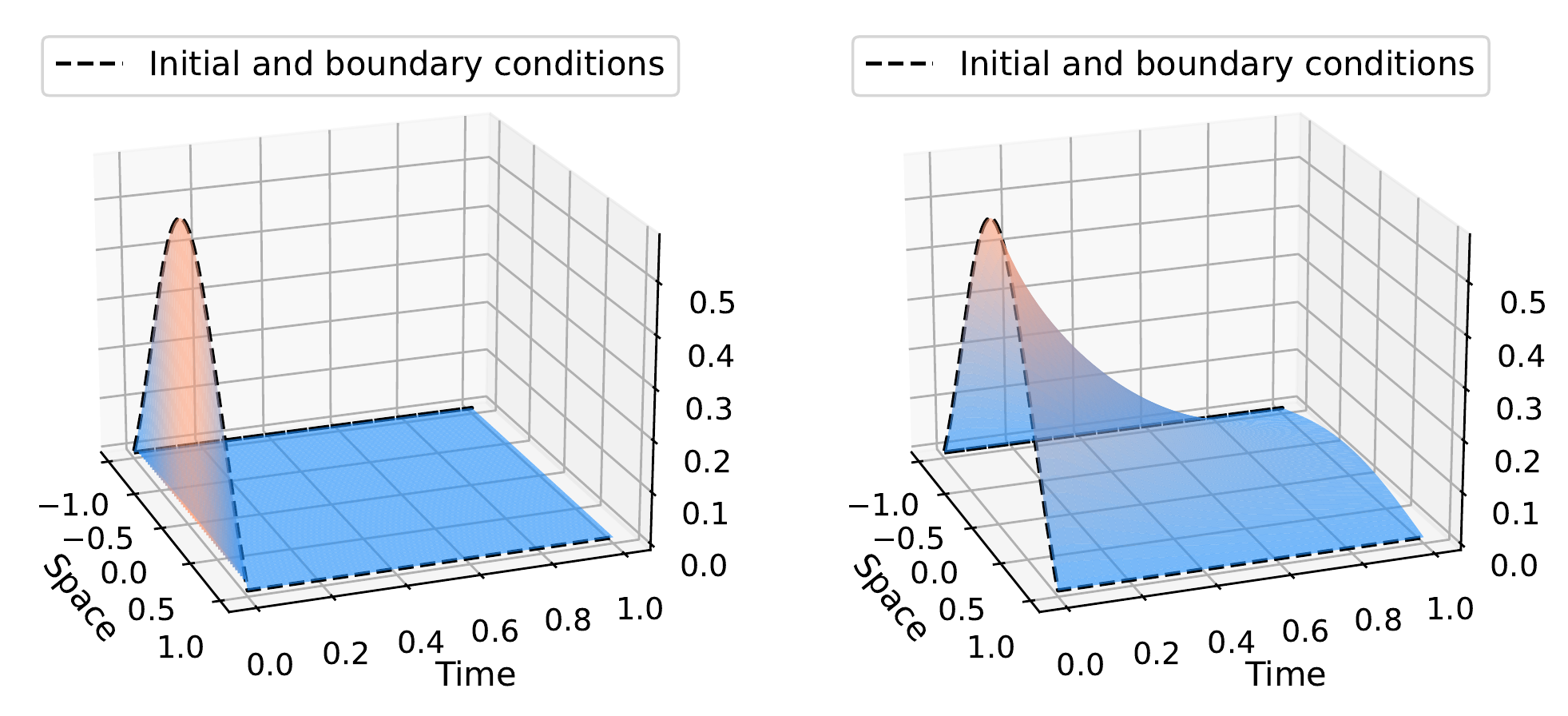}
    \caption{Inconsistent PINN (left) compared to the solution $u^\star$ of the PDE (right) for the heat propagation case.}
    \label{fig:failure_PDEsolving}
\end{figure}

\noindent\textbf{PDE solver: The heat propagation case} Consider the heat propagation differential operator defined on the domain $\Omega = ]-1, 1[ \times ]0,T[$ (with closure ${\bar \Omega} = [-1, 1] \times [0,T]$) by 
\begin{equation}
    \label{eq:wave_constraint}
    \forall  u \in C^2({\bar \Omega}, \mathbb{R}), \ \forall \bx \in \Omega, \quad \mathscr{F}(u,\bx) = \partial_{t}u(\bx) - \partial^2_{x,x}u(\bx),
\end{equation}
associated with the boundary conditions
\[\forall t \in [0,T], \quad u(-1, t) = u(1, t) = 0,\]
and the initial condition defined, for all $x \in [-1, 1]$, by
\[u(x,0) = \tanh^{\circ H}(x+0.5)-\tanh^{\circ H}(x-0.5) + \tanh^{\circ H}(0.5) - \tanh^{\circ H}(1.5). \]
The notation $\tanh^{\circ k}$ stands for the function recursively defined by $\tanh^{\circ 1} = \tanh$ and $\tanh^{\circ (k+1)} = \tanh \circ \tanh^{\circ k}$.  The unique solution $u^\star$ of the PDE is shown in Figure \ref{fig:failure_PDEsolving} (right). It models the time evolution of the temperature of a wire, whose extremities at $x=-1$ and $x=1$ are maintained at zero temperature. Note that the initial condition corresponds to a bell-shaped function, which belongs to $\text{NN}_H(2)$.
However, the setting can be extended to arbitrary initial conditions that take the form of a neural network function, given the boundary condition $u(\partial \Omega \times [0,T]) = \{0\}$. 

To solve the PDE \eqref{eq:wave_constraint}, we use $n_e$ i.i.d.~samples $\bX^{(e)}_1, \hdots, \bX^{(e)}_{n_e}$ on  $E = ([-1,1]\times\{0\}) \cup (\{-1, 1\}\times [0,T])$, distributed according to $\mu_E$, together with $n_r$ i.i.d.~samples $\bX^{(r)}_1, \hdots, \bX^{(r)}_{n_r}$, uniformly distributed on $\Omega$. Let $(\hat{\theta}(p, n_e, n_r, D))_{p \in \mathbb{N}}$ be a sequence of parameters  minimizing the empirical risk function
\[
R_{n_e, n_r}(u_\theta) = \frac{\lambda_e}{n_e}\sum_{j=1}^{n_e} |u_\theta(\bX^{(e)}_j)-h(\bX^{(e)}_j)|^2  + \frac{1}{n_r}\sum_{\ell=1}^{n_r} \mathscr{F}(u_\theta, \bX^{(r)}_\ell)^2,
\] 
over the space $\text{NN}_H(D)$. The theoretical counterpart of this empirical risk is
\[\mathscr{R}(u) =  \lambda_e \mathbb{E}|u(\bX^{(e)})-h(\bX^{(e)})|^2 + \frac{1}{|\Omega|}\int_\Omega \mathscr{F}(u, \bx)^2 d\bx.  \] 

\begin{prop}[PDE solver overfitting]
\label{prop:wave}
Consider the heat propagation model \eqref{eq:wave_constraint}. Then, 
 whenever $D \geqslant 4$, for any pair $(n_e, n_r)$, for all $\bX^{(e)}_1, \hdots, \bX^{(e)}_{n_e}$ and for all $\bX^{(r)}_1, \hdots, \bX^{(r)}_{n_r}$, there exists a minimizing sequence $(u_{\hat{\theta}(p, n_e, n_r, D)})_{p \in \mathbb{N}} \in \mathrm{NN}_H(D)^{\mathbb{N}}$ such that $\lim_{p\to \infty}R_{n_e, n_r}(u_{\hat{\theta}(p, n_e, n_r, D)}) = 0$ but $\lim_{p \to \infty}\mathscr{R}(u_{\hat{\theta}(p, n_e, n_r, D)}) = \infty$. So, this PINN estimator is not consistent. 
\end{prop}
Figure \ref{fig:failure_PDEsolving} (left) shows an example of an inconsistent PINN estimator.
Such an estimator corresponds to a function that equals zero on $\Omega$ (and thus satisfies the linear PDE), while satisfying the  initial condition on $\partial \Omega$. This function corresponds to a limit of neural networks $u_\theta$ such that $\|\theta\|_2\to \infty$.

The proof strategy of Propositions \ref{prop:friction} and \ref{prop:wave} does not depend on the geometry of the points $\bX^{(r)}$ and the points $\bX^{(e)}$, which could therefore be sampled along a grid, or by any quasi Monte Carlo method.
We emphasize that the two negative examples of Propositions \ref{prop:friction} and \ref{prop:wave} are no exceptions. In fact, their proofs can be easily generalized to differential operators $\mathscr{F}$ such that the following property holds: for all $\bx \in \Omega$, for all $u \in C^\infty(\Omega, \mathbb{R}^{d_2})$, if $\nabla u$ vanishes on an open set containing $\bx$, then $\mathscr{F}(u, \bx) = 0$.
This property is satisfied in the case of motion with friction, advection, heat, wave propagation, Schr\"odinger, Maxwell and Navier-Stokes equations, which are so as many cases that will suffer from overfitting.

\section{Consistency of regularized PINNs for linear and nonlinear PDE systems}
\label{sec:consistency}
Training PINNs can be tricky because it can lead to the type of pathological situations highlighted in Section \ref{POF}. To avoid such an overfitting behavior, a standard approach in machine learning is to resort to ridge regularization, where the empirical risk to be minimized is penalized by the $L^2$ norm of the parameters $\theta$. This technique has been shown to improve not only the optimization convergence 
during the training phase, but also the generalization ability of the resulting predictor \citep{krogh1991a, guo2017on}. Ridge regularization is available in most deep learning libraries (e.g., \texttt{pytorch} or \texttt{keras}),  where it is implemented using the so-called weight decay \citep{loshchilov2019decoupled}. 
Interestingly, the ridge re\-gu\-la\-ri\-za\-tion of a slight modification of PINNs, using adaptive activation functions, has been studied in \citet{jagtap2020adaptive}, which shows that gradient descent algorithms manage to generate an effective minimizing sequence of the penalized empirical risk. In this section, we formalize ridge PINNs and study their risk-consistency. 
\begin{defi}[Ridge PINNs]
    The ridge risk function is defined by
\begin{equation}
    R_{n, n_e,n_r}^{(\mathrm{ridge)}}(u_\theta) = R_{n, n_e,n_r}(u_\theta) + \lambda_{(\mathrm{ridge})} \|\theta\|_2^2, \label{eq:regPINN}
\end{equation}
where $\lambda_{(\mathrm{ridge})} >0$ is the ridge hyperparameter. We denote by $(\hat \theta_{(p, n_e, n_r, D)}^{(\mathrm{ridge})})_{p\in\mathbb{N}}$ a minimizing sequence of this risk, i.e., 
\[ 
\lim_{p \to \infty}R^{(\mathrm{ridge)}}_{n, n_e, n_r}(u_{\hat \theta_{(p, n_e, n_r, D)}^{(\mathrm{ridge})}}) = \inf_{\theta \in \Theta}\,R^{(\mathrm{ridge})}_{n, n_e, n_r}(u_\theta).
\]
\end{defi}

Our next Proposition \ref{prop:bounding} states that the $L^2$ norm of the parameters $\theta$ bounds the Hölder norm of the neural network $u_\theta$. This result is interesting in itself because it establishes a connection between the $L^2$ norm of a fully connected neural network and its regularity. (Note that, by equivalence of the norms, this result also holds if the ridge penalty is replaced by $\|\theta\|_p^p$.) In the present paper it plays a key role in the risk-consistency analysis.

\begin{prop}[Bounding the norm of a neural network by the norm of its parameter]
\label{prop:bounding}
    Consider the class $\mathrm{NN}_H(D)=\{u_\theta, \theta\in\Theta_{H,D}\}$. Let $K \in \mathbb{N}$.
    Then there exists a constant $C_{K,H} >0$, depending only on $K$ and $H$,
    such that, for all $ \theta \in \Theta_{H, D}$,
    \[
    \|u_\theta\|_{C^K(\mathbb{R}^{d_1})} \leqslant C_{K,H}(D+1)^{HK+1}(1+\|\theta\|_2)^{HK}\|\theta\|_2.
    \]
    Moreover, this bound is tight with respect to $\|\theta\|_2$, in the sense that, for all $H,D \geqslant 1$ and all $K\in \mathbb{N}$, there exists a sequence $(\theta_p)_{p \in \mathbb{N}} \in \mathrm{NN}_H(D)$ and a constant $\bar{C}_{K,H}>0$ such that $(i)$ $\lim_{p\to\infty}\|\theta_p\|_2 = \infty$ and $(ii)$ $\|u_{\theta_p}\|_{C^K(\mathbb{R}^{d_1})} \geqslant \bar{C}_{K,H}\|\theta_p\|_2^{HK+1}.$
\end{prop}
In order to study the generalization capabilities of regularized PINNs, we need to restrict the PDEs to a class of smooth differential operators, which we call polynomial operators (Definition \ref{defi:polyop} below). This class includes the most common PDE systems, as shown in the following example with the Navier-Stokes equations. 

\begin{ex}[Navier-Stokes equations]
\label{ex:navierS} Let $\Omega=\Omega_1 \times ]0,T[$, 
where $\Omega_1 \subseteq \mathbb{R}^{3}$ is a bounded Lipschitz domain and $T\geqslant 0$ is a finite time horizon.
The incompressible Navier-Stokes system of equations is defined for all $u = (u_x,u_y,u_z,p)\in C^2(\bar{\Omega}, \mathbb{R}^4)$ and for all $ \bx = (x,y,z,t) \in \Omega,$ by
\[ \left\{\begin{array}{rcl}
\mathscr{F}_1(u, \bx) &= &\partial_t u_x - (u_x \partial_x + u_y \partial_y + u_z \partial_z) u_x - \eta (\partial^2_{x,x}+\partial^2_{y,y}+\partial^2_{z,z}) u_x+ \rho^{-1} \partial_x p   \\
\mathscr{F}_2(u, \bx) &=& \partial_t u_y - (u_x \partial_x + u_y \partial_y + u_z \partial_z) u_y - \eta (\partial^2_{x,x}+\partial^2_{y,y}+\partial^2_{z,z}) u_y+ \rho^{-1} \partial_y p \\
         \mathscr{F}_3(u, \bx) &= &\partial_t u_z - (u_x \partial_x + u_y \partial_y + u_z \partial_z) u_z - \eta (\partial^2_{x,x}+\partial^2_{y,y}+\partial^2_{z,z}) u_z+ \rho^{-1} \partial_z p+g(\bx)\\
         \mathscr{F}_4(u, \bx) &= &\partial_x u_x + \partial_y u_y + \partial_z u_z,
    \end{array}\right. \]
where $\eta, \rho >0$ and $g\in C^{\infty}(\bar{\Omega}, \mathbb{R})$. Observe that $\mathscr{F}_1, \mathscr{F}_2, \mathscr{F}_3$, and $\mathscr{F}_4$ are polynomials in $u$ and its derivatives, with coefficients in $C^\infty(\bar{\Omega}, \mathbb{R})$. For example, $\mathscr{F}_3(u,\bx) = P_3(u_x, u_y, u_z, \partial_x u_z, \partial_y u_z, 
\partial_z u_z,$ $ \partial_t u_z, \partial^2_{x,x} u_z, \partial^2_{y,y} u_z, \partial^2_{z,z} u_z, \partial_z p)(\bx)$, where the polynomial $P_3\in C^\infty(\bar \Omega, \mathbb{R})[Z_1, \hdots, Z_{11}]$ is defined by $P_3(Z_1, \hdots, Z_{11}) = Z_7- Z_1Z_4 - Z_2Z_5 - Z_3Z_6- \eta (Z_8+Z_9+Z_{10}) + \rho^{-1}Z_{11} + g$.
\end{ex}

The above example can be generalized with the following definition. 
\begin{defi}[Polynomial operator]
\label{defi:polyop}
    An operator $\mathscr{F} : C^K(\bar{\Omega}, \mathbb{R}^{d_2})\times \Omega \to \mathbb{R}$  is  a polynomial operator
    of order $K \in \mathbb{N}$ if there exists an integer $s \in\mathbb{N}$ and multi-indexes $(\alpha_{i,j})_{1\leqslant i\leqslant d_2, 1\leqslant j\leqslant s} \in (\mathbb{N}^{d_1})^{sd_2}$ such that 
    \[\forall u=(u_1,\hdots,u_{d_2}) \in C^K(\bar{\Omega}, \mathbb{R}^{d_2}), \quad \mathscr{F}(u, \cdot) = P((\partial^{\alpha_{i,j}}u_i)_{1\leqslant i\leqslant d_2, 1\leqslant j\leqslant s}),\]
    where $P \in C^\infty(\bar{\Omega}, \mathbb{R})[Z_{1,1}, \hdots, Z_{d_2,s}]$ is a polynomial with smooth coefficients. 
\end{defi}
In other words, $\mathscr{F}$ is a polynomial operator if it is of the form
\[\mathscr{F}(u, \bx) = \sum_{k=1}^{N(P)} \phi_k \times \prod_{i=1}^{d_2}\prod_{j=1}^s (\partial^{\alpha_{i,j}}u_i(\bx))^{I(i,j,k)}, \]
where $N(P) \in \mathbb{N}^\star$, $\phi_k \in C^\infty(\bar{\Omega}, \mathbb{R})$, and $I(i,j,k) \in \mathbb{N}$. 
The associated polynomial is $P(Z_{1,1}, \hdots, Z_{d_2, s})$ $= \sum_{k=1}^{N(P)} \phi_k \times \prod_{i=1}^{d_2}\prod_{j=1}^s Z_{i,j}^{I(i,j,k)}$ (recall that $\partial^\alpha u_i = u_i$ when $\alpha = 0$). 
\begin{defi}[Degree] \label{defi:deg}
    The degree of the polynomial operator $\mathscr{F}$ is \[\mathrm{deg}(\mathscr{F}) = \max_{1\leqslant k \leqslant N(P)}\sum_{i=1}^{d_2}\sum_{j=1}^{s} (1+|\alpha_{i,j}|)I(i,j,k).\] 
\end{defi}
As an illustration, in Example \ref{ex:navierS}, one has $\deg(\mathscr{F}_3) = 3$, and this degree is reached in both the terms  $u_z\partial_z u_z$ and $\partial^2_{z,z}u_z$. 
 Note that $\deg(P_3) = 2$ but $\deg(\mathscr{F}_3) = 3$.
 To compute $\deg(\mathscr{F}_3)$, we first count the number of terms in each monomial ($u_z\partial_zu_z$ has two terms while $\partial^2_{z,z}u_z$ has one term), which is $\sum_{i=1}^{d_2}\sum_{j=1}^{s} I(i,j,k)$ for the $k$th monomial, and add the number of derivatives involved in the product ($u_z\partial_zu_z$ contains a single $\partial_z$ operator while $\partial^2_{z,z}u_z$ contains two derivatives in $\partial_z$), which corresponds to $\sum_{i=1}^{d_2}\sum_{j=1}^{s} |\alpha_{i,j}|I(i,j,k)$ for the $k$th monomial. Thus, for each monomial $k$, the total sum is $\sum_{i=1}^{d_2}\sum_{j=1}^{s} (1+|\alpha_{i,j}|)I(i,j,k)$.

We emphasize that this class includes a large number of PDEs, such as linear PDEs (e.g.,  advection, heat, and Maxwell equations), as well as some nonlinear PDEs (e.g., Blasius, Burger's, and Navier-Stokes equations). Proposition \ref{prop:bounding} is a key ingredient to uniformly bound the risk of PINNs involving polynomial PDE operators 
\supplementSecFive.
This in turn can be used to establish the risk-consistency of these PINNs when $n_e$ and $n_r$ tend to $\infty$, 
as follows.

\begin{thm}[Risk-consistency of ridge PINNs]
\label{thm:generalization_error}
Consider the ridge PINN problem \eqref{eq:regPINN}, over the  class $\mathrm{NN}_H(D)=\{u_\theta, \theta\in\Theta_{H,D}\}$,  where $H \geqslant 2$. 
Assume that the condition function $h$ is Lipschitz and that $\mathscr{F}_1, \hdots, \mathscr{F}_M$ are polynomial operators. 
Assume, in addition, that the ridge parameter is of the form
\[
\lambda_{(\mathrm{ridge})} = \min(n_e, n_r)^{-\kappa}, \quad \text{where} \quad  \kappa=\frac{1}{12+4H(1+(2+H)\max_k\deg(\mathscr{F}_k))}.
\] 

Then, almost surely, 
\[
\lim_{n_e, n_r \to \infty} \lim_{p\to \infty}\mathscr{R}_n(u_{\hat{\theta}^{(\mathrm{ridge})}(p, n_e, n_r, D)}) =  \inf_{u \in \mathrm{NN}_H(D)} \mathscr{R}_n(u).
\] 
\end{thm}
Thus, minimizing the ridge empirical risk \eqref{eq:regPINN} over $\Theta_{H,D}$ amounts to minimizing the theoretical risk \eqref{lossTheorical} over $\Theta_{H,D}$ in the asymptotic regime $n_e, n_r \to \infty$.
This fundamental result is complemented by the following one, which resorts to another asymptotics in the width $D$. This ensures that the choice of the neural architecture $\mathrm{NN}_H \subseteq C^\infty(\bar{\Omega}, \mathbb{R}^{d_2})$ does not introduce any asymptotic bias. 

\begin{thm}[The ridge PINN is asymptotically unbiased] 
    \label{thm:approximation}
    Under the same assumptions as in Theorem \ref{thm:generalization_error}, one has, almost surely, 
    \[
    \lim_{D \to \infty} \lim_{n_e, n_r \to \infty} \lim_{p\to \infty}\mathscr{R}_n(u_{\hat{\theta}^{(\mathrm{ridge})}(p, n_e, n_r, D)}) =  \inf_{u \in C^{\infty}(\bar{\Omega}, \mathbb{R}^{d_2})} \mathscr{R}_n(u).
    \]
\end{thm}
In other words, minimizing the ridge empirical risk over $\Theta_{H,D}$ and letting $D, n_e, n_r \to \infty$ amounts to minimizing the theoretical risk \eqref{lossTheorical} over the entire class $C^\infty(\bar{\Omega}, \mathbb{R}^{d_2})$.
We emphasize that these two theorems hold independently of the values of the hyperparameters $\lambda_d, \lambda_e \geqslant 0$. 
 Therefore, our results cover the general hybrid modeling framework \eqref{lossgenerale}, which includes the PDE solver. 
To the best of our knowledge, these are the first results that provide theoretical guarantees for PINNs regularized with a standard penalty. 
They complement the state-of-the-art approaches of \citet{shin2020convergence}, \citet{shin2023error}, \citet{mishra2022generalization}, and  \citet{wu2022convergence}, which consider re\-gu\-la\-ri\-za\-tion strategies that are unfortunately not feasible in practice.

It is worth noting that Theorem \ref{thm:approximation} still holds by choosing $D$ as a function of $n_e$ and $n_r$. 
In fact, an easy modification of the proofs reveals that one can take $D(n_e,n_r) = \min(n_e, n_r)^{\xi}$, where $\xi$ is a constant depending only on $H$ and $\max_k\deg(\mathscr{F}_k)$. Thus, in this setting, 
\[
\lim_{n_e, n_r \to \infty} \lim_{p\to \infty}\mathscr{R}_n(u_{\hat{\theta}^{(\mathrm{ridge})}(p, n_e, n_r, D(n_e,n_r)}) =  \inf_{u \in C^{\infty}(\bar{\Omega}, \mathbb{R}^{d_2})} \mathscr{R}_n(u).
\]

\begin{remark}[Dirichlet boundary conditions] Theorems \ref{thm:generalization_error} and \ref{thm:approximation} can be easily adapted to PINNs with Neumann conditions instead of Dirichlet boundary conditions. This is achieved by substituting the term $n_e^{-1}\sum_{j=1}^{n_e} \|u_\theta(\bX^{(e)}_j)-h(\bX^{(e)}_j)\|_2^2$ in \eqref{lossgenerale} by $n_e^{-1}\sum_{j=1}^{n_e}\|\partial_{\overrightarrow{n}}u_\theta(\bX^{(e)}_j)\|_2^2$, where $\overrightarrow{n}$ is the normal to $\partial \Omega$.
\end{remark}

\noindent\textbf{Practical considerations}
The decay rate of $\lambda_{(\mathrm{ridge})} = \min(n_e, n_r)^{-\kappa}$ does not depend on the dimension $d_1$ of $\Omega$. This is consistent with the results of \citet{karniadakis2021piml} and \citet{ryck2022kolmogorov}, which suggest that PINNs can overcome the curse of dimensio\-nality, opening up interesting perspectives for efficient solvers of high-dimensional PDEs.
We also emphasize that $\lambda_{(\mathrm{ridge})}$ depends only on the degree of the polynomial PDE operator,  the depth $H$, and the sample sizes $n_e$ and $n_r$.
All these quantities are known, which makes this hyperparameter immediately useful for practical applications.    
For example, in Navier-Stokes equations of Example \ref{ex:navierS}, one has $\max_k \deg(\mathscr{F}_k) = 3$. 
Thus, for a neural network of depth, say $H=2$, the ridge hyperparameter $\lambda_{(\mathrm{ridge})} = \min(n_e, n_r)^{-1/116}$ is sufficient to ensure consistency. 
It is also interesting to note that the bound on $\lambda_{(\mathrm{ridge})}$ in the theorems deteriorates with increasing  depth $H$. 
This confirms the preferential use of shallow neural networks in the experimental works of \citet{arzani2021uncovering}, \citet{karniadakis2021piml}, and \citet{xu2021extrapolation}. 
The bound also deteriorates as $\max_k \deg \mathscr{F}_k$ increases. 
This is in line with the empirical results of \citet{davini2021using}, which was able to improve the performance of PINNs by reformulating their polynomial differential equation of degree $3$ as a system of two polynomial differential equations of degree $2$.

It is also interesting to note that Theorems \ref{thm:generalization_error} and \ref{thm:approximation} hold for any ridge hyperparameter $\lambda_{(\mathrm{ridge})} \geqslant \min(n_e, n_r)^{-\kappa}$ such that $\lim_{n_e, n_r \to \infty} \lambda_{(\mathrm{ridge})} = 0$. 
However, if $n_e$ and $n_r$ are fixed, choosing too large a $\lambda_{(\mathrm{ridge})}$ will lead to a bias toward parameters of $\Theta_{H,D}$ with a low $L^2$ norm. 
Therefore, there is a trade-off between taking $\lambda_{(\mathrm{ridge})}$ as small as possible to reduce this bias, but large enough to avoid overfitting, as illustrated in Section \ref{POF}. 
Moreover, our choice of $\lambda_{(\mathrm{ridge})}$ may be suboptimal, since these results rely on inequalities involving a general class of polynomial operators. 
When studying a particular PDE, the consistency results of Theorems \ref{thm:generalization_error} and \ref{thm:approximation} should eventually hold with a smaller $\lambda_{(\mathrm{ridge})}$. 
To tune $\lambda_{(\mathrm{ridge})}$ in practice, one could, for example, monitor the overfitting gap  $\mathrm{OG}_{n, n_e, n_r} = |R_{n, n_e, n_r} - \mathscr{R}_n|$ for a ridge estimator $\hat{\theta}^{(\mathrm{ridge})}(p, n_e, n_r, D)$, by standard validation strategy (e.g., by sampling $\tilde{n}_r$ and $\tilde{n}_e$ new points to estimate $\mathscr{R}_n(u_{\hat{\theta}^{(\mathrm{ridge})}(p, n_e, n_r, D)})$ at a $\min(\tilde{n}_r, \tilde{n}_e)^{-1/2}$-rate given by the central limit theorem), and then choose the smallest parameter $\lambda_{(\mathrm{ridge})}$ to introduce as little bias as possible. More information about the relevance of $\mathrm{OG}_{n, n_e, n_r}$ is given in \supplementSecTwo.

\section{Strong convergence of PINNs for linear PDE systems}
\label{sec:functional}
Beyond risk-consistency concerns, the ultimate goal of PINNs is to learn a physics-informed regression function $u^\star$, or, in the PDE solver setting, to strongly approximate the unique solution $u^\star$ of a PDE system. 
Thus, what we want is to have guarantees regarding the convergence of $u_{\hat{\theta}^{(\mathrm{ridge})}(p, n_e, n_r, D)}$ to $u ^\star$ for an adapted norm. This requirement is called strong convergence in the functional analysis literature.
This is however not guaranteed under the sole convergence of the theo\-re\-ti\-cal risk $(\mathscr{R}_n(u_{\hat{\theta}^{(\mathrm{ridge})}(p, n_e, n_r, D)}))_{p,n_e, n_r,D \in \mathbb{N}}$, as shown in the following two examples.

\begin{ex}[Lack of data incorporation in the hybrid modeling setting]
\label{ex:dataIncorpo}
Suppose $M =1$, $d_1 = 2$, $d_2 = 1$, $\Omega = ]0,1[\times ]0,T[$, $h(x,0) = 1$ and $h(0, t) = 1$,  and let $\mathscr{F}(u, \bx) = \partial_x u(\bx) + \partial_t u(\bx)$. 
This corresponds to the assumption that the solution should approximately follow the advection equation and that it should be close to $1$. 
For any $\delta >0$, let the sequence $(u_{\delta, p})_{p \in \mathbb{N}}\in \mathrm{NN}_H(2n)^{\mathbb{N}}$ be defined by \[u_{\delta, p}(x,t) = 1+\sum_{i=1}^n \frac{Y_i}{2} \big(\tanh_p^{\circ H}(x-t - x_i + t_i + \delta)- \tanh_p^{\circ H}(x-t - x_i+t_i-\delta)\big), \] 
where $\tanh_p:=\tanh(p\,\cdot)$, and $\bX_i = (x_i, t_i)$. 
Then, as soon as $\delta \leqslant \frac{1}{2}\min_{i\neq j} |x_i-x_j + t_j-t_i|$, we have that $\lim_{p \to \infty}\mathscr{R}_n (u_{\delta, p}) = 0$.  
Thus, as long as $D \geq 2n$, $\inf_{u\in \mathrm{NN}_H(D)}\mathscr R_n(u) = 0$. Therefore, Theorem \ref{thm:approximation} shows that $\lim_{D \to \infty} \lim_{n_e, n_r \to \infty} \lim_{p\to \infty}\mathscr{R}_n(u_{\hat{\theta}^{(\mathrm{ridge})}(p, n_e, n_r, D)}) =  0.$ It is then easy to check that this implies that $u_{\hat{\theta}^{(\mathrm{ridge})}(p, n_e, n_r, D)}$ converges in $L^2(\Omega)$ to $1$, independently of $n$ and the function $u^{\star}$. This shows that the ridge PINNs fails to learn $u^\star$ whenever the model is inexact.
\end{ex}

In the PDE solver setting, one can consider the a priori favorable case where the PDE system admits a unique (strong) solution $u^\star$ in $C^K(\bar{\Omega}, \mathbb{R}^{d_2})$ (where $K$ is the maximum order of the differential operators $\mathscr{F}_1$, $\hdots$, $\mathscr{F}_{M}$). 
Note that $u^\star$ is the unique minimizer of $\mathscr{R}$ over $C^K(\bar{\Omega}, \mathbb{R}^{d_2})$, with $\mathscr{R}(u^\star) = 0$ (and $\mathscr{R}(u) = 0$ if and only if $u$ satisfies the initial conditions, the boundary conditions, and the system of differential equations). However, we describe below a situation where a minimizing sequence of $\mathscr{R}$ does not converge to the unique strong solution $u^\star$ of the PDE in question. 
\begin{ex}[Divergence in the PDE solver setting]
\label{ex:degeneratePINN}
Suppose $M=1$, $d_1 = d_2 = 1$, $\Omega = ]-1,1[$, $h(1) = 1$, $\lambda_e > 0$, and let the polynomial operator be $\mathscr{F}(u, \bx) = \bx u'(\bx)$. 
Clearly, $u^\star(\bx) = 1$ is the only strong solution of the PDE $\bx u'(\bx) = 0$ with $u(1) = 1$. 
Let the sequence $(u_p)_{p\in \mathbb{N}}\in \mathrm{NN}_H(D)^{\mathbb{N}}$ be defined by $u_p = \tanh_p\circ \tanh^{\circ (H-1)}$. According to  \supplementSecTwo, $\lim_{p \to \infty} \mathscr{R}(u_p) = \mathscr{R}(u^{\star}) = 0$.
However, the minimizing sequence $(u_p)_{p\in\mathbb{N}}$ does not converge to $u^\star$, since $ u_{\infty}(\bx) := \lim_{p \to \infty} u_p(\bx) = \mathbf{1}_{\bx > 0} -  \mathbf{1}_{\bx < 0}$. 
\end{ex}
We have therefore exhibited a sequence $(u_p)_{p \in \mathbb{N}}$ of neural networks that minimizes $\mathscr{R}$ and such that $(u_p)_{p\in\mathbb{N}}$ converges pointwise. However, its limit $u_\infty$ is not the unique strong solution of the PDE. In fact, $u_\infty$ is not differentiable at $0$, which is incompatible with the differential operators $\mathscr{F}$ used in $\mathscr{R}(u_\infty)$. Interestingly, the Cauchy-Schwarz inequality states that the pathological sequence $(u_p)_{p\in\mathbb{N}}$ satisfies $\lim_{p\to\infty}\|u_p'\|_{L^2(\Omega)}^2=\infty$, as in Example \ref{ex:dataIncorpo}.

\subsection*{Sobolev regularization}
The examples above illustrate how the convergence of the theoretical risk $\mathscr{R}_n(u_{\hat{\theta}^{(\mathrm{ridge})}(p, n_e, n_r, D)})$ to $\inf_{u\in C^\infty(\bar{\Omega}, \mathbb{R}^{d_2})}\mathscr{R}_n(u)$ (for any $n$) is not sufficient to guarantee the strong convergence to a PDE or hybrid modeling solution.
To ensure such a convergence, a different analysis is needed, mobilizing tools from functional analysis.
In the sequel, we build upon the regression estimation penalized by PDEs of \citet{azzimonti2015blood}, \citet{sangalli2021spatial}, \citet{ arnone2022spatialRegression}, and \citet{ferraccioli2022some}, and make use of the calculus of variations \citep[e.g.,][Theorems 1-4, Chapter 8]{evans2010partial}.
In the former references, the minimizer of $\mathscr{R}_n$ does not satisfy the PDE system injected in the PINN penalty, but another PDE system, known as the Euler-Lagrange equations.
Although interesting, the mathematical framework is different from ours. First, the authors do not study the convergence of neural networks, but rather methods in which the boundary conditions are hard-coded, such as the finite element method. Second, these frameworks are limited to special cases of theoretical risks. Indeed, only second-order PDEs with $\lambda_e=\infty$ are considered in \citet{azzimonti2015blood}, 
while \citet{evans2010partial} deal with first-order PDEs, echoing the case of $\lambda_d = 0$ and $ \lambda_e = \infty$.

It is worthwhile mentioning that the results of \citet{azzimonti2015blood}  rely on an important property of the theoretical risk function $\mathscr{R}_n$, called coercivity.  This is a common assumption of the calculus of variations \citep{evans2010partial}. The operator $\mathscr{R}_n$ is said to be coercive if there exist $K \in \mathbb{N}$ and $\lambda_t>0$ such that, for all $u\in H^K(\Omega, \mathbb{R}^{d_2})$, $\mathscr{R}_n(u) \geqslant \lambda_t \|u\|_{H^K(\Omega)}^2$ (the notation $H^K(\Omega, \mathbb{R}^{d_2}$) stands for the usual Sobolev space of order $K$---see \appendixlink. 
It turns out that the failures of Examples  \ref{ex:dataIncorpo} and \ref{ex:degeneratePINN} are due to a lack of coercivity, since, in both cases,  $\lim_{p \to\infty} \|u_p\|_{H^1(\Omega)}= \infty$ but $\lim_{p \to\infty} \mathscr{R}_n(u_p) \leqslant \mathscr{R}_n(u^\star)$. There are two ways to correct this problem: either one can restrict the study to coercive operators only, or one can resort to an explicit regularization of the risk to enforce its coercivity. We choose the latter, since most PDEs used in the practice of PINNs are not coercive. Note however that our results could be easily adapted to the coercive case. 

In the following, we restrict ourselves to affine operators, which exactly correspond to linear PDE systems, including  the advection, heat, wave, and Maxwell equations. 
\begin{defi}[Affine operator]
\label{defi:affine}
    The operator $\mathscr{F}$ is affine of order $K$ if there exists $A_{\alpha} \in C^\infty(\bar{\Omega}, \mathbb{R}^{d_2})$ and $B \in C^\infty(\bar{\Omega}, \mathbb{R})$  such that, for all $\bx \in \Omega$ and all $u \in H^K(\Omega, \mathbb{R}^{d_2})$, 
    \[\mathscr{F}(u, \bx) = \mathscr{F}^{(\mathrm{lin})}(u, \bx) + B(\bx),\]
    where  $\mathscr{F}^{(\mathrm{lin})}(u, \bx) = \sum_{|\alpha|\leqslant K} \langle A_{\alpha}(\bx), \partial^\alpha u(\bx)\rangle$ is linear.
\end{defi}

The source term $B$ is important, as it makes it possible to model a large variety of applied physical problems, as illustrated in \citet{song2021solving}. Note also that affine operators of order $K$ are in fact polynomial operators of degree $K+1$ (Definitions \ref{defi:polyop} and \ref{defi:deg}) that are extended from smooth functions to the whole Sobolev space $H^K(\Omega, \mathbb{R}^{d_2})$.

\begin{defi}[Regularized PINNs]
    The regularized theoretical risk function is
\begin{equation}
    \label{eq:regThRisk} \mathscr{R}^{(\mathrm{reg})}_n(u) = \mathscr{R}_n(u) + \lambda_t \|u\|_{H^{m+1}(\Omega)}^2,
\end{equation} 
where $\mathscr{R}_n$ is the original theoretical risk as defined in \eqref{lossTheorical}, and $m \in \mathbb{N}$. 
The corresponding regularized empirical risk function is
\begin{equation*}
    R_{n, n_e,n_r}^{(\mathrm{reg)}}(u_\theta) = R_{n, n_e,n_r}(u_\theta) + \lambda_{(\mathrm{ridge})} \|\theta\|_2^2 + \frac{\lambda_t}{n_\ell} \sum_{\ell=1}^{n_\ell} \sum_{|\alpha|\leqslant m+1}\|\partial^\alpha u_\theta(\bX_{\ell}^{(r)})\|_2^2.
\end{equation*} 
\end{defi}
It is noteworthy that $R_{n, n_e,n_r}^{(\mathrm{reg)}}$ can be straightforwardly implemented in the usual PINN framework and benefit from the computational scalability of the backpropagation algorithm, by encoding the regularization as supplementary PDE-type constraints $\mathscr{F}_\alpha(u, \bx) = \partial^\alpha u(\bx) = 0$. Since this discretized Sobolev penalty can be seen as additional physical priors $\mathscr{F}_\alpha$, the overfitting behavior observed for the unregularized PINNs can be transferred to Sobolev-regularized PINNs trained without ridge regularization. This is why the ridge penalty is still included in the risk.
Note also that the Sobolev regularization has been shown to avoid overfitting in machine learning, yet in different contexts \citep[e.g.,][]{fischer2020sobolev}.

The following proposition shows that the unique minimizer of \eqref{eq:regThRisk} can be interpreted as the unique minimizer of an optimization problem involving a weak formulation of the differential terms included in the risk. Its proof is based on the Lax-Milgram theorem \citep[e.g.,][Corollary 5.8]{brezis2010functional}.

\begin{prop}[Characterization of the unique minimizer of  $\mathscr{R}^{(\mathrm{reg})}_n$]
\label{prop:laxMLin}
    Assume that $\mathscr{F}_1, \hdots, \mathscr{F}_M$ are affine operators of order $K$. 
    Assume, in addition, that $\lambda_t >0$ and  $m \geqslant \max(\lfloor d_1/2\rfloor, K)$. Then the regularized theoretical risk $\mathscr{R}^{(\mathrm{reg})}_n$ has a unique minimizer $\hat u_n$ over $H^{m+1}(\Omega, \mathbb{R}^{d_2})$. This minimizer
    $\hat u_n$ is the unique element of $H^{m +1}(\Omega, \mathbb{R}^{d_2})$ that satisfies 
    \[\forall v \in H^{m+1}(\Omega, \mathbb{R}^{d_2}),\quad  \mathcal{A}_n(\hat u_n,v) = \mathcal{B}_n(v),\] 
     where
     \begin{align*}
        \mathcal{A}_n(\hat u_n,v) &= \frac{\lambda_d}{n} \sum_{i=1}^n \langle \tilde \Pi(\hat u_n)(\bX_i), \tilde \Pi(v)(\bX_i)\rangle +\lambda_e \mathbb{E}\langle\tilde \Pi(\hat u_n)(\bX^{(e)}),\tilde \Pi(v)(\bX^{(e)})\rangle\\
        & \quad +\frac{1}{|\Omega|}\sum_{k=1}^M\int_{\Omega} \mathscr{F}_k^{(\mathrm{lin})}(\hat u_n,\bx)\mathscr{F}_k^{(\mathrm{lin})}(v,\bx)d\bx\\
        & \quad + \frac{\lambda_t }{|\Omega|} \sum_{|\alpha|\leqslant m+1}\int_{\Omega}\langle \partial^\alpha \hat u_n(\bx) , \partial^\alpha v(\bx)\rangle d\bx,\\
        \mathcal{B}_n(v) &= \frac{\lambda_d}{n} \sum_{i=1}^n \langle Y_i, \tilde \Pi(v)(\bX_i)\rangle+ \lambda_e \mathbb{E}\langle\tilde \Pi(v)(\bX^{(e)}),h(\bX^{(e)})\rangle\\
        &\quad -\frac{1}{|\Omega|}\sum_{k=1}^M\int_{\Omega} B_k(\bx) \mathscr{F}_k^{(\mathrm{lin})}(v,\bx)d\bx,
    \end{align*}
    and where $\tilde{\Pi} : H^{m+1}(\Omega, \mathbb{R}^{d_2}) \to C^0(\Omega, \mathbb{R}^{d_2})$ is the so-called Sobolev embedding, such that $\tilde \Pi(u)$ is the unique continuous function that coincides with $u$ almost everywhere.
\end{prop}
The Sobolev embedding $\tilde{\Pi}$ is essential in order to give a precise meaning to the pointwise evaluation at the points $\bX_i$ of a function $u\in H^{m+1}(\Omega, \mathbb{R}^{d_2}) \subseteq L^2(\Omega, \mathbb{R}^{d_2})$, which is defined only almost everywhere.
The rationale behind Proposition \ref{prop:laxMLin} is that 
\[
\mathscr{R}_n^{(\mathrm{reg})}(u) = \mathcal{A}_n(u,u) -2\mathcal{B}_n(u) + \frac{\lambda_d}{n} \sum_{i=1}^n \|Y_i\|^2 +\lambda_e \mathbb{E}\|h(\bX^{(e)})\|_2^2 + \frac{1}{|\Omega|}\sum_{k=1}^M\int_{\Omega} B_k(\bx)^2d\bx.
\]
Therefore, minimizing $\mathscr{R}_n^{(\mathrm{reg})}$ amounts to minimizing $\mathcal{A}_n - 2 \mathcal{B}_n$.
It is also interesting to note that the  weak formulation $\mathcal{A}_n(\hat u,v) = \mathcal{B}_n(v)$ can be interpreted as a weak PDE on $H^{m+1}(\Omega, \mathbb{R}^{d_2})$.
In particular, if $\hat u_n \in H^{2(m+1)}(\Omega, \mathbb{R}^{d_2})$, then one has, almost everywhere,
\[
\sum_{k=1}^M (\mathscr{F}_k^{(\mathrm{lin})})^* \mathscr{F}_k(\hat u_n,\bx)+ \lambda_t  \sum_{|\alpha|\leqslant m+1} (-1)^{|\alpha|} (\partial^\alpha)^2 \hat u_n(\bx)  = 0.
\] 
$(\mathscr{F}_k^{(\mathrm{lin})})^*$ is the adjoint operator of $\mathscr{F}_k^{(\mathrm{lin})}$ such that, for all $ u, v \in C^\infty(\Omega, \mathbb{R})$ with $v|_{\partial \Omega} = 0$, 
\[
\int_\Omega u \mathscr{F}^{(\mathrm{lin})}(v, \bx) d\bx = \int_\Omega (\mathscr{F}_k^{(\mathrm{lin})})^*(u, \bx)vd\bx.
\]
Thus, even in the regime $\lambda_t \to 0$  (i.e., when the regularization becomes negligible), the solution of the PINN problem does not satisfy the constraints $\mathscr{F}_k(u,\bx) = 0$, but the following constraint $\sum_{k=1}^M (\mathscr{F}_k^{(\mathrm{lin})})^* \mathscr{F}_k(u,\bx) =0$. 
(Notice that, in the PDE solver setting, since $u^\star$ satisfies all the constraints, it satisfies in particular the constraint $\sum_{k=1}^M (\mathscr{F}_k^{(\mathrm{lin})})^* \mathscr{F}_k(u^\star,\bx) =0$.)
For instance, the advection equation constraint $\mathscr{F}(u, \bx) = (\partial_x + \partial_t) u(\bx)$ of Example \ref{ex:dataIncorpo} becomes $\mathscr{F}^* \mathscr{F}(u, \bx) = - (\partial_x + \partial_t)^2 u(\bx)$, and the constraint $\mathscr{F}(u, \bx) = \bx u'(\bx)$ of Example \ref{ex:degeneratePINN} becomes $\mathscr{F}^* \mathscr{F}(u, \bx) = -2\bx  u'(\bx)-\bx^2  u''(\bx)$.

Proposition \ref{prop:laxMLin} shows that the regularization in $\lambda_t$ is sufficient to make the PINN problem well-posed, i.e., to ensure that the theoretical risk function \eqref{eq:regThRisk} admits a unique minimizer.
The next natural requirement is that the regularized PINN estimator obtained by minimizing the regularized empirical risk function converges to this unique minimizer $\hat u_n$. 
Proposition \ref{prop:sequenceCvLin} and Theorem \ref{thm:functionalCv} show that this is true for linear PDE systems.
\begin{prop}[From risk-consistency to strong convergence]
    \label{prop:sequenceCvLin}
    Assume that $\lambda_t >0$ and  $m \geqslant \max(\lfloor d_1/2\rfloor, K)$. 
    Let $(u_p)_{p\in\mathbb{N}} \in C^\infty(\bar{\Omega}, \mathbb{R}^{d_2})$ be a sequence of smooth functions satisfying that $\lim_{p \to \infty}\mathscr{R}^{\mathrm{(reg)}}_n(u_p) = \inf_{u\in C^\infty(\bar \Omega, \mathbb{R}^{d_2})}\mathscr{R}^{\mathrm{(reg)}}_n$. Then $\lim_{p\to \infty} \|u_p - \hat u_n\|_{H^{m}(\Omega)}=0$, where $\hat u_n$ is the unique minimizer of $\mathscr{R}^{(\mathrm{reg})}_n$ over $H^{m+1}(\Omega, \mathbb{R}^{d_2})$. 
\end{prop}
The next theorem follows from Theorem \ref{thm:approximation} and Proposition \ref{prop:sequenceCvLin}, by simply observing that the Sobolev regularization is just an ordinary PINN regularization, taking the form of a polynomial operator of degree $(m+2)$.
\begin{thm}[Strong convergence of regularized PINNs]
\label{thm:functionalCv}
Assume that $\mathscr{F}_1, \hdots, \mathscr{F}_M$ are affine operators of order $K$. 
Assume, in addition, that $\lambda_t >0$, $m \geqslant \max(\lfloor d_1/2\rfloor, K)$, and the condition function $h$ is Lipschitz.
Let $(\hat{\theta}^{(\mathrm{reg})}(p, n_e, n_r, D))_{p \in \mathbb{N}}$ be a minimizing sequence of the regularized empirical risk function
    \begin{equation*}
        R_{n, n_e,n_r}^{(\mathrm{reg)}}(u_\theta) = R_{n, n_e,n_r}(u_\theta) + \lambda_{(\mathrm{ridge})} \|\theta\|_2^2 + \frac{\lambda_t}{n_\ell} \sum_{\ell=1}^{n_\ell} \sum_{|\alpha|\leqslant m+1}\|\partial^\alpha u_\theta(\bX_{\ell}^{(r)})\|_2^2
    \end{equation*}
    over the class $\mathrm{NN}_H(D)=\{u_\theta, \theta\in\Theta_{H,D}\}$, where $H \geqslant 2$.
    Then, with the choice
\[\lambda_{(\mathrm{ridge})} = \min(n_e, n_r)^{-\kappa}, \quad \text{where} \quad  \kappa=\frac{1}{12+4H(1+(2+H)(m+2))},\]
one has, almost surely, 
\[\lim_{D \to \infty} \lim_{n_e, n_r \to \infty} \lim_{p\to \infty} \|u_{\hat{\theta}^{(\mathrm{reg})}(p, n_e, n_r, D)} - \hat u_n\|_{H^m(\Omega)} = 0,\]
where $\hat u_n$ is the unique minimizer of $\mathscr{R}^{\mathrm{(reg)}}_n$ over $H^{m+1}(\Omega, \mathbb{R}^{d_2})$.
\end{thm}

Theorem \ref{thm:functionalCv} ensures 
that the sequence $u_{\hat{\theta}^{(\mathrm{reg})}(p, n_e, n_r, D)}$ of PINNs converges to the unique minimizer $\hat u_n$ of the regularized theoretical risk function \eqref{eq:regThRisk}, provided that the ridge hyperparameter $\lambda_{(\mathrm{ridge})}$ vanishes slowly enough.
However, it does not provide any information about the proximity between $u_{\hat{\theta}^{(\mathrm{reg})}(p, n_e, n_r, D)}$ and $u^\star$. 
On the other hand, since the regularized theoretical risk function is a small perturbation of the theoretical risk function \eqref{lossTheorical}, it is reasonable to think that its minimizer $\hat u_n$ should in some way converge to $u^\star$ as $\lambda_t \to 0$. This is encapsulated in Theorem \ref{prop:pdeSolverFunctional} for the PDE  solver setting and in Theorem \ref{cor:sPINNsConsistency} for the more general hybrid modeling setting.

\subsection*{The PDE solver case}
\begin{thm}[Strong convergence of linear PDE solvers]
    \label{prop:pdeSolverFunctional}
    Assume that $\mathscr{F}_1, \hdots, \mathscr{F}_M$ are affine operators of order $K$.
    Consider the PDE solver setting (i.e., $\lambda_e >0$ and $\lambda_d = 0$) and assume that the condition function $h$ is Lipschitz.  
    In addition, assume that the PDE system admits a unique solution $u^\star$ in $H^{m+1}(\Omega, \mathbb{R}^{d_2})$ for some $m \geqslant \max(\lfloor d_1/2\rfloor, K)$ (i.e., $u^\star$ is the unique function of $H^{m+1}(\Omega, \mathbb{R}^{d_2})$ such that 
    $\mathbb{E}\|u^\star(\bX^{(e)})-h(\bX^{(e)})\|_2^2 + \frac{1}{|\Omega|}\sum_{k=1}^M \int_\Omega \mathscr{F}_k(u^\star, \bx)^2 d\bx = 0$). Let $(\hat{\theta}^{(\mathrm{reg})}(p, n_e, n_r, D, \lambda_t))_{p \in \mathbb{N}}$ be a minimizing sequence of the regularized empirical risk function
    \begin{equation*}
        R_{n_e,n_r}^{(\mathrm{reg)}}(u_\theta) = R_{n_e,n_r}(u_\theta) + \lambda_{(\mathrm{ridge})} \|\theta\|_2^2 + \frac{\lambda_t}{n_\ell} \sum_{\ell=1}^{n_\ell} \sum_{|\alpha|\leqslant m+1}\|\partial^\alpha u_\theta(\bX_{\ell}^{(r)})\|_2^2
    \end{equation*}
    over the class $\mathrm{NN}_H(D)=\{u_\theta, \theta\in\Theta_{H,D}\}$, where $H \geqslant 2$. 
      Then, with the choice
    \[\lambda_{(\mathrm{ridge})} = \min(n_e, n_r)^{-\kappa}, \quad \text{where} \quad  \kappa=\frac{1}{12+4H(1+(2+H)(m+2))},\]
    one has, almost surely, 
    \[\lim_{\lambda_t \to 0}\lim_{D \to \infty} \lim_{n_e, n_r \to \infty} \lim_{p\to \infty} \|u_{\hat{\theta}^{(\mathrm{reg})}(p, n_e, n_r, D, \lambda_t)} -  u^\star\|_{H^m(\Omega)} = 0.\]
\end{thm} 
Back to Example \ref{ex:degeneratePINN}, one has $m = 1$.  
Recall that, in this setting, the unique minimizer of $\mathscr{R}$ over $C^0([-1,1], \mathbb{R})$ is $u^\star(\bx) = 1$, satisfying $u^\star \in H^2(]-1,1[, \mathbb{R})$. Therefore, by letting $\lambda_t \to 0$, this theorem shows that any sequence minimizing the regularized empirical risk function converges, with respect to the $H^2(\Omega)$ norm, to the unique strong solution $u^\star$ of the PDE $\bx u'(\bx) = 0$ and $u(1)=1$.  
\begin{remark}[Dimensionless hyperparameters and lower regularity assumptions on $u^\star$]
    The condition  $m \geqslant \lfloor d_1 / 2 \rfloor$ in Theorem \ref{thm:functionalCv} is necessary to make the pointwise evaluations $\tilde \Pi(u)(\bX_i)$ continuous.
    This condition does have an impact on $\lambda_{(\mathrm{ridge})}$, which grows exponentially fast with the dimension $d_1$. However, in the PDE solver setting, it is possible to get rid of this dimension problem, taking $m = \max_k\deg(\mathscr{F}_k)$. To see this, just note that there is no $\bX_i$, and so there is no need to resort to the $\tilde \Pi(u)(\bX_i)$.
    Indeed, the proof of 
    Theorem \ref{prop:pdeSolverFunctional} can be adapted by replacing the Sobolev inequalities in the proofs of Theorem \ref{thm:functionalCv} by the trace theorem for Lipschitz domains \citep[e.g.,][Theorem 1.5.1.10]{grisvard1985elliptic}. 
    In this case, it is enough to assume that $u^\star \in H^{K+1}(\Omega, \mathbb{R}^{d_2})$, which is less restrictive than $u^\star \in H^{\max(\lfloor d_1/2\rfloor, K)+1}(\Omega, \mathbb{R}^{d_2})$. However, this comes at the price of assuming that $\mu_E$ admits a density with respect to the hypersurface measure on $\partial \Omega$ (as it is often the case in practice).
\end{remark}

\subsection*{The hybrid modeling case}
To apply Theorem \ref{thm:functionalCv} to the general  hybrid modeling setting, it is necessary to measure the gap between $u^\star$ and the model specified by the constraints $\mathscr{F}_1, \hdots, \mathscr{F}_M$ and the condition function $h$. This is encapsulated in the next definition.
\begin{defi}[Physics inconsistency]
    For any $u \in H^{m+1}(\Omega, \mathbb{R}^{d_2})$, the physics inconsistency of $u$ is defined by
    \[\mathrm{PI}(u) = \lambda_e\mathbb{E}\|\tilde \Pi(u)(\bX^{(e)})-h(\bX^{(e)})\|_2^2 + \frac{1}{|\Omega|}\sum_{k=1}^{M}\int_\Omega \mathscr{F}_k(u,\bx)^2 d\bx.\] 
\end{defi}
Observe that $\mathscr{R}_n(u) = \frac{\lambda_d}{n} \sum_{i=1}^n \|\tilde \Pi(u)(\bX_i) - Y_i\|_2^2 + \mathrm{PI}(u)$.
In short, the quantity $\mathrm{PI}(u)$ measures how well the boundary/initial conditions, encoded by $h$, and the PDE system, encoded by the $\mathscr{F}_k$, describe the function $u$ \citep[see also][]{willard2023integrating}. In particular, $\mathrm{PI}(u^\star)$ measures the modeling error---the better the model, the lower $\mathrm{PI}(u^\star)$.

\begin{prop}[Strong convergence of hybrid modeling]
    \label{prop:consistencey} 
    Assume that the conditions of Theorem \ref{thm:functionalCv} are satisfied. 
    Then $\hat u_n\equiv \hat u_n(\bX_1, \hdots, \bX_n, \varepsilon_1, \hdots, \varepsilon_n)$ is a random variable such that $\mathbb{E}\|\hat u_n\|^2_{H^{m+1}(\Omega)} < \infty$.
    
   Suppose, in addition, that $u^\star \in H^{m+1}(\Omega, \mathbb{R}^{d_2})$,  that the noise $\varepsilon$ is independent from $\bX$, and that $\varepsilon$ has the same distribution as $-\varepsilon$. 
    Then there exists a constant $C_\Omega > 0$, depending only on $\Omega$, such that \begin{align*}
        \mathbb{E}\int_\Omega \|\tilde{\Pi} (\hat u_n - u^\star)\|_2^2d\mu_\bX &\leqslant \frac{1}{\lambda_d}\big(\mathrm{PI}(u^\star) + \lambda_t\|u^\star\|_{H^{m+1}(\Omega)}^2\big)\\
        &\quad + \frac{C_\Omega d_2^{1/2}}{n^{1/2}} \Big(2\|u^\star\|_{H^{m+1}(\Omega)}^2 + \frac{\mathrm{PI}(u^\star)}{\lambda_t}\Big)\\
        &\quad +\frac{8\mathbb{E}\|\varepsilon\|_2^2}{n}\Big(1+C_\Omega d_2^{3/2}\Big(\frac{\lambda_d}{\lambda_t} +\frac{\lambda_d^2}{\lambda_t^2n^{1/2}}\Big)\Big).
    \end{align*}
    In particular, with the choice $\lambda_e = 1$, $\lambda_t = (\log n)^{-1}$, and $\lambda_d = n^{1/2}/(\log n)$, one has
    \[\mathbb{E}\int_\Omega \|\tilde{\Pi} (\hat u_n - u^\star)\|_2^2d\mu_\bX \leqslant 
     \frac{\Lambda \log^2(n)}{n^{1/2} },\]
    where $\Lambda = 24d_2^{3/2}C_\Omega(\mathrm{PI}(u^\star) + \|u^\star\|_{H^{m+1}(\Omega)} + \mathbb{E}\|\varepsilon\|_2^2)$.
\end{prop}
This (nonasymptotic) proposition provides an insight into the scaling of the PINN hyperparameters. Indeed, the term $\frac{1}{\lambda_d}(\mathrm{PI}(u^\star) + \lambda_t\|u^\star\|_{H^{m+1}(\Omega)}) $ encapsulates the modeling error, damped by the weight $\lambda_d$. However, $\lambda_d$ cannot be arbitrarily large because of the term $\frac{8\mathbb{E}\|\varepsilon\|_2^2}{n}\big(1+C_\Omega d_2^{3/2}\big(\frac{\lambda_d}{\lambda_t} +\frac{\lambda_d^2}{\lambda_t^2n^{1/2}}\big)\big)$.
So, there is a trade-off between the modeling error and the random variation in the data.
Note also the other trade-off in the regularization hyperparameter $\lambda_t$, which should not converge to $0$ too quickly because of the term $\frac{C_\Omega d_2^{1/2}}{n^{1/2}} \big(2\|u^\star\|_{H^{m+1}(\Omega)}^2 + \frac{\mathrm{PI}(u^\star)}{\lambda_t}\big)$.
\begin{prop}[Physics consistency of hybrid modeling]
    \label{thm:phyCst}
     Under the conditions of Proposition \ref{prop:consistencey}, if $\lim_{n \to \infty} \frac{\lambda_d^2}{n \lambda_t} = 0$ and $\lim_{n \to \infty}\lambda_t = 0$, one has
    \[\mathbb{E}(\mathrm{PI}(\hat u_n)) \leqslant \mathrm{PI}(u^\star) + \oequivalent_{n\to \infty}(1).\]
    (Note that the conditions are satisfied with $\lambda_e = 1$, $\lambda_t = (\log n)^{-1}$, and $\lambda_d = n^{1/2}/(\log n)$.)
\end{prop}
As usual, we let $(u^{(n)}_{\hat{\theta}^{(\mathrm{reg})}(p, n_e, n_r, D)})_{p\in \mathbb{N}} \in \mathrm{NN}_H(D)^\mathbb{N}$ be a minimizing sequence of $R_{n, n_e,n_r}^{(\mathrm{reg)}}$, where the exponent $n$ indicates that the sample size $n$ is kept fixed along the sequence. 
Since $u^{(n)}_{\hat{\theta}^{(\mathrm{reg})}(p, n_e, n_r, D)} \in C^\infty(\bar \Omega, \mathbb{R}^{d_2})$, one has $\tilde \Pi(u^{(n)}_{\hat{\theta}^{(\mathrm{reg})}(p, n_e, n_r, D)}) = u^{(n)}_{\hat{\theta}^{(\mathrm{reg})}(p, n_e, n_r, D)}$. Thus, by combining Theorem \ref{thm:functionalCv} with Propositions \ref{prop:consistencey} and \ref{thm:phyCst}, we obtain the following important theorem.
\begin{thm}[Strong convergence of regularized PINNs]
\label{cor:sPINNsConsistency}
    Under the same assumptions as in 
    Theorem \ref{thm:functionalCv} and Proposition \ref{prop:consistencey}, with the choice $\lambda_e = 1$, $\lambda_t = (\log n)^{-1}$, and $\lambda_d = n^{1/2}/(\log n)$, one has 
    \[\lim_{D \to \infty} \lim_{n_e, n_r \to \infty} \lim_{p\to \infty} \mathbb{E}\int_\Omega \| u^{(n)}_{\hat{\theta}^{(\mathrm{reg})}(p, n_e, n_r, D)} -  u^\star\|_2^2d\mu_\bX \leqslant 
     \frac{\Lambda \log^2(n)}{n^{1/2} }\]
     and 
     \[\lim_{D \to \infty} \lim_{n_e, n_r \to \infty} \lim_{p\to \infty} \mathbb{E}(\mathrm{PI}(u^{(n)}_{\hat{\theta}^{(\mathrm{reg})}(p, n_e, n_r, D)})) \leqslant \mathrm{PI}(u^\star) + \oequivalent_{n\to \infty}(1).\]
\end{thm}
The minimax regression rate over any bounded class of functions in $C^{(m+1)}(\Omega, \mathbb{R}^{d_2})$ is known to be $n^{-2(m+1)/(2(m+1) +d_1)}$ \citep[][Theorem 1]{stone1982optimal}. Theorem \ref{cor:sPINNsConsistency} shows that the regularized PINN estimator achieves the rate $\log(n)/n^{1/2}$ over any \textit{larger} class bounded in $H^{(m+1)}(\Omega, \mathbb{R}^{d_2})$. Thus, the regularized PINN estimator has the nearly optimal rate, up to a log term, in the regime $d_1 \to \infty$ and $m = \lfloor d_1/2\rfloor$.

Theorem \ref{cor:sPINNsConsistency} shows that a properly regularized PINN estimator 
is both statistically \textit{and} physics consistent, in the sense that the error $\mathbb{E}\int_\Omega \| u^{(n)}_{\hat{\theta}^{(\mathrm{reg})}(p, n_e, n_r, D)} -  u^\star\|_2^2d\mu_\bX $ converges to zero with a physics inconsistency $ \mathbb{E}(\mathrm{PI}(u^{(n)}_{\hat{\theta}^{(\mathrm{reg})}(p, n_e, n_r, D)}))$ that is asymptotically no larger than $ \mathrm{PI}( u^\star)$.  It is also worth mentioning that in some applications, the physical measures $\bX_1, \hdots, \bX_n$ are forced to be sampled in certain subset of $\Omega$. 
An important application is when $\Omega$ is spatio-temporal and one wishes to extrapolate/transfer a model from a training dataset collected on $\mathrm{supp}(\mu_\bX) = \Omega_1 \times ]0,T_{\mathrm{train}}[$ to a test $\Omega_1 \times ]T_{\mathrm{train}},T_{\mathrm{test}}[$, using a temporal evolution PDE system to extrapolate \citep[e.g., ][]{cai2021physics}. On the other hand, the physical restriction on the data measurement can be also strictly spatial. 
This  is for example the case in some blood modeling problems, where the blood flow measures can only be taken in a specific region of a blood vessel, as illustrated in \citet{arzani2021uncovering}. Thus, in both these contexts, the support $\mathrm{supp}(\mu_\bX)$ of the distribution $\mu_\bX$ is strictly contained in $\Omega$.
Of course, this is compatible with Theorem \ref{cor:sPINNsConsistency}, which shows that the regularized PINN estimator consistently interpolates the function $u^\star$ on $\mathrm{supp}(\mu_\bX)$. Furthermore, Theorem \ref{cor:sPINNsConsistency} shows that the estimator uses the physical model to extrapolate on $\Omega \backslash \mathrm{supp}(\mu_\bX)$. 
In summary, the better the model, the lower the modeling error $\mathrm{PI}(u^\star)$, and the better the domain adaptation capabilities. 
This provides an interesting mathematical insight into the relevance of combining data-driven statistical models with the interpretability and extrapolation capabilities of physical modeling.

\noindent\textbf{Numerical illustration of imperfect modeling}  In the following experiments, we  
illustrate with a toy example the results of Theorem \ref{cor:sPINNsConsistency} and show  how the Sobolev regularization can be implemented directly in the PINN framework, taking advantage of the automatic differentiation and backpropagation. 
Let $\Omega = ]0,1[^2$ and assume that $Y =  u^\star(\bX) + \mathcal{N}(0,10^{-2})$, where 
$u^\star(x, t) = \exp(t-x) + 0.1 \cos(2\pi x)$. 
In this hybrid modeling setting, the goal is to reconstruct $u^\star$.
We consider  an advection model of the form $\mathscr{F}(u, \bx) = \partial_x u(\bx) + \partial_t u(\bx)$, with $h(x, 0) = \exp(-x)$ and $h(0, t) = \exp(t)$. The unique solution of this PDE is $u_{\mathrm{model}}(x, t) = \exp(t-x)$ (Figure \ref{fig:hybrid_modelling_nn}, left). Note that the function $u_{\mathrm{model}}$ is different from $u^\star$ (Figure \ref{fig:hybrid_modelling_nn}, middle), which casts our problem in the imperfect modeling setting. This PDE prior is relevant because
$\|u_{\mathrm{model}}-u^\star\|_{L^2(\Omega)}^2 \simeq \exp(-5.3)$ and $\mathrm{PI}(u^\star) \simeq \exp(-1.6)$, two quantities that are negligible with respect to $\|u^\star\|_{L^2(\Omega)}^2\simeq \exp(0.3)$. 
We randomly sample $n$ observations $\bX_1, \hdots, \bX_n$ uniformly on the rectangle $\mathrm{supp}(\mu_\bX) = ]0,0.5[\times ]0,1[ \subsetneq \Omega$ (note that this is a strict inclusion), and let $n$ vary from $n_{\min} = 10$ to $n_{\max} = 10^3$ (linearly in a log scale).  

\begin{figure}
    \centering
    \vspace{-0.4cm}
    \includegraphics[width = 0.45\textwidth]{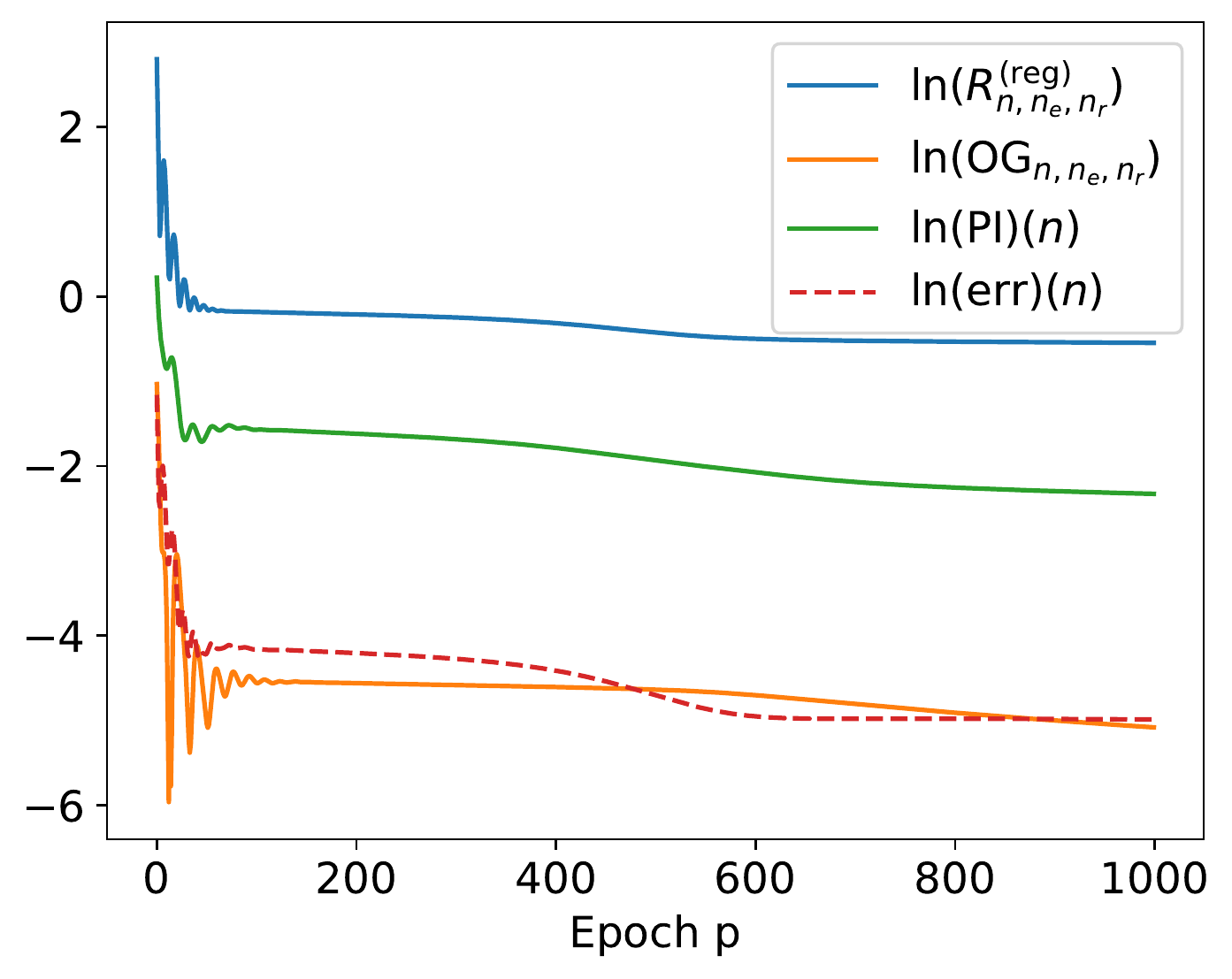}    \caption{Regularized empirical risk (blue) and overfitting gap $\mathrm{OG}$ (orange) with respect to the number $p$ of epochs for $n=10$. The physics inconsistency $\mathrm{PI}(n)$ (green) and the $L^2$ error $\mathrm{err}(n)$ (red) are also depicted.}
    \label{fig:monitoring}
\end{figure}

\begin{figure}
    \centering
    \begin{tabular}{cc}
    \includegraphics[width = 0.45\textwidth]{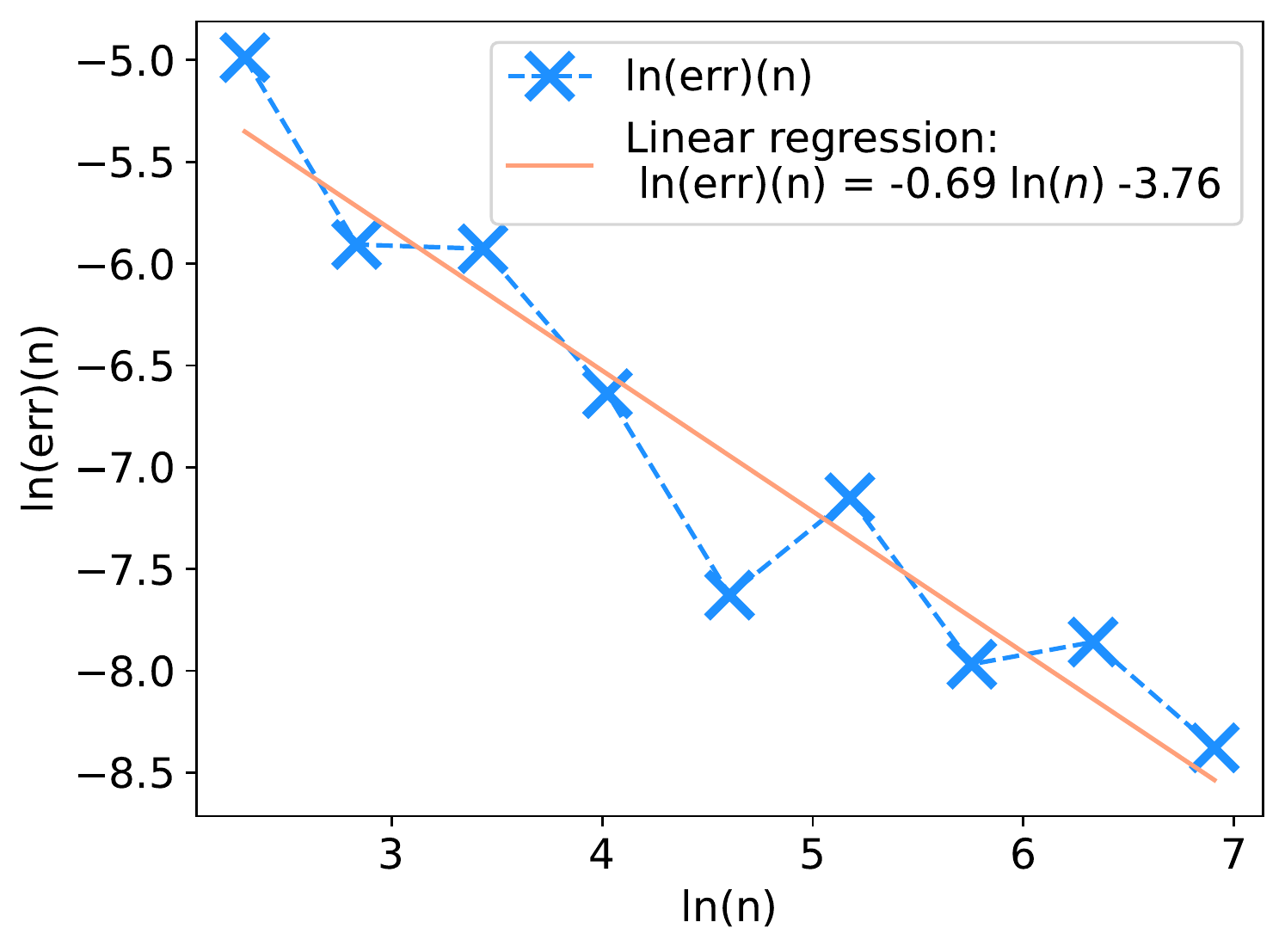} &
    \includegraphics[width = 0.45\textwidth]{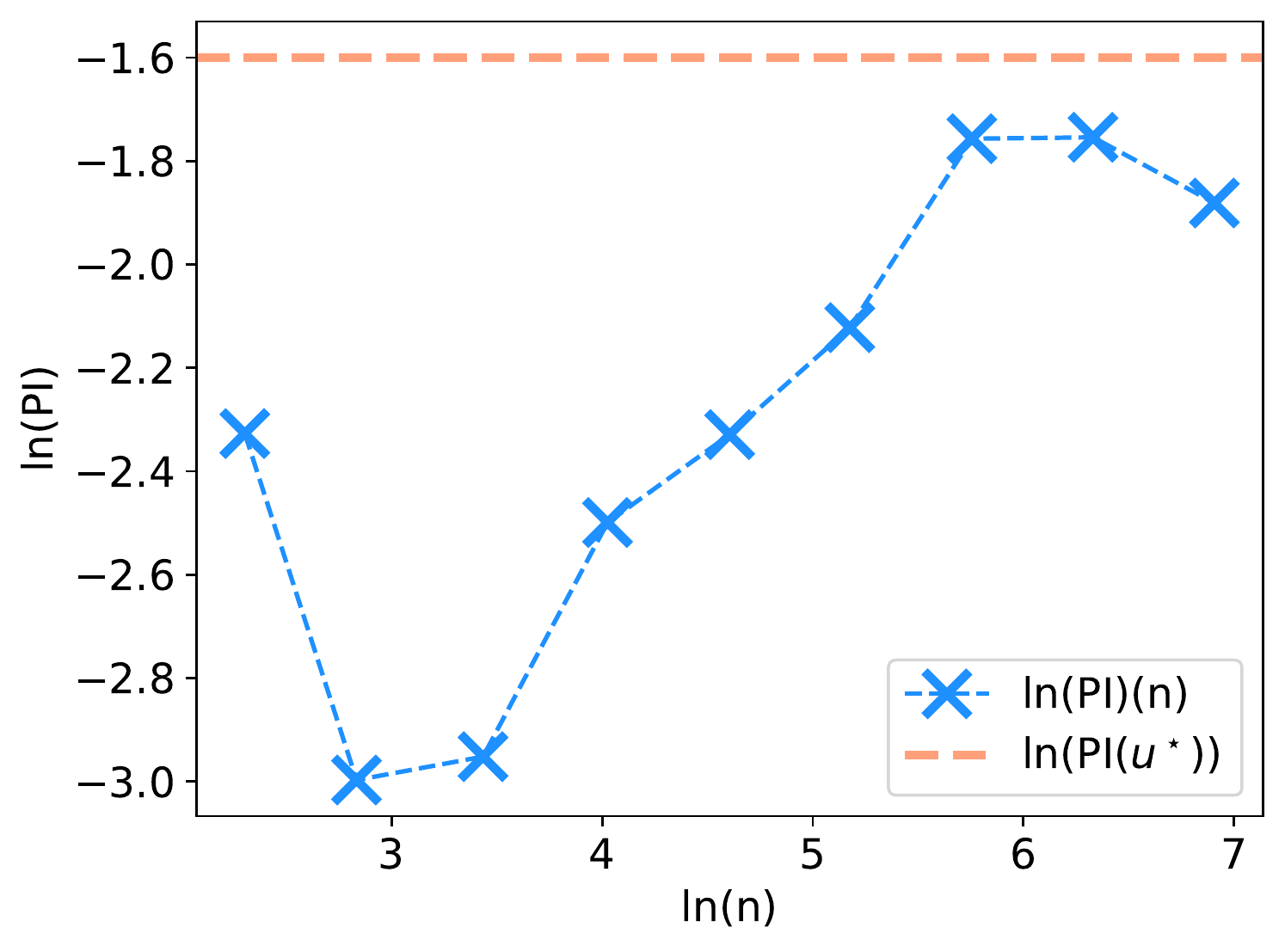} \\
    \end{tabular}
    \caption{Distance $\mathrm{err}(n)$ to  $u^\star$ (left) and physics inconsistency $\mathrm{PI}$ (right) of the regularized PINN estimator with respect to the number $n$ of observations in $\log$-$\log$ scale.}
    \label{fig:linReg}
\end{figure}

\begin{figure}
    \centering
    \includegraphics[width = 0.32\textwidth]{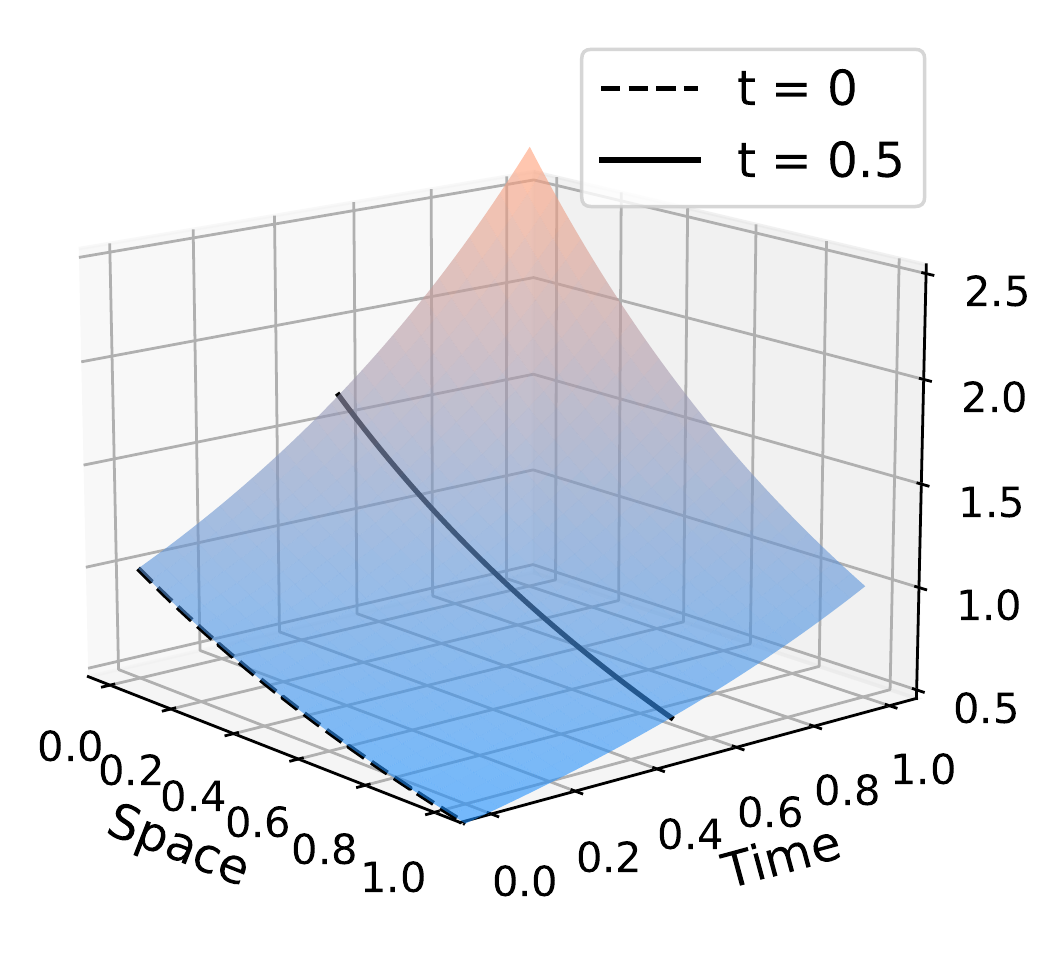}
    \includegraphics[width = 0.32\textwidth]{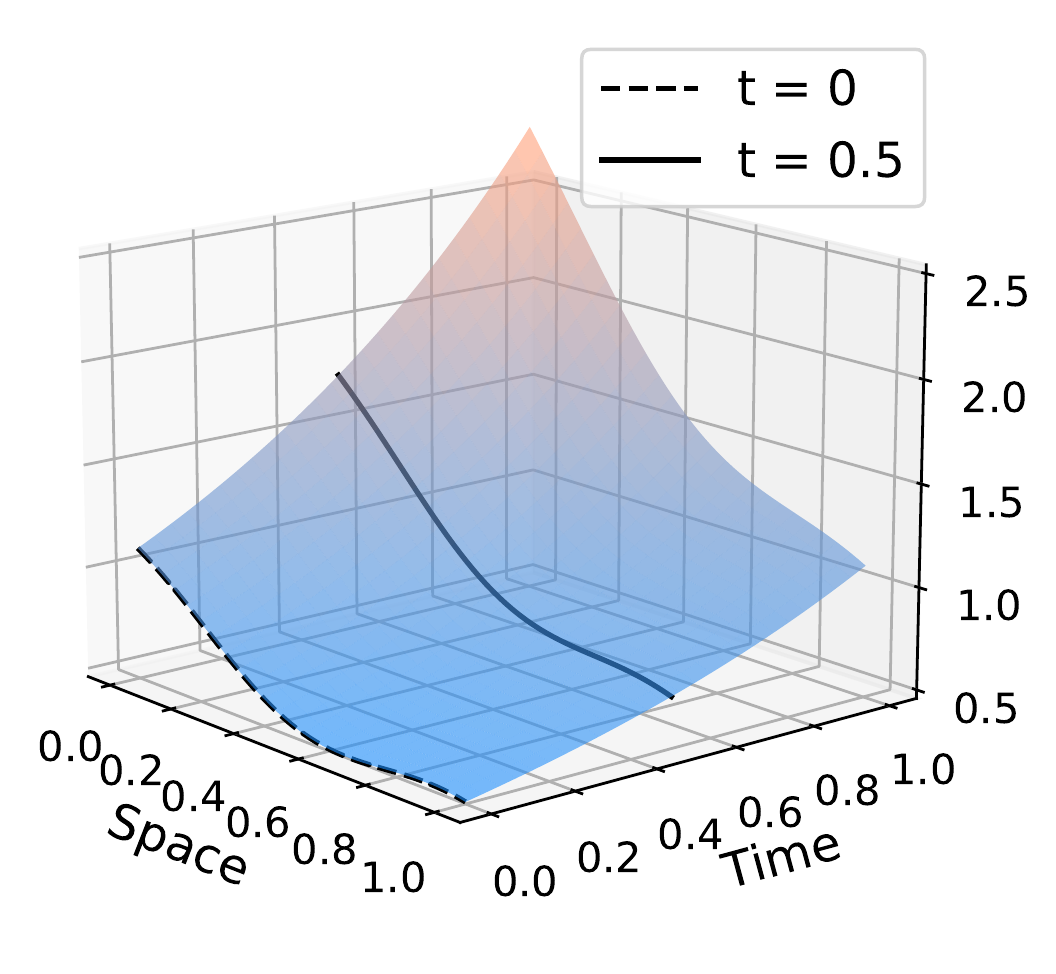}
    \includegraphics[width = 0.32\textwidth]{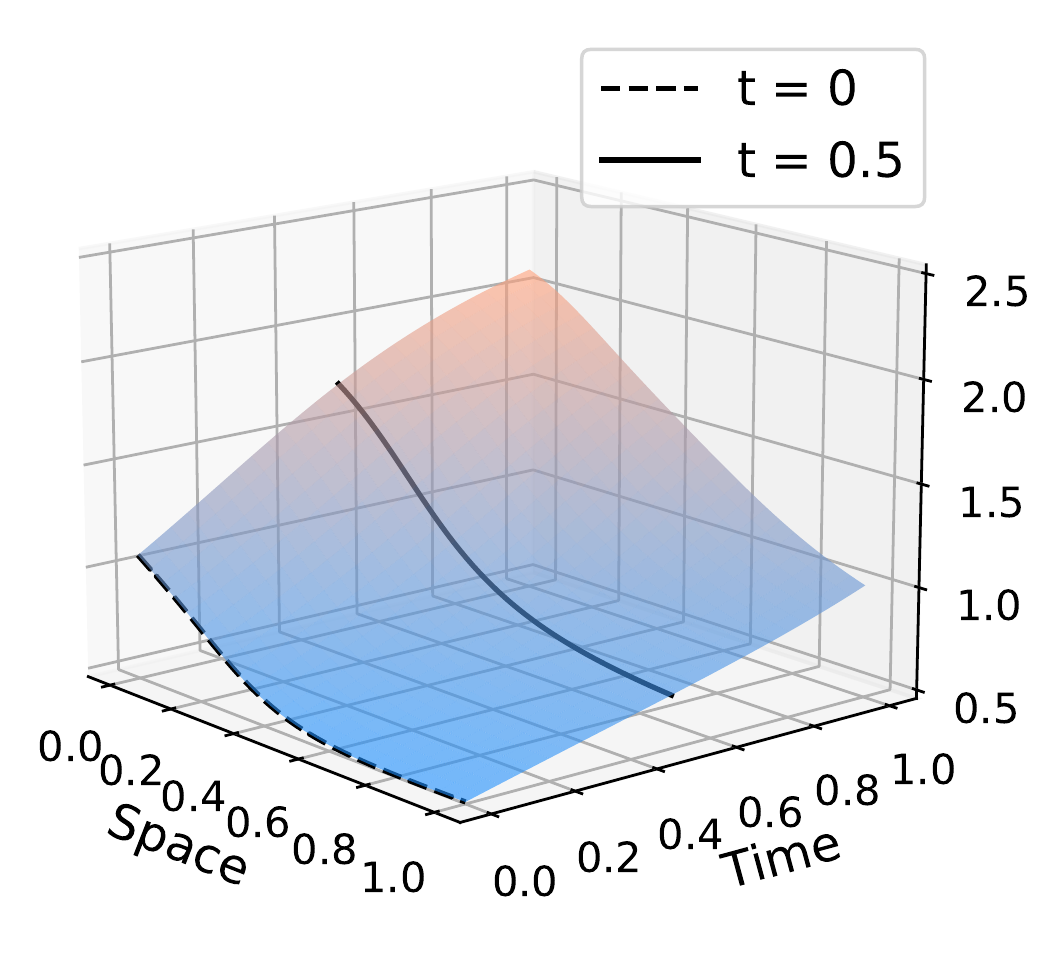}
    \caption{Functions $u_{\textrm{model}}$ (left), $u^\star$ (middle), and regularized PINN estimator with $n=10^3$ (right).}
    \label{fig:hybrid_modelling_nn}
\end{figure}

The architecture of the neural networks is set to $H= 2$ hidden layers with width $D = 100$, so that the total number of parameters is $10\,600 \gg n_{\max}$.
We fix $n_e, n_r = 10^4 \gg n_{\max}$ and $\lambda_{(\mathrm{ridge})} = \min(n_e, n_r)^{-1/2}$.  
Figure \ref{fig:monitoring} shows  the evolution of the regularized risk $R_{n, n_e,n_r}^{(\mathrm{reg)}}(u^{(n)}_{\hat \theta^{(\mathrm{reg})}(p, n_r, n_e, D)})$ in blue, with respect to the number $p$ of epochs in the gradient descent (for $n=10$). 
For a fixed number $n$ of observations,  the number $p_{\max}$ of epochs to stop training is determined by monitoring the evolution of the risk $R_{n, n_e,n_r}^{(\mathrm{reg})}(u^{(n)}_{\hat \theta^{(\mathrm{reg})}(p_{\max}, n_r, n_e, D)})$ (blue curve) and the overfitting gap $\mathrm{OG}_{n, n_e, n_r} = |R_{n, n_e, n_r}^{(\mathrm{reg})}- \mathscr{R}_n^{(\mathrm{reg})}|$ (orange curve). Both are required to be stable around a minimal value, so that the minimum of the risk is approximately reached, i.e., we require $R_{n, n_e,n_r}^{(\mathrm{reg})}(u^{(n)}_{\hat \theta^{(\mathrm{reg})}(p_{\max}, n_r, n_e, D)}) \simeq \inf_{u \in \mathrm{NN}_H(D)} R_{n, n_e,n_r}^{(\mathrm{reg})}(u)$ and $\mathscr{R}_n^{(\mathrm{reg})}(u^{(n)}_{\hat \theta^{(\mathrm{reg})}(p_{\max}, n_r, n_e, D)}) \simeq \inf_{u \in \mathrm{NN}_H(D)} \mathscr{R}_n^{(\mathrm{reg})}(u)$.
In this overparameterized regime ($D$ is large), one can consider that $\mathscr{R}_n^{(\mathrm{reg})}(u^{(n)}_{\hat \theta^{(\mathrm{reg})}(p_{\max}, n_r, n_e, D)}) \simeq \inf_{u \in C^\infty(\bar{\Omega}, \mathbb{R}^{d_2})} \mathscr{R}_n^{(\mathrm{reg})}(u)$ (Theorem \ref{thm:approximation}).
Keeping $n_e$, $n_r$, and $\lambda_{\text{ridge}}$ fixed, the proximity between the PINN  and $u^\star$ is measured by
\[\mathrm{err}(n)  = 2\int_0^{0.5}\int_0^1 \|u^{(n)}_{\hat \theta^{(\mathrm{reg})}(p_{\max}, n_r, n_e, D)}(x, t)-u^\star(x, t)\|_2^2 dxdt.\]
According to Theorem \ref{cor:sPINNsConsistency}, there exists some constant $\Lambda > 0$ such that, approximately, 
\[\ln\big(\mathbb{E}(\mathrm{err}(n))\big) \lesssim \ln(\Lambda) - \ln(n)/2.\]
This bound is validated numerically in Figure \ref{fig:linReg}, attesting a linear rate in log-log scale between $\mathrm{err}(n)$ and $n$ of $-0.69 \leqslant -0.5$. 
Furthermore, the second statement of Theorem \ref{cor:sPINNsConsistency} suggests that $\ln \mathrm{PI}(n) = \ln \mathrm{PI}(u^{(n)}_{\hat \theta^{(\mathrm{reg})}(p_{\max}, n_r, n_e, D)}) \leqslant \ln \mathrm{PI}(u^\star) = -1.6$, which is also verified in Figure \ref{fig:linReg}.
Interestingly, the regularized PINN estimator quickly becomes more accurate than the initial model, since $\mathrm{err}(n)$ is less than $ \int_\Omega\| u_{\mathrm{model}}-u^\star\|_2^2d\mu_\bX \simeq \exp(-5.3)$ as soon as $\ln(n) > 2.8$, i.e., $n\geqslant 17$.

The obtained regularized PINN estimator for $n=10^3$ is shown in Figure \ref{fig:hybrid_modelling_nn} (right). This estimator looks globally similar to the model $u_{\mathrm{model}}$ (Figure \ref{fig:hybrid_modelling_nn}, left) while managing to reconstruct the variation typical of the cosine perturbation of $u^\star$ (Figure \ref{fig:hybrid_modelling_nn}, middle) at $t = 0$. 
Of course, for $t\geqslant 0.5$, the estimator cannot approximate $u^\star$  
with an infinite precision, since the measurements $\bX_i$ are only sampled for $t < 0.5$. 
However, the regularized PINN estimator succeeds to follow the advection equation dynamics, as it does not vary much along the lines $x-t=\mathrm{cst}$--- despite some flattening effect of the Sobolev regularization for $t \geqslant 0.5$. 

\section{Conclusion}
We have shown that unregularized PINNs can overfit. To remedy this problem, we have proposed to add a ridge penalty to the empirical risk. This regularization ensures the consistency of the PINNs for both linear and nonlinear PDE systems. However, to enforce strong convergence to the target function, another layer of regularization is needed. For linear PDEs, we have proved that the addition of a Sobolev-type penalty is sufficient to ensure the strong convergence of the PINNs. Regarding future research, the next step would be to derive tighter bounds to better quantify the impact of the physical penalty on the convergence speed.

\setcounter{section}{0}
\renewcommand\thesection{\thechapter.\Alph{section}}
\section{Notations}
\label{sec:notation}
\paragraph{Composition of functions} Given two functions $u,v : \mathbb{R} \rightarrow \mathbb{R}$, we denote by $u\circ v$ the function $u\circ v(x) = u(v(x))$. For all $k \in \mathbb{N}$, the function $u^{\circ k}$ is defined by induction as $u^{\circ 0}(x) = x$ and $u^{\circ (k+1)} = u^{\circ k}\circ u = u \circ u^{\circ k}$. The composition symbol is placed before the derivative, so that the $k$th derivative of $u^{\circ H}$ is denoted by $(u^{\circ H})^{(k)}$.
\paragraph{Norms} The $p$-norm $\|x\|_p$ of  $x = (x_1,\hdots, x_{d}) \in \mathbb R^d$ is defined by $ \|x\|_p = (\frac{1}{d}\sum_{i=1}^{d} |x_i|^p)^{1/p}$. In addition, $\|x\|_\infty = \max_{1\leqslant i \leqslant d} |x_i|$.
For a function $u : \Omega \rightarrow \mathbb{R}^{d}$, we let $ \|u\|_{L^p(\Omega)} = (\frac{1}{|\Omega|}\int_\Omega \|u\|_p^p)^{1/p}$. Similarly, $\|u\|_{\infty, \Omega} = \sup_{x \in \Omega} \|u(x)\|_\infty$. For simplicity, we sometimes write $\|u\|_{\infty}$ instead of $\|u\|_{\infty, \Omega}$.

\paragraph{Multi-indices and partial derivatives} For a multi-index $\alpha = (\alpha_1, \hdots, \alpha_{d_1}) \in \mathbb{N}^{d}$ and a differentiable function $u:\mathbb R^{d_1}\to \mathbb R^{d_2}$, the $\alpha$ partial derivative of $u$ is defined by $\partial^\alpha u = (\partial_{1})^{\alpha_1}\hdots (\partial_{d_1})^{\alpha_{d_1}} u$. The set of multi-indices of sum less than $k$ is defined by \[\{|\alpha|\leqslant k\} = \{(\alpha_1, \hdots, \alpha_{d_1}) \in \mathbb{N}^{d}, \alpha_1 + \cdots +\alpha_{d_1} \leqslant k\}.\] If $\alpha = 0$, $\partial^\alpha u = u$. Given two multi-indices $\alpha$ and $\beta$, we write $\alpha \leqslant \beta$ when $\alpha_i \leqslant \beta_i$ for all $1\leqslant i \leqslant d_1$. 
The set of multi-indices less than $\alpha$ is denoted by $\{\beta \leqslant \alpha\}$. For a multi-index $\alpha$ such that $|\alpha|\leqslant k$, both sets $\{|\beta|\leqslant k\}$ and $\{\beta \leqslant \alpha\}$ are contained in $\{0, \hdots, k\}^{d_1}$ and are therefore finite.

\paragraph{Hölder norm} For $K \in \mathbb{N}$, the Hölder norm of order $K$ of a function $u \in C^K(\Omega, \mathbb{R}^{d})$, is defined by $\|u\|_{C^K(\Omega)} = \max_{|\alpha|\leqslant K} \|\partial^\alpha u\|_{\infty, \Omega}$. 
This norm allows to bound a function as well as its derivatives. 
The space $C^K(\Omega, \mathbb{R}^{d})$ endowed with the Hölder norm $\|\cdot\|_{C^K(\Omega)}$ is a Banach space. 
$C^\infty(\bar{\Omega}, \mathbb{R}^{d_2})$ is the space of continuous functions $u:\bar{\Omega} \to \mathbb{R}^{d_2}$ satisfying $u|_\Omega \in C^\infty(\Omega, \mathbb{R}^{d_2})$ and, for all $K\in \mathbb{N}$, $\|u\|_{C^K(\Omega)} < \infty$.

\paragraph{Lipschitz function} Given a normed space $(V, \|\cdot\|)$, the Lipschitz norm of a function $u : V \rightarrow \mathbb{R}^{d_1}$ is defined by 
$\|u\|_{\text{Lip}} = \sup_{x,y \in V}\|u(x)-u(y)\|_2/\|x-y\|$. A function $u$ is Lipschitz if $\|u\|_{\mathrm{Lip}}<\infty$. For all $u \in C^1(V, \mathbb{R})$, $\|u\|_{\text{Lip}} \leqslant \|u\|_{C^1(V)}$.

\paragraph{Lipschitz surface  and domain} A surface $\Gamma \subseteq \mathbb{R}^{d_1}$ is said to be Lipschitz if locally, in a neighborhood $U(x)$ of any point $x \in \Gamma$, an appropriate rotation $r_x$ of the coordinate system transforms $\Gamma$ into the graph of a Lipschitz function $\phi_{x}$, i.e., 
\[r_x(\Gamma \cap U(x)) = \{(x_1, \hdots, x_{d - 1}, \phi_x(x_1, \hdots, x_{d - 1})), \forall (x_1, \hdots, x_d)\in r_x(\Gamma \cap U_x)\}.\]
A domain $\Omega \subseteq \mathbb{R}^{d_1}$ is said to be Lipschitz if its has Lipschitz boundary and lies on one side of it, i.e., $\phi_x < 0$ or $\phi_x > 0$ on all intersections $\Omega \cap U_x$. All manifolds with $C^1$ boundary and all convex domains are Lipschitz domains \citep[e.g.,][]{agranovich2015lispchitz}.

\paragraph{Sobolev spaces} Let $\Omega \subseteq \mathbb{R}^{d_1}$ be an open set. A function $v \in L^2(\Omega, \mathbb{R}^{d_2})$ is said to be the $\alpha$th weak derivative of  $u \in L^2(\Omega, \mathbb{R}^{d_2})$ if, for any $\phi \in C^\infty(\bar{\Omega}, \mathbb{R}^{d_2})$ with compact support in $\Omega$, one has
$\int_\Omega \langle v, \phi\rangle = (-1)^{|\alpha|} \int_\Omega \langle u, \partial^\alpha \phi\rangle$. This is denoted by $v = \partial^\alpha u$. For $m \in\mathbb{N}$, the Sobolev space $H^m(\Omega, \mathbb{R}^{d_2})$ is the space of all functions $u \in L^2(\Omega, \mathbb{R}^{d_2})$ such that $\partial^\alpha u$ exists for all $|\alpha|\leqslant m$. This space is naturally endowed with the norm $\|u\|_{H^m(\Omega)} = (\sum_{|\alpha|\leqslant m} |\Omega|^{-1}\|\partial^\alpha u\|_{L^2(\Omega)} ^2)^{1/2}$. For example, the function $u : \, ]-1, 1[ \to \mathbb{R}$ such that $u(x) = |x|$ is not derivable on $]-1, 1[$, but it admits $u'(x) = \mathbf{1}_{x > 0} -  \mathbf{1}_{x < 0}$ as weak derivative. Since $u' \in L^2([-1, 1], \mathbb{R})$, $u$ belongs to the Sobolev space $H^1(]-1, 1[, \mathbb{R})$. However, $u'$ has no weak derivative, and so $u \notin H^2(]-1, 1[, \mathbb{R})$. Of course, if a function $u$ belongs to the Hölder space $C^K(\bar \Omega, \mathbb{R}^{d_2})$, then it belongs to the Sobolev space $H^K(\Omega, \mathbb{R}^{d_2})$, and its weak derivatives are the usual derivatives. For more on Sobolev spaces, we refer the reader to \citet[Chapter 5]{evans2010partial}.

\section{Some reminders of functional analysis on Lipschitz domains}
\label{sec:rem_functional_ana}

\noindent\textbf{Extension theorems} Let $\Omega \subseteq \mathbb{R}^{d_1}$ be an open set and let $K \in \mathbb{N}$ be an order of differentiation. It is not straightforward to extend a function $u \in H^K(\Omega, \mathbb{R}^{d_2})$ to a function $\tilde{u} \in H^K(\mathbb{R}^{d_1}, \mathbb{R}^{d_2})$ such that 
\[\tilde{u}|_\Omega = u|_\Omega \quad \text{and} \quad \|\tilde{u}\|_{H^K(\mathbb{R}^{d_1})} \leqslant C_\Omega \|u\|_{H^K(\Omega)},\]
for some constant $C_\Omega$ independent of $u$. This result is known as the extension theorem in \citet[][Chapter 5.4]{evans2010partial} when $\Omega$ is a manifold with $C^1$ boundary. 
However, the simplest domains in PDEs take the form $]0,L[^3\times ]0,T[$, the boundary of which is not $C^1$. Fortunately, \citet[][Theorem 5 Chapter VI.3.3]{stein1970lipschitz} provides an extension theorem for bounded Lipschitz domains.
We refer the reader to \citet{shvartzman2010sobolev} for a survey on extension theorems.

\paragraph{Example of a non-extendable domain} Let the domain $\Omega = ]-1, 1[^2 \backslash (\{0\}\times [0,1[)$ be the square $]-1, 1[^2$ from which the segment $\{0\}\times [0,1[$ has been removed. Then the function 
\[u(x, y) = \left\{\begin{array}{cl}
    0 & \quad  \text{if } x < 0  \text{ or if } y \leqslant 0\\
     \exp(-\frac{1}{y})& \quad  \text {if } x,y > 0,
\end{array}\right.\] belongs to $C^\infty(\Omega, \mathbb{R})$ but cannot be extended to $\mathbb{R}^2$, since it cannot be continuously extended to the segment $\{0\}\times [0,1[$. Notice that $\Omega$ is not a Lipschitz domain because it lies on both sides of the segment  $\{0\}\times [0,1[$, which belongs to its boundary $\partial \Omega$. 

\begin{thm}[Sobolev inequalities]
    \label{thm:sobIneq}
    Let $\Omega \subseteq \mathbb{R}^{d_1}$ be a bounded Lipschitz domain and let $m \in \mathbb{N}$. If $m  \geqslant d_1/2$, then there exists an operator $\tilde \Pi : H^{m}(\Omega, \mathbb{R}^{d_2}) \to C^0(\Omega, \mathbb{R}^{d_2})$ such that, for any $u \in H^{m}(\Omega, \mathbb{R}^{d_2})$, $\tilde \Pi(u) = u$ almost everywhere. Moreover, there exists a constant $C_\Omega >0$, depending only on $\Omega$, such that, $\|\tilde \Pi(u)\|_{\infty, \Omega} \leqslant C_\Omega \|u\|_{H^{m}(\Omega)}.$
\end{thm}
\begin{proof}
    Since $\Omega$ is a bounded Lipschitz domain, there exists a radius $r > 0$ such that $\Omega \subseteq B(0, r)$. According to the extension theorem \citep[][Theorem 5, Chapter VI.3.3]{stein1970lipschitz}, there exists a constant $C_{\Omega}>0$, depending only on $\Omega$, such that any $u\in H^{m}(\Omega, \mathbb{R}^{d_2})$ can be extended to $\tilde u\in H^{m}(B(0,r), \mathbb{R}^{d_2})$, with
    $\|\tilde u \|_{H^{m}(B(0,r))} \leqslant C_{\Omega} \| u \|_{H^{m}(\Omega)}$.
    Since $m  \geqslant d_1/2$, the Sobolev inequalities \citep[e.g.,][Chapter 5.6, Theorem 6]{evans2010partial} state that there exists a constant $\tilde C_{\Omega} >0$, depending only on $\Omega$, and a linear embedding $\Pi : H^{m}(B(0,r), \mathbb{R}^{d_2}) \to C^0(B(0,r), \mathbb{R}^{d_2})$ such that  $\|\Pi (\tilde u)\|_{\infty} \leqslant \tilde C_{\Omega} \|\tilde u \|_{H^{m}(B(0,r))}$ and  $\Pi (\tilde u) = \tilde u$ in $H^{m}(B(0,r), \mathbb{R}^{d_2})$.
    Therefore, $\tilde \Pi (u) = \Pi (\tilde u)|_{\Omega}$ and $\|\tilde \Pi (u)\|_{\infty, \Omega} \leqslant  C_{\Omega} \tilde C_{\Omega} \| u \|_{H^{m}(\Omega)}$. 
\end{proof}

\begin{defi}[Weak convergence in $L^2(\Omega)$]
    A sequence $(u_p)_{p\in \mathbb{N}} \in L^2(\Omega)^{\mathbb{N}}$ weakly converges to $u_\infty \in L^2(\Omega)$ if, for any $\phi \in L^2(\Omega)$, $ \lim_{p\to\infty }\int_\Omega \phi u_p = \int_\Omega \phi u_\infty$. This convergence is denoted by $u_p \rightharpoonup u_\infty$.
\end{defi}

The Cauchy-Schwarz inequality shows that the convergence with respect to the $L^2(\Omega)$ norm implies the weak convergence. However, the converse is not true. For example, the sequence of functions $u_p(x) = \cos(px)$ weakly converges to $0$ in $L^2([-\pi, \pi])$, whereas $\|u_p\|_{L^2([-\pi,\pi])} = 1/2$.

\begin{defi}[Weak convergence in $H^m(\Omega)$]
    A sequence $(u_p)_{p\in \mathbb{N}} \in H^m(\Omega)^{\mathbb{N}}$ weakly converges to $u_\infty \in H^m(\Omega)$  in $H^m(\Omega)$ if, for all $|\alpha|\leqslant m$, $\partial^\alpha u_p \rightharpoonup \partial^\alpha u_\infty$.
\end{defi}

\begin{thm}[Rellich-Kondrachov]
    \label{thm:rellichK}
    Let $\Omega \subseteq \mathbb{R}^{d_1}$ be a bounded Lipschitz domain and let $m \in \mathbb{N}$. Let $(u_p)_{p\in \mathbb{N}}\in H^{m+1}(\Omega, \mathbb{R}^{d_2})$ be a sequence such that $(\|u_p\|_{H^{m+1}(\Omega)})_{p\in \mathbb{N}}$ is bounded.
    There exists a function $u_\infty \in H^{m+1}(\Omega, \mathbb{R}^{d_2})$ and a subsequence of  $(u_p)_{p\in \mathbb{N}}$ that converges to $u_\infty$ both weakly in $H^{m+1}(\Omega, \mathbb{R}^{d_2})$ and with respect to the $H^m(\Omega)$ norm.
\end{thm}
\begin{proof}
    Let $r>0$ be such that $\Omega \subseteq B(0,r)$.
    According to the extension theorem of \citet[][Theorem 5, Chapter VI.3.3]{stein1970lipschitz}, there exists a constant $C_r >0$ such that each $u_p$ can be extended to $\tilde u_p \in H^{m+1}(B(0,r),\mathbb{R}^{d_2})$, with $\|\tilde u_p\|_{H^{m+1}(B(0,r))} \leqslant C_r \|u_p\|_{H^{m+1}(\Omega)}$.
    Observing that, for all $|\alpha|\leqslant m$, $\partial^\alpha \tilde u_p$ belongs to $H^1(B(0,r),\mathbb{R}^{d_2})$, the Rellich-Kondrachov compactness theorem \citep[Theorem 1, Chapter 5.7]{evans2010partial} ensures that there exists a subsequence of $(\tilde u_p)_{p \in \mathbb{N}}$ that converges to an extension of $u_\infty$ with respect to the $H^m(B(0,r))$ norm. 
    Since the subsequence is also bounded, upon passing to another subsequence, it also weakly converges in $H^{m+1}(B(0,r), \mathbb{R}^{d_2})$ to $u_\infty \in H^{m+1}(B(0,r), \mathbb{R}^{d_2})$ \citep[e.g.,][Chapter D.4]{evans2010partial}.
    Therefore, by considering the restrictions of all the previous functions to $\Omega$, we deduce that there exists a subsequence of $(u_p)_{p\in\mathbb{N}}$ that converges to $u_\infty$ both weakly in $H^{m+1}(\Omega)$ and with respect to the $H^m(\Omega)$ norm.
\end{proof}
\section{Some useful lemmas}
\label{app:lemmas}
The $n$th Bell number $B_n$ \citep{hardy2006combinatorics} corresponds to the number of partitions of the set $\{1,\hdots, n\}$. Bell numbers satisfy the relationship $B_0 = 1$ and 
\begin{equation}
    B_{n+1} = \sum_{k=0}^n \begin{pmatrix}
        n\\k
    \end{pmatrix} B_k \label{eq:bell}.
\end{equation}
For $K \geqslant 1$ and $u\in C^K(\mathbb{R}^{d_1}, \mathbb{R}^{d_2})$, the $K$th derivative of $u$ is denoted by $u^{(K)}$.

\begin{lem}[Bounding the partial derivatives of a composition of functions]
    \label{lem:boundPartialDer}
    Let $d_1, d_2 \geqslant 1$, $K \geqslant 0$,  $f \in C^{K}(\mathbb{R}^{d_1}, \mathbb{R})$, and $g \in C^{K}(\mathbb{R}, \mathbb{R}^{d_2})$. Then
    \[\|g\circ f\|_{C^K(\mathbb{R}^{d_1})} \leqslant B_{K} \|g\|_{C^K(\mathbb{R})} (1+\|f\|_{C^K(\mathbb{R}^{d_1})})^{K}.\]
\end{lem}
\begin{proof}
    Let $K_1 \leqslant K$ and let $\Pi(K_1)$ be the set of all partitions of $\{1, \hdots, K_1\}$. According to \citet[Proposition 1]{hardy2006combinatorics}, one has, for all $h\in C^{K_1}(\mathbb{R}^{K_1+d_1}, \mathbb{R})$,
    \[\partial^{K_1}_{1,2,3,\hdots, K_1} (g\circ h) = \sum_{P \in \Pi(K_1)} g^{(|P|)}\circ h \times \prod_{S \in P} \Big[\big(\prod_{j \in S}\partial_j\big) h\Big].\]
    Let  $ \alpha = (\alpha_1, \hdots, \alpha_{d_1})$ be a multi-index such that $|\alpha|=K_1$. Setting $\alpha_0 = 0$, $y_j = x_{K_1+j} + (x_{\alpha_1 + \cdots + \alpha_{j-1}}+\cdots+x_{\alpha_1 + \cdots + \alpha_{j}-1})$, and letting $h(x_1, \hdots, x_{K_1+d_1}) = f(y_1, \hdots, y_{d_1})$, 
    we are led to
    \begin{equation}
        \partial^\alpha (g\circ f) = \sum_{P \in \Pi(K_1)} g^{(|P|)}\circ f \times \prod_{S \in P} \partial^{\alpha(S)}f, \label{eq:fdbFormula}
    \end{equation} 
    where $\alpha(S) = ( |\{b \in S,\quad \alpha_1+\cdots+\alpha_{\ell-1}\leqslant b  \leqslant \alpha_1+\cdots+\alpha_\ell \}|)_{1\leqslant \ell \leqslant d_1}$. Moreover, by definition of the Bell number, $|\Pi(K_1)| = B_{K_1}$, and, by definition of a partition, $|P|\leqslant K_1$. So,
    \begin{align*}
        \|\partial^\alpha (g\circ f)\|_\infty & \leqslant B_{K_1}  \|g\|_{C^{K_1}(\mathbb{R}^{d_1})} \max_{i_1+2i_2+\cdots+ K_1 i_{K_1}=K_1}\prod_{j=1}^{K_1} \|f\|_{C^j(\mathbb{R}^{d_1})}^{i_j} \\
        &\leqslant B_{K_1} \|g\|_{C^{K_1}(\mathbb{R}^{d_1})} (1+\|f\|_{C^{K_1}(\mathbb{R}^{d_1})})^{K_1}.\nonumber
    \end{align*} 
    Since this inequality is true for all $K_1\leqslant K$ and for all $|\alpha| = K_1$, the lemma is proved.
\end{proof}

\begin{lem}[Bounding the partial derivatives of a changing of coordinates $f$]
\label{lem:boundPartialDer2}
    Let $d_1, d_2 \geqslant 1$, $K \geqslant 0$,  $f \in C^{K}(\mathbb{R}, \mathbb{R})$, and $g \in C^{K}(\mathbb{R}^{d_1}, \mathbb{R}^{d_2})$. Let $v\in C^K(\mathbb{R}^{d_1}, \mathbb{R}^{d_1})$ be defined by $v(\bx) = (f(x_1),\hdots, f(x_{d_1}))$.
    Then
    \[\|g\circ v\|_{C^K(\mathbb{R}^{d_1})} \leqslant B_{K}\times \|g\|_{C^K(\mathbb{R}^{d_1})}\times (1+\|f\|_{C^K(\mathbb{R})})^{K}.\]
\end{lem}
\begin{proof} Let $\alpha = (\alpha_1, \hdots, \alpha_{d_1})$ be a multi-index such that $|\alpha|=K$. For $\bx = (x_1, \hdots, x_{d_1})$ and a fixed $i \in \{1, \hdots, d_1\}$, we let $h(t)= g(f(x_1), \hdots, f(x_{i-1}), t, f(x_{i+1}), \hdots, f(x_{d_1}))$. Clearly, $(h\circ f)^{(\alpha_i)}(x_i) = (\partial_i)^{\alpha_i} (g\circ v)(\bx)$. Thus, according to Lemma \ref{lem:boundPartialDer},
\[(h\circ f)^{(\alpha_i)} = \sum_{P_i \in \Pi(\alpha_i)} h^{(|P_i|)}\circ f \times \prod_{S_i \in P_i} f^{(|S_i|)}.\]
Therefore,
\[(\partial_i)^{\alpha_i}(g\circ v)(\bx) = \sum_{P_i \in \Pi(\alpha_i)} (\partial_i)^{|P_i|}g\circ v(\bx) \prod_{S_i \in P_i} f^{(|S_i|)}(x_i).\]
Letting $i=1$ and observing that $\partial_j f^{(|S_1|)}(x_1) =0$ for $j \neq 1$, we see that
    \[\partial^{\alpha}(g\circ v)(\bx) = \sum_{P_1 \in \Pi(\alpha_1)} \Big[\prod_{S_1 \in P_1} f^{(|S_1|)}(x_1)\Big] \times (\partial_2)^{\alpha_2}\hdots (\partial_{d_1})^{\alpha_{d_1}}[(\partial_1)^{|P_1|}g\circ v](\bx) .\]
Repeating the same procedure for $(\partial_1)^{|P_1|}g\circ v, \hdots, (\partial_1)^{|P_1|}\hdots (\partial_{d_1})^{|P_{d_1}|} g\circ v$, we obtain
\begin{align*}
    \partial^{\alpha}(g\circ v)(\bx) &= \sum_{P_1 \in \Pi(\alpha_1)} \Big[\prod_{S_1 \in P_1} f^{(|S_1|)}(x_1)]\Big] \times \cdots \\
     & \quad\cdots \times \sum_{P_{d_1} \in \Pi(\alpha_{d_1})} \Big[\prod_{S_{d_1} \in P_{d_1}} f^{(|S_{d_1}|)}(x_{d_1})]\Big]\times  (\partial_1)^{|P_1|}\hdots (\partial_{d_1})^{|P_{d_1}|}g\circ v(\bx) .
\end{align*}
Since $\sum_{S_i \in P_i} |S_i| = \alpha_i$ and $ \sum_{i=1}^{d_1} \alpha_i = K$, we conclude that 
\[\|\partial^{\alpha}(g\circ v)\|_\infty \leqslant B_{\alpha_1}\times\cdots\times B_{\alpha_{d_1}}\times \|\partial^{\alpha}g\|_\infty  (1+\|f\|_{C^K(\mathbb{R})})^K. \] 
Using the injective map $\mathcal{M} : \Pi(\alpha_1)\times\cdots\times \Pi(\alpha_{d_1}) \to \Pi(K)$ such that $\mathcal{M}(P_1,\hdots,P_{d_1}) = \cup_{i=1}^{d_1}P_i$, we have $B_{\alpha_1}\times\cdots\times B_{\alpha_{d_1}} \leqslant B_K$. This concludes the proof.
\end{proof}

\begin{lem}[Bounding hyperbolic tangent and its derivatives]
\label{lem:derTanh}
    For all $K\in \mathbb{N}$, one has 
 \[\|\tanh^{(K)}\|_{\infty} \leqslant 2^{K-1} (K+2)!\]
\end{lem}
\begin{proof}
The $\tanh$ function is a solution of the equation $y' = 1 - y^2$. An elementary induction shows that there exists a sequence of polynomials $(P_K)_{K \in \mathbb{N}}$ such that $\tanh^{(K)} = P_K(\tanh)$, with $P_0(X) = X$ and $P_{K+1}(X) = (1-X^2)\times P_K'(X)$. Clearly, $P_K$ is a real polynomial of degree $K+1$, of the form
$P_K(X) = a^{(K)}_0 + a^{(K)}_1 X + \cdots + a^{(K)}_{K+1} X^{K+1}$. 
One verifies that $a^{(K+1)}_i = (i+1)a^{(K)}_{i+1} - (i-1)a_{i-1}^{(K)}$, with $a^{(K)}_{-1} = a^{(K)}_{K+2}=0$. The largest coefficient $M(P_K) = \max_{0 \leqslant i \leqslant K+1} |a^{(K)}_i|$ of $P_K$ satisfies $M(P_{K+1}) \leqslant 2(K+1) \times M(P_K)$.  Thus, since $M(P_1) = 1$, we see that $M(P_K) \leqslant 2^{K-1} K!$\ . Recalling that $0 \leqslant \tanh \leqslant 1$, we conclude that
\[\|\tanh^{(K)}\|_{\infty} = \|P_K(\tanh)\|_\infty \leqslant (K+2)  M(P_K)  \leqslant 2^{K-1} (K+2)!\]
\end{proof}

In the sequel, for all $\theta \in \mathbb{R}$, we write $\tanh_\theta(x) = \tanh(\theta x)$. We define the sign function such that $\mathrm{sgn}(x) = \mathbf{1}_{x > 0} - \mathbf{1}_{x < 0}$.
\begin{lem}[Characterizing the limit of hyperbolic tangent in Hölder norm]
    \label{lem:compTanh}
    Let $K \in \mathbb{N}$ and $H \in \mathbb{N}^\star$. Then, for all $\varepsilon > 0$,  
    $\lim_{\theta\to\infty}\|\tanh^{\circ H}_\theta- \mathrm{sgn}\|_{C^K(\mathbb{R}\backslash ]-\varepsilon, \varepsilon[)}=0$.
\end{lem}
\begin{proof}  Fix $\varepsilon > 0$. We prove the stronger statement that, for all $m \in \mathbb{N}$, one has
$$\lim_{\theta \to \infty}\theta^m \|\tanh^{\circ H}_\theta- \mathrm{sgn}\|_{C^K(\mathbb{R}\backslash ]-\varepsilon, \varepsilon[)}= 0.$$ 
We start with the case $H=1$ and then prove the result by induction on $H$. 
Observe first, since $\tanh_\theta^{\circ H} - \mathrm{sgn}$ is an odd function, that
\[\|\tanh^{\circ H}_\theta- \mathrm{sgn}\|_{C^K(\mathbb{R}\backslash ]-\varepsilon, \varepsilon[)} = \|\tanh^{\circ H}_\theta- \mathrm{sgn}\|_{C^K([\varepsilon, \infty[)}.\]
\paragraph{The case $H=1$} Assume, to start with, that $K=0$. For all $x \geqslant \varepsilon$, one has
\begin{align*}
    \theta^m|\tanh_\theta(x)- 1| &= \frac{2\theta^m}{1+\exp(-2\theta x)}\leqslant \frac{2\theta^m}{1+\exp(-2\theta \varepsilon)}.
\end{align*}
Therefore, for all $ m \in \mathbb{N}$,  
\begin{align*}
    \theta^m\|\tanh_\theta-\mathrm{sgn}\|_{\infty, \mathbb{R}\backslash ]-\varepsilon, \varepsilon[} &= \theta^m\|\tanh_\theta-\mathrm{sgn}\|_{\infty, [\varepsilon, \infty[}\leqslant \frac{2\theta^m}{1+\exp(-2\theta \varepsilon)}\xrightarrow{\theta \rightarrow \infty} 0.
\end{align*}  
Next, to prove that the result if true for all $K \geqslant 1$, it is enough to show that, for all $m$, $$\theta^m\|\tanh_\theta^{(K)}\|_{\infty,\mathbb{R}\backslash ]-\varepsilon, \varepsilon[}\xrightarrow{\theta \rightarrow \infty} 0.$$ 
According to the proof of Lemma \autoref{lem:derTanh}, there exists a sequence of polynomials $(P_K)_{K\in \mathbb{N}}$ such that $\tanh^{(K)} = P_K(\tanh)$ and $P_{K+1}(X) = (1-X^2)\times P_K'(X)$. Since $\tanh_\theta(x) = \tanh(\theta x)$, one has
\begin{align*}
    \tanh_\theta^{(K)}(x) & =  \theta^K \tanh^{(K)}(\theta x)\\
    & = \theta^K (1-\tanh^2(\theta x))\times P'_{K-1}(\tanh(\theta x))\\
    & = \theta^K (1-\tanh(\theta x)) (1+\tanh(\theta x))\times P'_{K-1}(\tanh(\theta x)).
\end{align*}
Fix $x \geqslant \varepsilon$. Then, letting $M_K = \|P'_{K-1}\|_{\infty, [-1,1]}$, we are led to
\begin{align*}
    |\tanh_\theta^{(K)}(x)|\leqslant 2 M_K  \theta^K (1-\tanh(\theta x)) &\leqslant 4 M_K \times \frac{\theta^K}{1+\exp(2\theta x)}\\
    &\leqslant 4 M_K \times \frac{\theta^K}{1+\exp(2\theta \varepsilon)}.
\end{align*}
This shows that $\theta^m\|\tanh_\theta^{(K)}\|_{\infty, [\varepsilon, \infty[}\leqslant 4 M_K \times \frac{\theta^{K+m}}{1+\exp(2\theta \varepsilon)}$. One proves with similar arguments that the same result holds for all $x \leqslant - \varepsilon$. Thus, 
\begin{equation*}
    \theta^m\|\tanh_\theta^{(K)}\|_{\infty, \mathbb{R}\backslash  ]-\varepsilon, \varepsilon[}\leqslant 4 M_K \times \frac{\theta^{K+m}}{1+\exp(2\theta \varepsilon)} \xrightarrow{\theta \rightarrow \infty}0,
\end{equation*}
and the lemma is proved for $H=1$.
\paragraph{Induction} 
Assume that that, for all $K$ and all $m$, 
\begin{equation}
\label{Hfixe}
\theta^m \|\tanh^{\circ H}_\theta- \mathrm{sgn}\|_{C^K(\mathbb{R}\backslash ]-\varepsilon, \varepsilon[)}\xrightarrow{\theta \rightarrow \infty} 0.
\end{equation}
Our objective is to prove that, for all $K_2$ and all $m_2$,
\[\theta^{m_2}\|\tanh^{\circ (H+1)}_\theta- \mathrm{sgn}\|_{C^{K_2}(\mathbb{R}\backslash ]-\varepsilon, \varepsilon[)}\xrightarrow{\theta \rightarrow \infty} 0.\]
If $K_2 = 0$, since, for all $(x,y) \in \mathbb{R}^2$,
  $|\tanh_\theta(x) - \tanh_\theta(y)| \leqslant \theta |x-y|\times \|\tanh'\|_\infty 
    \leqslant \theta |x-y|$.
We deduce that \[\theta^{m_2}\|\tanh_\theta^{\circ (H+1)}- \tanh_\theta(\mathrm{sgn})\|_{\infty, \mathbb{R}\backslash ]-\varepsilon, \varepsilon[} \leqslant \theta^{m_2+1} \|\tanh^{\circ H}_\theta- \mathrm{sgn}\|_{\infty, \mathbb{R}\backslash ]-\varepsilon, \varepsilon[}.\]
Therefore, according to \eqref{Hfixe}, we have that 
$\lim_{\theta\to\infty}\theta^{m_2}\|\tanh_\theta^{\circ (H+1)}- \tanh_\theta(\mathrm{sgn})\|_{\infty, \mathbb{R}\backslash ]-\varepsilon, \varepsilon[} = 0$.
Since $\tanh_\theta (\mathrm{sgn}) - \mathrm{sgn}= (\tanh(\theta)-1) \mathbf{1}_{x >0 } - (\tanh(\theta)-1) \mathbf{1}_{x <0 }$, we see that, for all $m_2$, 
$$\lim_{\theta\to\infty}\theta^{m_2}\|\tanh_\theta (\mathrm{sgn}) - \mathrm{sgn}\|_{\infty, \mathbb{R}\backslash ]-\varepsilon, \varepsilon[} =0.$$ 
Using the triangle inequality, we conclude as desired that, for all $m_2$, 
\begin{equation}
\label{eq:characterizing_K2_1}
    \theta^{m_2}\|\tanh_\theta^{\circ (H+1)}- \mathrm{sgn}\|_{\infty, \mathbb{R}\backslash ]-\varepsilon, \varepsilon[} \xrightarrow{\theta \rightarrow \infty} 0.
\end{equation}
Assume now that $K_2\geqslant 1$. Since $\tanh_\theta^{\circ (H+1)} = \tanh^{\circ H}(\tanh)$, the Faà di Bruno formula \citep[e.g.,][Chapter 3.4]{comtet1974advanced} states that 
\begin{align}
    (\tanh_\theta^{\circ (H+1)})^{(K_2)} &= \sum_{m_1 + 2m_2 + \cdots + K_2 m_{K_2} = K_2} \frac{K_2!}{\prod_{i=1}^{K_2} m_i! \times i!^{m_i}}  \nonumber \\
    &\times (\tanh_\theta^{\circ H})^{(m_1+\cdots+m_{K_2})}(\tanh_\theta)\times \prod_{j=1}^{K_2}(\tanh_\theta^{(j)})^{m_j}. \nonumber
\end{align}
Notice that if $|x| \leq \mathrm{arctanh}(1/\sqrt{2})$, $|\tanh(x)|\geqslant \frac{|x|}{2}$ because by calling $f(x) = \tanh(x) - \frac{x}{2}$, $f(0) = 0$ and $f'(x) = (1-\tanh(x)^2) - \frac{1}{2} \geqslant 0$. Therefore, if $|x|\geq \varepsilon$, $|\tanh(\theta x)| \geqslant \min(\frac{1}{\sqrt{2}},\frac{\theta}{2}\varepsilon)\geqslant \varepsilon$ if $\theta \geqslant 2$ and $\varepsilon \geqslant \frac{1}{\sqrt{2}}$. This is why for $\theta \geqslant 2$ and $\varepsilon \leqslant 1$, 
\[ \| (\tanh_\theta^{\circ H})^{(m_1+\cdots+m_{K_2})}(\tanh_\theta)\|_{\infty, \mathbb{R}\backslash ]-\varepsilon, \varepsilon[} \leqslant \|(\tanh_\theta^{\circ H})^{(m_1+\cdots+m_{K_2})}\|_{\infty, \mathbb{R}\backslash ]-\varepsilon, \varepsilon[} .\]  
Therefore, from the triangular inequality on $\|\cdot\|_{\infty, \mathbb{R}\backslash ]-\varepsilon, \varepsilon[}$,
\begin{align*}
    \|(\tanh_\theta^{\circ (H+1)})^{(K_2)}\|_{\infty, \mathbb{R}\backslash ]-\varepsilon, \varepsilon[} &\leqslant \sum_{m_1 + 2m_2 + \cdots + K_2 m_{K_2} = K_2} \frac{K_2!}{\prod_{i=1}^{K_2} m_i! \times i!^{m_i}} \\
    &\! \times  \| (\tanh_\theta^{\circ H})^{(m_1+\cdots+m_{K_2})}\|_{\infty, \mathbb{R}\backslash ]-\varepsilon, \varepsilon[}  \prod_{j=1}^{K_2}\|\tanh_\theta^{(j)}\|_{\infty, \mathbb{R}\backslash ]-\varepsilon, \varepsilon[}^{m_j}.
\end{align*}
 According to the induction hypothesis \eqref{Hfixe}, one has, for all $K \geqslant 1$ and all $m \in \mathbb{N}$,  
 $$\lim_{\theta\to\infty}\theta^m \|(\tanh^{\circ H}_\theta)^{(K)}\|_{\infty, \mathbb{R}\backslash ]-\varepsilon, \varepsilon[}=0.$$ 
 We deduce from the above that for all $K_2 \geqslant 1$ and all $m_2$, 
\begin{equation}
\label{eq:characterizin_K2_geq1}
    \theta^{m_2} \|(\tanh^{\circ (H+1)}_\theta)^{(K_2)}\|_{\infty, \mathbb{R}\backslash ]-\varepsilon, \varepsilon[}\xrightarrow{\theta \rightarrow \infty} 0.
\end{equation}
Combining \eqref{eq:characterizing_K2_1} and \eqref{eq:characterizin_K2_geq1}, it comes that
$\lim_{\theta\to\infty}\theta^{m_2}\|\tanh^{\circ (H+1)}_\theta- \mathrm{sgn}\|_{C^{K_2}(\mathbb{R}\backslash ]-\varepsilon, \varepsilon[)}= 0$.
\end{proof}
\begin{cor}[Bounding hyperbolic tangent compositions and their derivatives]
\label{cor:bounding_tanh}
    Let $K\in \mathbb{N}$ and $H\in \mathbb{N}^\star$. Then, for or all $\theta \in \mathbb R$, 
   $\|(\tanh_\theta^{\circ H})^{(K)}\|_{\infty} < \infty$.
\end{cor}
\begin{proof} 
An induction as the one of Lemma \ref{lem:compTanh} shows that
$\|(\tanh_\theta^{\circ H})^{(K)}\|_{\infty, \mathbb{R}\backslash ]-\varepsilon, \varepsilon[} < \infty$. In addition, since $\tanh_\theta^{\circ H} \in C^\infty(\mathbb{R}, \mathbb{R})$, $\|(\tanh_\theta^{\circ H})^{(K)}\|_{\infty, [-\varepsilon, \varepsilon]} < \infty$.  
\end{proof}

When $d_1=d_2=1$, the observations $(\bX_1,Y_1), \hdots,$ 
$(\bX_n, Y_n) \in\mathbb{R}^{2}$ can be reordered as $(\bX_{(1)}, Y_{(1)}), \hdots,$ $(\bX_{(n)}, Y_{(n)})$ according to increasing values of the $\bX_i$, that is,
$\bX_{(1)} \leqslant \cdots \leqslant \bX_{(n)}$. Moreover, we let 
 $\mathcal{G}(n, n_r) = \{(\bX_i, Y_i), 1\leqslant i \leqslant n\}\cup \{\bX^{(r)}_j, 1\leqslant j \leqslant n_r\}$,
 and denote by $\delta(n, n_r)$ the minimum distance between two distinct points in $\mathcal{G}(n, n_r) $, i.e., 
 \begin{equation}
 \label{def:delta}
     \delta(n,n_r) = \underset{z_1\neq z_2}{\min_{z_1, z_2 \in \mathcal{G}(n, n_r)}} |z_1-z_2|.
 \end{equation}
\begin{lem}[Exact estimation with hyperbolic tangent]
\label{lem:estimationTanh}
    Assume that $d_1=d_2=1$, and let $H \geqslant 1$. Let the neural network $u_\theta \in \mathrm{NN}_H(n-1)$ be defined by
    \[u_\theta(x) =  Y_{(1)} + \sum_{i=1}^{n-1} \frac{Y_{(i+1)}-Y_{(i)}}{2} \bigg[\tanh_\theta^{\circ H}\Big(x-\bX_{(i)}-\frac{\delta(n, n_r)}{2}\Big)+1\bigg].\]
    Then, for all $1 \leqslant i \leqslant n$, \[ \lim_{\theta \to \infty}u_\theta(\bX_i) = Y_i.\]
    Moreover, for all order $K \in \mathbb{N}^\star$ of differentiation and all $1 \leqslant j \leqslant n_r$, \[\lim_{\theta \to \infty} u^{(K)}_\theta(\bX^{(r)}_j) = 0.\]
\end{lem}
\begin{proof} Applying Lemma \ref{lem:compTanh} with $\varepsilon = \nicefrac{\delta(n, n_r)}{4}$ and letting \[G = \mathbb{R}\backslash \cup_{i=1}^n ]\bX_{(i)} + \frac{1}{4}\delta(n, n_r), \bX_{(i)} + \frac{3}{4}\delta(n, n_r)[,\] one has, for all $K$,
$\lim_{\theta\to\infty }\|u_\theta - u_\infty\|_{C^K(G)} = 0$,
where \[u_\infty(x) = Y_{(1)}+\sum_{i=1}^{n-1} \big[Y_{(i+1)}-Y_{(i)}\big]\times \mathbf{1}_{x>\bX_{(i)}+\frac{\delta(n, n_r)}{2}} . \]
Clearly, for all $1\leqslant i \leqslant n$, $u_\infty(\bX_{i}) = Y_i$. Since $u'_\infty(x) = 0$ for all $x \in G$, and since $\bX^{(r)}_j \in G$ for all $1 \leqslant j \leqslant n_r$, we deduce that $u_\infty^{(K)}(\bX^{(r)}_j) = 0$. This concludes the proof.
\end{proof} 
\begin{defi}[Overfitting gap]
    \label{defi:OG}
    For any $n, n_e, n_r \in \mathbb{N}^\star$ and $\lambda_{(\mathrm{ridge})} \geqslant 0$, the overfitting gap operator $\mathrm{OG}_{n, n_e, n_r}$ is defined, for all $u \in C^\infty(\bar{\Omega}, \mathbb{R}^{d_2})$, by \[\mathrm{OG}_{n, n_e, n_r}(u) = |R_{n, n_e, n_r}^{(\mathrm{ridge})}(u)- \mathscr{R}_n(u)|.\]
\end{defi}
\begin{lem}[Monitoring the overfitting gap]
    \label{lem:overfitting_gap}
    Let $\varepsilon > 0$, $\lambda_{(\mathrm{ridge})} \geqslant 0$, $H \geqslant 2$, and $D\in \mathbb{N}^\star$. Let $n , n_e, n_r \in \mathbb{N}^\star$. Let $\hat \theta \in \Theta_{H, D}$ be a parameter such that $(i)$ $R^{(\mathrm{ridge})}_{n, n_e,n_r}(u_{\hat \theta}) \leqslant  \inf_{u \in \mathrm{NN}_H(D)} R_{n, n_e,n_r}^{(\mathrm{ridge})}(u) + \varepsilon$ and $(ii)$ $\mathrm{OG}_{n, n_e, n_r}(u_{\hat \theta})\leqslant \varepsilon$. 
    Then
    \[\mathscr{R}_n(u_{\hat \theta}) \leqslant \inf_{u \in \mathrm{NN}_H(D)} \mathscr{R}_n(u) + 2 \varepsilon + o_{n_e,n_r \to \infty}(1).\]
\end{lem}
\begin{proof}
    On the one hand, since $\mathscr{R}_n \leqslant R_{n, n_e,n_r}^{(\mathrm{ridge})} + \mathrm{OG}_{n, n_e, n_r}$, assumptions $(i)$ and $(ii)$ imply that $\mathscr{R}_n(u_{\hat \theta}) \leqslant \inf_{u \in \mathrm{NN}_H(D)} R_{n, n_e,n_r}^{(\mathrm{ridge})}(u) + 2\varepsilon$.
    On the other hand, $R_{n, n_e,n_r}^{(\mathrm{ridge})} - \mathrm{OG}_{n, n_e, n_r} \leqslant \mathscr{R}_n$. 
   The proof of Theorem \ref{thm:generalization_error} reveals that there exists a sequence $(\theta(n_e, n_r))_{n_e, n_r \in \mathbb{N}} \in \Theta_{H,D}^{\mathbb{N}}$ such that $\lim_{n_e, n_r \to \infty}\mathrm{OG}_{n, n_e, n_r}(u_{\theta(n_e, n_r)}) = 0$ and $\lim_{n_e, n_r \to \infty }\mathscr{R}_n(u_{\theta(n_e, n_r)}) =  \inf_{u \in \mathrm{NN}_H(D)} \mathscr{R}_n(u)$. 
   Thus, $\inf_{u \in \mathrm{NN}_H(D)} R_{n, n_e,n_r}^{(\mathrm{ridge})}(u) \leqslant \inf_{\mathrm{NN}_H(D)} \mathscr{R}_n(u) + o_{n_e,n_r \to \infty}(1)$.
   We deduce that $$\mathscr{R}_n(u_{\hat \theta}) \leqslant \inf_{u\in \mathrm{NN}_H(D)} \mathscr{R}_n(u) + 2\varepsilon +  o_{n_e,n_r \to \infty}(1).$$ 
\end{proof}
\begin{lem}[Minimizing sequence of the theoretical risk.]
\label{lem:min}
    Let $H,D \in \mathbb{N}^\star$. Define the sequence $(v_p)_{p\in \mathbb{N}}\in \mathrm{NN}_H(D)^\mathbb{N}$ of neural networks by $v_p(\bx) = \tanh_p\circ \tanh^{\circ (H-1)}(\bx)$. Then, for any $\lambda_e > 0$,
    \[\lim_{p \to \infty} \lambda_e (1-v_p(1))^2 + \frac{1}{2}\int_{-1}^1 \bx^2 (v_p')^2(\bx)d\bx = 0.\]
\end{lem}
\begin{proof}
$\tanh^{\circ (H-1)}$ is an increasing $C^\infty$ function such that $\tanh^{\circ (H-1)}(0) = 0$. Therefore, Lemma \ref{lem:compTanh} shows that $\lim_{p \to \infty}v_p(1) = 1$, so that $\lim_{p\to\infty}\lambda_e(1-v_p(1))^2=0$. This shows the convergence of the left-hand term of the lemma.
    
To bound the right-hand term, we have, according to the chain rule, \[|v_p'(\bx)| \leqslant p \|\tanh^{\circ (H-1)}\|_{C^1(\mathbb{R})} |\tanh'(p\tanh^{\circ (H-1)}(\bx))|,\] with  $\|\tanh^{\circ (H-1)}\|_{C^1(\mathbb{R})} < \infty$ by Corollary \ref{cor:bounding_tanh}.
Thus, \[\int_{-1}^1 \bx^2 (v_p')^2(\bx) d\bx \leqslant  \|\tanh^{\circ (H-1)}\|_{C^1(\mathbb{R})}^2 \int_{-1}^1 p^2 \bx^2 (\tanh'(p\tanh^{\circ (H-1)}(\bx)))^2d\bx.\]
Notice that $\bx^2 (\tanh'(p\tanh^{\circ (H-1)}(\bx)))^2$ is an even function, so that 
\[\int_{-1}^1 \bx^2 (v_p')^2(\bx) d\bx \leqslant 2 \|\tanh^{\circ (H-1)}\|_{C^1(\mathbb{R})}^2 \int_{0}^1 p^2\bx^2 (\tanh'(p\tanh^{\circ (H-1)}(\bx)))^2d\bx.\]
Remark that $(\tanh')^2(\bx) = (1-\tanh(\bx))^2(1+\tanh(\bx))^2 \leqslant 16 \exp(-2\bx)$, 
so that 
\[\int_{-1}^1 \bx^2 (v_p')^2(\bx) d\bx \leqslant 32 \|\tanh^{\circ (H-1)}\|_{C^1(\mathbb{R})}^2 \int_{0}^1 p^2\bx^2 \exp(-2p\tanh^{\circ (H-1)}(\bx))d\bx.\]

If $H=1$, then the change of variable $\bar{\bx} = p\bx$ states that $$\int_{0}^1 p^2\bx^2 \exp{(-2p\bx)}d\bx \leqslant p^{-1}\int_{0}^\infty \bar{\bx}^2 \exp{(-2\bar{\bx})}d\bar{\bx} \xrightarrow{p\to \infty}0$$ and the lemma is proved.

If $H\geqslant 2$, notice that $\tanh(\bx) \geqslant \bx\mathbf{1}_{\bx\leqslant 1}/2 + \mathbf{1}_{\bx\geqslant 1}/2$ for all $\bx \geqslant 0$, and therefore we have that $\tanh^{\circ (H-1)}(\bx) \geqslant \bx\mathbf{1}_{\bx\leqslant 2^{H-1}}/2^H + \mathbf{1}_{\bx\geqslant 2^{H-1}}/2^H$. 
Thus, using the change of variable $\bar{\bx} = p\bx$,
\begin{align*}
    \int_{0}^1 p^2\bx^2 \exp(-2p\tanh^{\circ (H-1)}(\bx))d\bx &\leqslant \int_{0}^{1} p^2\bx^2 \exp(-2^{H-1}p\bx)d\bx\\  
    & \leqslant p^{-1} \int_0^\infty \bar{\bx}^2 \exp(-2^{H-1}\bar{\bx})d\bar{\bx}.
\end{align*}
Since this upper bound vanishes as $p\to \infty$, this concludes the proof when $H\geqslant 2$.

\end{proof}

\begin{defi}[Weak lower semi-continuity]
    A fonction $I: H^m(\Omega)\to \mathbb{R}$ is weakly lower semi-continuous on $H^m(\Omega)$ if, for any sequence $(u_p)_{p\in \mathbb{N}} \in H^m(\Omega)^{\mathbb{N}}$ that weakly converges to $u_\infty \in H^m(\Omega)$ in $H^m(\Omega)$, one has $I(u_\infty) \leqslant \liminf_{p \to \infty} I(u_p).$ 
\end{defi}

The following technical lemma will be useful for the proof of Proposition \ref{prop:sequenceCvLin}.
\begin{lem}[Weak lower semi-continuity with convex Lagrangians]
\label{lem:lowerSemiC0Lin}
    Let the Lagrangian $L\in C^\infty(\mathbb{R}^{\binom{d_1+m}{m}d_2}\times \cdots \times \mathbb{R}^{d_2}\times \mathbb{R}^{d_1}, \mathbb{R})$ be such that, for any $x^{(m)}, \hdots, x^{(0)}$, and $z$, the function $x^{(m+1)} \mapsto L(x^{(m+1)}, \hdots, x^{(0)}, z)$
     is convex and nonnegative.
    
    Then the function $I : u \mapsto \int_{\Omega}L((\partial^{m+1}_{i_1,\hdots, i_{m+1}}u(\bx))_{1\leqslant i_1,\hdots, i_{m+1} \leqslant d_1},\hdots,u(\bx), \bx)d\bx$ is lower-semi continuous for the weak topology on $H^{m+1}(\Omega,\mathbb{R}^{d_2})$.
\end{lem}

\begin{proof}
    This results generalizes \citet[Theorem 1, Chapter 8.2]{evans2010partial}, which treats the case $m=0$. Let $(u_p)_{p\in \mathbb{N}} \in H^{m+1}(\Omega, \mathbb{R}^{d_2})^{\mathbb{N}}$ be a sequence that weakly converges  to $u_\infty \in H^{m+1}(\Omega, \mathbb{R}^{d_2})$ in $H^{m+1}(\Omega, \mathbb{R}^{d_2})$. Our goal is to prove that $I(u_\infty) \leqslant \liminf_{p \to \infty} I(u_p)$.  Upon passing to a subsequence, we can suppose that $\lim_{p\to\infty} I(u_p) =\liminf_{p\to\infty} I(u_p)$.

    As a first step, we strengthen the convergence of $(u_p)_{p \in\mathbb{N}}$ by showing that for any $\varepsilon > 0$, there exists a subset $E_\varepsilon$ of $\Omega$ such that $|\Omega\backslash E_\varepsilon|\leqslant \varepsilon$ (the notation $|\cdot|$ stands for the Lebesgue measure), and such that there exists a subsequence that uniformly converges on $E_\varepsilon$, as well as its derivatives.
    Recalling that a weakly convergent sequence is bounded \citep[e.g.,][Chapter D.4]{evans2010partial}, one has $\sup_{p\in\mathbb{N}} \|u_p\|_{H^{m+1}(\Omega)} < \infty$. 
    Theorem \ref{thm:rellichK} ensures that a subsequence of $( u_p)_{p \in \mathbb{N}}$ converges to, say, $u_\infty \in H^{m+1}(\Omega, \mathbb{R}^{d_2})$ with respect to the $H^{m}(\Omega)$ norm. 
    Upon passing again to another subsequence, we conclude that for all $|\alpha|\leqslant m$ and for almost every $x$ in $\Omega$, $\lim_{p \to \infty} \partial^\alpha u_p(x) = \partial^\alpha u_\infty(x)$ \citep[see, e.g.][Theorem 4.9]{brezis2010functional}.
    Finally, by Egorov's theorem \citep[Chapter E.2]{evans2010partial}, for any $\varepsilon >0$, there exists a measurable set $E_\varepsilon$ such that $|\Omega\backslash E_\varepsilon| \leqslant \varepsilon$ and such that, for all $|\alpha| \leqslant m$, $\lim_{p\to\infty} \|\partial^\alpha u_p-\partial^\alpha u_\infty\|_{L^\infty(E_\varepsilon)} =0$. 
    
    Our next goal is to bound the function $L$.
    Let $F_\varepsilon = \{x \in \Omega, \sum_{|\alpha|\leqslant m+1}|\partial^\alpha u_\infty(x)| \leqslant \varepsilon^{-1}\}$ and $G_\varepsilon = E_\varepsilon \cap F_\varepsilon$. 
    Observe that $\lim_{\varepsilon \to 0}|\Omega \backslash G_\varepsilon| =0$.
    Since, for all $|\alpha|\leqslant m+1$,  $\|\partial^\alpha u_\infty\|_{\infty, G_\varepsilon}< \infty$,  
    and since $\lim_{p \to\infty} \|\partial^\alpha u_p-\partial^\alpha u_\infty\|_{L^\infty(G_\varepsilon)} =0$, 
    then, for $p$ large enough, $(\|\partial^\alpha u_p\|_{L^\infty(G_\varepsilon)})_{p\in\mathbb{N}}$ is bounded. 
    For now, to lighten notation, we denote $((\partial^{m+1}_{i_1,\hdots, i_{m+1}}u(z))_{1\leqslant i_1,\hdots, i_{m+1} \leqslant d_1},\hdots, u(z), z)$ by $(D^{m+1}u(z),\hdots, u(z), z)$.
    Therefore, since the Lagrangian $L$ is smooth and $\Omega$ is bounded, for all $p$ large enough, $(\| L(D^{m+1}u_p(\cdot), \hdots, Du_p(\cdot), u_p(\cdot), \cdot)\|_{L^\infty(G_\varepsilon)})_{p\in\mathbb{N}}$ is bounded as well.
    
    To conclude the proof, we take advantage of the convexity of the Lagrangian $L$.
    Let $J_{m+1}$ be the Jacobian matrix of $L$ along the vector $x^{(m+1)}$.
    The convexity of $L$ implies 
    \begin{align*}
        &L(D^{m+1}u_p(z), \hdots, u_p(z), z)\\
        &\quad \geqslant L(D^{m+1}u_\infty(z), D^{m}u_p(z)\hdots, u_p(z), z) \\
        & \qquad+ J_{m+1}(D^{m+1}u_\infty(z), D^{m}u_p(z)\hdots, u_p(z), z) \times (D^{m+1}u_p(z)-D^{m+1}u_\infty(z)). 
    \end{align*}
    Using the fact that $L\geqslant0$ and that $I(u_p) \geqslant \int_{G_\varepsilon}L(D^{m+1}u_p(z), \hdots, u_p(z), z)dz$, 
    we obtain
    \begin{align*}
        I(u_p) &\geqslant \int_{G_\varepsilon } L(D^{m+1}u_\infty(z), D^m u_p(z), \hdots, u_p(z), z) \\
        &\quad + J_{m+1}(D^{m+1}u_\infty(z), D^m u_p(z), \hdots, u_p(z), z) \times (D^{m+1}u_p(z)-D^{m+1}u_\infty(z))dz.
    \end{align*}
    Since $(\| L(D^{m+1}u_p(\cdot), \hdots, Du_p(\cdot), u_p(\cdot), \cdot)\|_{L^\infty(G_\varepsilon)})_{p\in\mathbb{N}}$ is bounded for $p$ large enough, and since, for all $|\alpha| \leqslant m$, $\lim_{p\to\infty} \|\partial^\alpha u_p-\partial^\alpha u_\infty\|_{L^\infty(G_\varepsilon)} =0$, the dominated convergence theorem ensures that  
    \begin{align*}
        &\lim_{p\to \infty}\int_{G_\varepsilon }  \!\!\! L(D^{m+1}u_\infty(z), D^m u_p(z), \hdots, u_p(z), z)dz = \int_{G_\varepsilon }  \!\!\! L(D^{m+1}u_\infty(z), \hdots, u_\infty(z), z)dz. 
    \end{align*}
    Since $(i)$ $L$ is smooth (and therefore Lipschitz on bounded domains), 
   $(ii)$ for all $p$ large enough,  $(\|\partial^\alpha u_p\|_{L^\infty(G_\varepsilon)})_{p\in\mathbb{N}}$ is bounded, and $(iii)$ for all $|\alpha| \leqslant m$, $\lim_p \|\partial^\alpha u_p-\partial^\alpha u_\infty\|_{L^\infty(G_\varepsilon)} =0$,  $\lim_{p \to \infty }\|J_{m+1}(D^{m+1}u_\infty(\cdot), D^m u_p(\cdot), \hdots, u_p(\cdot), \cdot) - J_{m+1}(D^{m+1}u_\infty(\cdot), \hdots, u_\infty(\cdot), \cdot) \|_{L^\infty(G_\varepsilon)} = 0$. Therefore, since $D^{m+1}u_p \rightharpoonup D^{m+1}u_\infty$, 
    \[\lim_{p\to \infty}\int_{G_\varepsilon}\!\!\!J_{m+1}(D^{m+1}u_\infty(z), D^m u_p(z), \hdots, u_p(z), z) \times (D^{m+1}u_p(z)-D^{m+1}u_\infty(z))dz = 0.\]
    Hence,
    $\lim_{p\to\infty} I(u_p) \geqslant \int_{G_\varepsilon } L(D^{m+1}u_\infty(z), \hdots, u_\infty(z), z)dz$.
    Finally, applying the monotone convergence theorem with $\varepsilon \to 0$ shows that $\lim_{p\to\infty} I(u_p) \geqslant I(u_\infty)$, which is the desired result.
\end{proof}

\begin{lem}[Measurability of $\hat u_n$]
    \label{lem:measurability}
    Let $\hat u_n = \argmin_{u \in H^{m+1}(\Omega, \mathbb{R}^{d_2})} \mathscr{R}_n^{(\mathrm{reg})}(u)$, where, for all $u \in H^{m+1}(\Omega, \mathbb{R}^{d_2})$, 
    \begin{align*} 
        \mathscr{R}^{(\mathrm{reg})}_n(u) &= \frac{\lambda_d}{n} \sum_{i=1}^n \|\tilde \Pi(u)(\bX_i) - Y_i\|_2^2 +\lambda_e\mathbb{E}\|\tilde \Pi(u)(\bX^{(e)})-h(\bX^{(e)})\|_2^2\\
        & \quad + \frac{1}{|\Omega|}\sum_{k=1}^M \|\mathscr{F}_k(u, \cdot) \|_{L^2(\Omega)} + \lambda_t \|u\|_{H^{m+1}(\Omega)}^2.
    \end{align*}
    Then $\hat u_n$ is a random variable.
\end{lem}
\begin{proof}
Recall that \[
\mathscr{R}_n^{(\mathrm{reg})}(u) = \mathcal{A}_n(u,u) -2\mathcal{B}_n(u) + \frac{\lambda_d}{n} \sum_{i=1}^n \|Y_i\|^2 +\lambda_e \mathbb{E}\|h(\bX^{(e)})\|_2^2 + \frac{1}{|\Omega|}\sum_{k=1}^M\int_{\Omega} B_k(\bx)^2d\bx.
\]
Throughout we use the notation $\mathcal{A}_{(\bx, e)}(u,u)$ instead of $\mathcal{A}_n(u,u)$, to make the dependence of $\mathcal{A}_n$ in the random variables $\bx = (\bX_1, \hdots, \bX_n)$ and $e=(\varepsilon_1, \hdots, \varepsilon_n)$ more explicit. We do the same with $\mathcal{B}_n$. 
For a given a normed space $(F, \|\cdot\|)$,  we let $\mathscr{B}(F, \|\cdot\|)$ be the Borel $\sigma$-algebra on $F$ induced by the norm $\|\cdot\|$. 

Our goal is to prove that the function 
\begin{align*}
    \hat  u_n : (\Omega^n\! \times \!\mathbb{R}^{nd_2}, \mathscr{B}(\Omega^n\!\times\! \mathbb{R}^{nd_2}, \|\!\cdot\!\|_2)) &\to (H^{m+1}(\Omega, \mathbb{R}^{d_2}), \mathscr{B}(H^{m+1}(\Omega, \mathbb{R}^{d_2}), \|\!\cdot\!\|_{H^{m+1}(\Omega)}))\\
    (\bx, e) & \mapsto \argmin_{u \in H^{m+1}(\Omega, \mathbb{R}^{d_2})} \mathcal{A}_{(\bx, e)}(u,u)  - 2 \mathcal{B}_{(\bx, e)}(u)
\end{align*} is measurable. 
Recall that  $H^{m+1}(\Omega, \mathbb{R}^{d_2})$ is a Banach space separable with respect to its norm $\|\cdot\|_{H^{m+1}(\Omega)}$. 
Let $(v_q)_{q \in \mathbb{N}} \in H^{m+1}(\Omega, \mathbb{R}^{d_2})^\mathbb{N}$ be a sequence dense in $H^{m+1}(\Omega, \mathbb{R}^{d_2})$. 
Note that, for any $\bx\in \Omega^n$ and any $e \in \mathbb{R}^{nd_2}$, one has $\min_{u \in H^{m+1}(\Omega, \mathbb{R}^{d_2})} \mathcal{A}_{(\bx, e)}(u,u) - 2\mathcal{B}_{(\bx, e)}(u) = \inf_{q \in \mathbb{N}} \mathcal{A}_{(\bx, e)}(v_q,v_q) - 2\mathcal{B}_{(\bx, e)}(v_q)$. 
This identity is a consequence of the fact that the function $u \mapsto \mathcal{A}_{(\bx, e)}(u,u) - 2\mathcal{B}_{(\bx, e)}(u)$ is continuous for the $H^{m+1}(\Omega)$ norm, as shown in the proof of Proposition \ref{prop:laxMLin}).
Moreover, according to this proof, each function $F_q(\bx, e) := \mathcal{A}_{(\bx,  e)}(u_q, u_q) - 2 \mathcal{B}_{(\bx, e)}(u_q)$ is a composition of  continuous functions, and is therefore measurable. 
Thus, the function 
\begin{align*}
    G(\bx, e) :=  \min_{u \in H^{m+1}(\Omega, \mathbb{R}^{d_2})} \mathcal{A}_{(\bx, e)}(u, u) - 2 \mathcal{B}_{(\bx, e)}(u)= \inf_{q \in \mathbb{N}} \mathcal{A}_{(\bx, e)}(u_q, u_q) - 2 \mathcal{B}_{(\bx, e)}(u_q)
\end{align*} is measurable.

Next, since $\Omega$, $\mathbb{R}$, and $H^{m+1}(\Omega, \mathbb{R}^{d_2})$ are separable, we know that  the $\sigma$-algebras
$\mathscr{B}(\Omega^n\times\mathbb{R}^{nd_2}\times H^{m+1}(\Omega, \mathbb{R}^{d_2}), \|\cdot\|_{\otimes})$ and $ \mathscr{B}(\Omega^n  \times \mathbb{R}^{nd_2}, \|\cdot\|_2)\otimes \mathscr{B}(H^{m+1}(\Omega, \mathbb{R}^{d_2}), \|\cdot\|_{H^{m+1}(\Omega)})$ are identical, 
where $\|(\bx,e, u)\|_{\otimes} = \|(\bx,e)\|_2 +\|u\|_{H^{m+1}(\Omega)}$ \citep[see, e.g.][Chapter II.13, E13.11c]{rogers1979diffusions}. 
This implies that the coordinate projections $\Pi_{\bx,e}$ and $\Pi_u$---defined for $(\bx,e) \in \Omega^n\times\mathbb{R}^{nd_2}$ and $u \in H^{m+1}(\Omega, \mathbb{R}^{d_2})$ by $\Pi_{\bx,e}(\bx,e,u) = (\bx,e)$ and $\Pi_u(\bx,e,u) = u$---are $\|\cdot\|_\otimes$ measurable.
It is easy to check that, for any $(\bx,e) \in \Omega^n\times \mathbb{R}^{nd_2}$ and $u\in H^{m+1}(\Omega, \mathbb{R}^{d_2})$, if $\lim_{p \to \infty }\|(\bx_p, e_p, u_p)-(\bx, e, u)\|_\otimes =0$, then $\lim_{p\to \infty}\|\tilde \Pi(u_p)-\tilde \Pi(u)\|_{\infty, \Omega} = 0$ and, since $\tilde \Pi (u) \in C^0(\Omega, \mathbb{R}^{d_2})$, we know that $\lim_{p\to \infty }\mathcal{A}_{\bx_p, e_p}(u_p,u_p) - 2\mathcal{B}_{\bx_p, e_p}(u_p) = \mathcal{A}_{\bx,e}(u,u) - 2\mathcal{B}_{\bx,e}(u)$.
This proves that $I : (\Omega^n\times \mathbb{R}^{nd_2} \times H^{m+1}(\Omega, \mathbb{R}^{d_2}), \mathscr{B}(\Omega^n\times \mathbb{R}^{nd_2}\times H^{m+1}(\Omega, \mathbb{R}^{d_2}), \|\cdot\|_{\otimes})) \to (\mathbb{R}, \mathscr{B}(\mathbb{R}))$ defined by 
\[I(\bx, e, u) = \mathcal{A}_{(\bx, e)}(u,u) - 2 \mathcal{B}_{(\bx, e)}(u)\] is continuous with respect to $\|\cdot\|_\otimes$ and therefore measurable. According to the above, the function \[\tilde I(\bx, e, u) = I(\bx, e, u) -  G\circ \Pi_{x,e}(\bx, e, u)\] 
is also measurable.
Observe that, by definition, $\hat u_n = J\circ (\bX_1, \hdots, \bX_n, \varepsilon_1, \hdots, \varepsilon_n)$, where $J(\bx, e) = \Pi_u(\tilde I^{-1}(\{0\})\cap (\{(\bx, e)\}\times H^{m+1}(\Omega, \mathbb{R}^{d_2})))$.
For any  set $S \in \mathscr{B}(H^{m+1}(\Omega, \mathbb{R}^{d_2}, \|\cdot\|_{H^{m+1}(\Omega)})$, $J^{-1}(S) = \Pi_{x,e}(\tilde{I}^{-1}(\{0\}) \cap (\Omega^n\times \mathbb{R}^{nd_2} \times S)) \in \mathscr{B}(\Omega^n\times \mathbb{R}^{nd_2})$. 
(Notice that $J^{-1}(S)$ is the collection of all pairs $(\bx, e)\in \Omega^n\times \mathbb{R}^{nd_2}$ satisfying $\argmin_{u \in H^{m+1}(\Omega, \mathbb{R}^{d_2})} \mathcal{A}_{(\bx, e)}(u,u)  - 2 \mathcal{B}_{(\bx, e)}(u) \in S$.)
To see this, jut note that for any set $\tilde S \in \mathscr{B}(\Omega^n \times \mathbb{R}^{nd_2}, \|\cdot\|_2)\otimes \mathscr{B}(H^{m+1}(\Omega, \mathbb{R}^{d_2}), \|\cdot\|_{H^{m+1}(\Omega, \mathbb{R}^{d_2})})$, one has $\Pi_{x,e}(\tilde S) \in \mathscr{B}(\Omega^n \times \mathbb{R}^{nd_2}, \|\cdot\|_2)$ \citep[see, e.g.][Lemma 11.4, Chapter II]{rogers1979diffusions}. We conclude that the function $J$ is measurable and so is $\hat u_n$.
\end{proof}

Let $B( 1, \|\cdot\|_{H^{m+1}(\Omega)}) = \{u \in H^{m+1}(\Omega, \mathbb{R}^{d_2}), \quad \|u\|_{H^{m+1}(\Omega)}\leqslant 1\}$ be the ball of radius $r$ centered at $0$. 
Let $N(B(1, \|\cdot\|_{H^{m+1}(\Omega)})), \|\cdot\|_{H^{m+1}(\Omega)}, r)$ be the minimum number of balls of radius $r$ according to the norm $\|\cdot\|_{H^{m+1}(\Omega)}$ needed to cover the space $B( 1, \|\cdot\|_{H^{m+1}(\Omega)})$.
\begin{lem}[Entropy of $H^{m+1}(\Omega, \mathbb{R}^{d_2})$]
Let $\Omega\subseteq \mathbb{R}^{d_1}$ be a Lipschitz domain. For $m \geqslant 1$, one has
    \label{lem:entropy}
    \[\log N(B( 1, \|\cdot\|_{H^{m+1}(\Omega)}), \|\cdot\|_{H^{m+1}(\Omega)}, r) = \Oequivalent_{r \to 0} (r^{-d_1/(m+1)}).\]
\end{lem}
\begin{proof}
According to the extension theorem \citep[][Theorem 5, Chapter VI.3.3]{stein1970lipschitz}, there exists a constant $C_{\Omega} >0$, depending only on $\Omega$, such that any $u\in H^{m+1}(\Omega, \mathbb{R}^{d_2})$ can be extended to $\tilde u\in H^{m+1}(\mathbb{R}^{d_1}, \mathbb{R}^{d_2})$, with
$\|\tilde u \|_{H^{m+1}(\mathbb{R}^{d_1})} \leqslant C_{\Omega} \| u \|_{H^{m+1}(\Omega)}$. 
Let $r > 0$ be such that $\Omega \subseteq B(r, \|\cdot\|_2)$ and let $\phi \in C^\infty(\mathbb{R}^{d_1}, \mathbb{R}^{d_2})$ be such that
\[\phi(\bx) = \left\{\begin{array}{ll}
    1  &\text{for } \bx \in \Omega\\
    0  &\text{for } \bx \in\mathbb{R}^{d_1}, |x|\geqslant r.
\end{array}\right. \]
Then, for any $ u \in H^{m+1}(\Omega, \mathbb{R}^{d_2})$, $(i)$ $\phi \tilde u \in H^{m+1}(\mathbb{R}^{d_1}, \mathbb{R}^{d_2})$, $(ii)$ $\phi \tilde u |_\Omega = u$, and $(iii)$ there exists a constant $\tilde C_\Omega > 0$ such that $\|\phi \tilde u\|_{H^{m+1}(\mathbb{R}^{d_1})} \leqslant \tilde C_\Omega \|u\|_{H^{m+1}(\Omega)}$.
The lemma follows from \citet[Corollary 4]{nickl2007bracketing}.     
\end{proof}

\begin{lem}[Empirical process $L^2$]
    \label{lem:empiricalL2}
    Let $\bX_1, \hdots, \bX_n$ be i.i.d.~random variables, with common distribution $\mu_\bX$ on $\Omega$. Then there exists a constant $C_\Omega >0$, depending only on $\Omega$, such that
    \[\mathbb{E}\Big(\sup_{\|u\|_{H^{m+1}(\Omega)}\leqslant 1} \mathbb{E}\|\tilde \Pi(u)(\bX)\|_2^2- \frac{1}{n}\sum_{i=1}^n \|\tilde \Pi(u)(\bX_i)\|_2^2\Big) \leqslant \frac{d_2^{1/2} C_\Omega}{n^{1/2}},\]
    and
    \[\mathbb{E}\Big(\Big(\sup_{\|u\|_{H^{m+1}(\Omega)}\leqslant 1} \mathbb{E}\|\tilde \Pi(u)(\bX)\|_2^2- \frac{1}{n}\sum_{i=1}^n \|\tilde \Pi(u)(\bX_i)\|_2^2\Big)^2\Big) \leqslant \frac{d_2 C_\Omega}{n},\]
    where $\tilde \Pi$ is the Sobolev embedding (see Theorem \ref{thm:sobIneq}).
\end{lem}
\begin{proof}
    For any $u \in H^{m+1}(\Omega, \mathbb{R}^{d_2})$, let 
\[Z_{n,u} =  \mathbb{E}\|\tilde \Pi(u)(\bX_i)\|_2^2 - \frac{1}{n} \sum_{j=1}^n \|\tilde \Pi(u)(\bX_i)\|_2^2 \quad \text{and} \quad Z_n = \sup_{\|u\|_{H^{m+1}(\Omega)}\leqslant 1} Z_{n,u}.\] 
For any $u,v \in  H^{m+1}(\Omega, \mathbb{R}^{d_2})$ such that $\|u\|_{H^{m+1}(\Omega)} \leqslant 1$ and $\|v\|_{H^{m+1}(\Omega)} \leqslant 1$, we have
\begin{align*}
    &\Big|\frac{1}{n}( \|\tilde \Pi(u)(\bX_i)\|_2^2 - \mathbb{E}\|\tilde \Pi(u)(\bX_i)\|_2^2)  - \frac{1}{n} (\|\tilde \Pi(v)(\bX_i)\|_2^2 - \mathbb{E}\|\tilde \Pi(v)(\bX_i)\|_2^2)\Big| \\
    &\quad  \leqslant  \frac{2}{n}(\| \tilde \Pi (u-v)(\bX_i)\|_2 + \mathbb{E}\|\tilde \Pi (u-v)(\bX_i)\|_2)\\
    &\quad \leqslant \frac{4 C_\Omega}{n}\sqrt{d_2} \|u-v\|_{H^{m+1}(\Omega)} \qquad\qquad\qquad \text{(by applying Theorem \ref{thm:sobIneq}).}
\end{align*}
Therefore, applying Hoeffding's, Azuma's and Dudley's theorem similarly as in the proof of Theorem \ref{thm:approx_integral} shows that
\[\mathbb{E}(Z_n) \leqslant 24 C_\Omega d_2^{1/2} n^{-1}  \int_0^\infty [\log N(B( 1, \|\cdot\|_{H^{m+1}(\Omega)}), \|\cdot\|_{H^{m+1}(\Omega)}, r)]^{1/2}dr.\]
Lemma \ref{lem:entropy} shows that there exists a constant $C_\Omega'$, depending only on $\Omega$, such that 
$\mathbb{E}(Z_n) \leqslant C'_\Omega d_2^{1/2} n^{-1/2}$. 
Applying McDiarmid's inequality  as in the proof of Theorem \ref{thm:approx_integral} shows that
$\mathrm{Var}(Z_n) \leqslant 16 C_\Omega^2 d_2 n^{-1}$.
Finally, since $\mathbb{E}(Z_n^2) \leqslant \mathrm{Var}(Z_n) + \mathbb{E}(Z_n)^2$, we deduce that
\[\mathbb{E}(Z_n^2)\leqslant \frac{d_2}{n}\big((C'_\Omega)^2+16 C_\Omega^2\big).\]
\end{proof}

\begin{lem}[Empirical process]
    \label{lem:empiricalProcess}
    Let $\bX_1, \hdots, \bX_n, \varepsilon_1, \hdots, \varepsilon_n$ be independent random variables, such that $\bX_i$ is distributed along $\mu_\bX$ and $\varepsilon_i$ is distributed along $\mu_\varepsilon$, such that $\mathbb{E}(\varepsilon)=0$. Then there exists a constant $C_\Omega >0$, depending only on $\Omega$, such that
    \[\mathbb{E}\Big(\Big(\sup_{\|u\|_{H^{m+1}(\Omega)}\leqslant 1} \frac{1}{n} \sum_{j=1}^n \langle \tilde \Pi (u)(\bX_j)- \mathbb{E}(\tilde \Pi (u)(\bX)),\varepsilon_j\rangle \Big)^2\Big) \leqslant \frac{d_2 \mathbb{E}\|\varepsilon\|_2^2}{n} C_\Omega,\]
    where $\tilde \Pi$ is the Sobolev embedding.
\end{lem}

\begin{proof}
First note, since $H^{m+1}(\Omega, \mathbb{R}^{d_2})$ is separable and since, for all $u \in H^{m+1}(\Omega, \mathbb{R}^{d_2})$, the function $(\bx_1, \hdots, \bx_n, e_1, \hdots, e_n) \mapsto \frac{1}{n} \sum_{j=1}^n \langle \tilde \Pi (u)(\bx_j)- \mathbb{E}(\tilde \Pi (u)(\bX)),e_j\rangle$ is continuous,  that the quantity $Z = \sup_{\|u\|_{H^{m+1}(\Omega)}\leqslant 1} \frac{1}{n} \sum_{j=1}^n \langle \tilde \Pi (u)(\bX_j) - \mathbb{E}(\tilde \Pi (u)(\bX)),\varepsilon_j\rangle$ is a random variable.
Moreover, $|Z|\leqslant 2 C_\Omega \sqrt{d_2} \sum_{j=1}^n \|\varepsilon_j\|_2/n$, where $C_\Omega$ is the constant of Theorem \ref{thm:sobIneq}. 
Thus, $\mathbb{E}(Z^2) < \infty$.

Define, for any $u \in H^{m+1}(\Omega, \mathbb{R}^{d_2})$, 
\[Z_{n,u} =  \frac{1}{n} \sum_{j=1}^n \langle \tilde \Pi (u)(\bX_j) - \mathbb{E}(\tilde \Pi (u)(\bX)),\varepsilon_j\rangle \quad \text{and} \quad Z_n = \sup_{\|u\|_{H^{m+1}(\Omega)}\leqslant 1} Z_{n,u}.\] 
For any $u,v \in  H^{m+1}(\Omega, \mathbb{R}^{d_2})$, we have
\begin{align*}
    &\Big|\frac{1}{n} \langle \tilde \Pi (u)(\bX_i) - \mathbb{E}(\tilde \Pi (u)(\bX)),\varepsilon_i\rangle  - \frac{1}{n} \langle \tilde \Pi (v)(\bX_i) - \mathbb{E}(\tilde \Pi (u)(\bX)),\varepsilon_i\rangle \Big| \\
    &\quad  =  \frac{1}{n}|\langle \tilde \Pi (u-v)(\bX_i) - \mathbb{E}(\tilde \Pi (u-v)(\bX)),\varepsilon_i\rangle|\\
    &\quad \leqslant \frac{2 C_\Omega}{n}\sqrt{d_2} \|u-v\|_{H^{m+1}(\Omega)}\|\varepsilon_i\|_2 \qquad\qquad\qquad \text{(by applying Theorem \ref{thm:sobIneq}).}
\end{align*}
Using that $\varepsilon$ is independent of $\bX$, so that the conditional expectation of $Z_n$ is indeed a real expectation with $\varepsilon_1, \hdots, \varepsilon_n$ fixed, we can apply Hoeffding's, Azuma's and Dudley's theorem similarly as in the proof of Theorem \ref{thm:approx_integral} to show that
\begin{align*}
    \mathbb{E}(Z_n\mid \varepsilon_1, \hdots, \varepsilon_n) &\leqslant \frac{24 C_\Omega}{n} \sqrt{d_2}\Big(\sum_{i=1}^n \|\varepsilon_i\|_2^2\Big)^{1/2} \\
    &\quad \times \int_0^\infty[\log N(B( 1, \|\cdot\|_{H^{m+1}(\Omega)}), \|\cdot\|_{H^{m+1}(\Omega)}, r)]^{1/2}dr.
\end{align*}
Hence, according to Lemma \ref{lem:entropy}, there exists a constant $C_\Omega'>0$, depending only on $\Omega$, such that $\mathbb{E}(Z_n \mid \varepsilon_1, \hdots, \varepsilon_n) \leqslant C'_\Omega n^{-1} \sqrt{d_2}\Big(\sum_{i=1}^n \|\varepsilon_i\|_2^2\Big)^{1/2}$.
We deduce that
\begin{align*}
    \mathbb{E}(Z_n) \leqslant C'_\Omega  \sqrt{d_2} \frac{(\mathbb E \|\varepsilon\|_2^2)^{1/2}}{n^{1/2}},
\end{align*}
and 
\begin{align*}
    \mathrm{Var}(\mathbb{E}(Z_n\mid \varepsilon_1, \hdots, \varepsilon_n)) \leqslant \mathbb E(\mathbb{E}(Z_n\mid \varepsilon_1, \hdots, \varepsilon_n)^2)  \leqslant (C'_\Omega)^2  d_2 \frac{\mathbb E \|\varepsilon\|_2^2}{n}.
\end{align*}
Applying McDiarmid's inequality as in the proof of Theorem \ref{thm:approx_integral} shows that
\[\mathrm{Var}(Z_n\mid \varepsilon_1, \hdots, \varepsilon_n) \leqslant 16 C_\Omega^2 d_2 \frac{1}{n^2}\sum_{i=1}^n \|\varepsilon_i\|_2^2. \]
The law of the total variance ensures that 
\begin{align*}
    \mathrm{Var}(Z_n) &= \mathrm{Var}(\mathbb{E}(Z_n\mid \varepsilon_1, \hdots, \varepsilon_n)) + \mathbb{E}(\mathrm{Var}(Z_n\mid \varepsilon_1, \hdots, \varepsilon_n))\\
    & \leqslant \frac{d_2 \mathbb E \|\varepsilon\|_2^2}{n}\big((C'_\Omega)^2+16 C_\Omega^2\big).
\end{align*}
Since $\mathbb{E}(Z_n^2) \leqslant \mathrm{Var}(Z_n) + \mathbb{E}(Z_n)^2$, we deduce that
\[\mathbb{E}(Z_n^2)\leqslant \frac{d_2 \mathbb E \|\varepsilon\|_2^2}{n}\big(2(C'_\Omega)^2+16 C_\Omega^2\big).\]
\end{proof}

\section{Proofs of Proposition \ref{prop:densite}} 
\label{proof:density}
\citet[][Theorem 5.1]{ryck2021approximation} ensures that $\text{NN}_2$ is dense in $(C^\infty([0,1]^{d_1}, \mathbb{R}), \|\cdot\|_{C^K([0,1]^{d_1})})$ for all $d_1 \geqslant 1$ and $K\in \mathbb{N}$. Note that the authors state the result for Hölder spaces $(W^{K+1,\infty}([0,1]^{d_1}), \|\cdot\|_{W^{K,\infty}(]0,1[^{d_1})})$ \citep[see][for a definition]{evans2010partial}. Clearly, $C^\infty([0,1]^{d_1}) \subseteq W^{K+1, \infty}([0,1]^{d_1})$ and the norms $\|\cdot\|_{C^{K}}$ and $\|\cdot\|_{W^{K,\infty}}$ coincide on $C^\infty([0,1]^{d_1})$.

Our proof generalizes this result to any bounded Lipschitz domain $\Omega$, to any number $H\geqslant 2$ of layers, and to any output dimension $d_2$. We stress that for any $U \subseteq \mathbb{R}^{d_1}$, the set $\mathrm{NN}_2 \subseteq C^\infty(\mathbb{R}^{d_1}, \mathbb{R}^{d_2})$ can of course be seen as a subset of $C^\infty(U, \mathbb{R}^{d_2})$.

\paragraph{Generalization to any bounded Lipschitz domain $\Omega$} 
In this and the next paragraph, $d_2=1$.
Our objective is to prove that $\text{NN}_2$ is dense in $(C^\infty(\bar{\Omega}, \mathbb{R}), \|\cdot\|_{C^K(\Omega)})$. 
Let $f \in C^\infty(\bar{\Omega}, \mathbb{R})$. Since $\Omega$ is bounded, there exists an affine transformation $\tau : x \mapsto A_\tau x +b_\tau$, with $A_\tau \in \mathbb{R}^\star$ and $b_\tau \in \mathbb{R}^{d_1}$, 
such that $\tau(\Omega) \subseteq [0,1]^d$. Set  $\hat{f} = f (\tau^{-1})$. 
According to the extension theorem for Lipschitz domains of \citet[][Theorem 5 Chapter VI.3.3]{stein1970lipschitz}, the function $\hat{f}$ can be extended to a function $\tilde{f} \in W^{K,\infty}([0,1]^{d_1})$ such that $\tilde{f}|_{\tau(\Omega)} = \hat{f}|_{\tau(\Omega)}$. 
Fix $\epsilon > 0$. According to \citet[][Theorem 5.1]{ryck2021approximation}, there exists $u_\theta \in \text{NN}_2$ such that 
$\|u_\theta - \hat{f}\|_{W^{K, \infty}([0,1]^d)} \leqslant \epsilon$. Since $\tilde{f}$ is an extension of 
$\hat{f}$,  $\tilde{f}|_{\tau(\Omega)} \in C^\infty(\bar{\Omega})$ and one also has $\|u_\theta - \hat{f}\|_{C^K(\tau(\Omega))} \leqslant \epsilon$. 

Now, let $m\in \mathbb{N}$ and let $\alpha$ be a multi-index such that $\sum_{i=1}^{d_1}\alpha_i = m$. Then, clearly,
$\partial^\alpha (\hat{f}(\tau))= A_\tau^m \times \partial^\alpha \hat{f}(\tau)$. 
Therefore,
$\|u_\theta(\tau) - \hat{f}(\tau)\|_{C^K(\Omega)} \leqslant \epsilon\times \max(1,A_\tau^K)$,
that is
\[\|u_\theta(\tau) - f\|_{C^K(\Omega)} \leqslant \epsilon\times \max(1,A_\tau^K).\] 
But, since $\tau$ is affine, $u_\theta(\tau)$ belongs to $\text{NN}_2$. This is the desires result.
\paragraph{Generalization to any number $H\geqslant 2$ of layers} We show in this paragraph that $\text{NN}_H$ is dense in $(C^\infty(\bar{\Omega}, \mathbb{R}), \|\cdot\|_{C^K(\Omega)})$ for all $H \geqslant 2$. The case $H=2$ has been treated above and it is therefore assumed that $H \geqslant 3$.

Let $f \in C^\infty(\bar{\Omega}, \mathbb{R})$. Introduce the function $v$ defined by
\[v(x_1, \hdots, x_{d_1}) = (\tanh^{\circ (H-2)}(x_1), \hdots, \tanh^{\circ (H-2)}(x_{d_1})),\] 
where $\tanh^{\circ (H-2)}$ stands for the $\tanh$ function composed $(H-2)$ times with itself.
For all $u_\theta \in \text{NN}_2$, $u_\theta(v) \in \text{NN}_H$ is a neural network such that the first weights matrices $(W_\ell)_{1 \leqslant \ell \leqslant H-2}$ are identity matrices and the first offsets $(b_\ell)_{1 \leqslant \ell \leqslant H-2}$ are equal to zero. 
Since $\tanh$ is an increasing $C^\infty$ function, $v$ is a $C^\infty$ diffeomorphism.
Therefore, $v(\Omega)$ is a bounded Lipschitz domain and $f(v^{-1}) \in C^\infty(v(\Omega), \mathbb{R})$. Lemma \ref{lem:boundPartialDer2} shows that $f(v^{-1}) \in C^\infty(\bar{v}(\Omega), \mathbb{R})$, where $\bar{v}(\Omega)$ is the closure of $v(\Omega)$. According to the previous paragraph, there exists a sequence $(\theta_m)_{m \in \mathbb{N}}$ of parameters such that $u_{\theta_m} \in \text{NN}_2$ and 
\begin{equation*}
    \lim_{m \to \infty} \|u_{\theta_m} - f(v^{-1})\|_{C^K(v(\Omega))} = 0.
\end{equation*}
Thus, $u_{\theta_m}$ approximates $f(v^{-1})$, and we would like $u_{\theta_m} (v)$ to approximate $f$.
From Lemma \ref{lem:boundPartialDer2}, \[ \|u_{\theta_m}(v)-f\|_{C^K(\Omega)} \leqslant B_K \times  \|u_{\theta_m}-f\circ v^{-1}\|_{C^K(\Omega)} \times (1+\|\tanh^{\circ H-2}\|_{C^K(\mathbb{R})})^K,\]
while Corollary \ref{cor:bounding_tanh} asserts that $\|\tanh^{\circ H-2}\|_{C^K(\mathbb{R})} < \infty$.
 Therefore, we deduce that $\lim_{m \to \infty }\|u_{\theta_m}(v)-f\|_{C^K(\Omega)}  = 0$ with $u_{\theta_m}(v) \in \text{NN}_H$, which proves the lemma for  $H \geqslant 2$.

\paragraph{Generalization to all output dimension $d_2$} We have shown so far that for all $H \geqslant 2$, $\text{NN}_H$ is dense in $(C^\infty(\bar{\Omega}, \mathbb{R}), \|\cdot\|_{C^K(\Omega)})$. It remains to establish that $\text{NN}_H$ is dense in $(C^\infty(\bar{\Omega}, \mathbb{R}^{d_2}), \|\cdot\|_{C^K(\Omega)})$ for any output dimension $d_2$.

Let $f = (f_1, \hdots, f_{d_2}) \in C^\infty(\Omega, \mathbb{R}^{d_2})$. For all $1\leqslant i \leqslant d_2$, let $(\theta_m^{(i)})_{m \in \mathbb{N}} \in (\text{NN}_H)^{\mathbb{N}}$ be a sequence of neural networks such that $\lim_{m \to \infty}\|u_{\theta^{(i)}_m}-f_i\|_{C^K(\Omega)} = 0$. Denote by $u_{\theta_m} = (u_{\theta_m^{(1)}}, \hdots, u_{\theta_m^{(d_2)}})$ the stacking of these sequences. For all $m\in \mathbb{N}$, $u_{\theta_m} \in \text{NN}_H$ and $\lim_{m \to \infty }\|u_{\theta_m}-f\|_{C^K(\Omega)} = 0$. Therefore, $\text{NN}_H$ is dense in $(C^\infty(\bar{\Omega}, \mathbb{R}), \|\cdot\|_{C^K(\Omega)})$.

\section{Proofs of Section \ref{POF}}

\subsection*{Proof of Proposition \ref{prop:friction}}
\label{proof:hybrid_modeling_failure}
Consider $u_{\hat{\theta}(p, n_r,D)} \in \text{NN}_H(D)$, the neural network defined by
    \[u_{\hat{\theta}(p, n_r,D)}(\bx) =  Y_{(1)} + \sum_{i=1}^{n-1} \frac{Y_{(i+1)}-Y_{(i)}}{2} \bigg[\tanh_p^{\circ H}\Big(\bx-\bX_{(i)}-\frac{\delta(n,n_r)}{2}\Big)+1\bigg],\]
where $\delta(n,n_r)$ is defined in \eqref{def:delta} and where the observations have been reordered according to increasing values of the $\bX_{(i)}$. 
According to Lemma \ref{lem:estimationTanh}, one has, for all $1 \leqslant i \leqslant n$, $\lim_{p \to \infty} u_{\hat{\theta}(p, n_r,D)}(\bX_i) = Y_i$. Moreover, for all order $K\geqslant 1$ of differentiation and all $1 \leqslant j \leqslant n_r$, $\lim_{p\to\infty}u^{(K)}_{\hat{\theta}(p, n_r,D)}(\bX^{(r)}_j) = 0$. Recalling that $\mathscr{F}(u, \bx) = mu''(\bx) + \gamma u'(\bx)$, we have
$\|\mathscr{F}(u, \bx)\|_2\leqslant m \|u''(\bx)\|_2 + \gamma \|u'(\bx)\|_2$. We therefore conclude that $\lim_{p\to\infty} R_{n, n_r}(u_{\hat{\theta}(p, n_r,D)}) = 0$, which is the first statement of the proposition.

Next, using the Cauchy-Schwarz inequality, we have that, for any function $f\in C^2(\mathbb{R})$ and any $\varepsilon >0$,
\[
2\varepsilon\int_{-\varepsilon}^\varepsilon (m f''+ \gamma f')^2  \geqslant \Big(\int_{-\varepsilon}^\varepsilon mf'' + \gamma f'\Big)^2= \big[m (f'(\varepsilon)-f'(-\varepsilon)) + \gamma (f(\varepsilon)-f(-\varepsilon))\big]^2. 
\]
Thus,
\begin{align*}
&\mathscr{R}_n(u_{\hat{\theta}(p, n_r,D)}) \\
&\quad \geqslant \frac{1}{T} \int_{[0,T]} \mathscr{F}(u_{\hat{\theta}(p, n_r,D)}, \bx)^2 d\bx\\
 &\quad \geqslant \frac{1}{T} \sum_{i=1}^{n}\int_{\bX_{(i)}+\delta(n, n_r)/2-\varepsilon}^{\bX_{(i)}+\delta(n, n_r)/2+\varepsilon } \mathscr{F}(u_{\hat{\theta}(p, n_r,D)}, \bx)^2 d\bx\\
 &\quad \geqslant  \frac{1}{T}\sum_{i=1}^{n} \frac{1}{2 \varepsilon}\big[m(u'_{\hat{\theta}(p, n_r,D)}(\bX_{(i)}+\delta(n, n_r)/2+\varepsilon)-u'_{\hat{\theta}(p, n_r,D)}(\bX_{(i)}+\delta(n, n_r)/2-\varepsilon))\\
 &\qquad \quad +\gamma(u_{\hat{\theta}(p, n_r,D)}(\bX_{(i)}+\delta(n, n_r)/2+\varepsilon)-u_{\hat{\theta}(p, n_r,D)}(\bX_{(i)}+\delta(n, n_r)/2-\varepsilon))\big]^2.
\end{align*}
Observe that, as soon as $\delta(n,n_r)/4 > \varepsilon$, one has, for all $1 \leqslant i\leqslant n-1$,
\[\lim_{p \to \infty}u_{\hat{\theta}(p, n_r,D)}(\bX_{(i)}+\delta(n, n_r)/2+\varepsilon ) - u_{\hat{\theta}(p, n_r,D)}(\bX_{(i)}+\delta(n, n_r)/2 -  \varepsilon ) = Y_{(i+1)}-Y_{(i)},\] and, for all $1 \leqslant i\leqslant n-1$,
\[\lim_{p\to\infty}u'_{\hat{\theta}(p, n_r,D)}(\bX_{(i)}+\delta(n, n_r)/2+\varepsilon ) - u'_{\hat{\theta}(p, n_r,D)}(\bX_{(i)}+\delta(n, n_r)/2 -  \varepsilon ) = 0.\] 
Hence, for any $ 0<\varepsilon<\delta(n,n_r)/4$,
\begin{align*}
    &\sum_{i=1}^{n} \frac{1}{2 \varepsilon}\big[m(u'_{\hat{\theta}(p, n_r,D)}(\bX_{(i)}+\delta(n, n_r)/2-\varepsilon)-u'_{\hat{\theta}(p, n_r,D)}(\bX_{(i)}+\delta(n, n_r)/2-\varepsilon))\\
    &\qquad +\gamma(u_{\hat{\theta}(p, n_r,D)}(\bX_{(i)}+\delta(n, n_r)/2-\varepsilon)-u_{\hat{\theta}(p, n_r,D)}(\bX_{(i)}+\delta(n, n_r)/2-\varepsilon))\big]
    ^2 \\
&\xrightarrow[p \rightarrow \infty]{} \gamma \times \frac{\sum_{i=1}^{n-1} (Y_{(i+1)}-Y_{(i)})^2}{2 \varepsilon}.
\end{align*}
We have just proved that, for any $ 0<\varepsilon<\delta(n,n_r)/4$, there exists $P \in \mathbb{N}$ such that, for all  $p \geqslant P$, 
\[\mathcal{R}_n(u_{\hat{\theta}(p, n_r,D)}) \geqslant  \gamma \times \frac{\sum_{i=1}^{n-1} (Y_{(i+1)}-Y_{(i)})^2}{2 \varepsilon T}.
\]
We conclude as desired that $\lim_{p\to\infty}\mathcal{R}_n(u_{\hat{\theta}(p, n_r,D)})  =\infty$, since we suppose that there exists two observations $Y_{(i)} \neq Y_{(j)}$.

\subsection*{Proof of Proposition \ref{prop:wave}}
\label{proof:PDE_failure}
Let $u_{\hat{\theta}(p,n_e, n_r,D)} \in \text{NN}_H(4)$ be the neural network defined by
\begin{align*}
    u_{\hat{\theta}(p,n_e, n_r,D)}(x, t) =& \tanh^{\circ H}(x+0.5+p t)-\tanh^{\circ H}(x-0.5+pt) \\
    &+\tanh^{\circ H}(0.5+pt) - \tanh^{\circ H}(1.5+pt).
\end{align*}
Clearly, for any $p\in \mathbb{N}$, $u_{\hat{\theta}(p,n_e, n_r,D)}$ satisfies the initial condition
\[u_{\hat{\theta}(p,n_e, n_r,D)}(x, 0) = \tanh^{\circ H}(x+0.5)-\tanh^{\circ H}(x-0.5) + \tanh^{\circ H}(0.5) - \tanh^{\circ H}(1.5).\]
We are going to prove in the next paragraphs that the derivatives of $u_{\hat{\theta}(p,n_e, n_r,D)}$ vanish as $p\to \infty$, starting with the temporal derivative and continuing with the spatial ones.
According to Lemma \ref{lem:compTanh}, for all $\varepsilon > 0$ and all $x \in [-1, 1]$, 
$\lim_{p\to\infty}\|u_{\hat{\theta}(p,n_e, n_r,D)}(x, \cdot)\|_{C^2([\varepsilon, T])} = 0$.
Therefore, for any $\bX^{(e)}_i\in \{-1, 1\}\times [0,T]$,
$\lim_{p\to\infty}\|u_{\hat{\theta}(p,n_e, n_r,D)}(\bX^{(e)}_i)\|_2 = 0$ and, 
for any  $\bX^{(r)}_j\in \Omega$,
$\lim_{p\to\infty}\|\partial_{t}u_{\hat{\theta}(p,n_e, n_r,D)}(\bX^{(r)}_j)\|_2 = 0$ (since $\bX^{(r)}_j \notin \partial \Omega$).

Letting $v(x,t) = \tanh^{\circ H}(x+0.5+p t)-\tanh^{\circ H}(x-0.5+pt)$,
it comes that $\partial^2_{x,x} u_{\hat{\theta}(p,n_e, n_r,D)} = p^{-2}\partial^2_{t,t} v$.
Thus, invoking again Lemma \ref{lem:compTanh}, for all $\varepsilon > 0$, and all $x \in [-1, 1]$,
\[\lim_{p\to\infty}p^{-2}\|\partial^2_{t,t} v(x, \cdot)\|_{\infty, [\varepsilon, T]} =  \lim_{p\to\infty}\|\partial^2_{x,x} u_{\hat{\theta}(p,n_e, n_r,D)}(x, \cdot)\|_{\infty, [\varepsilon, T]}  = 0.\]
Therefore, for any $\bX^{(r)}_j\in \Omega$, one has $\lim_{p\to\infty}\|\partial^2_{x,x}u_{\hat{\theta}(p,n_e, n_r,D)}(\bX^{(r)}_j)\|_2 = 0$ and, in turn, one has
$\lim_{p\to\infty}\|\mathscr{F}(u_{\hat{\theta}(p,n_e, n_r,D)}, \bX^{(r)}_j)\|_2 = 0$. Thus, for all $n_e, n_r \geqslant 0$,
$\lim_{p\to\infty}R_{n_e,n_r}(u_{\hat{\theta}(p,n_e, n_r,D)}) = 0$.

Next, observe that $\mathscr{R}(u_{\hat{\theta}(p,n_e, n_r,D)}) \geqslant \int_{[-1,1]\times[0,T]} (\partial_{t}u_{\hat{\theta}(p,n_e, n_r,D)}-\partial^2_{x,x}u_{\hat{\theta}(p,n_e, n_r,D)})^2$. 
By the Cauchy-Schwarz inequality, for any $\delta >0$, 
\begin{align*}
   & \int_{[-1,1]\times[0,T]} (\partial_{t}u_{\hat{\theta}(p,n_e, n_r,D)}-\partial^2_{x,x}u_{\hat{\theta}(p,n_e, n_r,D)})^2\\
   & \geqslant \delta^{-1}\int_{x =-1}^{1}\Big(\int_{t=0}^\delta \partial_{t}u_{\hat{\theta}(p,n_e, n_r,D)}(x,t)-\partial^2_{x,x}u_{\hat{\theta}(p,n_e, n_r,D)}(x,t)\Big)^2dx\\
    & \geqslant \delta^{-1}\int_{x =-1}^{1}\Big(u_{\hat{\theta}(p,n_e, n_r,D)}(x,\delta) - u_{\hat{\theta}(p,n_e, n_r,D)}(x,0) -  \int_{t=0}^\delta \partial^2_{x, x}u_{\hat{\theta}(p,n_e, n_r,D)}(x,t)dt\Big)^2dx.
\end{align*}
Invoking again Lemma \ref{lem:compTanh}, we know that $\lim_{p\to \infty}\|u_{\hat{\theta}(p,n_e, n_r,D)}(\cdot,\delta)\|_{[-1, 1], \infty} = 0$. Moreover, for all $t >0$ and all $-1\leqslant x \leqslant 1$, $\lim_{p\to\infty}\partial^2_{x,x} u_{\hat{\theta}(p,n_e, n_r,D)}(x, t)  = 0$. Besides, by Corollary 
\ref{cor:bounding_tanh}, $\|\partial^2_{x,x} u_{\hat{\theta}(p,n_e, n_r,D)}\|_{\infty, [0,1]\times [-1,1]} \leqslant 2\|\tanh^{\circ H}\|_{C^2(\mathbb{R})} < \infty$.
Thus, by the dominated convergence theorem, for any $\delta > 0$ and all $p$ large enough, 
\[\mathscr{R}(u_{\hat{\theta}(p,n_e, n_r,D)}) \geqslant  \frac{1}{2\delta}\int_{x =-1}^{1}\big(u_{\hat{\theta}(p,n_e, n_r,D)}(x,0)\big)^2dx.\]
Noticing that $u_{\hat{\theta}(p,n_e, n_r,D)}(x,0)$ corresponds to the initial condition, that does not depends on $p$, we conclude that $\lim_{p\to\infty}\mathscr{R}(u_{\hat{\theta}(p,n_e, n_r,D)}) = \infty$.

\section{Proofs of Section \ref{sec:consistency}}
\label{app:sec5}
\subsection*{Proof of Proposition \ref{prop:bounding}}
Recall that each neural network $u_\theta \in \mathrm{NN}_H(D)$ is written as $u_\theta = \mathcal{A}_{H+1} \circ (\tanh \circ \mathcal{A}_{H}) \circ \cdots \circ  (\tanh \circ \mathcal{A}_1)$,
where each $\mathcal{A}_k : \mathbb{R}^{L_{k-1}} \rightarrow \mathbb{R}^{L_{k}}$ is an affine function of the form $\mathcal{A}_k(x) = W_k x + b_k$, with $W_k$ a ($L_{k-1} \times L_k$)-matrix, $b_k \in \mathbb{R}^{L_k}$ a vector,  $L_0 = d_1$, $L_1=\cdots=L_H=D$, $L_{H+1} = d_2$, and $\theta = (W_1, b_1, \hdots, W_{H+1}, b_{H+1})\in \mathbb{R}^{\sum_{i=0}^H (L_i+1)\times L_i}$. For each $i\in\{1, \hdots, d_1\}$, we let $\pi_i$ be the projection operator on the $i$th coordinate, defined by $\pi_i(x_1, \hdots, x_{d_1}) = x_i$. Similarly, for a matrix $W = (W_{i,j})_{1\leqslant i \leqslant d_2, 1\leqslant j \leqslant d_1 }$, we let $\pi_{i,j}(W) = W_{i,j}$ and $\|W\|_\infty = \max_{1\leqslant i \leqslant d_2, 1\leqslant j \leqslant d_1 } |W_{i,j}|$. Note that $\|W_k\bx\|_\infty \leqslant L_{k-1} \|W_k\|_\infty \|\bx\|_\infty$.  Clearly,
$\max_{1\leqslant k \leqslant H+1}(\|W_k\|_\infty , \|b_k\|_\infty) \leqslant\|\theta\|_\infty \leqslant \|\theta\|_2$.
Finally, we recursively define  the constants $C_{K,H}$ for all $K\geqslant 0$ and all $H\geqslant 1$ by $C_{0,H} = 1$, $C_{K,1}= 2^{K-1}\times(K+2)!$, and 
\begin{equation}
    C_{K,H+1} = B_K 2^{K-1} (K+2)! \max_{\underset{i_1+2i_2+\cdots + Ki_K = K}{ i_1,\hdots, i_K \in \mathbb{N}}} \prod_{1\leqslant \ell \leqslant K}C_{\ell,H}, \label{eq:defC}
\end{equation}
where $B_K$ is the $K$th Bell number, defined in \eqref{eq:bell}.

We prove the proposition by induction  on $H$, starting with the case $H=1$. 
Clearly, for $H=1$, one has
\begin{equation}
    \label{eq:iniBounding}
    \|u_\theta\|_\infty \leqslant \|W_2 \times \tanh \circ \mathcal{A}_1\|_\infty + \|b_2\|_\infty \leqslant \|W_2\|_\infty D  + \|b_2\|_\infty \leqslant (D+1)  \|\theta\|_2.
\end{equation}
Next, for any multi-index $\alpha = (\alpha_1, \hdots, \alpha_{d_1})$ such that $|\alpha|\geqslant 1$,
\begin{equation}
    \label{eq:boundingNNIni}
    \partial^\alpha u_\theta(\bx) = W_2 \begin{pmatrix}
    \pi_{1,1}(W_1)^{\alpha_1} \times \cdots \times \pi_{1,d_1}(W_1)^{\alpha_{d_1}} \times \tanh^{(|\alpha|)}(\pi_1(\mathcal{A}_1(\bx)))\\
    \vdots\\
    \pi_{1,d_1}(W_1)^{\alpha_1} \times \cdots \times \pi_{d_1,d_1}(W_1)^{\alpha_{d_1}} \times \tanh^{(|\alpha|)}(\pi_{d_1}(\mathcal{A}_1(\bx)))
\end{pmatrix}.
\end{equation}
Upon noting that $|\pi_{1,d_1}(W_1)|\leqslant \|\theta\|_\infty$, we see that 
\begin{equation}
\label{eq:iniBoundingK}
    \|\partial^\alpha u_\theta\|_\infty \leqslant D  \|W_2\|_\infty  \|\theta\|_2^{|\alpha|}  \|\tanh^{(|\alpha|)}\|_\infty \leqslant D  \|\theta\|_2^{1+|\alpha|}  \|\tanh^{(|\alpha|)}\|_\infty.
\end{equation}
Therefore, combining \eqref{eq:iniBounding} and \eqref{eq:iniBoundingK},  we deduce that for any $K \geqslant 1$, $\|u_\theta\|_{C^K(\mathbb{R}^{d_1})} \leqslant (D+1) \max_{k \leq K}\|\tanh^{(k)}\|_\infty  (1+\|\theta\|_2)^{K} \|\theta\|_2$.
Applying Lemma \ref{lem:derTanh}, we conclude that, for all $u \in \mathrm{NN}_1(D)$ and for all $K \geqslant 0$,
\[\|u_\theta\|_{C^K(\mathbb{R}^{d_1})} \leqslant  C_{K,1} (D+1)  (1+\|\theta\|_2)^{K} \|\theta\|_2.\]
\paragraph{Induction} Assume that for a given $H \geqslant 1$, one has, for any neural network $u_\theta \in \mathrm{NN}_H(D)$ and any $K \geqslant 0$,
\begin{equation}
\label{eq:recBoundingNN}
    \|u_\theta\|_{C^K(\mathbb{R}^{d_1})} \leqslant C_{K,H} (D+1)^{1+KH}  (1+\|\theta\|_2)^{KH}\|\theta\|_2.
\end{equation}
Our objective is to show that for any $u_\theta \in \mathrm{NN}_{H+1}(D)$ and any $K \geqslant 0$,
\[\|u_\theta\|_{C^K(\mathbb{R}^{d_1})} \leqslant C_{K,H+1} (D+1)^{1+K(H+1)}  (1+\|\theta\|_2)^{K(H+1)} \|\theta\|_2.\] 
For such a $u_\theta$, we have, by definition, 
$u_\theta = \mathcal{A}_{H+2}\circ \tanh \circ v_\theta$, where $v_\theta \in \mathrm{NN}_{H}(D)$ (by a slight abuse of notation, the parameter of $v_\theta$ is in fact $\theta' = (W_1, b_1, \hdots, W_{H+1}, b_{H+1})$ while $\theta = (W_1, b_1, \hdots,$ $W_{H+2}, b_{H+2})$, so $\|\theta'\|_2 \leqslant \|\theta\|_2$ and $\|\theta'\|_\infty \leqslant \|\theta\|_\infty$).
Consequently, 
\begin{equation}
    \|u_\theta\|_\infty \leqslant \|W_{H+2}\|_\infty  D + \|b_{H+2}\|_\infty \leqslant (D+1) \|\theta\|_2.\label{eq:boundNN}
\end{equation}
In addition, for any multi-index $\alpha = (\alpha_1, \hdots, \alpha_{d_1})$ such that $|\alpha|\geqslant 1$,
\[
\partial^\alpha u_\theta(\bx) = W_{H+2} \begin{pmatrix}
    \partial^\alpha ( \tanh\circ \pi_1 \circ v_\theta(\bx))\\
    \vdots\\
    \partial^\alpha ( \tanh\circ \pi_{D}\circ v_\theta(\bx))
\end{pmatrix}.
\]
Thus,
$\|\partial^\alpha u_\theta\|_\infty \leqslant D \|W_{H+2}\|_\infty  \max_{j \leqslant D}\|\tanh \circ \pi_j \circ v_\theta\|_{C^K(\mathbb{R}^{d_1})}$. Invoking identity \eqref{eq:fdbFormula}, one has
\[\|\tanh \circ \pi_j\circ v\|_{C^K(\mathbb{R}^{d_1})} \leqslant B_K \|\tanh \|_{C^K(\mathbb{R})} \max_{i_1+2i_2+\cdots + Ki_{K}=K}\prod_{1\leqslant \ell \leqslant K}\|\pi_j \circ v_\theta\|_{C^\ell(\mathbb{R}^{d_1})}^{i_\ell}.\] 
Observing that $\pi_{j}\circ v_\theta $ belongs to $\mathrm{NN}_H(D)$, Lemma \ref{lem:derTanh} and inequality \eqref{eq:recBoundingNN} show that
\[\|\tanh \circ \pi_j\circ v_\theta\|_{C^\ell(\mathbb{R}^{d_1})} \leqslant C_{\ell,H+1} (D+1)^{1+\ell H}  (1+\|\theta\|_2)^{1+\ell H} \|\theta\|_2.\] 
Therefore, $\|\partial^\alpha u_\theta\|_\infty \leqslant C_{K,H+1} (D+1)^{1+KH}  (1+\|\theta\|_2)^{K(H+1)}  \|\theta\|_2$, which concludes the induction.

To complete the proof, it remains to show that the exponent of $\|\theta\|_2$ is optimal. To this aim, we let $d_1 = d_2 = 1$, $D=1$. For each $H \geqslant 1$, we consider the sequence $(\theta^{(H)}_m)_{m \in \mathbb{N}}$ defined by  
$\theta^{(H)}_m = (W_1^{(m)}, b_1^{(m)}, \hdots, W^{(m)}_{H+1}, b^{(m)}_{H+1})$, with $W_i^{m} = m$ and $b_i^m = 0$. Then, for all $\theta = (W_1, b_1, \hdots,$ $ W_{H+1}, b_{H+1}) \in \Theta_{H, 1}$, the associated neural network's derivatives satisfy
\[\|u_\theta^{(k)}\|_\infty =
\|(\tanh^{\circ H})^{(K)}\|_{\infty}  |W_{H+1}| \prod_{i=1}^{H} |W_i|^K. \]
Next, since 
$\|\theta^{(H)}_m\|_2 = m\sqrt{H+1}$, we have
\begin{equation*}
\|u_{\theta^{(H)}_m}\|_{C^K(\mathbb{R}^{d_1})} 
\geqslant \big\|u_{\theta^{(H)}_m}^{(K)}\big\|_\infty \geqslant \big\|(\tanh^{\circ H})^{(K)}\big\|_\infty   m^{1+HK}
\geqslant \bar{C}(H, K)   \|\theta^{(H)}_m\|_2^{1+HK},
\end{equation*} 
where $\bar{C}(H, K) = (H+1)^{-(1+HK)/2} \|(\tanh^{\circ H})^{(K)}\|_\infty $. Since $\lim_{m\to\infty}\|\theta^{(H)}_m\|_2 = \infty $, we conclude that the bound of inequality \eqref{eq:recBoundingNN} is tight.
\subsection*{Lipschitz dependence of the Hölder norm in the NN parameters}

\begin{prop}[Lipschitz dependence of the Hölder norm in the NN parameters]
\label{prop:lipschitzParam}
Consider the class $\mathrm{NN}_H(D)=\{u_\theta, \theta\in\Theta_{H,D}\}$.
Let $K \in \mathbb{N}$. Then there exists a constant $\tilde{C}_{K,H}>0$, depending only on $K$ and $H$, such that, for all $\theta, \theta' \in \Theta_{H, D}$, 
\[
\|u_\theta-u_{\theta'}\|_{C^K(\Omega)} \leqslant \tilde{C}_{K,H}  (1+d_1M(\Omega)) (D+1)^{H+KH^2}  (1+\|\theta\|_2)^{H+KH^2}  \|\theta-\theta'\|_2,
\]
where $M(\Omega) = \sup_{\bx\in\Omega} \|\bx\|_\infty$.
\end{prop}

\begin{proof}
    We recursively define the constants $\tilde{C}_{K,H}$ for all $K\geqslant 0$ and all $H\geqslant 1$ by $\tilde{C}_{K,1} = (K+2) 2^{2K-1}  (K+2)! (K+3)!$, and \[\tilde{C}_{K,H+1} = C_{K,H+1} [1 + (K+1) B_K 2^{2K-1} (K+3)!(K+2)! \tilde{C}_{K,H}].\]
    Recall that $\pi_i$ is the projection operator on the $i$th coordinate, defined by $\pi_i(x_1, \hdots, x_{d_1}) = x_i$.
Before embarking on the proof, observe that by identity \eqref{eq:fdbFormula}, we have, for all $u_1, u_2 \in C^K(\Omega, \mathbb{R}^{D})$, 
for all $1 \leqslant i \leqslant D,$
\begin{align*}
    \partial^\alpha (\tanh\circ \pi_i \circ u_1-\tanh\circ \pi_i \circ u_2) &= \sum_{P\in \Pi(K)}[\tanh^{(|P|)}\circ \pi_i \circ u_1]  \prod_{S \in P} \partial^{\alpha(S)}(\pi_i \circ u_1) \\
    &\quad - [\tanh^{(|P|)}\circ \pi_i \circ u_2]  \prod_{S \in P} \partial^{\alpha(S)} (\pi_i \circ u_2).
\end{align*}
In addition, for two sequences $(a_i)_{1\leqslant i \leqslant n}$ and $(b_i)_{1\leqslant i \leqslant n}$,
\begin{equation}
    \prod_{i=1}^n a_i - \prod_{i=1}^n b_i = \sum_{i=1}^n (a_i-b_i) \Big(\prod_{j=i+1}^n a_j\Big)\Big(\prod_{j=1}^{i-1}b_j\Big) \leqslant n \max_{1\leqslant i\leqslant n}\{|a_i-b_i|\} \prod_{i=1}^n \max(|a_i|, |b_i|). \label{eq:diffProd}
\end{equation}  
 Observe that for any $1 \leqslant i\leqslant d_2$ and $P \in \Pi(K)$, the term  $[\tanh^{(|P|)}\circ \pi_i \circ u_1]  \prod_{S \in P} \partial^{\alpha(S)}(\pi_i \circ u_1) - [\tanh^{(|P|)}\circ \pi_i \circ u_2]  \prod_{S \in P} \partial^{\alpha(S)} (\pi_i \circ u_2)$ is the difference of two products of $|P| +1$ terms to which we can apply \eqref{eq:diffProd}. So,
\begin{align}
    &\Big\|[\tanh^{(|\pi|)}\circ \pi_i \circ u_1] \prod_{S \in P} \partial^{\alpha(S)}(\pi_i \circ u_1) - [\tanh^{(|\pi|)}\circ \pi_i \circ u_2]  \prod_{S \in \pi} \partial^{\alpha(S)}(\pi_i \circ u_2)\Big\|_{\infty, \Omega}\nonumber\\
    &\quad \leqslant (|P|+1)  \big(\|\tanh^{(|P|)}\|_{\mathrm{Lip}} \|u_1-u_2\|_{\infty, \Omega} + \|u_1-u_2\|_{C^K(\Omega)}\big) \nonumber\\
    &\qquad \times \|\tanh^{(|P|)}\|_{\infty} \prod_{S \in P} \max(\|\partial^{\alpha(S)}u_1\|_{\infty, \Omega},\|\partial^{\alpha(S)}u_2\|_{\infty, \Omega}).\label{eq:fundRes}
\end{align}
Notice finally that $\|\tanh^{(|P|)}\|_{\mathrm{Lip}} = \|\tanh^{(|P|+1)}\|_{\infty}$.

With the preliminary results out of the way, we are now equipped to prove the statement of the proposition, by induction on $H$. Assume first that $H=1$. We start by examining the case $K= 0$ and then generalize to all $K \geqslant 1$. Let $u_\theta = \mathcal{A}_2 \circ \tanh \circ \mathcal{A}_1$ and $u_{\theta'} = \mathcal{A}'_2 \circ \tanh \circ \mathcal{A}'_1$. Notice that \[\|\mathcal{A}_1-\mathcal{A}'_1\|_{\infty, \Omega} \leqslant \|b_1-b'_1\|_\infty + d_1 M(\Omega) \|W_1-W_1'\|_\infty \leqslant \|\theta-\theta'\|_2 (1+d_1M(\Omega)),\]
where $M(\Omega)= \max_{\bx \in \Omega}\|\bx\|_\infty$. 
Since $\|\tanh\|_{\mathrm{Lip}}=1$, we deduce that $\|\tanh \circ \mathcal{A}_1-\tanh \circ \mathcal{A}'_1\|_\infty \leqslant \|\theta-\theta'\|_2 (1+d_1M(\Omega))$. Similarly, $\|\mathcal{A}_2-\mathcal{A}'_2\|_{\infty,  B(1, \|\cdot\|_\infty)} \leqslant \|\theta-\theta'\|_2 (1+D)$. Next, 
\begin{align*}
    \|u_\theta-u_{\theta'}\|_{\infty, \Omega} &\leqslant \|(\mathcal{A}_2-\mathcal{A}_2')\circ \tanh \circ \mathcal{A}_1\|_{\infty, \Omega} + \|\mathcal{A}_2' \circ \tanh \circ \mathcal{A}_1 - \mathcal{A}_2'\circ \tanh\circ \mathcal{A}_1')\|_{\infty, \Omega}\\
    & \leqslant \|\mathcal{A}_2-\mathcal{A}_2'\|_{\infty, B(1, \|\cdot\|_\infty)} + D \|W_2'\|_\infty \|\tanh \circ \mathcal{A}_1 - \tanh\circ \mathcal{A}_1'\|_{\infty, \Omega}\\
    & \leqslant \|\theta-\theta'\|_2  (1+D+ D  \|\theta'\|_2  (1+d_1M(\Omega)))\\
    &\leqslant \tilde{C}_{0,1}  (1+d_1M(\Omega)) (D+1) (1+\max(\|\theta\|_2,\|\theta'\|_2)) \|\theta-\theta'\|_2.
\end{align*}
This shows the result for $H=1$ and $K=0$. Assume now that $K\geqslant 1$, and let $\alpha$ be a multi-index such that $|\alpha| = K$. Observe that
\begin{align}
    \|\partial^\alpha(u_\theta-u_{\theta'})\|_{\infty, \Omega} &\leqslant \|(W_2-W_2') \partial^\alpha( \tanh \circ \mathcal{A}_1)\|_{\infty, \Omega} \nonumber\\
    &\quad + \|W_2' \partial^\alpha (\tanh \circ \mathcal{A}_1 - \tanh\circ \mathcal{A}_1')\|_{\infty, \Omega}.\label{eq:trig}
\end{align}
By Lemma \ref{lem:derTanh} and an argument similar to the inequality \eqref{eq:boundingNNIni}, we have
\begin{align}
    \|(W_2-W_2') \partial^\alpha( \tanh \circ \mathcal{A}_1)\|_{\infty, \Omega} &\leqslant (D+1)  \|\theta-\theta'\|_2  \|\theta\|_2^K  \|\tanh\|_{C^K(\mathbb{R})}\nonumber\\
    &\leqslant  2^{K-1}(K+2)!  (D+1)  \|\theta-\theta'\|_2  \|\theta\|_2^K\label{eq:trigPart1}.
\end{align}
 In order to bound the second term on the right-hand side of \eqref{eq:trig}, we use inequality \eqref{eq:fundRes} with $u_1 =\mathcal{A}_1$ and $u_2 =\mathcal{A}_1'$. In this case,  the only non-zero term on the right-hand side of \eqref{eq:fundRes} corresponds to the partition $\pi  = \{\{1\}, \{2\}, \hdots, \{K\}\}$. Recall that $\|\mathcal{A}_1-\mathcal{A}_1'\|_{\infty, \Omega} \leqslant \|\theta-\theta'\|_2 (1+d_1M(\Omega))$, and note that whenever $|\alpha| = 1$, $\|\partial^\alpha(\mathcal{A}_1-\mathcal{A}_1')\|_{\infty, \Omega} \leqslant \|\theta-\theta'\|_2$.  
Therefore, $\|\mathcal{A}_1-\mathcal{A}_1'\|_{C^K(\Omega)} = \|\mathcal{A}_1-\mathcal{A}_1'\|_{C^1(\Omega)} \leqslant  \|\theta-\theta'\|_2  (1+d_1M(\Omega))$. Observe that $\prod_{B \in \{\{1\}, \{2\}, \hdots, \{K\}\}} \max(\|\partial^{\alpha(B)}\mathcal{A}_1\|_{\infty, \Omega},\|\partial^{\alpha(B)}\mathcal{A}_1'\|_{\infty, \Omega}) \leqslant \max(\|\theta\|_2, \|\theta'\|_2)^{K}.$ Thus, putting all the pieces together, we are led to \begin{align*}
    &\|\partial^\alpha (\tanh\circ \mathcal{A}_1-\tanh\circ \mathcal{A}_1')\|_{\infty, \Omega}\\
    &\quad \leqslant (K+1) \|\tanh^{(K+1)}\|_{\infty} \|\theta-\theta'\|_2 (1+d_1 M(\Omega))  \|\tanh^{(K)}\|_{\infty}    \max(\|\theta\|_2, \|\theta'\|_2)^{K}.
\end{align*}
Now, by Lemma \ref{lem:derTanh}, $\|\tanh^{(K)}\|_{\infty} \leqslant 2^{K-1}(K+2)!$ So,
\begin{align}
    &\|\partial^\alpha (\tanh\circ \mathcal{A}_1-\tanh\circ \mathcal{A}_1')\|_{\infty, \Omega}\nonumber\\
    &\quad \leqslant (K+1)2^{2K-1}(K+2)!(K+3)!  \|\theta-\theta'\|_2 (1+d_1 M(\Omega)) \max(\|\theta\|_2, \|\theta'\|_2)^{K} \label{eq:trigPart2}.
\end{align}
Combining inequalities \eqref{eq:trig}, \eqref{eq:trigPart1}, and \eqref{eq:trigPart2}, we conclude that
\[\|\partial^\alpha(u_\theta-u_{\theta'})\|_{\infty, \Omega} \leqslant \tilde{C}_{K,1}  (1+d_1M(\Omega)) (D+1)(1+\max(\|\theta\|_2, \|\theta'\|_2))^{K+1} \|\theta-\theta'\|_2, \]
so that 
$\|u_\theta-u_{\theta'}\|_{C^K(\Omega)} \leqslant \tilde{C}_{K,1}  (1+d_1M(\Omega)) (D+1)(1+\max(\|\theta\|_2, \|\theta'\|_2))^{K+1} \|\theta-\theta'\|_2$.

\paragraph{Induction} Fix $H\geqslant 1$, and assume that for all $u_\theta, u_{\theta'} \in \mathrm{NN}_H(D)$ and all $K \geqslant 0$,
\begin{align}
    &\|u_\theta-u_{\theta'}\|_{C^K(\Omega)} \nonumber \\
    &\quad \leqslant \tilde{C}_{K,H}  (1+d_1M(\Omega)) (D+1)^{H+KH^2}(1+\max(\|\theta\|_2, \|\theta'\|_2))^{H+KH^2} \|\theta-\theta'\|_2. \label{eq:recLip}
\end{align}
Let $u_\theta, u_{\theta'} \in \mathrm{NN}_{H+1}(D)$. Observe that $u_\theta = \mathcal{A}_{H+2}\circ\tanh \circ v_\theta$ and $u_{\theta'} = \mathcal{A}_{H+2}'\circ\tanh \circ v_{\theta'}$, where  $v_\theta, v_{\theta'} \in \mathrm{NN}_H(D)$.
Moreover, 
\begin{align}
    &\|\partial^\alpha(u_\theta-u_{\theta'})\|_{\infty, \Omega} \nonumber \\
    &\quad \leqslant \|(W_{H+2}-W_{H+2}') \partial^\alpha( \tanh \circ v_\theta)\|_{\infty, \Omega} + \|W_{H+2}' \partial^\alpha (\tanh \circ v_\theta - \tanh\circ v_{\theta'})\|_{\infty, \Omega} \nonumber \\
    &\quad \leqslant D(\|\theta-\theta'\|_2 \times \|\partial^\alpha( \tanh \circ v_\theta)\|_{\infty, \Omega} + \|\theta'\|_2\times\|\partial^\alpha (\tanh \circ v_\theta - \tanh\circ v_{\theta'})\|_{\infty, \Omega}). \label{eq:lipTrig}
\end{align}
Since $\tanh\circ v_\theta \in \mathrm{NN}_{H+1}(D)$, we have, by Proposition \ref{prop:bounding},
\begin{equation}
    \|\partial^\alpha( \tanh \circ v_\theta)\|_{\infty, \Omega} \leqslant C_{K,H+1}(D+1)^{1+K(H+1)}(1+\|\theta\|_2)^{K(H+1)}\|\theta\|_2. \label{eq:boundTrig1}
\end{equation} 
Moreover, using \eqref{eq:fundRes}, Lemma \ref{lem:derTanh}, and the definition of $C_{K,H+1}$ in \eqref{eq:defC}, we have 
\begin{align}
      &\|\partial^\alpha (\tanh \circ v_\theta - \tanh\circ v_{\theta'})\|_{\infty, \Omega}
      \nonumber\\
      & \quad \leqslant B_K(K+1) \|\tanh^{(K+1)}\|_{\infty} \|v_\theta-v_{\theta'}\|_{C^K(\Omega)} \|\tanh^{(K)}\|_{\infty} \nonumber \\
      &\qquad \times C_{K,H+1} (D+1)^{KH}  (1+\max(\|\theta\|_2, \|\theta'\|_2))^{KH} \nonumber \\
      &\quad \leqslant 2^{2K-1} (K+3)!(K+2)! B_K(K+1) \|v_\theta-v_{\theta'}\|_{C^K(\Omega)} \nonumber \\
      &\qquad \times C_{K,H+1} (D+1)^{KH}  (1+\max(\|\theta\|_2, \|\theta'\|_2))^{KH}. \label{eq:boundingTrig2}
\end{align}
 The term $\|v_\theta-v_{\theta'}\|_{C^K(\Omega)}$ in \eqref{eq:boundingTrig2} can be upper bounded using the induction assumption \eqref{eq:recLip}. Thus, combining 
\eqref{eq:lipTrig}, \eqref{eq:boundTrig1}, and \eqref{eq:boundingTrig2}, we conclude as desired that for all $u_\theta, u_{\theta'} \in \mathrm{NN}_{H+1}(D)$ and all $K \in \mathbb{N}$,
\begin{align*}
\|u_\theta-u_{\theta'}\|_{C^K(\Omega)}&\leqslant \tilde{C}_{K,H+1}  (1+d_1M(\Omega)) (D+1)^{(H+1)+K(H+1)^2}\\
& \quad \times (1+\max(\|\theta\|_2, \|\theta'\|_2))^{(H+1)+K(H+1)^2} \|\theta-\theta'\|_2.
\end{align*}
\end{proof}

\subsection*{Uniform approximation of integrals}
\label{app:glivenko_cantelli}
Throughout this section, the parameters $H,D\in \mathbb{N}^\star$ are held fixed, as well as the neural architecture $\mathrm{NN}_H(D)$  parameterized by $\Theta_{H,D}$. We let $d$ be a metric in $\Theta_{H,D}$, and denote by $B(r, d)$ the closed ball in $\Theta_{H,D}$ centered at $0$ and of radius $r$ according to the metric $d$, that is, $B(r, d)= \{\theta \in \Theta_{H,D},\ d(0,\theta) \leqslant r\}$.
\begin{thm}[Uniform approximation of integrals]
\label{thm:approx_integral}
    Let $\Omega \subseteq \mathbb{R}^{d_1}$ be a bounded Lipschitz domain, let $\alpha_1 >0$, and let $\bX_1, \hdots, \bX_n$ be a sequence of i.i.d.~random variables in $\bar{\Omega}$, with distribution $\mu_{X}$. Let $f:C^\infty(\bar{\Omega}, \mathbb{R}^{d_2})\times\bar{\Omega}\to\mathbb{R}^{d_2}$ be an operator, and assume that the following two requirements are satisfied: 
    \begin{itemize}
        \item[$(i)$] there exist $C_1 >0$ and $\beta_1 \in [0, 1/2[$ such that, for all $n\geqslant 1$ and all $\theta, \theta'\in  B(n^{\alpha_1}, \|.\|_2)$,
    \begin{equation}
        \|f(u_\theta, \cdot)-f(u_{\theta'},\cdot)\|_{\infty, \bar{\Omega}} \leqslant C_1  n^{\beta_1}  \|\theta-\theta'\|_2;\label{eq:cdn1}
    \end{equation}
\item[$(ii)$] there exist $C_2 >0$ and $\beta_2 \in [0, 1/2[$ satisfying $\beta_2 > \alpha_1 + \beta_1$ such that, for all $n \geqslant 1$ and all $\theta \in B(n^{\alpha_1}, \|.\|_2)$,
    \begin{equation}
        \|f(u_\theta, \cdot)\|_{\infty, \bar{\Omega}} \leqslant C_2  n^{\beta_2}. \label{eq:cdn2}
    \end{equation}
    \end{itemize}
Then, almost surely, there exists $N \in \mathbb{N}^\star$ such that, for all $n\geqslant N$,
    \[\sup_{\theta\in B(n^{\alpha_1}, \|.\|_2)} \Big\|\frac{1}{n}\sum_{i=1}^n f(u_\theta, \bX_i) - \int_{\bar{\Omega}} f(u_\theta,\cdot) d\mu_{X}\Big\|_2 \leqslant \log^2(n) n^{\beta_2 - 1/2}.\]
    (Notice that the rank $N$ is random.)
\end{thm}
\begin{proof} Let us start the proof by considering the case $d_2=1$.
For a given $\theta \in B(n^{\alpha_1}, \|\cdot\|_2)$, we let 
\[
Z_{n,\theta} = \frac{1}{n}\sum_{i=1}^n f(u_\theta, \bX_i) - \int_{\bar{\Omega}} f(u_\theta, \cdot) d\mu_X.
\] 
We are interested in bounding the random variable 
\[
Z_n = \sup_{\theta \in B(n^{\alpha_1}, \|\cdot\|_2)} |Z_{n,\theta}| = \sup_{\theta \in B(n^{\alpha_1}, \|\cdot\|_2)} Z_{n,\theta}.
\] 
Note that there is no need of absolute value in the rightmost term since, for any $ \theta=(W_1, b_1,\hdots,$ $ W_{H+1}, b_{H+1}) \in B(n^{\alpha_1}, \|\cdot\|_2)$, it is clear that $\theta' = (W_1, b_1,\hdots, W_H, b_H$, $ -W_{H+1}, -b_{H+1}) \in B(n^{\alpha_1}, \|\cdot\|_2)$ and $ u_{\theta'} = -u_\theta$.
Let $M(\Omega) = \max_{\bx\in\bar{\Omega}} \|x\|_2$.
Using inequality \eqref{eq:cdn1}, we have, for any $\theta, \theta' \in B(n^{\alpha_1}, \|\cdot\|_2)$, 
\[
\Big|\frac{1}{n}\Big(f(u_\theta, \bX_i)-\int_{\bar{\Omega}} f(u_\theta,\cdot)d\mu_X\Big) - \frac{1}{n}\Big(f(u_\theta', \bX_i)-\int_{\bar{\Omega}} f(u_\theta',\cdot)d\mu_X\Big) \Big| \leqslant 2C_1 n^{\beta_1 - 1}\|\theta-\theta'\|_2.
\] 
According to Hoeffding's theorem \citep[][Lemma 3.6]{vanhandel2016lectures},  the random variable $n^{-1}(f(u_\theta, \bX_i) $ $-\int_{\bar{\Omega}} f(u_\theta,\cdot)d\mu_X) - n^{-1}(f(u_\theta', \bX_i)-\int_{\bar{\Omega}} f(u_\theta',\cdot)d\mu_X)$ is subgaussian with parameter $4C_1^2 n^{2\beta_1-2} \|\theta-\theta'\|_2^2$. 
Invoking Azuma's theorem \citep[][Lemma 3.7]{vanhandel2016lectures}, we deduce that $Z_{n,\theta}-Z_{n,\theta'}$, is also subgaussian, with parameter $4C_1^2n^{2\beta_1-1}\|\theta-\theta'\|_2^2$. 
Since $\mathbb{E}(Z_{n,\theta}) = 0$, we conclude that for all $n\geqslant 1$, $(Z_{n,\theta})_{\theta\in B(n^{\alpha_1}, \|\cdot\|_2)}$ is a subgaussian process on $B(n^{\alpha_1}, \|\cdot\|_2)$ for the metric $d(\theta,\theta') = 2C_1n^{\beta_1-1/2}\|\theta-\theta'\|_2$.
Moreover, since $\theta \mapsto Z_{n,\theta}$ is continuous for the topology induced by the metric $d$, $(Z_{n,\theta})_{\theta \in B(n^{\alpha_1}, \|\cdot\|_2)}$ is separable \citep[][Remark 5.23]{vanhandel2016lectures}.
Thus, by Dudley's theorem \citep[][Corollary 5.25]{vanhandel2016lectures}
\[\mathbb{E}(Z_n) \leqslant 12 \int_0^\infty [\log N(B( n^{\alpha_1}, \|\cdot\|_2), d, r)]^{1/2}dr,\] where $N(B(n^{\alpha_1}, \|\cdot\|_2), d, r)$ is the minimum number of balls  of radius $r$ according to the metric $d$ needed to cover the space $B( n^{\alpha_1}, \|\cdot\|_2)$. 
Clearly,
$N(B( n^{\alpha_1}, \|\cdot\|_2),\ d,\ r) = N(B( n^{\alpha_1}, \|\cdot\|_2),\ \|\cdot\|_2,\ n^{1/2-\beta_1}r/(2C_1))$.
Thus,
\[\mathbb{E}(Z_n) \leqslant 24 C_1 n^{\beta_1-1/2} \int_0^\infty [\log N(B( n^{\alpha_1}, \|\cdot\|_2), \|\cdot\|_2, r)]^{1/2}dr\] 
and, in turn,
\[
\mathbb{E}(Z_n) \leqslant 24 C_1 n^{\alpha_1+\beta_1-1/2} \int_0^\infty [\log N(B(1, \|\cdot\|_2),\ \|\cdot\|_2, r)]^{1/2}dr.
\]
Upon noting that $N(B(1, \|\cdot\|_2),\ \|\cdot\|_2, r) = 1$ for $r\geqslant 1$, we are led to
\[
\mathbb{E}(Z_n) \leqslant 24 C_1 n^{\alpha_1+\beta_1-1/2} \int_0^1 [\log N(B(1, \|\cdot\|_2),\ \|\cdot\|_2, r)]^{1/2}dr.
\] 
 Since $\Theta_{H,D} = \mathbb{R}^{(d_1+1)D+(H-1)D(D+1)+(D+1)d_2}$, according to \citet[][Lemma 5.13]{vanhandel2016lectures}, one has
\[
\log N(B(1, \|\cdot\|_2),\ \|\cdot\|_2, r) \leqslant [(d_1+1)D+(H-1)D(D+1)+(D+1)d_2]\log(3/r).
\] 
Notice that $\int_0^1 \log(3/r)^{1/2}dr \leqslant 3/2$. Therefore,  
\begin{equation}
    \mathbb{E}(Z_n) \leqslant 36 C_1[(d_1+1)D+(H-1)D(D+1)+(D+1)d_2]^{1/2} n^{\alpha_1+\beta_1-1/2}. \label{eq:expectancyZn}
\end{equation}
Next, observe that, by definition of $Z_n=Z_n(\bX_1, \hdots, \bX_n)$, 
\begin{align*}
    &\sup_{\bx_i \in \mathbb{R}^{d_1}}Z_n(\bX_1, \hdots, \bX_{i-1}, \bx_i, \bX_{i+1}, \hdots, \bX_n) - \inf_{\bx_i \in \mathbb{R}^{d_1}}Z_n(\bX_1, \hdots, \bX_{i-1}, \bx_i, \bX_{i+1}, \hdots, \bX_n)\\ 
    & \quad \leqslant 2n^{-1}\sup_{\theta \in B(n^{\alpha_1}, \|\cdot\|_2)}\Big\|f(u_\theta, \bX_i) - \int_{\bar{\Omega}} f(u_\theta,\cdot) d\mu_X\Big\|_2\\
    & \quad \leqslant 4n^{-1}\sup_{\theta \in B(n^{\alpha_1}, \|\cdot\|_2)}\|f(u_\theta, \cdot)\|_\infty.
\end{align*} 
Using inequality \eqref{eq:cdn2}, McDiarmid's inequality  \citep[Theorem 3.11]{vanhandel2016lectures} ensures that $Z_n$ is subgaussian with parameter $4C_2^2n^{2\beta_2-1}$. In particular, for all $t_n \geqslant 0$, $\mathbb{P}(|Z_n - \mathbb{E}(Z_n)|\geqslant t_n) \leqslant 2 \exp(-n^{1-2\beta_2}t_n^2/(8C_2^2))$, which is summable with $t_n = C_3 n^{\beta_2-1/2}\log^2(n)$, where $C_3$ is any positive constant.
Thus,  recalling that $\beta_2 > \alpha_1+\beta_1$, the Borel-Cantelli lemma and \eqref{eq:expectancyZn} ensure that,
almost surely, for all $n$ large enough,
$0 \leqslant Z_n \leqslant 2 C_3 n^{\beta_2-1/2}\log^2(n)$. Taking $C_3 = 1/2$ yields the desired result.

The generalization to the case $d_2 \geqslant 2$ is easy. Just note, letting $f = (f_1, \hdots, f_{d_2})$, that 
\begin{align*}
    &\sup_{\theta \in B(n^{\alpha_1}, \|\cdot\|_2)} \Big\|\frac{1}{n}\sum_{i=1}^n f(u_\theta, \bX_i) - \int_{\bar{\Omega}} f(u_\theta, \cdot) d\mu_X\Big\|_2 \\
    &\qquad \leqslant \sqrt{d_2} \max_{1 \leqslant j\leqslant d_2}\sup_{\theta \in B(n^{\alpha_1}, \|\cdot\|_2)} \Big\|\frac{1}{n}\sum_{i=1}^n f_j(u_\theta, \bX_i) - \int_{\bar{\Omega}} f_j(u_\theta, \cdot) d\mu_X\Big\|_2 .
\end{align*}
Taking $C_3 = d_2^{-1/2}/2$ as above leads to the result.
\end{proof}
\begin{prop}[Condition function]
\label{prop:GKopH}
    Let $\Omega$ be a bounded Lipschitz domain, let $E$ be a closed subset of $\partial\Omega$, and let $h \in \mathrm{Lip}(E,\mathbb{R}^{d_2})$. Then the operator $\mathscr{H}(u, \bx) = \mathbf{1}_{\bx \in E} \|u(\bx) -h(\bx)\|^2$ satisfies inequalities \eqref{eq:cdn1} and \eqref{eq:cdn2} with $\alpha_1 < (3+H)^{-1}/2$, $\beta_1=(1+H)\alpha_1$, and  $1/2 > \beta_2 \geqslant (3+H)\alpha_1$.  
\end{prop}
\begin{proof} First note, since $\mathrm{Lip}(E,\mathbb{R}^{d_2}) \subseteq C^0(E, \mathbb{R}^{d_2})$, that $\|h\|_\infty < \infty$. Observe also that for any $v,w \in \mathbb{R}^{d_2}$, $|\|v\|_2^2-\|w\|_2^2|= |\langle v+w, v-w\rangle|\leqslant \|v+w\|_2\|v-w\|_2 \leqslant d_2 \|v+w\|_\infty\|v-w\|_\infty$, where $\langle\cdot, \cdot\rangle$ denotes the canonical scalar product. Thus, 
we obtain, for all $\theta, \theta' \in B(n^{\alpha_1}, \|\cdot\|_2)$ and all $\bx\in E$,
\begin{align*}
    |\mathscr{H}(u_\theta,\bx)-\mathscr{H}(u_{\theta'},\bx)| &\leqslant (\|u_\theta(\bx)\|_2 + \|u_{\theta'}(\bx)\|_2 + 2\|h(\bx)\|_2) \|u_\theta(\bx) - u_{\theta'}(\bx)\|_2\\
    &\leqslant d_2 (\|u_\theta\|_{\infty, \bar{\Omega}} + \|u_{\theta'}\|_{\infty, \bar{\Omega}} + 2\|h\|_\infty) \|u_\theta - u_{\theta'}\|_{\infty, \bar{\Omega}} \\
    &\leqslant d_2 (2(D+1)n^{\alpha_1} + 2\|h\|_\infty) \|u_\theta - u_{\theta'}\|_{\infty, \bar{\Omega}}  \;\; \mbox{(by inequality \eqref{eq:boundNN})} \\
    &\leqslant 2 d_2 ((D+1)n^{\alpha_1} + \|h\|_\infty)\tilde{C}_{0,H}  (1+d_1M(\Omega)) \\
    &\qquad \times (D+1)^{H}(1+n^{\alpha_1})^{H} \|\theta-\theta'\|_2 \quad \mbox{(by Proposition \ref{prop:lipschitzParam})}\\
    &\leqslant C_1 n^{\beta_1} \|\theta- \theta'\|_2,
\end{align*}
where $\beta_1 = (1+H)\alpha_1$ and 
$C_1 = 2^{H+1} d_2 (D+1+ \|h\|_\infty)\tilde{C}_{0,H}  (1+d_1M(\Omega)) (D+1)^{H}$. 

Next,  using \eqref{eq:boundNN} once again, for all $\theta \in B(n^{\alpha_1}, \|.\|_2)$, 
$\|\mathscr{H}(u_\theta, \cdot)\|_{\infty, \bar{\Omega}} \leqslant d_2(\|u_\theta\|_{\infty, \bar{\Omega}}+\|h\|_\infty)^2\leqslant d_2 ((D+1)n^{\alpha_1} + \|h\|_\infty)^2 \leqslant C_2 n^{2 \alpha_1}$. Recall that for inequality \eqref{eq:cdn2}, $\beta_2$ must satisfy $\alpha_1 + \beta_1 < \beta_2 < 1/2$. This is true for $\beta_2 = (3+H)\alpha_1$, which completes the proof.
\end{proof}

\begin{prop}[Polynomial operator]
\label{prop:GKopF}
    Let $\Omega$ be a bounded Lipschitz domain, and let $\mathscr{F} \in {\mathscr P}_{\mathrm{op}}$. Then the operator $\mathbf{1}_{\bx \in \Omega} \mathscr{F}(u_\theta, \bx)^2$ satisfies inequalities \eqref{eq:cdn1} and \eqref{eq:cdn2} with $\alpha_1 < [2+H(1+ (2+H)\deg(\mathscr{F}))]^{-1}/2$, $\beta_1 = H(1+ (2+H)\deg(\mathscr{F}))\alpha_1$, and $1/2 > \beta_2 \geqslant [2+H(1+ (2+H)\deg(\mathscr{F}))]\alpha_1$. 
\end{prop}
\begin{proof} 
Let $\mathscr{F} \in {\mathscr P}_{\mathrm{op}}$ be a polynomial operator. 
By definition, there exist a degree $s\geqslant 1$, a polynomial $P \in C^\infty(\mathbb{R}^{d_1}, \mathbb{R})[Z_{1,1},\hdots,$ $ Z_{d_2, s}]$, and a sequence $(\alpha_{i,j})_{1\leqslant i\leqslant d_2, 1\leqslant j\leqslant s}$ of multi-indices such that, for any $u\in C^\infty(\bar{\Omega},\mathbb{R}^{d_2})$, $\mathscr{F}(u,\cdot) = P((\partial^{\alpha_{i,j}}u_i)_{1\leqslant i\leqslant d_2, 1\leqslant j\leqslant s})$. 
Namely, there exist $N(P) \in \mathbb{N}^\star$, exponents $ I(i,j,k) \in \mathbb{N}$, and functions $\phi_1, \hdots, \phi_{N(P)} \in C^\infty(\bar{\Omega}, \mathbb{R})$, such that $P(Z_{1,1}, \hdots, Z_{d_2,s}) = \sum_{k=1}^{N(P)} \phi_k \times \prod_{i=1}^{d_2}\prod_{j=1}^{s} Z_{i,j}^{I(i,j,k)}$. 
Recall, by Definition \ref{defi:deg}, that $\deg(\mathscr{F})= \max_k \sum_{i=1}^{d_2}\sum_{j=1}^{s} (1+|\alpha_{i,j}|)I(i,j,k)$.

Now, according to Proposition \ref{prop:bounding}, there exists a positive constant $C_{\mathrm{deg}(\mathscr{F}),H}$ such that 
\begin{align*}
    &\|\mathscr{F}(u_\theta,\cdot)^2\|_{\infty,\bar{\Omega}} \\
    &\quad \leqslant \bigg[\sum_{k=1}^{N(P)} \|\phi_k\|_{\infty, \bar{\Omega}}  \prod_{i=1}^{d_2}\prod_{j=1}^{s} \|\partial^{\alpha_{i,j}}u_\theta\|_{\infty, \bar{\Omega}}^{I(i,j,k)}\bigg]^2\\
    & \quad \leqslant N^2(P)  \big[\max_{1\leqslant k \leqslant N(P)}\|\phi_k\|_{\infty, \bar{\Omega}}\big]^2  C_{\mathrm{deg}(\mathscr{F}),H}^2(D+1)^{2 H \deg(\mathscr{F})} (1+\|\theta\|_2)^{2 H\deg(\mathscr{F})}.
\end{align*}
Thus, for any $\theta\in B(n^{\alpha_1}, \|\cdot\|_2)$, $\|\mathscr{F}(u_\theta,\cdot)^2\|_{\infty,\bar{\Omega}} \leqslant C_2 n^{\beta_2}$,  where 
\[C_2 = 2^{2H\deg(\mathscr{F})} N^2(P)  \big[\max_{1\leqslant k \leqslant N(P)}\|\phi_k\|_{\infty, \bar{\Omega}}\big]^2  C_{\mathrm{deg}(\mathscr{F}),H}^2(D+1)^{2H\deg(\mathscr{F})}, 
\]
and for any $\beta_2 \geqslant 2H\deg(\mathscr{F}) \alpha_1$.

Next, observe that, any $u$ and $v$, $||u|^2-|v|^2|= |(u+v)(u-v)|\leqslant |u+v||u-v|$. Therefore,
\begin{align*}
    |\mathscr{F}(u_\theta,\bx)^2 - \mathscr{F}(u_{\theta'},\bx)^2| &\leqslant \big(|\mathscr{F}(u_\theta,\bx)|+| \mathscr{F}(u_{\theta'},\bx)| \big) |\mathscr{F}(u_\theta,\bx) - \mathscr{F}(u_{\theta'},\bx)|\\
    &\leqslant 2C_2^{1/2} n^{H \deg(\mathscr{F})\alpha_1}|\mathscr{F}(u_\theta,\bx) - \mathscr{F}(u_{\theta'},\bx)|.
\end{align*}
Using inequality \eqref{eq:diffProd} (remark that the product $\prod_{i=1}^{d_2}\prod_{j=1}^{s} Z_{i,j}^{I(i,j,k)}$ has less than $\deg(\mathscr{F})$ terms different from $1$), it is easy to see that
\begin{align*}
    |\mathscr{F}(u_\theta,\bx) - \mathscr{F}(u_{\theta'},\bx)|&\leqslant N(P)  \big[\max_{1\leqslant k \leqslant N(P)}\|\phi_k\|_{\infty, \bar{\Omega}}\big] \deg(\mathscr{F}) \|u_\theta-u_{\theta'}\|_{C^{\deg(\mathscr{F})}(\Omega)}\\
    &\quad \times \max_{1\leqslant k \leqslant N(P)}\prod_{i,j} \max(\|u_\theta\|_{C^{|\alpha_{i,j}|}(\Omega)}, \|u_{\theta'}\|_{C^{|\alpha_{i,j}|}(\Omega)}) ^{I(i,j,k)}. 
\end{align*}
From Proposition \ref{prop:bounding}, we deduce that
\begin{align*}
    &\max_{1\leqslant k \leqslant N(P)}\prod_{i,j} \max(\|u_\theta\|_{C^{|\alpha_{i,j}|}(\Omega)}, \|u_{\theta'}\|_{C^{|\alpha_{i,j}|}(\Omega)}) ^{I(i,j,k)} \\
    &\quad\leqslant C_{\deg(\mathscr{F}),H} (D+1)^{H\deg(\mathscr{F})}(1+\max(\|\theta\|_2, \|\theta'\|_2))^{H\deg(\mathscr{F})}.
\end{align*}

Combining the last two inequalities with  Proposition \ref{prop:lipschitzParam} gives that
\begin{align*}
    &|\mathscr{F}(u_\theta,\bx) - \mathscr{F}(u_{\theta'},\bx)| \\
    &\quad \leqslant  N(P)  \big[\max_{1\leqslant k \leqslant N(P)}\|\phi_k\|_{\infty, \bar{\Omega}}\big] \deg(\mathscr{F})  \tilde{C}_{\deg(\mathscr{F}),H}(1+d_1M(\Omega))\|\theta-\theta'\|_2\\
    &\qquad \times  C_{\deg(\mathscr{F}),H} (D+1)^{H(1+ (1+H)\deg(\mathscr{F}))}(1+\max(\|\theta\|_2, \|\theta'\|_2))^{H(1+ (1+H)\deg(\mathscr{F}))}.
\end{align*}
Hence, for all $\theta, \theta'\in B(n^{\alpha_1}, \|\cdot\|_2)$, $ |\mathscr{F}(u_\theta,\bx)^2 - \mathscr{F}(u_{\theta'},\bx)^2| \leqslant C_1 n^{\beta_1} \|\theta-\theta'\|_2$, where 
\begin{align*}
C_1 &= 2C_2^{1/2}N(P)  \big[\max_{1\leqslant k \leqslant N(P)}\|\phi_k\|_{\infty, \bar{\Omega}}]\big] \deg(\mathscr{F}) \tilde{C}_{\deg(\mathscr{F}), H} (1+d_1M(\Omega)) \\
& \quad \times C_{\deg(\mathscr{F}), H}  (D+1)^{H(1+ (1+H)\deg(\mathscr{F}))} 2^{H(1+ (1+H)\deg(\mathscr{F}))}
\end{align*}
and $\beta_1 = H(1+ (2+H)\deg(\mathscr{F}))\alpha_1$. 

 Recall that for inequality \eqref{eq:cdn2}, $\beta_2$ must satisfy $\alpha_1 + \beta_1 < \beta_2 < 1/2$. This is true for $\beta_2 = [2+H(1+ (2+H)\deg(\mathscr{F}))]\alpha_1$ and $\alpha_1 < [2+H(1+ (2+H)\deg(\mathscr{F}))]^{-1}/2$.
\end{proof}

\subsection*{Proof of Theorem \ref{thm:generalization_error}}
Let $u_0 = 0 \in \mathrm{NN}_H(D)$ be the neural network with parameter $\theta = (0, \hdots, 0)$. 
Obviously, $R_{n, n_e, n_r}^{(\mathrm{ridge})}(u_0) = R_{n, n_e, n_r}(u_0)$. Also, 
\begin{equation*}
R_{n, n_e, n_r}(u_0)  \leqslant \frac{\lambda_d}{n}\sum_{i=1}^{n} \|Y_i\|_2^2 + {\lambda_e}\|h\|_\infty  + \frac{1}{n_r}\sum_{k=1}^M \sum_{\ell=1}^{n_r} \|\mathscr{F}_k(0, \bX^{(r)}_\ell)\|_2^2.
\end{equation*}
Since each $\mathscr{F}_k$
 is a polynomial operator (see Definition \ref{defi:polyop}), it takes the form \[\mathscr{F}_k(u, \bx) = \sum_{\ell=1}^{N(P_k)} \phi_{\ell,k} \prod_{i=1}^{d_2}\prod_{j=1}^{s_k} (\partial^{\alpha_{i,j,k}}u_i(\bx))^{I_k(i,j,\ell)}.\] Therefore,
\begin{align}
R_{n, n_e, n_r}(u_0) 
& \leqslant \frac{\lambda_d}{n}\sum_{i=1}^{n} \|Y_i\|_2^2 + {\lambda_e}\|h\|_\infty  + \sum_{k=1}^M  \sum_{\ell = 1}^{N(P_k)}\|\phi_{\ell,k}\|_{\infty, \bar{\Omega}} \nonumber \\
&:= I, \label{eq:borne0}
\end{align}
where $I$ does not depend on $\lambda_{(\mathrm{ridge})}$, $n_e$, and $n_r$.

Let $(\hat{\theta}^{(\mathrm{ridge})}(p, n_e, n_r, D))_{p\in\mathbb{N}}$ be any minimizing sequence of the empirical risk of the ridge PINN, i.e.,  $\lim_{p \to \infty}R^{(\mathrm{ridge})}_{n, n_e, n_r}(u_{\hat{\theta}^{(\mathrm{ridge})}(p, n_e, n_r, D)}) = \inf_{\theta \in \Theta_{H,D}}\,R^{(\mathrm{ridge})}_{n, n_e, n_r}(u_\theta)$.
In the rest of the proof, we let $n_{r,e} = \min(n_r, n_e)$.
We will make use of the following three sets:
$\mathcal{E}_1(n_{r,e}) = \{\theta \in \Theta_{H,D},\ \|\theta\|_2 \geq n_{r,e}^{\kappa}\}$,
$\mathcal{E}_2(n_{r,e}) = \{\theta \in \Theta_{H,D}, \  n_{r,e}^{\kappa/4}\leq \|\theta\|_2 \leq n_{r,e}^{\kappa}\}$, and
    $\mathcal{E}_3(n_{r,e}) = \{\theta \in \Theta_{H,D}, \  \|\theta\|_2 \leq n_{r,e}^{\kappa/4}\}$.
Clearly, $\Theta_{H,D} = \mathcal{E}_1 \cup \mathcal{E}_2 \cup \mathcal{E}_3$. 
The proof relies on the argument that almost surely, given any $n_r$ and $n_e$, for all $p$ large enough, $\hat{\theta}^{(\mathrm{ridge})}(p, n_e, n_r, D) \in \mathcal{E}_2 \cup \mathcal{E}_3$. 
Moreover, on $\mathcal{E}_2 \cup \mathcal{E}_3$, the empirical risk function $R_{n, n_e, n_r}^{(\mathrm{ridge})}$ is close to the theoretical risk $\mathscr{R}_n$, when $n_{r,e}$ is large enough. 
For clarity, the proof is divided into four steps.
 
\paragraph{Step 1}
We start by observing that, for any $\theta \in \mathcal{E}_1(n_{r,e})$,
$R_{n, n_e, n_r}^{(\mathrm{ridge})}(\theta) \geqslant \lambda_{(\mathrm{ridge})}\|\theta\|_2^2\geqslant  n_{r,e}^{\kappa}$.
Therefore, according to \eqref{eq:borne0}, once $n_{r,e} \geq (I+1)^{1/\kappa}$,
\[ \inf_{\theta \in \mathcal{E}_3(n_{r,e})}R_{n, n_e,n_r}^{(\mathrm{ridge})}(u_\theta) + 1 \leqslant R_{n, n_e,n_r}^{(\mathrm{ridge})}(u_0) +1 \leqslant \inf_{\theta \in \mathcal{E}_1(n_{r,e})}R_{n, n_e,n_r}^{(\mathrm{ridge})}(u_\theta).\] This shows that, for all $n_{r,e}$ large enough and for all $p$ large enough, $\hat{\theta}^{(\mathrm{ridge})}(p, n_e, n_r, D) \notin \mathcal{E}_1(n_{r,e})$. 

\paragraph{Step 2} Applying Proposition \ref{prop:GKopH} and Proposition \ref{prop:GKopF} with $\alpha_1 = \kappa$ and $\beta_2 = (2+H(1+(2+H)\max_k\deg(\mathscr{F}_k)))\alpha_1$, and then Theorem \ref{thm:approx_integral}, we know that, almost surely, there exists $N \in \mathbb{N}^\star$ such that, for all $n_{r,e}\geqslant N$,
\begin{align}
    &\sup_{\theta\in \mathcal{E}_2(n_{r,e})\cup \mathcal{E}_3(n_{r,e})} \Big|\frac{1}{n_e}\sum_{j=1}^{n_e} \|u_\theta(\bX_j^{(e)})-h(\bX_j^{(e)})\|_2^2 - \mathbb{E}\|u_\theta(\bX^{(e)})-h(\bX^{(e)})\|_2^2 \Big| \nonumber\\
    &\quad \leqslant \log^2(n_{r,e}) n_{r,e}^{\beta_2-1/2}
    \label{eq:GKh2}
\end{align}
and, for each $1 \leqslant k \leqslant M$,
\begin{equation}
    \sup_{\theta\in \mathcal{E}_2(n_{r,e})\cup \mathcal{E}_3(n_{r,e})} \Big|\frac{1}{n_r}\sum_{\ell=1}^{n_r}  \mathscr{F}_k(u_\theta, \bX^{(r)}_\ell)^2 - \frac{1}{|\Omega|}\int_\Omega \mathscr{F}_k(u_\theta, \bx)^2 d\bx\Big| \leqslant \log^2(n_{r,e}) n_{r,e}^{\beta_2-1/2}.
    \label{eq:GKF2}
\end{equation} 
Thus, almost surely, for all $n_{r,e}$ large enough and for all $\theta \in \mathcal{E}_2(n_{r,e})$,
\begin{align*}
    R_{n, n_e, n_r}^{(\mathrm{ridge})}(u_\theta) &\geqslant  \mathscr{R}_n(u_\theta) + \lambda_{(\mathrm{ridge})}\|\theta\|_2^2  - (M+1) \log^2(n_{r,e}) n_{r,e}^{\beta_2-1/2}.
\end{align*}
But, for all $\theta \in \mathcal{E}_2(n_{r,e})$,
$\lambda_{(\mathrm{ridge})}\|\theta\|_2^2 \geqslant n_{e,r}^{-\kappa / 2}$. Upon noting that $-\nicefrac{\kappa}{2}  > \beta_2 -\nicefrac{1}{2}$, we conclude that, almost surely, for all $n_{r,e}$ large enough and for all $\theta \in \mathcal{E}_2(n_{r,e})$, $ R_{n, n_e,n_r}^{(\mathrm{ridge})}(u_\theta) \geqslant \mathscr{R}_n(u_\theta)$.

\paragraph{Step 3}
Clearly, for all $\theta \in \mathcal{E}_3(n_{r,e})$,
$\lambda_{(\mathrm{ridge})}\|\theta\|_2^2 \leqslant n_{e,r}^{-\kappa / 2}$. Using inequalities \eqref{eq:GKh2} and \eqref{eq:GKF2}, we deduce that, almost surely, for all $n_{r,e}$ large enough and for all $\theta \in \mathcal{E}_3(n_{r,e})$, $|R_{n, n_e,n_r}^{(\mathrm{ridge})}(u_\theta) - \mathscr{R}_n(u_\theta)| \leqslant (M+2)\log^2(n_{r,e}) n_{r,e}^{-\kappa/2}$.
 
\paragraph{Step 4} Fix $\varepsilon > 0$.
Let $(\theta_p)_{p\in\mathbb{N}}$ be any minimizing sequence of the theoretical risk function $\mathscr{R}_n$, that is, 
$\lim_{p\to\infty} \mathscr{R}_n(u_{\theta_p}) = \inf_{\theta \in \Theta_{H,D}}\mathscr{R}_n(u_\theta)$.
Thus, by definition, there exists some $P_\varepsilon \in \mathbb{N}$ such that $|\mathscr{R}_n(u_{\theta_{P_\varepsilon}}) - \inf_{\theta \in \Theta_{H,D}}\mathscr{R}_n(u_\theta)|\leqslant \varepsilon$. 

For fixed $n_{r,e}$, according to Step 1, we have, for all $p$ large enough, $\hat{\theta}^{(\mathrm{ridge})}(p, n_e, n_r, D) \in  \mathcal{E}_2(n_{r,e}) \cup \mathcal{E}_3(n_{r,e})$. So, according to Step 2 and Step 3,
\[\mathscr{R}_{n}(u_{\hat{\theta}^{(\mathrm{ridge})}(p, n_e, n_r, D)}) \leqslant R_{n, n_e,n_r}^{(\mathrm{ridge})}(u_{\hat{\theta}^{(\mathrm{ridge})}(p, n_e, n_r, D)})+ (M+2)\log^2(n_{r,e}) n_{r,e}^{-\kappa/2}.\] 
Now, by definition of the minimizing sequence $(\hat{\theta}^{(\mathrm{ridge})}(p, n_e, n_r, D))_{p\in\mathbb{N}}$, for all $p$ large enough, $R_{n, n_e,n_r}^{(\mathrm{ridge})}(u_{\hat{\theta}^{(\mathrm{ridge})}(p, n_e, n_r, D)}) \leqslant  \inf_{\theta \in \Theta_{H,D}} R_{n, n_e,n_r}^{(\mathrm{ridge})}(u_\theta) + \varepsilon$. Also, according to Step 3,
\begin{align*}
    \inf_{\theta \in \mathcal{E}_2(n_{r,e}) \cup \mathcal{E}_3(n_{r,e})} R_{n, n_e,n_r}^{(\mathrm{ridge})}(u_\theta) &\leqslant \inf_{\theta \in \mathcal{E}_3(n_{r,e})} R_{n, n_e,n_r}^{(\mathrm{ridge})}(u_\theta)\\
    & \leqslant \inf_{\theta \in \mathcal{E}_3(n_{r,e})} \mathscr{R}_{n}(u_\theta) + (M+2)\log^2(n_{r,e}) n_{r,e}^{-\kappa/2}.
\end{align*}
Observe that, for all $n_{r,e}$ large enough, $\theta_{P_\varepsilon} \in \mathcal{E}_3(n_{r,e})$. Therefore, $\inf_{\theta \in \mathcal{E}_3(n_{r,e})} \mathscr{R}_{n}(u_\theta) \leqslant \mathscr{R}_{n}(u_{\theta_{P_\varepsilon}})$. Combining the previous inequalities, we conclude that, almost surely, for all $n_{r,e}$ large enough and for all $p$ large enough, 
\[\mathscr{R}_{n}(u_{\hat{\theta}^{(\mathrm{ridge})}(p, n_e, n_r, D)}) \leqslant \inf_{\theta \in \Theta_{H,D}}\mathscr{R}_n(u_\theta) + 3\varepsilon.\]
Since $\varepsilon$ is arbitrary, almost surely,
$\lim_{n_e,n_r \to \infty} \lim_{p\to\infty} \mathscr{R}_{n}(u_{\hat{\theta}^{(\mathrm{ridge})}(p, n_e, n_r, D)}) = \inf_{\theta \in \Theta_{H,D}}\mathscr{R}_n(u_\theta)$.

\subsection*{Proof of Theorem \ref{thm:approximation}}
The result is a direct consequence of Theorem \ref{thm:generalization_error}, Proposition \ref{prop:densite} and of the continuity of $\mathscr{R}_n$ with respect to the $C^K(\Omega)$ norm.

\section{Proofs of Section \ref{sec:functional}}
\subsection*{Proof of Proposition \ref{prop:laxMLin}}
Since the functions in $H^{m+1}(\Omega, \mathbb{R}^{d_2})$ are only defined almost everywhere, we first have to give a meaning to the pointwise evaluations $u(\bX_i)$ when $u\in H^{m+1}(\Omega, \mathbb{R}^{d_2})$. 
Since $\Omega$ is a bounded Lipschitz domain and $(m +1)  > d_1/2$, we can use the Sobolev embedding of Theorem \ref{thm:sobIneq}.
Clearly, $\tilde{\Pi}$ is linear and $\|\tilde \Pi (u)\|_{\infty} \leqslant  C_{\Omega} \| u \|_{H^{m+1}(\Omega)}$. The natural choice to evaluate $u \in H^{m+1}(\Omega,\mathbb{R}^{d_2})$ at the point $\bX_i$ is therefore to evaluate its unique continuous modification $\tilde \Pi (u)$ at $\bX_i$.
    
By assumption, $\mathscr{F}_k(u, \cdot) = \mathscr{F}_k^{(\mathrm{lin})}(u, \cdot) + B_k$, where $\mathscr{F}_k^{(\mathrm{lin})}(u, \cdot) = \sum_{|\alpha|\leqslant K} \langle A_{k,\alpha}, \partial^\alpha u\rangle$ and $A_{k,\alpha} \in C^\infty(\bar{\Omega}, \mathbb{R}^{d_1})$.
Next, consider the symmetric bilinear form, defined for all $u,v \in H^{m+1}(\Omega, \mathbb{R}^{d_2})$ by
\begin{align*}
    \mathcal{A}_n(u,v) &= \frac{\lambda_d}{n} \sum_{i=1}^n \langle\tilde \Pi (u)(\bX_i) , \tilde \Pi(v)(\bX_i) \rangle+\lambda_e \mathbb{E}\langle \tilde \Pi(u)(\bX^{(e)}), \tilde \Pi(v)(\bX^{(e)})\rangle\\
    &\quad +\frac{1}{|\Omega|}\sum_{k=1}^{M}\int_{\Omega} \mathscr{F}_k^{(\mathrm{lin})}(u,\bx)\mathscr{F}_k^{(\mathrm{lin})}(v,\bx)d\bx + \frac{\lambda_t }{|\Omega|} \!\sum_{|\alpha|\leqslant m+1}\!\int_{\Omega}\langle \partial^\alpha u(\bx), \partial^\alpha v(\bx)\rangle d\bx,
    \end{align*}
along with the linear form defined for all $u \in H^{m+1}(\Omega, \mathbb{R}^{d_2})$ by
\begin{align*}
    \mathcal{B}_n(u) &= \frac{\lambda_d}{n} \sum_{i=1}^n \langle Y_i, \tilde \Pi(u)(\bX_i)\rangle + \lambda_e \mathbb{E}\langle \tilde \Pi(u)(\bX^{(e)}), h(\bX^{(e)})\rangle\\
    &\quad - \frac{1}{|\Omega|}\sum_{k=1}^{M}\int_{\Omega} B_k(\bx)\mathscr{F}_k^{(\mathrm{lin})}(v,\bx)d\bx .
\end{align*}
Observe that \[\mathcal{A}_n(u,u) -2\mathcal{B}_n(u) = \mathscr{R}_n^{(\mathrm{reg})}(u) - \frac{\lambda_d}{n} \sum_{i=1}^n \|Y_i\|_2^2 -\lambda_e \mathbb{E}\|h(\bX^{(e)})\|_2^2 - \frac{1}{|\Omega|}\sum_{k=1}^{M}\int_{\Omega} B_k(\bx)^2d\bx.\] 
In addition, $\mathcal{A}_n(u,u) \geqslant \lambda_t \|u\|_{H^{m+1}(\Omega)}^2$, where $\lambda_t >0$, so that $\mathcal{A}_n$ is coercive on the normed space $(H^{m+1}(\Omega), \|\cdot\|_{H^{m+1}(\Omega)})$.  
Since $(m+1) > \max(d_1/2, K)$, one has that 
    \[|\mathcal{A}_n(u,v)| \leqslant ((\lambda_d+\lambda_e) C_{\Omega}^2 + \sum_{1\leqslant k \leqslant M}(\sum_{|\alpha|\leqslant K} \|A_{k,\alpha}\|_{\infty,\Omega})^2 +\lambda_t)\| u \|_{H^{m+1}(\Omega)}\| v \|_{H^{m+1}(\Omega)},\]
    and 
    \[|\mathcal{B}_n(u)| \leqslant C_\Omega \Big(\frac{\lambda_d}{n} \sum_{i=1}^n \|Y_i\|_2  + \lambda_e \|h\|_{\infty} + \sum_{k=1}^M(\|B_{k}\|_{\infty,\Omega}\sum_{|\alpha|\leqslant K} \|A_{k,\alpha}\|_{\infty,\Omega})\Big) \| u \|_{H^{m+1}(\Omega)}.\]
    This shows that the operators $\mathcal{A}_n$ and $\mathcal{B}_n$ are continuous. Therefore, by the Lax-Milgram theorem \citep[e.g.,][Corollary 5.8]{brezis2010functional}, there exists a unique $\hat u \in H^{m+1}(\Omega, \mathbb{R}^{d_2})$ such that $\mathcal{A}_n(\hat u,\hat u) - 2\mathcal{B}_n(\hat u) = \min_{u \in H^{m+1}(\Omega, \mathbb{R}^{d_2})}\mathcal{A}_n(u,u) - 2\mathcal{B}_n(u)$. This directly implies that $\hat u$ is the unique minimizer of $\mathscr{R}_n^{(\mathrm{reg})}$ over $H^{m+1}(\Omega, \mathbb{R}^{d_2})$. Furthermore, the Lax-Milgram theorem also states that $\hat u$ is the unique element of $H^{m+1}(\Omega, \mathbb{R}^{d_2})$ such that, for all $v \in H^{m+1}(\Omega, \mathbb{R}^{d_2})$, $\mathcal{A}_n(\hat u,v) = \mathcal{B}_n(v)$. This concludes the proof of the proposition.

\subsection*{Proof of Proposition \ref{prop:sequenceCvLin}}
Let $\hat u_n$ be the unique minimizer of the regularized theoretical risk $\mathscr{R}^{(\mathrm{reg})}_n$ over $H^{m+1}(\Omega, \mathbb{R}^{d_2})$ given by Proposition \ref{prop:laxMLin}. Notice that
\[\inf_{u \in C^{\infty}(\bar{\Omega}, \mathbb{R}^{d_2})} \mathscr{R}^{\mathrm{(reg)}}_n(u) = \inf_{u \in H^{m+1}(\Omega, \mathbb{R}^{d_2})} \mathscr{R}^{\mathrm{(reg)}}_n(u) = \mathscr{R}_n(\hat u_n).\]
The first equality is a consequence of the density of $ C^{\infty}(\bar{\Omega}, \mathbb{R}^{d_2})$ in $H^{m+1}(\Omega,\mathbb{R}^{d_2})$, together with the continuity of the function $\mathscr{R}^{\mathrm{(reg)}}_n : H^{m+1}(\Omega, \mathbb{R}^{d_2}) \to \mathbb{R}$ with respect to the $H^{m+1}(\Omega)$ norm (see the proof of Proposition \ref{prop:laxMLin}). The density argument follows from the extension theorem of \citet[][Chapter VI.3.3, Theorem 5]{stein1970lipschitz} and from \citet[Chapter 5.3, Theorem 3]{evans2010partial}.

Our goal is to show that the regularized theoretical risk satisfies some form of continuity, so that we can connect $\mathscr{R}^{(\mathrm{reg})}(u_p)$ and $\mathscr{R}^{(\mathrm{reg})}(\hat u_n)$. 
Recall that, by assumption, $\mathscr{F}_k(u, \cdot) = \mathscr{F}_k^{(\mathrm{lin})}(u, \cdot) + B_k$, where $\mathscr{F}_k^{(\mathrm{lin})}(u, \cdot) = \sum_{|\alpha|\leqslant K} \langle A_{k,\alpha}(\cdot), \partial^\alpha u(\cdot)\rangle$ and $A_{k,\alpha} \in C^\infty(\bar{\Omega}, \mathbb{R}^{d_1})$. Observe that 
    \begin{equation}
        \mathscr{R}_n^{(\mathrm{reg})}(u) = F(u)+ \frac{1}{|\Omega|}I(u),
        \label{eq:decomposition}
    \end{equation}
where \[F(u) =   \frac{\lambda_d}{n} \sum_{i=1}^n \|\tilde \Pi (u)(\bX_i) - Y_i\|_2^2 + \lambda_e \mathbb{E}\|\tilde \Pi (u)(\bX^{(e)})-h(\bX^{(e)})\|_2^2 ,\] 
\[I(u) = \int_{\Omega}L((\partial^{m+1}_{i_1,\hdots, i_{m+1}}u(\bx))_{1\leqslant i_1,\hdots, i_{m+1} \leqslant d_1},\hdots,u(\bx), \bx)d\bx,\] 
and where the function $L$ satisfies
\[L(x^{(m+1)}, \hdots, x^{(0)}, z) =  \sum_{k=1}^M\Big(B_k(z) + \sum_{|\alpha|\leqslant K} \langle A_{k,\alpha}(z), x^{(|\alpha|)}_{\alpha}\rangle\Big)^2 +  \lambda_t \sum_{j=0}^{m+1} \|x^{(j)}\|_2^2.\]
(The term $x^{(j)} \in \mathbb{R}^{\binom{d_1+j-1}{j-1}d_2}$ corresponds to the to the concatenation of all the partial derivatives of order $j$, i.e., to the term $(\partial^{j}_{i_1,\hdots, i_{j}}u(\bx))_{1\leqslant i_1,\hdots, i_{j} \leqslant d_1}$.)
Clearly, $L\geqslant 0$ and, since $(m+1) > K$, the Lagrangian $L$ is convex in $x^{(m+1)}$. 
Therefore, according to Lemma \ref{lem:lowerSemiC0Lin}, the function $I$ is weakly lower-semi continuous on $H^{m+1}(\Omega, \mathbb{R}^{d_2})$.

Now, let us proceed by contradiction and assume that there is a sequence $(u_p)_{p\in\mathbb{N}}$ of functions such that $(i)$ $u_p\in C^\infty(\bar{\Omega}, \mathbb{R}^{d_2})$, $(ii)$ $\lim_{p\to\infty}\mathscr{R}^{(\mathrm{reg})}_n(u_p)=\mathscr{R}^{(\mathrm{reg})}_n(\hat u_n)$, and $(iii)$ $(u_p)_{p\in\mathbb{N}}$ does not converge to $\hat u_n$ with respect to the $H^m(\Omega)$ norm. 
Therefore, upon passing to a subsequence, there exists $\varepsilon >0$ such that, for all $ p \geqslant 0$, $\|u_p-\hat u_n\|_{H^m(\Omega)} \geqslant \varepsilon$.

Since $\mathscr{R}^{(\mathrm{reg})}_n(u_p) \geqslant \lambda_t \|u_p\|_{H^{m+1}(\Omega)}$, $\lambda_t>0$, and $(u_p)_{p\in\mathbb{N}}$ is a minimizing sequence, $(u_p)_{p\in \mathbb{N}}$ is bounded in $H^{m+1}(\Omega, \mathbb{R}^{d_2})$.
Therefore, Theorem \ref{thm:rellichK} states that passing to a subsequence, $(u_p)_{p\in\mathbb{N}}$ converges to a limit, say $u_\infty$, both weakly in $H^{m+1}(\Omega, \mathbb{R}^{d_2})$ and with respect to the $H^m(\Omega)$ norm. 
Then, since $I$ is weakly lower-semi continuous on $H^{m+1}(\Omega, \mathbb{R}^{d_2})$, we deduce that 
\begin{equation}
    \label{eq:cvI}
    \lim_{p\to\infty}I(u_p) \geqslant  I(u_\infty).
\end{equation}
Recalling the definition of $\tilde \Pi$ in Theorem \ref{thm:sobIneq}, we know that there exists a constant $C_\Omega >0$ such that $\|u_p-\tilde \Pi(u_\infty)\|_{\infty, \Omega} = \|\tilde \Pi(u_p - u_\infty)\|_{\infty, \Omega} \leqslant C_\Omega \|u_p-u_\infty\|_{H^{m}(\Omega)}$.
We deduce that  $\lim_{p\to\infty}F(u_p) = F(u_\infty)$. Therefore, combining this result with \eqref{eq:decomposition} and \eqref{eq:cvI}, we deduce that $\lim_{p\to\infty}\mathscr{R}^{(\mathrm{reg})}_n(u_p) \geqslant  \mathscr{R}^{(\mathrm{reg})}_n(u_\infty)$.
However, recalling that $\lim_{p\to\infty}\mathscr{R}^{(\mathrm{reg})}_n(u_p)=\mathscr{R}^{(\mathrm{reg})}_n(\hat u_n)$ and that $\hat u_n$ is the unique minimizer of $\mathscr{R}^{(\mathrm{reg})}_n$ over $H^{m+1}(\Omega,\mathbb{R}^{d_2})$, we conclude that $u_\infty = \hat u_n$. 

We just proved that there exists a subsequence of $(u_p)_{p\in \mathbb{N}}$ which converges to $\hat u_n$ with respect to the $H^{m}(\Omega)$ norm.
This contradicts the assumption $\|u_p-\hat u_n\|_{H^m(\Omega)} \geqslant \varepsilon$ for all $ p \geqslant 0$.

\subsection*{Proof of Theorem \ref{thm:functionalCv}}

The result is an immediate consequence of Theorem \ref{thm:approximation}, Propositions \ref{prop:laxMLin}, and Proposition \ref{prop:sequenceCvLin}.

\subsection*{Proof of Theorem \ref{prop:pdeSolverFunctional}}
Throughout the proof, since no data are involved, we denote the regularized theoretical risk by $\mathscr{R}^{(\mathrm{reg})}$ instead of $\mathscr{R}_n^{(\mathrm{reg})}$. Also, to make the dependence in the hyperparameter $\lambda_t$ transparent, we denote by $u(\lambda_t)$ the unique minimizer of $\mathscr{R}^{(\mathrm{reg})}$ instead of $\hat u_n$. 

We proceed by contradiction and assume that 
$\lim_{\lambda_t \to 0}\| u(\lambda_t)-u^\star\|_{H^{m}(\Omega)} \neq 0$.
If this is true, then, upon passing to a subsequence $(\lambda_{t,p})_{p\in \mathbb{N}}$ such that $\lim_{p\to \infty} \lambda_{t,p} =0$, there exists $\varepsilon >0$ such that, for all $ p \geqslant 0$, $\| u(\lambda_{t,p})- u^\star\|_{H^m(\Omega)} \geqslant \varepsilon$.

Notice that $ \| u(\lambda_{t,p})\|_{H^{m+1}(\Omega)} \leqslant \mathscr{R}^{(\mathrm{reg})}(u^\star)/\lambda_{t,p} = \|u^\star\|_{H^{m+1}(\Omega)}$.
Theorem \ref{thm:rellichK} proves that upon passing to a subsequence, $(u(\lambda_{t,p}))_{p\in \mathbb{N}}$ converges with respect to the $H^m(\Omega)$ norm to a function $u_\infty \in H^{m+1}(\Omega, \mathbb{R}^{d_2})$.
Since $m \geqslant K$, the theoretical risk $\mathscr{R}$ is continuous with respect to the $H^m(\Omega)$ norm and we have that $\mathscr{R}(u_\infty) = \lim_{p\to \infty} \mathscr{R}(u(\lambda_{t,p}))$. 
Moreover, by definition of $u(\lambda_{t,p})$ and since $\mathscr{R}(u^\star) = 0$, we have that  $\mathscr{R}(u(\lambda_{t,p})) + \lambda_{t,p} \|u(\lambda_{t,p})\|_{H^{m+1}(\Omega)}\leqslant \lambda_{t,p} \|u^\star\|_{H^{m+1}(\Omega)}$.
Therefore, $\mathscr{R}(u_\infty) = 0$ and $u_\infty = u^\star$. 
This contradicts the assumption that for all $ p \geqslant 0$, $\| u(\lambda_{t,p})- u^\star\|_{H^m(\Omega)} \geqslant \varepsilon$.

\subsection*{Proof of Proposition \ref{prop:consistencey}}
We prove the proposition in several steps. In the sequel, given a measure $\mu$ on $\Omega$ and a function $u \in H^{m+1}(\Omega, \mathbb{R}^{d_2})$, we let $\|u\|^2_{L^2(\mu)} = \int_\Omega \|\tilde \Pi(u)(\bx)\|_2^2 d\mu(\bx)$, where, as usual, $\tilde \Pi(u)$ is the unique continuous function such that $\tilde \Pi(u) = u$ almost everywhere.

\paragraph{Step 1: Decomposing the problem into two simpler ones} 
Following the framework of \citet{arnone2022spatialRegression}, the core idea is to decompose the problem into two simpler ones thanks to the linearity in $\hat u_n$ and in $Y_i$ of the identity 
\[\forall v \in H^{m+1}(\Omega, \mathbb{R}^{d_2}),\quad  \mathcal{A}_n(\hat u_n,v) = \mathcal{B}_n(v)\] 
of Proposition \ref{prop:laxMLin}.
Thus, recalling that $Y_i = u^\star(\bX_i)+\varepsilon_i$, we 
let
\begin{align*}
    \mathcal{B}^{\star}_n(v) &= \frac{\lambda_d}{n} \sum_{i=1}^n \langle u^\star(\bX_i), \tilde \Pi(v)(\bX_i)\rangle+ \lambda_e \mathbb{E}\langle\tilde \Pi(v)(\bX^{(e)}),h(\bX^{(e)})\rangle\\
    &\quad - \frac{1}{|\Omega|}\sum_{k=1}^{M}\int_{\Omega} B_k(\bx)\mathscr{F}_k^{(\mathrm{lin})}(v,\bx)d\bx 
\end{align*}
and
\[\mathcal{B}^{\mathrm{(noise)}}_n(v) = \frac{\lambda_d}{n} \sum_{i=1}^n \langle \varepsilon_i, \tilde \Pi(v)(\bX_i)\rangle.\]
Clearly, $\mathcal{B}_n = \mathcal{B}^{\star}_n + \mathcal{B}^{\mathrm{(noise)}}_n.$
Using Proposition \ref{prop:laxMLin} with $Y_i$ instead of $\varepsilon_i$, and setting $\lambda_e = 0$, we see that there exists a unique $\hat u ^{\mathrm{(noise)}}_n \in H^{m+1}(\Omega, \mathbb{R}^{d_2})$ such that, for all $v \in H^{m+1}(\Omega, \mathbb{R}^{d_2})$, $\mathcal{A}_n( \hat u ^{\mathrm{(noise)}}_n,v) = \mathcal{B}^{\mathrm{(noise)}}_n(v)$. Furthermore, $\hat u ^{\mathrm{(noise)}}_n$ is the unique minimizer over $H^{m+1}(\Omega, \mathbb{R}^{d_2})$ of
\begin{align*}
    \mathscr{R}_n^{\mathrm{(noise)}}(u) &= \frac{\lambda_d}{n} \sum_{i=1}^n \|\tilde \Pi (u)(\bX_i)-\varepsilon_i\|_2^2+\lambda_e \mathbb{E}\| u(\bX^{(e)})\|_2^2+\frac{1}{|\Omega|}\sum_{k=1}^{M}\int_{\Omega} \mathscr{F}_k^{(\mathrm{lin})}(u,\bx)^2d\bx  \\
    &\quad + \lambda_t\|u\|_{H^{m+1}(\Omega)}^2.
\end{align*}
Similarly, Proposition \ref{prop:laxMLin} shows that there exists a unique $\hat u ^{\star}_n \in H^{m+1}(\Omega, \mathbb{R}^{d_2})$ such that, for all $v \in H^{m+1}(\Omega, \mathbb{R}^{d_2})$, 
$\mathcal{A}_n( \hat u ^{\star}_n,v) = \mathcal{B}_n^{\star}(v)$, and
$\hat u ^{\star}_n$ is the unique minimizer over $H^{m+1}(\Omega, \mathbb{R}^{d_2})$ of 
\begin{align*}
    \mathscr{R}_n^\star(u) &= \frac{\lambda_d}{n} \sum_{i=1}^n \|\tilde \Pi (u-u^\star)(\bX_i)\|_2^2+\lambda_e \mathbb{E}\| \tilde \Pi(u)(\bX^{(e)}) - h(\bX^{(e)})\|_2^2\\
    &\quad +\frac{1}{|\Omega|}\sum_{k=1}^{M}\int_{\Omega} \mathscr{F}_k(u,\bx)^2d\bx + \lambda_t\|u\|_{H^{m+1}(\Omega)}^2.
\end{align*}
By the bilinearity of $\mathcal{A}_n$, one has, for all $ v \in H^{m+1}(\Omega, \mathbb{R}^{d_2})$, $\mathcal{A}_n( \hat u_n ^{\star}+ \hat u_n ^{\mathrm{(noise)}} ,v) = \mathcal{B}_n(v)$. However, according to Proposition \ref{prop:laxMLin}, $\hat u_n$ is the unique element of $H^{m+1}(\Omega, \mathbb{R}^{d_2})$ satisfying this property. Therefore, 
$\hat u_n = \hat u_n ^{\star}+ \hat u_n ^{\mathrm{(noise)}}$.
\paragraph{Step 2: Some properties of the minimizers} According to Lemma \ref{lem:measurability}, $\hat u_n$, $\hat u_n^\star$, and $\hat u_n^{\mathrm{(noise)}}$ are random variables. Our goal in this paragraph is to prove that $\mathbb{E}\|\hat u_n\|_{H^{m+1}(\Omega)}^2$, $\mathbb{E}\|\hat u_n^\star\|_{H^{m+1}(\Omega)}^2$, and $\mathbb{E}\|\hat u_n^{\mathrm{(noise)}}\|_{H^{m+1}(\Omega)}^2 $ are finite, so that we can safely use conditional expectations on $\hat u_n$,  $\hat u_n^\star$, and  $\hat u_n^\mathrm{(noise)}$. 
Recall that, since  $\lambda_t \|\hat u_n\|_{H^{m+1}(\Omega)}^2 
    \leqslant \mathscr{R}^{(\mathrm{reg})}_n(\hat u_n)  \leqslant \mathscr{R}^{(\mathrm{reg})}_n(0) $, and since $\mathscr{F}_k^{(\mathrm{lin})}(0, \cdot) = 0$, 
\[\lambda_t \|\hat u_n\|_{H^{m+1}(\Omega)}^2 \leqslant \frac{\lambda_d}{n} \sum_{i=1}^n \|Y_i\|_2^2 + \lambda_e \mathbb{E}\|h(\bX^{(e)})\|_2^2 + \frac{1}{|\Omega|} \sum_{k=1}^M \int_\Omega B_k(\bx)^2 d\bx.\]
Hence, 
\[\mathbb{E}\|\hat u_n\|_{H^{m+1}(\Omega)}^2 \leqslant \lambda_t^{-1}\Big(\lambda_d\mathbb{E}\|u^\star(\bX) + \varepsilon\|_2^2 + \lambda_e \mathbb{E}\|h(\bX^{(e)})\|_2^2 + \frac{1}{|\Omega|}\sum_{k=1}^M \int_\Omega B_k(\bx)^2 d\bx\Big).\]
Similarly,
\begin{equation*}
    \mathbb{E}\|\hat u_n^\star\|_{H^{m+1}(\Omega)}^2 \leqslant \lambda_t^{-1}\Big(\lambda_d\mathbb{E}\|u^\star(\bX)\|_2^2 + \lambda_e \mathbb{E}\|h(\bX^{(e)})\|_2^2 + \frac{1}{|\Omega|}\sum_{k=1}^M \int_\Omega B_k(\bx)^2 d\bx\Big),
    \label{eq:boundingOptimizerExpectancy}
\end{equation*} 
and $\mathbb{E}\|\hat u_n^{\mathrm{(noise)}}\|_{H^{m+1}(\Omega)}^2 \leqslant \lambda_t^{-1}\lambda_d\mathbb{E}\|\varepsilon\|_2^2$.

\paragraph{Step 3: Bias-variance decomposition} In this paragraph, we use the notation $\mathcal{A}_{(\bx, e)}(u,u)$ instead of $\mathcal{A}_n(u,u)$, to make the dependence of $\mathcal{A}_n$ in the random variables $\bx = (\bX_1, \hdots, \bX_n)$ and $e=(\varepsilon_1, \hdots, \varepsilon_n)$ more explicit. We do the same with $\mathcal{B}_n$ and $\hat u_n^{\mathrm{(noise)}}$. 
 Observe that, for any $(\bx,e) \in\Omega^n \times \mathbb{R}^{nd_2}$ and for any $u \in H^{m+1}(\Omega, \mathbb{R}^{d_2})$, one has \[\mathcal{A}_{(\bx,-e)}(u,u) - 2 \mathcal{B}_{(\bx,e)}^{\mathrm{(noise)}}(u) = \mathcal{A}_{(\bx,e)}(-u,-u) - 2 \mathcal{B}_{(\bx,-e)}^{\mathrm{(noise)}}(-u).\]
Therefore, $\hat u^{\mathrm{(noise)}}_{(\bx,e)} = -\hat u^{\mathrm{(noise)}}_{(\bx,-e)}$. 

Since, by assumption, $\varepsilon$ has the same law as $-\varepsilon$, this implies $\mathbb{E}(\hat u_n^{\mathrm{(noise)}}\mid \bX_1, \hdots, \bX_n) = 0$, and so $\mathbb{E}(\hat u_n^{\mathrm{(noise)}}) = 0$.
Moreover, since $\hat u_n^\star$ is a measurable function of $\bX_1, \hdots, \bX_n$, we have $\mathbb{E}(\hat u_n^\star\mid\bX_1, \hdots, \bX_n) = \hat u_n^\star$. Recalling (Step 1) that $\hat u_n = \hat u_n ^{\star}+ \hat u_n ^{\mathrm{(noise)}}$, we deduce the following bias-variance decomposition:  
\begin{equation}
    \label{eq:biasVariance}
    \mathbb{E}\|\hat u_n-u^\star\|^2_{L^2(\mu_\bX)} = \mathbb{E}\|\hat u^\star_n-u^\star\|^2_{L^2(\mu_\bX)}+ \mathbb{E}\|\hat u^{\mathrm{(noise)}}_n\|^2_{L^2(\mu_\bX)}.
\end{equation}

\paragraph{Step 4: Bounding the bias} 
Recall that $\hat u_n^\star$ minimizes $\mathscr{R}_n^\star$ over $H^{m+1}(\Omega, \mathbb{R}^{d_2})$, so that $\mathscr{R}_n^\star(u^\star) \geqslant \mathscr{R}_n^\star(\hat u^\star_n)$.
Therefore, $\mathrm{PI}(u^\star)+ \lambda_t\|u^\star\|_{H^{m+1}(\Omega)}^2 \geqslant \frac{\lambda_d}{n} \sum_{i=1}^n \|\tilde \Pi (\hat u_n^\star-u^\star)(\bX_i)\|_2^2$. 
We deduce that
\begin{align*}
    &\frac{1}{\lambda_d}\big(\mathrm{PI}(u^\star)+ \lambda_t\|u^\star\|_{H^{m+1}(\Omega)}^2\big) \\
    &\geqslant  \frac{\|\hat u_n^\star-u^\star\|_{H^{m+1}(\Omega)}^2}{n} \sum_{i=1}^n \Big\|\tilde \Pi \Big (\frac{\hat u_n^\star-u^\star}{\|\hat u_n^\star-u^\star\|_{H^{m+1}(\Omega)}}\Big)(\bX_i)\Big\|_2^2 \\
    & \geqslant  \|\hat u_n^\star-u^\star\|_{L^2(\mu_\bX)}^2 \\
    & \quad - \|\hat u_n^\star-u^\star\|_{H^{m+1}(\Omega)}^2  \sup_{\|u\|_{H^{m+1}(\Omega)}\leqslant 1} \Big( \mathbb{E}\|\tilde \Pi(u)(\bX)\|_2^2 - \frac{1}{n}\sum_{i=1}^n \|\tilde \Pi(u)(\bX_i)\|_2^2\Big) \\
  & \geqslant  \|\hat u_n^\star-u^\star\|_{L^2(\mu_\bX)}^2\\
  & \quad 
   - 2  \big(\|\hat u_n^\star\|_{H^{m+1}(\Omega)}^2+\|u^\star\|_{H^{m+1}(\Omega)}^2\big) \!\sup_{\|u\|_{H^{m+1}(\Omega)}\leqslant 1}\! \Big( \mathbb{E}\|\tilde \Pi(u)(\bX)\|_2^2- \frac{1}{n}\sum_{i=1}^n \|\tilde \Pi(u)(\bX_i)\|_2^2\Big).
\end{align*}
Moreover, $\mathrm{PI}(u^\star)  + \lambda_t\|u^\star\|_{H^{m+1}(\Omega)}^2 \geqslant \lambda_t \|\hat u_n^\star\|_{H^{m+1}(\Omega)}^2$.
Taking expectations, we conclude by Lemma \ref{lem:empiricalL2} that there exists a constant $C_\Omega '$, depending only on $\Omega$, such that
\[\mathbb{E}\|\hat u_n^\star-u^\star\|_{L^2(\mu_\bX)}^2\leqslant \frac{1}{\lambda_d}\big(\mathrm{PI}(u^\star) + \lambda_t\|u^\star\|_{H^{m+1}(\Omega)}^2\big) + \frac{C_\Omega 'd_2^{1/2}}{n^{1/2}} \Big(2\|u^\star\|_{H^{m+1}(\Omega)}^2 + \frac{\mathrm{PI}(u^\star)}{\lambda_t}\Big) .\]

\paragraph{Step 5: Bounding the variance} 
Since $\hat u_n^{\mathrm{(noise)}}$ minimizes $\mathscr{R}_n^{\mathrm{(noise)}}$ over $H^{m+1}(\Omega, \mathbb{R}^{d_2})$, we have $\mathscr{R}_n^{\mathrm{(noise)}}(0) \geqslant \mathscr{R}_n^{\mathrm{(noise)}}(\hat u^{\mathrm{(noise)}}_n)$.
So, 
\[\frac{\lambda_d}{n} \sum_{i=1}^n \|\varepsilon_i\|_2^2  \geqslant \frac{\lambda_d}{n} \sum_{i=1}^n \|\tilde \Pi (\hat u_n^\mathrm{(noise)})(\bX_i)- \varepsilon_i\|_2^2.\]
Observing  that  $\|\tilde \Pi (\hat u_n^\mathrm{(noise)})(\bX_i)- \varepsilon_i\|_2^2 = \|\tilde \Pi (\hat u_n^\mathrm{(noise)})(\bX_i)\|_2^2 -2\langle \tilde \Pi (\hat u_n^\mathrm{(noise)})(\bX_i),\varepsilon_i\rangle + \|\varepsilon_i\|_2^2$, we deduce that
\[\frac{2}{n} \sum_{i=1}^n \langle \tilde \Pi (\hat u_n^\mathrm{(noise)})(\bX_i),\varepsilon_i\rangle  \geqslant  \frac{1}{n} \sum_{i=1}^n \|\tilde \Pi (\hat u_n^\mathrm{(noise)})(\bX_i)\|_2^2,\]
and  
\begin{align*}
&\Big\langle \int_\Omega \tilde \Pi(\hat u_n^\mathrm{(noise)})d \mu_\bX, \frac{2}{n} \sum_{i=1}^n \varepsilon_i\Big\rangle + \frac{2}{n} \sum_{i=1}^n \Big\langle \tilde \Pi (\hat u_n^\mathrm{(noise)})(\bX_i) - \int_\Omega \tilde \Pi(\hat u_n^\mathrm{(noise)})d \mu_\bX,\varepsilon_i\Big\rangle  \\
& \quad \geqslant  \frac{1}{n} \sum_{i=1}^n \|\tilde \Pi (\hat u_n^\mathrm{(noise)})(\bX_i)\|_2^2.
\end{align*}
Therefore, 
\begin{align*}
     \|\hat u_n^\mathrm{(noise)}\|_{L^2(\mu_\bX)}^2  &\leqslant \Big\langle \int_\Omega \tilde \Pi(\hat u_n^\mathrm{(noise)})d \mu_\bX, \frac{2}{n} \sum_{i=1}^n \varepsilon_i\Big\rangle  \\
    &+ \|\hat u_n^\mathrm{(noise)}\|_{H^{m+1}(\Omega)}\sup_{\|u\|_{H^{m+1}(\Omega)}\leqslant 1} \frac{1}{n} \sum_{j=1}^n \langle \tilde \Pi (u)(\bX_j) - \mathbb{E}(\tilde \Pi(u)(\bX)),\varepsilon_j\rangle \\ 
    &+ \|\hat u_n^\mathrm{(noise)}\|_{H^{m+1}(\Omega)}^2\sup_{\|u\|_{H^{m+1}(\Omega)}\leqslant 1} \Big(\mathbb{E}\|\tilde \Pi(u)(\bX_i)\|_2^2 - \frac{1}{n}\sum_{i=1}^n \|\tilde \Pi(u)(\bX_i)\|_2^2\Big)\\
    &:= A+B+C.
\end{align*}
According to the Cauchy-Schwarz inequality,
\[\mathbb{E}(A)  \leqslant \Big(\mathbb{E}\Big\| \int_\Omega \tilde \Pi(\hat u_n^\mathrm{(noise)})d \mu_\bX \Big\|_2^2\Big)^{1/2}  \times \frac{2 (\mathbb{E}\|\varepsilon\|_2^2)^{1/2}}{n^{1/2}},\]
and so, by Jensen's inequality,
\[\mathbb{E}(A) \leqslant \big(\mathbb{E}\|\hat u_n^\mathrm{(noise)}\|_{L^2(\mu_\bX)}^2\big)^{1/2}  \times \frac{2 (\mathbb{E}\|\varepsilon\|_2^2)^{1/2}}{n^{1/2}}.\]
The inequality $\mathscr{R}_n^{\mathrm{(noise)}}(0) \geqslant \mathscr{R}_n^{\mathrm{(noise)}}(\hat u^{\mathrm{(noise)}}_n)$ also implies that
\[\frac{\lambda_d}{n} \sum_{i=1}^n \|\varepsilon_i\|_2^2  \geqslant \frac{\lambda_d}{n} \sum_{i=1}^n \|\tilde \Pi (\hat u_n^\mathrm{(noise)})(\bX_i)- \varepsilon_i\|_2^2 + \lambda_t \|\hat u_n^\mathrm{(noise)}\|_{H^{m+1}(\Omega)}^2.\]
Therefore,
\[\frac{\lambda_d}{n \lambda_t} \sum_{i=1}^n 2\langle \tilde \Pi (\hat u_n^\mathrm{(noise)})(\bX_i),\varepsilon_i\rangle \geqslant  \|\hat u_n^\mathrm{(noise)}\|_{H^{m+1}(\Omega)}^2,\]
and
\[\frac{\lambda_d}{\lambda_t} \sup_{\|u\|_{H^{m+1}(\Omega)}\leqslant 1} \frac{1}{n} \sum_{j=1}^n \langle \tilde \Pi (u)(\bX_j),\varepsilon_j\rangle  \geqslant   \|\hat u_n^\mathrm{(noise)}\|_{H^{m+1}(\Omega)}.\]
By Theorem \ref{thm:sobIneq}, if $\|u\|_{H^{m+1}(\Omega)}\leqslant 1$, then  $ \langle \mathbb{E}(\tilde \Pi(u)(\bX)),\frac{1}{n} \sum_{j=1}^n\varepsilon_j\rangle \leqslant \frac{C_\Omega d_2^{1/2}}{n} \|\sum_{i=1}^n \varepsilon_i\|_2$. Thus, 
\begin{align*}
    &\|\hat u_n^\mathrm{(noise)}\|_{H^{m+1}(\Omega)}\\
    &\leqslant\frac{\lambda_d}{\lambda_t} \Big( \frac{C_\Omega d_2^{1/2}}{n} \|\sum_{i=1}^n \varepsilon_i\|_2 + \sup_{\|u\|_{H^{m+1}(\Omega)}\leqslant 1} \frac{1}{n} \sum_{j=1}^n \langle \tilde \Pi (u)(\bX_j) - \mathbb{E}(\tilde \Pi(u)(\bX)),\varepsilon_j\rangle \Big) .
\end{align*} 
Using Lemma \ref{lem:empiricalProcess} together with the fact that, for all $\bx, \by \in \mathbb{R}$, $(\bx+\by)^2 \leqslant 2(\bx^2 + \by^2)$, 
\begin{equation*}
    \mathbb E \|\hat u_n^\mathrm{(noise)}\|_{H^{m+1}(\Omega)}^2 \leqslant \frac{4\lambda_d^2}{n\lambda_t^2} C_\Omega^2 d_2 \mathbb E\|\varepsilon\|_2^2.
\end{equation*} 
Similarly, observing that for all random variables $X, Y \in \mathbb{R}$, $\mathbb E(XY)^2 \leqslant \mathbb E(X^2) \mathbb E(Y^2) $, 
\[\mathbb E(B) \leqslant \frac{4\lambda_d}{n\lambda_t} C_\Omega^2 d_2 \mathbb E\|\varepsilon\|_2^2.\]
Moreover, by Lemma \ref{lem:empiricalL2} and the inequality $\mathbb E(XYZ)^2 \leqslant \mathbb E(X^2) \mathbb E(Y^2) \mathbb E(Z^2) $, 
\[\mathbb E(C) \leqslant \frac{\lambda_d^2}{n^{3/2} \lambda_t^2}  C_\Omega^2 d_2^{3/2} \mathbb E\|\varepsilon\|_2^2.\]
Therefore, we conclude that there exists a constant $C_\Omega > 0$, depending only on $\Omega$, such that 
\begin{align*}
    \mathbb{E}\|\hat u_n^\mathrm{(noise)}\|_{L^2(\mu_\bX)}^2 &\leqslant \big(\mathbb{E}\|\hat u_n^\mathrm{(noise)}\|_{L^2(\mu_\bX)}^2\big)^{1/2}  \frac{2 (\mathbb{E}\|\varepsilon\|_2^2)^{1/2}}{n^{1/2}} \\
    &\quad + \frac{4\lambda_d}{n\lambda_t} C_\Omega^2 d_2 \mathbb E\|\varepsilon\|_2^2 +  \frac{\lambda_d^2}{n^{3/2} \lambda_t^2}  C_\Omega^2 d_2^{3/2} \mathbb E\|\varepsilon\|_2^2 .
\end{align*}
Hence, using elementary algebra,
\[\big(\mathbb{E}\|\hat u_n^\mathrm{(noise)}\|_{L^2(\mu_\bX)}^2\big)^{1/2} \leqslant \frac{(\mathbb{E}\|\varepsilon\|_2^2)^{1/2}}{n^{1/2}}\Big(2+2C_\Omega d_2^{3/4}\Big(\frac{\lambda_d^{1/2}}{\lambda_t^{1/2}} +\frac{\lambda_d}{\lambda_tn^{1/4}}\Big)\Big)\]
and 
\[\mathbb{E}\|\hat u_n^\mathrm{(noise)}\|_{L^2(\mu_\bX)}^2 \leqslant \frac{8\mathbb{E}\|\varepsilon\|_2^2}{n}\Big(1+C_\Omega d_2^{3/2}\Big(\frac{\lambda_d}{\lambda_t} +\frac{\lambda_d^2}{\lambda_t^2n^{1/2}}\Big)\Big).\]

\paragraph{Step 6: Putting everything together} Combining Steps 3, 4, and 5, we conclude that
\begin{align*}
    \mathbb{E}\|\hat u_n-u^\star\|^2_{L^2(\mu_\bX)} &\leqslant \frac{1}{\lambda_d}\big(\mathrm{PI}(u^\star) + \lambda_t\|u^\star\|_{H^{m+1}(\Omega)}^2\big)  + \frac{C_\Omega 'd_2^{1/2}}{n^{1/2}} \Big(2\|u^\star\|_{H^{m+1}(\Omega)}^2 + \frac{\mathrm{PI}(u^\star)}{\lambda_t}\Big)\\
    &\quad +\frac{8\mathbb{E}\|\varepsilon\|_2^2}{n}\Big(1+C_\Omega d_2^{3/2}\Big(\frac{\lambda_d}{\lambda_t} +\frac{\lambda_d^2}{\lambda_t^2n^{1/2}}\Big)\Big).
\end{align*}
\subsection*{Proof of Proposition \ref{thm:phyCst}}
By definition, $\hat u_n$ minimizes $\mathscr{R}_n^{(\mathrm{reg})}$ over $H^{m+1}(\Omega, \mathbb{R}^{d_2})$. So, $\mathscr{R}_n^{(\mathrm{reg})}(u^\star) \geqslant \mathscr{R}_n^{(\mathrm{reg})}(\hat u_n)$. 
Moreover, since 
\[
\|\tilde \Pi (\hat u_n)(\bX_i) - Y_i\|_2^2 = \|\tilde \Pi (\hat u_n-u^\star )(\bX_i)\|_2^2 -2 \langle \tilde \Pi (\hat u_n-u^\star )(\bX_i), \varepsilon_i\rangle + \|\varepsilon_i\|_2^2 ,\]
one has 
\begin{align*}
    &\frac{1}{n}\sum_{i=1}^n \|\tilde \Pi (\hat u_n)(\bX_i) - Y_i\|_2^2 \\
    &\quad \geqslant  -2 \|\hat u_n-u^\star\|_{H^{m+1}(\Omega)}\times \sup_{\|u\|_{H^{m+1}(\Omega)}\leqslant 1} \frac{1}{n} \sum_{j=1}^n \langle \tilde \Pi (u)(\bX_j) - \mathbb{E}(\tilde \Pi (u)(\bX)),\varepsilon_j\rangle\\
    &\qquad -2 \Big\langle \int_\Omega \tilde \Pi (\hat u_n-u^\star )d\mu_\bX, \frac{1}{n}\sum_{i=1}^n \varepsilon_i\Big\rangle + \frac{1}{n}\sum_{i=1}^n \|\varepsilon_i\|_2^2.
\end{align*}
Thus,
\begin{align}
    &\frac{1}{n}\sum_{i=1}^n \|\tilde \Pi (\hat u_n)(\bX_i) - Y_i\|_2^2 \nonumber\\
    &\quad \geqslant -2 (\|\hat u_n\|_{H^{m+1}(\Omega)}+\|u^\star\|_{H^{m+1}(\Omega)}) \sup_{\|u\|_{H^{m+1}(\Omega)}\leqslant 1} \frac{1}{n} \sum_{j=1}^n \langle \tilde \Pi (u)(\bX_j) - \mathbb{E}(\tilde \Pi (u)(\bX)),\varepsilon_j\rangle \nonumber\\
    &\qquad -2 \Big\langle \int_\Omega \tilde \Pi (\hat u_n-u^\star )d\mu_\bX, \frac{1}{n}\sum_{i=1}^n \varepsilon_i\Big\rangle + \frac{1}{n}\sum_{i=1}^n \|\varepsilon_i\|_2^2. \label{eq:phyConsistencyIneg}
\end{align}
Recall from Steps 4 and 5 of the proof of Theorem \ref{prop:consistencey} that
\begin{align*}
    \mathbb{E}\|\hat u_n\|_{H^{m+1}(\Omega)}^2 
    &\leqslant 2\mathbb{E}\|\hat u_n^\star\|_{H^{m+1}(\Omega)}^2 + 2\mathbb{E}\|\hat u_n^{\mathrm{(noise)}}\|_{H^{m+1}(\Omega)}^2\\
    & \leqslant 2\Big(\frac{\mathrm{PI}(u^\star)}{\lambda_t} + \|u^\star\|_{H^{m+1}(\Omega)}^2\Big) + \frac{8\lambda_d^2}{n\lambda_t^2} C_\Omega^2 d_2 \mathbb E\|\varepsilon\|_2^2 
\end{align*}
Therefore, Lemma \ref{lem:empiricalProcess} and the inequality $\mathbb E(XY)^2 \leqslant \mathbb E(X)^2 \mathbb E(Y)^2 $ show that 
\begin{align*}
    &\mathbb{E}\Big( \|\hat u_n\|_{H^{m+1}(\Omega)} \sup_{\|u\|_{H^{m+1}(\Omega)}\leqslant 1} \frac{1}{n} \sum_{j=1}^n \langle \tilde \Pi (u)(\bX_j) - \mathbb{E}(\tilde \Pi (u)(\bX)),\varepsilon_j\rangle \Big) = \Oequivalent_{n\to \infty}\Big(\frac{\lambda_d}{n\lambda_t}\Big). 
\end{align*}
By Theorem \ref{prop:consistencey},
\begin{align*}
    \mathbb{E}\Big|\Big\langle \int_\Omega \tilde \Pi (\hat u_n-u^\star )d\mu_\bX, \frac{1}{n}\sum_{i=1}^n \varepsilon_i\Big\rangle \Big| &\leqslant \big(\mathbb{E}\|u^\star-\hat u_n\|_{L^2(\mu_\bX)}^2\big)^{1/2} \frac{\mathbb{E}\|\varepsilon\|_2^2}{n^{1/2}}= \Oequivalent_{n\to \infty}\Big(\frac{\lambda_d}{n^{2}\lambda_t}\Big)^{1/2}.
\end{align*}
Combining these three results with \eqref{eq:phyConsistencyIneg}, we conclude that
\begin{align*}
    \mathbb{E}\Big(\frac{1}{n}\sum_{i=1}^n \|\tilde \Pi (\hat u_n)(\bX_i) - Y_i\|_2^2\Big) \geqslant \mathbb{E}\|\varepsilon\|_2^2 + \Oequivalent_{n\to \infty}\Big(\frac{\lambda_d}{n\lambda_t}\Big).
\end{align*}
Therefore, since $\lim_{n \to \infty }\frac{\lambda_d^2}{n\lambda_t} = 0$ and since $\mathscr{R}_n^{(\mathrm{reg})}(\hat u_n) = \frac{\lambda_d}{n}\sum_{i=1}^n \|\tilde \Pi (\hat u_n)(\bX_i) - Y_i\|_2^2+\mathrm{PI}(\hat u_n) + \lambda_t \|\hat u_n\|_{H^{m+1}(\Omega)}^2$,
\[\mathbb{E}\big(\mathscr{R}_n^{(\mathrm{reg})}(\hat u_n)\big) \geqslant  \lambda_d \mathbb{E}\|\varepsilon\|_2^2 + \mathbb{E}(\mathrm{PI}(\hat u_n)) + \oequivalent_{n\to \infty}(1).\]
Similarly, almost everywhere,
\[\frac{1}{n}\sum_{i=1}^n \|\tilde \Pi (\hat u^\star)(\bX_i) - Y_i\|_2^2 = \frac{1}{n}\sum_{i=1}^n \|\varepsilon_i\|_2^2.\]
Hence, 
\[\mathbb{E}\big(\mathscr{R}_n^{(\mathrm{reg})}(u^\star)\big) = \lambda_d \mathbb{E}\|\varepsilon\|_2^2 + \mathrm{PI}( u^\star) + \lambda_t \|u^\star\|_{H^{m+1}(\Omega)}^2.\]
Since $ \mathbb{E}(\mathscr{R}_n^{(\mathrm{reg})}(\hat u_n)) \leqslant \mathbb{E}(\mathscr{R}_n^{(\mathrm{reg})}(u^\star))$ and since $\lambda_t \to 0$, we are led to 
\[\mathbb{E}(\mathrm{PI}(\hat u_n)) \leqslant \mathrm{PI}(u^\star) + \oequivalent_{n\to \infty}(1),\]
which is the desired result.

\renewcommand\thesection{\thechapter.\arabic{section}}

\chapter{Physics-informed machine learning as a kernel method}
\label{ch:requirements}
This chapter corresponds to the following publication: \citet{doumeche2024physicsinformed}.

\section{Introduction}

\paragraph{Physics-informed machine learning.} Physics-informed machine learning (PIML) refers to a subdomain of machine learning that combines physical knowledge and empirical data to enhance performance of tasks involving a physical mechanism. Following the influential work of \citet{raissi2019PINN}, the field has experienced a notable surge in popularity, largely driven by scientific computing and engineering applications. We refer the reader to the surveys by \citet{rai2020review}, \citet{karniadakis2021piml}, \citet{cuomo2022scientific}, and \citet{Hao2022review}.
In a nutshell, the success of PIML relies on the smart interaction between machine learning and physics. In its most standard form, this achievement is realized by integrating physical equations into the loss function. Three common use cases include solving systems of partial differential equations (PDEs), addressing inverse problems (e.g., learning the PDE governing an observed phenomenon), and further improving the statistical performance of empirical risk minimization. 
This article focuses on the latter approach, known as hybrid modeling \citep[e.g.,][]{rai2020review}. 

\paragraph{Hybrid modeling.} Consider the classical regression model $Y = f^\star(X) + \varepsilon$, where the function $f^\star: \mathbb R^d \to \mathbb R$ is unknown. The random variable $Y \in \mathbb{R}$ is the target, the random variable $X\in \Omega \subseteq [-L,L]^d$ the vector of features, and $\varepsilon$ a random noise. Given a sample $\{(X_1, Y_1), \hdots, (X_n, Y_n)\}$ of i.i.d.~copies of $(X,Y)$, the goal is to construct an estimator $\hat f_n$ of $f^\star$ based on these $n$ observations. The distinctive element of PIML is the inclusion of a prior on $f^\star$, asserting its compliance with a known PDE. Therefore, it is assumed that $f^\star$ is at least weakly differentiable, belonging to the Sobolev space $H^s(\Omega)$ for some integer $s > d/2$, and that there is a known differential operator $\mathscr{D}$ such that $\mathscr{D}(f^\star) \simeq 0$. For instance, if the desired solution $f^\star$ is intended to conform to the wave equation, then $\mathscr{D}(f)(x,t) = \partial^2_{t,t} f(x,t) - \partial^2_{x,x} f(x,t)$ for $(x,t)\in \Omega$. Overall, we are interested in the minimizer of the empirical risk function 
\begin{equation}
    R_n(f) = \frac{1}{n}\sum_{i=1}^n |f(X_i) - Y_i|^2 + \lambda_n \|f\|_{H^s_{\mathrm{per}}([-2L, 2L]^d)}^2 + \mu_n \|\mathscr{D}(f)\|_{L^2(\Omega)}^2 
    \label{eq:risk_function}
\end{equation} 
over the class $\mathscr{F} = H^s_{\mathrm{per}}([-2L, 2L]^d)$ of candidate functions, where $\lambda_n > 0$ and $\mu_n \geqslant 0$ are hyperparameters that weigh the relative importance of each term. We refer to the appendix for a precise definition of the periodic Sobolev space $H^s_{\mathrm{per}}([-2L, 2L]^d)$, as well as the continuous extension $H^s(\Omega) \hookrightarrow H^s_{\mathrm{per}}([-2L, 2L]^d)$. 
It is stressed that the $\|\cdot\|_{H^s_{\mathrm{per}}([-2L, 2L]^d)}$ norm is the standard $\|\cdot\|_{H^s([-2L, 2L]^d)}$ norm---the symbol ``$\mathrm{per}$'' highlights that we consider functions belonging to a \emph{periodic} Sobolev space.
The choice of the periodic Sobolev space $H^s_{\mathrm{per}}([-2L,2L]^d)$ is merely technical---the reader can be confident that all subsequent results remain applicable to the standard Sobolev space $H^s(\Omega)$, as will be stressed later.

The first term in \eqref{eq:risk_function} is the standard component of supervised learning, corresponding to a least-squares criterion that measures the prediction error over the training sample. The second term $\|f\|_{H^s_{\mathrm{per}}([-2L, 2L]^d)}^2$ corresponds to a Sobolev penalty for $s>d/2$, which enforces the regularity of the estimator. 
Finally, the $L^2$ penalty $\|\mathscr{D}(f)\|_{L^2(\Omega)}^2$ on $\Omega$ quantifies the physical inconsistency of $f$ with respect to the differential prior on $f^\star$: the more $f$ aligns with the PDE, the lower the value of $\|\mathscr{D}(f)\|_{L^2(\Omega)}^2$. It is this last term that marks the originality of the hybrid modeling problem.

In this context, beyond classical statistical analyses, an interesting question is to quantify the impact of the physical regularization $\|\mathscr{D}(f)\|_{L^2(\Omega)}^2$ on the empirical risk \eqref{eq:risk_function}, typically in terms of convergence rate of the resulting estimator. It is intuitively clear, for example, that if the target $f^{\star}$ satisfies $\mathscr{D}(f^\star) =0$ (i.e., $f^{\star}$ is a solution of the underlying PDE), then, under appropriate conditions, the estimator $\hat f_n$ should have better properties than a standard estimator of the empirical risk. This is the challenging problem that we address in this contribution.

\paragraph{Contributions.} 
We are interested in the statistical properties of the minimizer of \eqref{eq:risk_function} over the space $H^s_{\mathrm{per}}([-2L,2L]^d)$, denoted by
\begin{equation}
    \hat f_n = \mathop{\mathrm{argmin}}_{f \in H^s_{\mathrm{per}}([-2L, 2L]^d)}\; R_n(f).
    \label{eq:estimator_sob}
\end{equation}
We show in Section \ref{sec:kernel} that problem \eqref{eq:estimator_sob} can be formulated as a kernel regression task, with a kernel~$K$ that we specify. This allows us, in Section \ref{sec:bounds}, to use tools from kernel theory to determine an upper bound on the rate of convergence of $\hat f_n$ to $f^\star$ in $L^2(\Omega, \mathbb{P}_X)$, where $\mathbb{P}_X$ is the distribution of~$X$ on $\Omega$. In particular, this rate can be evaluated by bounding the eigenvalues of the integral operator associated with the kernel. 
The latter problem is studied in detail in Theorem \ref{eq:weak_pde}, where the corresponding eigenfunctions are characterized through a weak formulation. Overall, we show that~$\hat f_n$ converges to $f^\star$ \emph{at least} at the Sobolev minimax rate. The complete mechanics are illustrated in Section \ref{sec:experiment} for the  operator $\mathscr{D} = \frac{d}{dx}$ in dimension $d=1$, showcasing a simple but instructive case.
In such a setting, the convergence rate is shown to be 
\begin{align*}
        \mathbb{E}\int_{[-L,L]} |\hat f_n-f^\star|^2 d{\mathbb P}_X = &\|\mathscr{D}(f^\star)\|_{L^2(\Omega)}\;\mathcal{O}_n \big(  n^{-2/3}\log^3(n)\big) +\|f^\star\|_{H^s(\Omega)}^2\mathcal{O}_n \big( n^{-1} \log^3(n)\big) .
    \end{align*}
Thus, the lower the modeling error $\|\mathscr{D}(f^\star)\|_{L^2(\Omega)}$, the lower the estimation error. In particular, if $f^{\star}$ exactly satisfies the PDE, i.e., $\|\mathscr{D}(f^\star)\|_{L^2(\Omega)} = 0$, then the rate is $n^{-1}$ (up a to log factor), significantly better than the Sobolev rate of $n^{-2/3}$. This shows that the use of physical knowledge in the PIML framework has a quantifiable impact on the estimation error.

\section{Related works}  

\paragraph{Approximation classes and Sobolev spaces.} Since Sobolev spaces are often considered too expensive for practical implementation, various alternative classes of functions over which to minimize the empirical risk function \eqref{eq:risk_function} have been suggested in the literature. 
In the case of a second-order and coercive PDE in dimension $d=2$, and with  an additional prior on the boundary conditions, \citet{azzimonti2015blood}, \citet{arnone2022spatialRegression}, and \citet{ferraccioli2022some} propose finite-element-based methods to optimize the minimization over $H^2(\Omega)$. However, the most commonly used approach to minimize the risk functional involves neural networks, which leverage the backpropagation algorithm for efficient computation of successive derivatives and optimize \eqref{eq:risk_function} through gradient descent. The so-called PINNs (for physics-informed neural networks---\citealp{raissi2019PINN}) have been successfully applied to a diverse range of physical phenomena, including  sea temperature modeling \citep{bezenac2017processes},
image denoising \citep{wang2020superResolution},
turbulence \citep{wang2020turbulence},
blood streams \citep{arzani2021uncovering},
glacier dynamics \citep{riel2021glacier},
and heat transfers \citep{ramezankhani2022multifidelity}, among others.
The neural architecture of PINNs is often designed to be large \citep[e.g.,][]{arzani2021uncovering, krishnapriyan2021characterizing, xu2021extrapolation}, allowing it to approximate any function in $H^s(\Omega)$ \citep{ryck2021approximation, doumeche2023convergence}.

\paragraph{Sobolev regularization.} In the PIML literature, the Sobolev regularization is either directly implemented as such \citep{shin2020convergence, doumeche2023convergence} or in a more implicit manner, by assuming that the operator $\mathscr{D}$ is inherently regular (e.g., second-order elliptic, parabolic, or hyperbolic) and specifying boundary conditions \citep{azzimonti2015blood, shin2020convergence, arnone2022spatialRegression, ferraccioli2022some, wu2022convergence, mishra2022generalization, shin2023error}.  
It turns out, however, that the specific form taken by the Sobolev regularization is unimportant. This will be enlightened by our Theorem \ref{thm:eq_reg}, which shows that using equivalent Sobolev norms does not alter the convergence rate of the estimators. 
From a theoretical perspective, much of the literature delves into the properties of PINNs in the realm of PDE solvers, usually through the analysis of their generalization error \citep{shin2020convergence, ryck2022kolmogorov, wu2022convergence, doumeche2023convergence, mishra2022generalization, qian2023error, deryck2023operator, shin2023error}.
Overall, there are few theoretical guarantees available regarding hybrid modeling, with the exception of \citet{azzimonti2015blood}, \citet{shin2020convergence}, \citet{arnone2022spatialRegression}, and \citet{doumeche2023convergence}. 

\paragraph{PIML and kernels.} Other studies have revealed interesting connections between PIML and kernel methods. In noiseless scenarios, the use of kernel methods to construct meshless PDE solvers under the Sobolev regularity hypothesis has long been explored by, for example, \citet{schaback2006from}. Recently, \citet{batlle2023error} uncovered convergence rates under regularity assumptions on the differential operator equivalent to a Sobolev regularization. For inverse PIML problems, \citet{lu2022sobolev} and \citet{de2023convergence} take advantage of a kernel reformulation of PIML to establish convergence rates for differential operator learning. This generalizes results obtained by \citet{nickl2019convergence} using Bayesian inference methods. However, none of these works has specifically addressed hybrid modeling. To the best of our knowledge, the present study is the first to show that the  physical regularization term $\|\mathscr{D}(f)\|_{L^2(\Omega)}$ in the PIML loss \eqref{eq:risk_function} may lead to improved convergence rates.

\section{PIML as a kernel method}
\label{sec:kernel}
\begin{figure}
    \centering
    \includegraphics[width=0.8\linewidth]{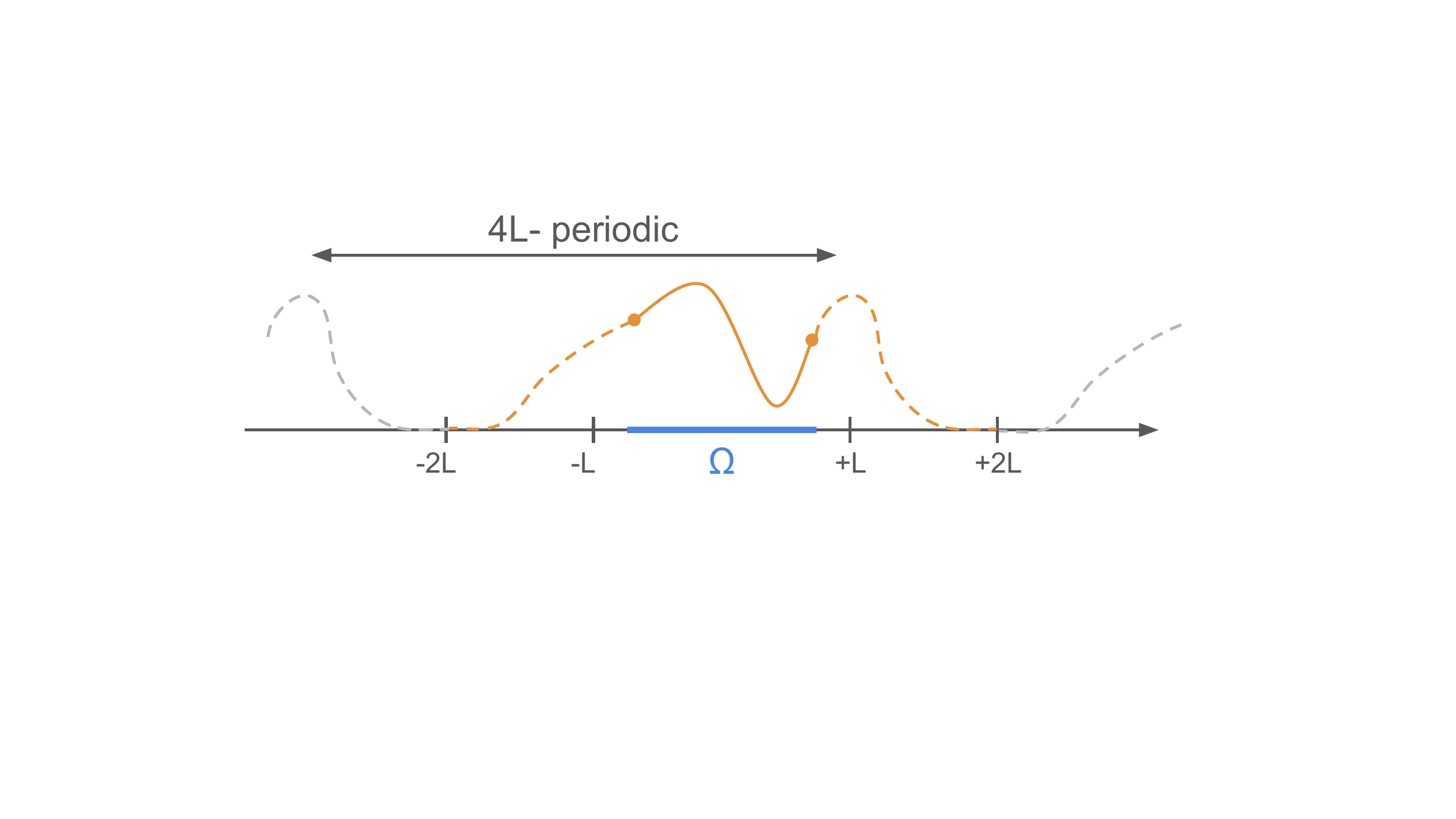}
    \caption{Illustration of a 4L-periodic extension of a function in $H^s(\Omega)$ to $H^s_{\mathrm{per}}([-2L,2L]^d)$ for $d=1$.}
    \label{fig:omega}
\end{figure}

Throughout the article, we let $\Omega \subseteq [-L, L]^d$ ($L >0$) be a bounded Lipschitz domain. Assuming that $\Omega$ is Lipschitz allows a high level of generality regarding its regularity, encompassing $C^1$-manifolds (such as the Euclidean ball $\{ x \in \mathbb{R}^d \; | \; \|x\|_2 \leqslant L\}$), as well as domains with non-differentiable boundaries (such as the hypercube $[-L, L]^d$). (A summary of the mathematical notation and functional analysis concepts used in this paper is to be found in Appendix \ref{sec:appA}.) 
The target function $f^\star: \mathbb R^d\to \mathbb R$ is assumed to belong to the Sobolev space $H^s(\Omega)$ for some positive integer $s>d/2$. Furthermore, this function is assumed to approximately satisfy a linear PDE on $\Omega$ (the coefficients of which are potentially non-constant) with derivatives of order less than or equal to $s$. In other words, one has $\mathscr{D}(f^\star) \simeq 0$ for some known operator $\mathscr D$ of the following form:
\begin{defi}[Linear differential operator]
\label{def:linearoperator}
    Let $s \in \mathbb{N}$. An operator $\mathscr{D}: H^s(\Omega) \to L^2(\Omega)$ is a linear differential operator if, for all $f \in H^s(\Omega)$,
    \[\mathscr{D}(f) = \sum_{|\alpha|\leqslant s} p_\alpha \partial^\alpha f,\]
    where  $p_\alpha: \Omega \to \mathbb{R}$ are \emph{functions} such that $\max_\alpha \|p_\alpha\|_\infty < \infty$. (By definition, $\{|\alpha|\leqslant s\}=\{\alpha\in \mathbb{N}^d \;|\; \|\alpha\|_1 \leqslant s\}$ and $\|\cdot\|_{\infty}$ stands for the supremum norm of functions.) 
\end{defi}
Given $s$, the linear differential operator $\mathscr D$, and a training sample $\{(X_1, Y_1), \hdots, (X_n, Y_n)\}$, we consider the estimator $\hat f_n$ that minimizes the regularized empirical risk  \eqref{eq:risk_function} over the periodic Sobolev space $H^s_{\mathrm{per}}([-2L, 2L]^d)$. Recall that $H^s_{\mathrm{per}}([-2L, 2L]^d)$ is the subspace of $H^s([-2L, 2L]^d)$ consisting of functions whose $4L$-periodic extension is still $s$-times weakly differentiable. 
The important point to keep in mind is that any function of $ H^s(\Omega)$ can be extended to a function in $H^s_{\mathrm{per}}([-2L, 2L]^d)$ (see Proposition \ref{prop:dec_four_lip} in the appendix), which makes it equivalent to suppose that $f^\star \in H^s(\Omega)$ or $f^\star \in H^s_{\mathrm{per}}([-2L,2L]^d)$, as shown in Theorem \ref{thm:eq_reg}. The extension mechanism is illustrated in Figure \ref{fig:omega}.  

The key step to turn the minimization of \eqref{eq:risk_function} into a kernel method is to observe that any function $f\in H^s_{\mathrm{per}}([-2L,2L]^d)$ can be linearly mapped in $L^2([-2L,2L]^d)$ in such a way that the norm $\|\cdot\|_{L^2([-2L,2L]^d)}$ of the embedding is equal to $\lambda_n \|f\|_{H^s_{\mathrm{per}}([-2L, 2L]^d)}^2 + \mu_n \|\mathscr{D}(f)\|_{L^2(\Omega)}^2$, i.e., the regularization term of \eqref{eq:risk_function}. Proposition \ref{prop:diff_op_main} below shows that this embedding takes the form of the inverse square root of a positive diagonalizable operator $\mathscr{O}_n$.
\begin{prop}[Differential operator]
    There exists a positive operator $\mathscr{O}_n$ on $L^2([-2L,2L]^d)$ such that $\mathscr{O}_n^{-1/2}: H^s_{\mathrm{per}}([-2L, 2L]^d) \to L^2([-2L,2L]^d)$ is well-defined and satisfies, for any $f \in H^s_{\mathrm{per}}([-2L,2L]^d)$, \[\|\mathscr{O}_n^{-1/2} (f)\|_{L^2([-2L,2L]^d)}^2 =  \lambda_n \|f\|_{H^s_{\mathrm{per}}([-2L, 2L]^d)}^2 + \mu_n \|\mathscr{D}(f)\|_{L^2(\Omega)}^2 .\]
    Moreover, there is an orthonormal basis of eigenfunctions $v_m \in H^s_{\mathrm{per}}([-2L,2L]^d)$ of $\mathscr{O}_n$ associated with eigenvalues $a_m > 0$ such that, for any $f \in L^2([-2L,2L]^d)$,
    \[\forall x \in [-2L,2L]^d, \quad \mathscr{O}_n(f)(x) = \sum_{m\in \mathbb{N}} a_m \langle f,  v_m\rangle_{L^2([-2L,2L]^d)} v_m(x).\]
    \label{prop:diff_op_main}
\end{prop}
Denote by $\delta_x$ the Dirac distribution at $x$. Informally, the properties of the embedding $\mathscr{O}_n^{-1/2}: H^s_{\mathrm{per}}([-2L, 2L]^d)\to L^2([-2L,2L]^d)$ in Proposition \ref{prop:diff_op_main} suggest that something like
\[f(x) \; ``="\; \langle f, \delta_x\rangle = \langle \mathscr{O}_n^{-1/2}(f), \mathscr{O}_n^{1/2}(\delta_x)\rangle_{L^2([-2L,2L]^d)}\] 
should be true. In other terms, still informally, we may write $f(x) = \langle z, \psi(x)\rangle_{L^2([-2L,2L]^d)}$, 
with $z = \mathscr{O}_n^{-1/2}(f)$, $\psi(x) = \mathscr{O}_n^{1/2}(\delta_x)$, and $\|z\|_{L^2([-2L,2L]^d)}^2 =   \lambda_n \|f\|_{H^s_{\mathrm{per}}([-2L, 2L]^d)}^2 + \mu_n \|\mathscr{D}(f)\|_{L^2(\Omega)}^2$. 
We recognize a reproducing property, turning $\psi$ into a kernel embedding associated with the risk~\eqref{eq:risk_function}. 
This mechanism is formalized in the following theorem. 
\begin{thm}[Kernel of linear PDEs]
    Assume that $s>d/2$, and let $\lambda_n >0, \mu_n \geqslant 0$. Let $a_m$ and $v_m$ be the eigenvalues and eigenfunctions of $\mathscr{O}_n$.
    Then the space $H^s_{\mathrm{per}}([-2L,2L]^d)$, equipped with the inner product $\langle f, g\rangle_{\mathrm{RKHS}} = \langle \mathscr{O}_n^{-1/2}f,  \mathscr{O}_n^{-1/2}g\rangle_{L^2([-2L,2L]^d)}$, is a reproducing kernel Hilbert space. 
    In particular, 
    \begin{enumerate}
        \item[$(i)$] The kernel $K: [-2L, 2L]^d \times [-2L, 2L]^d \to \mathbb{R}$ is defined by 
        \[K(x,y) = \sum_{m\in \mathbb{N}} a_m  v_m(x)v_m(y). \]
        \item[$(ii)$] For all $x \in [-2L, 2L]^d$, $K(x, \cdot) \in H^s_{\mathrm{per}}([-2L,2L]^d)$.
        \item[$(iii)$] For all $f \in H^s_{\mathrm{per}}([-2L, 2L]^d)$, \[\forall x \in [-2L,2L]^d, \quad f(x) = \langle f, K(x, \cdot)\rangle_{\mathrm{RKHS}}.\]
        \item[$(iv)$] For all $f\in H^s_{\mathrm{per}}([-2L, 2L]^d)$, 
        \[\|f\|^2_{\mathrm{RKHS}} = \lambda_n \|f\|_{H^s_{\mathrm{per}}([-2L, 2L]^d)}^2 + \mu_n \|\mathscr{D}(f)\|_{L^2(\Omega)}^2.\]
    \end{enumerate}
    \label{thm:PDE_kernel}
\end{thm}
\begin{proof}[{\bf sketch}]
    The complete proof is given in Appendix \ref{app:diif_of}.
    Only a rough sketch is given here by 
    examining the simplified case where  $L=\pi/2$, $\Omega = [-\pi,\pi]^d =[-2L,2L]^d$, 
    and $\mathscr{D}$ has constant coefficients. 
    This means that we consider functions with periodic derivatives on $\Omega$, penalized by the PDE on the whole domain $[-\pi,\pi]^d$. It turns out that, in this case, the corresponding operator $\mathscr{O}_n$, satisfying
    \[\| \mathscr{O}_n^{-1/2} (f)\|_{L^2([-\pi,\pi]^d)}^2 =  \lambda_n \|f\|_{H^s_{\mathrm{per}}([-\pi, \pi]^d)}^2 + \mu_n \|\mathscr{D}(f)\|_{L^2([-\pi,\pi]^d)}^2 := \|f\|^2_{\mathrm{RKHS}}\]
    has an explicit form. To see this, denote by $\mathrm{FS}$ the Fourier series operator. By the Parseval's theorem, for any frequency $k \in \mathbb{Z}^d$, one has $\mathrm{FS}( {\mathscr{O}}_n^{-1/2} (f))(k) = \sqrt{a_k} \; \mathrm{FS}(f)(k)$,
    where 
    \[a_k = \lambda_n \sum_{|\alpha|\leqslant s}   \prod_{j=1}^dk_j^{2\alpha_j} + \mu_n \Big(\sum_{|\alpha|\leqslant s}  p_\alpha \prod_{j=1}^dk_j^{\alpha_j}\Big)^2.\]
    Accordingly, $ {\mathscr{O}}_n$ is diagonalizable with eigenfunctions $v_k: x \mapsto \exp(i\langle k, x\rangle)$ associated with the eigenvalues $a_k^{-1}$. Next, using the Fourier decomposition of $f$, we have, for all $x \in [-\pi,\pi]^d$, 
    \[f(x) = \sum_{k\in \mathbb{Z}^d} \mathrm{FS}(f)(k) \exp(i \langle k, x\rangle) = \sum_{k\in \mathbb{Z}^d} \mathrm{FS}( {\mathscr{O}}_n^{-1/2} (f))(k) a_k^{-1/2}\exp(i \langle k, x\rangle).\]   
    Since $a_k^{-1} \leqslant \lambda_n^{-1} (\sum_{|\alpha|\leqslant s}   \prod_{j=1}^dk_j^{2\alpha_j})^{-1}$, it is easy to check that $\sum_{k \in \mathbb{Z}^d} a_k^{-1} < \infty$ and that the function $\psi_x$ such that $\mathrm{FS}(\psi_x)(k) = a_k^{-1/2} \exp(i \langle k, x\rangle)$ belongs to $H^s_{\mathrm{per}}([-\pi, \pi]^d)$. We therefore have the kernel formulation $f(x) = \langle  {\mathscr{O}}_n^{-1/2} (f), \psi_x\rangle_{L^2([-\pi,\pi]^d)}$, where $\| {\mathscr{O}}_n^{-1/2} (f)\|_{L^2([-\pi,\pi]^d)}^2 = \|f\|^2_{\mathrm{RKHS}}$. The corresponding kernel is then defined by \[K(x,y) = \langle \psi_x, \psi_y\rangle_{L^2([-\pi,\pi]^d)} = \sum_{k \in \mathbb{Z}^d} a_k v_k(x)\bar v_k(y).\] 
    The complete proof of Theorem \ref{thm:PDE_kernel} is more technical because, in our case $\Omega \subsetneq [-2L,2L]^d$ and $\mathscr{D}$ may have non-constant coefficients. Thus, the operator $\mathscr{O}_n$ is not diagonal in the Fourier space. To characterize its eigenvalues $a_m$ and eigenfunctions $v_m$, we resort to classical results of PDE theory building upon functional analysis.
\end{proof}
The message of Theorem \ref{thm:PDE_kernel} is that minimizing the empirical risk \eqref{eq:risk_function} can be cast as a kernel method associated with the regularization $\lambda_n \|f\|_{H^s_{\mathrm{per}}([-2L, 2L]^d)}^2 + \mu_n \|\mathscr{D}(f)\|_{L^2(\Omega)}^2$. In other words, \eqref{eq:risk_function} can be rewritten as
    \[\hat f_n = \mathop{\mathrm{argmin}}_{f \in H^s_{\mathrm{per}}([-2L, 2L]^d)}\; \frac{1}{n}\sum_{i=1}^n |f(X_i) - Y_i|^2 + \|f\|^2_{\mathrm{RKHS}}.\]
This result is interesting in itself because it fundamentally shows that a PIML estimator (and therefore its variants implemented in practice, such as PINNs) can be regarded as a kernel estimator. 
Note however that computing $K(x,y)$ is not always straightforward and may require the use of numerical techniques. This kernel is characterized by the following weak formulation.
\begin{prop}[Kernel characterization]
    The kernel $K$ is the unique solution to the following weak formulation, valid for all test functions $\phi \in H^s_{\mathrm{per}}([-2L,2L]^d)$, \[\forall x \in \Omega,\quad \lambda_n \sum_{|\alpha|\leqslant s} \int_{[-2L, 2L]^d}\partial^\alpha K(x, \cdot)\; \partial^\alpha \phi + \mu_n \int_{\Omega}\mathscr{D}(K(x, \cdot)) \; \mathscr{D}(\phi) = \phi(x).\]
    \label{prop:kernel_characterization}
\end{prop}
Regardless of the analytical computation of $K$, formulating the problem as a minimization in a reproducing kernel Hilbert space provides a way to  
quantify the impact of the physical regularization on the estimator's convergence rate, which is our primary goal. 

\section{Convergence rates}
\label{sec:bounds}
   The results of the previous section allow us to draw on the existing literature on kernel learning to gain a deeper understanding of the properties of the estimator $\hat f_n$ and the influence of the operator~$\mathscr{D}$ on the convergence rate. 
   
\subsection*{Eigenvalues of the integral operator}
The convergence rate of $\hat f_n$ to $f^\star$ is determined by the decay speed of the eigenvalues of the so-called integral operator $L_K: L^2(\Omega, \mathbb{P}_X) \to L^2(\Omega, \mathbb{P}_X),$ defined by
\[\forall f \in L^2(\Omega, \mathbb{P}_X), \forall x \in \Omega, \quad L_Kf(x) = \int_{\Omega} K(x,y) f(y) d\mathbb{P}_X(y),\]
where $\mathbb{P}_X$ is the distribution of $X$ on $\Omega$ \citep[e.g.,][]{caponnetto2007optimal}. Note that the integral in the definition of $L_K$ could also have been taken over $[-2L,2L]^d$ because the support of $\mathbb{P}_X$ is included in $\bar \Omega$.
However, finding the eigenvalues of $L_K$ is not an easy task---even when $X$ is uniformly distributed on $\Omega$---, not to mention the fact that $\mathbb{P}_X$ is usually unknown in real applications.
Nevertheless, we show in Theorem \ref{thm:eigenvalues} that these eigenvalues can be bounded by the eigenvalues of the operator $C\mathscr{O}_nC$, where $C$ is the projection on $\Omega$ defined below. Importantly, $C\mathscr{O}_nC$ no longer depends on $\mathbb{P}_X$. Moreover, its non-zero eigenvalues are characterized by a 
weak formulation, as we will see in Theorem \ref{prop:eigenfunction}. 
\begin{defi}[Projection on $\Omega$]
    Let $C$ be the operator on $L^2([-2L, 2L]^d)$ defined by $C f = f \mathbf{1}_\Omega$. 
    Then $C^2 = C$, i.e., $C$ is a projector, and 
    \[\langle f, C(g)\rangle_{L^2([-2L, 2L]^d)} = \int_{[-2L, 2L]^d} f g \mathbf{1}_\Omega =  \langle C(f), g\rangle_{L^2([-2L, 2L]^d)},\]
    i.e., $C$ is self-adjoint.
\end{defi} 
As for now, it is assumed that the distribution $\mathbb{P}_X$ of $X$ has a density $\frac{d\mathbb{P}_X}{dx}$ with respect to the Lebesgue measure on $\Omega$.
\begin{thm}[Kernels and eigenvalues]
\label{thm:eigenvalues}
     Let $K: [-2L, 2L]^d \times [-2L, 2L]^d \to \mathbb{R}$ be the kernel of Theorem \ref{thm:PDE_kernel}. Assume that there exists $\kappa>0$ such that $\frac{d\mathbb{P}_X}{dx} \leqslant \kappa$. Then the eigenvalues $a_m(L_K)$ of $L_K$ are bounded by the eigenvalues $a_m(C\mathscr{O}_nC)$ of $ C\mathscr{O}_nC$ on $L^2([-2L, 2L]^d)$ in such a way that $a_m(L_K) \leqslant \kappa a_m(C\mathscr{O}_nC)$.
\end{thm} 

\subsection*{Effective dimension and convergence rate} 
We will see in the next subsection how to compute the eigenvalues of $C\mathscr{O}_nC$. Yet, assuming we have them at hand, it is then possible to obtain a bound on the rate of convergence of $\hat f_n$ to $f^\star$ by bounding the so-called effective dimension of the kernel \citep{caponnetto2007optimal}, defined by
\[
\mathscr{N}(\lambda_n, \mu_n) = \mathrm{tr}(L_{K}  (\mathrm{Id} + L_{K})^{-1}),
\]
where $\mathrm{Id}$ is the identity operator, i.e., $\mathrm{Id}(f) = f$, and the symbol $\mathrm{tr}$ stands for the trace, i.e., the sum of the eigenvalues. Lemma \ref{lem:eff_dim} in the appendix shows that, whenever $\frac{d\mathbb{P}_X}{dx} \leqslant \kappa$, 
\begin{equation}
        \mathscr{N}(\lambda_n, \mu_n) 
    \leqslant \sum_{m \in \mathbb{N}} \frac{ 1}{1+(\kappa a_m(C\mathscr{O}_nC))^{-1}}.
    \label{eq:eff_dim_vp}
\end{equation}
Putting all the pieces together, we have the following theorem, which bounds the estimation error between $\hat f_n$ and $f^\star$.
\begin{thm}[Convergence rate]
    Assume that $s>d/2$, $f^\star \in H^s(\Omega)$, $\frac{d\mathbb{P}_X}{dx} \leqslant \kappa$ for some $\kappa > 0$, $\mu_n \geqslant 0$, $\lim_{n\to \infty}\lambda_n = \lim_{n\to \infty} \mu_n = \lim_{n\to \infty} \lambda_n / \mu_n = 0$, $\lambda_n \geqslant n^{-1}$, and  $\mathscr{N}(\lambda_n, \mu_n) \lambda_n^{-1} = o_n (n)$.
    Assume, in addition, that, for some $\sigma > 0$ and $M > 0$, the noise $\varepsilon$ satisfies 
    \begin{equation}
        \forall \ell \in \mathbb{N}, \quad \mathbb{E}(|\varepsilon|^\ell\; | \; X) \leqslant \frac{1}{2}\ell !\; \sigma^2\; M^{\ell-2}.
        \label{eq:conditionNoise}
    \end{equation}
    Then, for some constant $C_4 >0$ and $n$ large enough, 
    \begin{align*}
        &\mathbb{E}\int_\Omega |\hat f_n-f^\star|^2 d{\mathbb P}_X \\
        &\quad \leqslant C_4 \log^2(n)\Big(\lambda_n \|f^\star\|_{H^s(\Omega)}^2 + \mu_n \|\mathscr{D}(f^\star)\|_{L^2(\Omega)}^2 + \frac{M^2}{n^2 \lambda_n} + \frac{\sigma^2\mathscr{N}(\lambda_n, \mu_n)}{n}\Big).
    \end{align*}
\label{thm:boundexp}
\end{thm}
The sub-Gamma assumption \eqref{eq:conditionNoise} on the noise $\varepsilon$ is quite general and is satisfied in particular when $\varepsilon$ is bounded (possibly depending on $X$), or when $\varepsilon$ is Gaussian and independent of $X$ \citep[e.g.,][Theorem 2.10]{boucheron2013concentration}. We stress that the result of Theorem \ref{thm:boundexp} is general and holds regardless of the form of the linear differential operator $\mathscr{D}$. 
A simple bound on $\mathscr{N}(\lambda_n, \mu_n)$, neglecting the dependence in $\mathscr{D}$, allows to show that the PIML estimator converges at least at the Sobolev minimax rate over the class $H^s(\Omega)$.
\begin{prop}[Minimum rate]
    Suppose that the assumptions of Theorem \ref{eq:conditionNoise} are verified, and let
    $\lambda_n = n^{-2s/(2s+d)} \sqrt{\log(n)}$ and $\mu_n = \lambda_n \sqrt{\log(n)}$.
    Then the estimator $\hat f_n$ converges at a rate at least larger than the Sobolev minimax rate, up to a $\log$ term, i.e., 
    \begin{align*}
        \mathbb{E}\int_\Omega |\hat f_n-f^\star|^2 d{\mathbb P}_X = \mathcal{O}_n \big(  n^{-2s/(2s+d)}\log^3(n)\big).
    \end{align*}
    \label{prop:minSpeed}
\end{prop}
 However, this is only an upper bound, and we expect situations where $\hat f_n$ has a faster convergence rate thanks to the inclusion of the physical penalty $\| \mathscr{D}(f)\|_{L^2(\Omega)}$. 
 Such an improvement will depend on the magnitude of the modeling error $\|\mathscr{D}(f^{\star})\|_{L^2(\Omega)}$ and on the effective dimension $\mathscr{N}(\lambda_n,\mu_n)$. To achieve this goal, the eigenvalues $a_m$ of $C\mathscr{O}_nC$ must be characterized and then plugged into inequality \eqref{eq:eff_dim_vp}.
 This is the problem addressed in the next subsection. 

\subsection*{Characterizing the eigenvalues}
\label{sec:effectiveDim}
The goal of this section is to specify the spectrum of $C\mathscr{O}_n C$. 
It is worth noting that $\ker(C\mathscr{O}_n C)$ is not empty, as it encompasses every smooth function with compact support in $]\! -\! 2L, 2L[^d \backslash \bar\Omega$. 
The next theorem characterizes the eigenfunctions associated with non-zero eigenvalues and shows that they are in fact smooth functions on $\Omega$ and $(\bar \Omega)^c$ satisfying two PDEs.
\begin{thm}[Eigenfunction characterization]
    Assume that $s>d/2$ and that the functions $p_\alpha$ in Definition \ref{def:linearoperator} belong to $C^\infty(\Omega)$. Let $a_m >0$ be a positive eigenvalue of the  operator $C\mathscr{O}_n C$. Then the corresponding eigenfunction $v_m$ satisfies $v_m = a_m^{-1}C w_m$, where $w_m \in H^{s}_{\mathrm{per}}([-2L, 2L]^d)$. Moreover, for any test function $\phi \in H^s_{\mathrm{per}}([-2L,2L]^d)$,
    \begin{equation}
        \lambda_n \sum_{|\alpha|\leqslant s} \int_{[-2L, 2L]^d}\partial^\alpha w_m\; \partial^\alpha \phi + \mu_n \int_{\Omega}\mathscr{D}(w_m) \; \mathscr{D}(\phi) 
    = a_m^{-1} \int_{\Omega} w_m \phi.
    \label{eq:weak_pde}
    \end{equation} 
    In particular, 
     any solution of the weak formulation \eqref{eq:weak_pde} satisfies the following PDE system:
    \begin{itemize}
        \item[$(i)$] $w_m \in C^\infty(\Omega)$ and 
        \[\forall x \in \Omega, \quad \lambda_n\sum_{|\alpha|\leqslant s}(-1)^{|\alpha|} \partial^{2\alpha} w_m(x) + \mu_n \mathscr{D}^\ast \mathscr{D} w_m(x) = a_m^{-1} w_m(x),\]
        where $\mathscr{D}^\ast(f) := \sum_{|\alpha|\leqslant s} (-1)^{|\alpha|}\partial^\alpha (p_\alpha  f)$ is the adjoint operator of $\mathscr{D}$.
        \item[$(ii)$]  $w_m \in C^\infty([-2L, 2L]^d \backslash \bar\Omega)$ and 
        \[\forall x \in [-2L, 2L]^d \backslash \bar \Omega,\quad  \sum_{|\alpha|\leqslant s}(-1)^{|\alpha|} \partial^{2\alpha} w_m(x)  = 0.\]
    \end{itemize}
    Notice that $w_m$ might be irregular on the boundary $\partial \Omega$, but only there.
    \label{prop:eigenfunction}
\end{thm}
Theorem \ref{prop:eigenfunction} is important insofar as it allows to characterize the positive eigenvalues $a_m$ of the operator $C \mathscr{O}_n C$. Indeed, these eigenvalues are the only real numbers such that the weak formulation  \eqref{eq:weak_pde} admits a solution.
This weak formulation has to be solved in a case-by-case study, given the differential operator $\mathscr{D}$. As an illustration, an example is presented in the next section with $\mathscr{D}=\frac{d}{dx}$. 

\subsection*{The choice of Sobolev regularization is unimportant}
So far, we have considered problem \eqref{eq:risk_function} with the Sobolev regularization $\| f\|_{H^s_{\mathrm{per}}([-2L,2L]^d)}^2$. However, other choices of Sobolev norms, such as $\| f\|_{H^s(\Omega)}^2$, are also possible. Fortunately, this choice does not affect the effective dimension $\mathscr{N}(\lambda_n, \mu_n)$, and thus the convergence rate in Theorem \ref{thm:boundexp}. 
\begin{thm}[Equivalent regularities and effective dimension]
    Assume that $s > d/2$. Then the following three estimators correspond each to a kernel learning problem:
    \begin{align*}
        \hat f_n^{(1)} &= \mathop{\mathrm{argmin}}_{f \in H^s(\Omega)} \frac{1}{n}\sum_{i=1}^n |f(X_i) -Y_i|^2  + \lambda_n \| f\|_{H^s(\Omega)}^2 + \mu_n \| \mathscr{D}(f)\|_{L^2(\Omega)}^2,\\
   \hat f_n^{(2)} &= \mathop{\mathrm{argmin}}_{f \in H^s_{\mathrm{per}}([-2L,2L]^d)} \frac{1}{n}\sum_{i=1}^n |f(X_i) -Y_i|^2  + \lambda_n  \| f\|_{H^s_{\mathrm{per}}([-2L,2L]^d)}^2 + \mu_n \| \mathscr{D}(f)\|_{L^2(\Omega)}^2,\\
   \hat f_n^{(3)} &= \mathop{\mathrm{argmin}}_{f \in H^s(\Omega)} \frac{1}{n}\sum_{i=1}^n |f(X_i) -Y_i|^2 + \lambda_n  \| f\|^2 + \mu_n \| \mathscr{D}(f)\|_{L^2(\Omega)}^2,
    \end{align*}
    where $\| \cdot \|$ is any of the equivalent Sobolev norms.
    \label{thm:eq_reg}
     Moreover, these three estimators share equivalent effective dimensions ${\mathscr{N}(\lambda_n, \mu_n)}$. Accordingly, they share the same upper bound on the convergence rate given by Theorem \ref{thm:boundexp}.
\end{thm}
The incorporation of a Sobolev regularization in the empirical risk function is needed to guarantee that $\hat f_n$ has good statistical properties. For example, even with the simplest PDEs, the minimizer of $\frac{1}{n}\sum_{i=1}^n |f(X_i) -Y_i|^2 + \mu_n \| \mathscr{D}(f)\|_{L^2(\Omega)}^2$ might always be $0$, independently of the data points $(X_i, Y_i)$ \citep[see, e.g.,][Example 5.1]{doumeche2023convergence}.
A way to overcome these statistical issues is to specify the boundary conditions, and to consider regular differential operators $\mathscr{D}$ and smooth domain $\Omega$. For example, \citet{azzimonti2015blood},  \citet{arnone2022spatialRegression}, and \citet{ferraccioli2022some} consider models such that $f^\star|_{\partial \Omega} = 0$, where $\Omega$ is an Euclidean ball of $\mathbb{R}^d$ and $\mathscr{D}$ are second-order elliptic operators. However, these assumptions amount to adding a Sobolev penalty, since, in this case, $ \| \mathscr{D}(f)\|_{L^2(\Omega)} $ and $\|f\|_{H^2_0(\Omega)}$ are equivalent norms \citep[e.g.,][Chapter 6.3, Theorem~4]{evans2010partial}. 
Similar results hold for second order parabolic PDEs \citep[][Chapter 7.1, Theorem 5]{evans2010partial} and for second order hyperbolic PDEs \citep[][Chapter 7.2, Theorem 2]{evans2010partial}.
The need for a Sobolev regularization is explained by the fact that the  Sobolev embedding $H^s(\Omega) \hookrightarrow C^0(\Omega)$ only holds for $s>d/2$. 
In other words, the Sobolev regularization is needed to give a sense to the pointwise evaluations $|f(X_i) -Y_i|$. 
\label{rem:eq_reg}

\section{Application: speed-up effect of the physical penalty}
\label{sec:experiment}
\begin{wrapfigure}[16]{r}{0.35\textwidth}
    \vspace{-1.5em}
    \centering
    \includegraphics[width=0.35\textwidth]{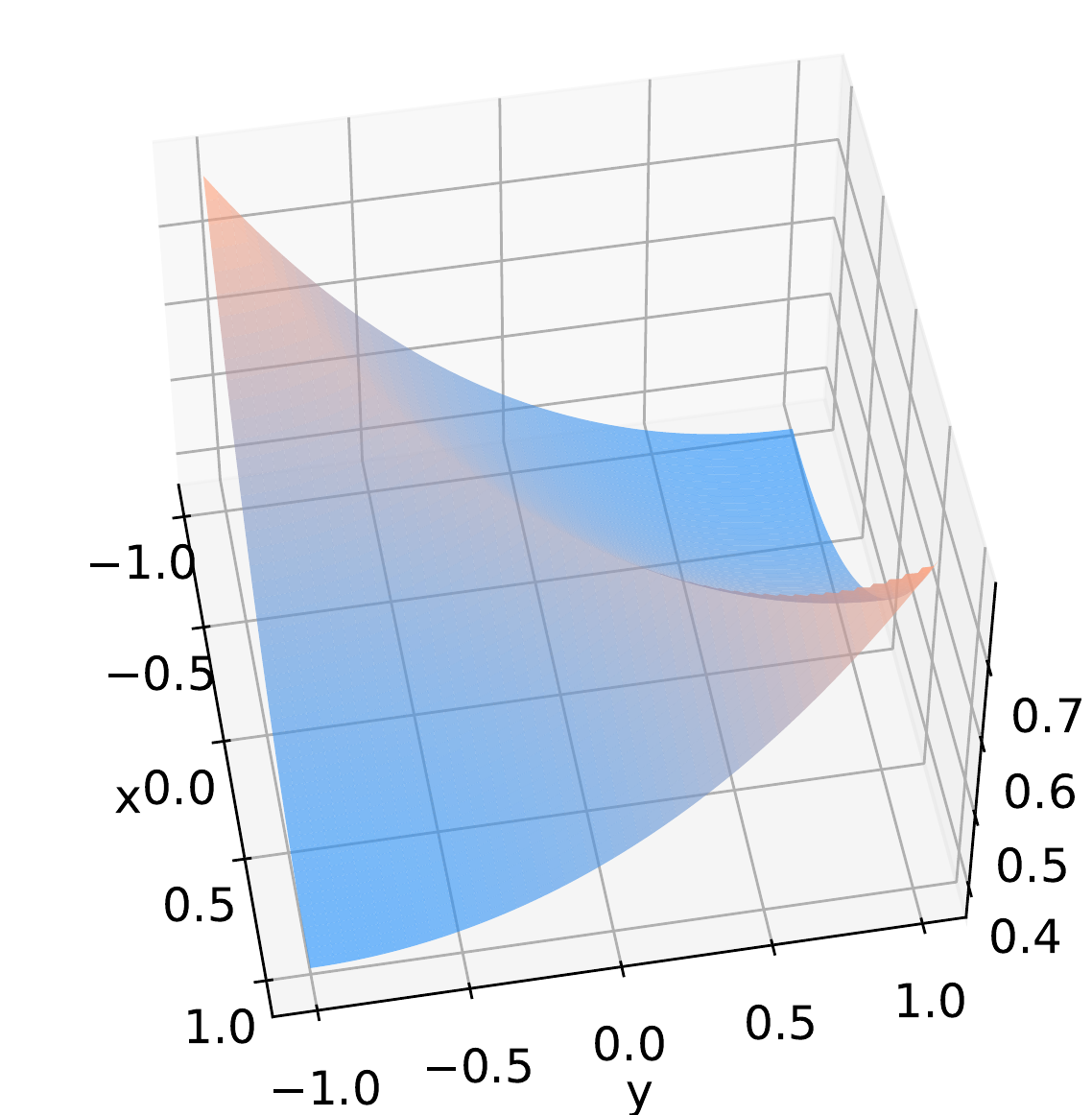}
    \caption{Kernel $K$ of Proposition \ref{prop:1dkernel} with $L=1$, $\lambda_n = \mu_n =1$.}
     \label{fig:2Dkernel}
 \end{wrapfigure}

Our objective is to apply the framework presented above to the case $d=1$, $\Omega = [-L, L]$, 
$s = 1$, $f^\star \in H^1(\Omega)$, and $\mathscr{D} = \frac{d}{dx}$.
Of course, assuming that $\mathscr{D}(f^\star) \simeq 0$ is a strong assumption, equivalent to assuming that $f^\star$ is approximately constant. However, the goal of this section is to provide a simple illustration where the kernel $K$ of Theorem \ref{thm:PDE_kernel} can be  analytically computed and
the eigenvalues of the operator $L_K$ can be effectively bounded.
The next result is a consequence of Proposition \ref{prop:kernel_characterization}.
\begin{prop}[One-dimensional kernel]
    Assume that $s = 1$, $\Omega = [-L, L]$, and $\mathscr{D} = \frac{d}{dx}$. Then, letting $\gamma_n = \sqrt{\frac{\lambda_n}{\lambda_n + \mu_n}}$, one has, for all $x,y \in [-L,L]$, 
    \begin{align*}
        K(x,y)& = \frac{\gamma_n}{2\lambda_n \sinh(2 \gamma_n L)}\Big( (\cosh(2\gamma_n L)+\cosh(2\gamma_n x))\cosh(\gamma_n (x-y))\\
        &\qquad \qquad + ((1-2 \times \mathbf{1}_{x > y})\sinh(2\gamma_n L) - \sinh(2 \gamma_n x)) \sinh(\gamma_n(x-y))\Big).
    \end{align*} 
    \label{prop:1dkernel}
\end{prop}
An example of kernel $K$ with $L=1$ and $\lambda_n=\mu_n=1$ is shown in Figure \ref{fig:2Dkernel}. Following the strategy of Section \ref{sec:bounds}, it remains to bound the positive eigenvalues $a_m$ of the operator $C\mathscr{O}_nC$ using Theorem~\ref{prop:eigenfunction}. According to the latter, this is achieved by solving the weak formulation
\begin{equation*}
        \forall \phi \in H^1_{\mathrm{per}}([-2L,2L]), \quad \lambda_n \int_{[-2L, 2L]^d}w_m\phi + (\lambda_n +\mu_n) \int_{\Omega}\frac{d}{dx}w_m\;\frac{d}{dx}\phi
    = a_m^{-1} \int_{\Omega} w_m \phi.
    \end{equation*} 
\begin{prop}[One-dimensional eigenvalues]
    Assume that $s = 1$, $\Omega = [-L, L]$, and $\mathscr{D} = \frac{d}{dx}$. Then, for all $m \geqslant 3$, 
    \[ \frac{4L^2}{(\lambda_n + \mu_n)(m+4)^2 \pi^2} \leqslant a_m \leqslant \frac{4L^2}{(\lambda_n + \mu_n)(m-2)^2 \pi^2},\]
    where $a_m$ are the eigenvalues of $C\mathscr{O}_nC$.
    \label{prop:1ddiff}
\end{prop}
Using inequality \eqref{eq:eff_dim_vp}, we can then bound the effective dimension of the kernel. This allows us, via Theorem \ref{thm:boundexp}, to specify the convergence rate of $\hat f_n$ to $f^{\star}$.
\begin{thm}[Kernel speed-up] 
   Assume that $f^\star \in H^1([-L,L])$, $\frac{d\mathbb{P}_X}{dx} \leqslant \kappa$ for some $\kappa > 0$, and the noise $\varepsilon$ satisfies the sub-Gamma condition \eqref{eq:conditionNoise}.
   Let $\lambda_n = n^{-1} \log(n)$ and 
    \begin{equation*}
        \mu_n = \left\{ 
    \begin{array}{llll}
        n^{-2/3}/ \|\mathscr{D}(f^\star)\|_{L^2(\Omega)} && \mathrm{ if}  &\|\mathscr{D}(f^\star)\|_{L^2(\Omega)} \neq 0 \\
        1/\log(n) && \mathrm{ if}  &\|\mathscr{D}(f^\star)\|_{L^2(\Omega)} = 0.  \\
    \end{array}\right.
    \end{equation*} 
    Then the estimator $\hat f_n$ of $f^\star$ minimizing the empirical risk function \eqref{eq:risk_function} with $s=1$ and $\mathscr{D} = \frac{d}{dx}$ satisfies
    \begin{align*}
        \mathbb{E}\int_{[-L,L]} |\hat f_n-f^\star|^2 d{\mathbb P}_X &= \|\mathscr{D}(f^\star)\|_{L^2(\Omega)}\;\mathcal{O}_n \big(  n^{-2/3}\log^3(n)\big) \\
        & \quad+ (\|f^\star\|_{H^1(\Omega)}^2 + \sigma^2 + M^2)\mathcal{O}_n \big( n^{-1} \log^3(n)\big) .
    \end{align*}
    \label{prop:kernel_speed_up}
\end{thm}
This bound reflects the benefit of the physical penalty $\|\mathscr{D}(f^\star)\|_{L^2(\Omega)}$ on the performance of the estimator $\hat f_n$. 
Indeed, when $\|\mathscr{D}(f^\star)\|_{L^2(\Omega)} = 0$ (i.e., the physical model is perfect), then $f^\star$ is a constant function, and the PIML method recovers the parametric convergence rate of $n^{-1}$. Here, the physical information directly improves the convergence rate. 
Otherwise, when $\|\mathscr{D}(f^\star)\|_{L^2(\Omega)}>0$, we recover the Sobolev minimax convergence rate in $H^1(\Omega)$ of $n^{-2/3}$ \citep[up to a log factor---see][Theorem 2.11]{tsybakov2009introduction}. 
We emphasize that this rate is also optimal for our problem, since $\|\mathscr{D}(f^\star)\|_{L^2(\Omega)} \leqslant \|f^\star\|_{H^1(\Omega)}$, i.e., it is as hard to learn a function of bounded $\|\mathscr{D}(\cdot)\|_{L^2(\Omega)}$ norm as it is to learn a function of bounded $H^1(\Omega)$ norm. 
In this case, the benefit of physical modeling is carried by the constant $\|\mathscr{D}(f^\star)\|_{L^2(\Omega)}$ in front of the convergence rate, i.e., the better the modeling, the smaller the estimation error. Note however that the parameter $\mu_n$ in Theorem \ref{prop:kernel_speed_up} depends on the unknown physical inconsistency $\|\mathscr{D}(f^\star)\|_{L^2(\Omega)}$. In practice, on may resort to a cross-validation-type strategy to estimate $\mu_n$.

\begin{figure}
    \centering
    \includegraphics[width=0.49\textwidth]{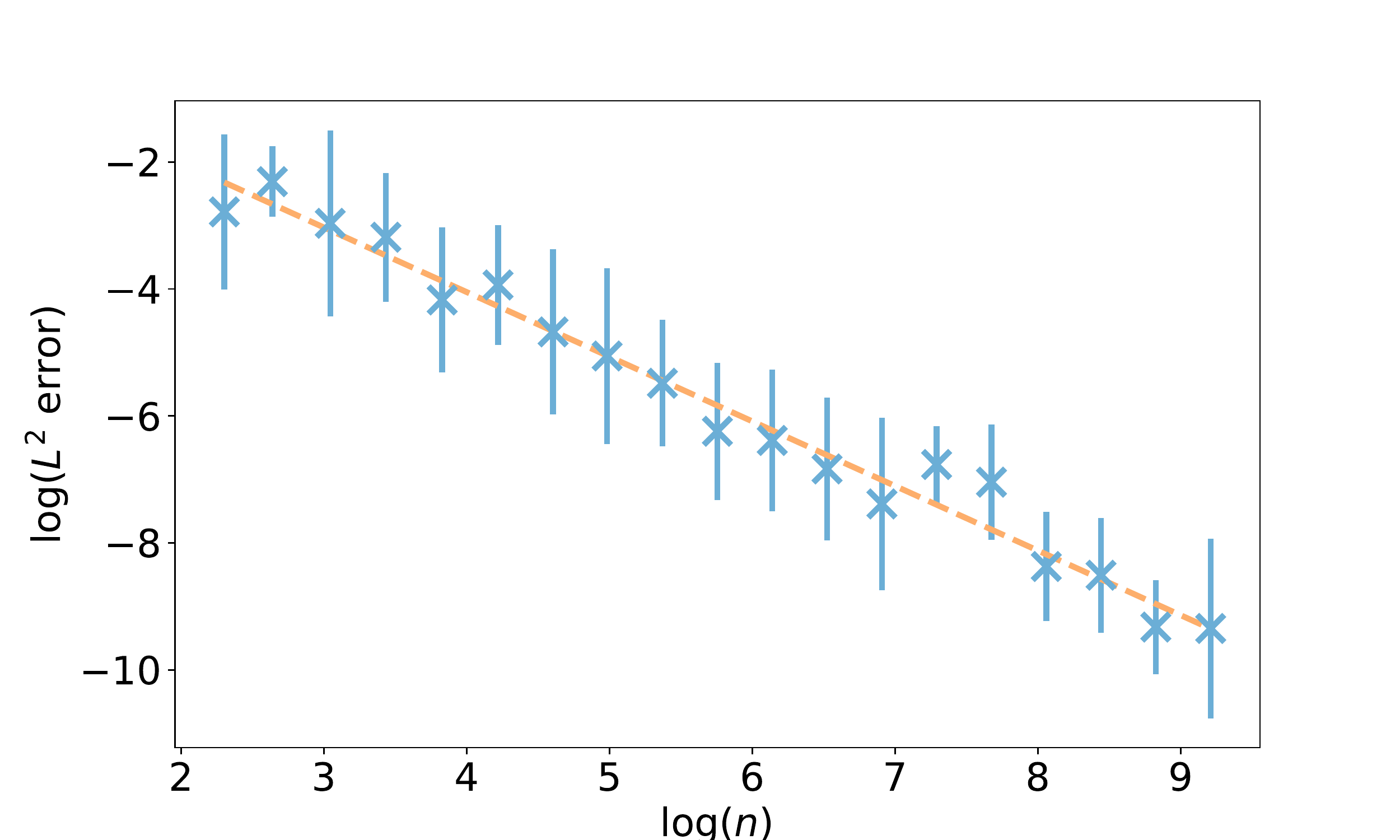}
    \includegraphics[width=0.49\textwidth]{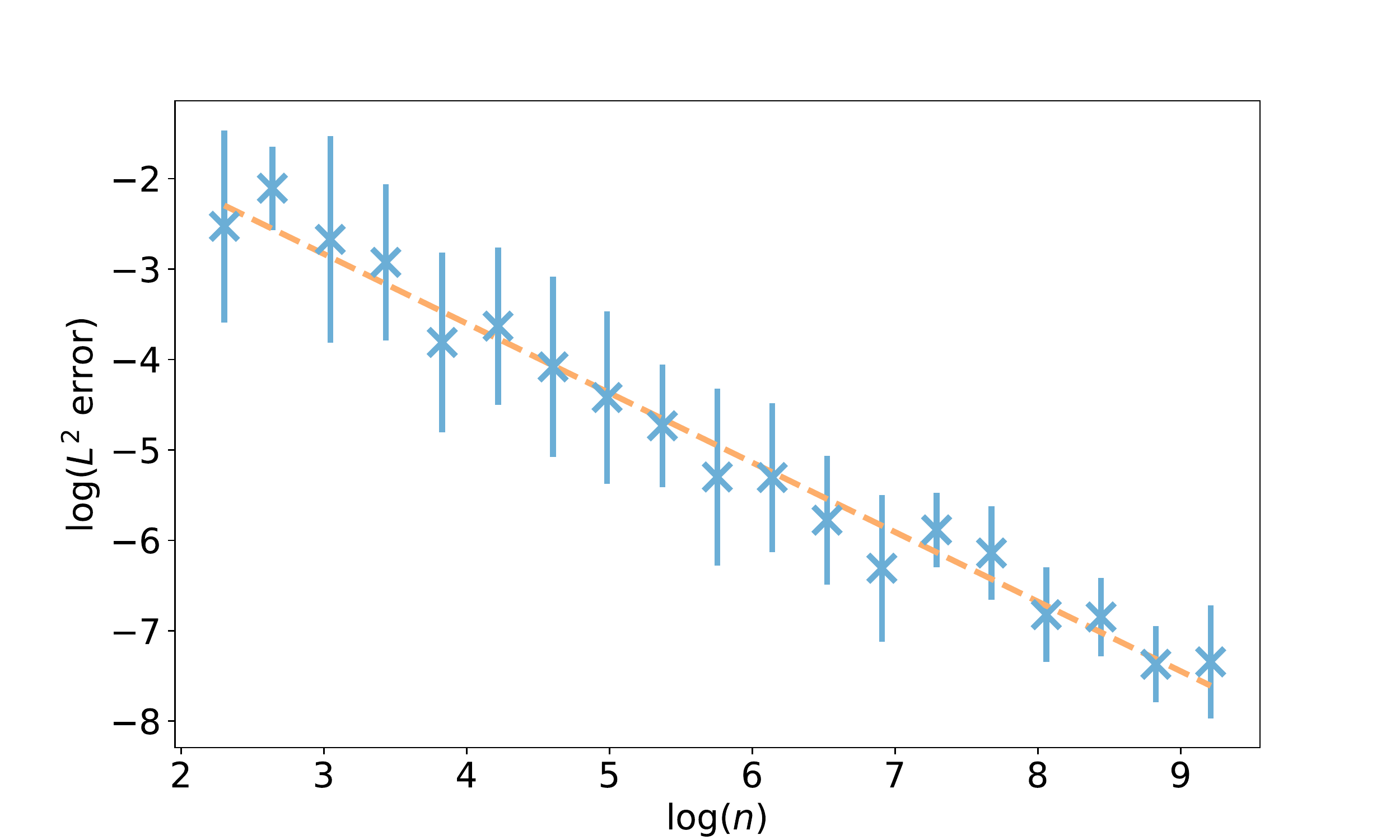}
    \caption{
    Error bounds $\mathrm{err}(n)$ (mean $\pm$ std over 10 runs) of the kernel estimator $\hat{f}_n$ with respect to the sample size $n$, in log-log scale, for the perfect modeling case (left) and the imperfect one (right). 
    The experimental convergence rates, obtained by fitting a linear regression, are displayed in orange dotted. 
    }
    \label{fig:numerical_experiment}
\end{figure}

We conclude this section with a small numerical experiment\footnote{The code to reproduce all numerical experiments can be found \href{https://github.com/NathanDoumeche/PIML_as_a_kernel_method}{here}.} illustrating Theorem \ref{prop:kernel_speed_up}. We consider two  problems: a perfect modeling situation  where $Y = 1 + \varepsilon$, and an imperfect modeling one where $Y = 1 + 0.1 |X| + \varepsilon$. In both cases, $X \sim \mathscr{U}([-1,1])$ and $\varepsilon \sim \mathcal{N}(0, 1)$. The difference is that in the perfect modeling case, $\mathscr{D}(f^\star) = 0$, whereas in the imperfect situation $\|\mathscr{D}(f^\star)\|_{L^2([-1,1])}^2 = 2/300$. For each $n$, we let $\mathrm{err}(n) = \mathbb{E}\int_\Omega |\hat f_n-f^\star|^2 d{\mathbb P}_X$.
Figure~\ref{fig:numerical_experiment} shows the values of 
$\log(\mathrm{err})(n)$ as a function of $\log(n)$, for $n$ ranging from $10$ to $10000$ (the quantity $\log(\mathrm{err})(n)$ is estimated by an empirical mean over 500-sample Monte Carlo estimations, repeated ten times). 
The experimental convergence rates obtained by fitting linear regressions are $-1.02$ in the perfect modeling case and $-0.77$ in the imperfect one. These experimental rates are consistent with the results of Theorem~\ref{prop:kernel_speed_up}, insofar as
$-1.02 \leqslant -1$ and $-0.77 \leqslant -2/3$. 
\section{Conclusion}
From the physics-informed machine learning point of view, we have shown that minimizing the empirical risk regularized by a PDE can be viewed as a kernel method. Leveraging kernel theory, we have explained how to derive convergence rates. In particular, the simple but instructive example $\mathscr{D}=\frac{d}{dx}$ illustrates how to compute both the kernel and the convergence rate of the associated estimator. To the best of our knowledge, this is the first contribution that demonstrates tangible improvements in convergence rates by including a physical penalty in the risk function. 
Thus, the take-home message is that physical information can be beneficial to the statistical performance of the estimators. Note that our work does not include boundary conditions $h$, but they could easily be considered. A first solution is to add another penalty to $R_n$ of the form $\|h-f\|^2_{L^2(\partial \Omega)}$, which would insert the extra term $\int_{\partial \Omega} (K(x,\cdot)-h)\phi$ in Proposition~\ref{prop:kernel_characterization}. 
A second solution is to enforce the conditions at new data points $X^{(b)}_j$ sampled on $\partial \Omega$ \citep[as done, for example, in][]{raissi2019PINN}. Our theorems hold for this extended training sample, provided that $f^\star_{|\partial \Omega}=h$.

An important future research direction is to implement numerical strategies for computing the kernel $K$ in the general case. If successful, such strategies can then be used directly to solve general physics-informed machine learning problems. In order to derive theoretical guarantees, we need to go further by obtaining bounds on the eigenvalues of the operator associated with the problem. The key lies in Theorem \ref{prop:eigenfunction}, which characterizes the eigenvalues by a weak formulation. Once established, such bounds can be employed to obtain accurate rates for related techniques, typically physics-informed neural networks. It would also be interesting to derive rates of convergence in the setting $s\leqslant d/2$ using the so-called source condition \citep[e.g.,][]{blanchard2020kernel}. An even more ambitious goal is to generalize the approach to nonlinear differential systems, for example polynomial. Overall, we believe that our results pave the way for a deeper understanding of the impact of physical regularization on empirical risk minimization performance.

\setcounter{section}{0}
\renewcommand\thesection{\thechapter.\Alph{section}}
\section{Some fundamentals of functional analysis}
\label{sec:appA}
\subsection*{Sobolev spaces}
\paragraph{Norms.} The $p$ norm $\|x\|_p$ of a $d$-dimensional vector $x = (x_1,\hdots, x_{d})$ is defined by $ \|x\|_p = (\frac{1}{d}\sum_{i=1}^{d} |x_i|^p)^{1/p}$.
For a function $f : \Omega \rightarrow \mathbb{R}$, we let $ \|f\|_{L^p(\Omega)} = (\frac{1}{|\Omega|}\int_\Omega |f|^p)^{1/p}$. Similarly, $\|f\|_{\infty, \Omega} = \sup_{x \in \Omega} |f(x)|$. For the sake of conciseness, we sometimes write $\|f\|_{\infty}$ instead of~$\|f\|_{\infty, \Omega}$.

\paragraph{Multi-indices and partial derivatives.} For a multi-index $\alpha = (\alpha_1, \hdots, \alpha_{d}) \in \mathbb{N}^{d}$ and a differentiable function $f:\mathbb R^{d}\to \mathbb R$, the $\alpha$ partial derivative of $f$ is defined by \[\partial^\alpha f = (\partial_{1})^{\alpha_1}\hdots (\partial_{d})^{\alpha_{d}} f.\] The set of multi-indices of sum less than $k$ is defined by \[\{|\alpha|\leqslant k\} = \{(\alpha_1, \hdots, \alpha_{d_1}) \in \mathbb{N}^{d}, \alpha_1 + \cdots +\alpha_{d_1} \leqslant k\}.\] If $\alpha = 0$, $\partial^\alpha f = f$. Given two multi-indices $\alpha$ and $\beta$, we write $\alpha \leqslant \beta$ when $\alpha_i \leqslant \beta_i$ for all $1\leqslant i \leqslant d$. 
The set of multi-indices less than $\alpha$ is denoted by $\{\beta \leqslant \alpha\}$. For a multi-index $\alpha$ such that $|\alpha|\leqslant k$, both sets $\{|\beta|\leqslant k\}$ and $\{\beta \leqslant \alpha\}$ are contained in $\{0, \hdots, k\}^{d}$ and are therefore finite.

\paragraph{Hölder norm.} For $K \in \mathbb{N}$, the Hölder norm of order $K$ of a function $f \in C^K(\Omega, \mathbb{R})$ is defined by $\|f\|_{C^K(\Omega)} = \max_{|\alpha|\leqslant K} \|\partial^\alpha f\|_{\infty, \Omega}$. 
This norm allows to bound a function as well as its derivatives. 
The space $C^K(\Omega, \mathbb{R})$ endowed with the Hölder norm $\|\cdot\|_{C^K(\Omega)}$ is a Banach space. 
The space $C^\infty(\bar{\Omega}, \mathbb{R}^{d_2})$ is defined as the subspace of continuous functions $f:\bar{\Omega} \to \mathbb{R}$ satisfying $f|_\Omega \in C^\infty(\Omega, \mathbb{R})$ and, for all $K\in \mathbb{N}$, $\|f\|_{C^K(\Omega)} < \infty$.

\paragraph{Lipschitz function.} Given a normed space $(V, \|\cdot\|)$, the Lipschitz norm of a function $f : V \rightarrow \mathbb{R}^{d}$ is defined by 
\[\|f\|_{\text{Lip}} = \sup_{x,y \in V}\frac{\|f(x)-f(y)\|_2}{\|x-y\|}.\] A function $f$ is Lipschitz if $\|f\|_{\mathrm{Lip}}<\infty$. The mean value theorem implies that for all $f \in C^1(V, \mathbb{R})$, $\|f\|_{\text{Lip}} \leqslant \|f\|_{C^1(V)}$.

\paragraph{Lipschitz surface  and domain.} A surface $\Gamma \subseteq \mathbb{R}^{d}$ is said to be Lipschitz if locally, in a neighborhood $U(x)$ of any point $x \in \Gamma$, an appropriate rotation $r_x$ of the coordinate system transforms $\Gamma$ into the graph of a Lipschitz function $\phi_{x}$, i.e., 
\[r_x(\Gamma \cap U(x)) = \{(x_1, \hdots, x_{d - 1}, \phi_x(x_1, \hdots, x_{d - 1})), \forall (x_1, \hdots, x_d)\in r_x(\Gamma \cap U_x)\}.\]
A domain $\Omega \subseteq \mathbb{R}^{d}$ is said to be Lipschitz if its has Lipschitz boundary and lies on one side of it, i.e., $\phi_x < 0$ or $\phi_x > 0$ on all intersections $\Omega \cap U_x$. All manifolds with $C^1$ boundary and all convex domains are Lipschitz domains \citep[e.g.,][]{agranovich2015lispchitz}.

\paragraph{Sobolev spaces.} Let $\Omega \subseteq \mathbb{R}^{d}$ be an open set. A function $g \in L^2(\Omega, \mathbb{R})$ is said to be the $\alpha$th weak derivative of  $f \in L^2(\Omega, \mathbb{R})$ if, for all $\phi \in C^\infty(\bar{\Omega}, \mathbb{R})$ with compact support in $\Omega$, one has
$\int_\Omega g \phi = (-1)^{|\alpha|} \int_\Omega f \partial^\alpha \phi$. 
This is denoted by $g = \partial^\alpha f$. For $s \in\mathbb{N}$, the Sobolev space $H^s(\Omega)$ is the space of all functions $f \in L^2(\Omega, \mathbb{R})$ such that $\partial^\alpha f$ exists for all $|\alpha|\leqslant s$. This space is naturally endowed with the norm \[\|f\|_{H^s(\Omega)} = \Big(\sum_{|\alpha|\leqslant s} \|\partial^\alpha u\|_{L^2(\Omega)} ^2\Big)^{1/2}.\] Of course, if a function $f$ belongs to the Hölder space $C^K(\bar \Omega, \mathbb{R})$, then it belongs to the Sobolev space $H^K(\Omega)$, and its weak derivatives are the usual derivatives. For more on Sobolev spaces, we refer the reader to \citet[Chapter 5]{evans2010partial}.

\paragraph{Fundamental results on Sobolev spaces.} Let $\Omega \subseteq \mathbb{R}^{d}$ be an open set and let $s \in \mathbb{N}$ be an order of differentiation. It is not straightforward to extend a function $f \in H^s(\Omega)$ to a function $\tilde{f} \in H^s(\mathbb{R}^{d})$ such that 
\[\tilde{f}|_\Omega = f|_\Omega \quad \text{and} \quad \|\tilde{f}\|_{H^s(\mathbb{R}^{d})} \leqslant C_\Omega \|f\|_{H^s(\Omega)},\]
for some constant $C_\Omega$ independent of $f$. This result is known as the extension theorem in \citet[][Chapter 5.4]{evans2010partial} when $\Omega$ is a manifold with $C^1$ boundary. 
However, the simplest domains in PDEs take the form $]0,L[^3\times ]0,T[$, the boundary of which is not $C^1$. Fortunately, \citet[][Theorem 5, Chapter VI.3.3]{stein1970lipschitz} provides an extension theorem for bounded Lipschitz domains. The following two theorems are proved in  \citet{doumeche2023convergence}.
\begin{thm}[Sobolev inequalities]
    \label{thm:sobIneq2}
    Let $\Omega \subseteq \mathbb{R}^{d}$ be a bounded Lipschitz domain and let $s \in \mathbb{N}$. If $s  > d_1/2$, then there is an operator $\tilde \Pi : H^{s}(\Omega) \to C^0(\Omega, \mathbb{R})$ such that, for all $f \in H^{s}(\Omega)$, $\tilde \Pi(f) = f$ almost everywhere. Moreover, there is a constant $C_\Omega >0$, depending only on $\Omega$, such that $\|\tilde \Pi(f)\|_{\infty, \Omega} \leqslant C_\Omega \|f\|_{H^{s}(\Omega)}.$
\end{thm}

\begin{thm}[Rellich-Kondrachov]
    \label{thm:rellichK2}
    Let $\Omega \subseteq \mathbb{R}^{d}$ be a bounded Lipschitz domain and let $s \in \mathbb{N}$. Let $(f_p)_{p\in \mathbb{N}}\in H^{s+1}(\Omega)$ be a sequence such that $(\|f_p\|_{H^{s+1}(\Omega)})_{p\in \mathbb{N}}$ is bounded.
    There exists a function $f_\infty \in H^{s+1}(\Omega)$ and a subsequence of $(f_p)_{p\in \mathbb{N}}$ that converges to $f_\infty$ with respect to the $H^s(\Omega)$ norm.
\end{thm}

\subsection*{Fourier series on complex periodic Sobolev spaces}
Let $L>0$.
\begin{defi}[Periodic extension operator]
    Let $d \in \mathbb{N}^\star$. The periodic extension operator $E_{\mathrm{per}}: L^2([-2L, 2L]^d) \to L^2([-4L, 4L]^d)$ is defined, for all function $f: [-2L, 2L[^d \to \mathbb{R}$ and all $x=(x_1, \hdots, x_d) \in [-4L, 4L]^d$, by
    \[E_{\mathrm{per}}(f)(x) = f\Big(x_1 - 4L\Big\lfloor \frac{x_1}{4L}\Big\rfloor, \hdots, x_d - 4L\Big\lfloor \frac{x_d}{4L}\Big\rfloor\Big).\] 
\end{defi}

\begin{defi}[Periodic Sobolev spaces]
    Let $s \in \mathbb{N}$. The space of functions $f$ such that $E_{\mathrm{per}}(f) \in H^s([-4L, 4L]^d)$ is denoted by $H_{\mathrm{per}}^s([-2L, 2L]^d)$.
\end{defi}
If $s>0$, then $H_{\mathrm{per}}^s([-2L, 2L]^d)$ is a strict linear subspace of $H^s([-2L, 2L]^d)$. For example, for all $s\geqslant 1$, the function $f(x) = x_1^2 + \cdots + x_d^2$ belongs to $H^s([-2L, 2L]^d)$, but $f \notin H_{\mathrm{per}}^s([-2L, 2L]^d)$. Indeed, though $E_{\mathrm{per}}(f)$ is continuous, it is not weakly differentiable. The following characterization of periodic Sobolev spaces in terms of Fourier series are well-known \citep[see, e.g.,][Chapter 2.1]{temam1995navier}.

\begin{prop}[Fourier decomposition on periodic Sobolev spaces]
    Let $s \in \mathbb{N}$ and $d \geqslant 1$. For all function $f \in H_{\mathrm{per}}^s([-2L, 2L]^d)$, there exists a unique vector $z \in \mathbb{C}^{\mathbb{Z}^d}$ such that
$f(x) = \sum_{k \in \mathbb{Z}^d} z_k \exp(i\frac{\pi}{2L}\langle k, x\rangle)$,
and
\[\forall |\alpha|\leqslant s, \quad \partial^\alpha f(x) = \Big(i\frac{\pi}{2L}\Big)^{|\alpha|}\sum_{k \in \mathbb{Z}^d} z_k \exp(i\frac{\pi}{2L}\langle k, x\rangle) \prod_{j=1}^d k_j^{\alpha_j}.\]
Moreover, for all multi-index $|\alpha| \leqslant s$,
$\|\partial^\alpha f\|_{L^2([-2L, 2L]^d))}^2 = (\frac{\pi}{2L})^{2|\alpha|}\sum_{k \in \mathbb{Z}^d}  |z_k|^2 \prod_{j=1}^d k_j^{2 \alpha_j}$. Therefore,
$\|f\|_{H^s([-2L, 2L]^d)}^2 = \sum_{k \in \mathbb{Z}^d}  |z_k|^2  
\sum_{|\alpha|\leqslant s} (\frac{\pi}{2L})^{2|\alpha|} \prod_{j=1}^d k_j^{2\alpha_j}
$.
\label{prop:dec_four}


\end{prop}

\begin{proof}  The uniqueness of the decomposition is a consequence of
\[z_k = \frac{1}{4^dL^d}\int_{[-2L, 2L]^d} f(x) \exp(-i\frac{\pi}{2L}\langle k, x\rangle) dx.\] 
To prove the existence of such a decomposition, consider $f \in H_{\mathrm{per}}^s([-2L, 2L]^d)$. Since $f \in L^2([-2L, 2L]^d)$ and its derivative with respect to the first variable $\partial_1 f \in L^2([-2L, 2L]^d)$, $f$ and $\partial_1 f$ can be decomposed into the following multidimensional Fourier series \citep[see, e.g.,][Chapter 5.4]{brezis2010functional}:
\[\forall x \in [-2L, 2L]^d, \quad f(x) = \sum_{k \in \mathbb{Z}^d} z_k \exp(i\frac{\pi}{2L}\langle k, x\rangle),\] 
\[\forall x \in [-2L, 2L]^d, \quad \partial_1 f(x) = \sum_{k\in \mathbb{Z}^d} \tilde z_k \exp(i\frac{\pi}{2L}\langle k,x\rangle).\] 
Observe that $E_{\mathrm{per}}(f)$ has the same Fourier decomposition as $f$ and that $E_{\mathrm{per}}(\partial_1 f)$ has the same decomposition as $\partial_1 f$. The goal is to show that $\tilde z_k = i \frac{\pi}{2L} k_1 z_k$.
By definition of the weak derivative $\partial_1 E_{\mathrm{per}}(f)$, for any test function $\phi \in C^\infty([-4L, 4L]^d)$ with compact support in $[-4L, 4L]^d$, one has 
\[\int_{[-4L, 4L]^d}  \phi\partial_1 E_{\mathrm{per}}(f)   = -\int_{[-4L, 4L]^d} E_{\mathrm{per}}(f)  \partial_1  \phi.\]
Let 
\[\psi(u) = \left\{ \begin{array}{cl}
     0 &\text{ if } -4L \leqslant u \leqslant -1-2L \\
      \frac{\int_{-1-2L}^u 
      \exp(\frac{-1}{(2L+1+v)^2}) \exp(\frac{-1}{(2L+v)^2})dv}{(\int_{-1-2L}^{-2L} \exp(\frac{-1}{(2L + 1+v)^2}) \exp(\frac{-1}{(2L+v)^2}dv)^{-1} },& \text{ if } -1-2L \leqslant u \leqslant -2L,\\
      1 & \text{ if } -2L \leqslant u \leqslant 2L-1,\\
       1 - \psi(u-4L)& \text{ if } 2L-1 \leqslant u \leqslant 2L, \\
       0 &\text{ if } 2L+1 \leqslant u \leqslant 4L.
\end{array}\right.\]
One easily verifies that $\psi \in C^\infty([-4L, 4L])$ and that it has a compact support in $[-4L, 4L]$. Moreover, $\|\psi\|_\infty =1$. Notice that, for all function $g \in L^2([-2L, 2L])$ and any $4L$-periodic function $\phi \in C^\infty([-4L, 4L])$ whose support is not necessary compact, 
\begin{equation}
    \int_{[-4L, 4L]} g \phi \psi = \int_{[-2L, 2L]} g \phi \quad \mbox{and} \quad \int_{[-4L, 4L]} g (\phi \psi)' = \int_{[-2L, 2L]} g \phi'.
    \label{eq:unityPartition}
\end{equation} 
To generalize such a property in dimension $d$, we let $\psi_d(x) = \prod_{j=1}^d \psi(x_j)$. Then, for all $k \in \mathbb{Z}^d$, $\phi_{k,d}(x) := \psi_d(x) \exp(-i\frac{\pi}{2L}\langle k,x\rangle)$ is a smooth function with compact support. Thus, by definition of the weak derivative,
\begin{align*}
    \int_{[-4L, 4L]^d}  \phi_{k,d} \partial_1 E_{\mathrm{per}}(f)   = -\int_{[-4L, 4L]^d} E_{\mathrm{per}}(f)  \partial_1 \phi_{k,d}.
\end{align*}
Moreover, using the left-hand side of \eqref{eq:unityPartition}, we have that
\[\int_{[-4L, 4L]^d}  \phi_{k,d} \partial_1 E_{\mathrm{per}}(f) = \int_{[-2L, 2L]^d} \exp(-i\frac{\pi}{2L}\langle k,x\rangle) \partial_1 E_{\mathrm{per}}(f)(x) dx = (4L)^d  \tilde z_k,\] 
while, using the right-hand side of \eqref{eq:unityPartition}, we have that 
\[\int_{[-4L, 4L]^d} E_{\mathrm{per}}(f)  \partial_1 \phi_{k,d} =  \frac{-i\pi}{2L} k_1\int_{[-2L, 2L]^d} E_{\mathrm{per}}(f)(x)  \exp(\frac{-i\pi}{2L} \langle k,x\rangle)dx = (4L)^d \frac{-i\pi}{2L} k_1 z_k.\]
Therefore, $\tilde z_k = i \frac{\pi}{2L} k_1 z_k$.

The exact same reasoning holds for $\partial_j f$, for all $1 \leqslant j \leqslant d$. By iterating on the successive derivatives, we obtain that for all $|\alpha|\leqslant s$,
$ \partial^\alpha f(x) = (i\frac{\pi}{2L})^{|\alpha|}\sum_{k \in \mathbb{Z}^d} z_k \exp(i \frac{\pi}{2L} \langle k, x\rangle) \prod_{j=1}^d k_j^{\alpha_j}$, as desired. The last two equations of the proposition are direct consequences of Parseval's theorem.
\end{proof} 
This proposition states that there is a one-to-one mapping between $H_{\mathrm{per}}^s([-2L, 2L]^d)$ and $\{z \in \mathbb{C}^{\mathbb{Z}^d}\; | \; \sum_k |z_k|^2 (1+\|k\|_2^2)^s  < \infty \;\mathrm{ and }\; \bar z_{k} = z_{-k}\}$. In particular, this shows that for $s >0$,  $H_{\mathrm{per}}^s([-2L, 2L]^d)$ is an Hilbert space for the norm $\|\cdot\|_{H^s([-2L, 2L]^d)}$. 

\subsection*{Fourier series on Lipschitz domains}
As for now, it is assumed that $\Omega \subseteq [-L, L]^d$ is a bounded Lipschitz domain. The objective of this section is to parameterize the Sobolev space $H^s(\Omega)$ by the space $\mathbb{C}^{\mathbb{Z}^d}$ of Fourier coefficients.
\begin{prop}[Fourier decomposition of $H^s(\Omega)$]
    Let $s\in \mathbb{N}$. For any function $ f \in H^s(\Omega)$, there is a vector $z \in \mathbb{C}^{\mathbb{Z}^d}$ such that $ \sum_{k \in \mathbb{Z}^d}  |z_k|^2 \|k\|_2^{2s} < \infty$ and
\begin{equation*}
    \forall |\alpha|\leqslant s, \forall x \in \Omega, \quad \partial^\alpha f(x)  =  \Big(i\frac{\pi}{2L}\Big)^{|\alpha|}\sum_{k \in \mathbb{Z}^d} z_k \exp(i\frac{\pi}{2L}\langle k, x\rangle) \prod_{j=1}^d k_j^{\alpha_j}.
\end{equation*}
Thus, $f$ can be linearly extended to the function  $\tilde E(f)(x) =  \sum_{k \in \mathbb{Z}^d} z_k \exp(i\frac{\pi}{2L}\langle k, x\rangle)$ which belongs to $H^s_{\mathrm{per}}([-2L,2L]^d)$.
Moreover, there is a constant $C_{s, \Omega}$, depending only on the domain $\Omega$ and the order of differentiation $s$, such that, for all $f \in H^s(\Omega)$,
\[\|\tilde E(f) \|_{H^s_{\mathrm{per}}([-2L, 2L]^d)}^2 = \sum_{k \in \mathbb{Z}^d}  |z_k|^2 \sum_{|\alpha|\leqslant s} \Big(\frac{\pi}{2L}\Big)^{2|\alpha|} \prod_{j=1}^d k_j^{2\alpha_j} \leqslant \tilde C_{s,\Omega} \|f \|_{H^s(\Omega)}^2.\]
\label{prop:dec_four_lip}
\end{prop}
\begin{proof}
    Let $ f \in H^s(\Omega)$. According to the Sobolev extension theorem \citep[][Chapter 5.4]{evans2010partial}, there is an extension operator $E:  H^s(\Omega) \to  H^s([-2L, 2L]^d)$ and a constant $C_{s, \Omega}$, depending only $\Omega$ and $s$, such that, for all $f \in H^s(\Omega)$,  $E(f) \in H^s([-2L, 2L]^d)$ and $\|E(f) \|_{H^s([-2L, 2L]^d)} \leqslant C_{s,\Omega} \|f \|_{H^s(\Omega)}$.
    Choose $\phi \in C^\infty([-2L, 2L]^d, [0,1])$ with compact support, and such that $\phi =1$ on $\Omega$ and $\phi = 0$ on $[-2L, 2L]^d \backslash [-3L/2, 3L/2]^d$. Then the extension operator $\tilde E(f) = \phi \times E(f)$ is such that $\tilde E(f) \in H^s_{\mathrm{per}}([-2L, 2L]^d)$. In addition, the Leibniz formula on weak derivatives shows that there is a constant $\tilde C_{s,\Omega}$ such that $\|\tilde E(f) \|_{H^s([-2L, 2L]^d)}^2 \leqslant \tilde C_{s,\Omega} \|f \|_{H^s(\Omega)}^2$. The result is then a direct consequence of Proposition \ref{prop:dec_four} applied to $\tilde E(f)$.
\end{proof}
Classical theorems on series differentiation show that given any vector 
$z \in \mathbb{C}^{\mathbb{Z}^d}$ satisfying 
$$ \sum_{k \in \mathbb{Z}^d}  |z_k|^2 \|k\|_2^{2s} < \infty\quad \text{and} \quad\bar z = -z,$$ the associated Fourier series belongs to $H^s(\Omega)$. This shows that one can identify $H^s(\Omega)$ with $\{z \in \mathbb{C}^{\mathbb{Z}^d}\; |\; \sum_{k \in \mathbb{Z}^d}  |z_k|^2 \|k\|_2^{2s} < \infty \; \mathrm{ and }\; \bar z = -z\}$, and the inner product  $\langle f, g\rangle_{H^s([-2L,2L]^d))} = \sum_{|\alpha|\leqslant s} \int_{[-2L,2L]^d} \partial^\alpha f \partial^\alpha g$ with $\langle \tilde z,  z\rangle_{\mathbb{C}^{\mathbb{Z}^d}} = \sum_{k \in \mathbb{Z}^d} \tilde z_k \bar z_k \sum_{|\alpha|\leqslant s} (\frac{\pi}{2L})^{2|\alpha|} \prod_{j=1}^d k_j^{2\alpha_j}$. 

\begin{prop}[Countable reindexing of $H^s(\Omega)$]
    There is a one-to-one mapping $k : \mathbb N \to \mathbb Z ^d$ such that, letting $e_j = (x \mapsto \exp(i \frac{\pi}{2L}\langle k(j), x\rangle))$, any function $f \in H^s(\Omega)$ can be written as $\sum_{j \in \mathbb{N}} z_j e_j$, with $ z \in \mathbb{C}^{\mathbb{N}}$ and $ \sum_{j\in\mathbb{N}} |z_j |^2 j^{2s/d} < \infty$.
    \label{prop:countable}
\end{prop}

\begin{proof}
    Let $ f \in L^2(\Omega)$. By Proposition \ref{prop:dec_four_lip}, we know that $ f \in H^s(\Omega)$ if and only if there is a vector $z \in \mathbb{C}^{\mathbb{Z}^d}$ such that $ \sum_{k \in \mathbb{Z}^d}  |z_k|^2 \|k\|_2^{2s} < \infty$, and  $f(x) = \sum_{k \in \mathbb{Z}^d} z_k \exp(i\frac{\pi}{2L}\langle k, x\rangle)$.
Let  $j \in \mathbb{N} \mapsto k(j) \in \mathbb{Z}^d$ be a one-to-one mapping such that $\|k(j)\|_1$ is increasing. 
Then, for all $K > 0$, \[ \begin{pmatrix}
    K + (d+1) - 1 \\ (d+1) - 1
\end{pmatrix} \leqslant \mathrm{argmin} \{j \in \mathbb{N} \; |\; \|k(j)\|_1 \geqslant K\} \leqslant 2^d \begin{pmatrix}
    K + (d+1) - 1 \\ (d+1) - 1
\end{pmatrix}.\] 
Indeed, $\begin{pmatrix}
    K + (d+1) - 1 \\ (d+1) - 1
\end{pmatrix}$ corresponds to the number of vectors $(n_0, \hdots, n_d) \in \mathbb{N}^{d+1} $ such that $n_0 + \cdots + n_d = K$, where $n_\ell$ represents the order of differentiation along the dimension $\ell$ and where $n_0$ is a fictive dimension to take into account derivatives of order less than $s$).
Since $\begin{pmatrix}
    K + (d+1) - 1 \\ (d+1) - 1
\end{pmatrix} \underset{j\to\infty}{\sim} \frac{K^d}{d!}$, we deduce that there are constants $C_1, C_2 > 0$ such that $C_1 j^{1/d} \leqslant \|k(j)\|_1 \leqslant C_2 j^{1/d}$.
Observe that 
$\|k\|_2^{2s} \geqslant (\max_{j=1}^d k_j)^{2s} \geqslant \|k\|_1^{2s}/ d^{2s}$, and that
$\|k\|_2^{2s} \leqslant (d \max_{j=1}^d k_j^2)^s \leqslant d^s \|k\|_1^{2s}$.
We conclude that $ f \in H^s(\Omega)$ if and only if $f$ can be written as $\sum_{j \in \mathbb{N}} z_{k(j)} \exp(i\frac{\pi}{2L}\langle k(j), x\rangle)$, where $\sum_{j \in \mathbb{N}} |z_{k(j)}|_2^2  j^{2s/d} < \infty$. 
\end{proof}

\subsection*{Operator theory}
An operator is a linear function between two Hilbert spaces, potentially of infinite dimensions. The objective of this section is to give conditions on the regularity of such an operator so that it behaves similarly to matrices in finite dimension spaces.  For more advanced material, the reader is referred to the textbooks by \citet[Chapter D.6]{evans2010partial} and \citet[Problem 37 (6)]{brezis2010functional}. 
\begin{defi}[Hermitian spaces and Hermitian basis]
     $(H, \langle \cdot, \cdot \rangle)$ is a Hermitian space when $H$ is a complex Hilbert space endowed with an Hermitian inner product $\langle \cdot, \cdot \rangle$. This Hermitian inner product is associated with the norm $\|u\|^2 = \langle u, u \rangle$, defining a topology on $H$. We say that $(v_n)_{n \in \mathbb{N}} \in H^{\mathbb{N}}$ is a Hermitian basis of $H$ if $\langle v_n, v_m\rangle = \delta_{n,m}$, and if for all $u \in H$, there exists a sequence $(z_n)_{n \in \mathbb{N}} \in  \mathbb{C}^{\mathbb{N}}$ such that $\lim_{n \to \infty}\|u - \sum_{j=1}^n z_j v_j\| =0$. $H$ is said to be separable if it admits an Hermitian basis.
\end{defi}
\begin{defi}[Self-adjoint operator]
    Let  $(H, \langle \cdot, \cdot \rangle)$ be a Hermitian space. Let $\mathscr{O}: H \to H$ be an operator. We say that $\mathscr{O}$ is self-adjoint if, for all $u,v \in H$, one has $\langle \mathscr{O} u, v\rangle = \langle  u, \mathscr{O} v\rangle$.
    
\end{defi}
\begin{defi}[Compact operator]
    Let  $(H, \langle \cdot, \cdot \rangle)$ be a Hermitian space.  Let $\mathscr{O}: H \to H$ be an operator. We say that $\mathscr{O}$ is compact if, for any bounded set $S \subseteq H$, the closure of $\mathscr{O}(S)$ is compact.
\end{defi}
\begin{thm}[Spectral theorem]
\label{thm:spectral}
    Let $\mathscr{O}$ be a compact self-adjoint operator on a separable Hermitian space $(H, \langle \cdot, \cdot \rangle)$. Then $\mathscr{O}$ is diagonalizable in an orthonormal basis with real eigenvalues, i.e., there is an Hermitian basis $(v_m)_{m \in \mathbb{N}}$ and real numbers $(a_m)_{m \in \mathbb{N}}$ such that, for all $u \in H$, $\mathscr{O}(u) = \sum_{m\in \mathbb{N}} a_m \langle v_m, u \rangle v_m$. 
\end{thm}
\begin{defi}[Positive operator]
    An operator $\mathscr{O}$ on a Hermitian space $(H, \langle \cdot, \cdot \rangle)$ is positive if, for all $u\in H$, $\langle u, \mathscr{O} u\rangle \geqslant 0$.
\end{defi}
\begin{thm}[Courant-Fischer min-max theorem]
\label{thm:courant_fischer}
    Let $\mathscr{O}$ be a positive compact self-adjoint operator on a separable Hermitian space $(H, \langle \cdot, \cdot \rangle)$. Then the eigenvalues of $\mathscr{O}$ are positive and, when reindexing them in a non-increasing order,
    \[a_m(\mathscr{O}) = \underset{\dim \Sigma = m}{\underset{\Sigma \subseteq H}{\max}} \underset{u \neq 0}{\min_{u \in \Sigma}}\|u\|^{-2}\langle u, \mathscr{O} u\rangle. \]
\end{thm}
\begin{defi}[Order on Hermitian operators]
    Let $\mathscr{O}_1$ and $\mathscr{O}_2$ be two positive compact self-adjoint operators on a separable Hermitian space $(H, \langle \cdot, \cdot \rangle)$. We say that $\mathscr{O}_1 \succeq \mathscr{O}_2$ if, for all $u \in H$, $\langle u, \mathscr{O}_1 u\rangle \geqslant\langle u, \mathscr{O}_2 u\rangle$. According to the Courant-Fischer min-max theorem, this implies that, for all $m\in \mathbb N$, $a_m(\mathscr{O}_1) \geqslant a_m(\mathscr{O}_2)$.
\end{defi}

\subsection*{Symmetry and PDEs}
The goal of this section is to recall various techniques useful for the determination of the eigenfunctions of a differential operator.
\begin{defi}[Symmetric operator]
    An operator $\mathscr{O}$ on a Hilbert space $\mathscr{H} \subseteq L^2([-2L,2L]^d)$ is said to be symmetric if, for all functions $f \in \mathscr{H}$ and for all $x \in [-2L,2L]^d$, $\mathscr{O}(f) (-x)= \mathscr{O}(f(-\cdot))(x)$, where $f(-\cdot)$ is the function such that $f(-\cdot)(x) = f(-x)$.
\end{defi}
For example, the Laplacian $\Delta$ in dimension $d=2$ is a symmetric operator, since $\Delta f(-\cdot)(x) = (\partial^2_{1,1}f(-\cdot))(x) + (\partial^2_{2,2}f(-\cdot))(x) = \Delta f(-x)$. However, $\partial_1$ is not symmetric, since $\partial_1 f(-\cdot)(x) = -(\partial_{1}f)(-x)$. 
\begin{prop}[Eigenfunctions of symmetric operators]
    Let $\mathscr{O}$ be a symmetric operator on a Hil\-bert space $\mathscr{H}$. Then, if $v$ is an eigenfunction of $\mathscr{O}$, $v^{\texttt{sym}} = v + v(-\cdot) $ and $v^{\texttt{antisym}} = v - v(-\cdot)$ are two eigenfunctions of $\mathscr{O}$ with the same eigenvalue as $v$, and $\int_{[-2L,2L]^d} v^{\texttt{sym}} v^{\texttt{antisym}} = 0$. Notice that $v = ({v^{\texttt{sym}} + v^{\texttt{antisym}}})/{2}$.
    \label{prop:sym}
\end{prop}
\begin{proof}
    Let $v$ be an eigenfunction of $\mathscr{O}$ for the eigenvalue $a \in \mathbb{R}$, i.e., $\mathscr{O}(v) = av$. Since $\mathscr{O}$ is symmetric, $\mathscr{O}(v(-\cdot)) = \mathscr{O}(v)(-\cdot) = a v(-\cdot)$. Therefore, $v^{\texttt{sym}}$ and $v^{\texttt{antisym}} $ are two eigenfunctions of $\mathscr{O}$ with $a$ as eigenvalue. Since $v^{\texttt{sym}}$ is symmetric and $v^{\texttt{antisym}} $ is antisymmetric, $\int_{[-2L,2L]^d} v^{\texttt{sym}} v^{\texttt{antisym}} = 0$, and so they are orthogonal.
\end{proof}

\section{The kernel point of view of PIML}
\label{app:diif_of}
This appendix is devoted to providing the tools of functional analysis relevant to our problem. 

\subsection*{Properties of the differential operator}
Let $\lambda_n > 0$ and $\mu_n \geqslant 0$. We study in this section some of the properties of the differential operator~$\mathscr{O}_n$ such  that, for all $f \in H^s_{\mathrm{per}}([-2L,2L]^d)$, $\|\mathscr{O}_n^{-1/2}(f)\|_{L^2([-2L, 2L]^d)}^2 = \lambda_n \|f\|_{H^s_{\mathrm{per}}([-2L, 2L]^d)}^2 + \mu_n \|\mathscr{D}(f)\|_{L^2(\Omega)}^2$.

\begin{prop}[Differential operator]
    There is an injective operator $\mathscr{O}_n: L^2([-2L, 2L]^d) \to H^s_{\mathrm{per}}([-2L, 2L]^d)$ defined as follows: for all $f \in L^2([-2L, 2L]^d)$, $\mathscr{O}_n(f)$ is the unique element of $H^s_{\mathrm{per}}([-2L, 2L]^d)$ such that, for any test function $\phi \in H^s_{\mathrm{per}}([-2L, 2L]^d)$,
        \[  \lambda_n \sum_{|\alpha|\leqslant s} \int_{[-2L, 2L]^d}\partial^\alpha \phi\; \partial^\alpha \mathscr{O}_n(f) + \mu_n \int_{\Omega}\mathscr{D} \phi \; \mathscr{D}  \mathscr{O}_n(f) = \int_{[-2L, 2L]^d} \phi f. \]
        Moreover, $\|\mathscr{O}_n f\|_{H^s_{\mathrm{per}}([-2L, 2L]^d)} \leqslant \lambda_n^{-1} \|f\|_{L^2([-2L, 2L]^d)}$, i.e., $\mathscr{O}_n$ is bounded.
        \label{prop:diff_op}
\end{prop}
\begin{proof}
     We use the framework provided by \citet[][page 304]{evans2010partial} to prove the result. Let the bilinear form $B: H^s_{\mathrm{per}}([-2L, 2L]^d) \times H^s_{\mathrm{per}}([-2L, 2L]^d) \to \mathbb{R}$ be defined by
    \[B[u,v] = \lambda_n \sum_{|\alpha|\leqslant s} \int_{[-2L, 2L]^d}\partial^\alpha u\; \partial^\alpha v + \mu_n \int_{\Omega}\mathscr{D} u \; \mathscr{D}  v.
    \]
    Observe that $B$ is coercive since $B[u,u] \geqslant \lambda_n \|u\|_{H^s_{\mathrm{per}}([-2L, 2L]^d)}^2$. Moreover, using the Cauchy-Schwarz inequality $(x_1 + \cdots + x_N)^2 \leqslant N(x_1^2 + \cdots + x_N^2)$, we see that
    \begin{align*}
        \int_{\Omega}|\mathscr{D} u|^2 &=  \int_{\Omega}\Big|\sum_{|\alpha|\leqslant s} p_\alpha \partial^\alpha u\Big|^2 \\
        &\leqslant (\max_\alpha \|p_\alpha\|_\infty)^2    \int_{[-2L, 2L]^d}\Big(\sum_{|\alpha|\leqslant s}| \partial^\alpha u|\Big)^2\\
        &\leqslant (\max_\alpha \|p_\alpha\|_\infty)^2\; 2^{s}\;  \|u\|_{H^s_{\mathrm{per}}([-2L, 2L]^d)}^2.
    \end{align*}
    Therefore, using the Cauchy-Schwarz inequality, we have
    \[\Big|\int_{\Omega}\mathscr{D} u \; \mathscr{D}  v\Big| \leqslant (\max_\alpha \|p_\alpha\|_\infty)^2\; 2^{s}\;  \|u\|_{H^s_{\mathrm{per}}([-2L, 2L]^d)}\|v\|_{H^s_{\mathrm{per}}([-2L, 2L]^d)}.\]
    Thus, 
    \[|B[u,v]| \leqslant (\lambda_n + (\max_\alpha \|p_\alpha\|_\infty)^2\; 2^{s} \mu_n)  \|u\|_{H^s_{\mathrm{per}}([-2L, 2L]^d)}\|v\|_{H^s_{\mathrm{per}}([-2L, 2L]^d)},\] 
    showing thereby the continuity of $B$.

    Next, for $f \in L^2([-2L, 2L]^d)$, observe that $ \phi \mapsto \int_{[-2L, 2L]^d} \phi f$ is a bounded linear form on $H^s_{\mathrm{per}}([-2L, 2L]^d)$, since 
    \[\Big|\int_{[-2L, 2L]^d} \phi f\Big| \leqslant \|\phi\|_{L^2([-2L, 2L]^d)} \|f\|_{L^2([-2L, 2L]^d)} \leqslant \|\phi\|_{H^s_{\mathrm{per}}([-2L, 2L]^d)} \|f\|_{L^2([-2L, 2L]^d)}.\] Thus, the Lax-Milgram theorem \citep[Chapter 6.2, Theorem 1]{evans2010partial} ensures that for all $f \in L^2([-2L, 2L]^d)$, there is a unique element $w \in H^s_{\mathrm{per}}([-2L, 2L]^d)$ such that, for any test function $\phi \in H^s_{\mathrm{per}}([-2L, 2L]^d)$,
        \[  \lambda_n \sum_{|\alpha|\leqslant s} \int_{[-2L, 2L]^d}\partial^\alpha \phi\; \partial^\alpha w + \mu_n \int_{\Omega}\mathscr{D} \phi \; \mathscr{D}  w = \int_{[-2L, 2L]^d} \phi f. \]
     Call $\mathscr{O}_n$ the function associating $w$ to $f$. Then, by the uniqueness of $w$ provided by the Lax-Milgram theorem, we deduce that $\mathscr{O}_n$ is injective and linear. Moreover, using the coercivity of $B$, we have
     \begin{align*}
         \|\mathscr{O}_n f\|_{H^s_{\mathrm{per}}([-2L, 2L])}^2 &\leqslant \lambda_n^{-1} B[\mathscr{O}_nf,\mathscr{O}_nf] = \lambda_n^{-1} \langle \mathscr{O}_n f, f\rangle_{L^2([-2L, 2L]^d)} \\
         & \leqslant \lambda_n^{-1} \|f\|_{L^2([-2L, 2L]^d)} \|\mathscr{O}_n f\|_{L^2([-2L, 2L]^d)}\\
         & \leqslant \lambda_n^{-1} \|f\|_{L^2([-2L, 2L]^d)} \|\mathscr{O}_n f\|_{H^s_{\mathrm{per}}([-2L, 2L]^d)}.
     \end{align*}
     In particular, $\|\mathscr{O}_n f\|_{H^s_{\mathrm{per}}([-2L, 2L]^d)} \leqslant \lambda_n^{-1} \|f\|_{L^2([-2L, 2L]^d)}$, and the proof is complete.
\end{proof}

\begin{prop}[Diagonalization on $L^2$]
    There exists an orthonormal basis $(v_m)_{m\in \mathbb{N}}$ of the space $L^2([-2L, 2L]^d)$ of eigenfunctions of $\mathscr{O}_n$, associated with non-increasing strictly positive eigenvalues $(a_m)_{m\in \mathbb{N}}$, such that $\mathscr{O}_n = \sum_{m \in \mathbb{N}} a_m \langle v_m, \cdot\rangle_{L^2([-2L, 2L]^d)} v_m$. 
    \label{prop:dzL2}
\end{prop}
\begin{proof}
    By the Rellich-Kondrakov theorem (Theorem \ref{thm:rellichK2}), the operator $\mathscr{O}_n: L^2([-2L, 2L]^d) \to L^2([-2L, 2L]^d)$ is compact. Moreover, by definition of $\mathscr{O}_n$, for all $f,g \in L^2([-2L, 2L]^d)$, one has $\langle f, \mathscr{O}_n g\rangle_{L^2([-2L, 2L]^d)} = B[\mathscr{O}_nf,\mathscr{O}_n g]  = \langle \mathscr{O}_n f,  g\rangle_{L^2([-2L, 2L]^d)}$. Therefore, $\mathscr{O}_n$ is self-adjoint. Furthermore, $\langle f, \mathscr{O}_n f\rangle_{L^2([-2L, 2L]^d)} = B[\mathscr{O}_nf,\mathscr{O}_n f] \geqslant \lambda_n \|\mathscr{O}_n f\|_{H^s_{\mathrm{per}}([-2L, 2L])} > 0$, since $\mathscr{O}_n$ is injective. This means that $\mathscr{O}_n$ is strictly positive.  The result is then a consequence of the spectral theorem (Theorem \ref{thm:spectral}).
\end{proof}

\begin{prop}[Diagonalization on $H^s$]
    The orthonormal basis $(v_m)_{m\in \mathbb{N}}$ of Proposition \ref{prop:dzL2} is in fact a basis of $H^s_{\mathrm{per}}([-2L, 2L]^d)$. 
    Moreover, letting $C_1 = (\lambda_n + (\max_\alpha \|p_\alpha\|_\infty)^2\; 2^{s} \mu_n)$, we have that for all $f\in H^s_{\mathrm{per}}([-2L, 2L]^d)$, \[\sum_{m \in \mathbb{N}} a_m^{-1}\langle f , v_m\rangle_{L^2([-2L, 2L]^d)}^2 \leqslant  C_1 \|f\|_{H^s_{\mathrm{per}}([-2L, 2L]^d)}^2.\]
    \label{prop:diagH}
\end{prop}
\begin{proof}
    We follow the framework of \citet[][page 337]{evans2010partial}. First observe that $\mathscr{O}_n v_m = a_m v_m$ and $\|v_m\|_{L^2([-2L, 2L]^d)} =1$, implies that $\|\mathscr{O}_n v_m\|_{H^s_{\mathrm{per}}([-2L, 2L]^d)} \leqslant \lambda_n^{-1} \|v_m\|_{L^2([-2L, 2L])}$ implies that $\|v_m\|_{H^s_{\mathrm{per}}([-2L, 2L]^d)} \leqslant \lambda_n^{-1} a_m^{-1}$. Therefore, $v_m \in H^s_{\mathrm{per}}([-2L, 2L]^d)$, and we can apply $B$ to it. For all $m\in \mathbb{N}$,
    \[B[v_m, v_m] = B[v_m, a_m^{-1}\mathscr{O}_n v_m] = a_m^{-1} \langle v_m, v_m\rangle_{L^2([-2L, 2L]^d)} =  a_m^{-1}.\]
    Similarly, if $m \neq \ell$, $B[v_m, v_\ell] = a_m^{-1} \langle v_m, v_\ell\rangle_{L^2([-2L, 2L]^d)} =  0$. Remark that $B$ is a inner product on $H^s_{\mathrm{per}}([-2L,2L]^d)$ and $(\sqrt{a_m}v_m)_{m\in \mathbb{N}}$ is an orthonormal family for the $B$-inner product. Indeed, notice that if, for a fixed $u \in H^s_{\mathrm{per}}([-2L, 2L]^d)$, one has $B[v_m, u]= 0$ for all $m \in \mathbb N$, then $\langle v_m, u\rangle_{L^2([-2L, 2L]^d)} = 0$. Thus, since $(v_m)_{m\in \mathbb{N}}$ is an orthonormal basis of $L^2([-2L, 2L]^d)$, $u=0$.

    Let, for $N \in \mathbb{N}$ and $u \in H^s_{\mathrm{per}}([-2L, 2L]^d)$,  
    \[u_N = \sum_{m=0}^N B[u, a_m^{1/2} v_m] a_m^{1/2} v_m.\]
    Since $v_m \in H^s_{\mathrm{per}}([-2L, 2L]^d)$, one has $u_N \in H^s_{\mathrm{per}}([-2L, 2L]^d)$.
    Upon noting that $B[u-u_N, u-u_N] \geqslant 0$ and that $B[u-u_N, u-u_N] = B[u, u] - \sum_{m=0}^N B[u, a_m^{1/2}v_m]^2$ (using the bilinearity of~$B$), we derive the following Bessel's inequality for $B$ 
    \begin{equation}
        \sum_{m=0}^\infty B[u, a_m^{1/2}v_m]^2 \leqslant B[u,u] \leqslant C_1 \|u\|_{H^s_{\mathrm{per}}([-2L, 2L]^d)}^2.
        \label{eq:bessel}
    \end{equation}
    Then, for all $\ell \geqslant p$, 
    \[B[u_\ell - u_p, u_\ell - u_p] = \sum_{m = p}^\ell B[u, a_m^{1/2} v_m]^2 \leqslant \sum_{m = p}^\infty B[u, a_m^{1/2} v_m]^2 \xrightarrow{p\to \infty} 0.\]
    This shows that $(u_N)_{N\in \mathbb{N}}$ is a Cauchy sequence for the $B$-inner product. Since, $B[u_\ell - u_p, u_\ell - u_p] \geqslant \lambda_n^{-1} \|u_\ell - u_p\|_{H^s([-2L, 2L]^d)}^2$, $(u_N)_{N\in \mathbb{N}}$ is also a Cauchy sequence for the $\|\cdot\|_{H^s_{\mathrm{per}}([-2L, 2L]^d)}$ norm. Recalling that $H^s_{\mathrm{per}}([-2L, 2L]^d)$ is a Banach space, we deduce that $u_\infty := \lim_{N\to \infty} u_N$ exists and belongs to $H^s_{\mathrm{per}}([-2L, 2L]^d)$. 
    Since $B$ is continuous with respect to the  $\|\cdot\|_{H^s_{\mathrm{per}}([-2L, 2L]^d)}$ norm, we also deduce that, for all $m \in \mathbb{N}$, $B[u-u_\infty, v_m] = 0$, i.e., $u = u_\infty$. 
    In conclusion, \[u = \sum_{m\in \mathbb{N}} B[u, a_m^{1/2} v_m] a_m^{1/2} v_m.\]
    This means that $(v_m)_{m\in \mathbb{N}}$ is a basis of $H^s_{\mathrm{per}}([-2L, 2L]^d)$. Moreover, using the Bessel's inequality \eqref{eq:bessel}, we have
    $\sum_{m=0}^\infty B[u, a_m^{1/2}v_m]^2 = \sum_{m\in \mathbb{N}} a_m^{-1}\langle u , v_m\rangle_{L^2([-2L,2L]^d)}^2 \leqslant C_1 \|u\|_{H^s_{\mathrm{per}}([-2L, 2L]^d)}^2$.
\end{proof}

\begin{prop}[Differential inner product]
    The operators
    \begin{itemize}
        \item $\mathscr{O}_n^{-1/2}:H^s_{\mathrm{per}}([-2L, 2L]^d) \to L^2([-2L, 2L]^d)$, defined by \[\mathscr{O}_n^{-1/2} = \sum_{m\in\mathbb{N}} a_m^{-1/2} \langle v_m, \cdot\rangle_{L^2([-2L,2L]^d)} v_m,\]
        \item \mbox{and } $\mathscr{O}_n^{1/2}:L^2([-2L, 2L]^d) \to H^s_{\mathrm{per}}([-2L, 2L]^d)$, defined by \[\mathscr{O}_n^{1/2} = \sum_{m\in\mathbb{N}} a_m^{1/2} \langle v_m, \cdot\rangle_{L^2([-2L,2L]^d)} v_m,\]
    \end{itemize} are well-defined and bounded. Moreover, for all $u \in H^s_{\mathrm{per}}([-2L, 2L]^d)$, \[\mathscr{O}_n^{1/2} \mathscr{O}_n^{-1/2} (u) = u,\] and
    \[\|\mathscr{O}_n^{-1/2}(u)\|_{L^2([-2L, 2L]^d)}^2 = \lambda_n \|u\|_{H^s_{\mathrm{per}}([-2L, 2L]^d)}^2 + \mu_n \|\mathscr{D}(u)\|_{L^2(\Omega)}^2.\]
    \label{prop:scalar_product}
\end{prop}
\begin{proof}
    Proposition \ref{prop:diagH} shows that, for all $u \in H^s_{\mathrm{per}}([-2L, 2L]^d)$, $$\sum_{m\in \mathbb{N}} a_m^{-1}\langle u , v_m\rangle_{L^2([-2L,2L]^d)}^2 \leqslant C_1 \|u\|_{H^s([-2L, 2L]^d)}^2.$$ 
    Thus, $(\sum_{m=0}^N a_m^{-1/2} \langle v_m, u\rangle_{L^2([-2L,2L]^d)} v_m)_{N\in \mathbb{N}}$ is a Cauchy sequence converging in the space $L^2([-2L, 2L]^d)$. Denote this limit by $\mathscr{O}_n^{-1/2} (u)$.
    Since $B[u,u] = \sum_{m\in \mathbb{N}} a_m^{-1}\langle u , v_m\rangle_{L^2([-2L,2L]^d)}^2$, we deduce that \[\|\mathscr{O}_n^{-1/2}(u)\|_{L^2([-2L, 2L]^d)}^2 = \lambda_n \|u\|_{H^s_{\mathrm{per}}([-2L, 2L]^d)}^2 + \mu_n \|\mathscr{D}(u)\|_{L^2(\Omega)}^2.\] 
    Finally, using $\|\mathscr{O}_n^{-1/2} (u)\|_{L^2([-2L, 2L]^d)}^2  \leqslant C_1 \|u\|_{H^s_{\mathrm{per}}([-2L, 2L]^d)}^2$, we conclude that the operator~$\mathscr{O}_n^{-1/2}$ is bounded.
    
    Moreover, $(\sum_{m=0}^N a_m^{1/2} \langle v_m, u\rangle_{L^2([-2L,2L]^d)} v_m)_{N\in \mathbb{N}}$ is also a Cauchy sequence in the space $H^s_{\mathrm{per}}([-2L, 2L]^d)$. To see this, 
    note that $\sum_{m=0}^N a_m^{1/2} \langle v_m, u\rangle_{L^2([-2L,2L]^d)} v_m \in H^s_{\mathrm{per}}([-2L, 2L]^d) $, and that this sequence is a Cauchy sequence for the $B$ inner product, because
    \begin{align*}
        &B\Big[\sum_{m=p}^\ell a_m^{1/2} \langle v_m, u\rangle_{L^2([-2L,2L]^d)} v_m, \sum_{m=p}^\ell a_m^{1/2} \langle v_m, u\rangle_{L^2([-2L,2L]^d)} v_m\Big] \\
        &\quad = \sum_{m=p}^\ell a_m \langle v_m, u\rangle_{L^2([-2L,2L]^d)} ^2 B[v_m, v_m]\\
        &\quad = \sum_{m=p}^\ell  \langle v_m, u\rangle_{L^2([-2L,2L]^d)} ^2 \\
        &\quad \leqslant \sum_{m = p}^\infty \langle v_m, u\rangle_{L^2([-2L,2L]^d)}^2  \xrightarrow{p\to \infty} 0.
    \end{align*}
    Thus, it $\sum_{m=0}^N a_m^{1/2} \langle v_m, u\rangle_{L^2([-2L,2L]^d)} v_m$ converges in $H^s_{\mathrm{per}}([-2L, 2L]^d)$ to a limit that we denote by $\mathscr{O}_n^{1/2}(u)$. By the continuity of $B$, $B[\mathscr{O}_n^{1/2} (u),\mathscr{O}_n^{1/2}(u)] \leqslant \|u\|_{L^2([-2L, 2L]^d)}^2$. Therefore, $\|\mathscr{O}_n^{1/2}(u)\|_{H^s_{\mathrm{per}}([-2L, 2L]^d)}^2 \leqslant \lambda_n^{-1} \|u\|_{L^2([-2L, 2L]^d)}^2$, i.e., $\mathscr{O}_n^{1/2}$ is bounded.
    
    To conclude the proof, observe that since $(\sum_{m=0}^N a_m^{-1/2} \langle v_m, u\rangle_{L^2([-2L,2L]^d)} v_m)_{N\in \mathbb{N}}$ converges to $ \mathscr{O}_n^{-1/2}(u)$ in $L^2([-2L, 2L]^d)$, and since the inner product $\langle\cdot, \cdot\rangle$ is continuous with respect to the $L^2([-2L, 2L]^d)$ norm (by the Cauchy-Schwarz inequality), then, for all $u \in H^s_{\mathrm{per}}([-2L, 2L]^d)$, one can write $\langle \mathscr{O}_n^{-1/2}(u), v_m\rangle_{L^2([-2L,2L]^d)} = a_m^{-1/2} \langle v_m, u\rangle_{L^2([-2L,2L]^d)}$. Besides, since the sequence $(\sum_{m=0}^N a_m^{1/2} \langle v_m, \mathscr{O}_n^{-1/2}(u)\rangle_{L^2([-2L,2L]^d)} v_m)_{N\in \mathbb{N}}$ converges in the space $L^2([-2L, 2L]^d)$ to $\mathscr{O}_n^{1/2}  \mathscr{O}_n^{-1/2}(u)$, one has that $\langle \mathscr{O}_n^{1/2} \mathscr{O}_n^{-1/2}(u), v_m\rangle_{L^2([-2L,2L]^d)} = \langle u, v_m\rangle_{L^2([-2L,2L]^d)}$. Finally, this shows that $\mathscr{O}_n^{1/2} \mathscr{O}_n^{-1/2}(u) = u$, and the proof is complete.
\end{proof}

Recall from Proposition \ref{prop:countable} that there exists a countable re-indexing $k: \mathbb{N} \to \mathbb{Z}^d$ such that $\|k\|_1$ is non-decreasing. Recall that we have let $e_{k(\ell)}(x) := \exp(i\langle k(\ell), x\rangle)$.

\begin{lem}[Non-empty intersection]
    Let $\mathscr{V}$ be a linear subspace of $H^s_{\mathrm{per}}([-2L,2L]^d)$ such that $\dim \mathscr{V} = m+1$.
    Then $\mathscr{V} \cap \mathrm{Span}( e_{k(\ell)})_{\ell \geqslant m} \neq \emptyset$.
    \label{lem:non_empty}
\end{lem}
\begin{proof}
    Let $z_0, \hdots, z_m$ be a basis of $\mathscr{V}$. 
    Let us consider the linear function \[T: (x_0, \hdots, x_m) \in \mathbb{R}^{m+1} \mapsto \Big( \langle e_{k(j)}, \sum_{\ell = 0}^m x_\ell z_\ell\rangle_{L^2([-2L,2L]^d)}\Big)_{0\leqslant j \leqslant m-1} \in \mathbb{R}^m.\]
    The rank–nullity theorem ensures that the dimension of the kernel of $T$ is at least 1. Thus, there is a linear combination $z = x_0 z_0 + \cdots + x_m z_m$ such that, for all $\ell \leqslant m - 1$, $\langle z, e_{k(\ell)}\rangle_{L^2([-2L,2L]^d)} = 0$ and $z \neq 0$. Since $(e_{k(\ell)})_{\ell \geqslant 0}$ is a basis of $H^s_{\mathrm{per}}([-2L,2L]^d)$, we conclude that $z \in \mathrm{Span}( e_{k(\ell)})_{\ell \geqslant m} \cap \mathscr{V}$.
\end{proof}

\begin{prop}[Eigenvalues of the differential operator]
    There is a constant $C_2>0$, depending only on $d$ and $s$, such that, for all $m\in \mathbb{N}$, 
    \[a_m \leqslant C_2 \lambda_n^{-1} m^{-2s/d}.\]
    In particular, $\sum_{m\in \mathbb{N}} a_m < \infty$ if $s > d/2$.
    \label{prop:eigenvalue}
\end{prop}
\begin{proof}
    From the proof of Proposition \ref{prop:countable}, we know that there exists a constant $C_1 > 0$, depending on $d$ and $s$ such that $(1+\|k(m)\|_2^2)^{s/2} \geqslant \|k(m)\|_2^{s} \geqslant \|k(m)\|_1^{s} / d^{s} \geqslant C_1 m^{s/d}$. 
    Therefore, there exists a constant $C_3 > 0$, depending on $d$ and $s$, such that  \begin{equation}
        \sum_{|\alpha|\leqslant s} \Big(\frac{\pi}{2L}\Big)^{2|\alpha|} \prod_{j=1}^d k_j(m)^{2\alpha_j} \geqslant C_3 m^{2s/d}.
        \label{eq:controlEigenValues}
    \end{equation}
    Let $C_2 = 2C_3^{-1}$, and let us prove Proposition \ref{prop:eigenvalue} by contradiction. Thus, suppose that there is an integer $m$ such that $a_m > 2C_3^{-1} \lambda_n^{-1} m^{-2s/d}$. Then, for all $\ell \leqslant m$, $a_\ell^{-1} < C_3 \lambda_n m^{2s/d}/2$ and $C_3 \lambda_n m^{2s/d}/2 > B[v_\ell, v_\ell] \geqslant \lambda_n \|v_\ell \|_{H^s([-2L,2L]^d)}^2$. Thus, $\mathscr{V} = \mathrm{Span}(v_0, \hdots, v_m)$ is a subspace of $H^s_{\mathrm{per}}([-2L,2L]^d)$ of dimension $m+1$. In particular, for all $z \in \mathscr{V}$, there are weights $\beta_\ell \in \mathbb{R}$ such that $z = \sum_{j=0}^m \beta_\ell v_\ell$, and so $\|z \|_{L^2([-2L,2L]^d)}^2 = \sum_{\ell=0}^m \beta_\ell^2 $. Hence, 
    \begin{equation}
        \lambda_n \|z \|_{H^s([-2L,2L]^d)}^2 \leqslant B[z,z] = \sum_{\ell=0}^m \beta_j^2 a_\ell^{-1} \leqslant  \|z \|_{L^2([-2L,2L]^d)}^2 \lambda_n C_3 m^{2s/d}/2.
        \label{eq:bounded_energy}
    \end{equation}
    Let $S_s : H^s_{\mathrm{per}}([-2L,2L]^d) \to L^2([-2L,2L]^d)$ be the operator such that \[S_s(e_{k(\ell)}) = \Big(\sum_{|\alpha|\leqslant s} \Big(\frac{\pi}{2L}\Big)^{2|\alpha|} \prod_{j=1}^d k_j(\ell)^{2\alpha_j} \Big)^{1/2} e_{k(\ell)}.\] Then, by definition, $S_s$
    is diagonalizable on $H^s_{\mathrm{per}}([-2L,2L]^d)$ with eigenfunctions $e_{k(\ell)}$ and, for all $f \in H^s_{per}([-2L,2L]^d)$, $\|S_s(f)\|_{L^2([-2L,2L]^d)} = \|f\|_{H^s_{per}([-2L,2L]^d)}$. 
    Since $\dim \mathscr{V} = m+1$, Lemma \ref{lem:non_empty} ensures that $\mathscr{V} \cap \mathrm{Span}( e_{k(\ell)})_{\ell \geqslant m} \neq \emptyset$. However, any $z \in \mathrm{Span}( e_{k(\ell)})_{\ell \geqslant m}$ can be written $z = \sum_{j\geqslant m} \beta_\ell e_{k(\ell)}$, for weights $\beta_\ell \in \mathbb{R}$. Thus, using \eqref{eq:controlEigenValues}, we have that
    \begin{align*}
        \|z \|_{H^s([-2L,2L]^d)}^2 = \sum_{j\geqslant m} \beta_\ell^2 \|e_{k(\ell)} \|_{H^s([-2L,2L]^d)}^2 &\geqslant C_3 m^{2s/d}\sum_{\ell \geqslant m} \beta_\ell ^2\\
        &=  C_3 m^{2s/d}\|z \|_{L^2([-2L,2L]^d)}^2.
    \end{align*}
    Since, by assumption, $z \in \mathscr{V}$, this contradicts \eqref{eq:bounded_energy}.
\end{proof}

\begin{remark}[Lower bound on $a_m^{-1}$]
    Using similar arguments but bounding the eigenvalues of $S_s$ by $\sum_{|\alpha|\leqslant s} \Big(\frac{\pi}{2L}\Big)^{2|\alpha|} \prod_{j=1}^d k_j(\ell)^{2\alpha_j} \geqslant 1$, or directly applying the so-called Rayleigh's formula \citep[Chapter 6.5, Theorem 2]{evans2010partial}, one shows that, for all $m \geqslant 0$, $a_m^{-1} \geqslant \lambda_n$.
    \label{rem:rayleigh}
\end{remark}
Let $x \in [-2L,2L]^d$. Let $\delta_x$ be the Dirac distribution, i.e., the linear form on $C^0([-2L,2L]^d)$ such that, for all $f \in C^0([-2L,2L]^d)$, $\langle \delta_x, f\rangle = f(x)$. Notice that $\delta_x$ is continuous with respect to the $\|\cdot \|_\infty$ norm. In the sequel, with a slight abuse of notation, we replace $\delta_x(f)$ by $\langle \delta_x, f\rangle$. In effect, $\delta_x$ can be approximated by a regularizing sequence $(\xi^x_m)_{m\in \mathbb{N}}$ with respect to the $L^2([-2L,2L]^d)$ inner product, i.e., \[\forall f \in C^0([-2L,2L]^d), \quad \lim_{m\to\infty} \langle \xi^x_m, f\rangle_{L^2([-2L,2L]^d)} = f(x).\] Therefore, the action of $\delta_x$ on $f$ behaves like an inner product on $L^2([-2L,2L]^d)$, and this intuition will be fruitful in the next Proposition. Moreover, since $H^s_{\mathrm{per}}([-2L,2L]^d) \subseteq H^s([-2L,2L]^d) \subseteq C^0([-2L,2L]^d)$, when ``applied" to any $f \in H^s_{\mathrm{per}}([-2L,2L]^d)$, $\delta_x$ can be considered as the evaluation at $x$ of the unique continuous representation of $f$. The following proposition shows that $\mathscr{O}_n^{1/2}$ can be extended to $\delta_x$ in such a way that this extension stays self-adjoint.
\begin{prop}[Self-adjoint operator extension]
    Let $s>d/2$ and, for $x \in [-2L, 2L]^d$, let $\mathscr{O}_n^{1/2} (\delta_x) = \sum_{m\in \mathbb{N}} a_m^{1/2} v_m(x) v_m$. Then, almost everywhere in $x$ according to the Lebesgue measure on $[-2L, 2L]^d$, $\mathscr{O}_n^{1/2} (\delta_x) \in L^2([-2L,2L]^d)$ and, for all $f \in L^2([-2L,2L]^d)$,  
    \[\langle \mathscr{O}_n^{1/2}(f), \delta_x\rangle = \langle   f, \mathscr{O}_n^{1/2}(\delta_x) \rangle_{L^2([-2L,2L]^d)}.\] 
    \label{prop:extension}
\end{prop}
\begin{proof}
    Let $\psi_N(x,y) = \sum_{m=0}^N \alpha_m^{1/2} v_m(x) v_m(y)$. Then, for all $N_1 \leqslant N_2$,
    \begin{align*}
        &\int_{[-2L, 2L]^d}\int_{[-2L, 2L]^d} |\psi_{N_2}(x,y)-\psi_{N_1}(x,y)|^2dx dy \\
        &= \int_{[-2L, 2L]^d}\int_{[-2L, 2L]^d} \Big|\sum_{m=N_1+1}^{N_2} a_m^{1/2} v_m(x) v_m(y)\Big|^2dx dy\\
        &= \sum_{m, \ell =N_1+1}^{N_2} a_m \int_{[-2L, 2L]^d} v_m(x)v_\ell(x) dx \int_{[-2L, 2L]^d}v_m(y) v_\ell(y) dy\\
        &= \sum_{m=N_1+1}^{N_2} a_m \leqslant \sum_{m=N_1+1}^{\infty} a_m.
    \end{align*}
    Proposition \ref{prop:eigenvalue} shows that $\lim_{N_1 \to \infty}\sum_{m=N_1}^{\infty} a_m = 0$, hence  $(\psi_N)_{N\in \mathbb{N}}$ is a Cauchy sequence. Therefore, $\psi_\infty (x,y) = \sum_{m\in \mathbb{N}} a_m^{1/2} v_m(x) v_m(y)$ converges in $L^2([-2L,2L]^d \times [-2L,2L]^d)$ and 
    \begin{equation*}
        \int_{[-2L, 2L]^{2d}} |\psi_\infty(x,y)|^2dx dy  = \sum_{m\in \mathbb{N}} a_m.
    \end{equation*}
    Thus, by the Fubini-Lebesgue theorem, almost everywhere in $x$ according to the Lebesgue measure on $[-2L, 2L]^d$, one has $\mathscr{O}_n^{1/2}(\delta_x):= \psi_\infty(x,\cdot) \in L^2([-2L,2L]^d)$. Recall that, by definition, \[\mathscr{O}_n^{1/2} (f) = \sum_{m\in \mathbb{N}} a_m^{1/2}\langle f, v_m\rangle_{L^2([-2L,2L]^d)} v_m \in H^s_{\mathrm{per}}([-2L,2L]^d),\] so that $\langle \mathscr{O}_n^{1/2} (f), \delta_x\rangle = \sum_{m\in \mathbb{N}} a_m^{1/2}\langle f, v_m\rangle_{L^2([-2L,2L]^d)} v_m(x)$. Moreover, for any function $f \in L^2([-2L,2L]^d)$, 
    \begin{align*}
        &\int_{[-2L, 2L]^d} |\langle f, \psi_N(x, \cdot)\rangle_{L^2([-2L,2L]^d)} - \langle \mathscr{O}_n^{1/2} (f), \delta_x\rangle| dx\\
        &\quad = \int_{[-2L, 2L]^d}  \Big|\int_{[-2L, 2L]^d} f(y) \sum_{m=0}^N a_m^{1/2} v_m(x) v_m(y)dy \\
        & \qquad \qquad \qquad \quad - \sum_{m\in \mathbb{N}} a_m^{1/2}\langle f, v_m\rangle_{L^2([-2L,2L]^d)} v_m(x)\Big|dx\\
         &\quad = \int_{[-2L, 2L]^d}  \Big|\int_{[-2L, 2L]^d} f(y) \sum_{m > N} a_m^{1/2} v_m(x) v_m(y)\Big|dx dy \xrightarrow{ N \to \infty} 0.
    \end{align*}
    Therefore, since
    \begin{align*}
        &\int_{[-2L, 2L]^d} |\langle f, \psi_\infty(x, \cdot)\rangle_{L^2([-2L,2L]^d)} - \langle \mathscr{O}_n^{1/2} (f), \delta_x\rangle| dx \\
        &\quad \leqslant \int_{[-2L, 2L]^d} |\langle f, \psi_N(x, \cdot)\rangle_{L^2([-2L,2L]^d)} - \langle \mathscr{O}_n^{1/2} (f), \delta_x\rangle| dx  \\
        &\qquad + \int_{[-2L, 2L]^d} |\langle f, \psi_\infty(x, \cdot) - \psi_N(x, \cdot)\rangle_{L^2([-2L,2L]^d)} | dx,
    \end{align*}
    and since 
    \begin{align*}
        &\int_{[-2L, 2L]^d} |\langle f, \psi_\infty(x, \cdot) - \psi_N(x, \cdot)\rangle_{L^2([-2L,2L]^d)} | dx \\
        &\quad \leqslant \Big(\int_{[-2L, 2L]^d}\int_{[-2L, 2L]^d} |f(y)|^2dydx\Big)^{1/2} \\
        &\qquad \times \Big(\int_{[-2L, 2L]^d}\int_{[-2L, 2L]^d}| \psi_\infty(x,y) - \psi_N(x,y)|^2 dydx\Big)^{1/2} \\
        & \quad \xrightarrow{N \to \infty} 0,
    \end{align*} 
    we deduce that $\int_{[-2L, 2L]^d} |\langle f, \psi_\infty(x, \cdot)\rangle_{L^2([-2L,2L]^d)} - \langle \mathscr{O}_n^{1/2} (f), \delta_x\rangle| dx = 0$. Hence, almost everywhere in $x$ according to the Lebesgue measure on $[-2L, 2L]^d$,  we get that  the operator $\mathscr{O}_n^{1/2}$ is self-adjoint, i.e., 
    $\langle \mathscr{O}_n^{1/2} (f), \delta_x\rangle = \langle   f, \mathscr{O}_n^{1/2}(\delta_x)\rangle$.
\end{proof}

\subsection*{Proof of Theorem \ref{thm:PDE_kernel}}
Let $s>d/2$, $n\in\mathbb{N}$, $\lambda_n >0$, $\mu_n \geqslant 0$, and consider a linear partial differential operator $\mathscr{D}(u) = \sum_{|\alpha|\leqslant s} p_\alpha \partial^\alpha u$ of order $s$ such that $\max_\alpha \|p_\alpha\|_\infty < \infty$. Proposition \ref{prop:diff_op} and \ref{prop:scalar_product} show that there exists a compact self-adjoint differential operator $\mathscr{O}_n$ such that, for all $f \in H^s_{\mathrm{per}}([-2L,2L]^d)$, \[\|\mathscr{O}_n^{-1/2}(f)\|_{L^2([-2L,2L]^d)}^2 = \lambda_n \|f\|_{H^s([-2L,2L]^d)}^2+ \mu_n \|\mathscr{D}(f)\|_{L^2(\Omega)}^2.\]
Consider any target function $f \in H^s_{\mathrm{per}}([-2L,2L]^d)$. The Sobolev embedding theorem states that $H^s_{\mathrm{per}}([-2L,2L]^d) \subseteq H^s([-2L,2L]^d) \subseteq C^0([-2L,2L]^d)$. Thus, for all $x \in \Omega$, we have that $f(x) = \langle f, \delta_x\rangle$.
Proposition \ref{prop:scalar_product} ensures that $f(x) = \langle \mathscr{O}_n^{1/2}\mathscr{O}_n^{-1/2}(f), \delta_x\rangle$ and Proposition \ref{prop:extension} that, for almost every $x \in \Omega$ with respect to the Lebesgue measure, 
\[f(x) = \langle \mathscr{O}_n^{-1/2}(f), \mathscr{O}_n^{1/2}(\delta_x)\rangle_{L^2([-2L,2L]^d)},\]
with $\mathscr{O}_n^{-1/2}(f) \in L^2([-2L,2L]^d)$ and $\mathscr{O}_n^{1/2}(\delta_x) \in L^2([-2L,2L]^d)$. 
Proposition \ref{prop:scalar_product} shows that 
$\mathscr{O}_n^{1/2}(\delta_x) = \mathscr{O}_n^{-1/2}\mathscr{O}_n(\delta_x)$. Thus, 
\[f(x) = \langle f, \mathscr{O}_n(\delta_x)\rangle_{\mathrm{RKHS}},\]
where the RKHS inner product is defined by $\langle g, h\rangle_{\mathrm{RKHS}} = \langle \mathscr{O}_n^{-1/2}(g), \mathscr{O}_n^{-1/2}(h)\rangle_{L^2([-2L,2L]^d)}$.
Since $\mathscr{O}_n^{1/2}(\delta_x) \in L^2([-2L,2L]^d)$, Proposition \ref{prop:scalar_product} shows that $\mathscr{O}_n(\delta_x) \in H^s_{\mathrm{per}}([-2L,2L]^d)$.
We can therefore define the kernel
\begin{align*}
    K(x,y) &= \langle \mathscr{O}_n(\delta_x), \mathscr{O}_n(\delta_y)\rangle_{\mathrm{RKHS}} \\
    &= \langle \mathscr{O}_n^{1/2}(\delta_x), \mathscr{O}_n^{1/2}(\delta_y)\rangle_{L^2([-2L,2L]^d)}.
\end{align*}
Proposition \ref{prop:extension} ensures that $K(x,y) = \langle \mathscr{O}_n(\delta_x), \delta_y\rangle = \mathscr{O}_n(\delta_x)(y) = \sum_{m\in \mathbb{N}} a_m  v_m(x)v_m(y)$. Therefore, we know that $K(x,\cdot) \in H^s_{\mathrm{per}}([-2L,2L]^d)$, and we recognize the reproducing property stating that, for all $f \in H^s_{\mathrm{per}}([-2L,2L]^d)$ and all $x \in [-2L,2L]^d$, $f(x) = \langle f, K(x, \cdot)\rangle_{\mathrm{RKHS}}$.

\subsection*{Proof of Proposition \ref{prop:kernel_characterization}}
 Recall that $K(x,\cdot) = \mathscr{O}_n(\delta_x)$.
 It was proven in Proposition \ref{prop:extension} that $\mathscr{O}_n^{1/2}(\delta_x) \in L^2([-2L,2L]^d)$. By Proposition \ref{prop:scalar_product}, $\sum_{m=0}^N a_m^{1/2}\langle v_m, \mathscr{O}_n^{1/2}(\delta_x)\rangle_{L^2([-2L,2L]^d)} v_m$ converges in $H^s_{\mathrm{per}}([-2L,2L]^d)$ to  $K(x, \cdot)$. Let $\phi \in H^s_{\mathrm{per}}([-2L,2L]^d)$ be a test function. Since $B$ is continuous on $H^s_{\mathrm{per}}([-2L,2L]^d)$, 
 \[\lim_{N\to \infty} B\Big[\sum_{m=0}^N a_m^{1/2}\langle v_m, \mathscr{O}_n^{1/2}(\delta_x)\rangle_{L^2([-2L,2L]^d)}v_m, \phi\Big] = B[K(x, \cdot), \phi].\]
    Then,
    \begin{align*}
        &B\Big[\sum_{m=0}^N a_m^{1/2}\langle v_m, \mathscr{O}_n^{1/2}(\delta_x)\rangle_{L^2([-2L,2L]^d)}v_m, \phi\Big] \\
        &\quad = \sum_{m=0}^N a_m^{1/2}\langle v_m, \mathscr{O}_n^{1/2}(\delta_x)\rangle_{L^2([-2L,2L]^d)}B[v_m, \phi] \quad  \hbox{(by bilinearity)}\\
         &\quad = \sum_{m=0}^N a_m^{-1/2}\langle v_m, \mathscr{O}_n^{1/2}(\delta_x)\rangle_{L^2([-2L,2L]^d)} \langle v_m, \phi \rangle_{L^2([-2L,2L]^d)} \quad  \hbox{(since $v_m$ is an eigenfunction)}\\
         &\quad = \sum_{m=0}^N a_m^{-1/2}\langle \mathscr{O}_n^{1/2}(v_m), \delta_x\rangle \langle v_m, \phi \rangle_{L^2([-2L,2L]^d)} \quad  \hbox{(by Proposition \ref{prop:extension})}\\
         &\quad = \sum_{m=0}^N \langle v_m, \phi \rangle_{L^2([-2L,2L]^d)} v_m(x).
    \end{align*}
    Notice that the expression above is the decomposition of $\phi$ on the $v_m$ basis. We conclude, as desired, that
    $B[K(x, \cdot), \phi] = \lim_{N\to \infty} B[\sum_{m=0}^N a_m^{1/2}\langle v_m, \mathscr{O}_n^{1/2}(\delta_x)\rangle_{L^2([-2L,2L]^d)}v_m, \phi] = \phi(x)$.

\section{Integral operator and eigenvalues}
\subsection*{Compactness of $C\mathscr{O}_nC$}
\begin{lem}[Compactness]
    The operator $C\mathscr{O}_nC: L^2([-2L, 2L]^d) \to L^2([-2L, 2L]^d)$ is positive, compact, and self-adjoint.
    \label{lem:compactness}
\end{lem}
\begin{proof}
    Since $C$ is a self-adjoint projector, then, for all $f \in L^2([-2L, 2L]^d)$, $\|C(f)\|_{L^2([-2L,2L]^d)}^2 \leqslant \|f\|_{L^2([-2L,2L]^d)}^2$. 
    Thus, for any bounded sequence $(f_m)_{m\in \mathbb{N}}$ in $L^2([-2L, 2L]^d)$, the sequence $(C(f_m))_{m\in \mathbb{N}}$ is bounded.
    Since $\mathscr{O}_n$ is compact, upon passing to a subsequence, $(\mathscr{O}_nC(f_m))_{m\in \mathbb{N}}$ converges to $f_\infty \in L^2([-2L, 2L]^d)$.
    Therefore, $\lim_{m\to \infty}\|C\mathscr{O}_nC(f_m) - C(f_\infty)\|_{L^2([-2L,2L]^d)}^2 \leqslant \lim_{m\to \infty}\|\mathscr{O}_nC(f_m) - f_\infty\|_{L^2([-2L,2L]^d)}^2 = 0$, i.e., $(C\mathscr{O}_nC(f_m))_{m\in \mathbb{N}}$ converges to the function $C(f_\infty) \in L^2([-2L, 2L]^d)$. 
    So, $C\mathscr{O}_nC: L^2([-2L, 2L]^d) \to L^2([-2L, 2L]^d)$ is a compact operator.
    Moreover, given any $f \in L^2([-2L, 2L]^d)$, we have that $\langle f, C\mathscr{O}_nC (f)\rangle_{L^2([-2L,2L]^d)} = \| \mathscr{O}_n^{1/2} C(f)\|_{L^2([-2L,2L]^d)}^2 \geqslant 0$, which means that $C\mathscr{O}_nC$ is positive. Finally, $C\mathscr{O}_nC$ is self-adjoint, since $C$ and $\mathscr{O}_n$ are self-adjoint.
\end{proof}

\subsection*{Proof of Theorem \ref{thm:eigenvalues}}
For clarity, the proof is divided into 4 steps. Steps 1 and 2 ensures that we can apply the Courant Fischer min-max theorem to the integral operator. 
Step 3 connects the Courant Fischer estimates of $L_K$ and $C\mathscr{O}_nC$. Finally, Step 4 establishes the result on the eigenvalues.

\paragraph{Step 1: Compactness of the integral operator.}
Let $L_{K, \mathscr{U}}$ be the integral operator
associated with the uniform distribution on $\Omega$, i.e.,
    \[\forall f \in L^2(\Omega), \forall x \in \Omega, \quad L_{K, \mathscr{U}}f(x) = \frac{1}{|\Omega|}\int_{\Omega} K(x,y) f(y) dy.\]
Since $K(x,y) = \sum_{m\in \mathbb{N}}a_m(\mathscr{O}_n) v_m(x) v_m(y)$, $\int_{[-2L,2L]^d} v_\ell v_m  = \mathbf{1}_{\ell = m}$, and $\sum_{m\in \mathbb{N}}a_m(\mathscr{O}_n) < \infty$, the Fubini-Lebesgue theorem states that
\[\int_{\Omega^2} |K(x,y)|^2 dx dy \leqslant \int_{[-2L,2L]^{2d}} |K(x,y)|^2 dx dy = \sum_{m\in \mathbb{N}}a_m^2(\mathscr{O}_n) < \infty,\] which implies that 
$L_{K, \mathscr{U}}$ is a Hilbert-Schmidt operator \citep[][Lemma 8.20]{renardy2004an}.
As a consequence, $L_{K, \mathscr{U}}$ is compact
\citep[][Theorem 8.83]{renardy2004an}.
Observe that $L_K f = L_{K, \mathscr{U}} f \frac{d\mathbb{P}_X}{dx}$. Let $C_2 >0$. Given any sequence $(f_n)_{n\in \mathbb{N}}$ such that $\|f_n\|_{L^2(\Omega, \mathbb{P}_X)} \leqslant C_2$, then, clearly, $\|f_n  \frac{d\mathbb{P}_X}{dx}\|_{L^2(\Omega)} \leqslant \kappa C_2$. This shows that the sequence $(f_n \frac{d\mathbb{P}_X}{dx})_{n\in \mathbb{N}}$ is bounded in $L^2(\Omega)$. Thus, since $L_{K, \mathscr{U}}$ is compact, upon passing to a  subsequence, $L_{K, \mathscr{U}}(f_n \frac{d\mathbb{P}_X}{dx}) = L_{K}(f_n)$ converges in $L^2(\Omega)$, and therefore in $L^2(\Omega, \mathbb{P}_X)$. This shows that the integral operator $L_K$ is compact.

\paragraph{Step 2: Courant Fischer min-max theorem.} Using $\frac{d\mathbb{P}_X}{dx} \leqslant \kappa$ and letting $\mathrm{FS}$ be the Fourier series operator, i.e., $\mathrm{FS}(f)(k) = \langle f, \exp(-i\frac{\pi}{2L}\langle k, \cdot\rangle)\rangle_{L^2([-2L,2L]^d)}$, we see that for all $f \in L^2(\Omega, \mathbb{P}_X)$,
\begin{align*}
&\lim_{n \to \infty} \Big\|f - \sum_{\|k\|_2 \leqslant n} \mathrm{FS}(f)(k) \exp(i \pi L^{-1} \langle k, \cdot\rangle)\Big\|_{L^2(\Omega, \mathbb{P}_X)} \\
& \quad \leqslant \kappa \lim_{n \to \infty}\Big\|f - \sum_{k \in \mathbb{Z}^d,\; \|k\| \leqslant n} \mathrm{FS}(f)(k) \exp(i \pi L^{-1} \langle k, \cdot\rangle)\Big\|_{L^2(\Omega)} = 0.
\end{align*}
Therefore, the Gram-Schmidt algorithm applied to the $(\exp(i \pi L^{-1} \langle k, \cdot\rangle))_{k \in \mathbb{Z}^d}$ family provides a Hermitian basis of $L^2(\Omega, \mathbb{P}_X)$. In particular, the space $L^2(\Omega, \mathbb{P}_X)$ is separable.
Since $L_K$ is a positive compact self-adjoint operator on $L^2(\Omega, \mathbb{P}_X)$, Theorem \ref{thm:spectral} and \ref{thm:courant_fischer} show that $L_K$ is diagonalizable with positive eigenvalues $(a_n(L_K))_{n \in \mathbb{N}}$, with 
\[a_n(L_K) = \underset{\dim \Sigma = n}{\underset{\Sigma \subseteq L^2(\Omega, \mathbb{P}_X)}{\max}} 
\underset{f \neq 0}{\min_{f \in \Sigma}} \|f\|_{L^2(\Omega, \mathbb{P}_X)}^{-2}\langle f, L_K f\rangle_{L^2(\Omega, \mathbb{P}_X)} . \]

\paragraph{Step 3: Switching integrals.}
Observe that, for all  $f\in L^2(\Omega, \mathbb{P}_X)$, 
\begin{align*}
    &\int_{\Omega^2} \sum_{m\in \mathbb{N}} a_m(\mathscr{O}_n) |f(x)| |f(y)| |v_m(x)| |v_m(y)| d\mathbb{P}_X(x) d\mathbb{P}_X(y)\\ &= \sum_{m \in \mathbb{N}}  a_m(\mathscr{O}_n) \Big(\int_{\Omega} |f(x)||v_m(x)|d\mathbb{P}_X(x)\Big)^2\\
    &\leqslant \sum_{m \in \mathbb{N}}  a_m(\mathscr{O}_n) \|f\|_{L^2(\Omega, \mathbb{P}_X)}^2 \int_{\Omega} |v_m(x)|^2 d\mathbb{P}_X(x)\\
    &\leqslant \|f\|_{L^2(\Omega, \mathbb{P}_X)}^2  \kappa \sum_{m \in \mathbb{N}}  a_m(\mathscr{O}_n) < \infty.
\end{align*}
In the last inequality, we used the fact that $\int_{\Omega} |v_m(x)|^2 d\mathbb{P}_X(x) \leqslant \kappa \int_{[-2L,2L]^d} |v_m(x)|^2 dx = \kappa$.
Therefore, according to the Fubini-Lebesgue theorem,
\begin{align*}
    \langle f, L_K f\rangle_{L^2(\Omega, \mathbb{P}_X)} &= \int_{\Omega^2} f(x) \Big(\sum_{m\in \mathbb{N}}  a_m(\mathscr{O}_n) v_m(x) v_m(y)\Big)  f(y)d\mathbb{P}_X(y) d\mathbb{P}_X(x)\\
    &= \sum_{m\in \mathbb{N}}  a_m(\mathscr{O}_n) \Big(\int_{\Omega^2} f(x)v_m(x) d\mathbb{P}_X(x)\Big)^2\\
    &= \Big\|\mathscr{O}_n^{1/2} \Big( f\frac{d\mathbb{P}_X}{dx}\Big)\Big\|_{L^2([-2L,2L]^d)}^2.
\end{align*}

\paragraph{Step 4: Comparison using Courant Fischer.}
Let $z =  f\frac{d\mathbb{P}_X}{dx}$. By noting that
 $f \frac{d\mathbb{P}}{dx} = f \frac{d\mathbb{P}}{dx} \mathbf{1}_\Omega$, we see that $Cz = z$ and $\langle f, L_K f\rangle_{L^2(\Omega, \mathbb{P}_X)} = \langle z, C\mathscr{O}_nC(z)\rangle_{L^2([-2L,2L]^d)}$. Therefore, for any $\Sigma \subseteq L^2(\Omega, \mathbb{P}_X)$, we have  
\begin{align*}
    \underset{f \neq 0}{\min_{f \in \Sigma}} \|f\|_{L^2(\Omega, \mathbb{P}_X)}^{-2}\langle f, L_K f \rangle_{L^2(\Omega, \mathbb{P}_X)} &= \underset{f \neq 0}{\min_{f \in \Sigma}}\|f\|_{L^2(\Omega, \mathbb{P}_X)}^{-2}\Big\langle f \frac{d\mathbb{P}_X}{dx}, C\mathscr{O}_nC\Big(f \frac{d\mathbb{P}_X}{dx} \Big)\Big\rangle_{L^2([-2L,2L]^d)}\\
    &\leqslant \underset{z \neq 0}{\min_{z \in \frac{d\mathbb{P}_X}{dx}\Sigma}} \kappa \|z\|_{L^2([-2L,2L]^d)}^{-2}\langle z, C\mathscr{O}_nC(z)\rangle_{L^2([-2L,2L]^d)},
\end{align*}
where the inequality is a consequence of $\|z\|_{L^2([-2L,2L]^d)}^2 = \int_\Omega |f|^2 (\frac{d\mathbb{P}_X}{dx})^2 \leqslant \kappa \|f\|_{L^2(\Omega, \mathbb{P}_X)}^{2}$.
Using $\frac{d\mathbb{P}_X}{dx} L^2([-2L,2L]^d) \subseteq L^2([-2L,2L]^d)$, we conclude that
\begin{align}
    &\underset{\dim \Sigma = m}{\underset{\Sigma \subseteq L^2(\Omega, \mathbb{P}_X)}{\max}} 
\underset{f \neq 0}{\min_{f \in \Sigma}}\|f\|_{L^2(\Omega, \mathbb{P}_X)}^{-2}\langle f
, L_K f\rangle_{L^2(\Omega, \mathbb{P}_X)} \nonumber\\
&\quad \leqslant \kappa \underset{\dim \Sigma = m}{\underset{\Sigma \subseteq L^2([-2L,2L]^d)}{\max}} 
\underset{z \neq 0}{\min_{z \in \Sigma}}\|z\|^{-2}_{L^2([-2L,2L]^d)}\langle z,  C\mathscr{O}_nC(z) \rangle_{L^2([-2L,2L]^d)}.
\label{eq:courantFischerComparison}
\end{align}
According to Lemma \ref{lem:compactness}, the operator $C\mathscr{O}_nC$ is compact, self-adjoint, and positive, and thus its eigenvalues are given by the Courant-Fischer min-max theorem. Remark that the left-hand side (resp.\ the right-hand term) of inequality \eqref{eq:courantFischerComparison} corresponds to the Courant-Fischer min-max characterization of the $m$th eigenvalue of $L_K$ (resp.\ $C\mathscr{O}_nC$).
Therefore, we deduce that $a_m(L_K) \leqslant \kappa a_m(C\mathscr{O}_nC)$.

\subsection*{Bounding the kernel}
The goal of this section is to upper bound the kernel $K(x,y)$ defined in Theorem \ref{thm:PDE_kernel}.

\begin{prop}[Partial continuity of the kernel]
    Let $x,y \in [-2L,2L]^d$. Both functions $K(x,\cdot)$ and $K(\cdot, y)$ are continuous. 
\end{prop}
    
\begin{proof}
     It is shown in the proof of  Proposition \ref{prop:extension} that $\psi_\infty(x,y) := \sum_{m\in\mathbb{N}} a_m^{1/2} v_m(x) v_m(y)$  converges in $L^2([-2L,2L]^d \times [-2L,2L]^d)$, that $\int_{[-2L, 2L]^{2d}} |\psi_\infty(x,y)|^2dx dy  = \sum_{m\in \mathbb{N}} a_m$, and that $\psi_\infty(x, \cdot) := \sum_{m\in\mathbb{N}} a_m^{1/2} v_m(x) v_m$ converges in $L^2([-2L,2L]^d)$ almost everywhere in $x$.
    By definition, $K(x, \cdot) = \mathscr{O}_n^{1/2}\psi_\infty(x, \cdot)$. Using Proposition \ref{prop:scalar_product}, this implies  \[\|K(x, \cdot)\|_{H^s_{\mathrm{per}}([-2L,2L]^d)}^2 \leqslant \lambda_n^{-1} \|\psi_\infty(x, \cdot)\|_{L^2([-2L,2L]^d)}^2.\]
    The Sobolev embedding theorem then ensures that $K(x, \cdot)$ is continuous for any $x$. One shows with the same argument that $K(\cdot, y)$ is continuous for any $y$.
\end{proof}

\begin{lem}[Trace reconstruction]
    Let $z \in [-2L,2L]^d$. Let $(\psi_\ell)_{\ell \in \mathbb{N}}$ be a sequence of functions in $L^2([-2L,2L]^d)$ such that $\int_{[-2L,2L]^d} \psi_\ell^2 =1$ and $\lim_{\ell \to \infty}\psi_\ell = \delta_z$. Then 
    \[\lim_{\ell \to \infty} \int_{[-2L,2L]^{2d}} K(x,y) \psi_\ell(x) \psi_\ell(y) dx dy = K(z,z). \]
    \label{lem:trace_rec}
\end{lem}
\begin{proof}
    \begin{align*}
        &\Big|\int_{[-2L,2L]^{2d}} K(x,y) \psi_\ell(x) \psi_\ell(y) dx dy - \int_{[-2L,2L]^{d}} \psi_\ell(x) K(x,z) dx \Big|\\
        &\quad =\Big| \int_{[-2L,2L]^{d}} \psi_\ell(x) \Big(K(x,z) -\int_{[-2L,2L]^{d}} K(x,y)  \psi_\ell(y)dy\Big) dx\Big|\\
        &\quad \leqslant \Big(\int_{[-2L,2L]^{d}} \psi_\ell^2(x) dx \Big)^{1/2} \Big(\int_{[-2L,2L]^{d}} \Big(K(x,z) -\int_{[-2L,2L]^{d}} K(x,y)  \psi_\ell(y)dy\Big)^2 dx \Big)^{1/2}.
    \end{align*}
    Recall that $\int_{[-2L,2L]^{d}} \psi_\ell^2(x) dx = 1$ and $\lim_{\ell \to \infty}\int_{[-2L,2L]^{d}} K(x,y)  \psi_\ell(y)dy = K(x,z)$.
    Let \[g_\ell(x) = \bigg(K(x,z) -\int_{[-2L,2L]^{d}} K(x,y)  \psi_\ell(y)dy\bigg)^2.\] Notice that 
    \begin{align*}
        |g_\ell(x)| &\leqslant 2 K^2(x,z) + 2 \Big|\int_{[-2L,2L]^{d}} K(x,y)  \psi_\ell(y)dy\Big|^2\\
        &\leqslant 2 K^2(x,z) + 2 \int_{[-2L,2L]^{d}} K^2(x,y)dy, 
    \end{align*}
    where we use the Cauchy-Schwarz inequality 
    \[\Big|\int_{[-2L,2L]^{d}} K(x,y)  \psi_\ell(y)dy\Big|^2 \leqslant \int_{[-2L,2L]^{d}} K^2(x,y) dy \times \int_{[-2L,2L]^{d}} \psi_\ell^2(y) dy\]
    and $\int_{[-2L,2L]^{d}} \psi_\ell^2(y) dy = 1$. Moreover, for almost every $z$, 
    \begin{align*}  
    &\int_{[-2L,2L]^{d}} \Big(2 K^2(x,z) + 2 \int_{[-2L,2L]^{d}} K^2(x,y)dy\Big)dx \\
 & \quad \leqslant 2 \int_{[-2L,2L]^{d}} K^2(x,z)dx + 2 \int_{[-2L,2L]^{2d}} K^2(x,y)dxdy 
 < \infty.
 \end{align*}
 Therefore, using the dominated convergence theorem, we see that $\lim_{\ell \to \infty} \int_{[-2L,2L]^{d}} g_\ell (x)dx = \int_{[-2L,2L]^{d}} \lim_{\ell \to \infty}  g_\ell (x)dx$. Since $\lim_{\ell \to \infty}\psi_\ell = \delta_z$, by the partial continuity of the kernel, we know that $\lim_{\ell \to \infty}  g_\ell (x) = 0$. So, 
 \[\lim_{\ell \to \infty} \Big|\int_{[-2L,2L]^{2d}} K(x,y) \psi_\ell(x) \psi_\ell(y) dx dy - \int_{[-2L,2L]^{d}} \psi_\ell(x) K(x,z) dx \Big| = 0,\]
 and
 \[\lim_{\ell \to \infty} \int_{[-2L,2L]^{2d}} K(x,y) \psi_\ell(x) \psi_\ell(y) dx dy = K(z,z). \]
\end{proof}
\begin{prop}[Bounding the kernel]
    Let $z \in [-2L,2L]^d$. One has $|K(z,z)|\leqslant \lambda_n^{-1}$.
    \label{prop:bound_kernel}
\end{prop}
\begin{proof}
    As in the proof of Theorem \ref{thm:eigenvalues}, it is easy to show that the operator $  L: f \mapsto (x \mapsto \int_{[-2L,2L]^d} K(x,y) f(y) dy)$ is compact and that $\langle f, L(f)\rangle_{L^2([-2L,2L]^d)}  = \langle f, \mathscr{O}_n(f)\rangle_{L^2([-2L,2L]^d)}$. Thus, the eigenvalues of $L$ are upper bounded by those of $\mathscr{O}_n$, and in turn, using Remark \ref{rem:rayleigh}, by $\lambda_n^{-1}$. Lemma \ref{lem:trace_rec} states that \[ \lim_{\ell \to \infty} \langle \psi_\ell, L(\psi_\ell)\rangle_{L^2([-2L,2L]^d)} = K(z,z).\]  Thus, the Courant-Fischer min-max theorem states that $\langle \psi_\ell, L(\psi_\ell)\rangle_{L^2([-2L,2L]^d)} \leqslant \lambda_n^{-1}$, and that $K(z,z) \leqslant \lambda_n^{-1}$.
\end{proof}

\subsection*{Proof of Theorem \ref{prop:eigenfunction}}

For clarity, the proof will be divided into three steps.

\paragraph{Step 1: Weak formulation.}
 According to Lemma \ref{lem:compactness}, the operator $C\mathscr{O}_n C$ can be diagonalized in an orthonormal basis. Therefore, there are eigenfunctions $v_m \in L^2([-2L,2L]^d)$ and eigenvalues $a_m$ such that 
    \[ C\mathscr{O}_n C(v_m) = a_m v_m.\]
    Define $w_m = \mathscr{O}_n C(v_m)$. Given that $C(v_m) \in L^2([-2L,2L]^d)$, Proposition \ref{prop:diff_op} shows that $w_m\in H^s_{\mathrm{per}}([-2L,2L]^d)$. 
    Notice that $Cw_m = a_m v_m$. Since $C^2 = C$, we have \[v_m = C(v_m) = a_m^{-1} C(w_m).\] 
    By definition of the operator $\mathscr{O}_n$, for any test function $\phi \in H^s_{\mathrm{per}}([-2L,2L]^d)$, 
    \[B[w_m, \phi] = \langle C(v_m), \phi\rangle_{L^2([-2L,2L]^d)} = a_m^{-1} \langle C(w_m), \phi\rangle_{L^2([-2L,2L]^d)}.\]
    This means that $w_m$ is a weak solution to the PDE
    \[  \lambda_n \sum_{|\alpha|\leqslant s} \int_{[-2L, 2L]^d}\partial^\alpha \phi\; \partial^\alpha w_m + \mu_n \int_{\Omega}\mathscr{D} \phi \; \mathscr{D}  w_m = a_m^{-1}\int_{\Omega} \phi w_m. \]
    This proves \eqref{eq:weak_pde}.

\paragraph{Step 2: PDE in $\Omega$.}
    Next, for any Euclidian ball $\mathscr{B} \subseteq \Omega$
    and any function $\phi \in C^\infty(\Omega)$ with compact support in $\Omega$,  $w_m$ is a weak solution to the PDE
    \[  \lambda_n \sum_{|\alpha|\leqslant s} \int_{\mathscr{B}}\partial^\alpha \phi\; \partial^\alpha w_m + \mu_n \int_{\mathscr{B}}\mathscr{D} \phi \; \mathscr{D}  w_m = a_m^{-1}\int_{\mathscr{B}} \phi w_m. \]
    Noting that the ball, as a smooth manifold, is already its own map  with the canonical coordinates. The principal symbol \citep[see, e.g., Chapter 2.9][]{taylor2010partial} of this PDE is defined for all $x \in \Omega$ and $\xi \in \mathbb{R}^d$ by \[\sigma(x, \xi) = \lambda_n (-1)^s \sum_{|\alpha|=2s}\xi^{2\alpha} + \mu_n (-1)^s\sum_{|\alpha|=2s} p_\alpha(x)^2 \xi^{2\alpha},\] where  $\xi^{2\alpha} = \prod_{j=1}^d \xi_j^{2\alpha_j}$. Clearly, $|\sigma(x, \xi)| \neq 0$ whenever $\xi \neq 0$. Hence, the symbol function defined by $u \mapsto \sigma(x,\xi) \times u$ is an isomorphism from $\mathbb R$ to $\mathbb R$ whenever $\xi \neq 0$. This is the definition of a general elliptic PDE. Since $\mathscr{B}$ is a smooth manifold with $C^\infty$-boundary and  $p_\alpha \in C^\infty(\bar \Omega)$, the elliptic regularity theorem \citep[][Chapter 5, Theorem 11.1]{taylor2010partial} states that $w_m \in C^\infty(\mathscr{B})$. Therefore, $w_m \in C^\infty(\Omega)$. Overall,
    \[\forall x \in \Omega, \quad  \lambda_n\sum_{|\alpha|\leqslant s}(-1)^{|\alpha|} \partial^{2\alpha} w_m(x) + \mu_n \mathscr{D}^\ast \mathscr{D} w_m(x) = a_m^{-1} w_m(x).\]
    This proves $(i)$.

\paragraph{Step 3: PDE outside $\Omega$.}
    To show the second statement of the proposition, fix $\varepsilon > 0$ such that $d(\Omega, \partial [-2L,2L]^d) > \varepsilon$. Observe that any function $\phi \in C^\infty(]-2L-\varepsilon, 2L+\varepsilon[^d \backslash \bar\Omega)$ with compact support in $]-2L-\varepsilon, 2L+\varepsilon[^d \backslash \bar\Omega$ can be linearly mapped into the function $\tilde \phi(x) = \sum_{k \in (4L\mathbb{Z})^d} \phi(x + k) $ in $H^s_{\mathrm{per}}([-2L,2L]^d)$. This function $\tilde \phi$ is such that, for any $u \in L^2([-2L,2L]^d)$, $\int_{]-2L-\varepsilon, 2L+\varepsilon[^d} \phi u = \int_{[-2L, 2L]^d} \tilde \phi u$.  
    We deduce that, for any ball $\mathscr{B}$ included in $]-2L-\varepsilon, 2L+\varepsilon[^d \backslash \bar\Omega$, for any function $\phi \in C^\infty(]-2L-\varepsilon, 2L+\varepsilon[^d \backslash \bar\Omega)$ with compact support in $]-2L-\varepsilon, 2L+\varepsilon[^d \backslash \bar\Omega$, $w_m$ is a weak solution to the PDE
    \[  \lambda_n \sum_{|\alpha|\leqslant s} \int_{\mathscr{B}}\partial^\alpha \phi\; \partial^\alpha w_m  = \lambda_n \sum_{|\alpha|\leqslant s} \int_{[-2L,2L]^d}\partial^\alpha \tilde \phi\; \partial^\alpha w_m = 0. \]
    This PDE is elliptic and $\mathscr{B}$ is a smooth manifold with $C^\infty$-boundary. Therefore, the elliptic regularity theorem \citep[][Chapter 5, Theorem 11.1]{taylor2010partial} states that $w_m \in C^\infty(\mathscr{B})$. So, $w_m \in C^\infty([-2L, 2L]^d \backslash \bar\Omega)$ and
    \[\forall x \in [-2L, 2L]^d \backslash \bar \Omega,\quad  \lambda_n\sum_{|\alpha|\leqslant s}(-1)^{|\alpha|} \partial^{2\alpha} w_m(x)  = 0.\]
    This proves $(ii)$.
   

\subsection*{High regularity in dimension 1}
In this section, we assume that $d=1$, $s \geqslant 1$, $p_\alpha \in C^\infty(\bar \Omega)$, and the domain $\Omega$ is a segment, i.e., $\Omega = [L_1, L_2] \subseteq [-L,L]$ for some $-L \leqslant L_1, L_2 \leqslant L $.
\begin{prop}[Regularity of the eigenfunctions of $C\mathscr{O}_nC$]
The functions $(w_m)_{N\in\mathbb{N}}$ of Theorem \ref{prop:eigenfunction} associated with non-zero eigenvalues satisfy the following properties:
     \begin{itemize}
         \item[$(i)$] $w_m \in C^{s-1}([-2L,2L])$,
         \item[$(ii)$] $w_m|_\Omega \in C^{\infty}(\bar \Omega)$,
         \item[$(iii)$] $w_m|_{\Omega^c} \in C^{\infty}(\bar{\Omega^c})$.
     \end{itemize}
     \label{prop:1dreg}
\end{prop}
\begin{proof}
    Since $d = 1$ and $w_m \in H^s([-2L,2L])$, the Sobolev embedding theorem states that $w_m \in C^{s-1}([-2L,2L])$. Moreover, since  $w_m \in C^\infty(\Omega)$, since 
    \[\mathscr{D}^\ast \mathscr{D} u = \sum_{\alpha= 0}^s p_\alpha \Big(\frac{d}{dt}\Big)^\alpha \Big(\sum_{\tilde \alpha = 0}^s p_{\tilde \alpha} \Big(\frac{d}{dt}\Big)^{\tilde \alpha} u\Big)\] 
    is a linear differential operator with coefficients in $C^\infty(\bar \Omega)$, and since $w_m$ is the solution to the ordinary differential equation
        \[\forall x \in \Omega, \quad  \lambda_n\sum_{j =1}^s(-1)^{j} \frac{d^j}{dt^j} w_m(x) + \mu_n \mathscr{D}^\ast \mathscr{D} w_m(x) = a_m^{-1} w_m(x),\]
        the Picard-Lindelöf theorem (or the Grönwall inequality) ensures that $w_m|_\Omega \in C^\infty(\bar \Omega)$. Similarly, since $w_m \in C^\infty([-2L, 2L]^d \backslash \bar\Omega)$ and 
        \[\forall x \in [-2L, 2L]^d \backslash \bar \Omega,\quad  \sum_{j =1}^s(-1)^{j} \frac{d^j}{dt^j}  w_m  = 0,\]
        we have $w_m|_{\Omega^c} \in C^{\infty}(\bar{\Omega^c})$.
\end{proof}
\begin{remark} 
   As a by-product, the limits $\lim_{\genfrac{}{}{0 pt}{2}{x \to L_1 }{ x > L_1}} w_m(x)$, $\lim_{\genfrac{}{}{0 pt}{2}{x \to L_2}{x < L_2}} w_m(x)$, $\lim_{\genfrac{}{}{0 pt}{2}{x \to L_1 }{ x < L_1}} w_m(x)$, and $\lim_{\genfrac{}{}{0 pt}{2}{x \to L_2 }{ x > L_2}} w_m(x)$ exist.
\end{remark}

\section{From eigenvalues of the integral operator to minimax convergence rates}
\subsection*{Effective dimension}
We recall that the effective dimension $\mathscr{N}$ of the kernel $K$ is defined by
\[
\mathscr{N}(\lambda_n, \mu_n) = \mathrm{tr}(L_{K}  (\mathrm{Id} + L_{K})^{-1}),
\]
where $\mathrm{Id}$ is the identity operator and the symbol $\mathrm{tr}$ stands for the trace, i.e., the sum of the eigenvalues \citep{caponnetto2007optimal}. So,
\begin{align*}
    \mathscr{N}(\lambda_n, \mu_n) &= \mathrm{tr}(L_{K} \times (\mathrm{Id} + L_{K})^{-1})\\
    &= \sum_{m\in \mathbb{N}} \frac{ a_m(L_K)}{1+a_m(L_K)}\\
    &= \sum_{m\in \mathbb{N}} \frac{ 1}{1+a_m(L_K)^{-1}},
\end{align*}
where $a_m(L_K)$ stands for the eigenvalues of the operator $L_K$. The second equality is a consequence of the fact that $\mathrm{Id}$ and $L_{K}$ are co-diagonalizable, and so are $\mathrm{Id}$, $L_{K}$, and $(\mathrm{Id} + L_{K})^{-1}$.
\begin{lem}
    Assume that $\frac{d\mathbb{P}}{dx} \leqslant \kappa$. Then 
\begin{equation*}
    \mathscr{N}(\lambda_n, \mu_n) \leqslant \sum_{m \in \mathbb N} \frac{ 1}{1+(\kappa a_m(C\mathscr{O}_nC))^{-1}}.
\end{equation*}
\label{lem:eff_dim}
\end{lem}
\begin{proof}
     Apply Theorem \ref{thm:eigenvalues} and observe that $0 < a_m(L_K) \leqslant \kappa a_m(C\mathscr{O}_nC) \Leftrightarrow a_m(L_K)^{-1} \geqslant  (\kappa a_m(C\mathscr{O}_nC))^{-1} \Leftrightarrow 1+ a_m(L_K)^{-1} \geqslant  1+ (\kappa a_m(C\mathscr{O}_nC)) ^{-1} \Leftrightarrow (1+ a_m(L_K))^{-1} \leqslant  (1+ (\kappa a_m(C\mathscr{O}_nC))^{-1})^{-1}$.
\end{proof}

\subsection*{Lower bound on the eigenvalues of the integral kernel}

\begin{lem}[Explicit computation of $\mathscr{O}_n^{-1}$]
    Let $f \in C^\infty(\Omega)$ with compact support in $\Omega$. Then 
    \[\mathscr{O}_n^{-1}(f) = \lambda_n \sum_{|\alpha|\leqslant s} (-1)^{|\alpha|} \partial^{2\alpha} f + \mu_n \mathscr{D}^\ast \mathscr{D} f. \]
    \label{lem:explicit_calculus}
\end{lem}
\begin{proof}
    Let $\phi \in H^s_{\mathrm{per}}([-2L,2L]^d)$ be a test function. Since  the successive derivatives of $f$ are smooth with compact support, by definition of the weak derivatives of $\phi$, we may write
    \begin{align*}
        \lambda_n \sum_{|\alpha|\leqslant s}\int_{[-2L,2L]^d} \partial^{\alpha} f \partial^{\alpha}\phi = \int_{[-2L,2L]^d} \Big(\lambda_n \sum_{|\alpha|\leqslant s} (-1)^{|\alpha|} \partial^{2\alpha} f\Big) \phi.
    \end{align*}
    Moreover, because the support of $f$ is included in $\Omega$, we have that
    \[  \mu_n  \int_{\Omega} \mathscr{D} f \;\mathscr{D} \phi = \mu_n  \int_{[-2L,2L]^d} \mathscr{D} f \;\mathscr{D} \phi = \mu_n\int_{[-2L,2L]^d} ( \mathscr{D}^\ast \mathscr{D} f)\; \phi. \]
    We deduce that $B[f, \phi] = \int_{[-2L,2L]^d} (\lambda_n \sum_{|\alpha|\leqslant s} (-1)^{|\alpha|} \partial^{2\alpha} f + \mu_n \mathscr{D}^\ast \mathscr{D} f) \phi$. Since this identity holds for all $\phi \in H^s_{\mathrm{per}}([-2L,2L]^d)$, and since there is a unique Lax-Milgram inverse satisfying this condition, we conclude that
    \[\lambda_n \sum_{|\alpha|\leqslant s} (-1)^{|\alpha|} \partial^{2\alpha} f + \mu_n \mathscr{D}^\ast \mathscr{D} f = \mathscr{O}_n^{-1}(f). \]
\end{proof}

\begin{lem}[Lower bound on the integral operator norm]
    Assume that \[\lim_{n\to \infty}\lambda_n = \lim_{n\to \infty} \mu_n = \lim_{n\to \infty} \lambda_n / \mu_n = 0.\] Then there is a constant $C_5 > 0$ such that 
    \[\|L_{K}\|_{\mathrm{op}, L^2(\Omega, \mathbb{P}_X)} := \sup_{\|f\|_{L^2(\Omega, \mathbb{P}_X)}=1}{\|L_{K}f\|_{L^2(\Omega, \mathbb{P}_X)}} \geqslant C_5 \mu_n^{-1} \rightarrow \infty.\]
    \label{lem:lowerbound}
\end{lem}
\begin{proof}
    The operator $L_K$ is diagonalizable according to Theorem \ref{thm:eigenvalues}, and thus its operator norm $\sup_{\|f\|_{L^2(\Omega, \mathbb{P}_X)}=1}{\|L_{K}f\|_{L^2(\Omega, \mathbb{P}_X)}}$ is larger than the largest eigenvalue of $L_{K}$. The Courant-Fischer min-max theorem states that 
    this eigenvalue is larger than $\langle f, L_K f\rangle$ for any function $f$ such that $\|f\|_{L^2(\Omega, \mathbb{P}_X)}=1$.
    By the proof of Theorem \ref{thm:eigenvalues}, we know that
    \begin{equation*}
    \langle f, L_K f\rangle_{L^2(\Omega, \mathbb{P}_X)}= \Big\|\mathscr{O}_n^{1/2} \Big( f\frac{d\mathbb{P}_X}{dx}\Big)\Big\|_{L^2([-2L,2L]^d)}^2 = \Big\langle \mathscr{O}_n\Big(f\frac{d\mathbb{P}_X}{dx}\Big), f\frac{d\mathbb{P}_X}{dx} \Big\rangle_{L^2([-2L,2L]^d)}. 
\end{equation*}
Consider a smooth function $g$ with compact support in the set $E = \{z \in [-2L,2L]^d \; |\; \frac{d\mathbb{P}_X}{dx} \geqslant (4L)^{-d} /2\}$. 
Let \begin{equation}
    f = \Big(\lambda_n \sum_{|\alpha|\leqslant s} (-1)^{|\alpha|} \partial^{2\alpha} g + \mu_n \mathscr{D}^\ast \mathscr{D} g\Big) \times \Big(\frac{d\mathbb{P}_X}{dx}\Big)^{-1}.
    \label{eq:def_f}
\end{equation} Since $g$ is smooth and, on $E$, $(\frac{d\mathbb{P}_X}{dx})^{-1} \leqslant 2 (4L)^d$, we deduce that $f \in L^2(\Omega, \mathbb{P}_X)$. According to Lemma \ref{lem:explicit_calculus}, $\mathscr{O}_n(f\frac{d\mathbb{P}_X}{dx}) = g$.
Thus,
\begin{align*}
    \langle f, L_K f\rangle_{L^2(\Omega, \mathbb{P}_X)} &=\Big\langle g, \lambda_n \sum_{|\alpha|\leqslant s} (-1)^{|\alpha|} \partial^{2\alpha} g + \mu_n \mathscr{D}^\ast \mathscr{D} g \Big\rangle_{L^2([-2L,2L]^d)}\\
    &=\lambda_n \|g\|_{H^s([-2L,2L]^d)}^2 +\mu_n \|\mathscr{D} g\|_{L^2(\Omega)}^2.
\end{align*}
Recall that 
\begin{align*}
    \|L_{K}\|_{\mathrm{op}, L^2(\Omega, \mathbb{P}_X)} &\geqslant (\|f\|_{L^2(\Omega, \mathbb{P}_X)})^{-2} \langle f, L_K f\rangle_{L^2(\Omega, \mathbb{P}_X)}.
\end{align*}
On the one hand, if $\mathscr{D}^\ast \mathscr{D} g = 0$, then identity \eqref{eq:def_f} implies that $\|f\|_{L^2(\Omega, \mathbb{P}_X)}^2 = \Theta_{n\to \infty}(\lambda_n^2)$, and thus
\[(\|f\|_{L^2(\Omega, \mathbb{P}_X)})^{-2} \langle f, L_K f\rangle_{L^2(\Omega, \mathbb{P}_X)} = \Theta_{n\to \infty}(\lambda_n^{-1}).\]
On the other hand, if $\mathscr{D}^\ast \mathscr{D} g \neq 0$, since $\mu_n/\lambda_n \to \infty$, \eqref{eq:def_f} implies that $\|f\|_{L^2(\Omega, \mathbb{P}_X)}^2 = \Theta_{n\to \infty}(\mu_n^2)$, and thus 
\[(\|f\|_{L^2(\Omega, \mathbb{P}_X)})^{-2} \langle f, L_K f\rangle_{L^2(\Omega, \mathbb{P}_X)} = \Theta_{n\to \infty}(\mu_n^{-1}).\]
Overall, we conclude that there is a constant $C_5 > 0$, such that 
\[\|L_{K}\|_{\mathrm{op}, L^2(\Omega, \mathbb{P}_X)}  \geqslant C_5 \mu_n^{-1}.\]
\end{proof}

\subsection*{Bounds on the convergence rate}
\begin{thm}[High-probability bound]
\label{thm:speedup}
    Assume that the following four assumptions are satisfied:
    \begin{itemize}
        \item[$(i)$] $\lim_{n\to \infty}\lambda_n = \lim_{n\to \infty} \mu_n = \lim_{n\to \infty} \lambda_n / \mu_n = 0$,
        \item[$(ii)$] $\lambda_n \geqslant n^{-1}$,  
        \item[$(iii)$]  $\mathscr{N}(\lambda_n, \mu_n) \lambda_n^{-1} = o_n (n)$,
        \item[$(iv)$] for some $\sigma > 0$ and $M > 0$, the noise $\varepsilon$ satisfies 
    \[\forall \ell \in \mathbb{N}, \quad \mathbb{E}(|\varepsilon|^\ell\; | \; X) \leqslant \frac{1}{2}\ell !\; \sigma^2\; M^{\ell-2}.\]
    \end{itemize} 
    Then, letting $C_3 = 96 \log(6)$, for $n$ large enough, for all $\eta > 0$, with probability at least $1-\eta$, 
    \begin{align*}
        &\int_\Omega\|\hat f_n(x) - f^\star(x)\|_2^2 d\mathbb{P}_X(x) 
        \\
        & \quad \leqslant C_3 \log^2(\eta)\Big(\lambda_n \|f^\star\|_{H^s_{\mathrm{per}}([-2L, 2L]^d)}^2 + \mu_n \|\mathscr{D}(f^\star)\|_{L^2(\Omega)}^2 + \frac{M^2}{n^2 \lambda_n} + \frac{\sigma^2\mathscr{N}(\lambda_n, \mu_n)}{n}\Big).
    \end{align*}
\end{thm}
\begin{proof}
Observe that the kernel $K$ of Theorem \ref{thm:PDE_kernel}  depends on $n$ and that the function $f^\star$ belongs to a ball of radius $R_n = (\lambda_n \|f^\star\|_{H^s_{\mathrm{per}}([-2L, 2L]^d)}^2 + \mu_n \|\mathscr{D}(f^\star)\|_{L^2(\Omega)}^2)^{1/2}$.
Consider the non-asymptotic bound of \citet[][Theorem 4]{caponnetto2007optimal} applied to $K$  (that can be interpreted as a regular kernel for the norm $\|f\|_{\mathrm{RKHS}}^2 = \lambda_n \|f\|_{H^s([-2L, 2L]^d)}^2 + \mu_n \|\mathscr{D}(f)\|_{L^2(\Omega)}^2$, with an hyperparameter set to 1). 
    Thus,  we have, with probability at least $1-\eta$,
    \begin{equation}
        \mathcal{E}(\hat f_n) - \mathcal{E}(f^\star) \leqslant 32 \log^2(6\eta^{-1})\Big(\mathcal{A}(1) + \frac{\kappa_n^2 \mathcal{B}(1)}{n^2} + \frac{\kappa_n\mathcal{A}(1)}{n} + \frac{\kappa_n M^2}{n^2} + \frac{\sigma^2 \mathscr{N}(\lambda_n, \mu_n)}{n}\Big),
        \label{eq:caponettoBound}
    \end{equation}
    where
    \begin{itemize}
        \item[$(i)$] $\mathcal{E}(f) = \int_\Omega\|f(x) - y\|_2^2 d\mathbb{P}_{(X,Y)}(x,y)$,
        \item[$(ii)$] $\kappa_n = \sup_{x \in \Omega} K(x,x) \leqslant \lambda_n^{-1}$, according to Proposition \ref{prop:bound_kernel},
        \item[$(iii)$] and $\mathscr{A}(1) \leqslant R_n^2$ and $\mathscr{B}(1) \leqslant R_n^2$ (take $c=1$ and $\lambda=1$ in  \citealp[Proposition 3]{caponnetto2007optimal}). 
    \end{itemize}
    Inequality \eqref{eq:caponettoBound} is true as long as
    \begin{itemize}
        \item[$(i)$] $n \geqslant 64 \log^2(6/\eta) \kappa_n \mathscr{N}(\lambda_n, \mu_n)$, which holds for $n$ large enough since $\kappa_n \mathscr{N}(\lambda_n, \mu_n) = \mathcal{O}_n(\lambda_n^{-1}\mathscr{N}(\lambda_n, \mu_n)) = o_n(n)$ by assumption,
        \item[$(ii)$] $\|L_{K}\|_{\mathrm{op}, L^2(\Omega, \mathbb{P}_X)} \geqslant 1$, which holds for $n$ large enough by Lemma \ref{lem:lowerbound}, because, by assumption, $\lim_{n\to \infty}\lambda_n = \lim_{n\to \infty} \mu_n = \lim_{n\to \infty} \lambda_n / \mu_n = 0$.
    \end{itemize}
    Since $\lambda_n \geqslant n^{-1}$, we deduce that $n^{-1}\kappa_n \leqslant 1$, and so \[\mathcal{A}(1) + \frac{\kappa_n^2 \mathcal{B}(1)}{n^2} + \frac{\kappa_n\mathcal{A}(1)}{n} \leqslant 3 (\lambda_n \|f^\star\|_{H^s_{\mathrm{per}}([-2L, 2L]^d)}^2  + \mu_n \|\mathscr{D}(f^\star)\|_{L^2(\Omega)}^2).\]
    It follows that, letting $C_3 = 96 \log(6)$, for $n$ large enough, for all $\eta > 0$, with probability at least $1-\eta$, 
    \begin{align*}
        &\mathcal{E}(\hat f_n) - \mathcal{E}(f^\star) \\
        & \quad \leqslant C_3 \log^2(\eta)\Big(\lambda_n \|f^\star\|_{H^s_{\mathrm{per}}([-2L, 2L]^d)}^2 + \mu_n \|\mathscr{D}(f^\star)\|_{L^2(\Omega)}^2 + \frac{M^2}{n^2 \lambda_n} + \frac{\sigma^2\mathscr{N}(\lambda_n, \mu_n)}{n}\Big).
    \end{align*}
    The conclusion is then a consequence of the identity $\mathcal{E}(\hat f_n) - \mathcal{E}(f^\star) = \int_\Omega\|\hat f_n(x) - f^\star(x)\|_2^2 d\mathbb{P}_X(x)$.
\end{proof} 

\subsection*{Proof of Theorem \ref{thm:boundexp}}
    Note that, for all $f \in H^s_{\mathrm{per}}([-2L,2L]^d)$, $ \lambda_n \|f\|_{L^2(\Omega)}^2 \leqslant R_n(f)$. Since,  $\hat f_n$ is defined as minimizing~$R_n$, we have that
     \[
         \lambda_n \|\hat f_n\|_{L^2(\Omega)}^2 \leqslant  R_n(\hat f_n) \leqslant R_n( f^\star)= \lambda_n \|f^\star\|_{H^s(\Omega)}^2+ \frac{1}{n}\sum_{j=1}^n \|f^\star(X_i) - Y_i\|_2^2.
     \]
     By taking the expectation on these inequalities, we obtained that \[\mathbb{E}\|\hat f_n\|_{L^2(\Omega)}^2 \leqslant \|f^\star\|_{H^s(\Omega)}^2+ \lambda_n^{-1}\mathbb{E}\|\varepsilon\|_2^2.\] 
     We therefore have the following bound on the risk, where the expectation is taken with respect to the distribution of $(\hat f_n, X)$, where $X$ is a random variable independent from $\hat f_n$ with distribution~$\mathbb{P}_X$:
    \begin{align*}
        &\mathbb{E}\|\hat f_n(X) - f^\star(X)\|_2^2 
        \\
        &\leqslant  C_3 \log^2(\eta)\Big(\lambda_n \|f^\star\|_{H^s_{\mathrm{per}}([-2L, 2L]^d)}^2 + \mu_n \|\mathscr{D}(f^\star)\|_{L^2(\Omega)}^2 + \frac{M^2}{n^2 \lambda_n} + \frac{\sigma^2\mathscr{N}(\lambda_n, \mu_n)}{n}\Big)\\
        &\quad + 2\eta (2\|f^\star\|_{H^s(\Omega)}^2+ \lambda_n^{-1}\mathbb{E}\|\varepsilon\|_2^2).
    \end{align*}
    Take 
    $\eta = n^{-2}$, i.e.,  $\log(1/\eta) = 2\log(n)$. Thus, letting $C_4 = 4C_3 = 384 \log(6)$, for $n$ large enough, 
    \begin{align*}
        &\mathbb{E}\|\hat f_n(X) - f^\star(X)\|_2^2\\
        &\quad \leqslant C_4 \log^2(n)\Big(\lambda_n \|f^\star\|_{H^s_{\mathrm{per}}([-2L, 2L]^d)}^2 + \mu_n \|\mathscr{D}(f^\star)\|_{L^2(\Omega)}^2 + \frac{M^2}{n^2 \lambda_n} + \frac{\sigma^2\mathscr{N}(\lambda_n, \mu_n)}{n}\Big)\\
        &\quad \leqslant C_4 C_{s, \Omega}\log^2(n)\Big(\lambda_n \|f^\star\|_{H^s(\Omega)}^2 + \mu_n \|\mathscr{D}(f^\star)\|_{L^2(\Omega)}^2 + \frac{M^2}{n^2 \lambda_n} + \frac{\sigma^2\mathscr{N}(\lambda_n, \mu_n)}{n}\Big),
    \end{align*}
    where $C_{s, \Omega}$ is the constant in the Sobolev extension.

\subsection*{Proof of Proposition \ref{prop:minSpeed}}

    According to \citet[Proposition 3]{caponnetto2007optimal}, if
$a_m = \mathcal{O}_m( m^{1/b})$, then  
\begin{equation}
    \label{eq:bornepoly}
    \sum_{m \in \mathbb{N}} \frac{1}{ 1+\lambda_n a_m} = \mathcal{O}_n(\lambda_n^{-b}).
\end{equation}
In particular, Proposition \ref{prop:eigenvalue} implies that
\[\mathscr{N}(\lambda_n, \mu_n)  = \mathcal{O}_n(\lambda_n^{-d/2s}) .\]
Combining this bound with Theorem \ref{thm:speedup} shows that the PDE kernel approaches $f^\star$ at least at the minimax rate on $H^s(\Omega)$, i.e., $n^{-2s/(2s+d)}$ (up to a log-term).

\section{About the choice of regularization}

\subsection*{Kernel equivalence}
\begin{lem}[Minimal Sobolev norm extension]
    Let $s \in \mathbb{N}$. There is an extension $E: H^s(\Omega) \to H^s_{\mathrm{per}}([-2L,2L]^d)$ such that 
    \[E(f) = \mathrm{argmin}_{g\in H^s_{\mathrm{per}}([-2L,2L]^d), \; g|_\Omega = f} \; \|g\|_{H^s_{\mathrm{per}}([-2L,2L]^d)}.\]
    Moreover, $E$ is linear and bounded, which means that $\|f\|_{H^s(\Omega)}$ and $\| E(f)\|_{H^s_{\mathrm{per}}([-2L,2L]^d)}$ are equivalent norms on $H^s(\Omega)$.
    \label{lem:minimal_norm}
\end{lem}
\begin{proof}
We have already constructed an extension $\tilde E: H^s(\Omega) \to H^s_{\mathrm{per}}([-2L,2L]^d)$ in Proposition~\ref{prop:dec_four_lip}. However, $\tilde E$ does not minimize the Sobolev norm on $\Omega^c$. 
Let $f \in H^s(\Omega)$ and $\mathscr{H}_0 = \{g\in H^s_{\mathrm{per}}([-2L,2L]^d), \; g|_\Omega = 0\}$. Clearly, $(\mathscr{H}_0, \|\cdot\|_{H^s_{\mathrm{per}}([-2L,2L]^d)})$ is a Banach space. One has 
\begin{align*}
    &\min_{g\in H^s_{\mathrm{per}}([-2L,2L]^d), \; g|_\Omega = f} \; \|g\|_{H^s_{\mathrm{per}}([-2L,2L]^d)} \\
    & \quad = \min_{g\in \mathscr{H}_0 } \; \|\tilde E(f) + g\|_{H^s_{\mathrm{per}}([-2L,2L]^d)}\\
    & \quad = \min_{g\in \mathscr{H}_0 } \; \|\tilde E(f) + g\|_{H^s_{\mathrm{per}}([-2L,2L]^d)}^2\\
    &\quad = \min_{g\in \mathscr{H}_0} \; \| g\|_{H^s_{\mathrm{per}}([-2L,2L]^d)}^2  + 2\langle \tilde E(f), g\rangle_{H^s_{\mathrm{per}}([-2L,2L]^d)}.
\end{align*}
The form $\langle \cdot, \cdot\rangle_{H^s_{\mathrm{per}}([-2L,2L]^d)}$ is bilinear, symmetric, continuous, and coercive on $\mathscr{H}_0 \times \mathscr{H}_0$.
Thus, according to the Lax-Milgram theorem \citep[e.g., Corollary 5.8]{brezis2010functional}, there exists a unique element $u(f)$ of $\mathscr{H}_0$ such that, for all $g \in \mathscr{H}_0$,
\begin{equation}
    \langle u(f), g\rangle_{H^s_{\mathrm{per}}([-2L,2L]^d)} = -\langle \tilde{E}(f), g\rangle_{H^s_{\mathrm{per}}([-2L,2L]^d)}.
    \label{eq:minSobExtension}
\end{equation}
Thus, $\langle u(f) + \tilde{E}(f), g\rangle_{H^s_{\mathrm{per}}([-2L,2L]^d)} = 0$.
Moreover, $u(f)$ is the unique minimum of $g \mapsto  \| g\|_{H^s_{\mathrm{per}}([-2L,2L]^d)}^2  + 2\langle \tilde E(f), g\rangle_{H^s_{\mathrm{per}}([-2L,2L]^d)}$. Therefore, $E(f) :=  \tilde{E}(f) + u(f)$ satisfies
 \[E(f)= \mathrm{argmin}_{g\in H^s_{\mathrm{per}}([-2L,2L]^d), \; g|_\Omega = f} \; \|g\|_{H^s_{\mathrm{per}}([-2L,2L]^d)}.\]

Let us now show that the extension $E$ is linear. Let $f_1 \in H^s(\Omega)$, $f_2 \in H^s(\Omega)$, and $\lambda \in \mathbb{R}$. We have shown that, for $g \in \mathscr{H}_0$,
\[\langle u(f_1) + \tilde{E}(f_1), g\rangle_{H^s_{\mathrm{per}}([-2L,2L]^d)} = 0,\]
\[\langle u(f_2) + \tilde{E}(f_2), g\rangle_{H^s_{\mathrm{per}}([-2L,2L]^d)} = 0,\]
\[\mbox{and }\langle u(f_1 + \lambda f_2) + \tilde{E}(f_1 + \lambda f_2), g\rangle_{H^s_{\mathrm{per}}([-2L,2L]^d)} = 0.\]
By subtracting the third identity to the first two ones, and observing that, since $\tilde E$ is linear, $\tilde{E}(f_1 + \lambda f_2) = \tilde{E}(f_1) + \tilde{E}(\lambda f_2)$, we deduce that
\[\langle u(f_1) + \lambda u(f_2) - u(f_1 + \lambda f_2) , g\rangle_{H^s_{\mathrm{per}}([-2L,2L]^d)} = 0.\]
As $u(f) \in \mathscr{H}_0$ for all $f\in H^s(\Omega)$, we deduce that $ u(f_1) + u(f_2) - u(f_1 + f_2) \in \mathscr{H}_0$. Therefore, taking $g = u(f_1) + u(f_2) - u(f_1 + f_2)$, we have 
\[\| u(f_1) + \lambda u(f_2) - u(f_1 + \lambda f_2) \|_{H^s_{\mathrm{per}}([-2L,2L]^d)} = 0,\]
i.e., $ u(f_1 + \lambda f_2) = u(f_1) + \lambda u(f_2) $. Thus, $E$ is linear.

Proposition \ref{prop:dec_four_lip} shows that $\|\tilde E(f)\|_{H^s([-2L,2L]^d)}^2 \leqslant \tilde C_{s, \Omega} \|f\|_{H^s(\Omega)}^2$. Moreover, by definition of~$E$, $\| E(f)\|_{H^s([-2L,2L]^d)}^2 \leqslant \|\tilde E(f)\|_{H^s([-2L,2L]^d)}^2$. Thus, $\| E(f)\|_{H^s([-2L,2L]^d)}^2 \leqslant \tilde C_{s, \Omega} \|f\|_{H^s(\Omega)}^2$, i.e., the extension $E$ is bounded. Clearly, $\| E(f)\|_{H^s([-2L,2L]^d)}^2 \geqslant \|f\|_{H^s(\Omega)}^2$. We conclude that $\|f\|_{H^s(\Omega)}$ and $\| E(f)\|_{H^s([-2L,2L]^d)}$ are equivalent norms.
\end{proof}

\begin{prop}[Kernel equivalence]
    Assume that $s>d/2$. Let $\lambda_n > 0$ and $\mu_n \geqslant 0$.  Let $\langle \cdot,  \cdot\rangle_n$ be inner products associated with kernels on $H^s(\Omega)$. Assume that there exist constants $C_1 > 0$ and $C_2 >0$ such that, for all $n \in \mathbb{N}$ and all $f \in H^s(\Omega)$,
    \[ C_1 (\lambda_n \| f\|_{H^s(\Omega)}^2 +\mu_n \| \mathscr{D}(f)\|_{L^2(\Omega)}^2 )\leqslant \langle f, f\rangle_n \leqslant C_2 (\lambda_n \| f\|_{H^s(\Omega)}^2 + \mu_n \| \mathscr{D}(f)\|_{L^2(\Omega)}^2).\]
    Then the kernels associated with $\langle \cdot, \cdot\rangle_n$ on $H^s(\Omega)$ have the same convergence rate as the kernel of Theorem \ref{thm:PDE_kernel} associated with the $\lambda_n \| f\|_{H^s_{\mathrm{per}}([-2L,2L]^d)}^2 + \mu_n \| \mathscr{D}(f)\|_{L^2(\Omega)}^2$ norm.
    \label{prop:kernel_eq}
\end{prop}
\begin{proof}
For clarity, the proof is divided into four steps. 

\paragraph{Step1: From $H^s(\Omega)$ to $H^s_{\mathrm{per}}([-2L,2L]^d)$.}
    Observe that 
\begin{align*}
    \hat f_n &= \mathrm{argmin}_{f \in H^s_{\mathrm{per}}([-2L,2L]^d)} \sum_{i=1}^n |f(X_i) -Y_i|^2  + \lambda_n  \| f\|_{H^s_{\mathrm{per}}([-2L,2L]^d)}^2 + \mu_n \| \mathscr{D}(f)\|_{L^2(\Omega)}^2\\
    &= E\Big(\mathrm{argmin}_{f \in H^s(\Omega)} \sum_{i=1}^n |f(X_i) -Y_i|^2  + \lambda_n  \| E(f)\|_{H^s_{\mathrm{per}}([-2L,2L]^d)}^2 + \mu_n \| \mathscr{D}(f)\|_{L^2(\Omega)}^2\Big),
\end{align*}
where $E(f)$ is the extension $H^s(\Omega) \to H^s_{\mathrm{per}}([-2L,2L]^d)$ with minimal $H^s_{\mathrm{per}}([-2L,2L]^d)$ norm (see Lemma \ref{lem:minimal_norm}).
Define 
\begin{align*}
    \hat f_n^{(5)} &= \mathrm{argmin}_{f \in H^s(\Omega)} \sum_{i=1}^n |f(X_i) -Y_i|^2  + \lambda_n  \| E(f)\|_{H^s_{\mathrm{per}}([-2L,2L]^d)}^2 + \mu_n \| \mathscr{D}(f)\|_{L^2(\Omega)}^2.
\end{align*}
Then $\hat f_n = E(\hat f_n^{(5)})$, which means that for all $x \in \Omega$, $\hat f_n(x) = \hat f_n^{(5)}(x)$. Thus, $\hat f_n$ and $\hat f_n^{(5)}$ have the same convergence rate to $u^\star$. 

\paragraph{Step 2: inner products equivalence.}
Lemma \ref{lem:minimal_norm} states that  $\|f\|_{H^s(\Omega)}$ and $\| E(f)\|_{H^s_{\mathrm{per}}([-2L,2L]^d)}$ are equivalent norms on $H^s(\Omega)$. Therefore, there are constants $C_3$ and $C_4$ such that
\begin{align*}
&C_3 (\lambda_n \| E(f)\|_{H^s_{\mathrm{per}}([-2L,2L]^d)}^2 + \mu_n \| \mathscr{D}(f)\|_{L^2(\Omega)}^2) \\
&\quad \leqslant \|f\|_n^2 \leqslant C_4 ( \lambda_n E(f)\|_{H^s_{\mathrm{per}}([-2L,2L]^d)}^2 + \mu_n \| \mathscr{D}(f)\|_{L^2(\Omega)}^2).
\end{align*}
This shows that the function $\langle \cdot, \cdot\rangle_n: H^s(\Omega) \times H^s(\Omega) \to \mathbb{R}$ is coercive with respect to the $(\lambda_n \| E(f)\|_{H^s_{\mathrm{per}}([-2L,2L]^d)}^2+ \mu_n \| \mathscr{D}(f)\|_{L^2(\Omega)}^2)$ norm. By the Cauchy-Schwarz inequality, $\langle \cdot, \cdot\rangle_n$ is continuous with respect to the same norm. Set $\langle f, g \rangle_n^{\mathrm{per}} =  \lambda_n\sum_{|\alpha|\leqslant s} \int_{[-2L,2L]^d} \partial^\alpha E(f) \partial^\alpha E(g) + \mu_n \int_\Omega \mathscr{D}(f)\;\mathscr{D}(g)$. 
Thus, by the Lax-Milgram theorem, there exists a linear operator $\mathscr{O}: H^s(\Omega) \to H^s(\Omega)$ such that, for all $f$, $g\in H^s(\Omega)$, 
\begin{equation}
    \langle \mathscr{O} f,g\rangle_n = \langle f, g\rangle_n^{\mathrm{per}}.
    \label{eq:transition_op}
\end{equation}
Since \[C_3 (\|\mathscr{O} f\|_n^{\mathrm{per}})^2 \leqslant \|\mathscr{O} f\|_n^2 = \langle \mathscr{O} f, f\rangle_n^{\mathrm{per}} \leqslant \|\mathscr{O} f\|_n^{\mathrm{per}} \|f\|_n^{\mathrm{per}},\]
we deduce that $\|\mathscr{O} f\|_n^{\mathrm{per}} \leqslant C_3^{-1} \| f\|_n^{\mathrm{per}}$. Similarly, the coercivity and continuity of $\langle \cdot, \cdot \rangle_n^{\mathrm{per}}$ with respect to $\langle \cdot, \cdot \rangle_n$ shows that $\|\mathscr{O}^{-1} f\|_n \leqslant C_4 \| f\|_n$, so that $\|\mathscr{O}^{-1} f\|_n^{\mathrm{per}} \leqslant C_3^{-1} C_4^2 \| f\|_n^{\mathrm{per}}$. All in all, \[ C_3 C_4^{-2} \| f\|_n^{\mathrm{per}} \leqslant \|\mathscr{O} f\|_n^{\mathrm{per}} \leqslant C_3^{-1} \| f\|_n^{\mathrm{per}}.\]
One easily verifies that $\mathscr{O}$ is self-adjoint.

\paragraph{Step 3: Link between kernels.} 
Let $f \in H^s(\Omega)$. Remember that, for all $x \in \Omega$, $K(x, \cdot) = \mathscr{O}_n(\delta_x)$ satisfies a weak formulation consistent with the weak formulation of the minimal-Sobolev norm extension in \eqref{eq:minSobExtension}. Thus, $E(K(x, \cdot)) = K(x, \cdot)$, and according to Theorem \ref{thm:PDE_kernel}, we have $f(x) = \langle f, K(x,\cdot) \rangle_n^{\mathrm{per}}$. In this proof, to distinguish between kernels, we denote the associated kernel by $K_n^{\mathrm{per}}(x,y) := K(x,y)$. 
Using the spectral theorem for bounded operators, we have that~$\mathscr{O}^{-1}$ admits a square root $\mathscr{O}^{-1/2}$ which is self-adjoint for the $\langle \cdot, \cdot \rangle_n^{\mathrm{per}}$ inner product.
Therefore, using \eqref{eq:transition_op}, we know that, for all $x \in \Omega$, $f(x) = \langle {\mathscr{O}}^{-1/2}(f), {\mathscr{O}}^{1/2}K(x, \cdot) \rangle_n^{\mathrm{per}}$. Since $\|{\mathscr{O}}^{-1/2}(f)\|_n^{\mathrm{per}} = \|f\|_n$, we deduce that $H^s(\Omega)$ is also a kernel space for the $\|\cdot\|_n$ norm, with kernel $K_n(x,y) = \langle {\mathscr{O}}(K(x, \cdot)), K(y, \cdot)\rangle_n^{\mathrm{per}}$.

\paragraph{Step 4: Eigenvalues of the integral operator.} Define the integral operators $L_n$ and $L^{\mathrm{per}}_n$ on $L^2(\Omega, \mathbb{P}_X)$ by 
\[L_n^{\mathrm{per}}(f) : x \mapsto \int_\Omega K_n^{\mathrm{per}}(x,y) f(y) d\mathbb{P}_X(y) \quad \mbox{and} \quad L_n(f) : x \mapsto \int_\Omega K_n(x,y) f(y) d\mathbb{P}_X(y).\]
Recalling that $K_n^{\mathrm{per}}(x, y) = \sum_{m\in \mathbb{N}} a_m v_m(x) v_m(y)$, we can use the same technique as in the proof of Theorem \ref{thm:eigenvalues} to apply the Fubini-Lebesgue theorem, and show that 
\[\langle f, L_n(f)\rangle_{L^2(\Omega, \mathbb{P}_X)} = (\|\mathscr{O}^{1/2} \mathscr{O}_n(f)\|_n^{\mathrm{per}})^2.\]
Thus, $C_3 C_4^{-2} \langle f, L_n^{\mathrm{per}}(f)\rangle \leqslant \langle f, L_n(f)\rangle \leqslant C_3^{-1}  \langle f, L_n^{\mathrm{per}}(f)\rangle$. The Courant-Fischer min-max theorem guarantees that the eigenvalues of $L_n^{\mathrm{per}}$ are upper and lower bounded by those of $L_n$. In particular, the effective dimensions $\mathscr{N}(\lambda_n, \mu_n)$ related to $\|\cdot\|_n^{\mathrm{per}}$ and $\mathscr{N}^{\mathrm{per}}(\lambda_n, \mu_n)$ satisfy
\[ C_3 C_4^{-2}\mathscr{N}^{\mathrm{per}}(\lambda_n, \mu_n) \leqslant \mathscr{N}(\lambda_n, \mu_n) \leqslant C_3^{-1}\mathscr{N}^{\mathrm{per}}(\lambda_n, \mu_n).\]
This implies that both kernels have equivalent effective dimensions.
\end{proof}

\subsection*{Proof of Theorem \ref{thm:eq_reg}}

Proposition \ref{prop:kernel_eq} ensures that $\hat f_n^{(1)}$ and $\hat f_n^{(2)}$ converge at the same rate. 
If $\|\cdot\|$ and $\|\cdot\|_{H^s(\Omega)}$ are equivalent, then there are constants $0 <C_1 < 1$, $C_2 > 1$ such that, for all $f \in H^s(\Omega)$, $C_1 \|f\|_{H^s(\Omega)}^2 \leqslant \|f\|_2^2 \leqslant C_2 \|f\|_{H^s(\Omega)}^2$. Thus, $C_1 (\mu_n \|\mathscr{D}(f)\|_{L^2(\Omega)}^2 + \lambda_n \|f\|_{H^s(\Omega)}^2) \leqslant \mu_n \|\mathscr{D}(f)\|_{L^2(\Omega)}^2 + \lambda_n \|f\| \leqslant C_2 (\mu_n \|\mathscr{D}(f)\|_{L^2(\Omega)}^2 + \lambda_n \|f\|_{H^s(\Omega)}^2).$ Proposition \ref{prop:kernel_eq} then shows that $\hat f_n^{(2)}$ and $\hat f_n^{(3)}$ converge at the same rate.

\section{Application: the case $\mathscr{D} = \frac{d}{dx}$}
\subsection*{Boundary conditions}
\begin{prop}
    Let $s = 1$, $\Omega = [-L,L]$, and $\mathscr{D} = \frac{d}{dx}$. Then any weak solution $w_m$ of the weak formulation \eqref{eq:weak_pde} satisfies  
    \begin{align*}
             (\lambda_n+\mu_n)\lim_{x \to -L, x > -L}\frac{d}{dx}w_m(x) &= \lambda_n \lim_{x \to -L, x < -L}\frac{d}{dx}w_m(x),\\
            (\lambda_n+\mu_n)\lim_{x \to L, x < L}\frac{d}{dx}w_m(x) &= \lambda_n \lim_{x \to L, x >L}\frac{d}{dx}w_m(x).
    \end{align*}
    \label{prop:raccord1d}
\end{prop}
\begin{proof}
    The proof uses the framework of distribution theory. By the inclusion $C^\infty([-2L,2L]) \subseteq H^s_{\mathrm{per}}([-2L,2L])$, we know that, considering any test function $\phi \in C^\infty([-2L,2L])$ with compact support in $]-2L,2L[$, one has $B[w_m, \phi] = a_m^{-1} \langle w_m \mathbf{1}_\Omega, \phi\rangle$. 
     Moreover, standard results of functional analysis (using the mollification of $y \mapsto \mathbf{1}_{|y-x| < 3\varepsilon/2}$ with a parameter $\eta = \varepsilon /8$ as in \citealt[Appendix C, Theorem 6]{evans2010partial}) ensures that, for any $x \in [-2L,2L]$, there exists a sequence of functions $(\xi^{x}_\varepsilon)_{\varepsilon > 0}$ such that, for all $m$,
    \begin{itemize}
        \item[$(i)$] $\xi^{x}_\varepsilon \in C^\infty([-2L,2L])$ with compact support in $D$, 
        \item[$(ii)$] $\|\xi^{x}_\varepsilon\|_\infty = 1$,
        \item[$(iii)$] and for all $y \in [-2L,2L]$,
        \begin{align*}
            |y-x| \geqslant 2\varepsilon &\Rightarrow \xi^{x}_\varepsilon(y) = 0\\
            |y-x| \leqslant \varepsilon\;\; &\Rightarrow \xi^{x}_\varepsilon(y) = 1.
        \end{align*}
    \end{itemize}
    Fix two of such sequences with $x = -L$ and $x=L$, and let $\phi_\varepsilon = \phi \times (\xi^{-L}_\varepsilon + \xi^{L}_\varepsilon)$. Notice that the following is true: 
    \begin{itemize}
        \item[$(i)$] $\phi_\varepsilon \in C^\infty([-2L,2L])$ has compact support and $\mathrm{supp}(\phi_\varepsilon) \subseteq \mathrm{supp}(\phi)$,
        \item[$(ii)$]  for all $r \geqslant 0$, $\frac{d^r}{dx^r}\phi_\varepsilon(-L) = \frac{d^r}{dx^r}\phi(-L)$, 
        \item[$(iii)$] for any function $f \in L^2([-2L,2L])$, $\lim_{\varepsilon \to 0} \langle f, \phi_\varepsilon\rangle = 0$,
        \item[$(iv)$] and $B[w_m, \phi_\varepsilon] = a_m^{-1} \langle w_m \mathbf{1}_\Omega, \phi_\varepsilon\rangle$.
    \end{itemize}
    Choose $\mathrm{supp}(\phi) \subseteq [-3L/2, -L/2]$. Clearly, $\int_{-2L}^{2L} (\frac{d}{dx}w_m) (\frac{d}{dx} \phi_\varepsilon) = \int_{-3L/2}^{-L} (\frac{d}{dx}w_m) (\frac{d}{dx} \phi_\varepsilon) + \int_{-L}^{-L/2} (\frac{d}{dx}w_m) (\frac{d}{dx} \phi_\varepsilon)$. The integration by parts formula implies 
    \begin{align*}
        \int_{-3L/2}^{-L} \Big(\frac{d}{dx}w_m\Big) \Big(\frac{d}{dx} \phi_\varepsilon\Big) &=  -\int_{-3L/2}^{-L} \Big(\frac{d^2}{dx^2}w_m\Big) \phi_\varepsilon + \lim_{x \to -L, x < -L} \phi_\varepsilon(x) \frac{d}{dx}w_m(x) \\
        &\xrightarrow{ \varepsilon \to 0 } \lim_{x \to -L, x < -L} \phi_\varepsilon(x) \frac{d}{dx}w_m(x).
    \end{align*}
    Similarly, 
    \begin{align*}
        \int_{-L}^{-L/2} \Big(\frac{d}{dx}w_m\Big)\Big (\frac{d}{dx} \phi_\varepsilon\Big) &=  -\int_{-L}^{-L/2} \Big(\frac{d^2}{dx^2}w_m\Big) \phi_\varepsilon - \lim_{x \to -L, x > -L} \phi_\varepsilon(x) \frac{d}{dx}w_m(x) \\
        &\xrightarrow{ \varepsilon \to 0 } -\lim_{x \to -L, x > -L} \phi_\varepsilon(x) \frac{d}{dx}w_m(x).
    \end{align*}
    Note that  $\lim_{x \to -L, x < -L} \phi_\varepsilon(x) = \lim_{x \to -L, x > -L} \phi_\varepsilon(x)  = \phi(-L)$. Therefore,
    \begin{equation*}
        \lim_{\varepsilon \to 0}\int_{-2L}^{2L} \Big(\frac{d}{dx}w_m\Big) \Big(\frac{d}{dx} \phi_\varepsilon\Big) 
        = \phi(-L)\Big(\lim_{x \to -L, x < -L}\frac{d}{dx}w_m(x) -\lim_{x \to -L, x > -L}\frac{d}{dx}w_m(x)\Big).
    \end{equation*}
    This means that the integral $\int_{-2L}^{2L} (\frac{d}{dx}w_m) (\frac{d}{dx} \phi_\varepsilon)$ quantifies the discontinuity in the derivative of $w_m$ at $-L$.
    Thus, since 
    \[B[w_m, \phi_\varepsilon] = a_m^{-1} \int_{-L}^L w_m\phi_\varepsilon,\]
    we obtain, letting $\varepsilon \to 0$ that
    \[(\lambda_n+\mu_n)\lim_{x \to -L, x > -L}\frac{d}{dx}w_m(x) = \lambda_n \lim_{x \to -L, x < -L}\frac{d}{dx}w_m(x).\]
    The same analysis holds in a neighborhood of $L$, and leads to
    \[(\lambda_n+\mu_n)\lim_{x \to L, x < L}\frac{d}{dx}w_m(x) = \lambda_n \lim_{x \to L, x >L}\frac{d}{dx}w_m(x).\]
\end{proof}

\subsection*{Proof of Proposition \ref{prop:1dkernel}}
Combining Theorem \ref{thm:eq_reg} and  Proposition \ref{prop:kernel_eq}, we know that
\[
        \hat f_n^{(1)} = \mathrm{argmin}_{f \in H^1([-L,L])} \sum_{i=1}^n |f(X_i) -Y_i|^2 + \lambda_n \| f\|_{H^1([-L,L])}^2 + \mu_n \| \mathscr{D}(f)\|_{L^2([-L,L])}^2 
\]
and
\[
 \hat f_n^{(2)} = \mathrm{argmin}_{f \in H^1_{\mathrm{per}}([-2L,2L])} \sum_{i=1}^n |f(X_i) -Y_i|^2 + \lambda_n  \| f\|_{H^1_{\mathrm{per}}([-2L,2L])}^2 + \mu_n \| \mathscr{D}(f)\|_{L^2([-L,L])}^2,
\]
converge at the same rate to $f^\star$. Moreover, the $\lambda_n \| f\|_{H^1([-L,L])}^2 + \mu_n \| \mathscr{D}(f)\|_{L^2([-L,L])}^2 $ norm on $H^1([-L,L])$ defines a kernel. This is this particular kernel, denoted by $K$, that we compute in the remaining of the proof. Employing the exact same arguments as for the kernel on $H^1_{\mathrm{per}}([-2L,2L])$, we know that for all $x \in [-L,L]$, the function $f_x: y \mapsto K(x,y) \in H^1([-L,L])$ is a solution to the weak PDE
\[\forall \phi \in H^1([-L,L]), \quad \lambda_n \int_{[-L,L]} f_x \phi + (\lambda_n + \mu_n)\int_{[-L,L]} \frac{d}{dy}f_x \frac{d}{dy}\phi = \phi(x).\]
Using the elliptic regularity theorem as in the proof of Theorem \ref{prop:eigenfunction} and computing the boundary conditions as in Proposition \ref{prop:raccord1d} shows that
$f_x \in C^\infty([-L, x]) \cap C^\infty([x, L])$, $\frac{d}{dy}f_x(-L) = \frac{d}{dy}f_x(L) = 0$, and 
\[\lambda_n  f_x  - (\lambda_n + \mu_n)\frac{d^2}{dy^2}f_x = \delta_x,\]
where $\delta_x$ is the Dirac distribution.
Thus, since $f_x \in H^1([-L,L]) \subseteq C^0([-L,L])$, there are constants $A$ and $B$ such that
\begin{equation}
        \left\{ \begin{array}{cc}
             \forall -L \leqslant y \leqslant x,    &f_x(y) = A \cosh(\gamma_n (x-y)) + B \sinh(\gamma_n (x-y)) ,\\
             \forall x \leqslant y \leqslant L,    &f_x(y) = A \cosh(\gamma_n (x-y)) + (B + \frac{\gamma_n}{\lambda_n}) \sinh(\gamma_n (x-y)).
        \end{array}\right.
        \label{eq:kernelForm}
        \end{equation}
The boundary conditions   $\frac{d}{dy}f_x(-L) = \frac{d}{dy}f_x(L) = 0$ lead to
\[P \begin{pmatrix}
    A \\ B
\end{pmatrix} =   \begin{pmatrix}
    0 \\ - \frac{\gamma_n}{\lambda_n} \cosh(\gamma_n(x-L))
\end{pmatrix}, \]
where \[P = \begin{pmatrix}
    \sinh(\gamma_n(x+L)) & \cosh(\gamma_n(x+L))\\
    \sinh(\gamma_n(x-L)) & \cosh(\gamma_n(x-L))
\end{pmatrix}.\] Notice that $\det P = \sinh(\gamma_n(x+L)) \cosh(\gamma_n(x-L)) - \sinh(\gamma_n(x-L))\cosh(\gamma_n(x+L)) =  \sinh(2 \gamma_n L)$. Thus, 
\[P^{-1} = \sinh(2 \gamma_n L)^{-1} \begin{pmatrix}
    \cosh(\gamma_n(x-L)) & -\cosh(\gamma_n(x+L))\\
    -\sinh(\gamma_n(x-L)) & \sinh(\gamma_n(x+L))
\end{pmatrix}.\]
This leads to 
\begin{align}
    \begin{pmatrix}
    A \\ B
\end{pmatrix} &= \frac{\gamma_n}{\lambda_n\sinh(2 \gamma_n L)} \begin{pmatrix}
    \cosh(\gamma_n(x+L)) \cosh(\gamma_n(x-L)) \\ -\sinh(\gamma_n(x+L))\cosh(\gamma_n(x-L))
\end{pmatrix} \nonumber\\
&= \frac{\gamma_n}{2\lambda_n\sinh(2 \gamma_n L)} \begin{pmatrix}
    \cosh(2\gamma_nL) + \cosh(2\gamma_nx) \\ \sinh(2\gamma_nL))-\sinh(2\gamma_nx)
\end{pmatrix}. \label{eq:invMatrix}
\end{align}  
Combining \eqref{eq:kernelForm} and \eqref{eq:invMatrix}, we are led to
\begin{align*}
        K(x,y)& = \frac{\gamma_n}{2\lambda_n \sinh(2 \gamma_n L)}\Big( (\cosh(2\gamma_n L)+\cosh(2\gamma_n x))\cosh(\gamma_n (x-y))\\
        &\qquad  + ((1-2 \times \mathbf{1}_{x > y})\sinh(2\gamma_n L) - \sinh(2 \gamma_n x)) \sinh(\gamma_n(x-y))\Big).
    \end{align*}
One easily checks that $K(x,y) = K(y,x)$ and that $K(x,x) \geqslant 0$.

\subsection*{Proof of Proposition \ref{prop:1ddiff}}
 The strategy of the proof is to characterize the solutions $w_m$ to the weak formulation \eqref{eq:weak_pde} with $\mathscr{D} = \frac{d}{dt}$ and $s = 1 > d/2 = 1/2$. For clarity, the proof is divided into 5 steps.

\paragraph{Step 1: Symmetry.} Recall that $\Omega = [-L, L]$. Using the Lax-Milgram theorem, let us define the operator $\tilde{\mathscr{O}}_n$ as follows. For all $f \in L^2([-2L,2L])$, $\tilde{\mathscr{O}}_n(f)$ is  
the unique function of $H^2_{\mathrm{per}}([-2L,2L])$ such that, for all $\phi \in H^2_{\mathrm{per}}([-2L,2L])$, $B[\tilde{\mathscr{O}}_n(f), \phi] = \langle C f, C \phi \rangle$. Clearly, the eigenfunctions of $\tilde{\mathscr{O}}_n$ associated to non-zero eigenvalues are the $w_m$. Let $\phi \in H^2_{\mathrm{per}}([-2L,2L])$ be a test function. Using \begin{align*}
        \int_{-2L}^{2L} \partial^\alpha \phi(-\cdot)(x) \partial^\alpha \tilde{\mathscr{O}}_n(f)(-\cdot)(x)dx &= (-1)^{2\alpha}\int_{-2L}^{2L} \partial^\alpha \phi(-x) \partial^\alpha \tilde{\mathscr{O}}_n(f)(-x)dx\\
        & = -\int_{-2L}^{2L} \partial^\alpha \phi(x) \partial^\alpha \tilde{\mathscr{O}}_n(f)(x)dx,
    \end{align*} 
    we see that $B[\tilde{\mathscr{O}}_n(f)(-\cdot), \phi(-\cdot)] = \langle Cf(-\cdot), C \phi(-\cdot)\rangle $. Therefore, since $H^2_{\mathrm{per}}([-2L,2L])$ is stable by the action $\phi \mapsto \phi(-\cdot)$, using the uniqueness statement provided by the Lax-Milgram theorem, we deduce that $\tilde{\mathscr{O}}_n(f)(-x) = \tilde{\mathscr{O}}_n(f(-\cdot))(x)$, so that $\tilde{\mathscr{O}}_n(f)$ is symmetric. According to Proposition \ref{prop:sym}, we can therefore assume that $w_m$ is either symmetric or antisymmetric. 

\paragraph{Step 2: PDE system.}
According to Theorem \ref{prop:eigenfunction}, \ref{prop:1dreg}, and \ref{prop:raccord1d}, the following statements are verified: 
 \begin{itemize}
        \item[$(i)$] The function $w_m \in C^\infty([-L,L])$ and 
        \[\forall x \in \Omega, \quad \lambda_n\Big(1- \frac{d^2}{dx^2}\Big)w_m(x)  - \mu_n \frac{d^2}{dx^2} w_m(x) = a_m^{-1} w_m(x).\]
        Since $a_m^{-1} \geqslant \lambda_n$ (see Remark \ref{rem:rayleigh}), the solutions of this ODE are  linear combinations of $\cos(\sqrt{\frac{a_m^{-1}-\lambda_n}{\lambda_n + \mu_n}}x)$ and $\sin(\sqrt{\frac{a_m^{-1}-\lambda_n}{\lambda_n + \mu_n}}x)$.
        \item[$(ii)$] The function $w_m \in C^\infty([-2L, 2L] \backslash [-L,L])$, with a $C^\infty$ junction condition at $-2L$, and 
        \[\forall x \in [-2L, 2L]^d \backslash \bar \Omega,\quad \Big(1- \frac{d^2}{dx^2}\Big)w_m(x)  = 0.\]
        The solutions of this ODE are linear combinations of $\cosh(x)$ and $\sinh(x)$. The $C^\infty$ $4L$-periodic junction condition at $-2L$ guarantees that there are two constants $A$ and $B$ such that 
\begin{align*}
             \forall -2L \leqslant x \leqslant -L,    &w_m(x) = A \cosh(x+2L) + B \sinh(x+2L),\\
              \forall L \leqslant x \leqslant 2L,   &w_m(x) = A \cosh(x-2L) + B \sinh(x-2L).
        \end{align*}
        \item[$(iii)$] The function $w_m \in C^0_{\mathrm{per}}([-2L,2L])$.
        \item[$(iv)$] One has
        \begin{align*}
        (\lambda_n+\mu_n)\lim_{x \to -L, x > -L}\frac{d}{dx}w_m(x) &= \lambda_n \lim_{x \to -L, x < -L}\frac{d}{dx}w_m(x),\\
              (\lambda_n+\mu_n)\lim_{x \to L, x < L}\frac{d}{dx}w_m(x) &= \lambda_n \lim_{x \to L, x >L}\frac{d}{dx}w_m(x).
    \end{align*}
        \item[$(v)$] One has $\int_{-2L}^{2L} w_m^2 = 1$.
    \end{itemize}

\paragraph{Step 3: Symmetric eigenfunctions.} Our goal in this paragraph is to describe the symmetric eigenfunctions, i.e., $w_m(-x) = w_m(x)$. We denote by $a_m^{\mathrm{sym}}$ the eigenvalues of such eigenfunctions. From statements $(i)$ and $(ii)$ above, we deduce that there are two constant $A$ and $C$ such that
\begin{align*}
        \forall -2L \leqslant x \leqslant -L,\;    &w_m(x) = A \cosh(x+2L) ,\\
             \forall -L \leqslant x \leqslant L,\;    &w_m(x) = C\cos\Big(\sqrt{\frac{a_m^{-1}-\lambda_n}{\lambda_n + \mu_n}}x\Big),\\
              \forall L \leqslant x \leqslant 2L,\;    &w_m(x) = A \cosh(x-2L).
        \end{align*}
Applying $(iii)$ at $x = -L$ leads to
\begin{equation}
    A \cosh(L) = C \cos\Big(\sqrt{\frac{(a_m^{\mathrm{sym}})^{-1}-\lambda_n}{\lambda_n + \mu_n}}L\Big).
    \label{eq:continuite}
\end{equation} 
Similarly, statement $(iv)$ applied at $x = -L$ shows that
\begin{equation}
    \lambda_n  A \sinh(L) = - (\lambda_n  + \mu_n) C \sqrt{\frac{(a_m^{\mathrm{sym}})^{-1}-\lambda_n}{\lambda_n + \mu_n}} \sin\Big(-\sqrt{\frac{(a_m^{\mathrm{sym}})^{-1}-\lambda_n}{\lambda_n + \mu_n}}L\Big).
    \label{eq:derivation}
\end{equation}
Dividing \eqref{eq:derivation} by \eqref{eq:continuite} leads to
\[L \sqrt{\frac{(a_m^{\mathrm{sym}})^{-1}-\lambda_n}{\lambda_n + \mu_n}} \tan\Big(\sqrt{\frac{(a_m^{\mathrm{sym}})^{-1}-\lambda_n}{\lambda_n + \mu_n}}L\Big) = L \frac{\lambda_n}{\lambda_n + \mu_n}   \tanh(L).\]
The equation $x \tan(x) = \tilde C$, where $\tilde C$ is constant, has exactly one solution in any interval $[\pi(k - 1/2), \pi(k+1/2)]$ for $k \in \mathbb{Z}$. Therefore, there is only one admissible value of $\sqrt{\frac{(a_m^{\mathrm{sym}})^{-1}-\lambda_n}{\lambda_n + \mu_n}}$ in each of these interval. So,
\[ \lambda_n + (\lambda_n + \mu_n)(m-1/2)^2 \pi^2/L^2 \leqslant (a_m^{\mathrm{sym}})^{-1} \leqslant \lambda_n + (\lambda_n + \mu_n)(m+1/2)^2 \pi^2/L^2.\]

\paragraph{Step 4: Antisymmetric eigenfunctions.} Our goal in this paragraph is to describe the antisymmetric eigenfunctions, i.e., $w_m(-x) = -w_m(x)$. We denote by $a_m^{\mathrm{anti}}$ the eigenvalues of such eigenfunctions. From statements $(i)$ and $(ii)$, we deduce that there are two constant $B$ and $D$ such that
\begin{equation*}
        \left\{ \begin{array}{cc}
             \forall -2L \leqslant x \leqslant -L,    &w_m(x) = B \sinh(x+2L) ,\\
             \forall -L \leqslant x \leqslant L,    &w_m(x) = D\sin\Big(\sqrt{\frac{a_m^{-1}-\lambda_n}{\lambda_n + \mu_n}}x\Big),\\
              \forall L \leqslant x \leqslant 2L,    &w_m(x) = B \sinh(x-2L).
        \end{array}\right.
        \end{equation*}
Applying $(iii)$ at $x = -L$, one has
\begin{equation}
    B \sinh(L) = D \sin\Big(\sqrt{\frac{(a_m^{\mathrm{anti}})^{-1}-\lambda_n}{\lambda_n + \mu_n}}L\Big).
    \label{eq:continuite2}
\end{equation} 
Similarly, applying $(iv)$ at $x = -L$ shows that
\begin{equation}
    \lambda_n  B \cosh(L) =  (\lambda_n  + \mu_n) D \sqrt{\frac{(a_m^{\mathrm{anti}})^{-1}-\lambda_n}{\lambda_n + \mu_n}} \cos\Big(-\sqrt{\frac{(a_m^{\mathrm{anti}})^{-1}-\lambda_n}{\lambda_n + \mu_n}}L\Big).
    \label{eq:derivation2}
\end{equation}
Dividing  \eqref{eq:continuite2} by \eqref{eq:derivation2} leads to
\[L \Big({\frac{(a_m^{\mathrm{anti}})^{-1}-\lambda_n}{\lambda_n + \mu_n}}\Big)^{-1/2} \tan\Big(\sqrt{\frac{(a_m^{\mathrm{anti}})^{-1}-\lambda_n}{\lambda_n + \mu_n}}L\Big) = L (1+\frac{\mu}{\lambda_n})   \tanh(L).\]
The equation $\tan(x)/x = \tilde C$, where $\tilde C$ is constant, has exactly one solution in any interval $[\pi(k - 1/2), \pi(k+1/2)]$ for $k \in \mathbb{Z}$. Therefore, there is only one admissible value of $\sqrt{\frac{(a_m^{\mathrm{anti}})^{-1}-\lambda_n}{\lambda_n + \mu_n}}$ in each of these interval. So,
\[ \lambda_n + (\lambda_n + \mu_n)(m-1/2)^2 \pi^2/L^2 \leqslant (a_m^{\mathrm{anti}})^{-1} \leqslant \lambda_n + (\lambda_n + \mu_n)(m+1/2)^2 \pi^2/L^2.\]
\paragraph{Step 5: Conclusion.} Recall that the sequence 
$(a_m)_{m\in\mathbb{N}}$ is a non-increasing re-indexing of the sequences $(a_m^{\mathrm{sym}})_{m\in\mathbb{N}}$ and $(a_m^{\mathrm{anti}})_{m\in\mathbb{N}}$. Putting the bounds obtained for $a_m^{\mathrm{sym}}$ and $a_m^{\mathrm{anti}}$ together, we obtain 
\[ \lambda_n + (\lambda_n + \mu_n)(m/2-1)^2 \pi^2/L^2 \leqslant a_m^{-1} \leqslant \lambda_n + (\lambda_n + \mu_n)(m/2+1)^2 \pi^2/L^2,\]
and 
\[ (\lambda_n + \mu_n)(m-2)^2 \pi^2/(4L^2) \leqslant a_m^{-1} \leqslant (\lambda_n + \mu_n)(m+4)^2 \pi^2/(4L^2).\]
We conclude that
\[ \frac{4L^2}{(\lambda_n + \mu_n)(m+4)^2 \pi^2} \leqslant a_m \leqslant \frac{4L^2}{(\lambda_n + \mu_n)(m-2)^2 \pi^2}.\]

\subsection*{Proof of Theorem \ref{prop:kernel_speed_up}}
This is a straightforward consequence of Proposition \ref{prop:1ddiff}, identity \eqref{eq:bornepoly}, and Theorem \ref{thm:boundexp}.

    


\renewcommand\thesection{\thechapter.\arabic{section}}

\chapter{Physics-informed kernel learning}
\label{ch:related-work}

This chapter corresponds to the following publication: \citet{doumèche2024physicsinformedkernellearning}.

\section{Introduction}

\paragraph{Physics-informed machine learning.} Physics-informed machine learning (PIML), as described by \citet{raissi2019PINN}, is a promising framework that combines statistical and physical principles to leverage the strengths of both fields. PIML can be applied to a variety of problems, such as solving partial differential equations (PDEs) using machine learning techniques, leveraging PDEs to accelerate the learning of unknown functions (hybrid modeling), and learning PDEs directly from data (inverse problems). For an introduction to the field and a literature review, we refer to \citet{karniadakis2021piml} and \citet{cuomo2022scientific}. 

\paragraph{Hybrid modeling setting.}
We consider in this paper the classical regression model, which aims at learning the unknown function $f^\star : \mathbb{R}^d \to \mathbb{R}$ such that $Y = f^\star(X) + \varepsilon$, where $Y \in \mathbb{R}$ is the output, $X \in \Omega$ are the features with $\Omega \subseteq [-L, L]^d$ the input domain, and $\varepsilon$ is a random noise. 
Using $n$ observations $(X_1, Y_1), \ldots, (X_n, Y_n)$, independent copies of $(X, Y)$, the goal is to construct an estimator $ \hat{f}_n$ of $f^\star$. What makes PIML special compared to other regression settings is the prior knowledge that $f^\star$ approximately follows a PDE. Therefore, we assume that $f^\star$ is weakly differentiable up to the order $s > \frac{d}{2}$ and that there exists a known differential operator~$\mathscr{D}$ such that $\mathscr{D}(f^\star) \simeq 0$. This framework typically accounts for modeling error by recognizing that $\mathscr{D}(f^\star)$ may not be exactly zero, since most PDEs in physics are derived under ideal conditions and may not hold exactly in practice. 
For example, if $f^\star$ is expected to satisfy the wave equation $\partial^2_{t} f(x,t) \simeq \partial^2_{x} f(x,t)$, we define the operator $\mathscr{D}(f)(x,t) = \partial^2_{t} f(x,t) - \partial^2_{x} f(x,t)$ for $(x,t) \in \Omega$.

To estimate $f^\star$, we consider the minimizer of the physics-informed empirical risk 
\begin{equation}
    R_n(f) = \frac{1}{n}\sum_{i=1}^n |f(X_i) - Y_i|^2 + \lambda_n \|f\|^2_{H^s(\Omega)} + \mu_n \|\mathscr{D}(f)\|^2_{L^2(\Omega)}
    \label{eq:riskBase}
\end{equation}
over the class $\mathscr{F} = H^s(\Omega)$ of candidate functions, where $\lambda_n > 0 $ and $\mu_n \geqslant  0$ are hyperparameters that weight the relative importance of each term. Here, $H^s(\Omega)$ denotes the Sobolev space of functions with weak derivatives up to order $s$.
The empirical risk function $R_n(f)$ is characteristic of hybrid modeling, as it is composed of:
\begin{itemize}
\item A data fidelity term $\frac{1}{n}\sum_{i=1}^n |f(X_i) - Y_i|^2$, which is standard in supervised learning and measures the discrepancy between the predicted values $f(X_i)$ and the observed targets $Y_i$;
\item A	regularization term $\lambda_n \|f\|_{H^s(\Omega)}^2$, which penalizes the regularity of the estimator; 
\item A model error term $\mu_n \|\mathscr{D}(f)\|_{L^2(\Omega)}^2$, which measures the deviation of $f$ from the physical prior encoded in the differential operator $\mathscr{D}$.
To put it simply, the lower this term, the more closely the estimator aligns with the underlying physical principles.
\end{itemize}
Throughout the paper, we refer to $\hat{f}_n$ as the unique minimizer of the empirical risk function, i.e.,
\begin{equation}
    \hat f_n = \mathop{\mathrm{argmin}}_{f \in H^s(\Omega)}\; R_n(f).
    \label{eq:estimator_sob_0}
\end{equation}

\paragraph{Algorithms to solve the PIML problem.} 
Various algorithms have been proposed to compute the estimator $\hat f_n$, and physics-informed neural networks (PINNs) have emerged as a leading approach \citep[e.g.,][]{raissi2019PINN, arzani2021uncovering, karniadakis2021piml, kurz2022hybrid, Agharafeie2023from}. 
PINNs are usually trained by minimizing a discretized version of the risk over a class of neural networks using gradient descent strategies.
Leveraging the good approximation properties of neural networks, as the size of the PINN grows, this type of estimator typically converges to the unique minimizer over the entire space $H^s(\Omega)$ \citep{shin2020convergence,  doumeche2023convergence, mishra2022generalization, shin2023error, bonito2024convergenceerrorcontrolconsistent}. 
However, apart from the fact that optimizing PINNs by gradient descent is an art in itself, the theoretical understanding of the estimators derived through this approach is far from complete \citep{bonfanti2024challengesnonlinearregimephysicsinformed,  rathore2024challengestrainingpinnsloss}, and only a few initial studies have begun to outline their theoretical contours \citep[][]{krishnapriyan2021characterizing, wang2022when,doumeche2023convergence}.
Alternative algorithms for physics-informed learning have since been developed, primarily based on kernel methods, and are seen as promising candidates for bridging the gap between machine learning and PDEs. 
The connections between PDEs and kernel methods are now well established \citep[e.g.,][]{schaback2006from, chen2021solving, batlle2023error}. 
Recently, a kernel method has been adapted to perform operator learning \citep{nelsen2024operator}. It consists of solving a PDE using samples of the initial condition (with a purely data driven empirical risk).

\paragraph{Quantifying the impact of physics.} 
Understanding how physics can enhance learning is of critical importance to the PIML community. 
\citet{arnone2022spatialRegression} show that for second-order elliptic PDEs in dimension $d=2$, 
the PIML estimator converges at a rate of $n^{-4/5}$, outperforming the Sobolev minimax rate of $n^{-2/3}$.

\paragraph{Kernel formulation in the Fourier space.}
This work builds on the results of \citet{doumeche2024physicsinformed}, which shows that, in the case of hybrid modeling (potentially including a noise $\varepsilon \neq 0$ and a modeling error, i.e., $\mathscr D(f^\star) \neq 0$), the PIML problem \eqref{eq:estimator_sob_0} can be reformulated as a kernel regression task. 
Provided the associated kernel $K$ is made explicit, this reformulation allows to obtain a closed-form estimator that converges at least at the Sobolev minimax rate. 
However, the kernel $K$ is highly dependent on the underlying PDE, and its computation can be tedious even for simple priors, such as $\mathscr{D} = \frac{d}{dx}$ in one dimension.
Thus, one of the goals of the present paper is to propose an approximation of $K$, making it possible to implement this kernel method in practice.

For general linear PDEs in dimension $d$, \citet{doumeche2024physicsinformed} have adapted to PIML the notion of effective dimension, a central idea in kernel methods that quantify their convergence rate. 
As a result, for $d=1$, $s = 1$, $\Omega = [-L, L]$, and $\mathscr D = \frac{d}{dx}$, the authors show that the $L^2$-error of the physics-informed kernel method is of the order of ${\log(n)^2}/{n}$ when $\mathscr D(f^\star) = 0$, and achieves the Sobolev minimax rate $n^{-2/3}$ otherwise. However, extending this type of results to more complex differential operators $\mathscr D$ remains a challenge.
In this context, we show how to approximate the effective dimension, making it possible to experimentally estimate  the convergence rate of a given PIML problem.

\paragraph{Contributions.}
Building on the characterization of the PIML problem as a kernel regression task, we use Fourier methods to approximate the associated kernel $K$ and, in turn, propose a tractable estimator minimizing the physics-informed risk function.
The approach involves developing the kernel $K$ along the Fourier modes with frequencies bounded my $m$, and then taking $m$ as large as possible.
We refer to this approach as the physics-informed kernel learning (PIKL) method. 
Subsequently, for general linear operators $\mathscr{D}$, a numerical strategy is developed to estimate the effective dimension of the kernel problem, allowing for the quantification of the expected statistical convergence rate when incorporating the physics prior into the learning process.   
Finally, we demonstrate the numerical performance of the PIKL estimator through simulations, both in the context of hybrid modeling and in solving partial differential equations. In short, the PIKL algorithm consistently outperforms specialized PINNs from the literature, which were specifically designed for the applications under consideration.

\section{The PIKL estimator}
\label{sec:finite_approx}
In this section, we detail the construction of the PIKL estimator, our approximate kernel method for physics-informed learning.
Throughout this paper, we assume that the differential operator $\mathscr D$ is linear with constant coefficients, as stated in the following assumption.
\begin{assumption}[Linear differential operator with constant coefficients]
    The differential operator $\mathscr{D}: H^s(\Omega) \to L^2(\Omega)$ is linear with constant coefficients, i.e.,  $\mathscr{D}(f) = \sum_{|\alpha| \leqslant s} a_\alpha \partial^\alpha f$ for some $s \in \mathbb{N}^\star$ and $a_\alpha \in \mathbb{R}$.
\end{assumption}
We begin by observing that solving the PIML problem \eqref{eq:estimator_sob_0} is equivalent to performing a kernel regression task, as shown by \citet[][Theorem 3.3]{doumeche2024physicsinformed}. Thus, leveraging the extensive literature on kernel methods, it follows that the estimator $\hat{f}_n$ has the closed-form expression
\[
\hat f_n = \Big(x\mapsto (K(x, X_1), \hdots, K(x, X_n)) (\mathbb K + n I_n)^{-1} \mathbb Y\Big),
\]
where $K: \Omega^2 \to \mathbb{R}$ is the kernel associated with the squared norm $\lambda_n \|f\|^2_{H^s(\Omega)} + \mu_n \|\mathscr{D}(f)\|^2_{L^2(\Omega)}$ of~\eqref{eq:riskBase}, and $\mathbb{K}$ is the $n\times n$ kernel matrix defined by $\mathbb K_{i,j} = K(X_i, X_j)$.

\begin{figure}
    \centering
    \includegraphics[width = 0.45\textwidth]{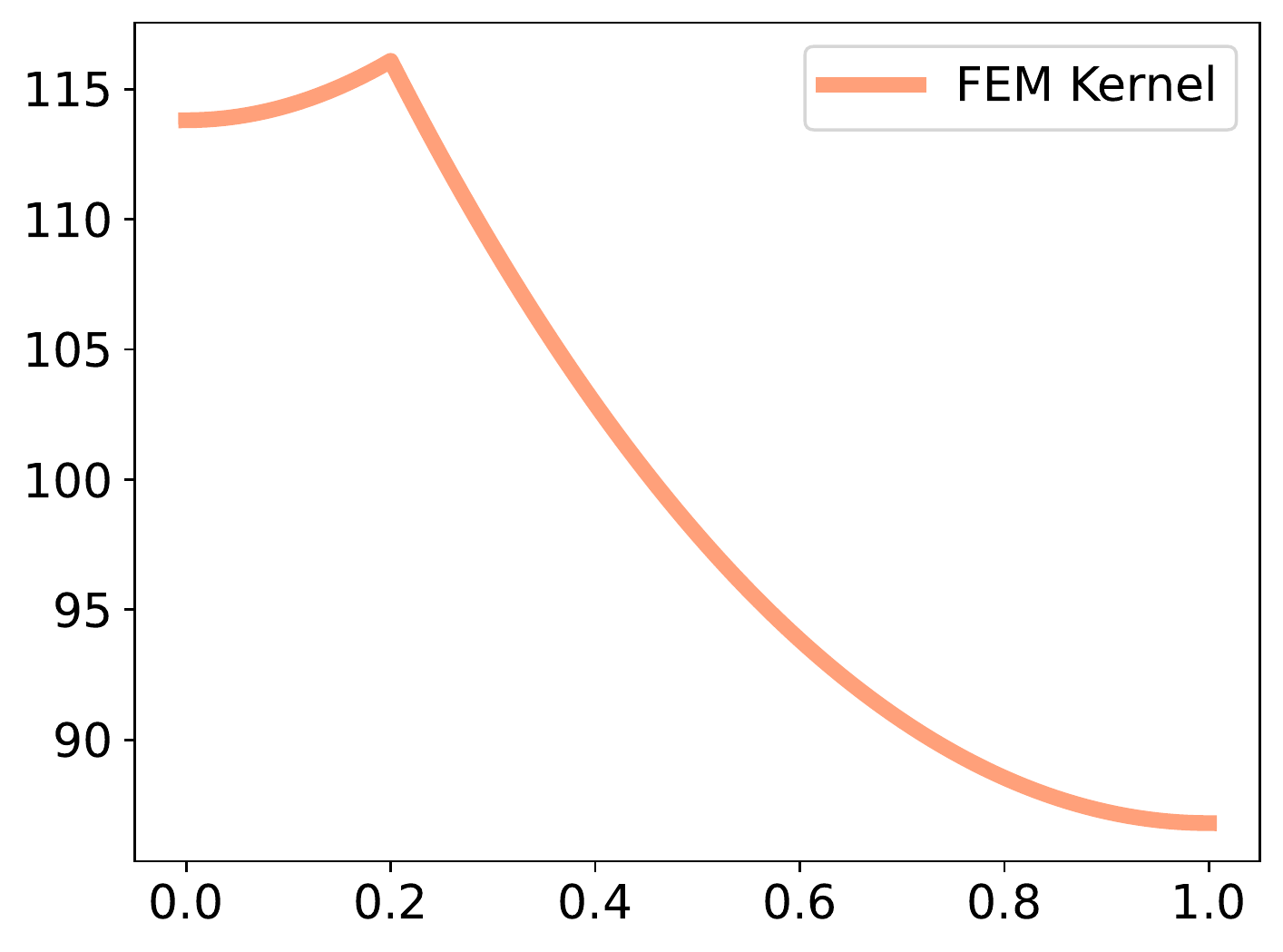}
    \includegraphics[width = 0.45\textwidth]{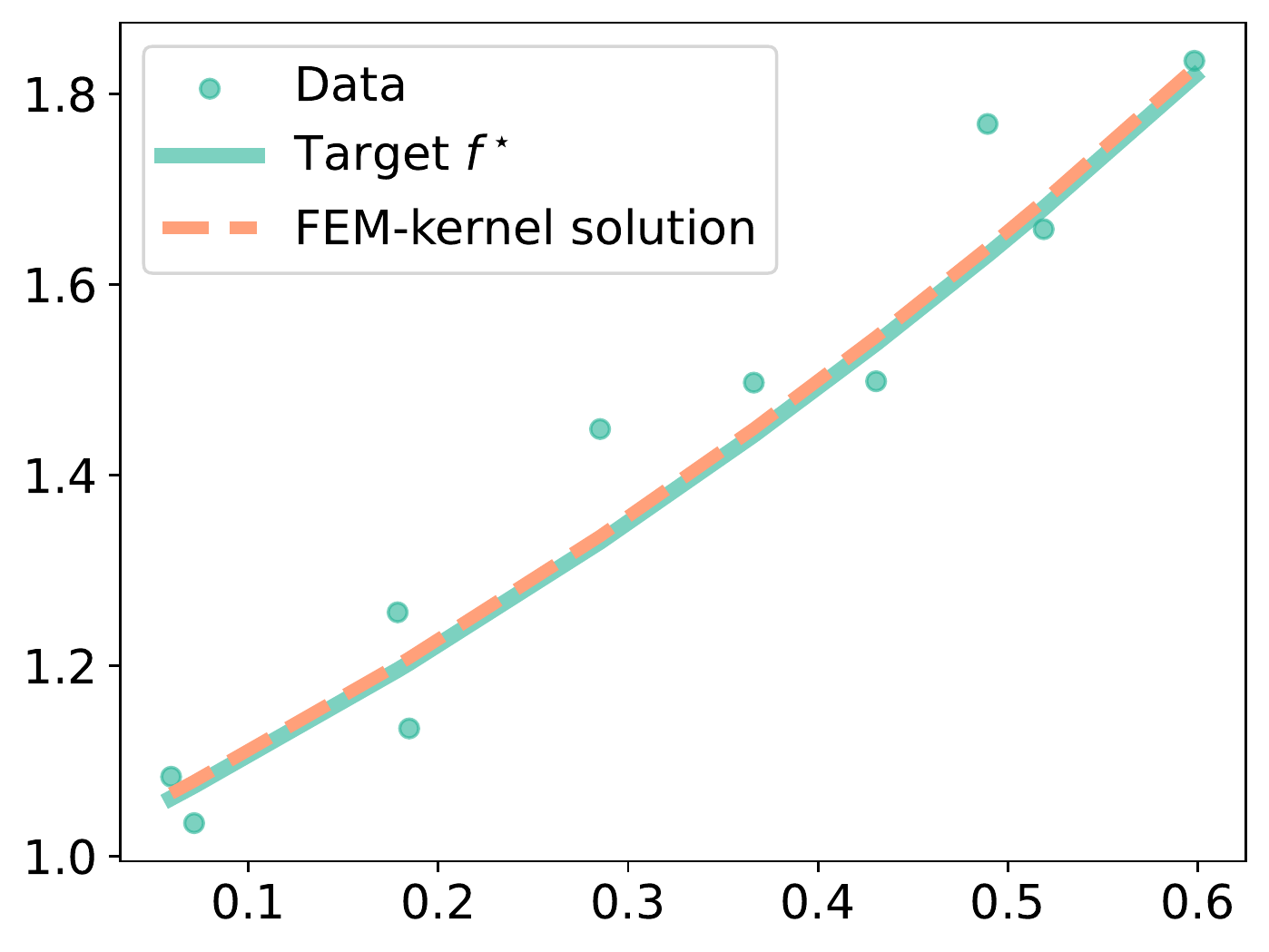}
    \caption{\textbf{Left:} Kernel function $K(0.4, \cdot)$ estimated by the FEM. \textbf{Right:} Kernel method $\hat f_n$ combined with the FEM.}
    \label{fig:fem}
\end{figure}

\paragraph{A finite-element-method approach.} 
The analysis of \citet[][Proposition 3.4]{doumeche2024physicsinformed} reveals that the kernel related to the PIML problem is uniquely characterized as the solution to a weak PDE. Indeed, for all $x \in \Omega$, the function $y \mapsto K(x, y)$ is the unique solution in $H^s(\Omega)$ to the weak formulation
\begin{equation}
    \forall \phi \in H^s(\Omega), \quad \lambda_n  \int_{\Omega}\big[K(x, \cdot)\; \phi +\sum_{|\alpha| = s}\partial^\alpha K(x, \cdot)\; \partial^\alpha  \phi\big] + \mu_n \int_{\Omega}\mathscr{D}(K(x, \cdot)) \; \mathscr{D}(\phi) = \phi(x).
    \label{eq:weakPDE}
\end{equation}
\citep[This is a consequence of ][Proposition 3.4, applied to the risk $R_n$.]{doumeche2024physicsinformed}
A spontaneous idea is to approximate the kernel $K$ using finite element methods (FEM). For illustrative purposes, we have applied this approach in numerical experiments with $d=1$, $\Omega = [0,1]$, and $\mathscr{D}(f) = \frac{d}{dx}f - f$. Figure \ref{fig:fem} (Left) depicts the associated kernel function $K(0.4, \cdot)$ with $\lambda_n = 10^{-2}$, $\mu_n = 0$, and $100$ nodes. Figure \ref{fig:fem} (Right) shows that the PIML method \eqref{eq:estimator_sob_0} successfully reconstructs $f^\star(x) = \exp(x)$ using $n = 10$ data points, $\varepsilon \sim \mathcal{N}(0, 10^{-2})$, $\lambda_n = 10^{-10}$, and $\mu_n = 1000$. However, solving the weak formulation \eqref{eq:weakPDE} in full generality is quite challenging, particularly when dealing with arbitrary domains $\Omega$ in dimension $d>1$. In fact, FEM strategies need to be specifically tailored to the PDE and the domain in question. Additionally, standard kernel methods combined with FEM approaches come at a high computational cost, since storing the matrix $\mathbb{K}$ requires $O(n^2)$ memory. This becomes prohibitive for large amounts of data, as $n = 10^4$ already requires several gigabytes of RAM. 

\paragraph{Fourier approximation.}
Our primary objective in this article is to develop a more agile, flexible, and efficient method capable of handling arbitrary domains $\Omega$. 
Following \citet{doumeche2024physicsinformed}, our methodology first requires extending the learning problem from $\Omega\subseteq [-L,L]^d$ to the torus $[-2L,2L]^d$. 
Indeed, any function of $H^s(\Omega)$ can be periodically extended into a function of $H^s_{\mathrm{per}}([-2L,2L]^d)$ \citep[see][Proposition~A.6 and Figure~1]{doumeche2024physicsinformed}.
The choice of $[-2L,2L]^d$ is commented in Appendix~\ref{sec:extended_domain}.
This initial technical step allows us to use approximations with the standard Fourier basis, given for $k\in\mathbb{Z}^d$ and $x\in [-2L,2L]^d$ by
\[
\phi_k(x)=(4L)^{-d/2} e^{\frac{i \pi}{2L}\langle k, x\rangle},
\]
particularly adapted to periodic functions on $[-2L,2L]^d$. 
Therefore, the minimization of the risk $R_n$ defined in \eqref{eq:riskBase} over $H^s(\Omega)$ can then be transferred into the minimization of the PIML risk 
\begin{equation*}
    \bar R_n(f) = \frac{1}{n}\sum_{i=1}^n |f(X_i) - Y_i|^2 + \lambda_n \|f\|^2_{H^s_{\mathrm{per}}([-2L,2L]^d)} + \mu_n \|\mathscr{D}(f)\|^2_{L^2(\Omega)}
\end{equation*}
over the periodic Sobolev space $H^s_{\mathrm{per}}([-2L,2L]^d)$ (see Appendix~\ref{sec:sobolev_intro} for an introduction to periodic Sobolev spaces).
This results in a slightly modified kernel, 
determined by the RKHS norm
\[\|f\|^2_{\mathrm{RKHS}} = \lambda_n \|f\|_{H^s_{\mathrm{per}}([-2L,2L]^d)}^2 + \mu_n \|\mathscr D(f)\|_{L^2(\Omega)}^2.\] 
It is important to note that the estimators derived from the minimization of either $R_n$ or $\bar{R}_n$ share the same statistical guarantees, as both kernel methods have been shown to converge to $f^\star$ at the same rate \citep[][Theorem 4.6]{doumeche2024physicsinformed}.

To implement this kernel method, a natural approach is to expand the kernel using a truncated Fourier series, i.e.,
$K_m(x,y) = \sum_{\|k\|_\infty \leqslant m} a_k \phi_k(x)\phi_k(y)$, 
with $(\phi_k)_{\|k\|_\infty \leqslant m}$ the Fourier basis, $(a_k)_{\|k\|_\infty \leqslant m}$ the kernel coefficients in this basis, and $m$ the order of approximation.
This idea is at the core of techniques such as random Fourier features (RFF) \citep[e.g.,][]{rahimi2007random, yang2012nystrom}. 
However, unlike RFF, the Fourier features in our problem are not random quantities, as they systematically correspond to the low-frequency modes. 
This low-frequency approximation is particularly well-suited to the Sobolev penalty, which more strongly regularizes high frequencies (the analogous RFF algorithm would involve sampling random frequencies $k$ according to a density that is proportional to the Sobolev decay). 
In addition, and more importantly, the use of such approximations bypasses the need to discretize the domain into finite elements and requires only the knowledge of the (partial) Fourier transform of $\mathbf{1}_\Omega$, as will be explained later.

A key milestone in the development of our method is to 
minimize $\bar{R}_n$ not over the entire space $H^s_\mathrm{per}([-2L,2L]^d)$, but rather on the finite-dimensional Fourier subspace $H_m=\mathrm{Span}((\phi_k)_{\|k\|_\infty \leqslant m})$. This leads to the PIKL estimator, defined by
\begin{align}
\hat{f}^{\mathrm{PIKL}} = \mathop{\mathrm{argmin}}_{f \in H_m}\; \bar R_n(f).
\label{pb:PIML_on_Hm}
\end{align}
This naturally transforms the PIML problem into a finite-dimensional kernel regression task, where the associated kernel $K_m$ corresponds to a Fourier expansion of $K$, as will be clarified in the following paragraph. Of course, $H_m$ provides better approximates of $H^s_\mathrm{per}([-2L,2L]^d)$
as $m$ increases, since for any function $f \in H^s_\mathrm{per}([-2L,2L]^d)$, $\lim_{m\to\infty}\min_{g\in H_m}\|f-g\|_{H^s_\mathrm{per}([-2L,2L]^d)} = 0$. 
Remarkably, the key advantage of using Fourier approximations in our PIKL algorithm lies in the fact that
both the squared Sobolev norm $\|f\|_{H^s_\mathrm{per}([-2L,2L]^d)}^2$ and the PDE penalty $\|\mathscr{D}(f)\|_{L^2(\Omega)}^2$ are bilinear functions of the Fourier coefficients of $f$. As shown below, these bilinear forms can be represented as closed-form matrices, easing the computation of the estimator.

\paragraph{RKHS norm in Fourier space.} Suppose that the differential operator $\mathscr{D}$ is linear with constant coefficients, i.e., it can be expressed as
$\mathscr{D}(f) = \sum_{|\alpha| \leqslant s} a_\alpha \partial^\alpha f$ for some $s \in \mathbb{N}^\star$ and $a_\alpha \in \mathbb{R}$.
If $f \in H_m$, then $f$ can be rewritten in terms of its Fourier coefficients as
\[
f(x) = \langle z, \Phi_m(x) \rangle_{\mathbb C^{(2m+1)^d}},
\]
where $\langle \cdot, \cdot \rangle_{\mathbb C^{(2m+1)^d}}$ denotes the canonical inner product on $\mathbb C^{(2m+1)^d}$, $z$ is the vector of Fourier coefficients of $f$, and 
\[
\Phi_m(x) = \Big((4L)^{-d/2} e^{\frac{i \pi}{2L}\langle k, x\rangle}\ \Big)_{\|k\|_\infty\leqslant m}.
\] 
According to Parseval's theorem, the $L^2$-norm of the derivatives of $f\in H^s_{\mathrm{per}}([-2L,2L]^d)$ can be expressed using the Fourier coefficients of $f$ as follows: for $r\leqslant s$ and $1 \leqslant i_1, \hdots, i_r \leqslant d$, 
\[ 
\|\partial^r_{i_1, \hdots, i_r}f\|_{L^2([-2L,2L]^d)}^2 = (2L)^{-2k}\sum_{\|j\|_\infty\leqslant m} |z_j|^2 \prod_{\ell=1}^r j_{i_\ell}^2.
\]
With this notation, the Sobolev norm reads
\[
\|f\|_{H^s_{\mathrm{per}}([-2L,2L]^d)}^2 
= \sum_{\|j\|_\infty\leqslant m,\|k\|_\infty\leqslant m} z_j \bar{z_k} \Big(1+\Big(\frac{\|k\|_2^2}{(2L)^d}\Big)^{s}\Big)\delta_{j,k},
\]
and, similarly,
\[\|\mathscr D(f)\|_{L^2(\Omega)}^2 = \sum_{\|j\|_\infty\leqslant m,\|k\|_\infty\leqslant m} z_j \bar{z_k} \frac{P(j)\bar P(k)}{(4L)^d}\int_\Omega e^{\frac{i \pi}{2L}\langle k-j, x\rangle}dx,\]
where $P(k) = \sum_{|\alpha| \leqslant s} a_\alpha (\frac{-i\pi}{2L})^{|\alpha|}\prod_{\ell =1}^d  (k_\ell)^{\alpha_\ell} $. 
Therefore, introducing $M_m$ the 
 $(2m+1)^d\times (2m+1)^d$ matrix with coefficients indexed by $j,k\in \{-m, \hdots, m\}^d$,
\begin{equation}
    (M_m)_{j,k} = \lambda_n \Big(1+\Big(\frac{\|k\|_2^2}{(2L)^d}\Big)^{s}\Big)\delta_{j,k}+\mu_n \frac{P(j)\bar P(k)}{(4L)^d}\int_\Omega e^{\frac{i \pi}{2L}\langle k-j, x\rangle}dx,
    \label{eq:Mm}
\end{equation}
we obtain that the RKHS norm of $f$ is expressed as a bilinear form of its Fourier coefficients $z$, i.e.,  
\[
\|f\|^2_{\mathrm{RKHS}} = \langle z, M_mz\rangle_{\mathbb C^{(2m+1)^d}}.
\]
It is important to note that $M_m$ is Hermitian,\footnote{since $M_m^\star = \bar M_m^\top = M_m$.} positive,\footnote{since $\langle z, M_mz\rangle_{\mathbb C^{(2m+1)^d}} = \|f\|^2_{\mathrm{RKHS}} \geqslant  0$.} and definite.\footnote{since $\langle z, M_mz\rangle_{\mathbb C^{(2m+1)^d}} = 0$ implies $ \|f\|_{H^s([-2L,2L]^d)} = 0$, i.e., $f = 0$.} Therefore, the spectral theorem (see Theorem~\ref{thm:spectral2}) ensures that $M_m$ is invertible, and that its positive inverse square root $M_m^{-1/2}$ is unique and well-defined. We have now all the ingredients to define the PIKL algorithm.

\begin{remark}[Linear PDEs with non-constant coefficients]
    This framework could be adapted to PDEs with non-constant coefficients, i.e., to operators $\mathscr{D}(f) = \sum_{|\alpha| \leqslant s} a_\alpha \partial^\alpha f$ for some $s \in \mathbb{N}^\star$ and $a_\alpha \in C^0(\mathbb{R})$. In this case, the polynomial $P$ in \eqref{eq:Mm} should be replaced by convolutions involving the Fourier coefficients of the functions $a_\alpha$.
    \end{remark}

\paragraph{Computing the PIKL estimator.} 
For a function $f\in H_m$, one can evaluate $f$ at $x$ by $f(x) = \langle M_m^{1/2}z, M_m^{-1/2}\Phi_m(x)\rangle_{\mathbb C^{(2m+1)^d}}$. This reproducing property indicates that minimizing the risk $\bar R_n$ on $H_m$ is a kernel method governed by the kernel
\[
K_m(x,y) = \langle  M_m^{-1/2} \Phi_m(x), M_m^{-1/2}\Phi_m(y)\rangle_{\mathbb C^{(2m+1)^d}}.
\]
Define $\mathbb Y = (Y_1,\hdots,Y_n)^\top$ and $\mathbb K_m \in \mathcal M_n(\mathbb C)$ to be the matrix such that $(\mathbb K_m)_{i,j} = K_m(X_i, X_j)$ for all $1 \leqslant i, j \leqslant n$. 
The PIKL estimator \eqref{pb:PIML_on_Hm}, minimizer of $\bar{R}_n$ restricted to $H_m$, is therefore given by
\begin{align}           \hat{f}^{\mathrm{PIKL}} (x)
&= (K_m(x, X_1), \hdots, K_m(x, X_n)) (\mathbb K_m + n I_n)^{-1} \mathbb Y \nonumber\\
&= \Phi_m(x)^\star (\mathbb{\Phi}^\star\mathbb{\Phi} + n M_m)^{-1} \mathbb \Phi^\star \mathbb Y,\label{eq:PIKL}
\end{align}
where 
$
\mathbb{\Phi} = \begin{pmatrix}
    \Phi_m(X_1)^\star\\
    \vdots\\
    \Phi_m(X_n)^\star
\end{pmatrix}$ is an $n\times  (2m+1)^d$ matrix. The formula obtained in \eqref{eq:PIKL} is provided by the so-called kernel trick.  
This step offers a significant advantage to the PIKL estimator as it reduces the computational burden in large sample regimes: instead of storing and inverting the $n \times n$ matrix $\mathbb{K}_m + nI_n$, we only need to store and invert the $(2m+1)^d \times (2m+1)^d$ matrix $\mathbb{\Phi}^\star \mathbb{\Phi} + n M_m$. Moreover, the computation of $\mathbb{\Phi}^\star \mathbb{\Phi}$ and $\mathbb \Phi^\star \mathbb Y$ can be performed online and in parallel as $n$ grows. Of course, this approach is subject to the curse of dimensionality. However, it is unreasonable to try to learn more parameters than the sample complexity $n$. Therefore, in practice, $(2m + 1)^d \ll n$, which justifies the preference of the $(2m + 1)^{2d} $ storage complexity over the $n^2$ storage complexity of the FEM-based algorithm. In addition, similar to PINNs, the PIKL estimator has the advantage that its training phase takes longer than its evaluation at certain points. In fact, once the $(2m + 1)^d$ Fourier modes of $\hat{f}^{\mathrm{PIKL}}$ (given by $(\mathbb{\Phi}^\star\mathbb{\Phi} + nM_m)^{-1}\mathbb{\Phi}^\star \mathbb Y$) are computed, the evaluation of $\Phi^\star_m(x)$ is straightforward. This is in sharp contrast to the FEM-based strategy, which requires approximating the kernel vector $(K(x, X_1), \dots, K(x, X_n))$ at each query point $x$.

We also emphasize that the PIKL predictor is characterized by low-frequency Fourier coefficients, which, in turn, enhance its interpretability.  This methodology differs significantly from PINNs, which are less interpretable and rely on gradient descent for optimization \citep[see, e.g.,][]{wang2022when}.

\begin{remark}[PIKL vs.\ spectral methods] In the PIML context, the RFF approach resembles a well-known class of powerful tools for solving PDEs, known as spectral and pseudo-spectral methods \citep[e.g.,][]{canuto2007spectral}. These methods solve PDEs by selecting a basis of orthogonal functions and computing the coefficients of the solution on that basis to satisfy both the boundary conditions and the PDE itself. For example, the Fourier basis $(x \mapsto \exp(\frac{i \pi}{2L} \langle k, x\rangle ))_{k \in \mathbb{Z}^d}$ already used in this paper is particularly well suited for solving linear PDEs on the square domain $[-2L, 2L]^d$ with periodic boundary conditions. Spectral methods such as these have already been used in the PIML community to integrate PDEs with machine learning techniques \citep[e.g.,][]{meuris2023machine}. However, the basis functions used in spectral and pseudo-spectral methods must be specifically tailored to the domain $\Omega$, the differential operator $\mathscr{D}$, and the boundary conditions.
For more information on this topic, please refer to Appendix~\ref{sec:discussion_spec}.
\end{remark}

\paragraph{Computing $M_m$ for specific domains.} Computing the matrix $M_m$ requires the evaluation of the integrals $(j,k) \mapsto \int_\Omega e^{\frac{i \pi}{2L}\langle k-j, x\rangle}dx$.
In general, these integrals can be approximated using numerical integration schemes or Monte Carlo methods. However, it is possible to provide closed-form expressions for specific domains $\Omega$. 
To do so, for $d \in \mathbb{N}^\star$, $L > 0$, and $\Omega \subseteq [-L, L]^d$, we define the characteristic function $F_\Omega $ of $\Omega$ by 
    \[
    F_\Omega(k) =  \frac{1}{(4L)^{d}}\int_\Omega e^{\frac{i \pi}{2L}\langle k, x\rangle}dx.
    \] 
\begin{prop}[Closed-form characteristic functions]
The characteristic functions associated with the cube and the Euclidean ball can be analytically obtained as follows.
\begin{itemize}
\item (Cube) Let $\Omega = [-L,L]^d$. Then, for $k\in\mathbb{Z}^d$,
    \[F_\Omega(k) =  \prod_{j=1}^d\frac{\sin(\pi k_j/2)}{\pi k_j}.\]
    \label{prop:geom_cube}
\item (Euclidean ball) Let $d=2$ and $\Omega = \{x \in [-L,L], \|x\|_2 \leqslant L\}$. Then, for $k\in\mathbb{Z}^d$,
    \[F_\Omega(k) =   \frac{J_1(\pi \|k\|_2/2)}{4\|k\|_2},\]
where $J_1$ is the  Bessel function of the first kind of parameter 1.   
\end{itemize}
\end{prop}
This proposition, along with similar analytical results for other domains, can be found in \citet[][Table 13.4]{bracewell}, noting that
$F_\Omega$
is the Fourier transform of the indicator function $\mathbf{1}_\Omega$ and is also the characteristic function of the uniform distribution on $\Omega$ evaluated at $\frac{k}{2L}$.
We can extend these computations further since, given the characteristic functions of elementary domains $\Omega$, it is easy to compute the characteristic functions of translation, dilation, disjoint unions, and Cartesian products of such domains
(see Proposition~\ref{prop:op_char} in Appendix~\ref{app:Theory}). 
For instance, it is straightforward to obtain the characteristic function of the three-dimensional  cylinder $\Omega = \{x \in [-L,L], \|x\|_2 \leqslant L\} \times [-L,L]$ as
\[
F_\Omega(k_1, k_2, k_3) = \frac{J_1(\pi (k_1^2+k_2^2)^{1/2}/2)}{4(k_1^2+k_2^2)^{1/2}} \times \frac{\sin(\pi k_3/2)}{\pi k_3}.
\]

\section{The PIKL algorithm in practice}

To enhance the reproducibility of our work, we provide a \texttt{Python} package that implements the PIKL estimator, designed to handle any linear PDE prior with constant coefficients in dimensions $d = 1$ and $d = 2$. This package is available at \url{https://github.com/NathanDoumeche/numerical_PIML_kernel}. $\;\;$ Note that this package implements the matrix inversion of the PIKL formula \eqref{eq:PIKL} by solving a linear system using the LU decomposition. Of course, any other efficient method to avoid direct matrix inversion could be used  instead, such as solving a linear system with the conjugate gradient method. 

Through numerical experiments, we demonstrate the performance of our approach in simulations for hybrid modeling (Subsection~\ref{sec:hybrid_mod}),
and derive experimental convergence rates that quantify the benefits of incorporating PDE knowledge into a learning regression task (Subsection~\ref{sec:eff_dim}).

\subsection{Hybrid modeling}
\label{sec:hybrid_mod}

\begin{wrapfigure}[15]{r}{0.4\textwidth}
    \centering
    \vspace{-1.4em}
    \includegraphics[width=\linewidth]{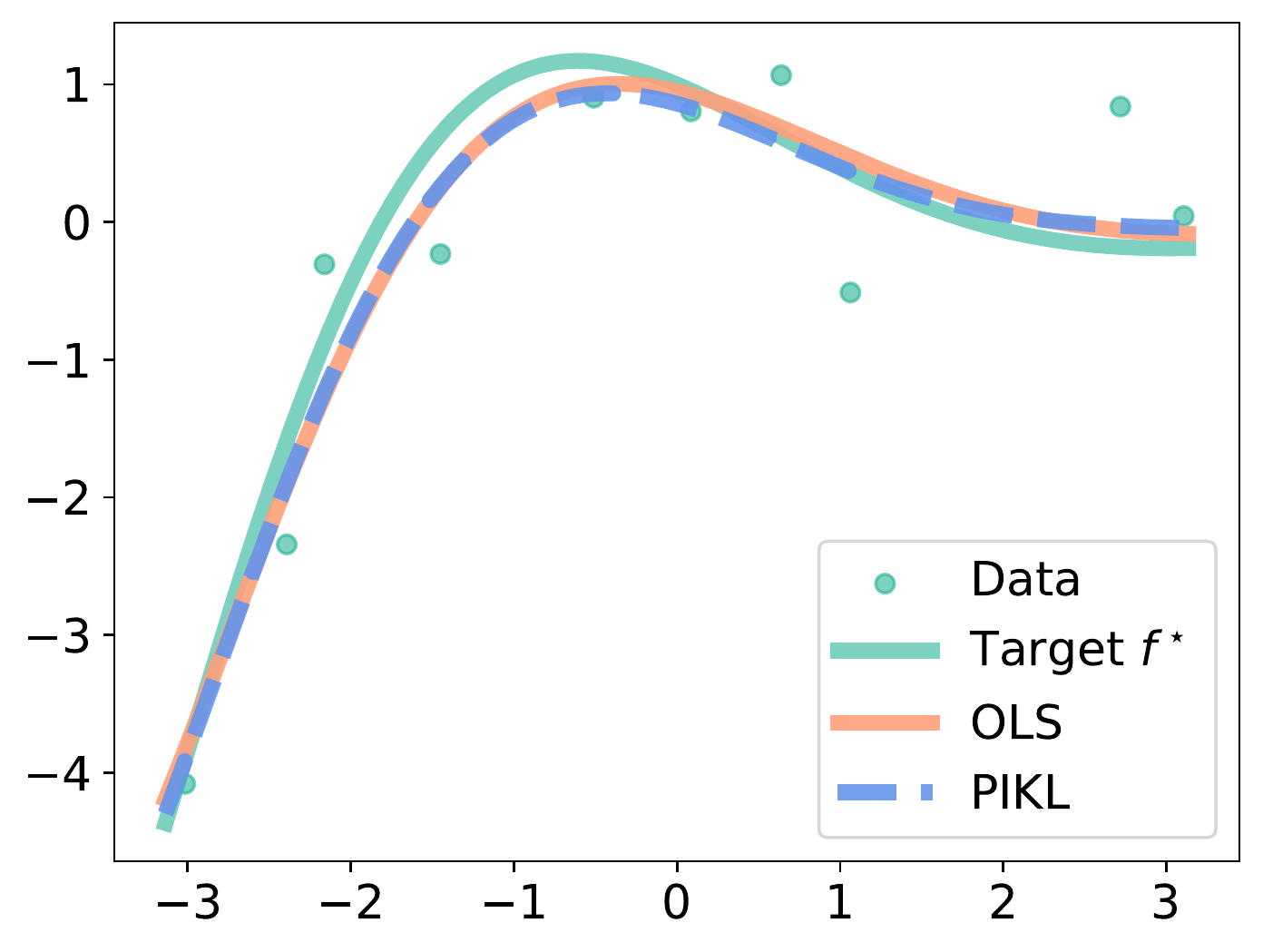}
    \caption{OLS and PIKL estimators for the harmonic oscillator with $d=1$, sample size $n=10$.
    \label{fig:ressort_least_square}}
\end{wrapfigure}

\paragraph{Perfect modeling with closed-form PDE solutions.} 

We start by assessing the performance of the PIKL estimator in a perfect modeling situation (i.e., $\mathscr D(f^\star) = 0$), where the solutions of the PDE $\mathscr D(f) = 0$ can be decomposed on a basis $(f_k)_{k\in \mathbb N}$ of closed-form solution functions. 
In this ideal case, the spectral method suggests an alternative estimator, which involves learning the coefficients $a_k \in \mathbb R$ of $f^\star = \sum_{k\in \mathbb N} a_k f_k$ in this basis.
For example, consider the one-dimensional case ($d = 1$) with domain $\Omega = [-\pi, \pi]$, and the harmonic oscillator differential prior $\mathscr D(f) = \frac{d^2 f}{dx^2} + \frac{df}{dx} + f$. In this case, the solutions of $\mathscr D(f) = 0$ are the linear combinations $f = a_1 f_1 + a_2 f_2$, where  $(a_1, a_2) \in \mathbb R^2$, $f_1(x) = \exp(-x/2)\cos(\sqrt{3}x/2)$, and $f_2(x) = \exp(-x/2)\sin(\sqrt{3}x/2)$. 
Thus, the spectral method focuses on learning the vector $(a_1, a_2)\in \mathbb R^2$, instead of learning the Fourier coefficients of $f^\star$, which is the approach taken by the PIKL algorithm.

A baseline that exactly leverages the particular structure of this problem, referred to as the ordinary least squares (OLS) estimator, is therefore
$\hat g_n = \hat a_1 f_1 + \hat a_2 f_2$, where 
\[
(\hat a_1, \hat a_2) = \mathop{\mathrm{argmin}}_{(a_1, a_2)\in \mathbb R^2}\; \frac{1}{n}\sum_{i=1}^n |a_1 f_1(X_i)+ a_2 f_2(X_i)-Y_i|^2.
\]
\begin{wrapfigure}[16]{r}{0.4\textwidth}
    \centering
    \vspace{-0.2em}
\includegraphics[width=\linewidth]{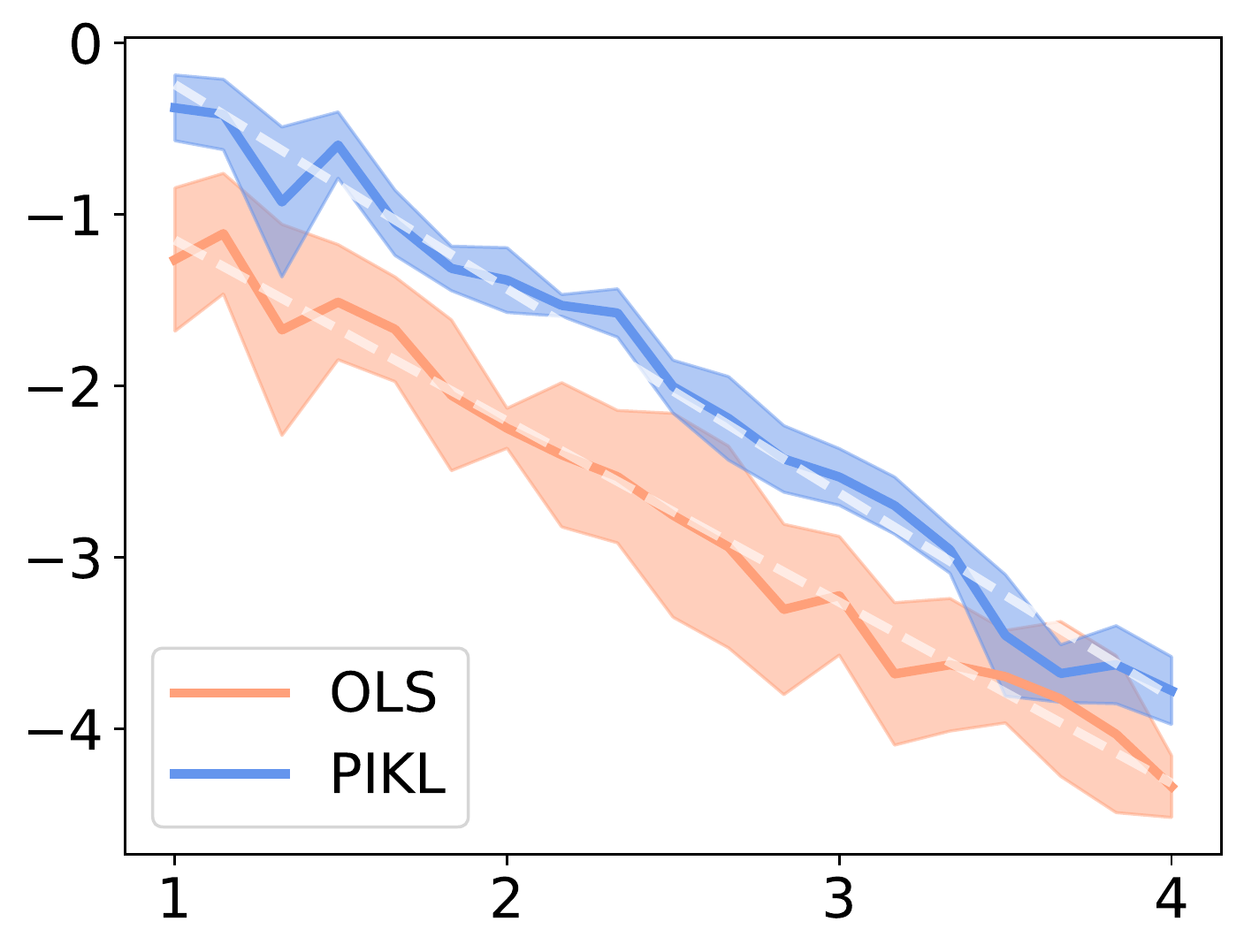}
     \caption{$L^2$-error (mean $\pm$ std over 5 runs) of the OLS and PIKL estimators for the harmonic oscillator with $d=1$, w.r.t.\ $n$ in $\log_{10}-\log_{10}$ scale.  The dashed lines represent adjusted linear models w.r.t.\ $n$, for both $L^2$-errors.\label{fig:perf_ressort_least_square}}
\end{wrapfigure}To compare the PIKL and OLS estimators, we generate data such that $Y = f^\star(X) + \varepsilon$, where $X \sim \mathcal{U}(\Omega)$, $\varepsilon \sim \mathcal{N}(0, \sigma^2)$ with $\sigma = 0.5$, and the target function is $f^\star = f_1$ (corresponding to $(a_1, a_2) = (1, 0)$). We implement the PIKL algorithm with $601$ Fourier modes ($m = 300$) and $s = 2$. 
Figure~\ref{fig:ressort_least_square} shows that even with very few data points ($n = 10$) and high noise levels, both the OLS and PIKL methods effectively reconstruct $f^\star$, both incorporating physical knowledge in their own way. In Figure~\ref{fig:perf_ressort_least_square}, we display the $L^2$-error of both estimators for different sample sizes $n$. The two methods have an experimental convergence rate of $n^{-1.1}$, which is consistent with the expected parametric rate of $n^{-1}$.
This sanity check shows that under perfect modeling conditions, the PIKL estimator with $m=300$ performs as well as the OLS estimator specifically designed to explore the space of PDE solutions.

\paragraph{Combining the best of physics and data in imperfect modeling.}
In this paragraph, we deal with an imperfect modeling scenario using the heat differential operator $\mathscr D(f) = \partial_1 f - \partial^2_{2,2} f$ in dimension $d=2$ over the domain $\Omega = [-\pi, \pi]^2$. The data are generated according to the model $Y = f^\star(X) + \varepsilon$, where $\|\mathscr D(f^\star)\|_{L^2(\Omega)} \neq 0$. 
We assume, however, that the PDE serves as a good physical prior, meaning that $\|f^\star\|_{L^2(\Omega)}^2$ is significantly larger than the modeling error $\|\mathscr D(f^\star)\|_{L^2(\Omega)}^2$.
The hybrid model is implemented using the PIKL estimator with parameters $s=2$, $\lambda_n = n^{-2/3}/10$, and $\mu_n = 100/n$. These hyperparameters are selected to ensure that, when only a small amount of data is available, the model relies heavily on the PDE. 
Yet, as more data become available, the model can use the data to correct the modeling error. 
The performance of the PIKL estimator is compared with that of a purely data-driven estimator, referred to as the Sobolev estimator, and a strongly PDE-penalized estimator, referred to as the PDE estimator. 
The Sobolev estimator uses the same parameter $s=2$ and $\lambda_n = n^{-2/3}/10$, but sets $\mu_n = 0$. This configuration ensures that the estimator relies entirely on the data without considering the PDE as a prior. On the other hand, the PDE estimator is configured with parameters $s=2$, $\lambda_n = 10^{-10}$, and $\mu_n = 10^{10}$. These hyperparameters are set to ensure that the resulting PDE estimator effectively satisfies the heat equation, making it highly dependent on the physical model.

\begin{wrapfigure}[15]{r}{0.5\linewidth}
    \centering
    \vspace{-3em}
    \includegraphics[width=\linewidth]{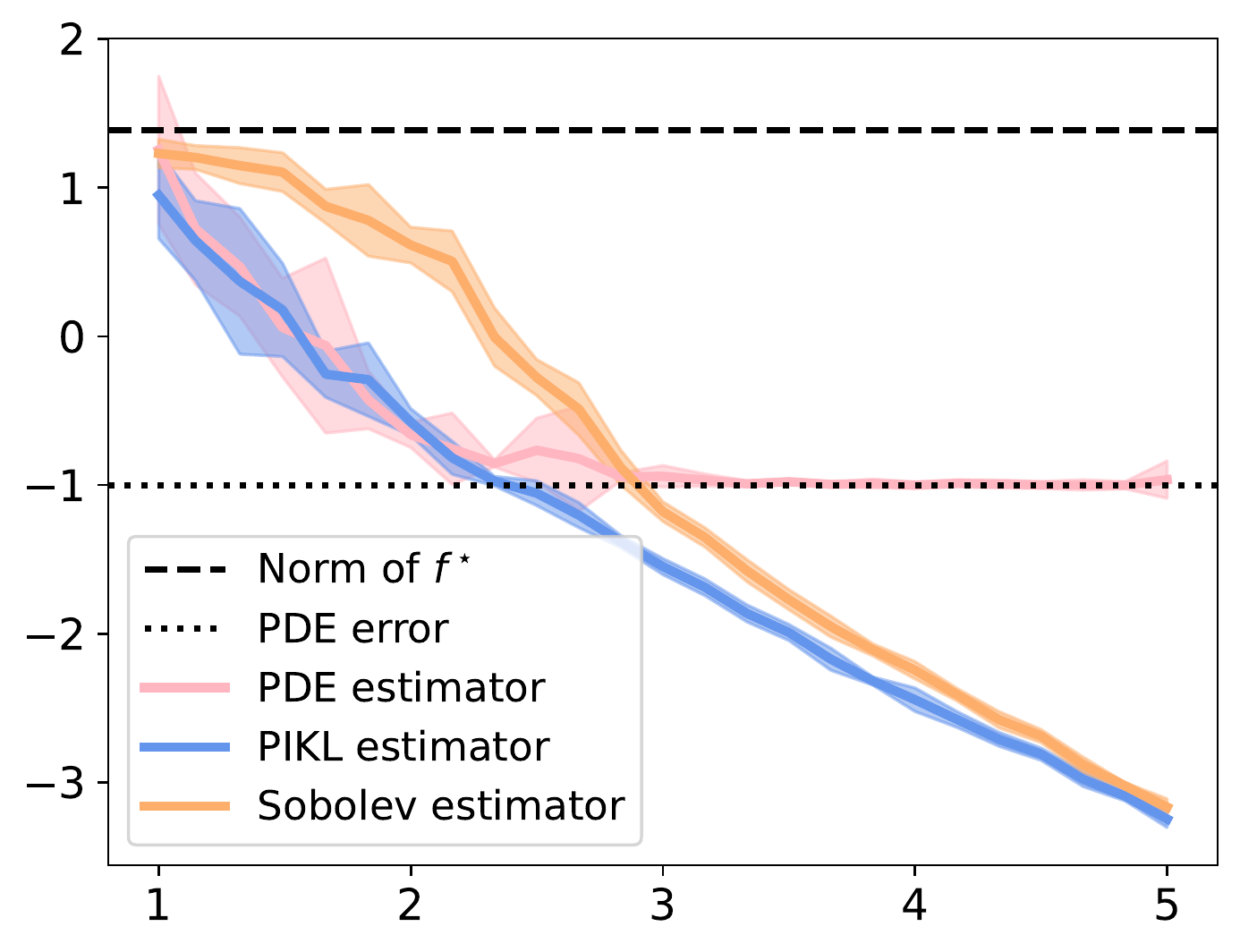}
    \caption{$L^2$-error  (mean $\pm$ std over 5 runs) of the PDE, PIKL, and Sobolev estimators  for imperfect modeling with the heat equation, as a function of $n$ in $\log_{10}-\log_{10}$ scale. The PDE error is the $L^2$-norm between $f^\star$ and the PDE solution that is closest to $f^\star$.}
    \label{fig:expe_heat}
\end{wrapfigure}
We perform an experiment where $\varepsilon \sim \mathcal{N}(0, \sigma^2)$ with $\sigma = 0.5$, and $f^\star(t,x) = \exp(-t)\cos(x) + 0.5 \sin(2x)$. This scenario is an example of imperfect modeling, since $\|\mathscr{D}(f^\star)\|_{L^2(\Omega)}^2 = \pi > 0$. However, the heat equation serves as a strong physical prior, since 
$\|\mathscr{D}(f^\star)\|_{L^2(\Omega)}^2/ \|f^\star\|_{L^2(\Omega)}^{2}\simeq 4 \times 10^{-3}$.
Figure~\ref{fig:expe_heat} illustrates the performance of the different estimators.

Clearly, the PDE estimator outperforms the Sobolev estimator when the data set is small ($n \leqslant 10^2$). 
As expected, the performance of the Sobolev estimator improves as the sample size increases ($n \geqslant  10^3$), but it remains consistently inferior to that of the PIKL.
When only a small amount of data is available, the PDE provides significant benefits, and the $L^2$-error decreases at the super-parametric rate of $n^{-2}$ for both the PIKL and the PDE estimators. However, in the context of imperfect modeling, the PDE estimator cannot overcome the PDE error, resulting in no further improvement beyond $n \geqslant  100$. In addition, when a large amount of data is available, the data become more reliable than the PDE. In this case, the errors for both the PIKL and the Sobolev estimators decrease at the Sobolev minimax rate of $n^{-2/3}$. Overall, the PIKL estimator successfully combines the strengths of both approaches, using the PDE when data is scarce and relying more on data when it becomes abundant.

\subsection{Measuring the impact of physics with the effective dimension}
\label{sec:eff_dim}
The important question of measuring the impact of the differential operator $\mathscr D$ on the convergence rate of the PIML estimator has not yet found a clear answer in the literature. In this subsection, we propose an approach to experimentally compare the PIKL convergence rate to the Sobolev minimax rate in $H^s(\Omega)$, which is $n^{-2s/(2s+d)}$ \citep[e.g.,][Theorem 2.1]{tsybakov2009introduction}. 
\paragraph{Theoretical backbone.}
According to \citet[Theorem 4.3]{doumeche2024physicsinformed}, if $X$ has a bounded density and the noise $\varepsilon$ is sub-Gamma with parameters $(\sigma, M)$, the $L^2$-error of both estimators \eqref{eq:estimator_sob_0} and \eqref{pb:PIML_on_Hm} satisfies
\begin{align}
     &\mathbb{E}\int_\Omega |\hat f_n-f^\star|^2 d{\mathbb P}_X \nonumber\\
        &\quad \leqslant  C_4 \log^2(n)\Big(\lambda_n \|f^\star\|_{H^s(\Omega)}^2 + \mu_n \|\mathscr{D}(f^\star)\|_{L^2(\Omega)}^2 + \frac{M^2}{n^2 \lambda_n} + \frac{\sigma^2\mathscr{N}(\lambda_n, \mu_n)}{n}\Big), \label{eq:err_l2}
\end{align} 
where $\mathbb P_X$ is the distribution of $X$. The quantity $\mathscr{N}(\lambda_n, \mu_n)$ on the right-hand side of~\eqref{eq:err_l2} is referred to as the effective dimension. 
Given the integral kernel operator $L_K$, defined by $L_K:f\in L^2(\mathbb P_X) \mapsto (x\mapsto\int_\Omega K(x,y) f(y)d\mathbb P_X(y))\in  L^2(\mathbb P_X)$, the effective dimension is the trace of the operator $(L_K+\hbox{Id})^{-1}L_K$ \citep[see, e.g.,][]{caponnetto2007optimal}.
Since $\lambda_n$ and $\mu_n$ can be freely chosen by the practitioner, the effective dimension $\mathscr{N}(\lambda_n, \mu_n)$ becomes a key consideration that help quantify the impact of the physics on the learning problem. 
Unfortunately, bounding $\mathscr{N}(\lambda_n, \mu_n)$ is not trivial. 
\citet{doumeche2024physicsinformed} have shown that, whenever $\frac{d\mathbb{P}_X}{dx} \leqslant \kappa$ with $\kappa \geq 1$,
\[
\mathscr N(\lambda_n, \mu_n) \leqslant  \sum_{\lambda\in \sigma(C\mathscr O_nC)} \frac{1}{1+(\kappa\lambda)^{-1}} \leqslant  \kappa \sum_{\lambda\in \sigma(C\mathscr O_nC)} \frac{1}{1+\lambda^{-1}} ,
\]
 where $\mathscr O_n$ is the operator $\mathscr O_n = \lim_{m\to \infty} M_m^{-1}$ (where the limit is taken in the sense of the operator norm---see Definition~\ref{defi:op_norm}) and $C$ is the operator $C(f) = 1_\Omega f$ (that is, $C(f)(x) = f(x)$ if $x\in \Omega$, and $C(f)(x) =0 $ otherwise).
Therefore, a natural idea to assess the effective dimension is to replace $C\mathscr O_nC$ by $C_mM_m^{-1}C_m$, where $C_m: H_m \to H_m$ is defined by \[\forall j,k\in \{-m, \hdots, m\}^d, \quad (C_m)_{j,k} = \frac{1}{(4L)^{d}}\int_\Omega e^{\frac{i \pi}{2L}\langle k-j, x\rangle}dx.\]
The following theorem shows that this is a sound strategy, in the sense that computing the effective dimension using the eigenvalues of $C_mM_m^{-1}C_m$ becomes increasingly accurate as $m$ grows.

\begin{thm}[Convergence of the effective dimension]
\label{thm:convergence_eff_dim}\hfill
    \begin{itemize}
        \item[(i)] One has
        \[\lim_{m\to\infty}\sum_{\lambda\in \sigma(C_mM_m^{-1}C_m)} \frac{1}{1+\lambda^{-1}} = \sum_{\lambda\in \sigma(C\mathscr O_nC)} \frac{1}{1+\lambda^{-1}}.\]
        \item[(ii)] Let $\sigma^\downarrow_k(C_mM_m^{-1}C_m)$ be the $k$-th highest eigenvalue of $C_mM_m^{-1}C_m$. The spectrum of the matrix $C_mM_m^{-1}C_m$ converges to the spectrum of $C\mathscr O_nC$ in the following sense:
        \[\forall k \in \mathbb N^\star,\quad \lim_{m\to\infty} \sigma^\downarrow_k(C_mM_m^{-1}C_m) = \sigma^\downarrow_k(C\mathscr O_nC).\]
    \end{itemize}
\end{thm}

The provided \texttt{Python} package\footnote{\url{https://github.com/NathanDoumeche/numerical_PIML_kernel}}$^{,}$\footnote{\url{https://pypi.org/project/pikernel}}   numerical approximations of the effective dimension in dimensions $d=1$ and $d=2$ for any linear operator $\mathscr{D}$ with constant coefficients, when $\Omega$ is either a cube or a Euclidean ball. The code is available is designed to run on both CPU and GPU. The convergence of the effective dimension as $m$ grows is studied in greater detail in Appendix \ref{sec:eff_dim_m}.

\begin{figure}
    \centering
    \includegraphics[scale=0.45]{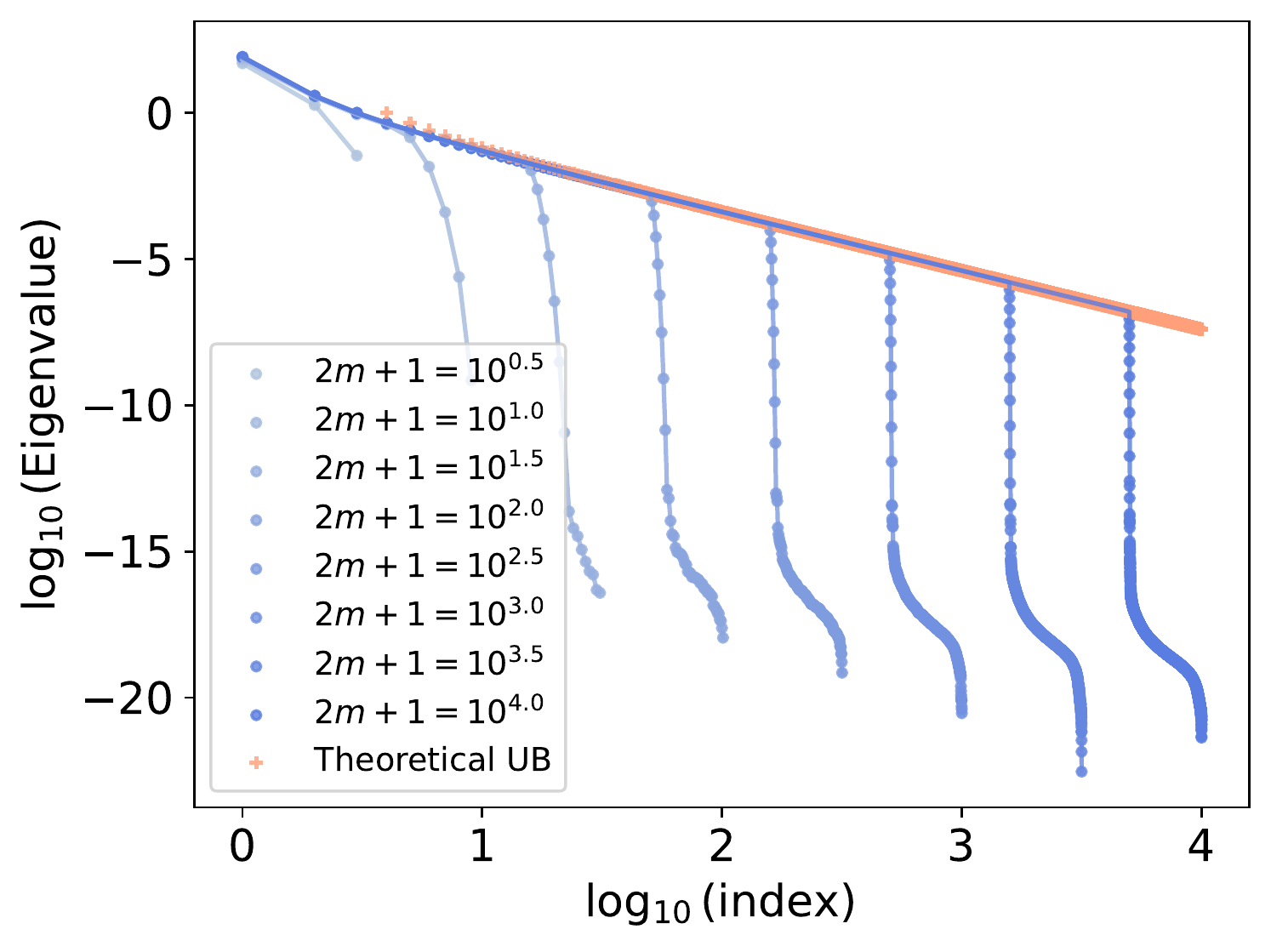}
    \includegraphics[scale=0.45]{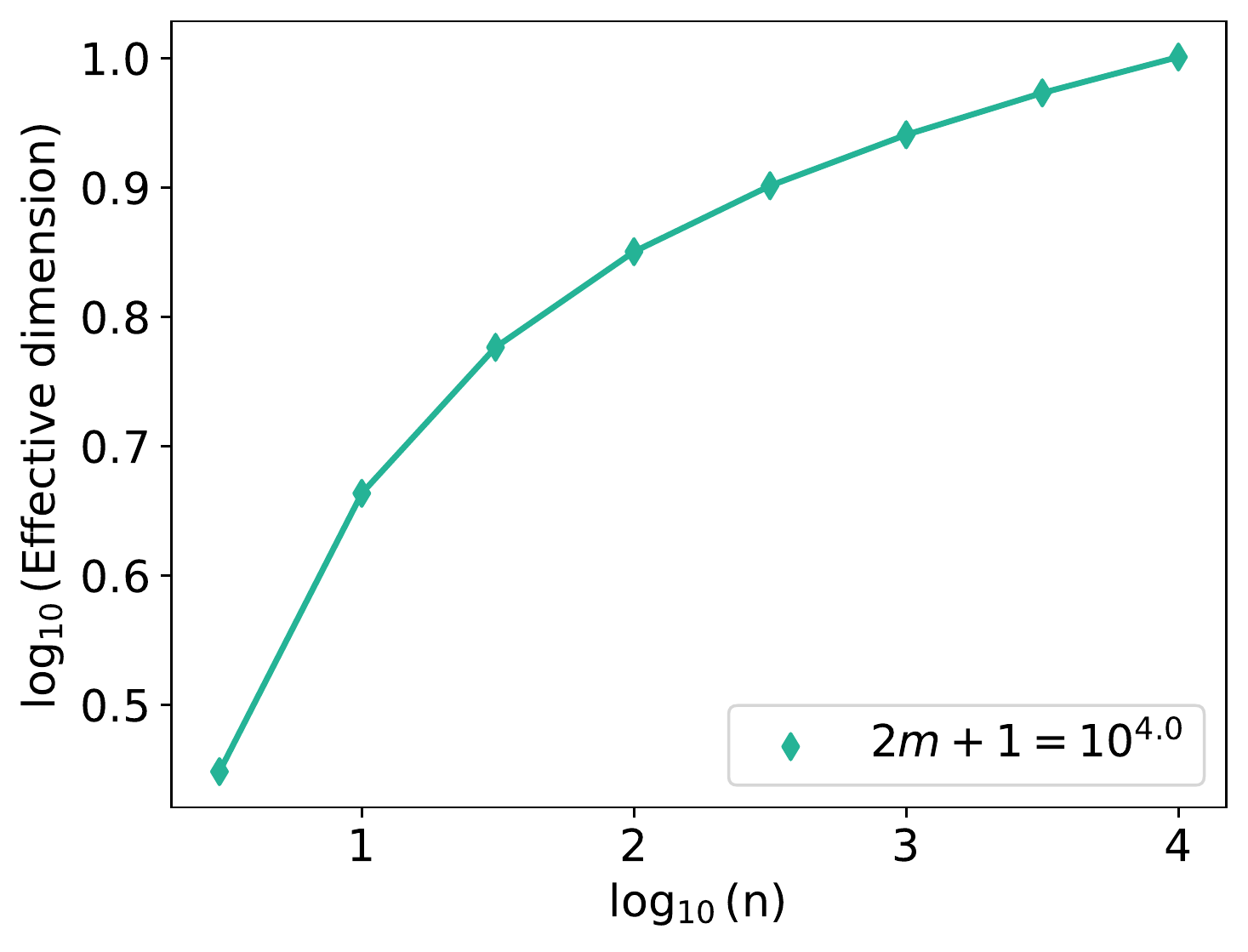}
    \caption{The case of $\mathscr D = \frac{d}{dx}$. \textbf{Left:} Spectrum of $C_mM_m^{-1}C_m$. \textbf{Right:} Estimation of the effective dimension $n \mapsto \mathscr N(\frac{\log(n)}{n}, \frac{1}{\log(n)})$. 
    }
    \label{fig:1d_spectrum}
\end{figure}

\paragraph{Comparison to the closed-form case.} We start by assessing the quality of the approximation encapsulated in Theorem \ref{thm:convergence_eff_dim} in a scenario where the eigenvalues can be theoretically bounded. 
When $d=1$, $s=1$, $\mathscr D = \frac{d}{dx}$, and $\Omega = [-\pi,\pi]$, one has \citep[Proposition 5.2]{doumeche2024physicsinformed} 
\[\frac{4}{(\lambda_n+\mu_n)(k+4)^2} \leqslant \sigma^\downarrow_k(C\mathscr O_nC)\leqslant \frac{4}{(\lambda_n+\mu_n)(k-2)^2}.\] 
This shows that $\log \sigma^\downarrow_k(C\mathscr O_nC) \sim_{k\to \infty} - 2\log(k)$.
Figure \ref{fig:1d_spectrum} (Left) represents the eigenvalues of $C_mM_m^{-1}C_m$ in decreasing order, for increasing values of $m$, with $\lambda_n = 0.01$ and $\mu_n = 1$. 
For any fixed $m$, two distinct regimes can be clearly distinguished: initially, the eigenvalues decrease linearly on a $\log-\log$ scale and align with the theoretical values of $- 2\log(k)$.
Afterward, the eigenvalues suddenly drop to zero. As $m$ increases, the spectrum progressively approaches the theoretical bound.

In Appendix~\ref{sec:eff_dim_m}, we show that $m = 10^{2}$ Fourier modes are sufficient to accurately approximate the effective dimension when $n \leqslant 10^4$. It is evident from Figure \ref{fig:1d_spectrum} (Right) that the effective dimension exhibits a sub-linear behavior in the $\log-\log$ scale, experimentally confirming the findings of \citet{doumeche2024physicsinformed}, which show that $\mathscr{N}(\frac{\log(n)}{n}, \frac{1}{\log(n)}) = o_{n\to\infty}(n^\gamma)$ for all $\gamma > 0$. 
So, plugging this into \eqref{eq:err_l2} with $\lambda_n = n^{-1}\log(n)$ and $\mu_n = \log(n)^{-1}$ leads to
\begin{equation*}
        \mathbb{E}\int_{[-L,L]} |\hat f_n-f^\star|^2 d{\mathbb P}_X = (\|f^\star\|_{H^1(\Omega)}^2 + \sigma^2 + M^2)O_n \big( n^{-1} \log^3(n)\big) 
\end{equation*}
when $\mathscr D(f^\star) = 0$, i.e., when the modeling is perfect.
The Sobolev minimax rate on $H^1(\Omega)$ is $n^{-2/3}$, whereas the experimental bound in this context gives a rate of $n^{-1}$. This indicates that when the target $f^{\star}$ satisfies the underlying PDE, the gain in terms of speed from incorporating the physics into the learning problem is  $n^{-1/3}$.

\begin{figure}
    \centering
    \includegraphics[scale=0.45]{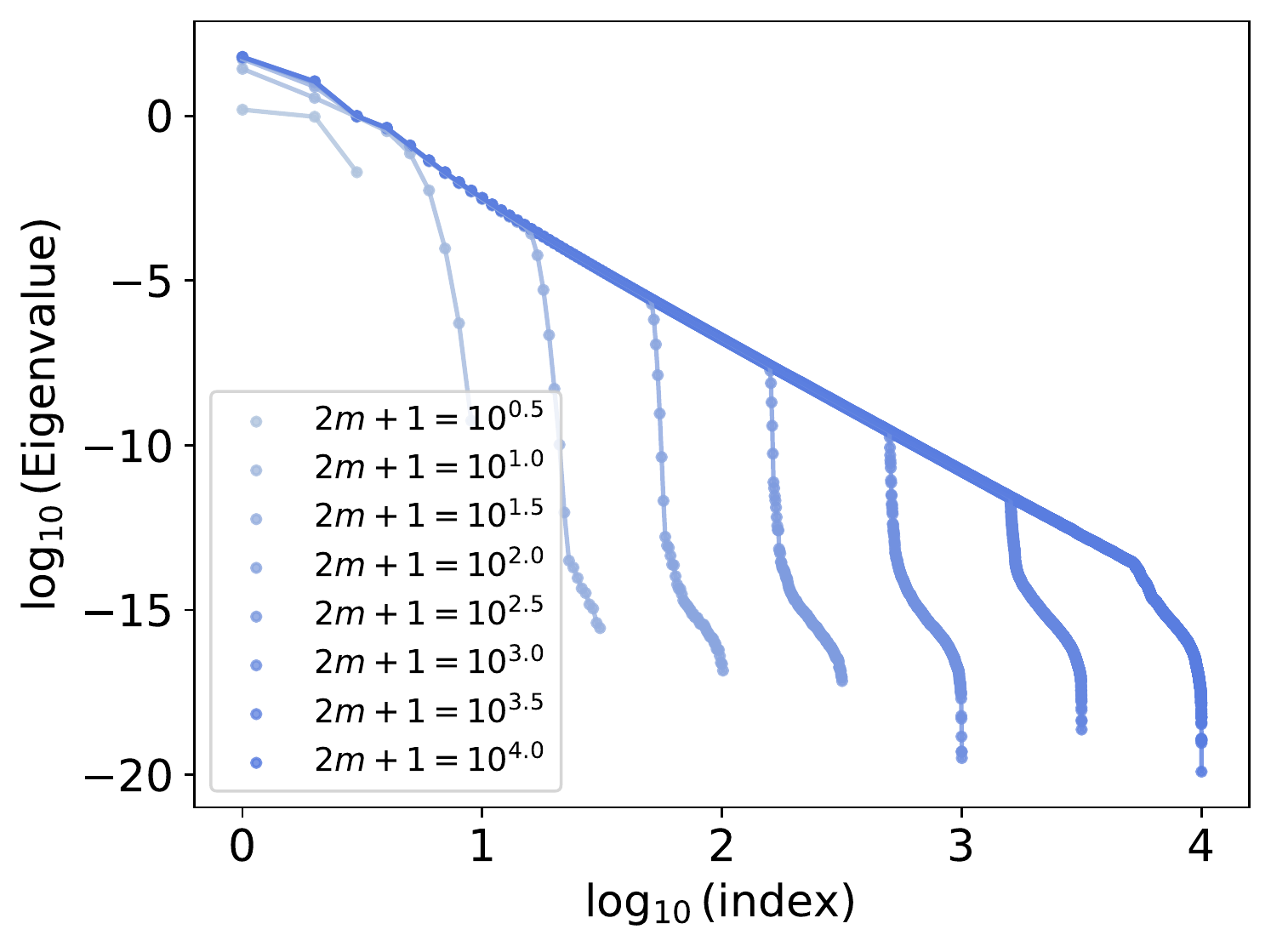}
    \includegraphics[scale=0.45]{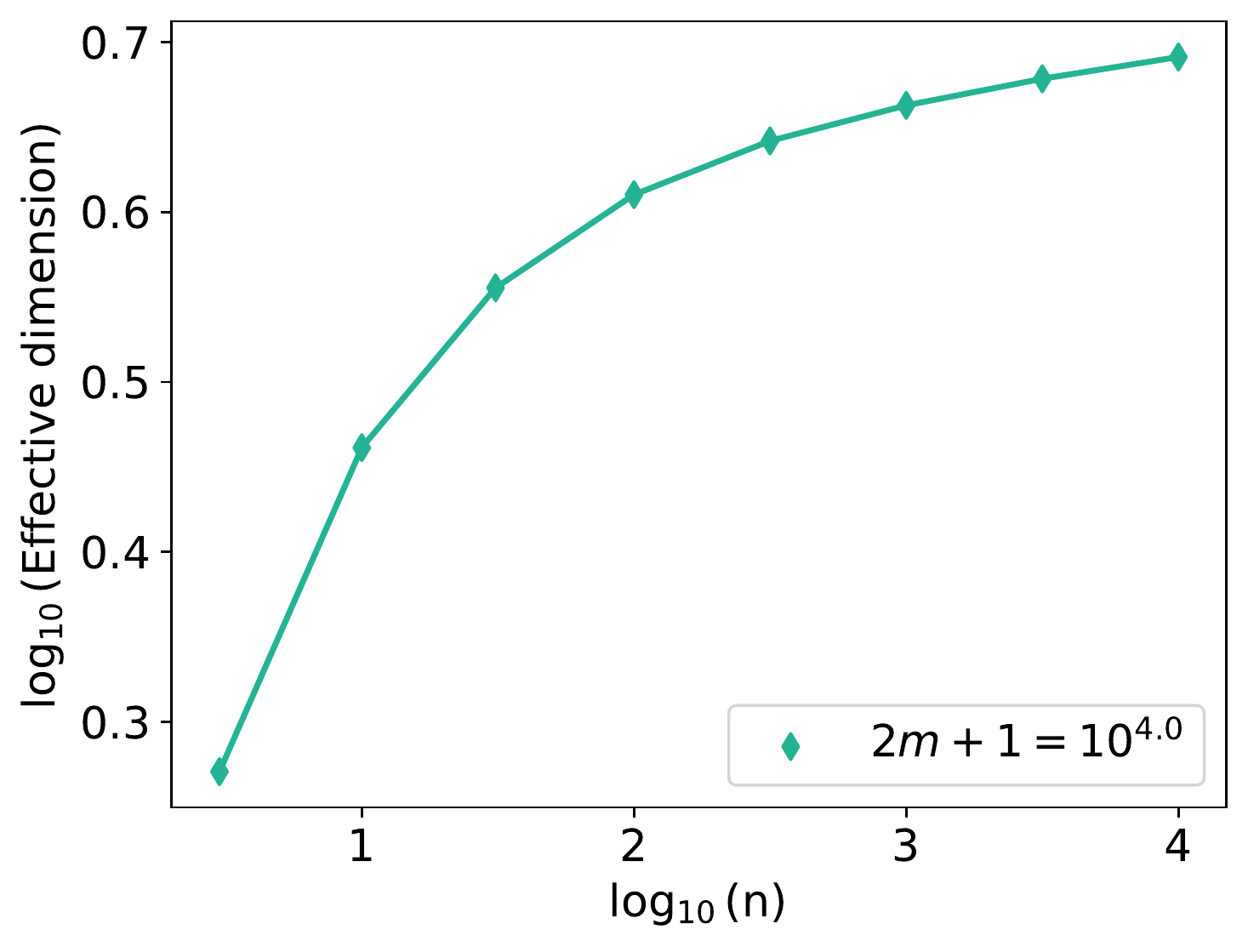}
    \caption{Harmonic oscillator. \textbf{Left:} Spectrum of $C_mM_m^{-1}C_m$. \textbf{Right:} Estimation of the effective dimension $n \mapsto \mathscr N(\frac{\log(n)}{n}, \frac{1}{\log(n)})$.}
    \label{fig:oscillator_spectrum}
\end{figure}

\paragraph{Harmonic oscillator equation.} Here, we follow up on the example of Subsection~\ref{sec:hybrid_mod}, as presented in Figures~\ref{fig:ressort_least_square} and \ref{fig:perf_ressort_least_square}. 
Thus, we set $d=1$, $s=2$, $\mathscr D(u) = \frac{d^2}{dx^2}u+\frac{d}{dx}u+u$, and $\Omega=[-\pi,\pi]$. Recall that in this perfect modeling experiment, we observed a parametric convergence rate of $n^{-1}$, which is not surprising since the regression problem essentially involves learning the two parameters $a_1$ and $a_2$. 
Figure \ref{fig:oscillator_spectrum} (Left) shows the eigenvalues of $C_m M_M^{-1} C_m$, while Figure \ref{fig:oscillator_spectrum} (Right) shows the effective dimension as a function of $n$. Similarly to the previous closed-form case, we observe that $\mathscr{N}(\frac{\log(n)}{n}, \frac{1}{\log(n)}) = o_{n\to\infty}(n^\gamma)$ for all $\gamma > 0$. 
The same argument as in the paragraph above shows that this results in a parametric convergence rate, provided $\mathscr D(f^\star) = 0$. 

\paragraph{Heat equation on the disk.} Let us now consider the one-dimensional heat equation $\mathscr D = \frac{\partial}{\partial x}-\frac{\partial^2}{\partial y^2}$, with $d=2$, $s=2$, and the disk  $\Omega=\{x\in \mathbb R^2, \|x\|_2\leq\pi\}$. Since the heat equation is known to have $C^\infty$ solutions with bounded energy \citep[see, e.g.,][Chapter~2.3, Theorem~8]{evans2010partial}, we expect the convergence rate to match that of $H^{\infty}(\Omega)$, which corresponds to the parametric rate of $n^{-1}$.
\begin{figure}
    \centering
    \includegraphics[scale=0.45]{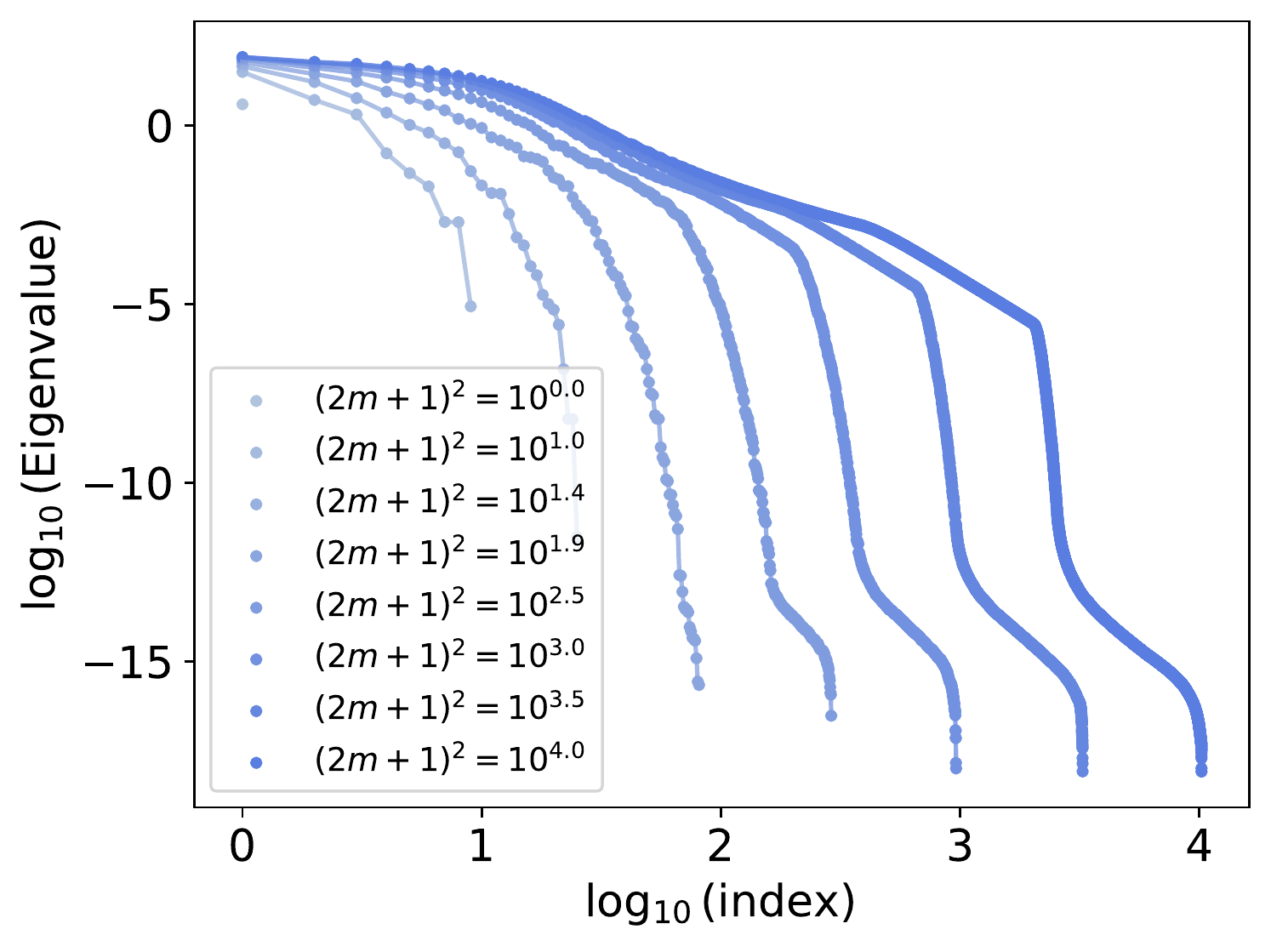}
    \includegraphics[scale=0.45]{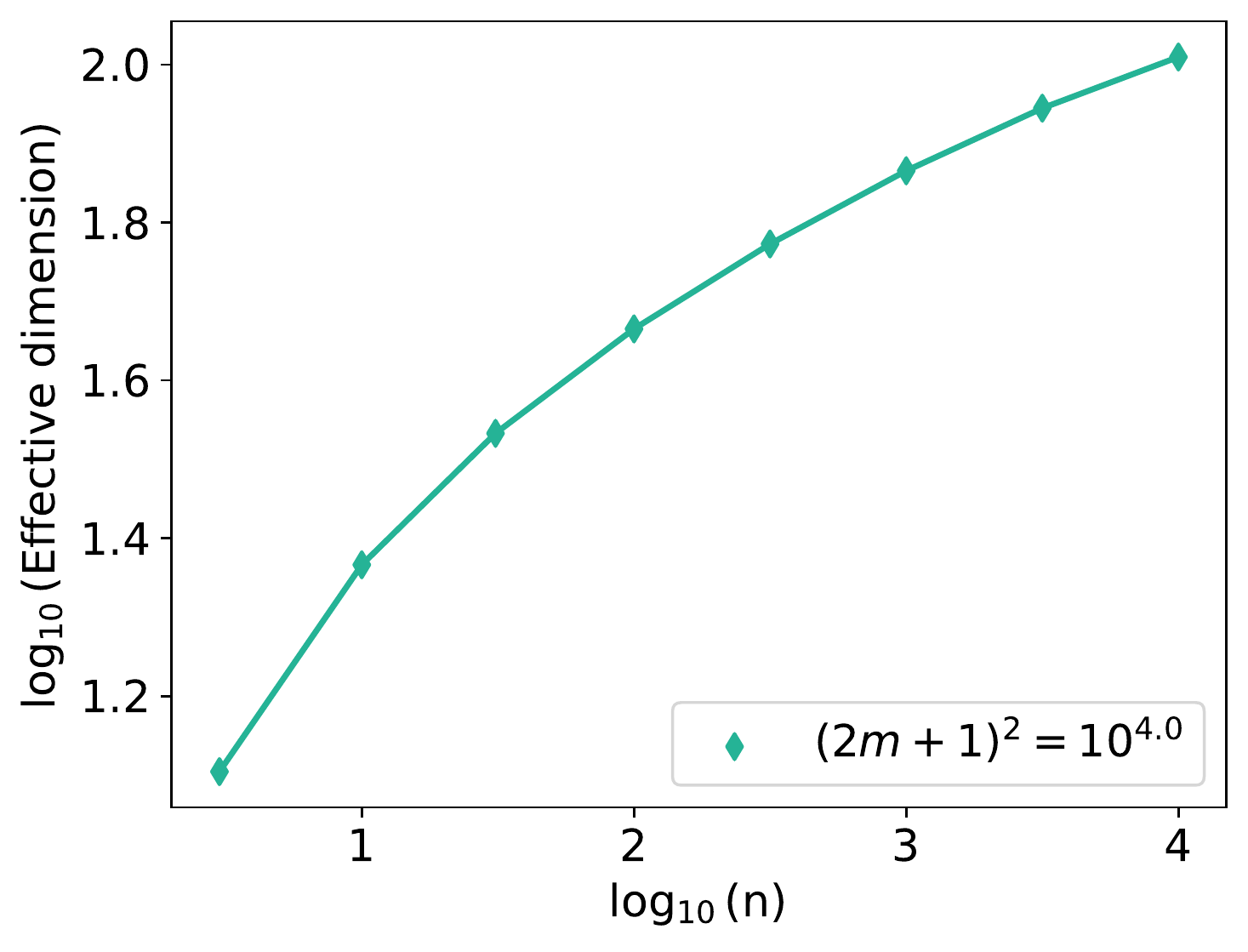}
    \caption{Heat equation. \textbf{Left:} Spectrum of $C_mM_m^{-1}C_m$. \textbf{Right:} Estimation of the effective dimension $n \mapsto \mathscr N(\frac{\log(n)}{n}, \frac{1}{\log(n)})$.}
    \label{fig:heat_spectrum}
\end{figure}
Once again, we observe $\mathscr{N}(\frac{\log(n)}{n}, \frac{1}{\log(n)}) = o_{n\to\infty}(n^\gamma)$ for all $\gamma > 0$, and thus an improvement over the $n^{-2/3}$-Sobolev minimax rate on $H^2(\Omega)$ when $\mathscr D(f^\star) = 0$.

\paragraph{Quantifying the impact of physics.} The three examples above show how incorporating physics can enhance the learning process by reducing the effective dimension, leading to a faster convergence rate. In all cases, the rate becomes parametric due to the PDE, achieving the fastest possible speed, as predicted by the central limit theorem. Our package can be directly applied to any linear PDE with constant coefficients to compute the effective convergence rate given a scaling of $\lambda_n$ and~$\mu_n$. By identifying  the optimal convergence rate, this approach can assist in determining the best parameters $\lambda_n$ and $\mu_n$ for use in other PIML techniques, such as PINNs.

\section{PDE solving: Mitigating the difficulties of PINNs with PIKL}
\label{sec:PDE_solving}
It turns out that our PIKL algorithm can be effectively used as a PDE solver. In this scenario, there is no noise (i.e., $\varepsilon = 0$), no modeling error (i.e., $\mathscr{D}(f^\star) = 0$), and the data consist of samples of boundary and initial conditions, as is typical for PINNs. Assume for example that the objective is to solve the Laplacian equation $\Delta(f^\star) = 0$ on a domain $\Omega \subseteq [-1,1]^2$ with the Dirichlet boundary condition $f^\star|_{\partial \Omega} = g$, where $g$ is a known function. Then this problem can be addressed by implementing the PIKL estimator, which minimizes the risk $\bar R_n(f) = \frac{1}{n}\sum_{i=1}^n |f(X_i) - Y_i|^2 + \lambda_n \|f\|^2_{H^2_{\mathrm{per}}([-1,1]^2)} + \mu_n \|\Delta(f)\|^2_{L^2(\Omega)}$, where the $X_i$ are uniformly sampled on $\partial \Omega$ and $Y_i = g(X_i)$.
Of course, this example focuses on Dirichlet boundary conditions, but PIKL is a highly flexible framework that can incorporate a wide variety of boundary conditions, such as periodic and Neumann boundary conditions, as the next two examples will illustrate.

\paragraph{Comparison with PINNs for the convection equation.} 
To begin, we compare the performance of our PIKL algorithm with the PINN approach developed by \citet{krishnapriyan2021characterizing} for solving the one-dimensional convection equation $\mathscr{D}(f) = \partial_{t} f + \beta \partial_{x} f$ on the domain $\Omega = [0,1]\times [0,2\pi]$. The problem is subject to the following periodic boundary conditions:
\[
\left\{
\begin{array}{l}
\forall x \in [0,2\pi], \quad f(0, x) = \sin(x),\\
\forall t \in [0,1], \quad f(t, 0) = f(t, 2\pi) = 0.
\end{array}
\right.
\] 

The solution of this PDE is given by $f^\star(t,x) = \sin(x - \beta t)$. \citet{krishnapriyan2021characterizing} show that for high values of $\beta$, PINNs struggle to solve the PDE effectively. To address this challenge, we train our PIML kernel method using $n = 100$ data points and $1681$ Fourier modes (i.e., $m = 20$). The training data set $(X_i, Y_i)_{1 \leqslant i \leqslant n}$ is constructed such that $X_i = (0, 2\pi U_i)$ and $Y_i = \sin(U_i)$, where $(U_i)_{1 \leqslant i \leqslant n}$ are i.i.d.~uniform random variables. To enforce the periodic boundary conditions,  we center $\Omega$ at $\tilde \Omega = \Omega-(0.5, \pi)$, extend it to $[-1, 1]\times[-\pi, \pi]$, and consider $\tilde H_m = \mathrm{Span}((t,x)\mapsto e^{i (\frac{\pi}{2}k_1 t + k_2 x)})_{\|k\|_\infty \leq m}$. Noting that for all $(j_1, k_1), (j_2, k_2)\in \mathbb Z^2$,
\[
\int_{[-1, 1]\times[-\pi, \pi]}e^{i (\frac{\pi}{2}(k_1-j_1) t + (k_2 - j_2) x)}dx = \frac{\sin(\pi (k_1-j_1)/2)}{\pi } \delta_{k_2,j_2},
\]
we let the matrix $(M_m)_{j,k}$ be as follows:
\begin{equation*}
    (M_m)_{j,k} = \lambda_n \Big(1+\frac{\|k\|_2^2}{(2L)^2}\Big)\delta_{j,k}+\mu_n \frac{P(j)\bar P(k)}{(4L)^2}\frac{\sin(\pi (k_1-j_1)/2)}{\pi } \delta_{k_2,j_2},
\end{equation*}
where $P$ is the polynomial associated with the operator $\mathscr D$.  Notice that, although $f^\star$ is a sinusoidal function, the frequency vector of $f^\star$ is $(-\beta, 1)$, which does not belong to $\frac{\pi}{2}\mathbb Z \oplus \mathbb Z$. As a result, $f^\star$ does not lie in $\tilde H_m$ for any $m$.

Table~\ref{table_perf_PDE_convection} compares the performance of various PIML methods using a sample of $n = 100$ initial condition points. The performance of an estimator $\hat{f}_n$ on a test set ($\mathrm{Test}$) is evaluated based on the $L^2$ relative error $(\sum_{x \in \mathrm{Test}} \|\hat f_n(x)-f^\star(x)\|_2^2/\sum_{y \in \mathrm{Test}} \|f^\star(y)\|_2^2)^{1/2}$.
Standard deviations are computed across 10 trials. The results show that the PIML kernel estimator clearly outperforms PINNs in terms of accuracy. 

\begin{table}[ht]
\centering
\caption{\textbf{$L^2$ relative error of the kernel method in solving the advection equation.} } 

\begin{tabular*}{\textwidth}{@{\extracolsep{\fill}}lccc}
  \toprule
  & Vanilla PINNs$^{\diamond}$ & Curriculum-trained PINNs$^{\diamond}$ & PIKL estimator\\
  \midrule
  \textit{$\beta = 20$ } & $7.50 \times 10^{-1}$&   $9.84 \times 10^{-3}$& $\mathbf{(1.56{\scriptstyle \pm 3.46}) \times 10^{-8}}$\\
  \textit{$\beta = 30$} & $8.97 \times 10^{-1}$ & $2.02 \times 10^{-2}$ & $\mathbf{(0.91{\scriptstyle \pm 2.20})\times 10^{-7}}$\\
  \textit{$\beta = 40$} & $9.61 \times 10^{-1}$ & $5.33 \times 10^{-2}$ & $\mathbf{(7.31{\scriptstyle \pm 6.44}) \times 10^{-9}}$\\
   \bottomrule
\end{tabular*}

\flushleft 
\textit{$^\diamond$ \citet[][Table 1]{krishnapriyan2021characterizing}}

\label{table_perf_PDE_convection}
\end{table}

\paragraph{Comparison with PINNs for the 1d-wave equation.} The performance of the PIKL algorithm is compared to the PINN methodology of \citet[Section 7.3]{wang2022when} for solving the one-dimensional wave equation $\mathscr{D}(f) = \partial^2_{t,t} f - 4 \partial_{x,x}^2 f$ on the square domain $[0,1]^2$, with the following boundary conditions: 
\[
\left\{
\begin{array}{l}
\forall x \in [0,1], \quad f(0, x) = \sin(\pi x) + \sin(4\pi x)/2,\\
\forall x \in [0,1], \quad \partial_tf(0, x) = 0, \\
\forall t \in [0,1], \quad f(t, 0) = f(t, 1) = 0.
\end{array} 
\right.
\]

\begin{figure}
    \centering
    \includegraphics[width=0.3\textwidth]{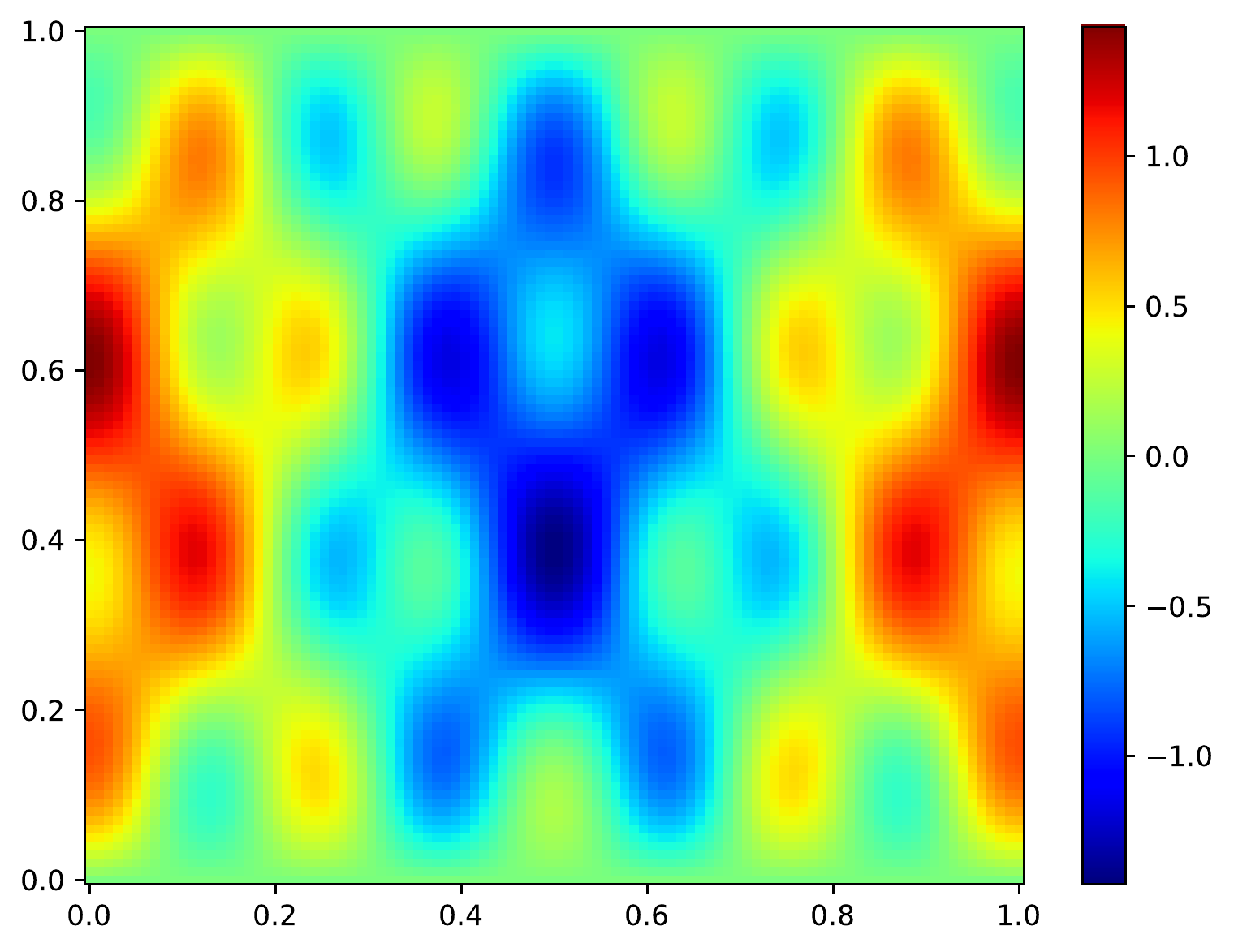}
    \includegraphics[width=0.3\textwidth]{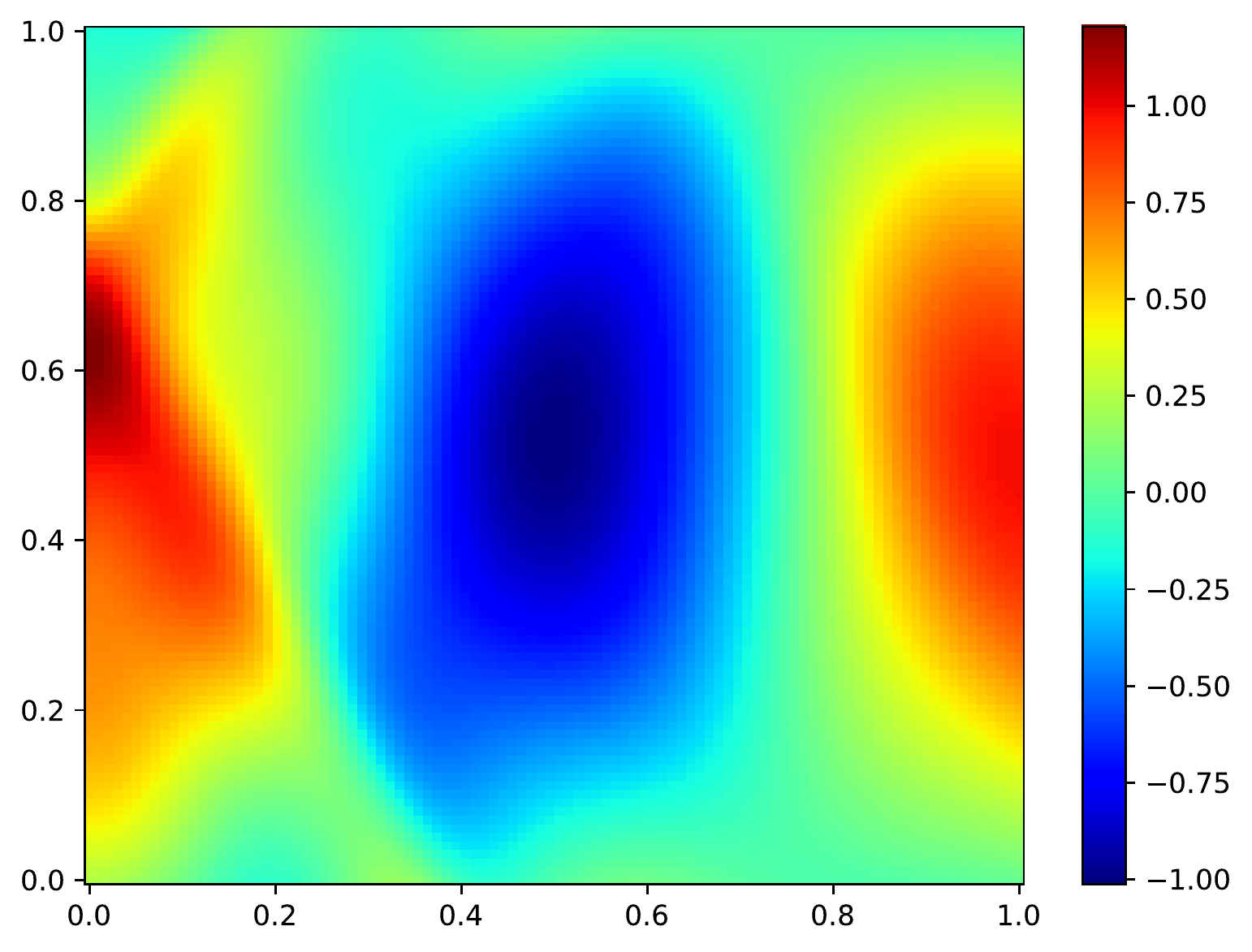}
    \includegraphics[width=0.3\textwidth, trim={15.6cm 0.65cm 0 0},clip]{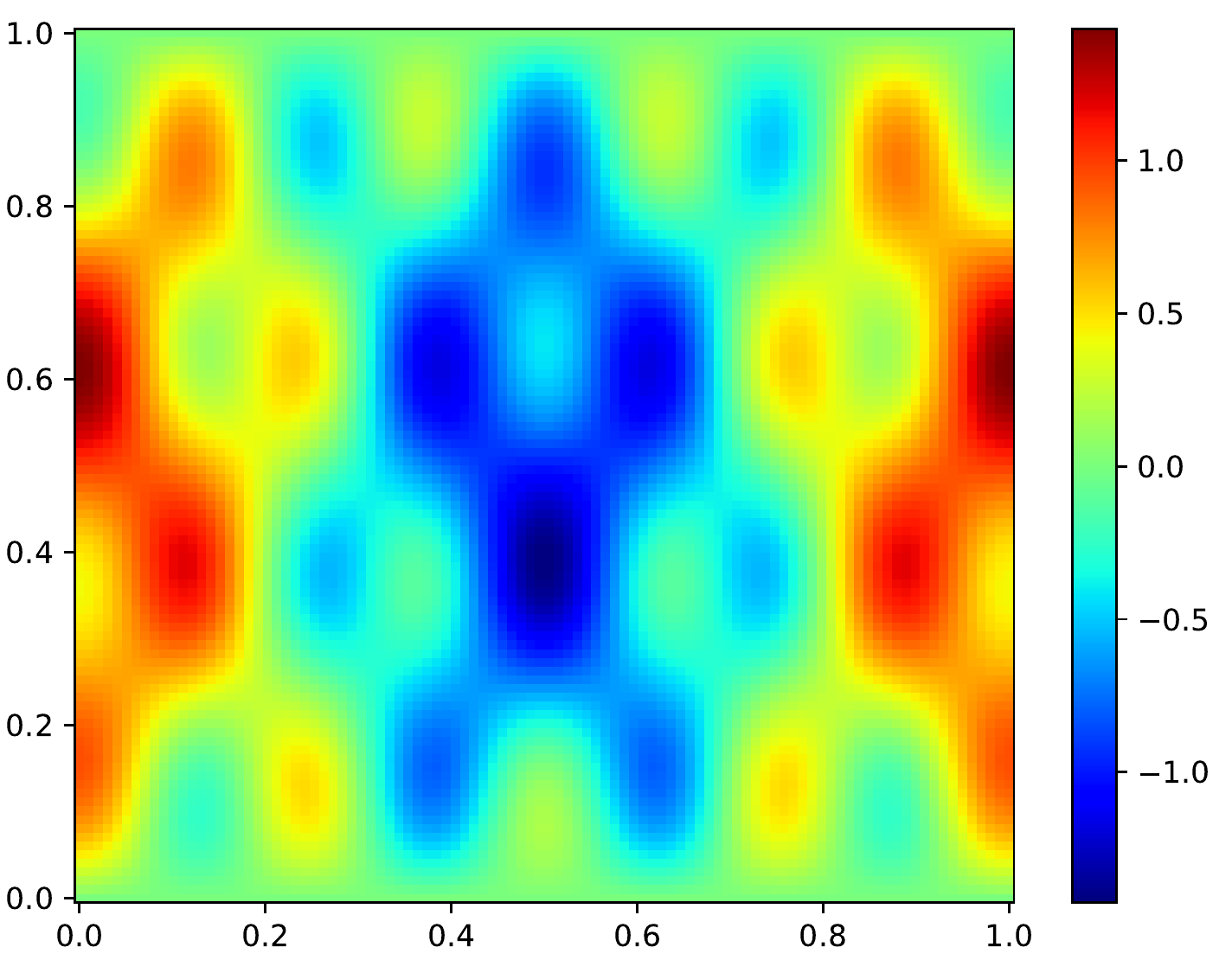}
    \caption{\textbf{Left:} ground truth solution $f^\star$ to the wave equation \citep[taken from][Figure~6]{wang2022when}. \textbf{Middle:} Vanilla PINNs from \citet{wang2022when}. \textbf{Right:} PIKL estimator. }
    \label{fig:wave_comp}
\end{figure}

The solution of the PDE is $f^\star(t,x) = \sin(\pi x)\cos(2\pi t) + \sin(4\pi x)\cos(8\pi t)/2$. This solution serves as an interesting benchmark since $f^\star$ exhibits significant variations, with $\|\partial_t f^\star\|_2^2/ \|f^\star \|_2^2 = 16 \pi^2$ (Figure~\ref{fig:wave_comp}, Left). Meanwhile, PINNs are known to have a spectral bias toward low frequencies \citep[e.g.,][]{Deshpande2022investigations, wang2022physics}. The optimization of the PINNs in \citet{wang2022physics} is carried out using stochastic gradient descent with $80,000$ steps, each drawing $300$ points at random, resulting in a sample size of $n = 2.4 \times 10^6$. The architecture of the PINNs these authors employ is a dense neural network with $\tanh$ activation functions and layers of sizes $(2, 500, 500, 500, 1)$, resulting in $m = (2 \times 500 + 500) + 2 \times (500 \times 500 + 500) + (500 \times 1 + 1) = 503,001$ parameters.
The training time for Vanilla PINNs is 7 minutes on an Nvidia L4 GPU (24 GB of RAM, 30.3 teraFLOPs for Float32). We obtain an $L^2$ relative error of $4.21\times 10^{-1}$, which is consistent with the results of \citet{wang2022when}, who report a $L^2$ relative error of $4.52\times 10^{-1}$. Figure~\ref{fig:wave_comp} (Middle) shows the Vanilla PINNs. 

We train our PIKL method using $n = 10^5$ data points and $1681$ Fourier modes (i.e., $m = 20$). Let $(U_i)_{1 \leqslant i \leqslant n}$ be i.i.d.~random variables uniformly distributed on 
$[0,1]$. The training data set $(X_i, Y_i)_{1 \leqslant i \leqslant n}$ is constructed such that
\begin{itemize}
    \item if $\ 1\leqslant i\leqslant \lfloor n/4\rfloor$, then $X_i = (0, U_i)$  and $Y_i = \sin(\pi U_i) + \sin(4\pi U_i)/2$,
    \item if $ \lfloor n/4\rfloor+1\leqslant i\leqslant 2\lfloor n/4\rfloor$, then $X_i = (U_i, 0)$ and $Y_i = 0$,
    \item if $ 2\lfloor n/4\rfloor+1\leqslant i\leqslant 3\lfloor n/4\rfloor$, then $X_i = (U_i, 1)$ and $Y_i = 0$,
    \item if $ 3\lfloor n/4\rfloor+1\leqslant i\leqslant  n$, then  $X_i = (1/n, U_i)$ and 
    \begin{align*}
        Y_i &= f(0, U_i) + \frac{1}{2n^2}\partial^2_{t,t}f(0, U_i) = f(0, U_i) + \frac{2}{n^2}\partial^2_{x,x}f(0, U_i) \\
        &= \Big(1-\frac{2\pi^2}{n^2}\Big)\sin(\pi U_i) + \Big(\frac{1}{2}-\frac{16\pi^2}{n^2}\Big)\sin(4\pi U_i).
    \end{align*}
\end{itemize}
The final requirement enforces the initial condition $\partial_t f = 0$ in a manner similar to that of a second-order numerical scheme.

Table~\ref{table_perf_PDE} compares the performance of the PINN approach from \citet{wang2022when} with the PIKL estimator. Across $10$ trials, the PIKL method achieves an $L^2$ relative error of $(8.70 \pm 0.08) \times 10^{-4}$, which is $50\%$ better than the performance of the PINNs. This demonstrates that the kernel approach is more accurate, requiring fewer data points and parameters than the PINNs. The training time for the PIKL estimator is 6 seconds on an Nvidia L4 GPU. Thus, the PIKL estimator can be computed 70 times faster than the Vanilla PINNs. Figure~\ref{fig:wave_comp} (Right) shows the PIKL estimator. 
Note that in this case, the solution $f^\star$ can be represented by a sum of complex exponential functions ($f^\star\in H_{16}$), which could have biased the result in favor of the PIKL estimator by canceling its approximation error. However, the results remain unchanged when altering the frequencies in $H_m$ (e.g., taking $L=0.55$ in \eqref{eq:Mm} instead of $L=0.5$ yields an $L^2$ relative error of $(9.6\pm 0.3)\times 10^{-4} $).
\begin{table}[ht]
\centering
\caption{\textbf{Performance of PINN/PIKL methods for solving the wave equation on $\Omega=[0,1]^2$}} 

\begin{tabular*}{\textwidth}{@{\extracolsep{\fill}}lccc}
  \toprule
  & Vanilla PINNs$^{\diamond}$ & NTK-optimized PINNs$^{\diamond}$ & PIKL estimator\\
  \midrule
  \textit{$L^2$ relative error} & $4.52 \times 10^{-1}$&   $1.73 \times 10^{-3}$& $\mathbf{(8.70 {\scriptstyle \pm 0.08})  \times 10^{-4}}$\\
  \textit{Training data (n)} & $2.4 \times 10^{6}$ & $2.4 \times 10^{6}$ & $\mathbf{10^5}$\\
  \textit{Number of parameters} & $5.03 \times 10^5$ & $5.03 \times 10^5$ & $\mathbf{1.68 \times 10^3}$\\
   \bottomrule
\end{tabular*}

\flushleft 
\textit{$^\diamond$ \citet[][Figure 6]{wang2022when}}

\label{table_perf_PDE}
\end{table}

\section{PDE solving with noisy boundary conditions}
\label{sec:noisy}

\subsection{Wave equation in dimension 2}

\paragraph{Comparison with traditional PDE solvers.} PIML is a promising framework for solving PDEs, particularly due to its adaptability to domains $\Omega$ with complex geometries, where most traditional PDE solvers tend to be highly domain-dependent. However, its comparative performance against traditional PDE solvers remains unclear in scenarios where both approaches can be easily implemented. The meta-analysis by \citet{mcgreivy2024weakbaselinesreportingbiases} indicates that, in some cases, PINNs may be faster than traditional PDE solvers, although they are often less accurate. In our study, solving the wave equation on a simple square domain represents a setting where traditional numerical methods are straightforward to implement and are known to perform well. Table~\ref{table_perf_PDE_solvers} summarizes the performance of classical techniques, including the explicit Euler, Runge-Kutta 4 (RK4), and Crank-Nicolson (CN) schemes (see Appendix~\ref{sec:numerical} for a brief presentation of these methods). These methods clearly outperform both PINNs and the PIKL algorithm, even with fewer data points.
\begin{table}[ht]
\centering
\caption{\textbf{Performance of traditional PDE solvers for the wave equation on $\Omega=[0,1]^2$}.} 

\begin{tabular*}{\textwidth}{@{\extracolsep{\fill}}lccc}
  \toprule
  & Euler explicit  & RK4 & CN\\
  \midrule
  \textit{$L^2$ relative error} & $\mathbf{3.8 \times 10^{-6}}$&   $6.8 \times 10^{-6}$& $ 5.6 \times 10^{-3}$\\
  \textit{Training data (n)} & $10^{4}$ & $10^{4}$ & $10^4$\\
   \bottomrule
\end{tabular*}
\label{table_perf_PDE_solvers}
\end{table}

\paragraph{Noisy boundary conditions.} However, a more relevant setting for comparing the performance of these methods arises when noise is introduced into the boundary conditions. This situation is common, for instance, when the initial condition of the wave is measured by a noisy sensor. Such a setting aligns with hybrid modeling, where $\varepsilon \neq 0$, but there is no modeling error (i.e., $\mathscr{D}(f^\star) = 0$). Table~\ref{table_perf_PDE_solvers_noisy} compares the performance of all methods with Gaussian noise of variance of $10^{-2}$. In this case, the PIKL estimator outperforms all other approaches.
\begin{table}[ht]
\centering
\caption{\textbf{Performance for the wave equation with noisy boundary conditions}.} 
\begin{tabular*}{\textwidth}{@{\extracolsep{\fill}}lcccccc}
  \toprule
  & PINNs & Euler explicit & RK4 & CN  & PIKL estimator\\
  \midrule
  \textit{$L^2$ relative error} & $4.61 \times 10^{-1}$& $1.25 \times 10^{-1}$&   $6.05\times 10^{-2}$ & $2.01  \times 10^{-2}$&    $\mathbf{1.87 \times 10^{-2}}$\\
  \textit{Training data (n)} & $2.4\times10^{6}$ & $4\times10^{4}$ & $4\times 10^{4}$ & $4\times 10^{4}$& $4\times 10^4$\\
   \bottomrule
\end{tabular*}
\label{table_perf_PDE_solvers_noisy}
\end{table}\\
Such PDEs with noisy boundary conditions are special cases of the hybrid modeling framework, where the data is located on the boundary of the domain. This situation arises, for example, in \citet{cai2021physics} which models the temperature in the core of a nuclear reactor.
\subsection{Heat equation in dimension 4}
\begin{figure}
    \centering
    \includegraphics[width=0.5\linewidth]{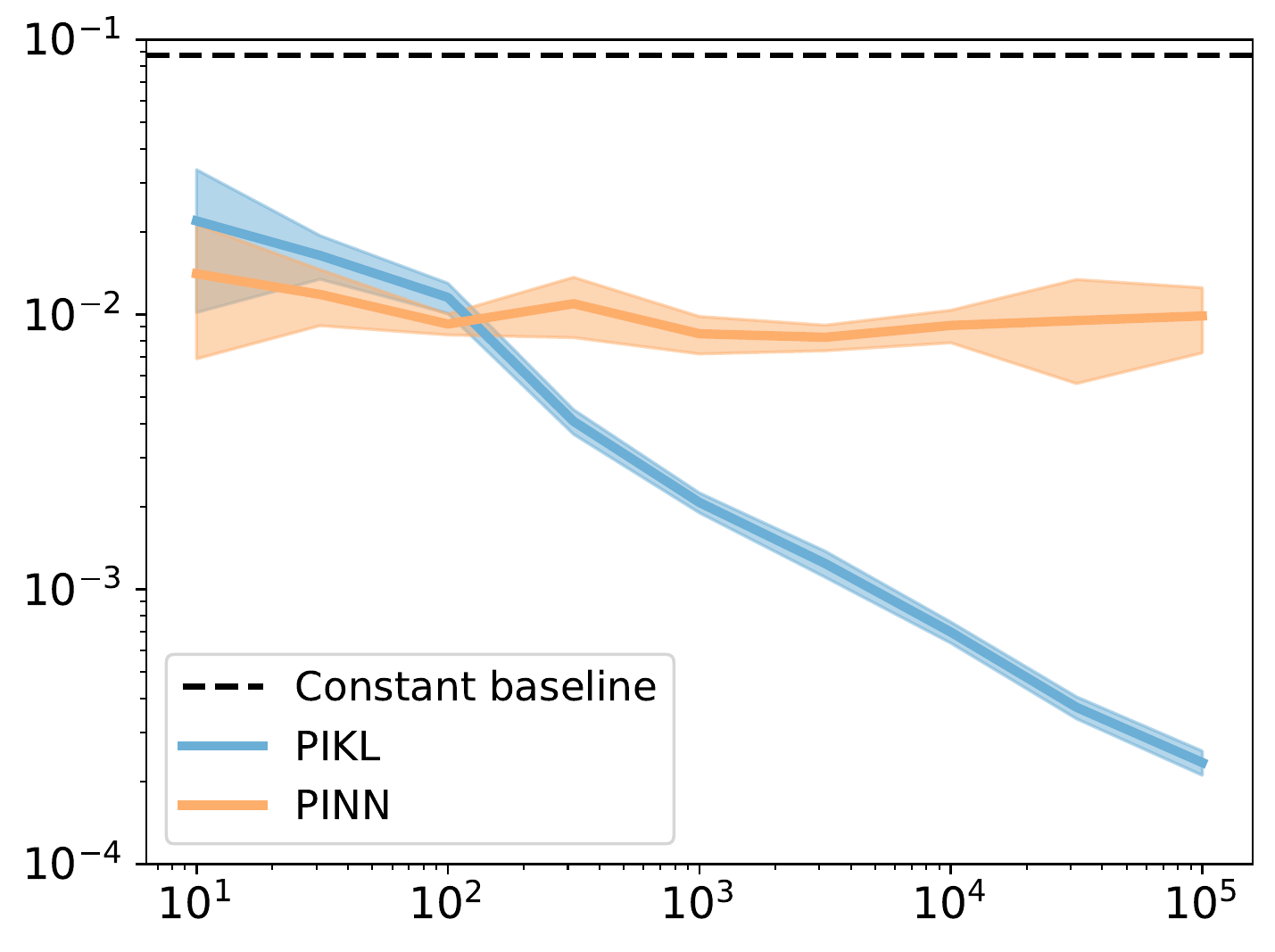}
    \caption{$L^2$ relative error of the models for the 4-dimensional heat equation with noisy boundary conditions as a function of the number of training points. Standard deviations are estimated over 5 runs.}
    \label{fig:4D}
\end{figure}

Still in the context of noisy boundary conditions, we study the feasibility and limitations of the PIKL estimator in higher dimensions.
To this aim, we consider the task of learning a solution to the heat equation in dimension 4 with noisy boundary conditions.
In this setting, the goal is to learn \[f^\star(x_1, x_2, x_3, x_4) = \exp(-3 x_1 /\pi^{2}) \cos(x_2 / \pi)  \cos(x_3 / \pi)  \cos(x_4 / \pi)\] on $\Omega = [-0.5, 0.5]^4$ given $n$ i.i.d. observations $(X_1, Y_1)$, $\dots$, $(X_n, Y_n)$ such that $Y = f^\star(X) +\mathcal N(0, 10^{-2})$, and $X$ is sampled on $(\{0\}\times [-0.5, 0.5]^3) \cup ([-0.5, 0.5]\times \partial[-0.5, 0.5]^3)$ to respectively encode the initial and boundary conditions.
The function $f^\star$ is the unique solution to the heat equation $(\partial_1- \partial^2_{2,2}-\partial^2_{3,3}-\partial^2_{4,4})f^\star = 0$ satisfying this set of initial and boundary conditions \citep[see, e.g.,][Chapter~2.3, Theorem~5]{evans2010partial}.
The function $f^\star$ is flat, and close to the constant function equal to $1$.
We compare the performance of our PIKL estimator with a PINN.
Here, the PIKL estimator is computed with $m=3$, leading to 2401 Fourier modes. 
The PINN is a fully-connected neural network with three hidden layers of size 10, using $\tanh$ as activation function, and optimized on $2\times 10^5$ collocation points by $2000$ gradient descent steps.
In this high-dimensional setting, Figure~\ref{fig:4D} shows that the PIKL estimator clearly outperforms the PINN both in terms of performance. 
Note that both methods outperform the constant model equal to $\int_\Omega f^\star$ (dotted line). 
Moreover, the PIKL estimator is more than $100$ times faster than the PINN.
The experimental convergence rate of the PIKL estimator is $n^{-0.53}$, which matches the parametric rate of $n^{-1/2}$.

\section{Conclusion and future directions}
In this article, we developed an efficient algorithm to solve the PIML hybrid problem \eqref{eq:estimator_sob_0}. The PIKL estimator can be computed exactly through matrix inversion and possesses strong theoretical properties. Specifically, we demonstrated how to estimate its convergence rate based on the PDE prior $\mathscr{D}$. Moreover, through various examples, we showed that it outperforms PINNs in terms of performance, stability, and training time in certain PDE-solving tasks where PINNs struggle to escape local minima during optimization.
Future work could focus on comparing PIKL with the implementation of RFF and exploring its performance against PINNs in the case of PDEs with non-constant coefficients. Another avenue for future research is to assess the effectiveness of the kernel approach compared to traditional PDE solvers, as discussed in Section~\ref{sec:noisy}.

\paragraph{Extension to nonlinear PDEs.} 
Extending the PIKL framework to accommodate nonlinear PDEs is an important and interesting direction for future research. 
From a PDE theory point of view, it is expected to be harder to deal with nonlinear PDEs, since nonlinear differential operators are challenging compared to linear operators. Even in dimension $d=1$, the solution of the ODE $y' = y^2$ with initial condition $y(0) = y_0$ is $y(t) = (1/y_0-t)^{-1}$, which explodes at $t=y_0$. Note that the domain of the solution $y$ is given by the set $\{t\leq y_0\}$, which intricately depends on the initial condition $y_0$. This prevents us from using a systematic methodology to solve this problem, and would require to design specific algorithms tailored to the condition-dependent geometry of the domain $\Omega$. 

\setcounter{section}{0}
\renewcommand\thesection{\thechapter.\Alph{section}}
\section{Comments on the PIKL estimator}
\subsection{Spectral methods and PIKL}
\label{sec:discussion_spec}
The Fourier approximation on which the PIKL algorithm relies resembles usual spectral methods. Spectral methods are a class of numerical techniques used to solve PDEs by representing the solution as a sum of basis functions, typically trigonometric (Fourier series) or polynomial (Chebyshev or Legendre polynomials). These methods are particularly powerful for problems with smooth solutions and periodic or well-behaved boundary conditions \citep[e.g.,][]{canuto2007spectral}. However the basis functions used in spectral and pseudo-spectral methods must be
specifically tailored to the domain $\Omega$, the differential operator $\mathscr D$, and the boundary conditions. 
This customization ensures that the method effectively captures the characteristics of the problem being solved. For example, the Fourier basis is unable to accurately reconstruct non-periodic functions on a square domain, leading to the Gibbs phenomenon at points of periodic discontinuity. A natural solution to this problem is to extend the solution of the PDE from the domain $\Omega$ to a simpler domain that admits a known spectral basis \citep[e.g.,][for Fourier basis extension]{Matthysen2016fast}. If the solution of the PDE on $\Omega$ can be extended to a solution of the same PDE on the extended domain, it becomes possible to apply a spectral method directly to the extended domain \citep[e.g.,][]{badea2001on, LUI2009541}. However, the PDE must satisfy certain regularity conditions (e.g., ellipticity), and there must be a method to implement the boundary conditions on $\partial \Omega$ instead of on the boundary of the extended domain.

In this article, we take a slightly different approach. Although we extend $\Omega \subseteq [-L, L]^d$ to $[-2L, 2L]^d$, we impose the PDE only on $\Omega$ and not on the entire extended domain $[-2L, 2L]^d$. Also, unlike spectral methods, we do not require that $\mathscr{D}(\hat{f}^{\mathrm{PIKL}}) = 0$. Instead, to ensure that the problem is well-posed, we regularize the PIML problem using the Sobolev norm of the periodic extension. This Tikhonov regularization is a conventional approach in kernel learning and is known to resemble spectral methods because it acts as a low-pass filter \citep[see, e.g.,][]{caponnetto2007optimal}. However, given a kernel, it is non-trivial to identify the basis of orthogonal functions that diagonalize it. The main contribution of this article is to establish an explicit connection between the Fourier basis and the PIML kernel, leading to surprisingly simple formulas for the kernel matrix $M_m$.

\subsection{Choice of the extended domain}
\label{sec:extended_domain}

Embedding $\Omega$ in a toroidal structure is necessary to consider periodic functions, which is a requirement for our Fourier expansion. 
Let $L_\mathbb T$ define the length of the torus, i.e., such that an extension to $[-L_\mathbb T, L_\mathbb T]$ is considered. Since we assume that $\Omega \subseteq [-L,L]^d$, we necessarily have $L_\mathbb T > L$. This condition being satisfied, any value of $L_\mathbb T > L$ would be admissible for the PIKL framework. 

However, the choice of $L_\mathbb T$ may impact the algorithmic performances. In particular, the basis functions $\Phi_k(x) = \exp(i\frac{\pi}{L_\mathbb T}\langle k,x\rangle)$ depend on $L_\mathbb T$. 
With a fixed number $m$ of Fourier modes, we see that for all $k$ such that $\|k\|_\infty\leq m$, and for all $x \in \Omega$, $\lim_{L_\mathbb T\to \infty}\Phi_k(x) = 1$. This means that, in the case of an excessively large torus (when $L_\mathbb T\to \infty$ grows), our PIKL estimator will only be able to learn constant functions. Therefore, $L_\mathbb T $ should not be too large relative to $L$.

Letting $L_\mathbb T$ be close to $L$, i.e., $L_\mathbb T = (1+\varepsilon)L$ is not a good idea. To see this, consider the case where $d=1$ and $\Omega = [-1,1]$. Then, if $\varepsilon$ is small, the periodic extension of $f^\star$ to $[-1-\varepsilon, 1+\varepsilon]$ will admit an exploding Sobolev norm. Indeed, for any periodic function $f\in H^1_{\mathrm{per}}([-1-\varepsilon, 1+\varepsilon]),$ we have  \[\|f\|_{H^1_{\mathrm{per}}([-1-\varepsilon, 1+\varepsilon])}^2 = \int_{[-1-\varepsilon, 1+\varepsilon]}f^2 + (f')^2 \geq \int_{[-1-\varepsilon, 1+\varepsilon]} (f')^2 \geq \int_{[-\varepsilon, \varepsilon]} (f'(x+1+\varepsilon))^2dx.\]
Then, the Cauchy-Schwarz inequality states that 
\[\int_{[-\varepsilon, \varepsilon]} (f'(x+1+\varepsilon))^2dx \int_{[-\varepsilon, \varepsilon]} 1dx \geq \Big(\int_{[-\varepsilon, \varepsilon]} f'(x+1+\varepsilon)dx\Big)^2,\]
meaning that 
\[\int_{[-\varepsilon, \varepsilon]} (f'(x+1))^2dx  \geq \varepsilon^{-1}[f(1)-f(-1)]^2.\]
All in all, denoting by $E(f^\star)$ the extension of $f^\star$ to $[-1-\varepsilon, 1+\varepsilon]$, we deduce that 
\[\|E(f^\star)\|_{H^1_{\mathrm{per}}([-1-\varepsilon, 1+\varepsilon])}^2 \geq \varepsilon^{-1} (f^\star(1) - f^\star(-1))^2. \]
This computation shows that taking the extension torus to be $[-1-\varepsilon, 1+\varepsilon]$ results in the Sobolev norm $\|f^\star\|_{H^1_{\mathrm{per}}([-1-\varepsilon, 1+\varepsilon])}^2$ introducing a bias towards functions satisfying $f^\star(1) = f^\star(-1)$, i.e., favoring periodic functions on $\Omega$.

The choice of an optimal constant $L_{\mathbb T}$ depending on $L$ is non-trivial, and closely relates to computing the constant in the Sobolev embedding $H^s(\Omega) \to H^s_{\mathrm{per}}([-L_{\mathbb T}, L_{\mathbb T}])$, which is known to be a difficult problem in functional analysis.
Given that $L_{\mathbb T}=2L$ strikes a balance by avoiding both pathological behaviors discussed earlier, we have adopted this choice throughout the paper.

\subsection{Reproducing property}

Here, we formally prove that both properties 
\begin{itemize}
    \item[$(i)$] $f(x) = \langle M_m^{1/2}z, M_m^{-1/2}\Phi_m(x)\rangle_{\mathbb C^{(2m+1)^d}}$, and
    \item[$(ii)$] $\|f\|^2_{\mathrm{RKHS}} = \langle z, M_mz\rangle_{\mathbb C^{(2m+1)^d}} = \|M_m^{1/2}z\|_2^2$
\end{itemize}
are sufficient to show that minimizing $\bar R_n$ is a kernel method.
From $(i)$ and $(ii)$, we deduce that the feature map is $x\mapsto M_m^{-1/2}\Phi_m(x)$. The kernel is thus necessarily given by $K(x,y) = \langle M_m^{-1/2}\Phi_m(x), M_m^{-1/2}\Phi_m(y)\rangle_{\mathbb C^{(2d+1)^d}}.$
From $(ii)$, we deduce that the $\mathrm{RKHS}$ inner product is given by $\langle z,\tilde z\rangle_{\mathrm{RKHS}} = \langle M_m^{1/2}  z, M_m^{1/2}  \tilde z\rangle_{\mathbb C^{(2d+1)^d}}$, so that $\|z\|^2_{\mathrm{RKHS}} = \|M_m^{1/2}z\|_2^2$. 

The reproducing property is then a consequence of $(i)$ and $(ii)$. Indeed, let $x\in \Omega$. We know that $\langle f, K(x, \cdot)\rangle_{\mathrm{RKHS}} = \langle M_m^{1/2}  z, M_m^{1/2}  \tilde z\rangle_{\mathbb C^{(2d+1)^d}}$,
where $z$ is the Fourier vector of $f$ and $\tilde z$ the Fourier vector of $K(x,\cdot)$.
Since the kernel $K$ results from 
\[K(x,y) = \langle M_m^{-1/2}\Phi_m(x), M_m^{-1/2}\Phi_m(y)\rangle_{\mathbb C^{(2d+1)^d}} = \langle M_m^{-1}\Phi_m(x), \Phi_m(y)\rangle_{\mathbb C^{(2d+1)^d}},\] we know that $\tilde z = M_m^{-1}\Phi_m(x)$.
This means that 
\begin{align*}
    \langle  f, K(x, \cdot)\rangle_{\mathrm{RKHS}} &= \langle  M_m^{1/2}  z, M_m^{1/2}  M_m^{-1}\Phi_m(x)\rangle_{\mathbb C^{(2d+1)^d}} = \langle  z, \Phi_m(x)\rangle_{\mathbb C^{(2d+1)^d}} = f(x),
\end{align*}
which is the reproducing property.

\section{Fundamentals of functional analysis on complex Hilbert spaces}
\label{sec:sobolev_intro}
Let $L>0$ and $d\in \mathbb N^\star$. We define $L^2([-2L,2L]^d, \mathbb C)$ as the space of complex-valued functions $f$ on the hypercube $[-2L,2L]^d$ such that $\int_{[-2L,2L]^d} |f|^2 < \infty$. The real part of $f$ is denoted by $\Re(f)$, and the imaginary part by $\Im(f)$, such that $f = \Re(f) + i \Im(f)$. Throughout the appendix, for the sake of clarity, we use the dot symbol $\cdot$ to represent functions. For example, $\|\cdot\| $ denotes the function $x\mapsto \|x\|$, and $\langle\cdot, \cdot\rangle$ stands for the function $( x,y) \mapsto \langle x, y\rangle$.
\begin{defi}[$L^2$-space and $\|\cdot\|_2$-norm]
    The separable Hilbert space $L^2([-2L,2L]^d, \mathbb C)$ is associated with the inner product $\langle f,g\rangle = \int_{[-2L,2L]^d} f\bar g$ and the norm $\|f\|_2^2 = \int_{[-2L,2L]^d} |f|^2$.
\end{defi}
Let $s\in\mathbb N$. 
\begin{defi}[Periodic Sobolev spaces]
    The periodic Sobolev space $H^s_{\mathrm{per}}([-2L,2L]^d, \mathbb R)$ is the space of real functions $f(x) = \sum_{k\in\mathbb Z^d} z_k \exp(\frac{i\pi}{2L}\langle k,x\rangle)$ such that the Fourier coefficients $z_k$ satisfy $\sum_k |z_k|^2(1+\|k\|^{2s}) <\infty$. The corresponding complex periodic Sobolev space is defined by \[H^s_{\mathrm{per}}([-2L,2L]^d, \mathbb C) = H^s_{\mathrm{per}}([-2L,2L]^d, \mathbb R) \oplus i\; H^s_{\mathrm{per}}([-2L,2L]^d, \mathbb R).\] It is the space of complex-valued functions $f(x) = \sum_{k\in\mathbb Z^d} z_k \exp(\frac{i\pi}{2L}\langle k,x\rangle)$ such that $\sum_k |z_k|^2(1+\|k\|^{2s}) <\infty$.
\end{defi}
We recall that, given two Hilbert spaces $\mathcal H_1$ and $\mathcal H_2$, an operator is a linear function from $\mathcal H_1$ to $\mathcal H_2$.
\begin{defi}[Operator norm]\citep[e.g.,][Section 2.6]{brezis2010functional}
\label{defi:op_norm}
    Let $\mathscr O: L^2([-2L, 2L]^d, \mathbb C)\to L^2([-2L, 2L]^d, \mathbb C)$ be an operator. Its operator norm $|||\mathscr O|||_2$ is defined by
    \[|||\mathscr O|||_2 =\underset{\|g\|_2 = 1}{\sup_{g \in L^2([-2L, 2L]^d, \mathbb C)}} \|\mathscr O g\|_2 = \underset{g \neq 0}{\sup_{g \in L^2([-2L, 2L]^d, \mathbb C)}} \|g\|_2^{-1}\|\mathscr O g\|_2. \]
\end{defi}
The operator norm is sub-multiplicative, i.e., $|||\mathscr O_1 \circ \mathscr O_2|||_2 \leqslant |||\mathscr O_1|||_2 \times |||\mathscr O_2|||_2$. 
\begin{defi}[Adjoint]
    Let $(\mathcal H, \langle\cdot, \cdot\rangle_{\mathcal H})$ be an Hilbert space and $\mathscr O:\mathcal H \to \mathcal H$ be an operator. The adjoint $\mathscr O^\star$ of $\mathscr O$ is the unique operator such that $\forall f,g \in \mathcal H$, $\langle f, \mathscr O g\rangle_{\mathcal H} = \langle \mathscr O^\star f,  g\rangle_{\mathcal H}$.
\end{defi}
If $\mathcal H = \mathbb R^d$ with the canonical scalar product, then $\mathscr O^\star$ is the $d\times d$ matrix $\mathscr O^\star = \mathscr O^T$. If $\mathcal H = \mathbb C^d$ with the canonical sesquilinear inner product, then $\mathscr O^\star$ is the $d\times d$ matrix $\mathscr O^\star = \bar{\mathscr O}^T$.
\begin{defi}[Hermitian operator]
    Let $\mathcal H$ be an Hilbert space and $\mathscr O:\mathcal H \to \mathcal H$ be an operator. The operator $\mathscr O$ is said to be Hermitian if $\mathscr O = \mathscr O^\star$.
\end{defi}
\begin{thm}[Spectral theorem]\citep[e.g.][Theorems 12.29 and 12.30]{rudin1991functional}
    Let $\mathscr O$ be a positive Hermitian compact operator. Then $\mathscr O$ is diagonalizable on an Hilbert basis with positive eigenvalues that tend to zero. We denote its eigenvalues, ordered in decreasing order, by $\sigma(\mathscr O) = (\sigma_k^\downarrow(\mathscr O))_{k\in \mathbb N^\star}$.
    \label{thm:spectral2}
\end{thm}
We emphasize that, given an invertible positive self-adjoint compact operator $\mathscr O$ and its inverse $\mathscr O^{-1}$, the eigenvalues of $\mathscr O^{-1}$ can also be ordered in increasing order, i.e., 
\begin{equation}
    \sigma(\mathscr O^{-1}) = (\sigma_k^\uparrow(\mathscr O^{-1}))_{k\in \mathbb N^\star} = (\sigma_k^\downarrow(\mathscr O)^{-1})_{k\in \mathbb N^\star}. \label{eq:order}
\end{equation}
\begin{thm}[Courant-Fischer minmax theorem ]\citep[][Problem 37]{brezis2010functional}
    Let $\mathscr O: \mathcal H \to \mathcal H$ be a positive Hermitian compact operator. Then
    \[\sigma_k^\downarrow(\mathscr O) = \underset{\dim H = k}{\max_{H\subseteq \mathcal H}} \;  \underset{\|g\|_2=1}{\min_{g\in H }} \langle g, \mathscr O g\rangle_{\mathcal H}.\]
    If $\mathscr O$ is injective, then
\[\sigma_k^\uparrow(\mathscr O^{-1}) = \underset{\dim H = k}{\min_{H\subseteq \mathscr O(\mathcal H)}} \;  \underset{\|g\|_2=1}{\max_{g\in H }} \langle g, \mathscr O^{-1} g\rangle_{\mathcal H}.\]
\label{thm:minmax}
\end{thm}
Interestingly, if $\mathscr O$ is a positive Hermitian compact operator, then Theorem~\ref{thm:minmax} shows that $|||\mathscr O|||_2$ equals its largest eigenvalue. 
\begin{defi}[Orthogonal projection on $H_m$]
    We let $\Pi_m: H^s_{\mathrm{per}}([-2L,2L]^d, \mathbb C) \to H_m$ be the orthogonal projection with respect to $\langle\cdot, \cdot\rangle$, i.e., for all $f \in H^s_{\mathrm{per}}([-2L,2L]^d, \mathbb C),$ \[ \Pi_m f(y) = \sum_{\|{k}\|_\infty \leqslant m}\Big(\frac{1}{(4L)^{d}}\int_{[-2L,2L]^d}\exp\Big(-\frac{i\pi}{2L}\langle {k}, x\rangle\Big)f(x)dx\Big)\exp\Big(\frac{i\pi}{2L}\langle {k}, y\rangle\Big).\]
    Note that $\Pi_m$ is Hermitian and that, for all $f \in L^2([-2L,2L]^d, \mathbb C)$, $\lim_{m\to\infty}\|f-\Pi_m f\|_2 = 0$.
\end{defi}

\section{Theoretical results for PIKL}
\label{app:Theory}
\subsection{Detailed computation of the Fourier expansion of the differential penalty}
The formula relating $\|\mathscr D(f)\|_{L^2(\Omega)}$ to the Fourier coefficients of $f$ follows from $(i)$ expanding $f$ in the Fourier basis, $(ii)$ leveraging the linearity of $\mathscr{D}$, $(iii)$ applying the property $\mathscr{D}(x\mapsto e^{-\frac{i \pi}{2L}\langle k, x\rangle}) =  (x\mapsto P(k)e^{-\frac{i \pi}{2L}\langle k, x\rangle})$, and (iv) recognizing that $|z|^2 = z \bar z$. From these steps, we deduce that
\begin{align*}
    \|\mathscr D(f)\|_{L^2(\Omega)}^2 
    &= \int_\Omega |\mathscr{D}(f)(x)|^2dx\\
    &= \int_\Omega \Big|\sum_{\|j\|_\infty\leqslant m} z_j \frac{P(j)}{(4L)^{d/2}} e^{-\frac{i \pi}{2L}\langle j, x\rangle}\Big|^2dx,\\
    &= \int_\Omega \Big(\sum_{\|j\|_\infty\leqslant m} z_j \frac{P(j)}{(4L)^{d/2}} e^{-\frac{i \pi}{2L}\langle j, x\rangle}\Big)\Big(\sum_{\|k\|_\infty\leqslant m} \bar{z}_k \frac{\bar{P}(k)}{(4L)^{d/2}} e^{\frac{i \pi}{2L}\langle k, x\rangle}dx\Big)dx\\
    &= \sum_{\|j\|_\infty\leqslant m,\|k\|_\infty\leqslant m} z_j \bar{z_k} \frac{P(j)\bar P(k)}{(4L)^d}\int_\Omega e^{\frac{i \pi}{2L}\langle k-j, x\rangle}dx.
\end{align*}

\subsection{Proof of Proposition \ref{prop:geom_cube}}
We have
\begin{align*}
    \frac{1}{(4L)^{d}}\int_{[-L,L]^d} e^{\frac{i \pi}{2L}\langle k, x\rangle}dx &= \frac{1}{(4L)^{d}}\prod_{j=1}^d\int_{[-L,L]} e^{\frac{i \pi}{2L} k_j x}dx = \prod_{j=1}^d\Big[\frac{1}{2i\pi} e^{\frac{i \pi}{2L} k_j x}\Big]_{x=-L}^L\\
    &= \prod_{j=1}^d\frac{e^{\frac{i \pi}{2} k_j }-e^{-\frac{i \pi}{2} k_j }}{2i\pi k_j} = \prod_{j=1}^d\frac{\sin(\frac{\pi}{2} k_j)}{\pi k_j}.
\end{align*}
The characteristic function of the Euclidean ball is computed in \citet[][Table 13.4]{bracewell}.
\subsection{Operations on characteristic functions}
\begin{prop}[Operations on characteristic functions]
\label{prop:op_char}
    Consider $d \in \mathbb N^\star$, $L >0$, and $\Omega \subseteq [-L,L]^d$.
    \begin{itemize}
        \item Let $a \in [-1,1]$. Then $a\cdot\Omega \subseteq [-L,L]^d$ and 
        \[F_{a\cdot\Omega}(k) = |a|^d\times F_{\Omega}(a\cdot k).\]
        \item Let $\tilde \Omega \subseteq [-L,L]^d$ be a domain such that $\Omega \cap \tilde \Omega = \emptyset$. Then $\Omega \sqcup \tilde \Omega \subseteq [-L,L]^d$ and
        \[F_{\Omega \sqcup \tilde \Omega}(k) = F_{\Omega}(k) + F_{\tilde \Omega}(k).\]
        \item Assume that $\Omega \subseteq [-L/2,L/2]^d$, and let $z \in \mathbb R^d$ be such that $\|z\|_\infty < L/2$. Then $\Omega + z \subseteq [-L,L]^d$ and \[F_{\Omega + z}(k) = F_{\Omega}(k) \times \exp\Big(\frac{i \pi}{2L} \langle k,z\rangle\Big).\]
        \item Assume that $\Omega = \Omega_1 \times \Omega_2$, where $\Omega_1\subseteq [-L,L]^{d_1}$, $\Omega_2 \subseteq [-L,L]^{d_2}$, and $d_1+d_2 = d$.  Then
        \[F_{\Omega}(k) = F_{\Omega_1}(k_1, \hdots, k_{d_1}) \times F_{\Omega_2}(k_{d_1+1}, \hdots, k_{d}).\]   
    \end{itemize}
    \end{prop}

\subsection{Operator extensions}
\begin{defi}[Projection on $\Omega$]
    The projection $C: L^2([-2L,2L]^d, \mathbb C) \to L^2([-2L,2L]^d, \mathbb C)$ on $\Omega$  is defined by $C(f) = 1_\Omega f$.
\end{defi}

\begin{defi}[Operator extensions]
    The operators $C_m: H_m \to H_m$, $M_m: H_m \to H_m$, and $M_m^{-1}: H_m \to H_m$ can be extended to $L^2([-2L,2L]^d,  \mathbb C)$ by $C_m = \Pi_m C_m \Pi_m$, $M_m = \Pi_m M_m \Pi_m$, and $M_m^{-1} = \Pi_m M_m^{-1} \Pi_m$.
\end{defi}
From now on, we consider the extensions of these operators, allowing us to express equivalently \[|||M_m^{-1}|||_2 = \underset{\|g\|_2 = 1}{\sup_{g \in H_m}} \|M_m^{-1}g\|_2  = \underset{\|g\|_2 = 1}{\sup_{g \in L^2([-2L, 2L]^d)}} \|M_m^{-1}g\|_2.\]
It is important to note that the extended operator $M_m^{-1}$ is no longer the inverse of the extended operator $M_m$.

\begin{prop}[Compact operator extension]
    Let $\mathscr O$ be a positive Hermitian compact  operator on $L^2([-2L,2L]^d, \mathbb R)$. Then its unique extension $\tilde{\mathscr O}$ to $L^2([-2L,2L]^d, \mathbb C)$ is a positive Hermitian compact operator with the same real eigenfunctions and positive eigenvalues.
    \label{prop:extension2}
\end{prop}
\begin{proof}
    Since $\tilde{\mathscr O}$ is $\mathbb C$-linear, we necessarily have $\tilde{\mathscr O} (f) = \mathscr O (\Re(f)) + i \mathscr O (\Im(f))$. Therefore, the extension is unique. Since $\mathscr O$ is compact, $\tilde{\mathscr O}$ is also compact. According to Theorem \ref{thm:spectral2}, the operator $\mathscr O$ is diagonalizable in a Hermitian basis $(f_k)_{k\in \mathbb N^\star}$. Thus, for all $f \in L^2([-2L,2L], \mathbb R)$,
    \[\mathscr O(f) = \sum_{k\in\mathbb N^\star} \sigma_k^\downarrow(\mathscr O) \langle f, f_k\rangle_{L^2([-2L,2L], \mathbb R)}f_k.\] 
    Thus, for all $f\in L^2([-2L,2L], \mathbb C)$
    \begin{align*}
        \tilde{\mathscr O} (f) &= \sum_{k\in\mathbb N^\star} \sigma_k^\downarrow(\mathscr O) (\langle \Re(f), f_k\rangle_{L^2([-2L,2L], \mathbb R)}+ i\langle \Im(f), f_k\rangle_{L^2([-2L,2L], \mathbb R)})f_k\\
        &= \sum_{k\in\mathbb N^\star} \sigma_k^\downarrow(\mathscr O) \langle f, f_k\rangle_{L^2([-2L,2L], \mathbb C)}f_k.
    \end{align*}
    This formula shows that $\tilde{\mathscr O}$ is Hermitian and diagonalizable with the same real eigenfunctions and positive eigenvalues as $\mathscr O$.
\end{proof}
 Recall that $\mathscr O_n$ is the operator $\mathscr O_n = \lim_{m\to \infty} M_m^{-1}$, where the limit is taken in the sense of the operator norm \citep[see Proposition~B.2][]{doumeche2024physicsinformed}.  
\begin{defi}[Operator $M$]
    Proposition \ref{prop:extension2} shows that the operator $\mathscr O_n$ can be extended to $L^2([-2L,2L], \mathbb C)$. We denote the extension of $\mathscr O_n$ by $M^{-1}$. 
\end{defi}
The uniqueness of the extension in Proposition \ref{prop:extension2} implies that the extension of the operator $C\mathscr O_nC: L^2([-2L,2L]^d)\to L^2([-2L,2L]^d)$ to $\mathbb C$ is indeed $CM^{-1}C: L^2([-2L,2L]^d, \mathbb C)\to L^2([-2L,2L]^d, \mathbb C)$.
Proposition \ref{prop:extension2} shows that $C\mathscr O_nC$ has the same eigenvalues as $CM^{-1}C$.

\subsection{Convergence of $M_m^{-1}$}
\begin{lem}[Bounding the spectrum of $M_m^{-1}$]
    Let $m \in \mathbb N^\star$. Then, for all $k\in \mathbb N^\star$, 
    \[\sigma_k^\downarrow(M_m^{-1}) \leqslant  \sigma_k^\downarrow(M^{-1}).\]
    \label{lem:bounding_spectrum}
\end{lem}
\begin{proof}
    Let $f \in H_m$. Then, \[\langle f, M_m f\rangle = \langle f, M f\rangle = \lambda_n \|f\|_{H^s_{\mathrm{per}}([-2L,2L]^d)}^2 + \mu_n \|\mathscr D(f)\|_{L^2(\Omega)}^2.\]
    Thus, using Theorem \ref{thm:minmax}, we deduce that 
    \begin{align*}
    \sigma_k^\uparrow(M_m) &= \underset{\dim H = k}{\min_{H\subseteq H_m}} \;  \underset{\|g\|_2=1}{\max_{g\in H }} \langle g, M_m g\rangle\\
    &= \underset{\dim H = k}{\min_{H\subseteq H_m}} \; \underset{\|g\|_2=1}{\max_{g\in H }} \langle g, M g\rangle\\
    &\geqslant  \underset{\dim H = k}{\min_{H\subseteq H^s_{\mathrm{per}}([-2L,2L]^d)}} \; \underset{\|g\|_2=1}{\max_{g\in H }} \langle g, M g\rangle\\
    &=\sigma_k^\uparrow(M).
    \end{align*} 
    From \eqref{eq:order}, we deduce that $\sigma_k^\downarrow(M_m^{-1}) \leqslant \sigma_k^\downarrow(M^{-1})$. 
\end{proof}

\begin{lem}[Spectral convergence of $M_m$]
    Let $m \in \mathbb N^\star$. Then, for all $k\in \mathbb N^\star$, one has
    \[\lim_{m\to\infty}\sigma_k^\downarrow(M_m^{-1}) = \sigma_k^\downarrow(M^{-1}).\]
    \label{lem:convergence_spectrum}
\end{lem}
\begin{proof}
    By continuity of the RKHS norm $f\mapsto \langle f, Mf\rangle$ on $H^s_{\mathrm{per}}([-2L,2L]^d, \mathbb C)$ \citep[][Proposition B.1]{doumeche2024physicsinformed}, we deduce that, for all function $f\in H^s_{\mathrm{per}}([-2L,2L]^d, \mathbb C)$, the quantity $ \lambda_n \|\Pi_m(f)\|_{H^s_{\mathrm{per}}([-2L,2L]^d, \mathbb C)}^2 + \mu_n \|\mathscr D(\Pi_m(f))\|_{L^2(\Omega, \mathbb C)}^2$ converges, as $m$ goes to the infinity, to $ \lambda_n \|f\|_{H^s_{\mathrm{per}}([-2L,2L]^d, \mathbb C)}^2 + \mu_n \|\mathscr D(f)\|_{L^2(\Omega, \mathbb C)}^2$. Thus, 
    \begin{equation*}
    \forall f\in H^s_{\mathrm{per}}([-2L,2L]^d, \mathbb C),\quad \lim_{m\to \infty}\langle f, (M-M_m)f\rangle = 0.
    \end{equation*}

Next, consider $f_1, \hdots, f_k$ to be the eigenfunctions of $M$ associated with the ordered eigenvalue $\sigma_1^\uparrow(M), \hdots,\sigma_k^\uparrow(M)$. 
    Since, for any $1\leqslant j,\ell \leqslant k$, we have that $\lim_{m\to \infty}\langle f_j + f_\ell, M_m(f_j+f_\ell)\rangle = \langle f_j + f_\ell, M(f_j+f_\ell)\rangle$ and $\lim_{m\to \infty}\langle f_j + f_\ell, M_m(f_j+f_\ell)\rangle = \lim_{m\to \infty}\langle f_j, M_m f_j\rangle + \lim_{m\to \infty}\langle f_\ell, M_m f_\ell\rangle + 2\lim_{m\to \infty}\Re(\langle f_j, M_m f_\ell\rangle)$, we deduce that $\lim_{m\to \infty}\;\Re(\langle f_j, M_m f_\ell\rangle) = $ $\Re(\langle f_j, Mf_\ell\rangle)$. Using the same argument by developing $\langle f_j + if_\ell, M_m(f_j+if_\ell)\rangle$ shows that $\lim_{m\to \infty}\Im(\langle f_j, M_m f_\ell\rangle) = \Im(\langle f_j, Mf_\ell\rangle)$. Overall,
    \begin{equation}
        \forall 1\leqslant j,\ell \leqslant k,\quad \lim_{m\to \infty}\langle f_j, M_m f_\ell\rangle = \langle f_j, Mf_\ell\rangle.
    \label{eq:pointwise_cv}
    \end{equation}
     Now, observe that \[(g\in \mathrm{Span}(f_1, \hdots, f_k)  \hbox{ and } \|g\|_2=1) \Leftrightarrow (\exists (a_1, \hdots, a_k)\in \mathbb C^k, \;g = \sum_{j=1}^k a_j f_j \hbox{ and } \sum_{j=1}^k |a_j|^2 = 1).\] Thus,
     \begin{align*}
         \underset{\|g\|_2=1}{\max_{g\in \mathrm{Span}(f_1, \hdots, f_k) }} |\langle g, M_m g\rangle -\langle g, M g\rangle|&\leqslant \max_{\|a\|_2=1} \sum_{i,j=1}^k |a_i a_j| |\langle f_i, (M_m-M) f_j\rangle|\\
         & \leqslant k \max_{1\leqslant i,j\leqslant k}|\langle f_i, (M_m-M) f_j\rangle|\\
        &\xrightarrow{m\to \infty}0\quad \hbox{according to \eqref{eq:pointwise_cv}}.
     \end{align*}
     So,
     \begin{align}
         \lim_{m\to\infty}\underset{\|g\|_2=1}{\max_{g\in \mathrm{Span}(\Pi_m(f_1), \hdots, \Pi_m(f_k)) }} \langle g, M_m g\rangle &= \lim_{m\to\infty}\underset{\|g\|_2=1}{\max_{g\in \mathrm{Span}(f_1, \hdots, f_k) }} \langle g, M_m g\rangle\nonumber\\
         &= \underset{\|g\|_2=1}{\max_{g\in \mathrm{Span}(f_1, \hdots, f_k) }} \langle g, M g\rangle\nonumber\\
         &= \sigma_k^\uparrow(M)\label{eq:bound_vp}.
     \end{align}
     Note that $\mathrm{Span}(\Pi_m(f_1), \hdots, \Pi_m(f_k)) \subseteq H_m$. Moreover, for $m$ large enough, we have that $\dim \mathrm{Span}(\Pi_m(f_1), \hdots, \Pi_m(f_k)) = k$.
     Therefore, according to Theorem \ref{thm:minmax}, \[\sigma_k^\uparrow(M_m) = \underset{\dim H = k}{\min_{H\subseteq H_m}} \;  \underset{\|g\|_2=1}{\max_{g\in H }} \langle g, M_m g\rangle \leqslant \underset{\|g\|_2=1}{\max_{g\in \mathrm{Span}(\Pi_m(f_1), \hdots, \Pi_m(f_k)) }} \langle g, M_m g\rangle .\] Combining this inequality with identity \eqref{eq:bound_vp} shows that 
     $\limsup_{m\to\infty}\sigma_k^\uparrow(M_m) \leqslant \sigma_k^\uparrow(M)$.
     Equivalently, \[\liminf_{m\to\infty}\sigma_k^\downarrow(M_m^{-1}) \geqslant  \sigma_k^\downarrow(M^{-1}).\] 
     Finally, by Lemma \ref{lem:bounding_spectrum}, we have $\sigma_k^\downarrow(M_m^{-1}) \leqslant \sigma_k^\downarrow(M^{-1})$. We conclude that $\lim_{m\to\infty}\sigma_k^\downarrow(M_m^{-1}) = \sigma_k^\downarrow(M^{-1}).$
\end{proof}

\begin{lem}[Eigenfunctions convergence]
    Let $(f_{j,m})_{j\in\mathbb N^\star}$ be the eigenvectors of $M_m$ associated with the eigenvalues $(\sigma^\uparrow_j(M_m))_{j\in\mathbb N^\star}$. Let $E_j = \ker(M-\sigma^\uparrow_j(M)\mathrm{Id})$.
    Then
    \[\forall j\in \mathbb N^\star, \quad \lim_{m\to\infty} \min_{y\in E_j}\|f_{j,m}-y\|_2 = 0.\]
    \label{lem:eigenfunction_cv}
\end{lem}
\begin{proof}
Let $(f_j)_{j\in\mathbb N^\star}$ be the eigenvectors of $M$ associated with the eigenvalues $(\sigma^\uparrow_j(M))_{j\in\mathbb N^\star}$.

    The proof proceeds by contradiction. Assume that the lemma is false, and 
    consider the minimum integer $p\in \mathbb N^\star$ such that $\limsup_{m\to\infty} \min_{y\in E_p}\|f_{p,m}-y\|_2  > 0$. 
    Let $k_1 < p$ be the largest integer such that $\sigma^\uparrow_{k_1}(M_m) < \sigma^\uparrow_p(M_m)$, and let $k_2 > p$ be the smallest integer such that $\sigma^\uparrow_{k_2}(M_m) > \sigma^\uparrow_p(M_m)$. Observe that $E_p = \mathrm{Span}(f_{k_1+1}, \hdots, f_{k_2-1})$.

    Let $k_1 < j < k_2$. We know that
    $\langle f_j, M_m f_j\rangle = \sum_{\ell \in \mathbb N^\star} \sigma^\uparrow_j(M_m) |\langle f_{\ell, m}, f_j\rangle|^2$ from diagonalizing $M_m$. 
    The minimality assumption on $j$ ensures that, for all $\ell \leqslant k_1$, $\lim_{m\to\infty} \min_{y\in E_\ell}\|f_{\ell,m}-y\|_2 = 0$. Since $E_j \subseteq \mathrm{Span}(f_1, \hdots, f_{k_1})$, we deduce that $\forall \ell \leqslant k_1$, $\lim_{m\to\infty} \langle f_{\ell, m}, f_j\rangle = 0$. Thus, 
    \begin{equation}
        \sum_{\ell\geqslant  k_1}  |\langle f_{\ell, m}, f_j\rangle|^2 = 1+ o_{m\to \infty}(1),\label{eq:technical1}
    \end{equation}
    and
    \begin{equation}
        \sum_{k_1\leqslant \ell < k_2} \sigma^\uparrow_j(M_m) |\langle f_{\ell, m}, f_j\rangle|^2 = o_{m\to \infty}(1).\label{eq:technical2}
    \end{equation}
    By Lemma \ref{lem:convergence_spectrum}, using $|\langle f_{\ell, m}, f_j\rangle|^2 \leqslant 1$, we have 
    \begin{equation}
        \sum_{k_1\leqslant \ell \leqslant k_2} \sigma^\uparrow_j(M_m) |\langle f_{\ell, m}, f_j\rangle|^2 = \sigma^\uparrow_j(M) \sum_{k_1\leqslant \ell < k_2}  |\langle f_{\ell, m}, f_j\rangle|^2 + o_{m\to \infty}(1).\label{eq:technical3}
    \end{equation}
    Combining \eqref{eq:technical2} and \eqref{eq:technical3}, we deduce that 
    \[\langle f_j, M_m f_j\rangle = o_{m\to \infty}(1) +\sigma^\uparrow_j(M)\sum_{k_1\leqslant \ell < k_2} |\langle f_{\ell, m}, f_j\rangle|^2 + \sum_{ \ell \geqslant  k_2} \sigma^\uparrow_j(M_m) |\langle f_{\ell, m}, f_j\rangle|^2.\]
    Moreover, identity
    \eqref{eq:pointwise_cv} ensures that $\langle f_j, M_m f_j\rangle =  \sigma^\uparrow_j(M) +  o_{m\to \infty}(1).$
    Thus, 
    \begin{equation*}
        \sigma^\uparrow_j(M) (1 - \sum_{k_1\leqslant \ell < k_2} |\langle f_{\ell, m}, f_j\rangle|^2) = o_{m\to \infty}(1) +  \sum_{ \ell \geqslant  k_2} \sigma^\uparrow_j(M_m) |\langle f_{\ell, m}, f_j\rangle|^2.
    \end{equation*}
    However, according to Lemma \ref{lem:convergence_spectrum}, there is $\varepsilon > 0$ such that, for $m$ large enough,
    \[\forall \ell \geqslant  k_2, \quad \sigma^\uparrow_j(M_m) \geqslant  \sigma^\uparrow_k(M) + \varepsilon.\]
    Hence, 
    \begin{equation*}
        \sigma^\uparrow_j(M) (1 - \sum_{k_1\leqslant \ell < k_2} |\langle f_{\ell, m}, f_j\rangle|^2) \geqslant  o_{m\to \infty}(1) +  (\sigma^\uparrow_j(M)+\varepsilon)\sum_{ \ell \geqslant  k_2}  |\langle f_{\ell, m}, f_j\rangle|^2.
    \end{equation*}
    Combining this inequality with \eqref{eq:technical1}, this means that
    \begin{equation*}
        0 \geqslant  o_{m\to \infty}(1) +  \varepsilon\sum_{ \ell \geqslant  k_2}  |\langle f_{\ell, m}, f_j\rangle|^2.
    \end{equation*}
    Thus, $\lim_{m\to\infty} \sum_{ \ell \geqslant  k_2}  |\langle f_{\ell, m}, f_j\rangle|^2 = 0$ and $\lim_{m\to\infty} \sum_{ k_1 \leqslant \ell < k_2}  |\langle f_{\ell, m}, f_j\rangle|^2 = 1$. 

    We deduce that, for all $k_1 < j < k_2$, $\lim_{m\to\infty} \min_{y\in \mathrm{Span}(f_{k_1+1, m}, \hdots, f_{k_2-1, m})}\|f_{j}-y\|_2 = 0$. By symmetry of the $\ell^2$-distance between two spaces of the same dimension $k_2-k_1-1$, for all $k_1 < j < k_2$, $\lim_{m\to\infty} \min_{y\in \mathrm{Span}(f_{k_1+1}, \hdots, f_{k_2-1})}\|f_{j,m}-y\|_2 = 0$. This contradicts the fact that $\limsup_{m\to\infty} \min_{y\in E_p}\|f_{p,m}-y\|_2  > 0$.
\end{proof}

\begin{lem}[Convergence of $M_m^{-1}$] One has
    \[\lim_{m \to \infty} |||M^{-1}-M_m^{-1}|||_2 = 0.\]
    \label{lem:operator_norm}
\end{lem}
\begin{proof}
    Let $(f_{j,m})_{j\in\mathbb N^\star}$ be the eigenvectors of $M_m$, each associated with the corresponding eigenvalues $(\sigma^\uparrow_j(M_m))_{j\in\mathbb N^\star}$. Let $(f_j)_{j\in\mathbb N^\star}$ be the eigenvectors of $M$, each associated with the eigenvalues $(\sigma^\uparrow_j(M))_{j\in\mathbb N^\star}$.
    By Lemma \ref{lem:bounding_spectrum}, $\sigma^\downarrow_j(M_m^{-1}) \leqslant \sigma^\downarrow_j(M^{-1})$; by Lemma \ref{lem:convergence_spectrum}, $\lim_{m\to\infty} \sigma^\downarrow_j(M_m^{-1}) = \sigma^\downarrow_j(M^{-1})$; and by Lemma \ref{lem:eigenfunction_cv} $\lim_{m\to\infty} \min_{y\in E_j}\|f_{j,m}-y\|_2 = 0.$

    Notice that $M_m^{-1} = \sum_{\ell \in \mathbb N } \sigma^\downarrow_j(M_m^{-1}) \langle f_{j,m}, \cdot\rangle f_{j,m}$ and that $M^{-1} = \sum_{\ell \in \mathbb N } \sigma^\downarrow_j(M^{-1}) \langle f_{j}, \cdot\rangle f_{j}$. Let $g\in L^2([-2L,2L]^d, \mathbb C)$ be such that $\|g\|_2 = 1$.
    Then,
    \begin{align*}
        \|(M^{-1}-M^{-1}_m)g\|_2 &\leqslant \sum_{j \in \mathbb N^\star} \|\sigma^\downarrow_j(M_m^{-1}) \langle f_{j,m}, g\rangle f_{j,m} - \sigma^\downarrow_j(M^{-1}) \langle f_{j}, g\rangle f_{j}\|_2\\
        &\leqslant \sum_{j \in \mathbb N^\star} (\sigma^\downarrow_j(M_m^{-1})-\sigma^\downarrow_j(M^{-1}))\| \langle f_{j,m}, g\rangle f_{j,m}\|_2 \\
        &\quad + \sum_{j \in \mathbb N^\star}\sigma^\downarrow_j(M^{-1})\| \langle f_{j,m}- f_{j}, g\rangle f_{j,m}\|_2\\
        &\quad + \sum_{j \in \mathbb N^\star}\sigma^\downarrow_j(M^{-1})\| \langle f_{j}, g\rangle (f_{j,m}-f_{j})\|_2.
    \end{align*}
    Since $\|f_{j,m}\|_2 = \|f_j\|_2 = 1$, it follows that $| \langle f_{j,m}, g\rangle| \leqslant 1$. Additionally, by the Cauchy-Schwarz inequality, $| \langle f_{j}, g\rangle| \leqslant 1$ and $| \langle f_{j,m}- f_{j}, g\rangle | \leqslant \| (f_{j,m}-f_{j})\|_2$. Thus, the above inequality can be simplified as 
    \begin{align*}
        \|(M^{-1}-M^{-1}_m)g\|_2 
        &\leqslant \sum_{j \in \mathbb N^\star} (\sigma^\downarrow_j(M_m^{-1})-\sigma^\downarrow_j(M^{-1}) + 2\sigma^\downarrow_j(M^{-1}) \| f_{j,m}-f_{j}\|_2).
    \end{align*}
    Thus, 
    \begin{align*}
        |||M^{-1}-M^{-1}_m|||_2 
        &\leqslant \sum_{j \in \mathbb N^\star} (\sigma^\downarrow_j(M_m^{-1})-\sigma^\downarrow_j(M^{-1}) + 2\sigma^\downarrow_j(M^{-1}) \| f_{j,m}-f_{j}\|_2).
    \end{align*}
    Clearly, since $|\sigma^\downarrow_j(M_m^{-1})-\sigma^\downarrow_j(M^{-1})| \leqslant 2\sigma^\downarrow_j(M^{-1})$ and $\| f_{j,m}-f_{j}\|_2 \leqslant 2$,  \[|\sigma^\downarrow_j(M_m^{-1})-\sigma^\downarrow_j(M^{-1}) + 2\sigma^\downarrow_j(M^{-1}) \| f_{j,m}-f_{j}\|_2|\leqslant 4\sigma^\downarrow_j(M^{-1}).\] Moreover, $\sum_{j \in \mathbb N^\star} \sigma^\downarrow_j(M^{-1}) < \infty$ \citep[] [Proposition B.6]{doumeche2024physicsinformed}. 
    Thus, since we have that $\lim_{m\to \infty}|\sigma^\downarrow_j(M_m^{-1})-\sigma^\downarrow_j(M^{-1})| = \lim_{m\to\infty }\| (f_{j,m}-f_{j})\|_2 = 0$, we conclude with the dominated convergence theorem that $\lim_{m\to\infty} |||M^{-1}-M^{-1}_m|||_2 = 0$, as desired.
\end{proof}

\subsection{Operator norms of $C_m$ and $C$}
\begin{lem}
    One has $|||C|||_2 \leqslant 1$ and $|||C_m|||_2\leqslant 1$, for all $m\in \mathbb N^\star$.
    \label{lem:C_op_norm}
\end{lem}
\begin{proof}
    Let $g\in L^2([-2L,2L]^d, \mathbb C)$. Then, by definition, 
    $Cg = 1_\Omega g$, and $\|Cg\|_2 =  \|1_\Omega  g\|_2 \leqslant \|g\|_2$. Therefore, 
    $|||C|||_2 \leqslant 1$.
    
    Let $m\in \mathbb N^\star$. Then, since $C_m: L^2([-2L,2L]^d, \mathbb C) \to L^2([-2L,2L]^d, \mathbb C)$ is a positive Hermitian compact operator, Theorem \ref{thm:minmax} states that 
    \begin{align*}
        \sigma_1^\downarrow( C_m) &= \underset{\|h\|_2=1}{\max_{h\in L^2([-2L,2L]^d, \mathbb C)}}  \langle h,  C_m h\rangle =\underset{\|h\|_2=1}{\max_{h\in L^2([-2L,2L]^d, \mathbb C)}} \|\Pi_m h\|_2^2 \leqslant \underset{\|h\|_2=1}{\max_{h\in L^2([-2L,2L]^d, \mathbb C)}} \|h\|_2^2=1.
    \end{align*}
    Since $\sigma_1^\downarrow( C_m^2) = \sigma_1^\downarrow( C_m)^2 \leqslant 1$, we deduce that
    \begin{align*}
        1 &\geqslant  \sigma_1^\downarrow( C_m)^2 = \underset{\|h\|_2=1}{\max_{h\in L^2([-2L,2L]^d, \mathbb C)}}  \langle h,  C_m^2 h\rangle =\underset{\|h\|_2=1}{\max_{h\in L^2([-2L,2L]^d, \mathbb C)}}  \|C_m h\|_2^2.
    \end{align*}
    This shows that $|||C_m|||_2\leqslant 1$.
\end{proof}

\subsection{Proof of Theorem \ref{thm:convergence_eff_dim}}

Note that if $H$ is a linear subspace of $H_m$, then $C_mH$ is also a subspace of $H_m$, and $\dim H \geqslant  \dim C_mH$. Therefore, 
    \begin{align*}
    \sigma_k^\downarrow(C_mM_m^{-1}C_m) &= \underset{\dim H = k}{\max_{H\subseteq H_m}} \;\underset{\|g\|_2=1}{\min_{g\in H }} \langle g, C_mM_m^{-1}C_m g\rangle\nonumber\\
    &= \underset{\dim H = k}{\max_{H\subseteq H_m}}\; \underset{\|g\|_2=1}{\min_{g\in H }} \langle (C_mg), M_m^{-1} (C_mg)\rangle\nonumber\\
    &\leqslant \underset{\dim H = k}{\max_{H\subseteq H_m}}\; \underset{\|g\|_2=1}{\min_{g\in H }} \langle g, M_m^{-1} g\rangle\nonumber\\
    &= \sigma_k^\downarrow(M_m^{-1})\nonumber\\
    &\leqslant \sigma_k^\downarrow(M^{-1}).
    \end{align*} 
Moreover, according to Lemma \ref{lem:operator_norm}, one has $|||M_m^{-1}-M^{-1}|||_2 \to 0$. Thus,  $\sup_{\|g\|_2=1}||(M_m^{-1}-M^{-1})(g)||_2 \to 0$. Using Lemma \ref{lem:C_op_norm}, we see that 
\begin{align*}
    &\|C_mM_m^{-1}C_mg - CM^{-1}Cg\|_2 \\
    &\leqslant  \|(C_m-C)M^{-1}Cg\|_2 +\|C_m(M_m^{-1}C_m - M^{-1}C)g \|_2 \\
    &\leqslant  |||(C_m-C)M^{-1}|||_2 +\|(M_m^{-1}C_m - M^{-1}C)g \|_2\\
    &\leqslant  |||(C_m-C)M^{-1}|||_2 +\|(M_m^{-1}-M^{-1})C_m g \|_2 + \|M^{-1}(C_m -C)g \|_2\\
    &\leqslant  |||(C_m-C)M^{-1}|||_2 +|||M_m^{-1}-M^{-1} |||_2 + |||M^{-1}(C_m -C)|||_2.
\end{align*}
Thus, \[|||C_mM_m^{-1}C_m - CM^{-1}C|||_2\leqslant |||(C_m-C)M^{-1}|||_2 +|||M_m^{-1}-M^{-1} |||_2 + |||M^{-1}(C_m -C)|||_2.\]
By diagonalizing $M^{-1}$ and using the facts that $\sum_{\ell\in\mathbb N^\star}\sigma^\downarrow_\ell(M^{-1}) < \infty$ and that $\lim_{m\to\infty}\|(C_m-C)f\|_2 = 0$ $\forall f \in L^2([-2L,2L]^d)$, it is easy to see that \[\lim_{m\to\infty} |||(C_m-C)M^{-1}|||_2 = \lim_{m\to\infty} |||M^{-1}(C_m-C)|||_2 = 0.\] Applying Lemma \ref{lem:operator_norm}, we deduce that 
\[\lim_{m\to\infty}|||C_mM_m^{-1}C_m - CM^{-1}C|||_2 = 0.\]
But, by Theorem \ref{thm:minmax},
\[\sigma_k^\downarrow(C_mM_m^{-1}C_m) = \underset{\dim H = k}{\max_{H\subseteq L^2([-2L,2L]^d, \mathbb C)}} \;\underset{\|g\|_2=1}{\min_{g\in H }} \langle g, C_mM_m^{-1}C_m g\rangle,\]
and 
\[\sigma_k^\downarrow(CM^{-1}C) = \underset{\dim H = k}{\max_{H\subseteq L^2([-2L,2L]^d, \mathbb C)}} \;\underset{\|g\|_2=1}{\min_{g\in H }} \langle g, CM^{-1}C g\rangle.\]
Clearly, for all $g\in L^2([-2L,2L]^d, \mathbb C)$,
\begin{align*}
|\langle g, CM^{-1}C g\rangle - \langle g, C_mM_m^{-1}C_m g\rangle| &= |\langle g, (CM^{-1}C-C_mM_m^{-1}C_m) g\rangle|\\&
\leqslant |||C_mM_m^{-1}C_m - CM^{-1}C|||_2.
\end{align*}
Therefore,
\[|\sigma_k^\downarrow(C_mM_m^{-1}C_m) - \sigma_k^\downarrow(CM^{-1}C)|\leqslant |||C_mM_m^{-1}C_m - CM^{-1}C|||_2,\]
and, in turn,
$\lim_{m\to\infty}\sigma_k^\downarrow(C_mM_m^{-1}C_m) = \sigma_k^\downarrow(CM^{-1}C)$.

To conclude the proof, observe that, on the one hand, 
\[
\frac{1}{1+\sigma_k^\downarrow(C_mM_m^{-1}C_m)^{-1}} = \frac{\sigma_k^\downarrow(C_mM_m^{-1}C_m)}{1+\sigma_k^\downarrow(C_mM_m^{-1}C_m)} \leqslant \sigma_k^\downarrow(C_mM_m^{-1}C_m) \leqslant \sigma_k^\downarrow(M^{-1}),
\] with $\sum_{k\in\mathbb N^\star} \sigma_k^\downarrow(M^{-1}) < \infty$.
On the other hand, 
\[\lim_{m\to\infty}\frac{1}{1+\sigma_k^\downarrow(C_mM_m^{-1}C_m)^{-1}} = \frac{1}{1+\sigma_k^\downarrow(CM^{-1}C)^{-1}},
\]
by continuity on $\mathbb R^+$ of the function $x\mapsto \frac{x}{1+x}$. Thus, applying the dominated convergence theorem, we are led to
\[\lim_{m\to\infty}\sum_{\lambda\in \sigma(C_mM_m^{-1}C_m)} \frac{1}{1+\lambda^{-1}} = \sum_{\lambda\in \sigma(CM^{-1}C)} \frac{1}{1+\lambda^{-1}}.\]

\section{Experiments}
\subsection{Numerical precision}
\begin{figure}
    \centering
    \includegraphics[scale=0.45]{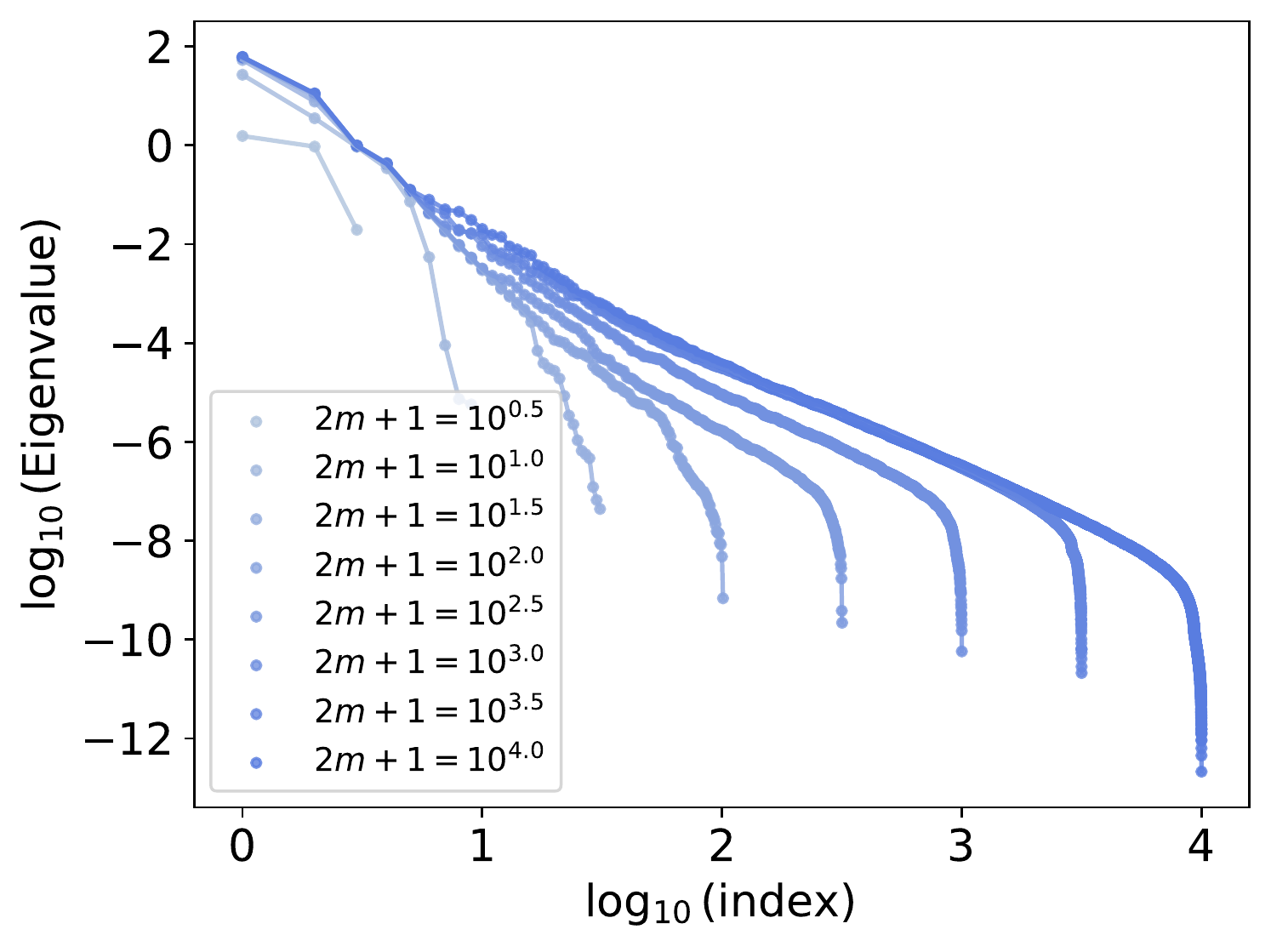}
    \includegraphics[scale=0.45]{figures/paper_3/oscillator_spectrum.pdf}
    \caption{Spectrum of $C_mM_m^{-1}C_m$. \textbf{Left:} \textit{Float32} precision. \textbf{Right:} \textit{Float64} precision.}
    \label{fig:float_precision}
\end{figure}

Enabling high numerical precision is crucial for efficient kernel inversion. Setting the default precision to \textit{Float32} and \textit{Complex64} can lead to significant numerical errors when approximating the kernel. For example, consider the harmonic oscillator case with $d=1$, $s=2$, and the operator $\mathscr D = \frac{d^2}{dx^2}u+\frac{d}{dx}u+u$, with $\lambda_n=0.01$ and $\mu_n=1$. Figure \ref{fig:float_precision} (left) shows the spectrum of $C_mM_m^{-1}C_m$ using \textit{Float32} precision, while Figure \ref{fig:float_precision} (right) shows the same spectrum with \textit{Float64} precision.
It is evident that with \textit{Float32}, the diagonalization results in lower eigenvalues compared to \textit{Float64}. 
In more physical terms, some energy of the matrix is lost when the last digits of the matrix coefficients are ignored. This leads to a problematic interpretation of the situation, as the \textit{Float64} estimation of the eigenvalues shows a clear convergence of the spectrum of $C_m M_m^{-1} C_m$, whereas the \textit{Float32} estimation appears to indicate divergence.

\subsection{Convergence of the effective dimension approximation}
\label{sec:eff_dim_m}
The PIKL algorithm relies on a Fourier approximation of the PIML kernel, as developed in Section~\ref{sec:finite_approx}. 
The precision of this approximation is determined by the number $m$ of Fourier modes used to compute the kernel.
However, determining an \textit{optimal} value of $m$ for a specific regression problem is challenging. There is a trade-off between the accuracy of the kernel estimation, which improves with higher values of $m$, and the computational complexity of the algorithm, which also increases with m. There is a trade-off between the accuracy of the kernel approximation, which improves with higher values of $m$, and the computational complexity of the algorithm, which also increases with $m$. 

An interesting tool to leverage here is the effective dimension, as it captures the underlying degrees of freedom of the PIML problem and, consequently, the precision of the method.  
Theorem~\ref{thm:convergence_eff_dim} states that the estimation of the effective dimension on $H_m$ converges to the effective dimension on $H^s(\Omega)$ as $m$ increases to infinity. Therefore, the smallest value $m^\star$ at which the effective dimension stabilizes is a strong candidate for balancing accuracy and computational complexity.

\begin{figure}
    \centering
    \includegraphics[width=0.5\linewidth]{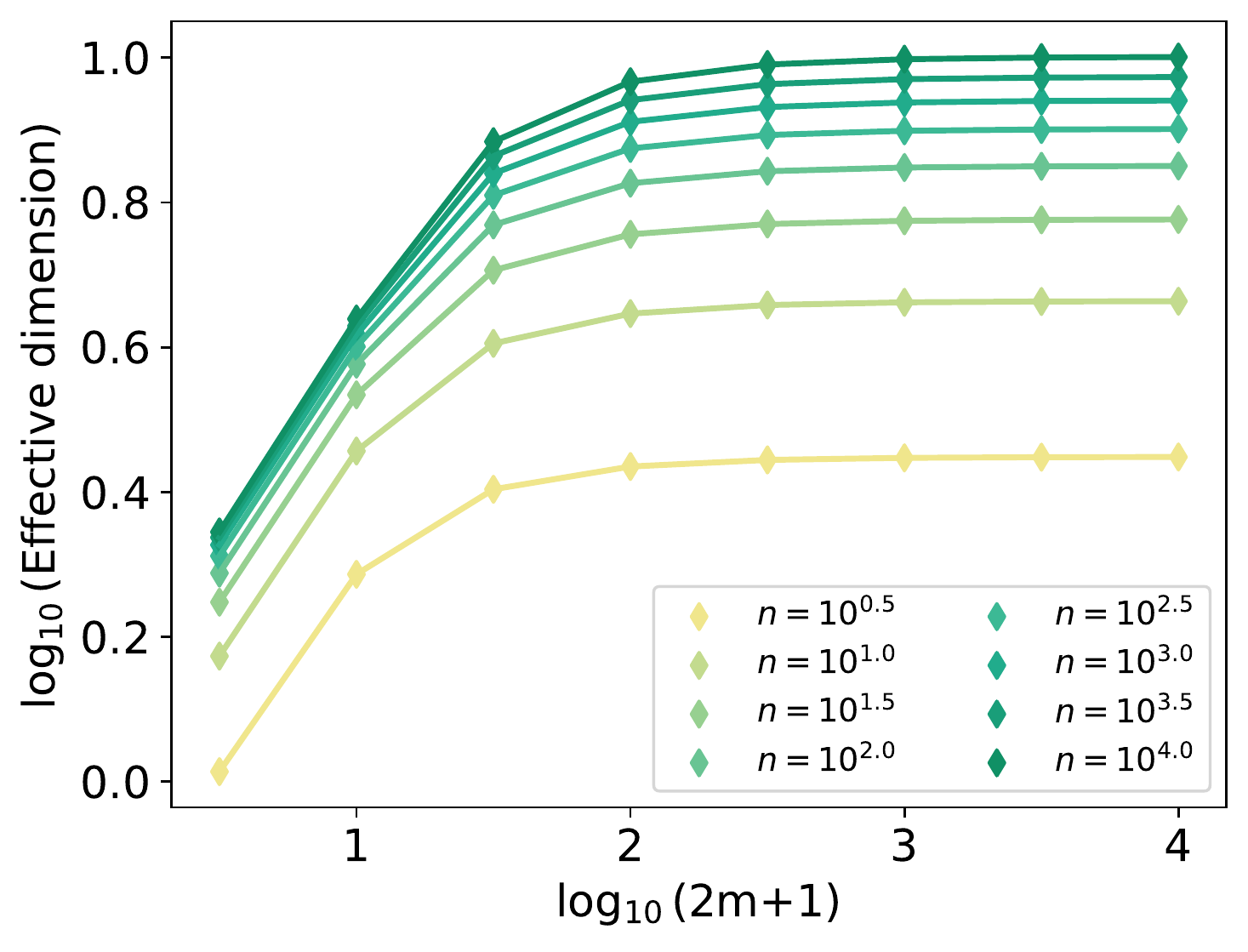}
    \caption{Convergence of the effective dimension as $m$ grows for $\mathscr D = \frac{d}{dx}$.}
    \label{fig:eff_dim_cv_1d}
\end{figure}

\begin{figure}
    \centering
    \includegraphics[width=0.5\linewidth]{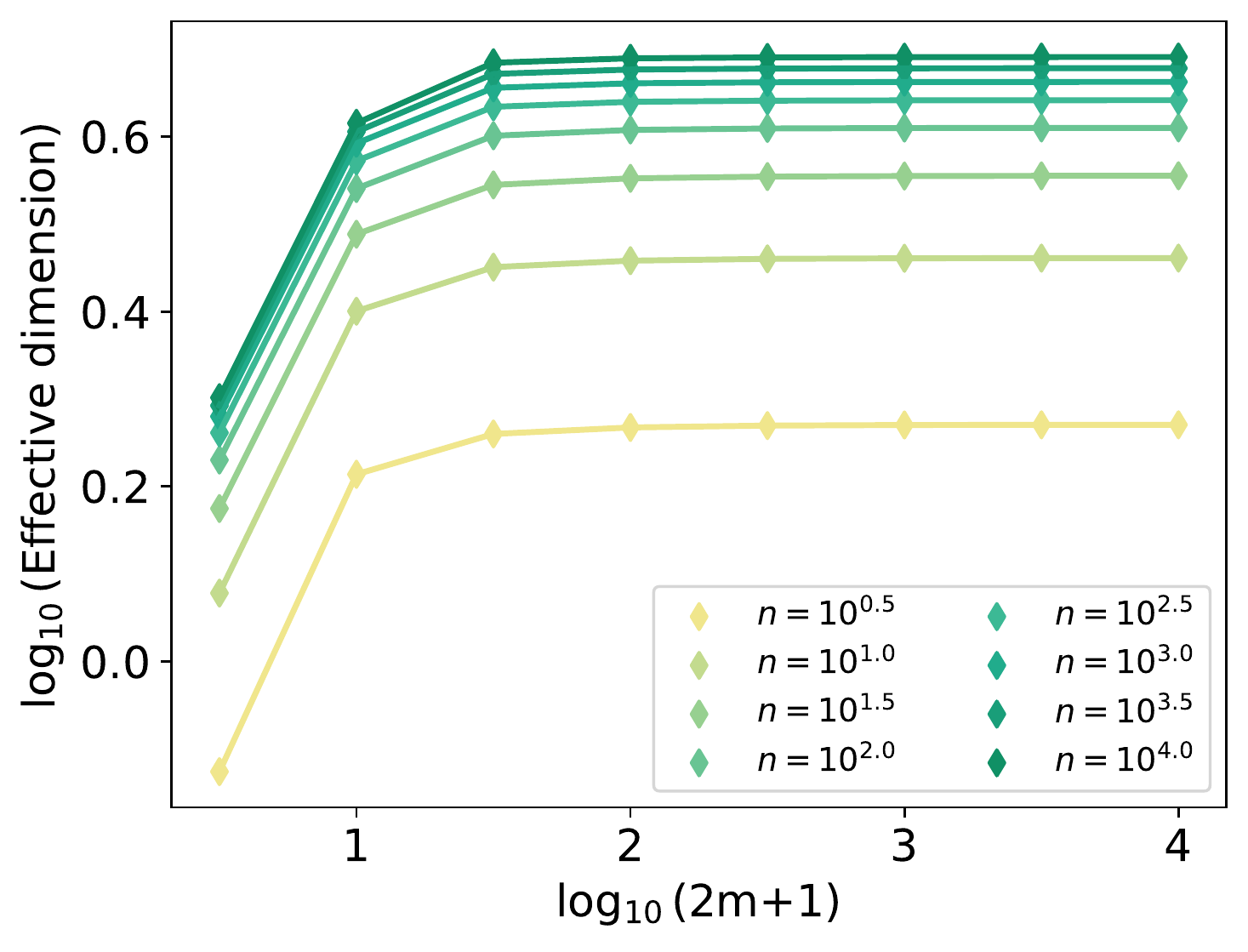}
    \caption{Convergence of the effective dimension as $m$ grows for the harmonic oscillator}
    \label{fig:eff_dim_cv_osc}
\end{figure}

\begin{figure}
    \centering
    \includegraphics[width=0.5\linewidth]{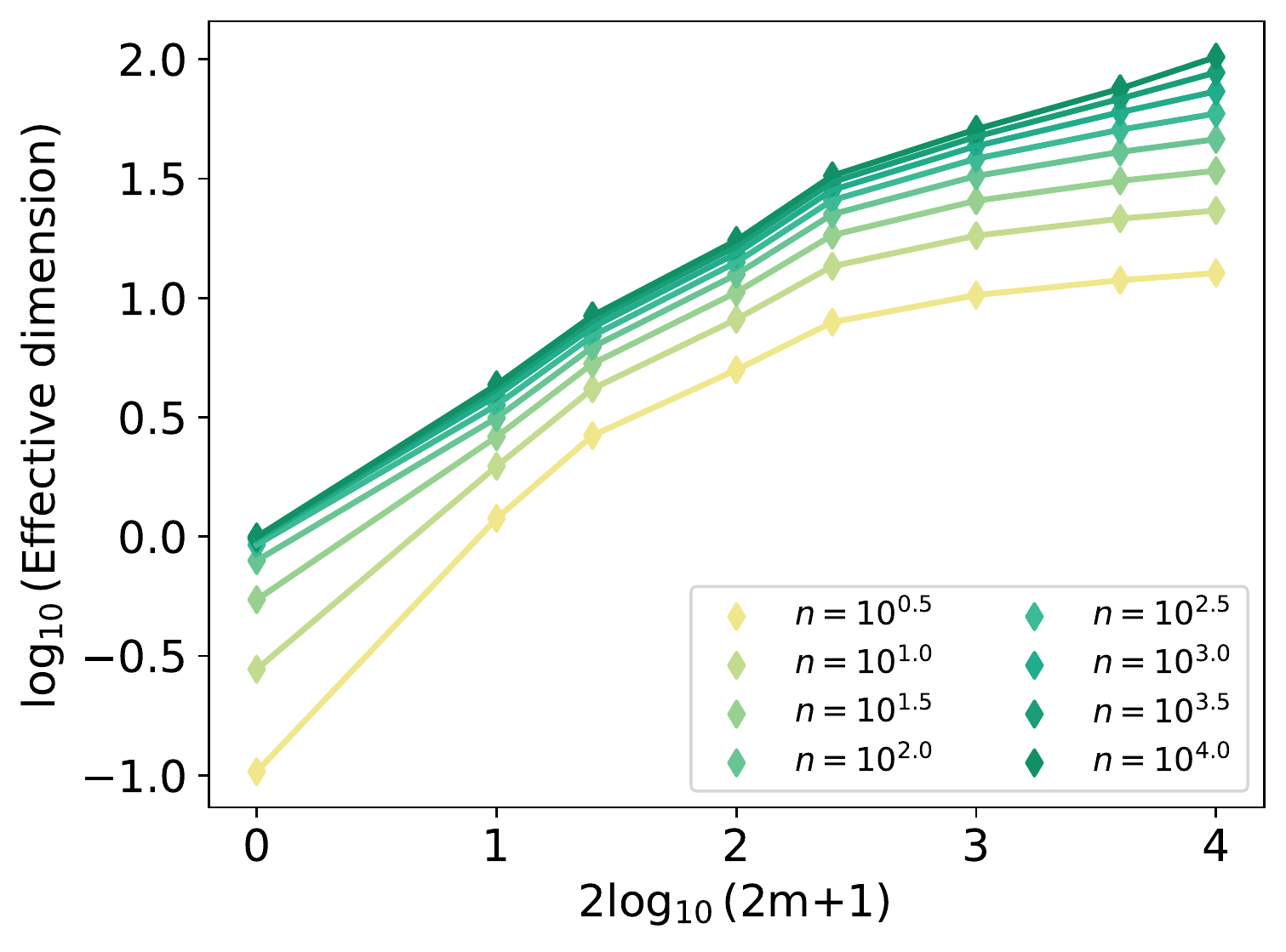}
    \caption{Convergence of the effective dimension as $m$ grows for the heat equation on the disk.}
    \label{fig:eff_dim_cv_heat_sm}
\end{figure}

Figures~\ref{fig:eff_dim_cv_1d}, \ref{fig:eff_dim_cv_osc}, and \ref{fig:eff_dim_cv_heat_sm} illustrate the convergence of the effective dimension estimation, using the eigenvalues of $C_m M_M^{-1}C_m$, as $m$ increases, for different values of $n$.
These figures provide insights into the PIK algorithm. As expected, the Fourier approximations converge more slowly as the dimension $d$ increases. Specifically, Figures ~\ref{fig:eff_dim_cv_1d} and \ref{fig:eff_dim_cv_osc} show that in dimension $d=1$, $m^\star \simeq 10^2$ for $n \leqslant 10^4$, while Figure~\ref{fig:eff_dim_cv_heat_sm} indicates that in dimension $d=2$, $m^\star \simeq 10^2$ for $n \leqslant 10^3$.

\subsection{Numerical schemes}
\label{sec:numerical}
We detail below the numerical schemes used as benchmarks in Section~\ref{sec:PDE_solving} for solving the wave equation. All these numerical schemes are constructed by discretizing the domain $\Omega = [0,1]^2$ into the grid $(\ell_1^{-1} \mathbb Z/\ell_1 \mathbb Z) \times (\ell_2^{-1} \mathbb Z/\ell_2 \mathbb Z)$. The initial and boundary conditions are then enforced on $n = 2\ell_1 + \ell_2$ points, with the approximation $\hat{f}_{n}$ defined accordingly as
\begin{itemize}
    \item for all $0\leqslant \ell \leqslant \ell_2$, $\hat f_{n}(0, \ell / \ell_2) = \sin(\pi \ell / \ell_2) + \sin(4 \pi \ell / \ell_2)/2$,
    \item for all $0\leqslant \ell \leqslant \ell_1$, $\hat f_{n}(\ell / \ell_1, 0) = 0$,
    \item for all $0\leqslant \ell \leqslant \ell_1$, $\hat f_{n}(\ell / \ell_1, 1) = 0$.
\end{itemize}
Let the discrete Laplacian $\Delta_{(\ell_1,\ell_2)}$ be defined for all $(a,b) \in \mathbb Z/\ell_1 \mathbb Z \times \mathbb Z/\ell_2 \mathbb Z$ by 
\[(\Delta_{(\ell_1,\ell_2)} \hat f_{n})(a/\ell_1, b / \ell_2) = \ell_2^{2}(\hat f_{n}(a/\ell_1, (b+1) / \ell_2) - 2\hat f_{n}(a/\ell_1, b / \ell_2) + \hat f_{n}(a/\ell_1, (b-1) / \ell_2)).\]
If $f^\star \in C^2([0,1]^2)$, its  Taylor expansion leads to $(\Delta_{(\ell_1,\ell_2)} f^\star)(a/\ell_1, b / \ell_2) = \partial^2_{x,x} f^\star(a/\ell_1, b / \ell_2) + o_{\ell_2 \to 0}(1)$.
Similarly, let the second-order time partial derivative operative $\partial^2_{t,t, (\ell_1,\ell_2)}$ be defined for all $(a,b) \in \mathbb Z/\ell_1 \mathbb Z \times \mathbb Z/\ell_2 \mathbb Z$ by \[(\partial^2_{t,t, (\ell_1,\ell_2)} \hat f_{n})(a/\ell_1, b / \ell_2) = \ell_2^{2}(\hat f_{n}((a+1)/\ell_1, b / \ell_2) - 2\hat f_{n}(a/\ell_1, b / \ell_2) + \hat f_{n}((a-1)/\ell_1, b / \ell_2)).\]
\paragraph{Euler explicit.} The Euler explicit scheme is initialized using the Taylor expansion $f(t, x) = f(0, x) + t \partial_t f(0, x) + t^2 \partial^2_{t,t} f(0, x)/2 + o_{t\to 0}(t^2)$. With the initial condition $\partial_t f(0, x) = 0$ and the wave equation $\partial^2_{t,t} f(0, x) = 4 \partial^2_{x,x}f(0,x)$, this simplifies to $f(t, x) = f(0, x) +  2 t^2 \partial^2_{x,x} f(0, x) + o_{t\to 0}(t^2)$. This leads to the initialization 
\[\forall 0\leqslant b \leqslant \ell_2, \quad \hat f_{n}(1/\ell_1, b / \ell_2) = \hat f_{n}(0, b / \ell_2) + 2\ell_1^{-2}(\Delta_{(\ell_1,\ell_2)} \hat f_{n})(0, b / \ell_2).\]
The wave equation $\partial^2_{t,t} f^\star = 4 \partial^2_{x,x} f^\star$ can then be discretized as $\partial^2_{t,t, (\ell_1,\ell_2)} \hat f_n = 4 \Delta_{(\ell_1,\ell_2)} \hat f_n$. This leads to the explicit Euler recursive formula
\[\hat f_{n}((a+1)/\ell_1, b / \ell_2) = 2\hat f_{n}(a/\ell_1, b / \ell_2) - \hat f_{n}((a-1)/\ell_1, b / \ell_2) + 4\ell_1^{-2} (\Delta_{(\ell_1,\ell_2)} \hat f_{n})(a/\ell_1, b / \ell_2).\]
This formula allows to compute $\hat f_{n}((a+1)/\ell_1, \cdot)$ given the values of $\hat f_{n}(0, \cdot), \hdots, \hat f_{n}(a/\ell_1, \cdot)$.
\paragraph{Runge-Kutta 4.} The RK4 scheme is a numerical scheme applied on both $f^\star$ and its derivative $\partial_t f^\star$. Here, $\hat g_n$ represents the approximation of $\partial_t f^\star$. The initial condition $\partial_t f(0, \cdot)=0$ translates into 
\[\forall 0\leqslant b \leqslant \ell_2, \quad \hat g_{n}(0, b / \ell_2) = 0.\]
To infer $\hat f_{n}((a+1)/\ell_1, \cdot)$ and $\hat g_{n}((a+1)/\ell_1, \cdot)$ given the values of $\hat f_{n}(0, \cdot), \hdots, \hat f_{n}(a/\ell_1, \cdot)$ and $\hat g_{n}(0, \cdot), \hdots, \hat g_{n}(a/\ell_1, \cdot)$, the RK4 scheme introduces intermediate estimates as follows:
\begin{itemize}
    \item $\tilde f_1 = \hat g_{n}(a/\ell_1, \cdot) /\ell_1$,
    \item $\tilde g_1 = 4 (\Delta_{(\ell_1,\ell_2)} \hat f_{n})(a/\ell_1, \cdot)/\ell_1$,
    \item $\tilde f_2 = (\hat g_{n}(a/\ell_1, \cdot) +0.5 \tilde f_1)/\ell_1$,
    \item $\tilde g_2 = 4 (0.5 \tilde g_1 + (\Delta_{(\ell_1,\ell_2)} \hat f_{n})(a/\ell_1, \cdot))/\ell_1$,
    \item $\tilde f_3 = (\hat g_{n}(a/\ell_1, \cdot) +0.5 \tilde f_2)/\ell_1$,
    \item $\tilde g_3 = 4 (0.5 \tilde g_2 + (\Delta_{(\ell_1,\ell_2)} \hat f_{n})(a/\ell_1, \cdot))/\ell_1$,
    \item $\tilde f_4 = (\hat g_{n}(a/\ell_1, \cdot) + \tilde f_3)/\ell_1$,
    \item $\tilde g_4 = 4 (\tilde g_3 + (\Delta_{(\ell_1,\ell_2)} \hat f_{n})(a/\ell_1, \cdot))/\ell_1$,
    \item $\hat f_{n}((a+1)/\ell_1, \cdot) = \hat f_{n}(a/\ell_1, \cdot) + (\tilde f_1+2\tilde f_2+2\tilde f_3+\tilde f_4)/6$,
    \item $\hat g_{n}((a+1)/\ell_1, \cdot) = \hat g_{n}(a/\ell_1, \cdot) + (\tilde g_1+2\tilde g_2+2\tilde g_3+\tilde g_4)/6$.
\end{itemize}
Similarly to the Euler explicit scheme, the RK4 relies on a recursive formulas to compute $\hat f_n$.

\paragraph{Crank-Nicolson}. The CN scheme is an implicit scheme defined as follows. Similar to the Euler explicit scheme, the $\partial_t f(0, \cdot) = 0$ initial condition is implemented as
 \[\forall 0\leqslant \ell \leqslant \ell_2, \quad \hat f_{n}(1/\ell_1, \ell / \ell_2) = \hat f_{n}(0, \ell / \ell_2) + 2\ell_1^{-2}(\Delta_{(\ell_1,\ell_2)} \hat f_{n})(0, \ell / \ell_2).\]
 Then, the recursive formula of this scheme takes the form 
 \[\partial^2_{t,t, (\ell_1,\ell_2)} \hat f_n(a/\ell_1, \cdot) = 2 ((\Delta_{(\ell_1,\ell_2)} \hat f_n)(a/\ell_1, \cdot) + (\Delta_{(\ell_1,\ell_2)} \hat f_n)((a+1)/\ell_1, \cdot)).\]
 This leads to the recursion
 \[\hat f_n((a+1)/\ell_1, \cdot) = (\mathrm{Id} + 2\ell_2^2\ell_1^{-2} \Delta)^{-1}(3 \hat f_n(a/\ell_1, \cdot)- \hat f_n((a-1)/\ell_1, \cdot)) - \hat f_n((a-1)/\ell_1, \cdot),\]
 where $\Delta = \begin{pmatrix}
     2 & -1 & 0 & \hdots & 0\\
     -1 & 2 & -1 & \ddots & \vdots\\
     0 & -1& 2 & \ddots &  0\\
     \vdots & \ddots & \ddots & \ddots & -1 \\
     0 & \hdots & 0 & -1 & 2& 
 \end{pmatrix}$ is the discrete Laplacian matrix.
 
 \subsection{PINN training}
\begin{figure}
    \centering
    \includegraphics[width=\linewidth]{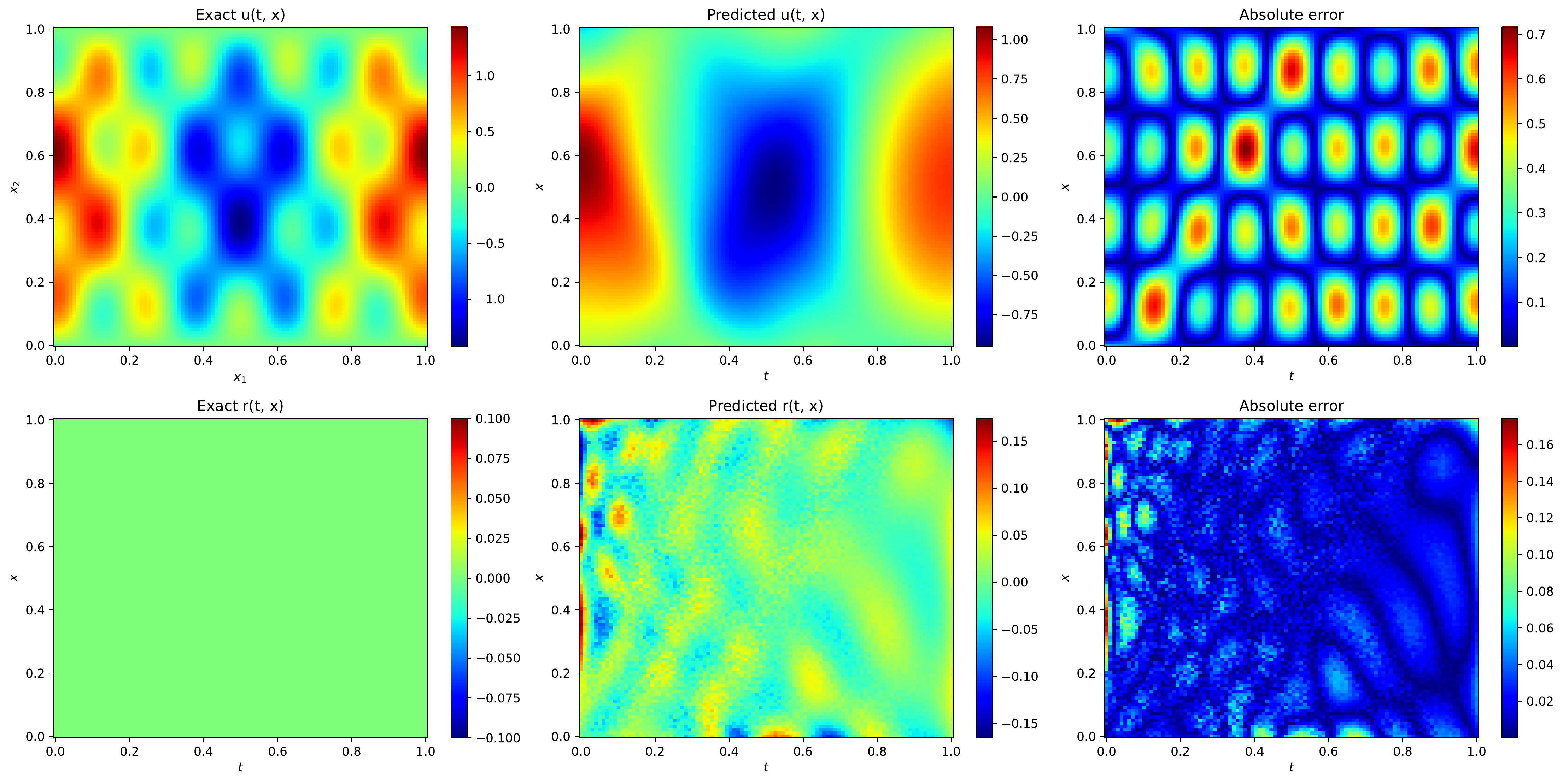}
    \caption{\textbf{Left}: Ground truth. \textbf{Middle}: PINN estimator. \textbf{Right}: Error $=$ PINN - Ground truth.}
    \label{fig:nn_noise}
\end{figure}

\begin{figure}
    \centering
    
    \includegraphics[width=0.7\linewidth]{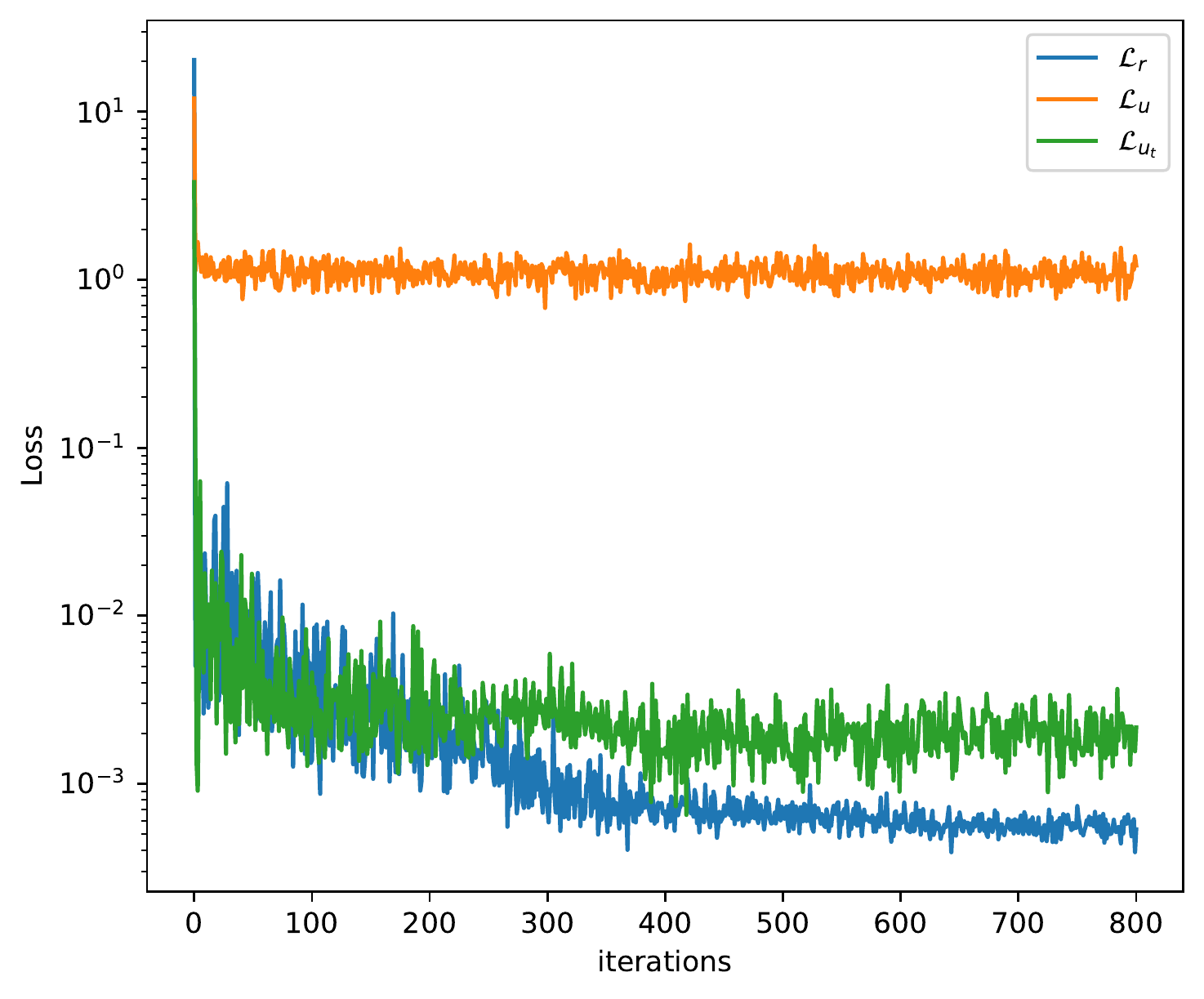}
    \caption{PINN training with noisy boundary conditions.}
    \label{fig:PINN_BC}
\end{figure}

Figures~\ref{fig:nn_noise} and \ref{fig:PINN_BC} illustrate the performance of the PINNs during training while solving the 1d wave equation with noisy boundary conditions.


\renewcommand\thesection{\thechapter.\arabic{section}}

\part{Time series forecasting in atypical periods}
\label{part:part2}

\chapter{Forecasting Electric Vehicle Charging Station Occupancy:
Smarter Mobility Data Challenge}
\label{ch:contrib-1}
This chapter corresponds to the following publication: \citet{amara-ouali2024forecasting}.

\section{Introduction}

\label{intro}

\paragraph{Electric mobility} The transportation sector is currently one of the main contributors to greenhouse gas emissions in Europe \citep{EEA_2022}. To reduce these emissions, an interesting avenue has been to foster the development of EVs. In 2021, China led global EV sales with 3.3 million units, tripling its 2020 sales, followed by Europe with 2.3 million units, up from 1.4 million in 2020 \citep{IEA_EV_2022}. The U.S. market share of electric vehicles doubled to 4.5\%, with 630,000 units sold. Meanwhile, electric vehicle sales in emerging markets more than doubled \citep{IEA_EV_2022}. As a consequence, electric mobility development entails new needs for energy providers and consumers \citep{rte2022}. Companies and researchers are proposing a large amount of innovative solutions including pricing strategies and smart charging \citep{dallinger2012grid, wang2016smart, alizadeh2017optimal, Moghaddam2018smart, crozier2020a} to couple it with renewable production \cite{hafeez2023utilization}.  
However, their implementation requires a precise understanding of charging behaviours and better EV charging models are necessary to grasp the impact of EVs on the grid \citep{Gopalakrishnan2016DemandPA, kaya2022electric, ciociola2023data, Andrenacci2023}. 
In particular, forecasting the occupancy of a charging station can be a critical need for utilities to optimise their production units according to charging demand \citep{ZHANG2023}. On the user side, knowing when and where a charging station will be available is critical, but large-scale datasets on EVs are rare \citep{calearo2021a, amara-ouali2021a}.  

\paragraph{Summary of the challenge} This article presents the Smarter Mobility Data Challenge, which aims at testing statistical and machine learning forecasting models to predict the states of a set of charging stations in the Paris area at different geographical resolutions. This challenge was held from October, 3rd 2022 to December 5th, 2022 on the CodaLab platform \url{https://codalab.lisn.upsaclay.fr/competitions/7192}. It was organised by the \textit{Manifeste IA}, a network of 16 French industrials and TAILOR, a European project which aims to provide the scientific foundations for Trustworthy AI. It has been pioneered following the ‘AI for Humanity’ French government plan launched in 2019. The challenge gathered 169 participants and was open to students from the EU. The authors (except the participants of the challenge) have collected and prepared the dataset, and organised the data challenge.

\paragraph{Time series models} Forecasting time series data is essential for businesses and governments to make informed decisions. However, the temporal structure in time series comes with specific challenges, such as non-stationarity and missing values.
This is why, in addition to standard machine learning models, a wide range of models have been tailored for time series. These include auto-regressive models \citep{box2015time}, tree-based models \citep{friedman2001greedy}, and deep learning models such as recurrent neural networks \citep{jordan1997serial, hochreiter1997long}, temporal convolutional networks \citep{bai2018empirical} and transformers \citep{wen2022transformers}. 
However, no one model has proven to be better than the others at predicting time series.
On the one hand, although deep learning models are known to perform well with large datasets, it is still unclear how they compare to other models on small datasets, how they handle non-stationary data or how they deal with with exogenous information \citep{zeng2023transformers, Tayal2024}. In fact, modern machine learning models still struggle to deal with missing values and time-dependent patterns such as trends or breaks.
On the other hand, tree-based models such as gradient-boosted trees are known to perform well on tabular data \citep{NEURIPS2023_f06d5ebd}, and to sometimes outperform complex deep learning models \citep{MAKRIDAKIS20221346}. Therefore, practical insights from datasets and benchmarks are valuable \citep{petropoulos2022forecasting}. In particular, a recent comprehensive  benchmark  \citep{godahewa2021monash} has regrouped 26 time series datasets on various domains, including energy and transport, taken from challenges \citep[see, e.g.,][]{makridakis2022m5} and the public domain. Other works have proposed synthetic datasets to evaluate specific properties of forecast algorithms, such as interpretability \citep{ismail2020benchmarking}, outlier detection \citep{Lai2021RevisitingTS}, and forecast performance \citep{Kang2020GRATIS}. 

\paragraph{Hierarchical forecasting}
The data of the Smarter Mobility Data Challenge has a hierarchical structure because it EV charging stations can be regrouped at different scales (stations, areas, and global). Hierarchical time series forecasting has been studied on various other applications where the data is directly or indirectly hierarchically organised. For example, in the retail industry, goods are often classified into categories (such as food or clothing) and inventory management can be done at different geographical (national, regional, shop) or temporal (week, month, season) scales. Moreover, electricity systems often have an explicit (electricity network) or implicit (e.g., customer types, tariff options) hierarchy. Recent work shows that exploiting this structure can improve forecasting performance at different levels of hierarchy. For instance, \cite{hyndman2011optimal} focuses on tourism demand,  \cite{Athanasopoulos2019HierarchicalF} on macroeconomic forecasting, and \cite{Hong2019GlobalEF, Brgre2020OnlineHF, Taieb2020HierarchicalPF, nespoli2022multivariate} on electricity consumption data.

\paragraph{Related works} Similar to energy and transport forecasting, EV demand forecasting has received a lot of attention. The survey by \citet{amara-ouali2022a} compares the classical time series methods, the statistical models, the machine learning methods and the deep learning methods that have been used to capture the temporal dependencies in EV charging data. Overall, it shows that both tree-based models and deep learning models are able to capture the complex non-linear temporal relationships in EV charging data.
More recently, \citet{ma2022multistep} proposed a hybrid LSTM model that outperformed classical machine learning approaches (support vector machine, random forest, and Adaboost) and other deep learning architectures (LSTM, Bi-LSTM, and GRU)  in forecasting the occupancy of 9 fast chargers in the city of Dundee.
\citet{wang2023predicting} have investigated the use of spatial correlations to predict EV charging behaviour. They proposed a spatio-temporal graph  convolutional network incorporating both geographical and temporal dependencies to predict the short-term charging demand in Beijing  using a dataset of 76774 private EVs. 
However, such individual data is expensive and often kept private, and \citet{wang2023predicting} only had access to data for the month of January 2018. 
In fact, although datasets describing the development of EV infrastructures are common \cite[see, e.g.,][]{falchetta2021electric, yi2022electric}, fewer datasets document the actual use of EVs and they are often of lower spatial resolution \cite[see, e.g.,][]{lee2019acn}.
In fact, open datasets at the scale of individual stations, such as the one presented in this article, are still very rare \citep{amara-ouali2021a}. 
Such so-called EVSE-centric (for Electric Vehicle Supply Equipment) datasets are more informative and hierarchical forecasting could be useful for users and operators interested in specific EV stations. 
However, even with EV datasets spanning multiple years, ruptures are common and models require specific adjustments \cite[see, e.g.,][]{koohfar2023prediction}.

\paragraph{Main Contributions}
The main contributions of the paper can be summarised as follows:
\vspace{1em}
\begin{enumerate}[nosep]
    \item An open dataset on electric vehicle behaviors gathering both spatial and hierarchical features, available at   \url{https://gitlab.com/smarter-mobility-data-challenge/additional_materials}. Datasets with such features are rare and valuable for electric network management.
    \item An in-depth descriptive analysis of this dataset revealing meaningful user behaviors (work behaviors, daily and weekly patterns...). 
    \item A detailed and reproducible benchmark for forecasting the EV charging station occupancy. This benchmark compares the winning solutions of a data challenge and state-of-the-art predictive models.
\end{enumerate}

\paragraph{Overview}

The paper is structured as follows. Section \ref{EV_charging} describes the dataset. Section \ref{sec:problem} details the forecasting problem at hand and baseline models. Section \ref{winning} presents the methods proposed by the three winning teams. Finally, Section \ref{summary} summarizes the findings and discusses our results. The full dataset, baseline models, winning solutions, and aggregations, are available at \href{https://gitlab.com/smarter-mobility-data-challenge/tutorials}{https://gitlab.com/smarter-mobility-data-challenge/tutorials} and distributed under the Open Database License (ODbL). 
A supplementary material presents the Belib's pricing and park history in Section 1, a detailed data description (collection, preprocessing, explanatory data analysis) in Section 2, some complements about the winning strategies of the challenge in Section 3, future perspectives about new datasets and benchmarks in Section 4 and a Datasheet in Section 5.

\section{EV charging dataset}
\label{EV_charging}

In this section we present how the raw dataset was collected and how it was then preprocessed to make it suitable for the data challenge. 

\paragraph{General description}
\label{sec:dataset}
The dataset is based on the real-time charging station occupancy information of the Belib network, available on the Paris Data platform (ODbL) \citep{paris_data}. The Belib network was composed of 91 charging stations in Paris at the time of the challenge, each offering 3 plugs for a total of 273 charging points. A process to store the data was initiated by the EDF R\&D team since daily data was not stored by Paris Data. A pipeline was set up to collect this data every 15 minutes, starting July 2020, on the platform's dedicated API \footnote{\href{https://parisdata.opendatasoft.com/explore/dataset/belib-points-de-recharge-pour-vehicules-electriques-disponibilite-temps-reel/api}{parisdata.opendatasoft.com/explore/dataset/belib-points-de-recharge-pour-vehicules-electriques-disponibilite-temps-reel/api}}. The data was then stored in a data lake based on Hadoop technologies (HDFS, PySpark, Hive, and Zeppelin).
The storage of this information over time allows, for example, to estimate the usage of the charging stations depending on their location.

\paragraph{Belib's history: pricing mechanism and park evolution}
\label{par:history}
89\% of EV users living in a house mainly charge their vehicle at home, compared to only 54\% of EV users living in an apartment in 2020 \citep{2021_ENEDIS_VE}. Paris is a very dense city that allows limited access to private residential charging points, hence the need for public charging stations. The first 5 stations of the Belib network were commissioned on 12 January, 2016 \citep{2016automobileBornes, 2016stageBornes}. The network grew progressively in 2016 to reach 60 stations all around Paris. Users needed to buy a 15 euro badge to connect to the network. Different pricing strategies were applied depending on the time of the day and plugs.
The "normal charge" of 3kW was free at night (between 8 p.m. and 8 a.m.) and cost 1 euro per hour on daytime (between 8 a.m. and 8 p.m.).
The "quick charge" of 22kW cost 25 cents every 15 minutes during the first hour of charge. After the first hour, the first 15 minutes cost 2 euros. After this 1h and 15 minutes, each 15 minutes cost 4 euros.
Each station contained 3 parking spots:
\begin{itemize}
    \item one dedicated to "normal charge" with an E/F electric plug,
    \item one dedicated to "quick charge" with a ChaDeMo and a Combo2 plugs,
    \item one where both "normal charge" and "quick charge" were possible, with an E/F, a T2, and a T3 plugs.
\end{itemize}
The pricing strategy was intended to allow the usage of "normal charge" plugs as a free parking spot overnight, while "quick charge" became expensive after one hour of usage.

In 2021, the city of Paris allowed the  company TotalEnergies to run the Belib network for a period of 10 years. The goal is to enhance the network, increasing from 90 stations and 270 charging points, to 2300 charging points \citep{2021totalVehicules, 2021leparisienLe}. We elaborate on the new pricing mechanism in the supplementary material.

\paragraph{Data preprocessing} In the raw data, each observation reflects the status of the plugs (up to 6) within a charging point. The structure of the raw dataset is misleading as only one of these plugs can be in use at a time. Therefore, we processed the dataset to only keep the relevant rows, i.e., the rows containing the plugs in use, and we treated a charging point as a single plug. In addition, charging points are clustered in groups of three according to their geographic location in the raw data. This  charging point structure was confirmed by the data provider. We grouped the three adjacent charging points into a single charging station and aggregated the data accordingly. To account for differences in timestamp synchronization between stations, we have adjusted timestamps to match the nearest 15-minute interval. The \textit{available}, \textit{charging}, and \textit{passive} states are taken directly from the raw data. The last state \textit{other} regroups several statuses including \textit{reserved} (a user has booked the charging point), \textit{offline} (the charging point is not able to send information to the server), and \textit{out of order} (the charging point is out of order). We made this choice because of the relatively small number of \textit{reserved} and \textit{out of order} records. This way, the \textit{other} state could be interpreted as a noisy version of the \textit{offline} state. Missing timestamps in the dataset have not been filled so there is room for missing data imputation techniques.

\begin{figure}
    \centering
    \includegraphics[width=0.35\linewidth]{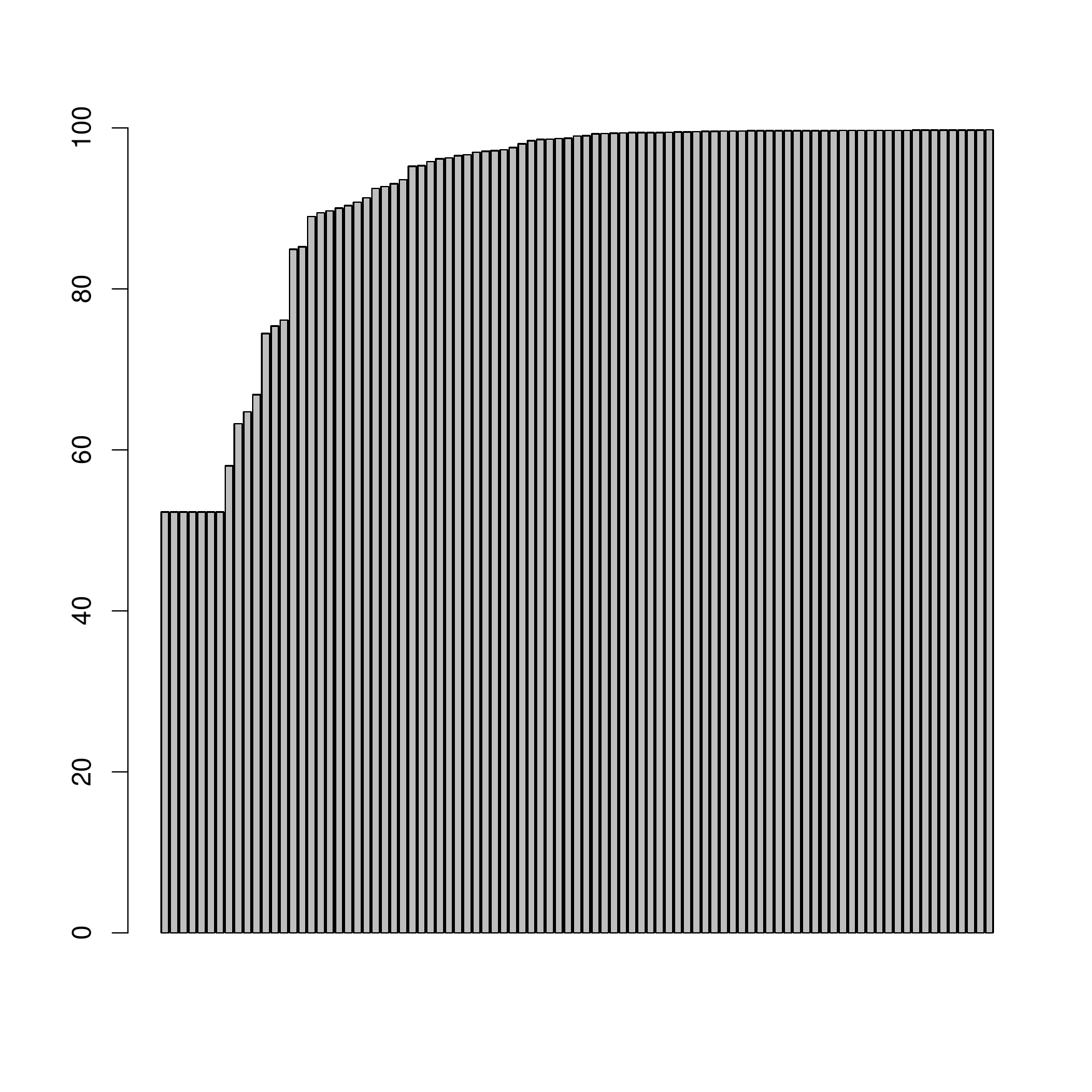}
    \includegraphics[width=0.35\linewidth]{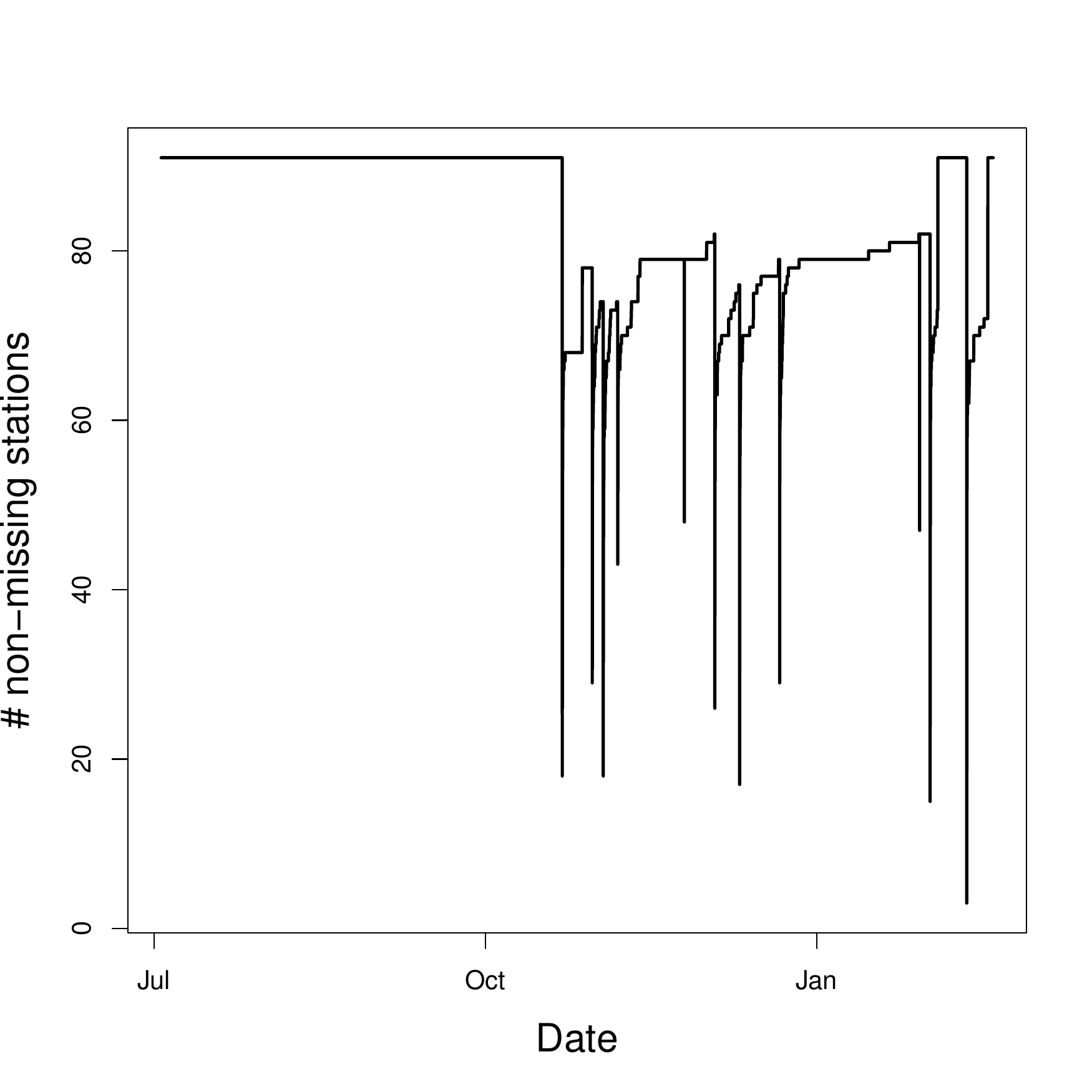}
    \caption{Left: Percentage of non missing observations per station. Right: Number of non missing stations in function of time on the train set.}
    \label{nb_observations_per_station}
\end{figure}

\paragraph{Missing values} There is a significant number of missing values in the data. To illustrate, the records of the following five days are very incomplete as they contain less than 96 observations for all the 91 stations: 2020-08-06 (with 92 data points), 2020-10-27 (95), 2020-11-20 (54), 2020-12-29 (95), and 2021-01-04 (95). 
The distribution of missing values is highly station-dependent, as illustrated in Figure~\ref{nb_observations_per_station}. We note that half of the stations have almost no missing data (except for the five days documented above), whereas 7 stations have around 50\% missing observations. This suggests that malfunctioning behaviors are specific to some stations and could be learned.
In addition, the number of non-missing stations is depicted in Figure~\ref{nb_observations_per_station}. Note that we excluded timestamps from the plot when all stations were missing. We also note that the number of missing stations starts to fluctuate a lot after October.

\begin{figure}
    \centering
    \begin{tabular}{c|c}
    \includegraphics[width=0.45\linewidth]{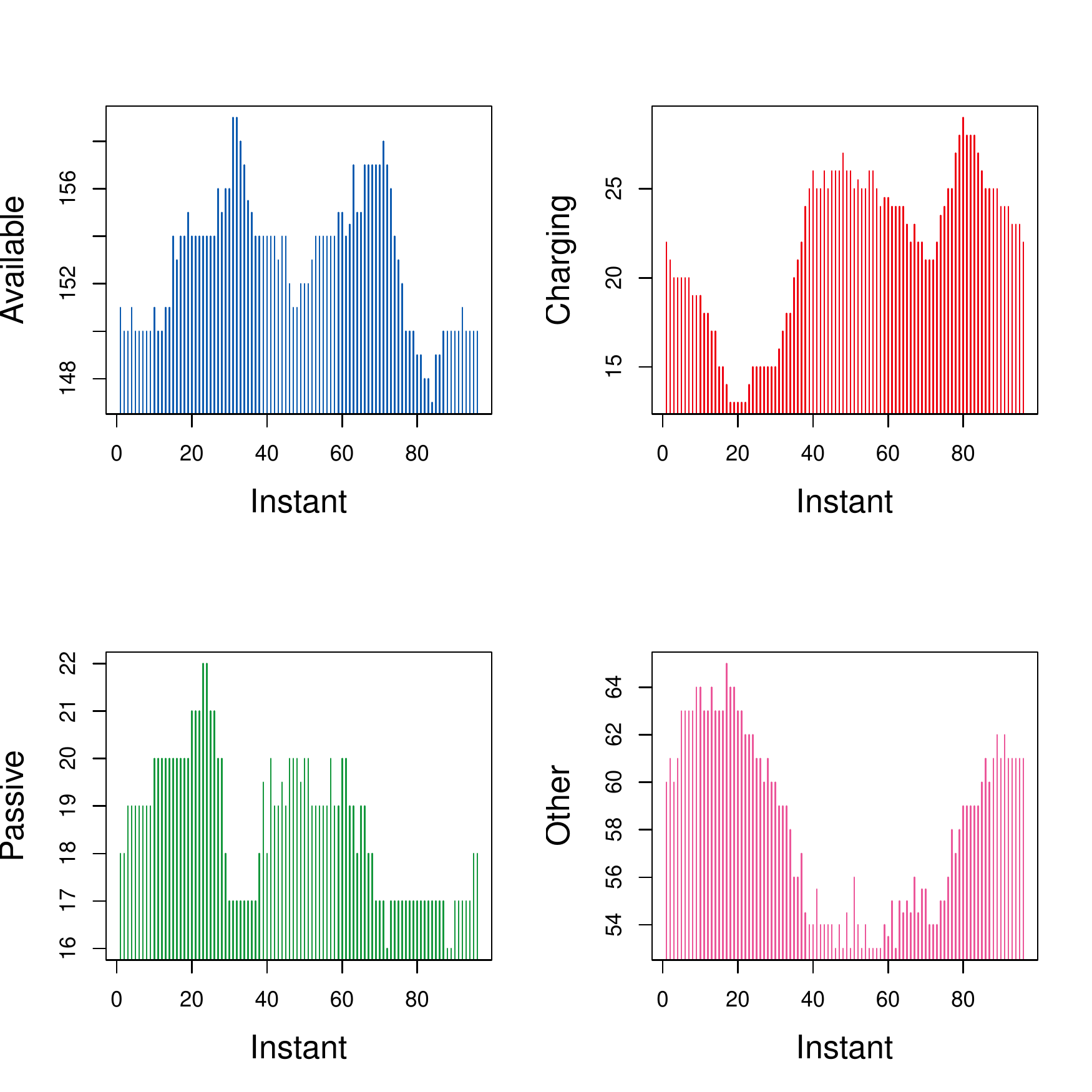} &
    \includegraphics[width=0.45\linewidth]{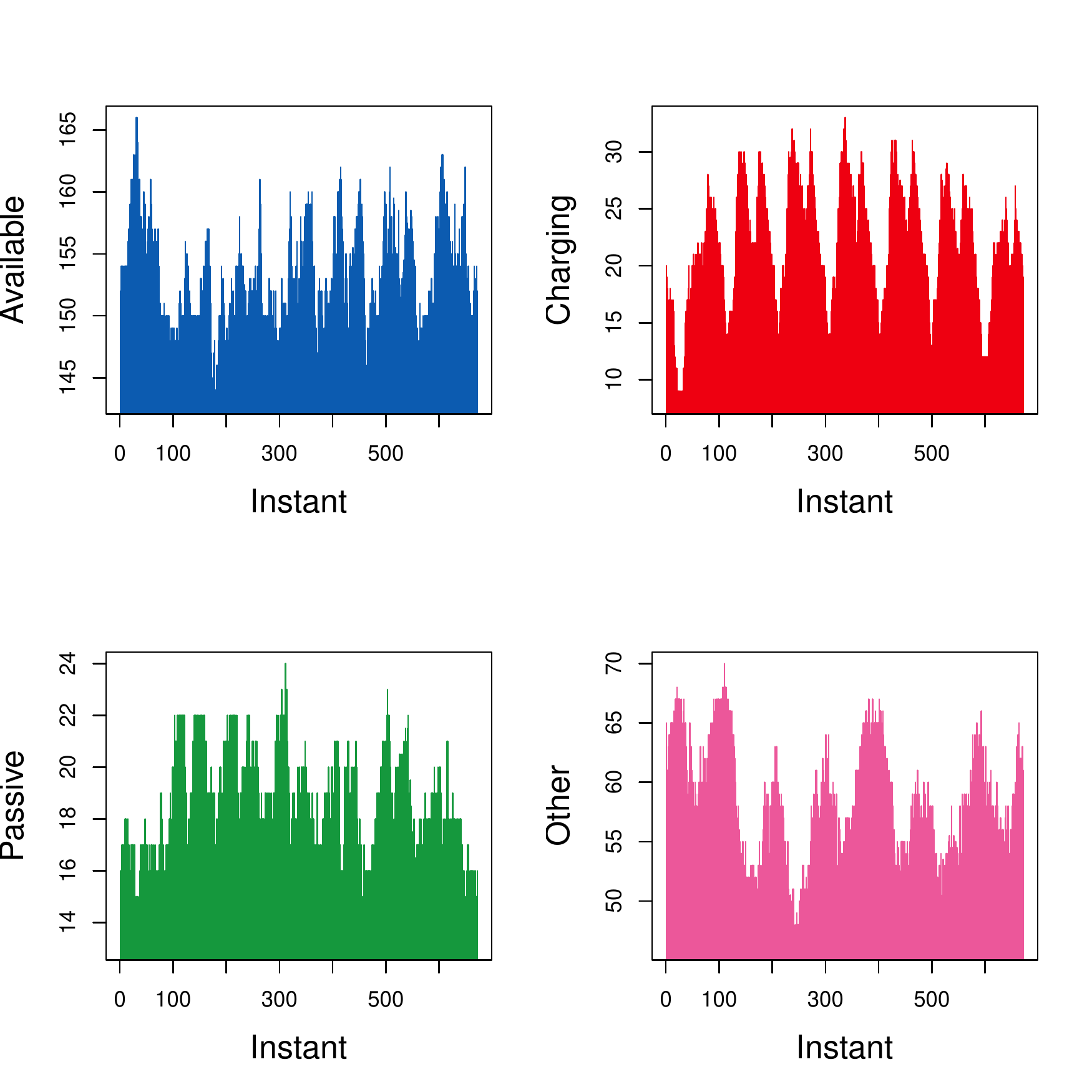}
    \end{tabular}
    \caption{Daily (left) and Weekly (right) profiles for each status at the Global level.}
    \label{calendar_profile}
\end{figure}

\paragraph{Exploratory Data Analysis}
\label{EDA}
We show daily and weekly profiles with the median number of plugs as a function of time (an instant corresponding to a 15-minute interval) per status at the Global level on Figure~\ref{calendar_profile}. From these graphs, We observe the presence of a daily pattern in the data and a change in the pattern between weekdays and weekends.
What we observe in Figure~\ref{calendar_profile} matches with the pricing strategy used from 2016 to 2021, detailed in Paragraph \ref{par:history}.
The pricing changed twice a day: at 8 a.m. and 8 p.m. At night, the free "normal charge regime" (7kW) explains the peak in \textit{charging} states at instant 80 (corresponding to 8 p.m.) and the drop of \textit{available} at the same hour.
This "normal charge" mode provides a low electrical power, hence the slowness of charging. 
Therefore, overnight, as EV batteries become fully charged, the number of \textit{charging} states decrease in favour of the number of \textit{passive} states.
The proportion of \textit{other} states is more important at night, mainly because there is more maintenance jobs at night.
Since users tend to be repelled by or ignore malfunctioning stations it explains why this excess in \textit{other} comes with a slight increase of \textit{available} at night. On the other hand, the price increase, after 8 a.m., induces a decrease of \textit{passive} spots from 7p.m. to 9 p.m. and an increase of \textit{available} (drivers parking on regular parking spots) and \textit{charging} spots (drivers  charging their car in front of their office after the morning ride).
This analysis is consistent with the weekly scale in Figure~\ref{calendar_profile}.
The number of \textit{charging} stations is greater during work days, while the number of \textit{available} stations is greater during week-ends, reflecting commuting behaviors. 
We note that the daily peaks at 8 a.m. and 8 p.m. are pronounced on the weekly \textit{charging} profile.

\begin{figure}[ht]
    \centering
    \includegraphics[width=0.6\linewidth]{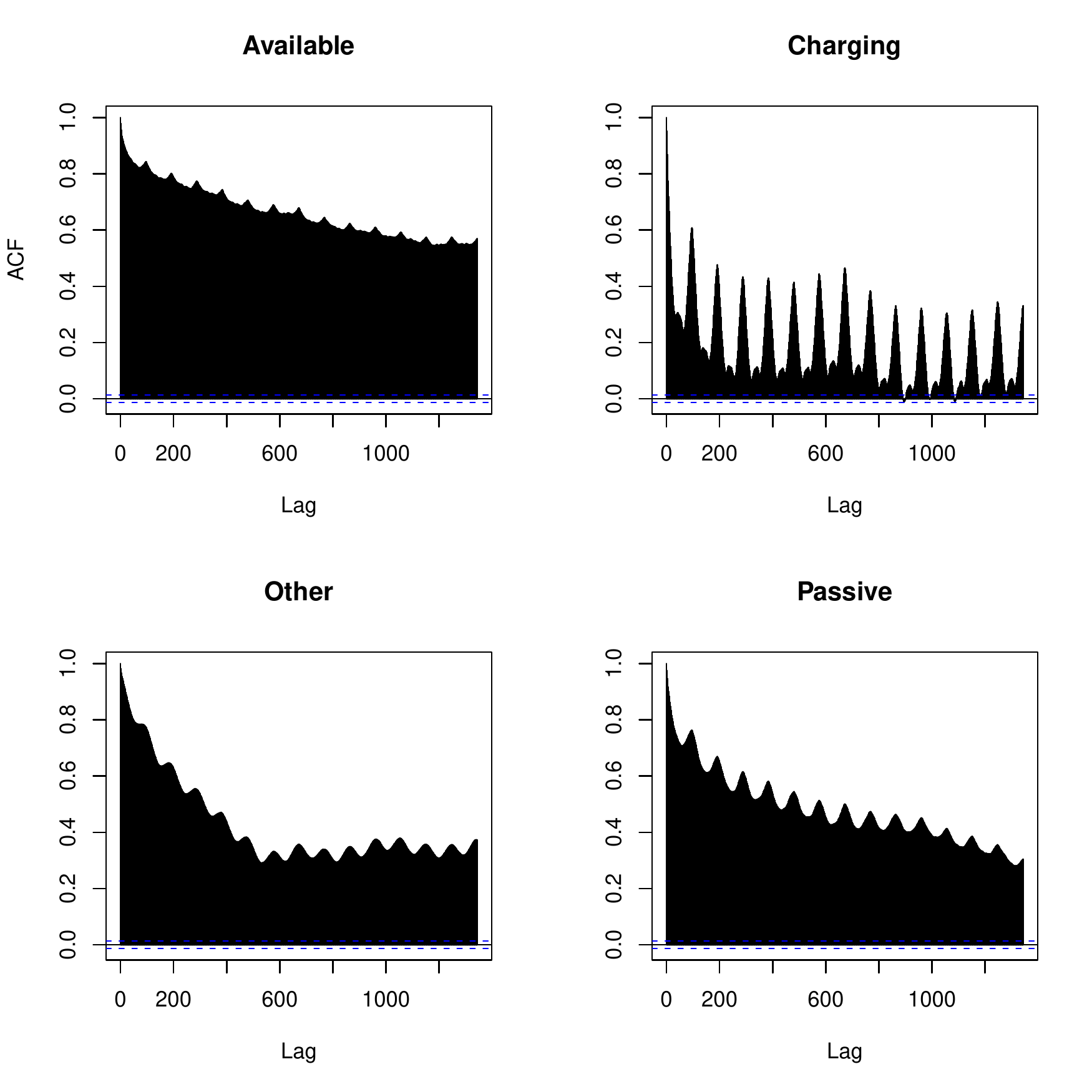}
    \caption{Empirical ACF of the 4 status at the global level.}
    \label{ACF}
\end{figure}

\begin{figure}[ht]
    \centering
    \includegraphics[width=0.45\linewidth]{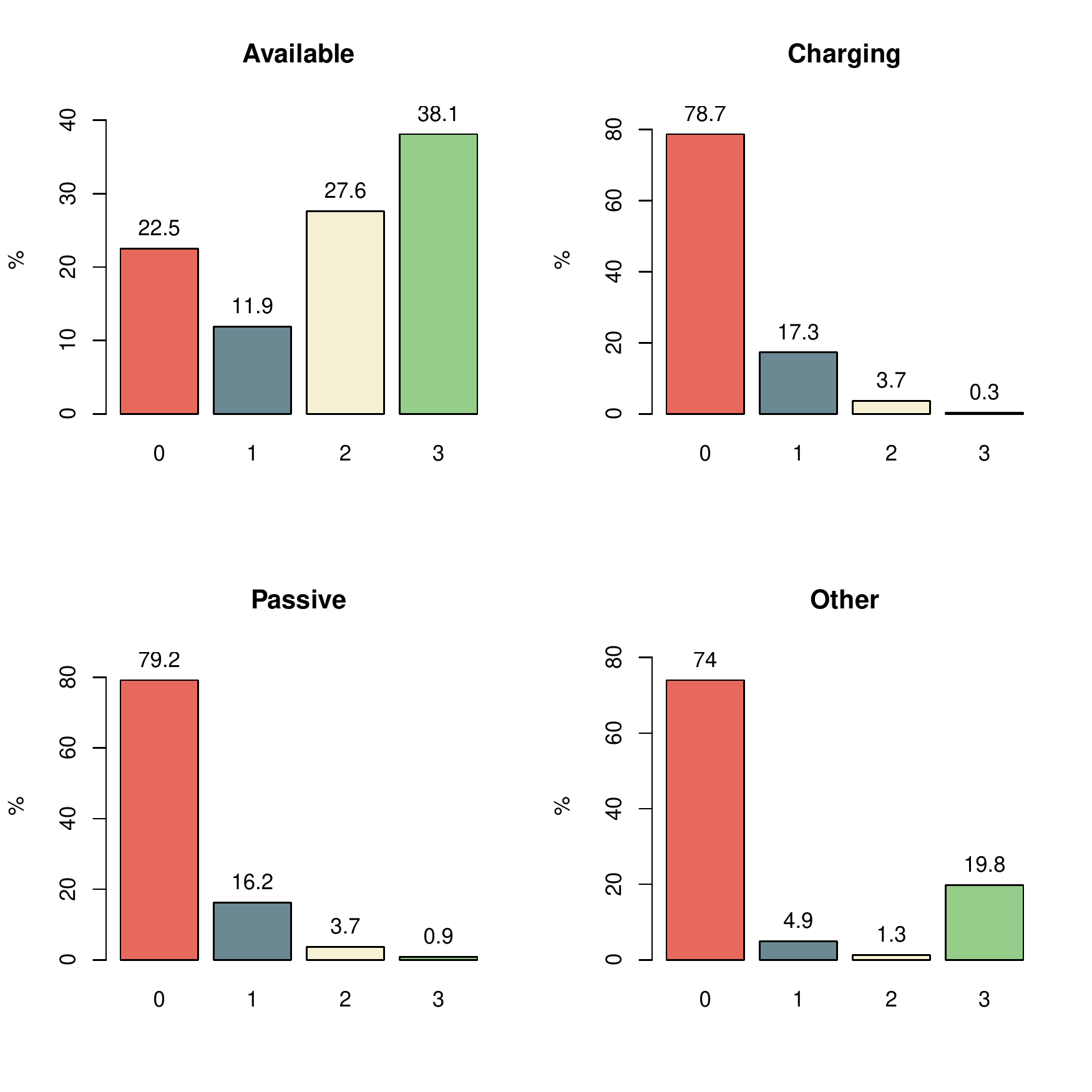}
    \includegraphics[width=0.45\linewidth]{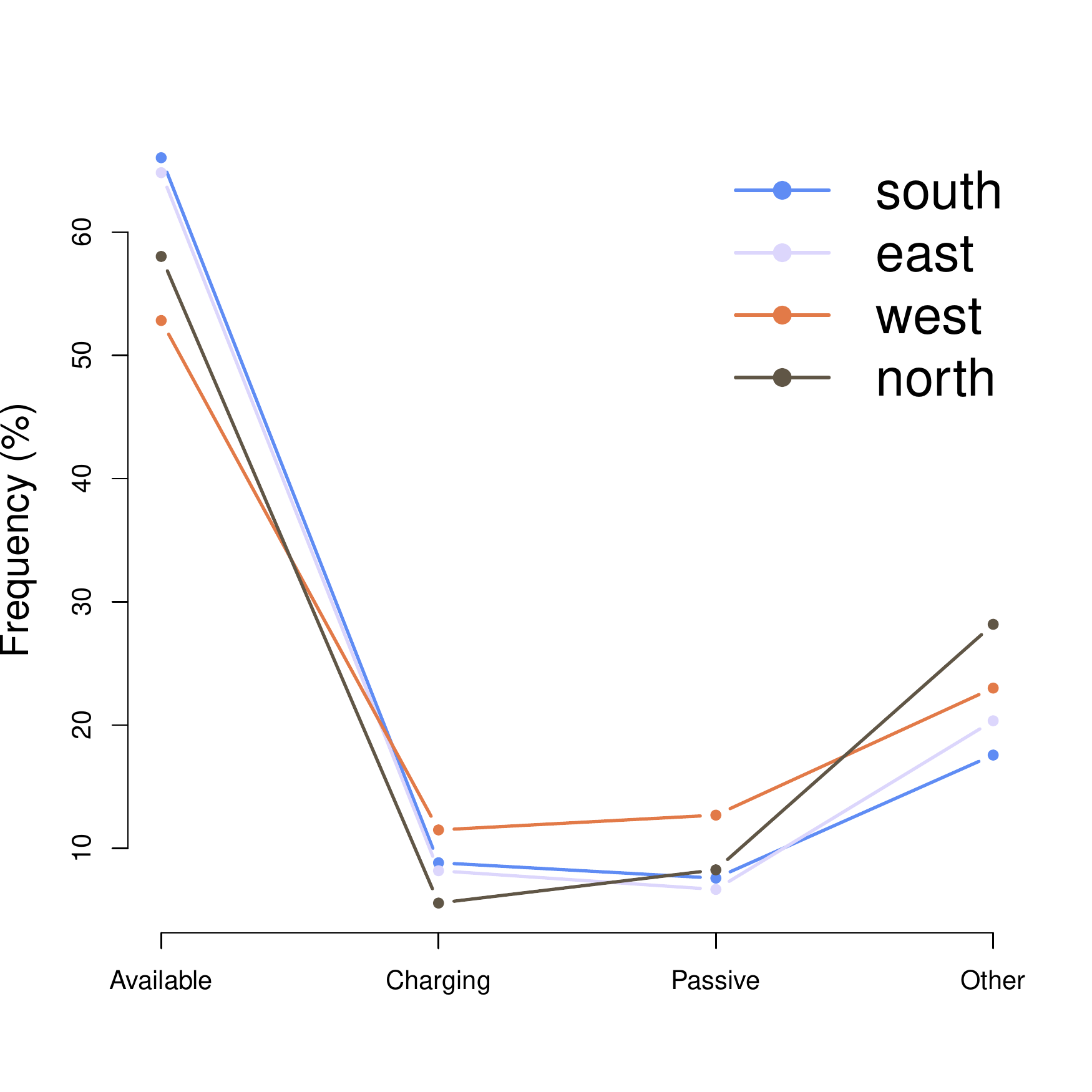}
    \caption{Left: distribution of the 4 states over all the stations and instants. Right: distribution of the 4 states by area.}
    \label{status_distribution}
\end{figure}

Figure~\ref{ACF} shows the empirical autocorrelation functions (ACF) at the global level. As excepted, we observe daily and weekly cycles. The daily cycle depends on the state of the plug. The non-stationnarity of the data is visible on the ACF: the \textit{available} status slowly decreases on its ACF due to the low frequency component of the data. We study the distribution of the states with respect to time and stations. The barplots of the corresponding frequencies (in percent) are shown in Figure~\ref{status_distribution} (left). We note a major difference between the \textit{available} status and the 3 others; the stations' plugs are most often available than in any other state. The distribution of the 4 states by area is shown in Figure~\ref{status_distribution} (right). The distribution profile is similar in all areas with a high frequency of \textit{available} status, followed by \textit{other}, \textit{passive} then \textit{charging}. We note that the \textit{other} status is over represented in the north area. The west area has lower availability due to higher charging activities as well as a high representation of Other. The south and east area are very similar, with higher representation of the \textit{available} status.

\section{Problem description}
\label{sec:problem}
In this section, we introduce the hierarchical forecasting challenge proposed to the contestants of the Smarter Mobility Challenge. The overall goal is to forecast the occupancy of charging stations at different geographical resolutions: single station, regional and global Paris area. Accurate prediction of a single station typically benefits to EV drivers looking for available charging points, whereas forecasting the occupancy of a network of charging stations allows utility providers to optimise their production units. This can lead to significant savings for the electricity system (around 1 billion euros per year, see \citet[Sections~5.4 and 5.5,][]{rte2019} and \citet[][]{LAUVERGNE2022120030}).

\begin{figure}[!ht]
    \centering
    \includegraphics[width=0.45\linewidth]{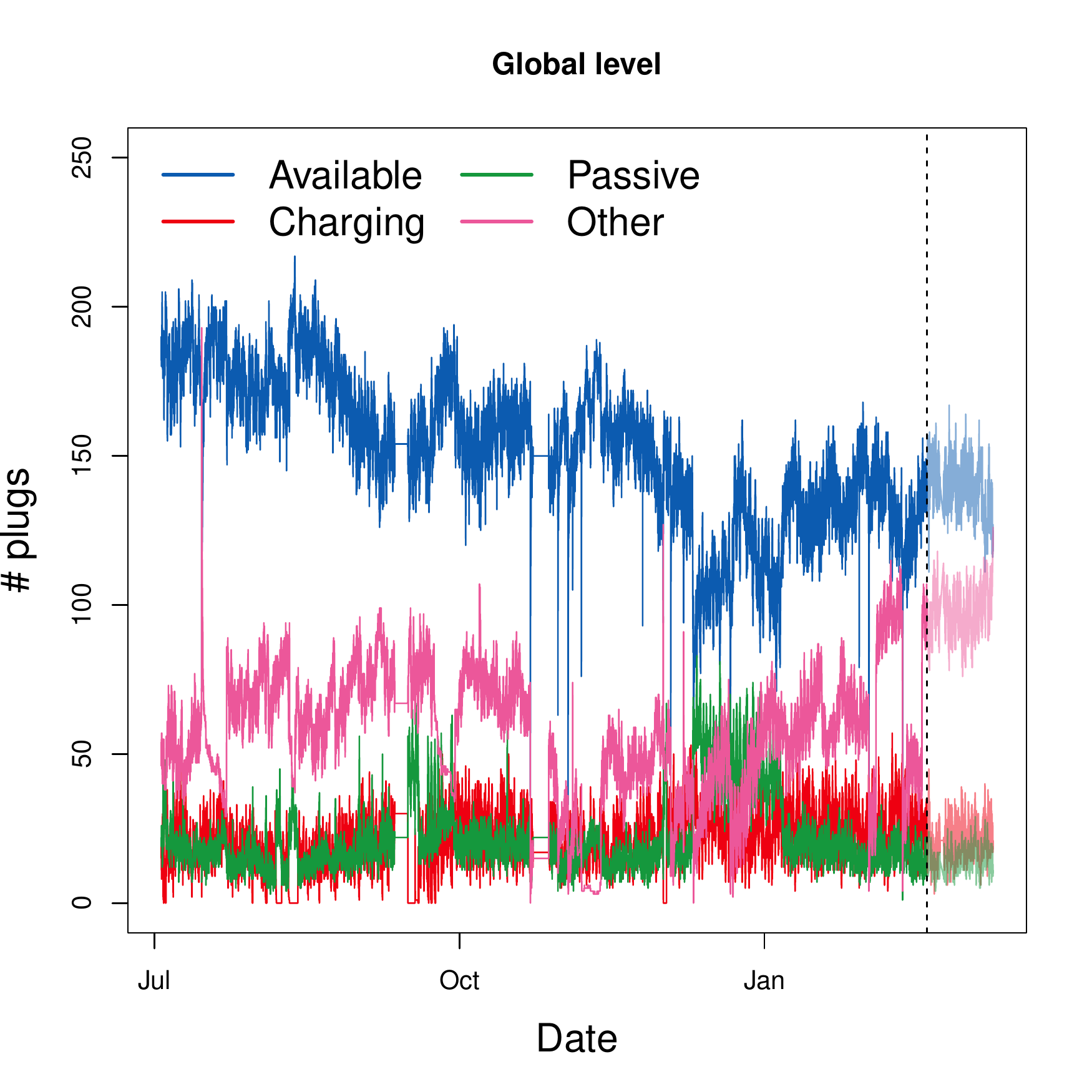}
    \includegraphics[width=0.45\linewidth]{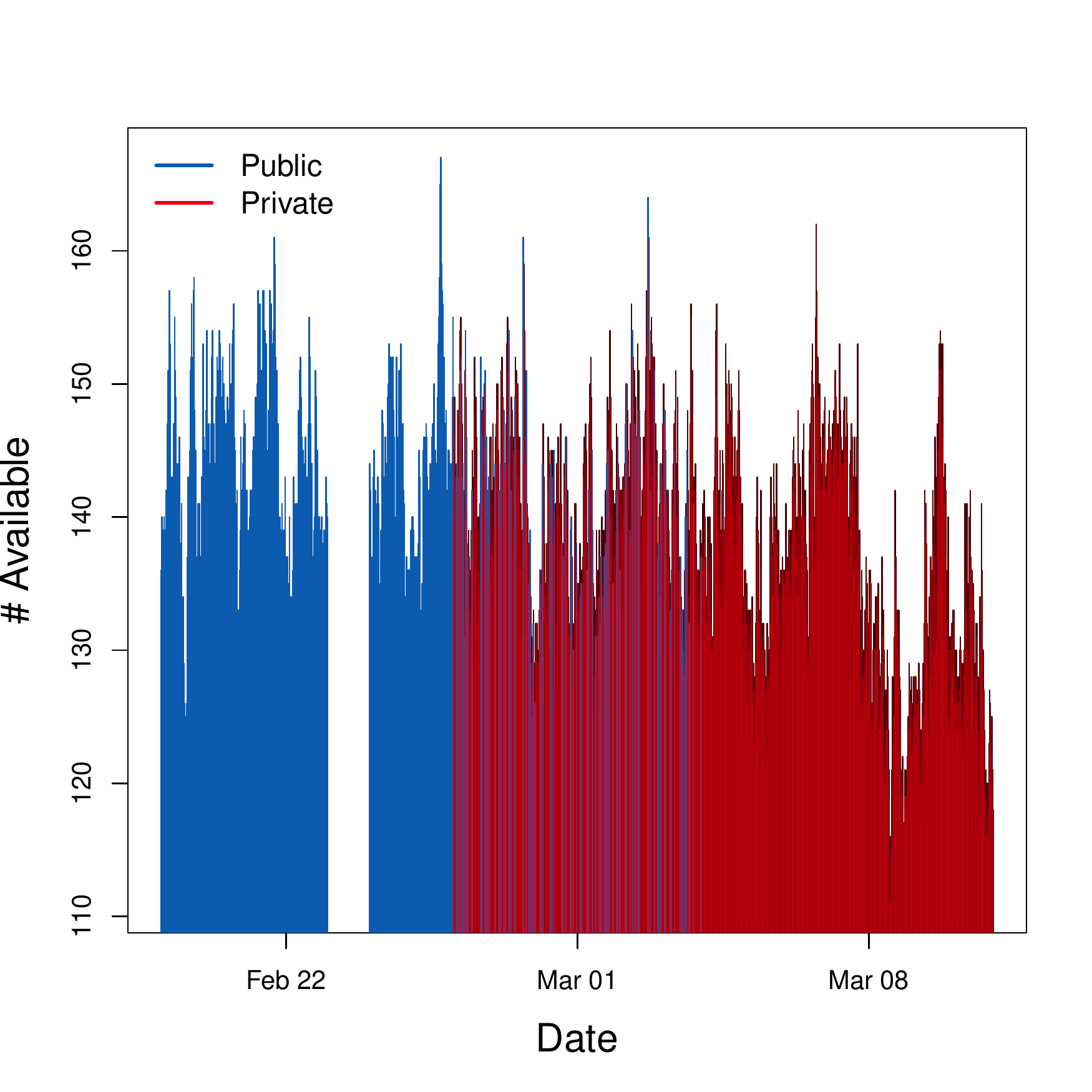}
    \caption{Left: total number of plugs in each state in function of time on train and test (transparent color). The vertical dashed lines represents the end of the train set. Right: total of \textit{available} plugs in function of time on the test set. In blue: public set. In red: private set.}
    \label{train_test}
\end{figure}

\paragraph{Data splitting} For this data challenge, we split the data between a training and a testing set. Because of the change of operator and pricing (see Section \ref{par:history}) on March 25th, 2021, we decided to study the following period: from July 3rd, 2020 to March 10th, 2021, when both the EV park and the pricing stayed unchanged. To mimic a genuine time-series forecasting problem, we preserved the time structure when partitioning the data and selected a test set of three weeks. The test set is a stable period that does not include significant changes in the data on the global level (Figure~\ref{train_test}).

The training set contains $D_{train}$ points from 2020-07-03 00:00 to 2021-02-18 23:45. The test set contains $D_{test}$ points from 2021-02-19 00:00 to 2021-03-10 23:45. As most EVSE stakeholders (e.g., EDF Group) receive the data with a delay of one to two weeks, we designed the challenge to match the operational perspective, hence the two-week forecast horizon. The test set has been divided into two subsets: a public set for validation purposes and a private set  $D_{private}$. The latter being used to quantify the performance of the solution while minimising the risk of overfitting.

To create the public and the private sets, the test set was split into three subsets of one week each. The first week was assigned to the public set, and the third one to the private set. We randomly assigned $20\%$ of second week to the public set and the rest to the private, as illustrated in Figure~\ref{train_test}. February 23 was excluded of the test set as it contains outliers. The public and private test sets were structured to balance the preservation of the temporal structure of the data and to avoid overfitting on short forecast horizons.

\paragraph{Target description} At any given time, a plug is in one of the four states.

\begin{itemize}
  \item A station is in state $c$ (\textit{charging}) when it is plugged into a car and provides electricity.
  \item In state $p$ (\textit{passive}) when connected to a car that is already fully charged.
  \item In state $a$ (\textit{available}) when the plug is free.
  \item In state $o$ (\textit{other}) when the plug is malfunctioning.
\end{itemize}

 We denote by $y_{t,k} = (a_{t,k} , c_{t,k} , p_{t,k} , o_{t,k}) \in \{0,1,2,3\}^4$ the vector representing the state of station $k\in \{1, \hdots, 91\}$ at time $t$, where $a_{t,k}$ is the number of available plugs, $c_{t,k}$ the number of charging plugs, $p_{t,k}$ the number of passive plugs, and $o_{t,k}$ the number of other plugs, at station $k$ and time $t$. By definition, eq. \ref{eq:sum} is always valid,

\begin{equation}\label{eq:sum}
    a_{t,k}+c_{t,k}+p_{t,k}+o_{t,k}=3. 
\end{equation}

\paragraph{Features description} To predict the state of station $k$ at time $t$, the dataset contains the following variables:

\begin{itemize}
    \item Temporal information: \textit{date}, \textit{tod} (time of day), \textit{dow} (day of week), and  \textit{trend} (a temporal index).
    \item Spatial information: \textit{latitude}, \textit{longitude}, and \textit{area} (south, north, east, and west) of the station.
\end{itemize}

\begin{figure}
    \centering
    \includegraphics[width=\linewidth]{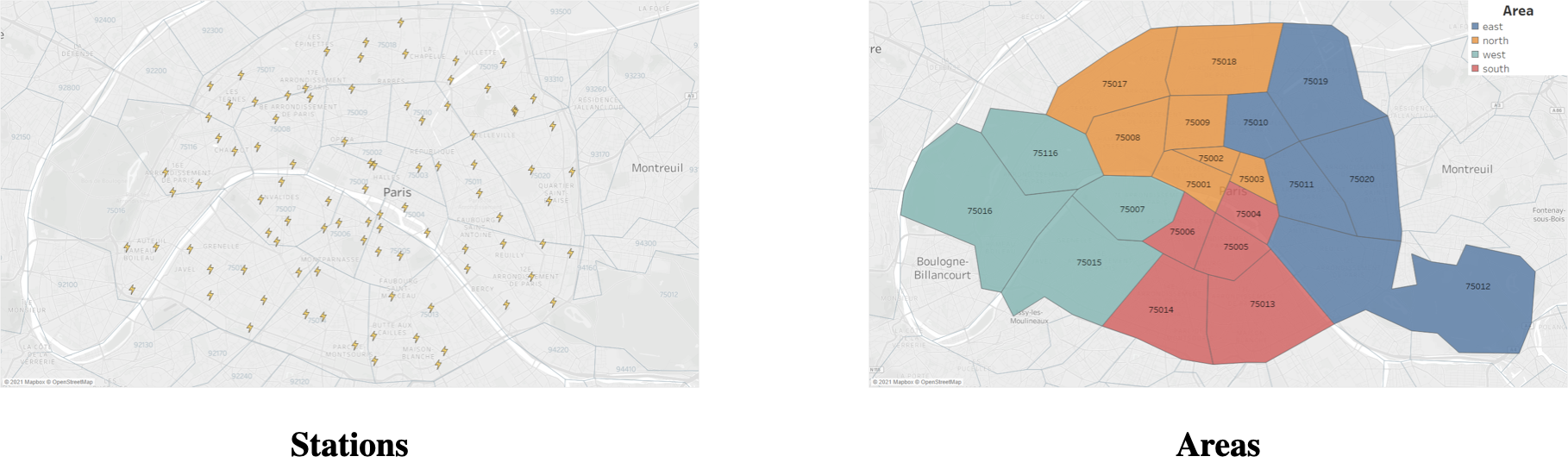}
    \caption{The 91 stations (yellow dots on the left) and the 4 areas of Paris (colored on the right)}
    \label{areas_stations}
\end{figure}

\textit{dow} is the day of week (from $1$ for Monday to $7$ for Sunday) and \textit{tod} the time of day, by interval of 15 minutes (0 for 00:00:00 to 95 for 23:45:00). The \textit{trend} feature is the numerical conversion of the time index, and \textit{date} is the corresponding string, in the ISO 8601 format. The data is then aggregated into 4 areas of about 20 stations each, as shown in Figure~\ref{areas_stations}.

\begin{figure}
    \centering
    \includegraphics[width=0.4\linewidth]{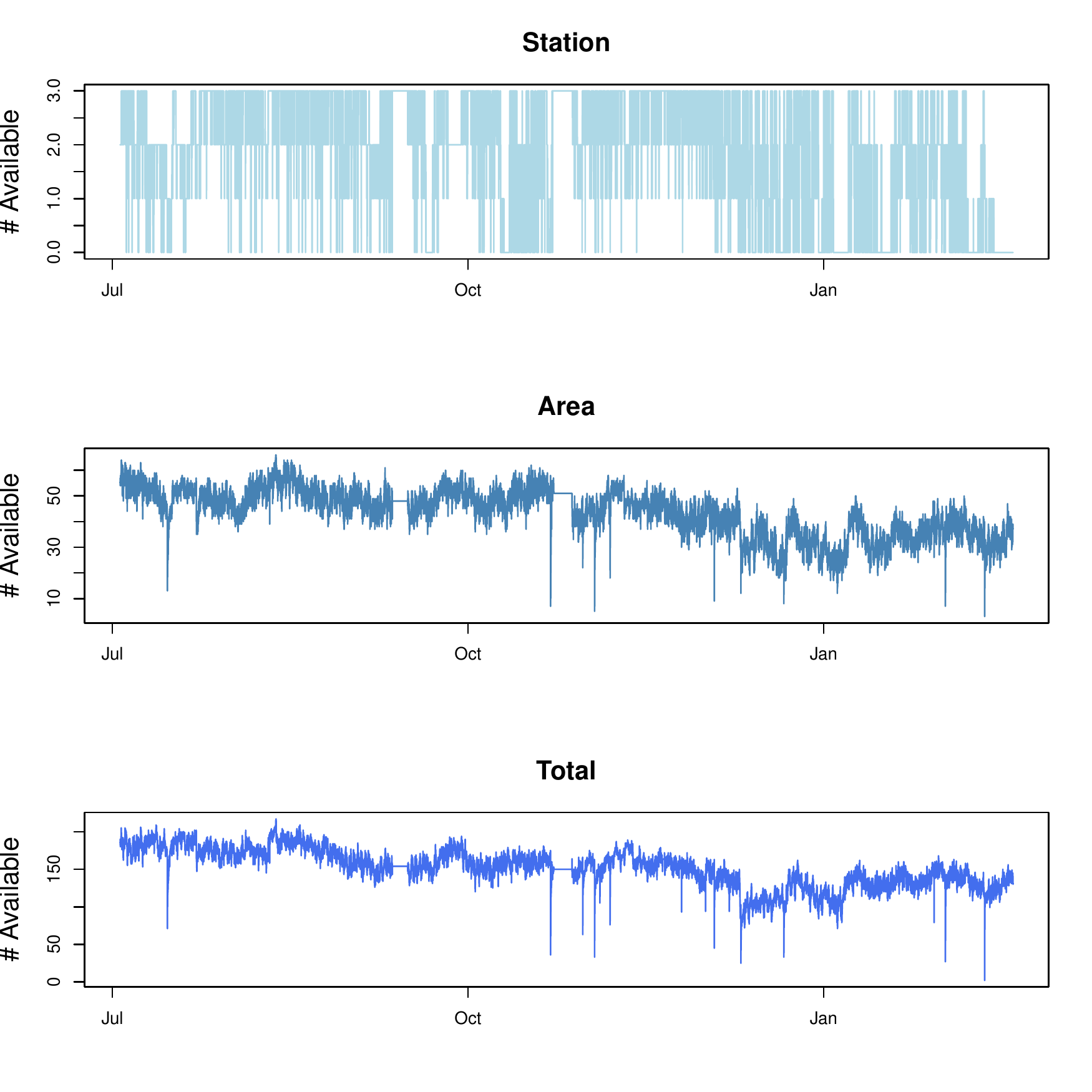}
        \includegraphics[width=0.4\linewidth]{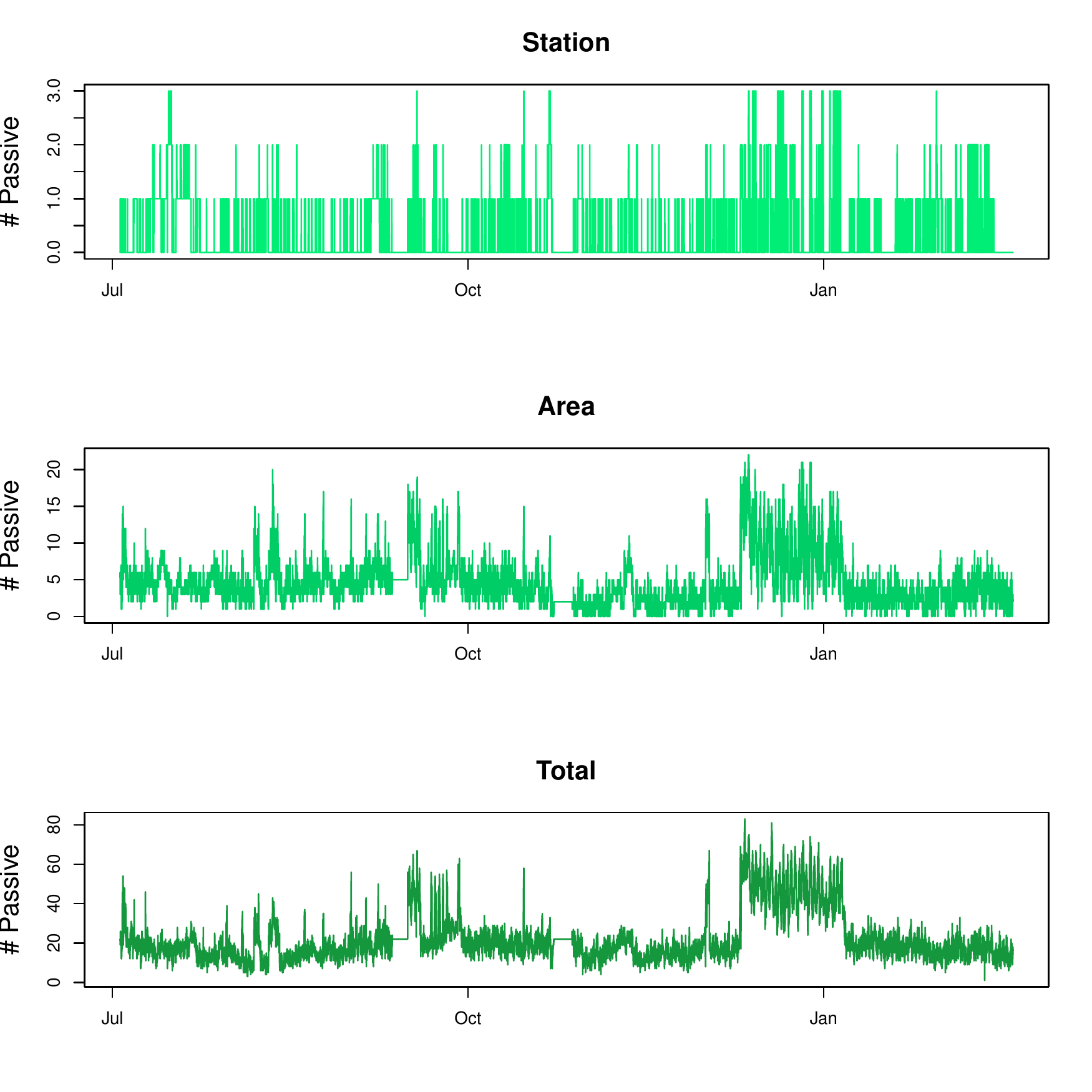}
    \caption{Number of available (left) an passive (right) plugs in function of time for one station, its corresponding area and at the global level.}
    \label{hierarchical}
\end{figure}

\paragraph{Evaluation}
\label{eval}
We aim to forecast the state of the different plugs at 3 hierarchical levels:

\begin{itemize}
    \item Individual stations: denoted by $y_{t,i}$, for $i \in \{1 \ldots 91\}$.
    \item Areas, corresponding to the cardinal points: $y_{t,\text{south}}$, $y_{t,\text{north}}$, $y_{t,\text{east}}$, and $y_{t,\text{west}}$
    \item At the global level: $y_{t,\text{global}}$
\end{itemize}

we also introduce $y_{t,\text{zone}} = \sum_{i \in \text{zone}} y_{t,i}$ as the sum of the plugs per state in a zone (south, north, east, west, or global). Let $\displaystyle z_t= (y_{t,1},\ldots, y_{t,91}, y_{t \text{south}},  y_{t,\text{north}}, y_{t,\text{east}}, y_{t,\text{west}},  y_{t,\text{global}})$ be the aggregated matrix containing the statutes of all stations at the different hierarchical levels at time $t$. The goal is to provide the best estimator $\hat z$ of $z$. Performance is evaluated using the following score, encoding each hierarchical level as a penalty.
\begin{equation}
    L(z, \hat{z}) = |D_{\mathrm{private}}|^{-1} \sum_{t \in D_{\mathrm{private}}}\big(\ell_{\text{station}}(z_t, \hat{z}_t) + \ell_{\text{area}}(z_t, \hat{z}_t) + \ell_{\text{global}}(z_t, \hat{z}_t)\big),
    \label{eq:testLoss}
\end{equation}
with the different terms defined as follows:
\begin{align*}
    \ell_{\text{station}}(z_t, \hat{z}_t) &= \sum_{k=1}^{91} \|y_{t,k}- \hat{y}_{t,k}\|_1, \\
    \ell_{\text{area}}(z_t, \hat{z}_t) &= \sum_{\text{zone} \, \in \, \mathcal{C}} \|y_{t,\text{zone}}- \hat{y}_{t,\text{zone}}\|_1, \\
    \ell_{\text{global}}(z_t, \hat{z}_t) &= \|y_{t,\text{global}}- \hat{y}_{t,\text{global}}\|_1,
\end{align*}
where $\mathcal{C} = \{ \text{south},  \text{north}, \text{east}, \text{west}\}$ is the set of cardinal points and $\|x\|_1 = \sum_{k=1}^{p} |x_k|$ is the usual $\ell^1$ norm on $\mathbb{R}^p$. We illustrate the different hierarchical level of the data in Figure~\ref{hierarchical}. We observe that spatial aggregation increases the signal-to-noise ratio, as the variance tends to decrease when the spatial aggregation is broader.

\paragraph{Baseline models}
As a baseline, we provided two models. A first naive estimator of 
$z_t$ is the median per day of week and quarter-hour over the training set, in which we removed the missing values:
\begin{equation} \label{eq:median}
    \hat{z}_{t} = \underset{t' \in Cal_{t}}{\mathrm{median}}\{z_{t'}\},
\end{equation}
where 
\begin{equation*} 
    Cal_{t} = \{t' \in D_{train}, \; \textrm{dow}(t') = \textrm{dow}(t)\} \cap \{ t' \in D_{train}, \;  \textrm{tod}(t') = \textrm{tod}(t)\}.
\end{equation*}
Notice that the $Cal_t$ corresponds to the timestamps of the same day of the week and the same hour of the day.

\label{CatBoostSection}
The second baseline model is the parametric model called 
\href{https://CatBoost.ai/}{(CatBoost)}. It is a tree-based gradient boosting algorithm designed to solve regression problems on categorical data. We used its implementation in the python library \texttt{CatBoost} \citep{CatBoost} and it has demonstrated excellent performance for a great variety of regression tasks \citep{daoud2019comparison, huang2019evaluation, hancock2020CatBoost} and forecasting challenges \citep{makridakis2022m5}. The performance of these two baselines on the private test set is shown by the dotted lines in Figure~\ref{fig:ranking}, next to the solutions of the winning team.

\section{Solutions of the winning teams}
\label{winning}
\begin{figure}[ht]
    \centering
    \includegraphics[width=\textwidth]{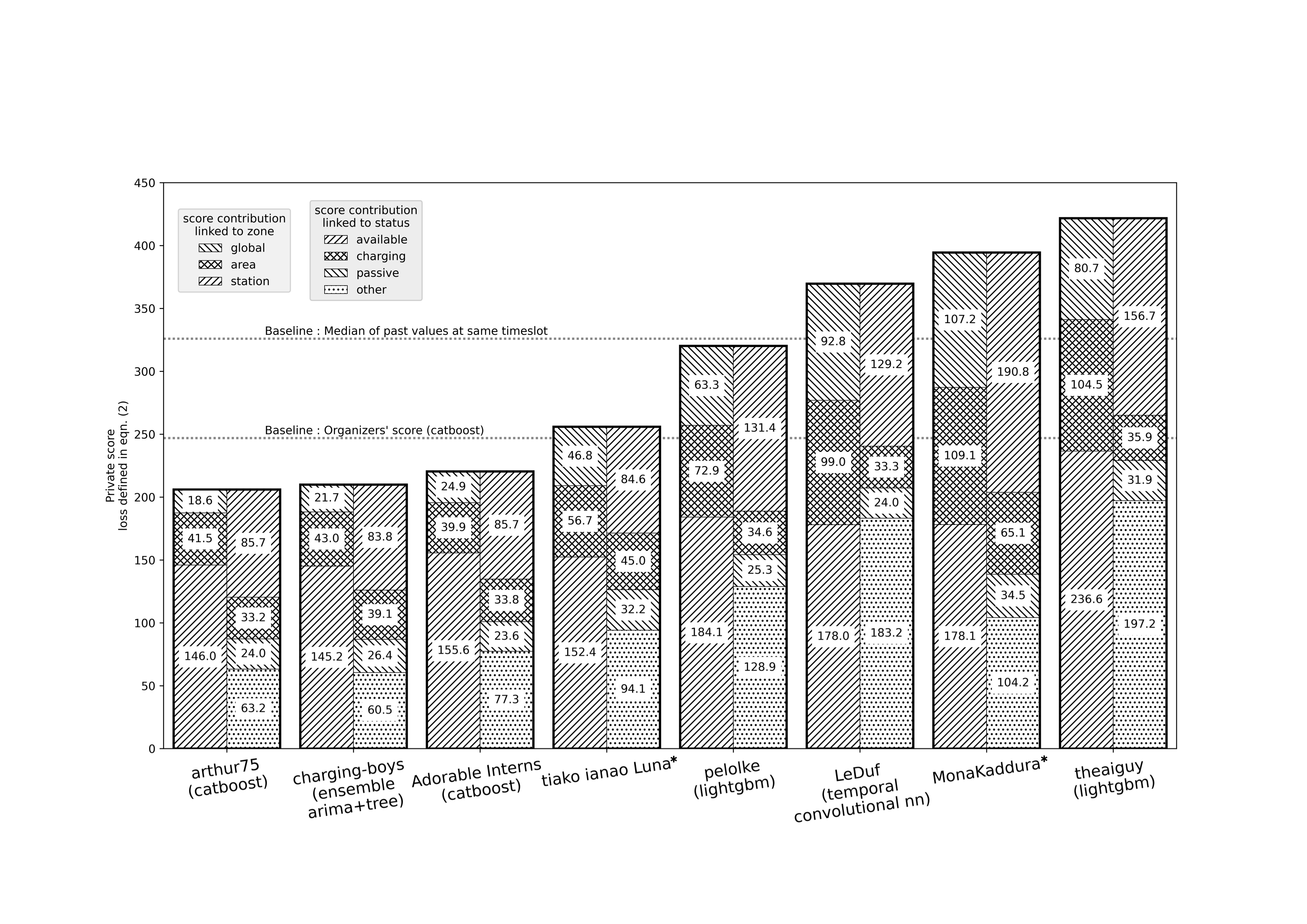}
    \vspace{-4em}
    \caption{Ranking of the top competitors.}
    \centering
    {\footnotesize $^*$ No information about these methods were provided by these competitors.}
    \label{fig:ranking}
\end{figure}

This section describes the methods used by the three winning teams. The ranking 
of the top competitors is shown in Figure~\ref{fig:ranking}. The confidence intervals are constructed by time series bootstrapping (non-overlapping moving block bootstrap) \citep{kunsch1989jackknife, politis1994stationary}. 
One subsection is dedicated to each of the winning teams, as their approaches are informative for the analysis of the dataset. In the last subsection, their strengths are combined using aggregation methods.

\subsection*{Arthur Satouf (team Arthur75)}
\label{satouf}


 \paragraph{Data exploration}$\;$ As shown in Figure \ref{nb_observations_per_station}, the dataset presents a lot of missing data. Common techniques were considered to impute these \cite{pratama2016review}, including computing the mean by station, forward and backward filling, simple moving average, weighted moving average, and exponential moving weighted average (EMW) \cite{sec:ewm}. These techniques are  evaluated by measuring the mean absolute error (MAE) on a validation subset of the training set. As a result, the EMW is the most effective technique, and it is thus implemented for both forward and backward filling approaches. Specifically, we use the last 8 known values to forward fill the first 8 missing values. The same procedure is applied to backward filling.
 
\paragraph{Model description}$\;$
We compare usual forecasting models \cite{ahmed2010empirical, chen2016xgboost, ribeiro2020ensemble}, such as SARIMAX, LSTM, XGBoost, random forest, and CatBoost.
The evaluation metric used is the MAE, and the time series cross-validation technique is applied to evaluate the performance of the models \cite{kreiss2011bootstrap, sklearn}.  
The CatBoost algorithm is ultimately chosen for its fast optimization relying on parallelization and its ability to handle categorical data without preprocessing.
As explained in Section \ref{sec:dataset}, the states of any station $k$ satisfy at any time $t$ the equation $a_{t,k}+c_{t,k}+p_{t,k}+o_{t,k}=3$, which is enforced in the CatBoost estimator as follows.

\begin{figure}[ht]
    \centering
        \begin{tikzpicture}[>=stealth,auto, node distance=0cm]
        \node [block] (avail1) {$Y_{avail.}$};
        \node [block, left=0cm of avail1] (train1)     {$X_{train}$};

        \node [block, right= 1.5cm of avail1] (model1)     {CatBoost 1};

        \node [block, below= 0.5cm of train1] (train2)     {$X_{train}$};
        \node [block, right= of train2] (avail2) {$Y_{avail.}$};
        \node [block, right=of avail2] (char2) {$Y_{char.}$};

        \node [block, right= 1.5cm of char2] (model2)     {CatBoost 2};

        \node [block, below= 0.5cm of train2] (train3)     {$X_{train}$};
        \node [block, right= of train3] (avail3) {$Y_{avail.}$};
        \node [block, right=of avail3] (char3) {$Y_{char.}$};
        \node [block, right=of char3] (pass3) {$Y_{pass.}$};

        \node [block, right= 1.5cm of pass3] (model3)     {CatBoost 3};

        \node [block, below= 0.5cm of train3] (train4)     {$X_{train}$};
        \node [block, right= of train4] (avail4) {$Y_{avail.}$};
        \node [block, right=of avail4] (char4) {$Y_{char.}$};
        \node [block, right=of char4] (pass4) {$Y_{pass.}$};
        \node [block, right=of pass4] (other4) {$Y_{other.}$};

        \node [block, right= 1.5cm of other4] (model4)     {CatBoost 4};
        \draw [->] (avail1) -- node {Train} (model1);
        \draw [->] (char2) -- node {Train} (model2);
        \draw [->] (pass3) -- node {Train} (model3);
        \draw [->] (other4) -- node {Train} (model4);
    \end{tikzpicture}
    
    \caption{Training process of the regressor Chain with CatBoost-Regressor.}
    \label{fig:training_catboost}
\end{figure}
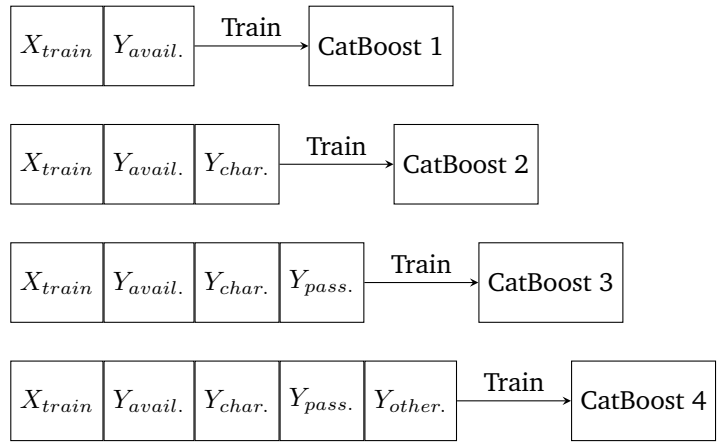

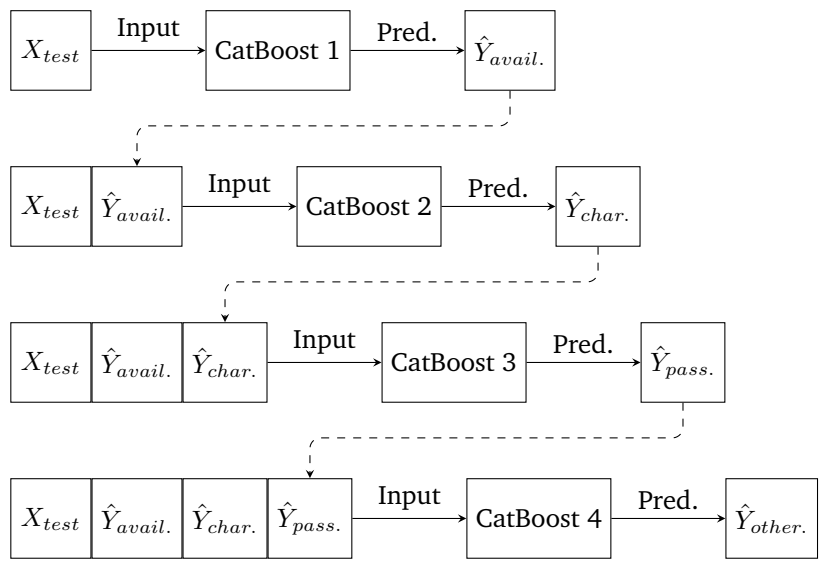
\begin{figure}[ht]
    \centering
        \begin{tikzpicture}[>=stealth,auto, node distance=0cm]
        \node [block] (test1)     {$X_{test}$};
        \node [block, right=1.5cm of test1] (model1)     {CatBoost 1};
        \node [block, right= 1.5cm of model1] (avail1) {$\hat{Y}_{avail.}$};

        \node [block, below=1cm of test1] (test2)     {$X_{test}$};
        \node [block, right= 0cm of test2] (avail2) {$\hat{Y}_{avail.}$};
        \node [block, right = 1.5cm of avail2] (model2)     {CatBoost 2};
        \node [block, right=1.5cm of model2] (char2) {$\hat{Y}_{char.}$};

        \node [block, below=1cm of test2] (test3)     {$X_{test}$};
        \node [block, right= 0cm of test3] (avail3) {$\hat{Y}_{avail.}$};
        \node [block, right= of avail3] (char3) {$\hat{Y}_{char.}$};
        \node [block, right = 1.5cm of char3] (model3)     {CatBoost 3};
        \node [block, right= 1.5cm of model3] (pass3) {$\hat{Y}_{pass.}$};

        \node [block, below=1cm of test3] (test4)     {$X_{test}$};
        \node [block, right= 0cm of test4] (avail4) {$\hat{Y}_{avail.}$};
        \node [block, right= of avail4] (char4) {$\hat{Y}_{char.}$};
        \node [block, right= of char4] (pass4) {$\hat{Y}_{pass.}$};
        \node [block, right = 1.5cm of pass4] (model4)     {CatBoost 4};
        
        \node [block, right= 1.5cm of model4] (other4) {$\hat{Y}_{other.}$};

        \draw [->] (test1) -- node {Input} (model1);
        \draw [->] (model1) -- node {Pred.} (avail1);
        \draw [dashed, rounded corners, ->] (avail1.270) |- ([yshift=-1cm]model1.west) -| node {} (avail2.90);

        \draw [->] (avail2) -- node {Input} (model2);
        \draw [->] (model2) -- node {Pred.} (char2);
        \draw [dashed, rounded corners, ->] (char2.270) |- ([yshift=-1cm]model2.west) -| node {} (char3.90);

        \draw [->] (char3) -- node {Input} (model3);
        \draw [->] (model3) -- node {Pred.} (pass3);
        \draw [dashed, rounded corners, ->] (pass3.270) |- ([yshift=-1cm]model3.west) -| node {} (pass4.90);

        \draw [->] (pass4) -- node {Input} (model4);
        \draw [->] (model4) -- node {Pred.} (other4);
    \end{tikzpicture}
    
    \caption{Inference process of the regressor Chain with CatBoost-Regressor.}
    \label{fig:inference_catboost}
\end{figure}

\begin{itemize}
    \item At the station level, the problem is transformed from a multi-task regression problem to a classification problem. This is achieved by concatenating the values of each task as a string, resulting in 20 unique classes. In this approach, the sum of the four vectors always equals three, given that there are three plugs. After predicting a given target, the target is decomposed into four values. Table \ref{data_example} provides an example.

\begin{table}
\centering
\caption{Example of a data conversion to a string} 
    \begin{tabular}{ cccccc } 
    \toprule
 Given station at a given time  & Available & Charging & Passive & Other & Target \\ 
    \midrule
 14h15-16/08/2021 & 1 & 2 & 0 & 0 & 1200 \\ 
 14h30-16/08/2021 & 0 & 1 & 1 & 1 & 0111 \\ 
 14h45-16/08/2021 & 0 & 0 & 3 & 0 & 0030 \\ 
    \bottomrule
\end{tabular}
\label{data_example}
\end{table}

    \item  At the area level, CatBoost was also used as a regression problem,  as shown in Figure \ref{fig:training_catboost} and Figure \ref{fig:inference_catboost}. However, each area had its own model, and each area used a combination of CatBoost regressor  and Regressor-Chain \cite{src:chain}. Regressor-Chain involves building a unique model for each task and using the result of each task as an input for the next prediction model. The output of each model, along with the previous output, is then used as input for the next task. This approach helps to keep the sum of plug equal to the right number and takes into account the correlation between tasks, making the prediction more robust. 

    \item At the global level, the approach is similar to the one applied to the area level, with only 4 models as there are no longer areas. 
\end{itemize} 
A time series cross validation is used once again to tune the hyperparameters and to validate the models. It relies on the mean absolute percentage error \cite{MAPE} at the area and the global levels, and on the F-measure \cite{chen2004statistical} at the station level. In total, 21 CatBoost models are used to forecast the private datasets.


\subsection*{Thomas Wedenig and Daniel Hebenstreit (team Charging-Boys)}

\paragraph{Data exploration}$\;$
Exploratory experiments did not show any signs of a trend within the time series. Regarding stationarity, we run the Augmented Dickey–Fuller test \cite{dickey1979distribution} on the daily averages of the target values for each station and find inconclusive results. Therefore, we cannot assume stationarity for all target-station pairs, which is why we employ differencing in the construction of our ARIMA model.
As usual in statistical frameworks, we assume that the noise interferes with the high frequencies of the signal.
To denoise, we preprocess the time series by computing a rolling window average with a window size of $2.5$ hours \cite{hyndman2018forecasting}.
During our data exploration, we encounter a significant change in the behavior of the individual stations in the end of October 2020, just before the COVID-19 regulations were enforced in Paris. We also assume that several stations were turned off after this event, as labels were missing over large time intervals. Thus, we experiment with different methods of missing value imputation, but find that simply dropping the timestamps with missing values performs best. We add custom features, namely a column indicating whether the current date is a French holiday, as well as sine and cosine transforms of \textit{tod}, \textit{dow}, the month, and the position of the day in the year. To ensure that our regression models return integer outputs that sum to $3$ for each station and timestamp (since stations have exactly $3$ plugs), we round and rescale these predictions in a post-processing step.

\paragraph{Model description} We train different models and then aggregate them. First, we start by considering a tree-based regression model.
Using \texttt{skforecast} \cite{skforecast}, we train an autoregressive XGBoost model \cite{chen2016xgboost} with $100$ estimators. We train it on all of the 91 stations individually, each having 4 targets, resulting in $364$ models. Each model receives the last $20$ target values, as well as the sine/cosine transformed time information as input, and predicts the next target value. We also discard all features that are constant per station (e.g., station name, longitude, and latitude). The final regression model achieves a public leaderboard score of $177.67$. 

Then, we consider a tree-based classification model.
To effectively enforce structure in the predictions (i.e., that they sum to $3$), we transform the regression problem discussed above into a classification problem. For a given station and timestamp, consider the set of possible target values $\mathcal{C} = \left\{ \mathbf{x} \in \{0,1,2,3\}^4 \quad \text{s.t.} \ \sum_{i=1}^4 x_i = 3 \right\}$.
We treat each element $c \in \mathcal{C}$ as a separate class and only predict class indices $\in \mathcal{I} = \{0, \dots, 19\}$ (since $|\mathcal{C}| = 20$). While $\mathcal{I}$ loses the ordinal information present in $\mathcal{C}$, this approach empirically shows competitive performance. When training a single XGBoost classifier with 300 estimators for all stations, we achieve a public leaderboard score of $178.9$. We also experiment with autoregressive classification (i.e.,including predictions of previous timestamps), but find no improvement in the validation error.

Finally, we fit a non-seasonal autoregressive integrated moving average (ARIMA) model \cite{box2015time} for each target-station combination.

To predict the value of a given target, we only consider the last $p = 2$ past values of the same target (in the preprocessed time series) and do not use any exogenous variables for prediction (e.g., time information).
We apply first-order differencing to the time series ($d=1$) and design the moving average part of the model to be of first-order ($q=1$).
On the validation and training sets, forecasts were applied recursively, using past forecasts as ground truth. 

We observe that the forecasts using these models have very low variance, i.e., each model outputs an approximately constant time series.
These predictions achieve a competitive score on the public leaderboard (third place).

The final model is an ensemble of the tree-based regression model, the tree-based classification model, and the ARIMA model. For a single target, we compute the weighted average of the individual model predictions (per timestamp). The ensemble weights are chosen to be roughly proportional to the public leaderboard score ($w_{\text{reg}} = 0.35$, $w_{\text{class}} = 0.25$,  $w_{\text{ARIMA}} = 0.4$).
Since the predictions of the tree-based models have high variance, we can interpret mixing in the ARIMA model's predictions as a regularizer, which decreases the variance of the final model.
As the tree-based models also use time information for their predictions, we use the entirety of the available features.

\subsection*{Nathan Doumèche and Alexis Thomas (team Adorable Interns)}
\label{teamai}
\begin{figure}
        \centering
        \includegraphics[width=0.8\textwidth]{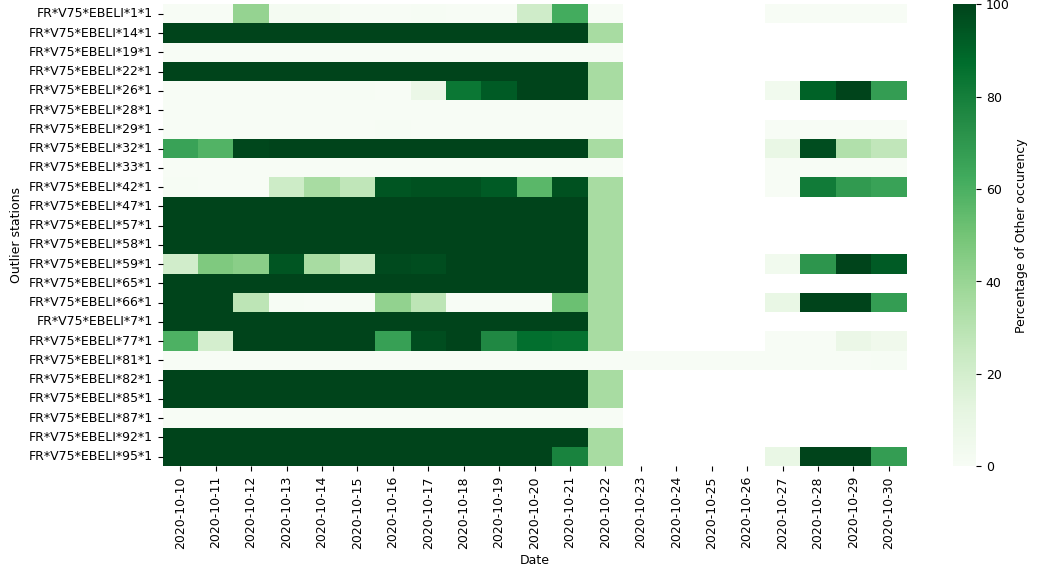}
        \caption{Percentage of state $o$ occurrences per outlier per day around 2020-10-22}
        \label{fig:Other2}
\end{figure}

\paragraph{Data exploration}$\;$ Several challenges arise from the data, as shown in Figure \ref{nb_observations_per_station}. 
An interesting phenomenon is the emergence of a change in the data distribution on 2020-10-22, characterized by the appearance of missing data.  
A reasonable explanation is that the detection of missing values is due to an update in the software that communicates with the stations.
The update would have taken place on 2020-10-22, allowing the software to detect new situations in which stations were malfunctioning. 
This hypothesis is supported by the fact that the stations with missing values are those that were stuck in states corresponding to the absence of a car, i.e., either the state $a$  or the state $o$ (see Figure \ref{fig:Other2}). In fact, 88\% of the stations that were stuck in either $a$  or $o$  for the entire week before 2020-10-22 had missing values on  2020-10-22.
Perhaps the users avoided the malfunctioning stations, or perhaps the users tried to connect to the station, but the plug was unresponsive, so the users went undetected. 
An important implication of this hypothesis is that the data before the change  should not be invalidated, since the behaviour of the well-functioning stations did not change. 
Another challenge of the dataset was its shortness. In fact, we expect a yearly seasonal effect due to holidays \citep{Xing2019charging} that cannot be distinguished from a potential trend because there is less than one year of data. 
All these observations suggest giving more weight to the most recent data.

As usual in the supervised learning setting, we need to choose a model $\mathcal{F}$ to construct the estimator $\hat{z}_t \in \mathcal{F}$. 
To estimate the entire $D_{test}$ period at once, we cannot rely on online models such as autoregressive models or hidden-state neural networks (RNN, LSTM, transformers...), although they perform well for time series forecasting \citep{bryan2021survey}, and in particular for EV charging station occupancy forecasts \cite{ma2022multistep, mohammad2023energy}.

Once a model $\mathcal{F}$ is chosen, we define an empirical loss $L$ on the training data. Then, a learning procedure, such as a gradient descent, fits the estimator $\hat{z}$ that minimizes $L$, with the hope that $\hat{z}$ will minimize the expectation of the test loss \eqref{eq:testLoss} \citep{vapnik1991principles, hastie2017elements}. Given a training set $T_{train} \subseteq D_{train}$, we consider two empirical losses.

The first one corresponds to Eq. \ref{eq:first_loss_nathan}, this loss gives equal weight to all data points.

\begin{equation} \label{eq:first_loss_nathan}
    L_{equal}(\hat{z}) = | T_{train}|^{-1}\sum_{t \in T_{train}} \|z_t - \hat{z}_t\|_1
\end{equation}

The second one is given in Eq. \ref{eq:second_loss_nathan}. 

\begin{equation} \label{eq:second_loss_nathan}
    L_{exp}(\hat{z}) = \sum_{t \in T_{train}} \exp((t-t_{max})/\tau)\|z_t - \hat{z}_t\|_1,
\end{equation}

where $\tau = 30$ days and $t_{max} =$ 2021-02-19 00:00:00.

This time-adjusted loss function is common for non-stationary processes \cite{ditzler2015learning} because it gives more weight to the most recent observations. This makes it possible to give more credit to the data after the change in the data distribution and to capture the latest effect of the trend, while using as much data as possible.

\paragraph{Model description} $\;$ To compare the performance of the models, we defined a training period $T_{train}$, covering the first $95\%$ of $D_{train}$, and a validation period $T_{val}$, covering the last $5\%$. In this benchmark phase, models are trained on $T_{train}$ to minimize $L_{equal}$ or $L_{exp}$, and then their performance is evaluated on $T_{val}$ by $L_{val}(\hat{z}) = |T_{val}|^{-1}\sum_{t \in T_{val}} \|z_t-\hat{z}_t\|_1$. 

The $Mean$ model estimates $\hat{y}_{t, k}$, $\hat{A}_{t, k}$ and $\hat{G}_t$ by their mean over the training period for each value of $(tod, dow)$. Idem for the $Median$ model. They are robust to missing values since the malfunctioning of a station $k$ only affects $\hat{y}_{t,k}$. 

We compare them with the CatBoost model presented in Section \ref{CatBoostSection}. Let $C(d, i)$ be the CatBoost model of depth $d$ trained with $i$ iterations using $L_{equal}$, and $C_{exp}(d, i)$ the same model trained using $L_{exp}$. 
In this setting, we train twelve CatBoost models: one for each pair of state ($a, c, p, o$) and hierarchical level.

After hyperparameter tuning, we found $C(4, 150)$ and $C_{exp}(5,200)$ to be the best models in terms of tradeoff between performance and number of parameters, knowing that early stopping and a small number of parameters prevent overfitting \citep[see, e.g.,][]{ying2019An}. All of these models take advantage of the fact that malfunctioning stations tend to stay in specific states.

The contest organizers allowed participants to test their models on a subset $T_{val}$ of $D_{test}$. In this validation phase, we trained  our best models on the entire $D_{train}$ period and  tested them with the test loss \eqref{eq:testLoss}. Table \ref{fig:perf} shows that the ranking of the models is preserved.
\begin{table}
    \centering
    \caption{Evaluation of the performance of the Adorable Interns' models  in both phases}
    \begin{tabular}{ccccc}
        \toprule & Mean & Median &$ C(4, 150)$ & $ C_{exp}(5, 200)$\\
        \midrule
         Benchmark Phase & 316 & 309 & 292 & \bf{261}\\
         Validation Phase& 323 & 303 & 233 & \bf{189}\\
            \bottomrule
    \end{tabular}
    \label{fig:perf}
\end{table}
The submitted model  was therefore $C_{exp}(5, 200)$. Note that this model is also interesting because its small number of parameters ensures robustness and scalability. In addition, tree-based models are quite interpretable, which is paramount for operational use \cite{jabeur2021CatBoost}.

\subsection*{Aggregation of forecasts from the winning teams}

Naive aggregations of uncorrelated estimators are known to have good asymptotic \citep{tsybakov2003optimal} and online \citep{cesa2006prediction} properties. In practice, they often achieve better performance than the individual estimators \citep[see, e.g.,][]{bojer2021kaggle, mcandrew2021aggregating}. 

Table \ref{table_score_target_agg2} shows the performance of the top 3 teams compared with two aggregation techniques. The \textit{Total} score is the result of Equation \eqref{eq:testLoss}, while the other scores are straightforward  subdivisions of the loss by hierarchical level and by state. Standard deviations are estimated by moving block bootstrap. The uniform aggregation --denoted by \textit{Uniform agg.}-- corresponds to the mean of each team's prediction, while the weighted aggregation --denoted by \textit{Weighted agg.}-- is computed by gradient descent using the MLpol algorithm \citep{gaillard2014second} to minimise the error on the training set. Notice how the weighted aggregation outperforms the other forecasts for the total loss, as well as for all the subdivisions of the loss. Note that the weighted aggregation of the 3 teams forecasts performs better than the weighted aggregation of any subsets of it (Arthur75+Charging Boys: $199$, Arthur75+Adorable Interns: $203$, Charging Boys+Adorable Interns:$200$). From these results, each team brings a significant contribution to the final score.

\begin{table}[ht]
\centering
\caption{Score by target of the top 3 teams and aggregations. } 
\resizebox{\textwidth}{!}{
\begin{tabular}{rllllllll}
  \toprule
 & Available & Charging & Passive & Other & Stations & Area & Global & Total \\ 
  \hline
Arthur75 & 85.7 (2.7) & 33.1 (0.7) & 24 (0.6) & 63.3 (2.8) & 145.6 (1.4) & 41.8 (2.5) & 18.7 (4.8) & 206.1 (5.7) \\ 
  Charging Boys & 83.9 (3.3) & 38.9 (0.6) & 26.3 (0.4) & 60.7 (3.4) & 145.3 (1.8) & 42.9 (3) & 21.7 (5.7) & 209.9 (6.8) \\ 
  Adorable Interns & 85.7 (2) & 33.8 (0.7) & 23.6 (0.6) & 77.4 (2.7) & 155.4 (1.5) & 40.1 (2.8) & 25 (3.8) & 220.5 (5.1) \\ 
  Uniform Aggregation & 82.9 (2.5) & 33.1 (0.7) & 22.1 (0.5) & 63.4 (2.7) & 141.1 (1.4) & 40.5 (2.9) & 20 (4.4) & 201.5 (5.4) \\ 
  Weighted Aggregation & 82.3 (2.7) & 33 (0.7) & 22.4 (0.5) & 58.5 (2.9) & 137.1 (1.4) & 40.3 (2.9) & 18.7 (4.4) & 196.2 (5.4) \\ 
   \hline
\end{tabular}}
\label{table_score_target_agg2}
\end{table}

\subsection*{Neural networks}

Although participants proposed a wide variety of models, they mainly focused on classical time series models like ARIMA (see, e.g., charging-boys) and tree-based models (see, e.g., arthur75). Indeed, the only neural network proposed in the challenge was LeDuf's temporal convolutional neural network, inspired by \citet{bai2018empirical}, and it performed poorly (see Figure~\ref{fig:ranking}).
Therefore, in order to get a better overview of their potential strengths, we completed our benchmark with neural networks after the challenge. The code to reproduce these experiments is available at \url{https://gitlab.com/smarter-mobility-data-challenge/tutorials/-/tree/master/2.\%20Model\%20Benchmark}.

Indeed, Fully Connected Neural Networks (FCNNs) are known to be able to forecast EV demand \citep{Boulakhbar2022a, Ahmadian2023artificial}.
The FCNN model we implemented predicts the status of individual stations. The forecasts for the area and the global levels are then derived in a bottom-up manner by summing the forecasts of the individual stations.
In contrast to the CatBoost models, this bottom-up approach performed better than training a FCNN for each hierarchical level (station, area and global).
The hyperparameters of the FCNN were then optimised using the \texttt{optuna} package in \texttt{Python} \citep{akiba2019optuna}.
As a result, the package selected a FCNN with one hidden layer, 155 neurons, a learning rate of $7.8e-4$, a dropout of $0.012$, a batch size of $480$ and $14$ epochs.
Similar to \citet{Ahmadian2023artificial}, we found out that FCNNs with a single hidden layer were the ones that performed best.
The performance of the FCNN on the test set for the hierarchical loss is 250.5 $\pm$ 3.1.
The standard deviation of the score is estimated by moving block bootstrap.
Thus, the FCNN is outperformed by the CatBoost model which has a loss of 246.1 $\pm$ 2.3.

Graph Neural Networks (GNNs) are neural networks that encode the spatial dependencies in a dataset as a graph to capture spatial correlations. 
GNNs are natural candidates among neural networks for EV charging forecasting because they inherently encode the spatial hierarchical structure of the dataset \citep{wang2023predicting, Qu2024a}. 
Among GNNs, Graph Attention Networks (GATs) are models designed for time series forecasting that exploit both temporal and spatial dependencies \citep{velickovic2018graphattentionnetworks}. 
Contrary to \citet{wang2023predicting} and \citet{Qu2024a}, the optimisation of this GNN did not converge, and its loss on the test set did not go below 400. We believe that this is due to the fact that  we only had access to 91 charging stations, which is not a big data regime, as compared to  \citet{wang2023predicting} who fitted their GNN on 76774 EVs and to \citet{Qu2024a} who fitted their GNN on 18061 EV charging piles. Both \citet{wang2023predicting} and \citet{Qu2024a} only had access to one month of data and focused on short-term forecasting, which may also explain this difference.
\section{Summary of findings and discussion}
\label{summary}

This paper presents a dataset in the context of hierarchical time series forecasting of EV charging station occupancy, providing valuable insights for energy providers and EV users alike. 

\paragraph{Models}
Contestants were able to train models that significantly outperformed the baseline performance (see Figure~\ref{fig:ranking}).
This dataset contains many practical problems related to time series, including missing values, non-stationarity, and outliers. 
This explains why most contestants relied on tree-based models, which are robust enough to outperform more sophisticated machine learning methods.

\paragraph{Data cleaning}
Specific techniques were developed to deal with missing data and outliers  (see, e.g., Section \ref{satouf}). 
Data preprocessing is a crucial step, and the addition of relevant exogenous features, such as the national holidays calendar, significantly improved the results.

\paragraph{Time dependant loss function}
All three of the winning solutions described in this paper were robust enough to maintain a high private test score, showing good generalization of the models. The choice of the empirical cost function to drive the training process produced the best results when more recent data points were given greater weight (see, e.g., Section \ref{teamai}).  

\paragraph{Aggregation}
Aggregating the forecasts of the three winning teams even yielded a better global score, with a notable improvement at the station level. The hierarchical models presented are promising and could help improve the overall EV charging network. 

\paragraph{Why publishing this dataset?}
This open dataset is interesting for research purpose because it encompasses many real-world problems related to time series matters, such as missing values, non-stationarities, and spatio-temporal correlations. In addition, we strongly believe that sharing the benchmark models derived from this challenge will be useful for making comparisons in future research.
Two more complete datasets using new features and spanning from July 2020 to July 2022 are available at  \href{https://doi.org/10.5281/zenodo.8280566}{doi.org/10.5281/zenodo.8280566} and at \href{https://gitlab.com/smarter-mobility-data-challenge/additional_materials/}{gitlab.com/smarter-mobility-data-challenge/additional\_materials}. A primary analysis is presented in the supplementary material. 

\paragraph{Perspectives}
Managing a fleet of EVs in the context of an increasing renewable production amount open new challenges for forecasters. We hope this dataset will allow other researchers to work on topics such as probabilistic forecasts, online learning (our challenge was "offline") or graphical models.

\paragraph{Limitations} The deployment of electric vehicles (EVs) is progressing at a remarkable pace \citep{sathiyan2022comprehensive}, making any dataset merely a snapshot of a swiftly evolving world \citep[see also][]{hecht2021predicting}. 
To enhance forecasting accuracy, additional features could be incorporated into a dataset. Numerous covariates, such as mobility and traffic information, meteorological data, and vehicle characteristics, could be included. In a forthcoming release of the dataset, in addition to extending the observation period, we intend to incorporate traffic and meteorological data. A first attempt is proposed in the Section 4 of the supplementary material.


\paragraph{Ethical concerns} To the best of our knowledge, our work does not pose any risk of security threats or human rights violations. Knowing when and where someone plugs in their EV could lead to a risk of surveillance.
However, this dataset does not contain any personal information about the user of the plug or their car, so there is no risk of consent or privacy.

\setcounter{section}{0}
\renewcommand\thesection{\thechapter.\Alph{section}}

\section{Belib's history: pricing mechanism and park evolution}
\label{sec:history}
Though there is no official document relating the evolution of the Belib pricing mechanism, it is possible to reconstruct its history through the press. 
Belib pricing strategy evolved twice (on 25 March 2021 and on January 2023), and the press releases explicitly state that they do not vary between these dates \cite{2021lesnumeriquestotal, 2023lesnumeriquesvoiture}.
This ensures that both the Belib EV park and the pricing strategy did not change during the period studied in the challenge, from 2020-07-03  to  2021-03-10. 

\paragraph{Belib creation in 2016} 
The 5 first stations of the Belib network were first operational on 12 January 2016 \cite{2016automobileBornes, 2016stageBornes}. The network grew progressively during the year 2016 to reach 60 stations all around Paris. Users needed to buy a 15 euro badge to connect to the network. Different pricing strategies were applied depending on the time of the day and the plugs electric power.
The "normal charge" of 3kW was free at night (between 8 p.m. and 8 a.m.) and cost 1 euro per hour on daytime (between 8 a.m. and 8 p.m.).
The "quick charge" of 22kW cost 0.25 euro per 15 minutes  the first hour of charge. After the first hour, the first 15 minutes cost 2 euros. After this 1h and 15 minutes, each 15 minutes slot cost 4 euros. 
Each station contained 3 parking spots:
\begin{itemize}
    \item one dedicated to "normal charge" with an E/F electric plug,
    \item one dedicated to "quick charge" with a ChaDeMo and a Combo2 plugs,
    \item one where both "normal charge" and "quick charge" were possible, with an E/F, a T2, and a T3 plugs.
\end{itemize}
Therefore, this pricing strategy meant that "normal charge" plugs could serve as free parking spot at night, while "quick charge" became expensive after one hour of usage.

\paragraph{Belib under the TotalEnergies supervision begining on 25 March 2021} In 2021, the city of Paris allowed the TotalEnergies company to run the Belib network for a period of 10 years. 
The goal is to develop the network from its 90 stations of 270 charging points, to 2300 charging points \cite{2021totalVehicules, 2021leparisienLe}.
More precisely, our dataset accounts for 91 stations corresponding to 273 charging points.
This change of the network's operator was accompanied with the following change in pricing on 25 March 2021 \cite{2021automotoParis}. 
Four programmes became available (Flex, Moto, Boost, and Boost+), with pricing depending on station's location and on the frequency of use.
For occasional users, 
\begin{itemize}
    \item the Flex programme allows the usage of 3.7kW and 7kW plugs. In the districts 1 to 11 of Paris, the pricing goes as follows. 
    The first 2 hours of charging, each 15 minutes cost 0.90 euro.
    Then,  each 15 minutes cost 1.00 euro. Then, after 3 hours of charging, each 15 minutes cost 1.10 euro. In the districts 12 to 20 of Paris, the pricing goes as follows. 
    The first 2 hours of charging, each 15 minutes cost 0.55 euro.
    Then,  each 15 minutes cost 0.65 euro. Then, after 3 hours of charging, each 15 minutes cost 0.75 euro.
    \item the Moto programme allows the usage of 3.7 kW plugs for motorcycles at the cost of 0.35 euro per 15 minutes. 
    \item the Boost programme allows the usage of 22kW plugs  at the cost of 1.90 euro per 15 minutes. 
    \item the Boost+  programme allows the usage of 50 kW plugs at the cost of 4.80 euros per 15 minutes. 
\end{itemize}
For regular users, at the condition of a yearly-7-euro subscription,
\begin{itemize}
    \item the Flex programme allows the usage of 3.7kW and 7kW plugs. In the districts 1 to 11 of Paris, the pricing goes as follows. 
    The first 2 hours of charging, each 15 minutes cost 0.75 euro.
    Then,  each 15 minutes cost 0.80 euro. Then, after 3 hours of charging, each 15 minutes cost 0.85 euro. In the districts 12 to 20 of Paris, the pricing goes as follows. 
    The first 2 hours of charging, each 15 minutes cost 0.50 euro.
    Then,  each 15 minutes cost 0.55 euro. Then, after 3 hours of charging, each 15 minutes cost 0.60 euro.
    \item the Moto programme allows the usage of 3.7 kWh plugs for motorcycles at the cost of 0.30 euro per 15 minutes. 
    \item the Boost programme allows the usage of 22kWh plugs  at the cost of 1.70 euro per 15 minutes. 
    \item the Boost+  programme allows the usage of 50 kWh plugs at the cost of 4.40 euros per 15 minutes. 
    \item at nighttime, pricing was more advantageous. The Flex programme cost 3.90 euros for the whole night, plus  0.20 euros for each consumed kWh after a 19,5 kWh consumption.
    The Moto programme cost 2.90 euros for the whole night, and then 0.20 euros for each kWh after a 19.5 kWh consumption.
\end{itemize}
In all cases, after 14 consecutive hours of parking on a charging spot, any programme would then cost 10 euros per hour, the charging starting at the beginning of each hour.
This more complex pricing was perceived as misleading for the consumers, and resulted in overall higher expenses.
The pricing did not changed until 1 January 2023 \cite{2023lesnumeriquesvoiture}.

\paragraph{An price increase on January 2023}
On 1 January 2023, the Belib pricing strategy evolved. 
It was decided to take into account that some stations were malfunctioning and did not deliver the expected electric power.
Indeed, with the previous pricing strategy depending only on the charging time, some drivers were paying too much for charging their cars.
Therefore, the pricing evolve to take into account both the time spent in the EV charging spot and the energy transmitted to the car.
Moreover, because of the energy crisis in Europe, electricity prices had increased and TotalEnergies raised their charging price.
This resulted in higher expenses for EV users, leading to a significant drop in the usage of Beilb stations \cite{2023energeekVoiture}.
On 25 January 2023, to cope with the decreasing numbers of users, Belib prices were decreased \cite{2023leparisienParis}.
The same pricing segmentation was kept, but with lower prices. 
This pricing is the same as today, on August 2023.
Both the pricing on 1 January 2023 and on 25 January 2023 (in bold) are detailed  in Tables \ref{table_pricing_visitor_1}, \ref{table_pricing_visitor_2}, and \ref{table_pricing_parisian} \cite{2023lesnumeriquesvoiture, 2023belib}. Table \ref{table_pricing_visitor_1} details the pricing for occasional users, without subscription. Table \ref{table_pricing_visitor_2} details the pricing for regular users not living in Paris, with a yearly 7-euro subscription. Table \ref{table_pricing_parisian} details the pricing for regular users living in Paris, with a yearly 7-euro subscription. Notice that the distinctions between districts was abandoned. 
In all cases, after 14 consecutive hours of parking on a charging spot, any programme would then cost 10 euros per hour, the charging starting at the beginning of each hour.

\begin{table} 
\centering
\begin{tabular}{rrrrr}
  \hline
  Pricing &  Moto (3.7kW) & Flex (7kW) & Boost (22 kW) & Boost+ (50kW) \\ 
  \hline
 kWh on 01/01/2023 & 0.55€ & 0.55€  & 0
  & 0 \\
 Parking on 01/01/2023 & 0.35€ / 15min & 0.78€ / 15min &
 2.30€ / 15min & 0.50€ / min \\
 \hline
 kWh on 01/25/2023 & 0.35€ & 0.35€  & 0
  & 0 \\
 Parking on 01/25/2023 & 0.20€ / 15min & 0.55€ / 15min &
 2.30€ / 15min & 0.38€ / min \\
   \hline
\end{tabular}
\caption{Pricing for occasional users} 
\label{table_pricing_visitor_1}
\end{table}

\begin{table} 
\centering
\begin{tabular}{rrrrr}
  \hline
  Pricing &  Moto (3.7kW) & Flex (7kW) & Boost (22 kW) & Boost+ (50kW) \\ 
  \hline
 kWh on 01/01/2023 & 0.55€ & 0.55€  & 0
  & 0 \\
 Parking on 01/01/2023 & 0.30€ / 15min & 0.60€ / 15min &
 2.15€ / 15min & 0.45€ / min \\
 \hline
 kWh on 01/25/2023 & 0.35€ & 0.35€  & 0
  & 0 \\
 Parking on 01/25/2023 & 0.15€ / 15min & 0.35€ / 15min &
 2.05€ / 15min & 0.35€ / min \\
   \hline
\end{tabular}
\caption{Pricing for regular users not living in Paris} 
\label{table_pricing_visitor_2}
\end{table}

\begin{table} 
\centering
\begin{tabular}{rrrrr}
  \hline
  Pricing &  Moto (3.7kW) & Flex (7kW) & Boost (22 kW) & Boost+ (50kW) \\ 
  \hline
  8 a.m. to 8 p.m. & & & &\\
 kWh on 01/01/2023 & 0.55€ & 0.55€  & 0
  & 0 \\
 Parking on 01/01/2023 & 0.30€ / 15min & 0.60€ / 15min &
 2.15€ / 15min & 0.45€ / min \\
 kWh on 01/25/2023 & 0.35€ & 0.35€  & 0
  & 0 \\
 Parking on 01/25/2023 & 0.15€ / 15min & 0.35€ / 15min &
 2.05€ / 15min & 0.35€ / min \\
 \hline 
  8 p.m. to 10 p.m. & & & &\\
 kWh on 01/01/2023 & 0.55€ & 0.55€  & 0
  & 0 \\
 Parking on 01/01/2023 & 0.15€ / 15min & 0.20€ / 15min &
 2.15€ / 15min & 0.45€ / min \\
 kWh on 01/25/2023 & 0.35€ & 0.35€  & 0
  & 0 \\
 Parking on 01/25/2023 & 0.10€ / 15min & 0.15€ / 15min &
 2.05€ / 15min & 0.35€ / min \\
 \hline 
  10 p.m. to 8 a.m. & & & &\\
 kWh on 01/01/2023 & 0.30€ & 0.30€  & 0
  & 0 \\
 Parking on 01/01/2023 & 0.05€ / 15min & 0.05€ / 15min &
 2.15€ / 15min & 0.45€ / min \\
 kWh on 01/25/2023 & 0.25€ & 0.25€  & 0
  & 0 \\
 Parking on 01/25/2023 & 0.05€ / 15min & 0.05€ / 15min &
 2.05€ / 15min & 0.35€ / min \\
   \hline
\end{tabular}
\caption{Pricing for regular users living in Paris} 
\label{table_pricing_parisian}
\end{table}

\section{Data description}
\label{sec:dataDescr}
\subsection*{Data set collection}

We set up a DataLake to collect and make available all types of data related to electric mobility. This dataset is informative about the charging stations (static data) and their use in real time (dynamic data) everywhere in France and in particular in Paris, where the operating network is called Belib.
The DataLake has set up an automatic and real-time collect of Belib data as it is published on the supplier's site: \url{https://parisdata.opendatasoft.com/explore/dataset/belib-points-de-recharge-pour-vehicules-electriques-disponibilite-temps-reel/api}.
The storage of this information over time allows, for example, to estimate the frequentation of the charging stations according to their location.
The DataLake Mobility uses a big data infrastructure based on Hadoop technologies (HDFS, PySpark, Hive and Zeppelin), allowing to manipulate large volumes of data and to process massive data, including to launch machine learning algorithms.

\subsection*{Data preprocessing}

\paragraph{Aggregation}

The raw data is structured so that each observation reflects the status of the plugs (up to 6) within a charging point. This is misleading because only one of these plugs can be in use at a time. Therefore, we have kept the relevant rows only for plugs in use and treated a charging point as a single plug. In addition, in the raw data, charging points appear in fixed geographic locations and in groups of three. This  charging point structure was confirmed by the data provider. 
Therefore, it makes sense to regroup three adjacent charging points into a single charging station and we have aggregated the data to make each observation the status of the 3 charging points every 15 minutes. To account for discrepancies in timestamp synchronisation between stations we adjusted the timestamps to match the closest 15 minute interval.

\paragraph{States}
The \textit{available}, \textit{charging}, and \textit{passive} states are taken directly from the raw data. The last state \textit{other} is regroups several statuses including \textit{reserved} (a user has booked the charging point), \textit{offline} (the charging point is not able to send information to the server), and \textit{out of order} (the charging point is out of order). This choice was made because of the relatively small number of \textit{reserved} and \textit{out of order} records. Therefore, the \textit{other} state could be interpreted as a noisy version of the \textit{offline} state.
Missing data were not filled.

\section{Further insights on the winning strategies}
\label{sec:further}
\subsection*{Arthur75: time-series cross validation}

To select the best model, Arthur75 relied on a 4-fold time-series cross validation. 
More precisely, the training data is separated in six equally long subsets.
The nth cross-validation step consists in training the model on the n+1 th first subsets and evaluating it on the n+2 th subset.
Then, the test loss are averaged and the parameters of  the models are chosen to minimize this averaged test loss.

\subsection*{Charging Boys: ablation study}

The Charging Boys' forecast is an ensemble of models. 
Therefore, it is more complex than what the other teams have proposed. 
To better understand what each model brings to the ensemble, we conduct an ablation study. 
In Table \ref{table_score_charging}, we compute the scores on the private test of each model of the ensemble.
To get a better grasp of the strengths of each model, we also compute these score on each hierarchical level (station, area, and global).
Interestingly, the ARIMA outperforms the other individual models at each hierarchical level. It is also clear that for each hierarchical level the ensemble outperforms the individual models meaning that each component of this complex strategy brings its contribution to the final score.

\begin{table}[ht]
\centering
\begin{tabular}{rrrrr}
  \hline
 & Stations & Area & Global & Total \\ 
  \hline
ARIMA & 148 & 44 & 22 & 214 \\ 
  XGB-reg & 150 & 46 & 26 & 222 \\ 
  XGB-class & 149 & 49 & 23 & 221 \\ 
  Ensemble & 145 & 43 & 22 & 210 \\ 
   \hline
\end{tabular}
\caption{Score by zones of Charging Boys models: ARIMA, XGB-reg, XGB-class, Ensemble.} 
\label{table_score_charging}
\end{table}

\section{Future perpectives: a longer dataset with more features}
\label{sec:perspectives}

To continue the endeavour initiated by the smarter mobility challenge, exogeneous data and additional occupancy records have been collected. This section details the additional resources gathered at \href{https://gitlab.com/smarter-mobility-data-challenge/additional_materials/}{gitlab.com/smarter-mobility-data-challenge/additional\_materials} for the code and \href{https://doi.org/10.5281/zenodo.8280566}{https://doi.org/10.5281/zenodo.8280566} for the datasets.

\subsection*{Adding new features}
We present here an updated version of the data which was conducted after the smarter mobility challenge. This dataset was obtained by merging the Belib dataset with weather \cite{salmon_riem_2016} and traffic datasets \cite{paris_data}. 

\paragraph{Weather}

Weather data covering the same period as the charging point occupancy used for the Smarter Mobility Data Challenge, i.e., from July 2020 to March 2021, was collected using the \texttt{riem} package available in R \cite{salmon_riem_2016}. 3 weather stations were selected from the 3 existing Paris airports: \textit{Orly}, \textit{Le Bourget}, and \textit{Roissy-Charles de Gaulle}. We selected the 4 weather variables the most relevant to the problem at hand, knowingly
\begin{itemize}
    \item Temperature (tmpf) in Fahrenheit,
    \item Relative humidity (relh) in percentage,
    \item Wind speed (sknt) in knots,
    \item Visibility (vsby) in miles.
\end{itemize}
These variables were collected on a half-hourly timestep. To match the 15 minutes timestep of the occupancy data, the missing values were filled in with linear interpolation. Furthermore, the last and first non-NA values were carried forward and backward respectively to fill in missing data at the beginning and end of the dataset.

\paragraph{Traffic}

The traffic data collected on the \textit{Paris Data} platform \cite{paris_data} contains information on the number of vehicles on the road at various key locations in Paris. Every hour, sensors record the number of vehicles passing through a specific location in the city of Paris. Similarly to the weather data, the data has therefore been interpolated and filled to match the 15 minute timestamp of the occupancy data. Only one sensor was kept for each charging station: the closest one according to the Haversine formula that calculates the distance between two latitudes and longitudes \cite{robusto1957cosine}.The traffic variables available are:

\begin{itemize}
    \item flow (q): number of vehicles counted
    \item occupancy rate (k) 
    \item traffic state: 0 for \textit{unknown}, 1 for \textit{fluid}, 2 for \textit{pre-saturated}, 3 for \textit{saturated} and 4 for \textit{blocked}
    \item traffic stopped: 0 for \textit{unknown}, 1 for \textit{open}, 2 for \textit{blocked} and 3 for \textit{invalid} 
\end{itemize}

\paragraph{Benchmark}

We computed the catboost benchmark including the new features over the same period as the challenge. The results are shown in Table \ref{Tab:exo}. As expected, the addition of weather and traffic information marginally improves the prediction performance when the models are trained with enough iterations.

\begin{table}[htbp]
    \centering
    \begin{subtable}{0.4\textwidth}
        \centering
        \begin{tabular}{@{}lll@{}}
        \toprule
        Features                & Public set & Private set \\ \midrule
        Temporal                & 233    & 246     \\
        All                     & 249    & 263     \\ \bottomrule
        \end{tabular}
        \caption{150 iterations}
    \end{subtable}%
    \hspace{1cm} 
    \begin{subtable}{0.4\textwidth}
        \centering
        \begin{tabular}{@{}lll@{}}
        \toprule
        Features                & Public set & Private set \\ \midrule
        Temporal                & 238        & 260         \\
        All                     & 223        & 244         \\ \bottomrule
\end{tabular}
        \caption{300 iterations}
    \end{subtable}
    \caption{Losses calculated for the \texttt{catboost} benchmark model with 150 and 300 iterations only with temporal features and with all features (temporal, weather, traffic)}
    \label{Tab:exo}
\end{table}


\subsection*{Adding new observations}
Regarding the observation period, more data have been collected since the smarter mobility challenge. We provide a raw extraction and some initial data preparation at the following link \href{https://doi.org/10.5281/zenodo.8280566}{https://doi.org/10.5281/zenodo.8280566}. 
This new dataset runs from 28 June 2021 to 15 July 2022 at a 5-minute frequency.

\paragraph{Data Preparation}

This new dataset covers the period from June 2021 to July 2022. The data preprocessing was kept minimal. 
It consists in simply concatenating the monthly extracts from the database, translating the statuses in English and converting the timestamp (in seconds from 1970) to a UTC datetime field.
Both raw and preprocessed data are available.

\paragraph{Statuses and missing values}

\begin{figure}
    \centering
    \includegraphics[width=.5\linewidth]{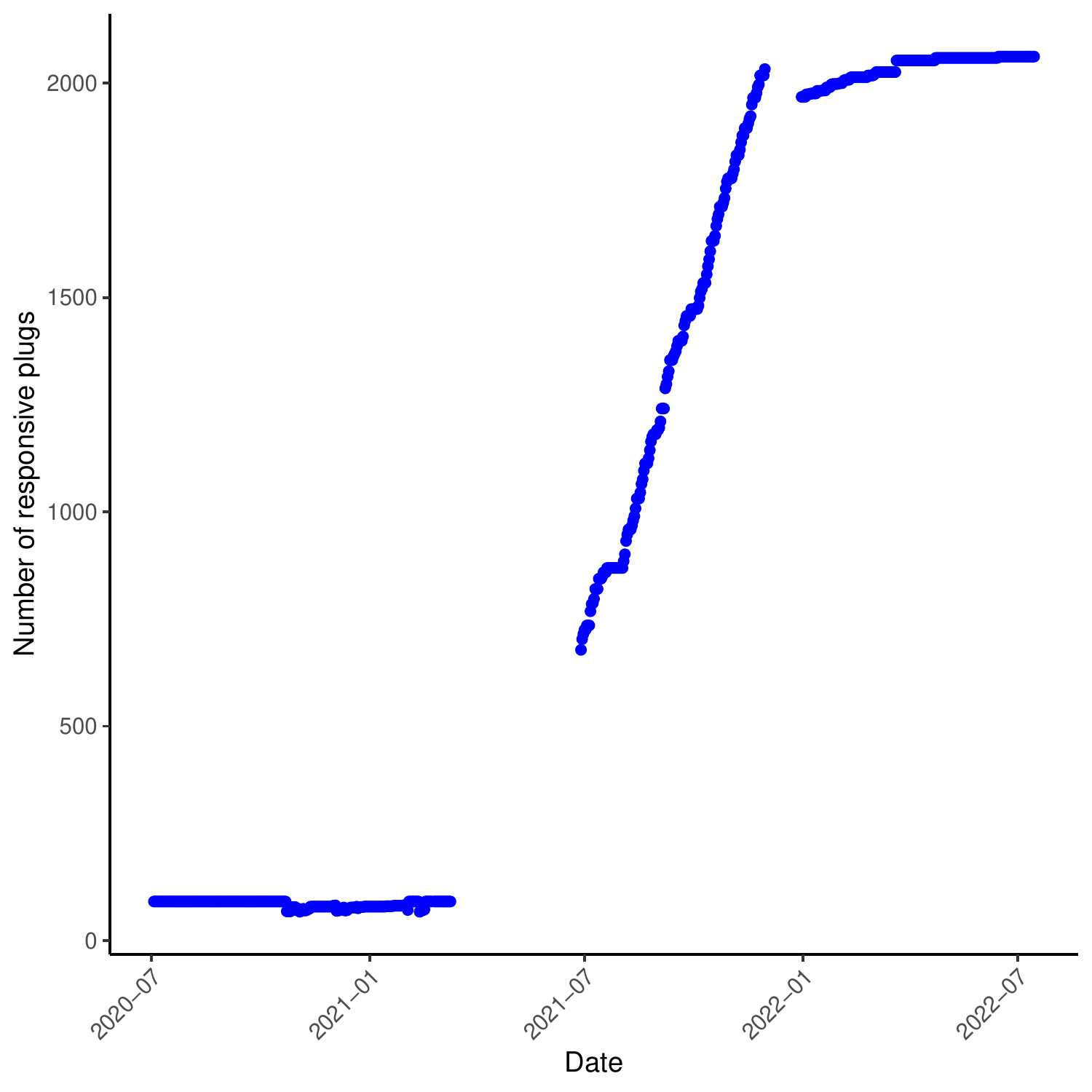}
    \caption{Daily EVSE count}
    \label{fig:evse-count}
\end{figure}

The statuses are different between the version V1 of the dataset (from July 2020 to March 2021) and V2 (from June 2021 to July 2022), as shown in Table \ref{Tab:statuses}.
\begin{table}[htbp]
    \begin{subtable}{0.35\textwidth}
        \flushleft
        \begin{tabular}{@{}ll@{}}
        \toprule
        Status                  & Description \\ \midrule
        Available               & Free EVSE      \\
        Charging                & EV charging    \\
        Passive                 & EV plugged      \\
        Other                   & EVSE out of order     \\
        &\\
        \bottomrule
        \end{tabular}
        \caption{Smarter Mobility Challenge (V1)}
    \end{subtable}%
    \hfill
    \begin{subtable}{0.65\textwidth}
        \centering
            \begin{tabular}{@{}lll@{}}
                \toprule
                Status                         & Description \\ \midrule
                Available                      & Free EVSE   \\
                Charging                       & EV charging at EVSE \\
                In maintenance                 & EVSE being fixed     \\
                Currently being commissioned    & EVSE being deployed     \\
                Unknown                        & EVSE 
                offline 
                \\ \bottomrule
            \end{tabular}
        \caption{Current version (V2)}
    \end{subtable}
    \caption{Possible statuses of the Electric Vehicle Supply Equipment (EVSE) in the previous (V1) and current (V2) versions }
    \label{Tab:statuses}
\end{table}
Notice that the \textit{Passive} state is now part of the \textit{Available} state.
Indeed, after TotalEnergies became the operator of the Belib network on March 2021 (see Section \ref{sec:history}), the information released regarding the stations was updated. 
This explains the missing values between April 2021 and June 2021 and the evolution of the states detailed in Table \ref{Tab:statuses}.
For our analysis, we regroup the \textit{In maintenance} and \textit{Unknown} states and consider them as the \textit{Other} state. 
The \textit{Currently being commissioned} state is related to the extension of the EV park, as shown in Figure \ref{fig:evse-count}, and correspond to the fact that there were some delay between the moments when new stations were built and sent signals, and the moments when they became open to use for EV drivers.
In what follows, we left this state apart.

\begin{figure}
    \centering
    \includegraphics[width=0.7\linewidth]{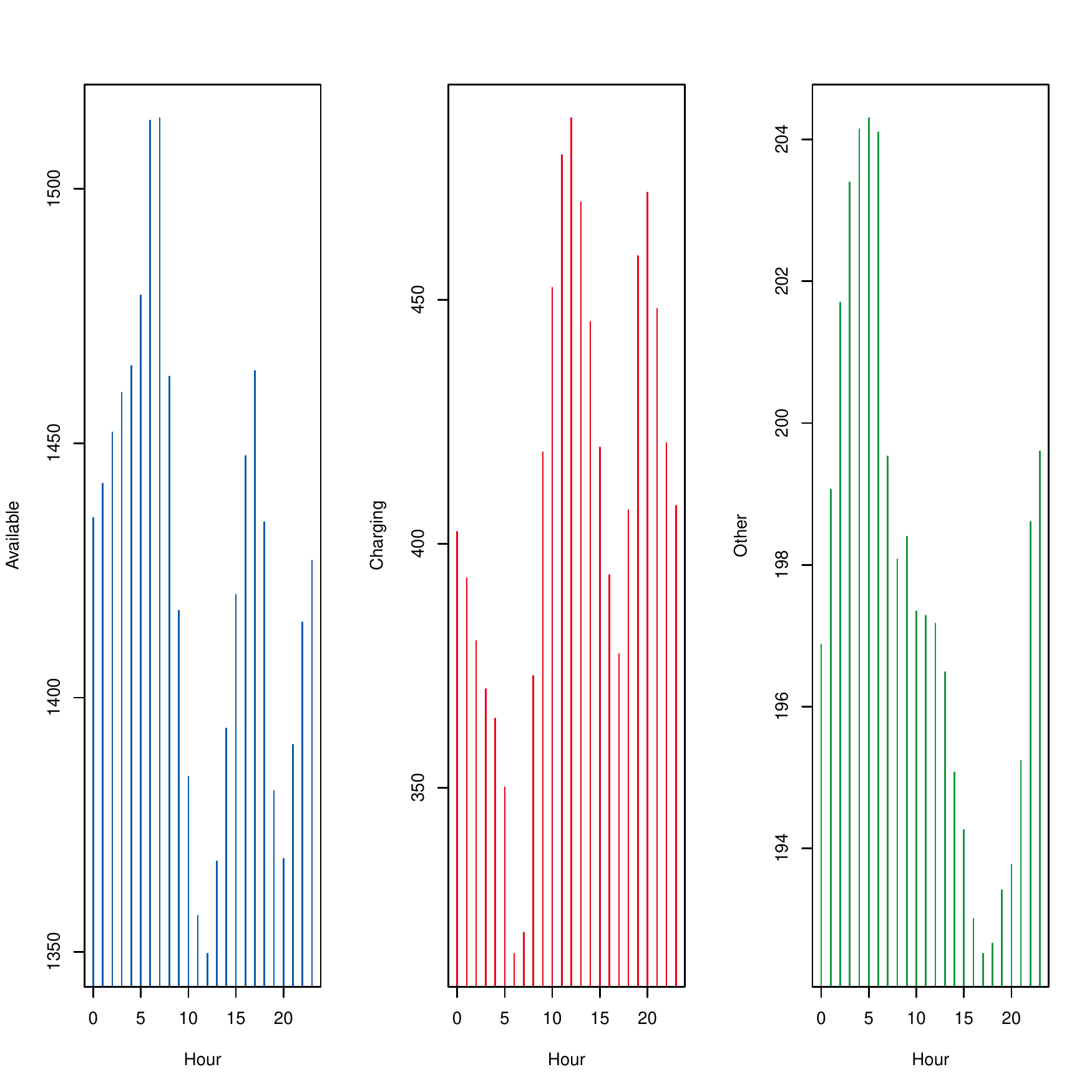}
    \caption{Daily profile at the Global level between July 2021 and July 2022}
    \label{fig:daily_new}
\end{figure}

\begin{figure}
    \centering
    \includegraphics[width=0.7\linewidth]{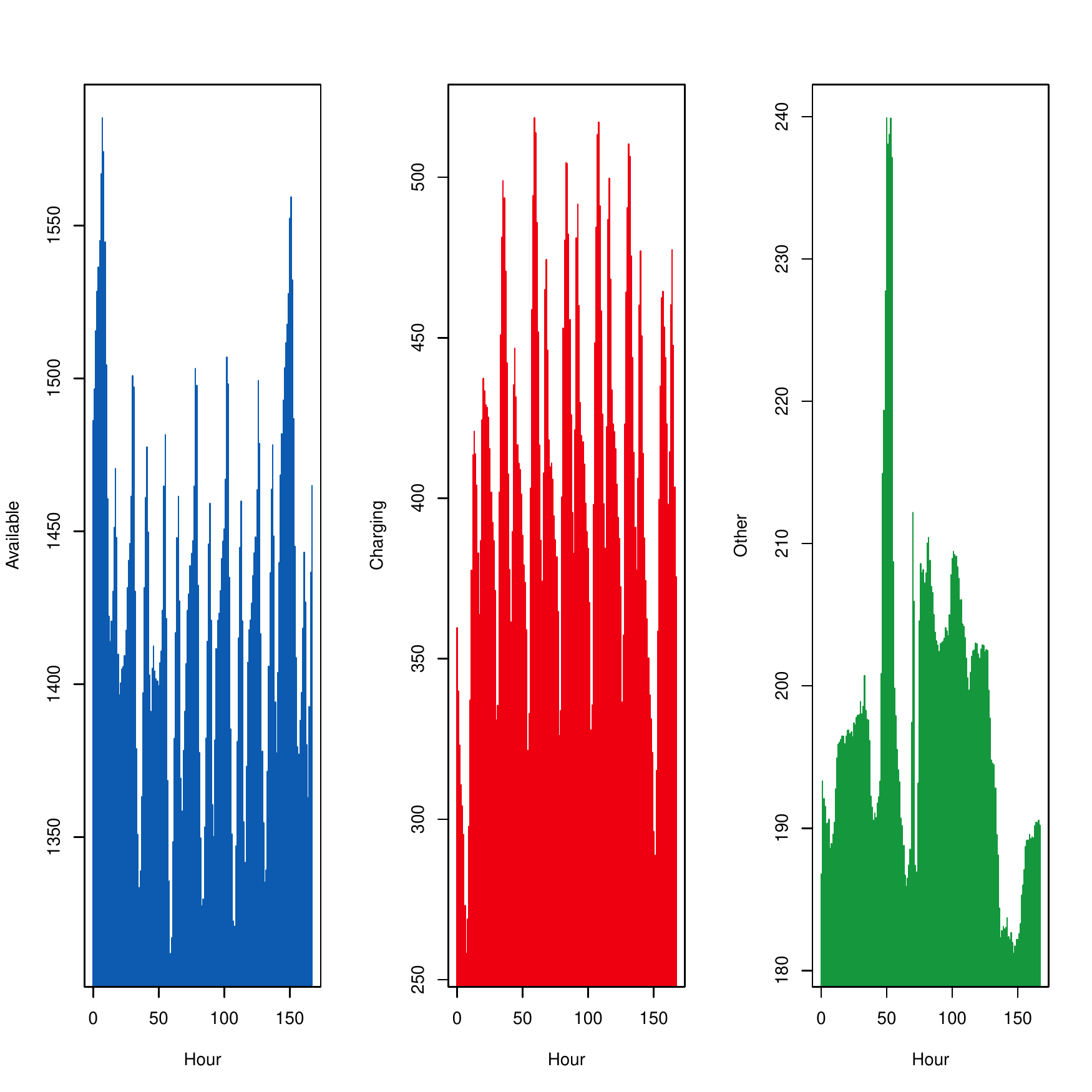}
    \caption{Weekly profile at the Global level between July 2021 and July 2022}
    \label{fig:weekly_new}
\end{figure}

\paragraph{Exploratory data analysis} Similarly to what we did in the EDA for the period spanning from July 2020 to March 2021, we plot the daily and weekly profiles of each status for this new period running from June 2021 to July 2022 in Figures \ref{fig:daily_new} and  \ref{fig:weekly_new}.
Notice that, though the pricing mechanism change on 25 March 2021 (see Section \ref{sec:history}), it still was more advantageous at night (between 8 p.m. and 8 a.m.) for regular users.
This explains why the daily pattern of  the \textit{Charging} state of the new period on Figure \ref{fig:daily_new} is very similar to the one on Figure \ref{calendar_profile}.
In both cases, there are a notable peak at 8 p.m. in \textit{charging} stations, a slow decay in charging stations over night until 5 a.m., and another peak in the morning around 10 p.m. corresponding to commuting behaviors.
The daily pattern of the new \textit{available} state is consistent with the  sum of the former \textit{available} and \textit{passive} states.
Indeed, EV charging spots becoming very attractive around 8 p.m., we observe a drop in the number of \textit{available} plugs.
Then, overnight, as batteries get filled, the stations come from \textit{charging} to \textit{available} (formerly \textit{passive}), which explains the slow increase in the number of \textit{available} plugs overnight.
At 10 a.m., because of commuting behaviors, EV drivers move their cars, which corresponds to  the drop in \textit{Available} plugs.
Once again, commuting behaviors are very visible for the \textit{Charging} state on Figure \ref{fig:weekly_new}, with EV cars being more used during weekdays.
Once again, the daily peaks and drops are very distinctive for  the \textit{charging} state  on Figure \ref{fig:weekly_new}.

\begin{figure}
    \centering
    \includegraphics[width=0.7\linewidth]{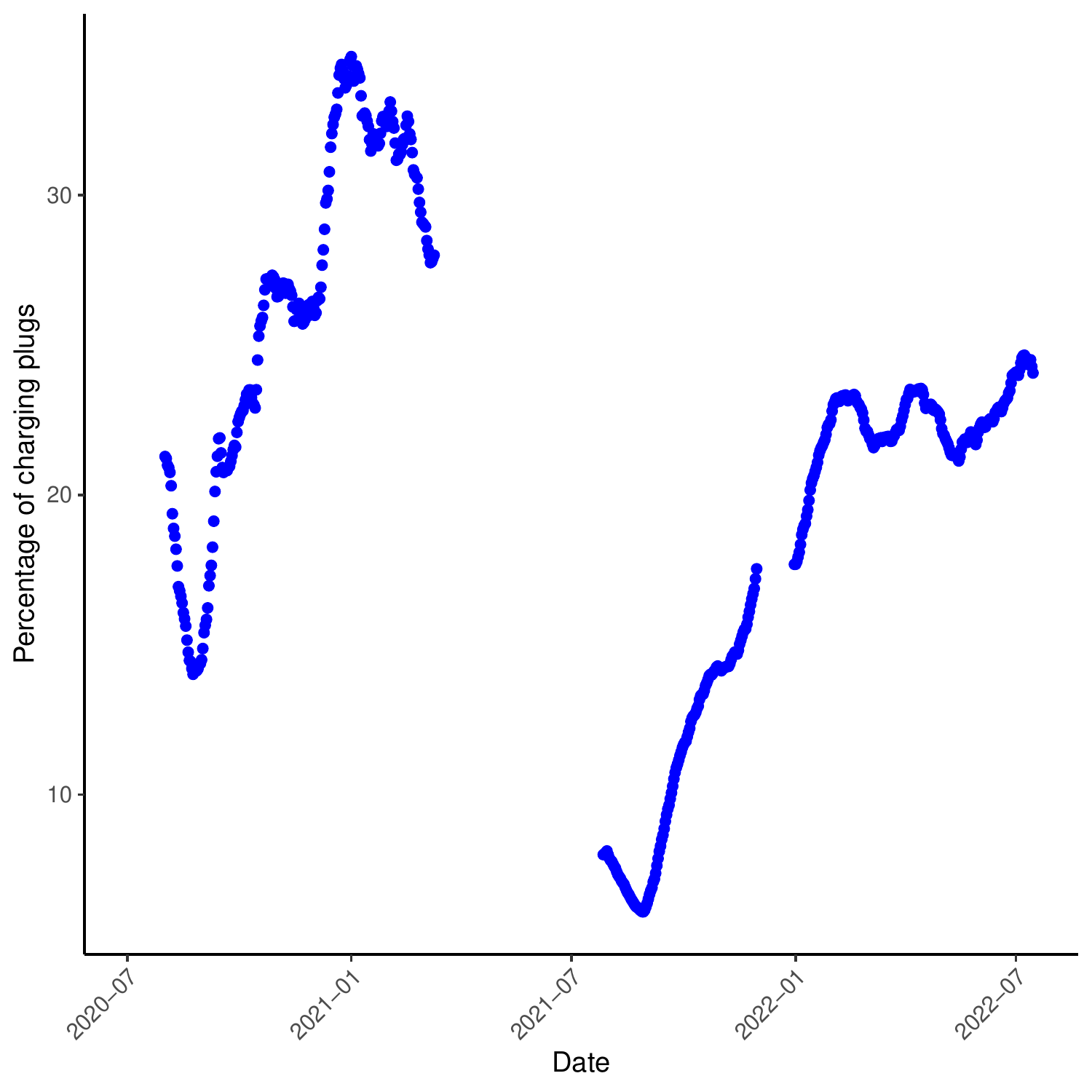}
    \caption{Daily percentage of \textit{Charging} state smoothed on a 30-day window}
    \label{fig:seasonality}
\end{figure}

\paragraph{Seasonality} Since the new dataset spans over two years, it allows to get a grasp of the seasonality of the EV charging demand.
However, the EV charging station park has grown a lot during this period, as evidenced by Figure \ref{fig:evse-count}.
Therefore, one needs to divide the charging demand by the number of plugs.
The evolution of the daily percentage of charging station smoothed with a 30-day window is shown in Figure \ref{fig:seasonality}.
Though  there are  missing values between April 2021 and June 2021, Figure \ref{fig:seasonality} suggests a seasonality in the EV charging demand. Indeed, both in the end of August 2020 and August 2021, the EV charging demand is at its yearly lowest. Then, it increases until it reaches a first peak in October, and a second higher peak in the middle of December.
Notice that both these peaks corresponds to holidays in France.

\renewcommand\thesection{\thechapter.\arabic{section}}

\chapter{Human spatial dynamics for electricity demand forecasting}
\label{ch:contrib-2}
This chapter corresponds to the following paper: \citet{doumeche2023human}.

\section{Introduction}
From 2021 to 2023, Europe has experienced a major energy crisis with energy prices reaching levels not seen in decades \cite{ferrani2023the}. 
Prices rose rapidly in the summer of 2021 as the global economy picked up following the easing of COVID-19 restrictions. Subsequently, the war in Ukraine led to a significant reduction in gas supplies, pushing gas prices even higher \cite{ruhnau2023natural}. 
In this context, the European Union adopted a series of emergency measures to mitigate the effects of this crisis, mainly by reducing electricity demand, with a binding reduction target of 5\% during peak hours \cite{european2022council}.
In France, the government called for a voluntary effort to reduce energy consumption by 10\% over two years and launched its own energy sobriety plan \cite{gouvernement2022plan}.
Various media  documented a subsequent drop in France's electricity demand in the winter of 2022-2023 
\cite{technique2022consommation,thenyt2022asrussia, lemonde2023labaisse}.
Energy saving is also part of France's long-term policy of ecological transition and energy sovereignty.
Indeed, the energy sector's impact on climate change is forcing changes in consumption patterns, which is fueling a growing interest in energy savings and the transition to sustainable energy sources \cite{abdelaziz2011a, rockstrom2017a, hoegh2019the, demaere2020the}.
In France, electricity is one of the most important components of the energy mix, accounting for 25\% of its final energy consumption, and the French Ecological Transition Plan is based on massive electrification driven by decarbonised energy, coupled with energy savings \cite{omar2022decarbonizing, rte2022energy}. 
While modifying human behaviour (e.g., by encouraging remote working) has been identified as an important axis of the sobriety plan, a better understanding of how this relates to energy savings is crucial for energy planning.

 Recently, machine learning techniques have been applied to electricity load forecasting to ensure the electricity grid remains balanced \cite{pinheiro2023short} and to reduce electricity wastage. 
As France's electricity storage capacity is limited and expensive to run, electricity supply must match demand at all times. 
As a result, electricity load forecasting at different forecast horizons has attracted increasing interest over the last few years \cite{hong2020energy}. 
This article focuses on  so-called short-term load forecasting, or 24-hour ahead load forecasting, which is particularly relevant for operational usage in industry and the electricity market  \cite{nti2020review, hammad2020methods}. 
We address this problem both in terms of feature selection and model design.
Most state-of-the-art models rely on historical electricity load data, seasonal data such as holidays or the position of the day in the week, and meteorological data such as temperature and humidity \cite{nti2020review}. 
However, such data cannot accurately account for the complex human behaviours that affect the variability of energy demand, such as holidays and remote working. 
As a result, traditional models struggle to account for unexpected large-scale societal events such as the COVID-19 lockdowns or energy savings following economic, geopolitical, and environmental crises \cite{obst2021adaptative}.
New data capturing consumption behaviours is therefore needed to better model electricity demand. 
Over recent decades, datasets generated from mobile networks, location-based services, and remote sensors in general, have been used to study human behaviour \cite{blondel_understanding_2015}.
Indeed, geolocation from mobile phones makes it possible to precisely characterise human flows \cite{deville_dynamic_2014, blumenstock_predicting_2015, lorenzo_allaboard_2016}. 
For example, such data have been used to study disease propagation \cite{bengtsson_improved_2011, blumenstock_inferring_2012, rubrichi_comparison_2018, pullano_evaluating_2020}, traffic
\cite{xu2021understanding}, the impact of human activities on biodiversity \cite{filazzola2022using}, and water consumption  \cite{terroso2021human, smolak2020applying}. 
In terms of day-ahead load forecasting, mobility data from SafeGraph, Google, and Apple mobility reports were strongly correlated with electricity load drops in the US during the COVID-19 outbreaks \cite{chen2020using, ruan2020cross}, as well as in Ireland \cite{zarbakhsh2022human} and in  France \cite{antoniadis2021hierarchical}. 
These works show that social behaviors like lockdowns and remote working affect significantly the intensity and daily patterns of the electricity load consumption, and that these changes can be predicted by using mobility data.
Although such data is quite informative about activity in urban areas, e.g., in retail stores and train stations, it does not precisely account for human presence and flows. Indeed, there is  intrinsic bias in such data collection, corresponding for example to that of using (or not) a specific application. It is therefore necessary to take into account such biases when building models using this kind of data. 

In this context, the originality of this paper relies on the incorporation of high-quality human presence data provided by the mobile network operator Orange---representing about $40$\%  of the French market--- in adaptive models to forecast the short-term electricity demand during France's 2022--2023 sobriety period  \cite{fluxVision}.
This dataset is based on adjusted mobile phone traffic volume measurements collected continuously and passively at the mobile network level, unlike most location-based services data where the user is required to opt-in, which may introduce biases. 
As a result, our mobile network-based signal can be considered representative of the underlying population's data.
Similar datasets have been used to perform dynamic census with the aim of planning the development of long-term electricity infrastructures in emerging economies \cite{martinezcesena2015usingmobilephonedata, salat2020a, salat2021analysing}, however these models were prospective and were not tested against the state-of-the-art in highly competitive tasks such as  short-term electricity demand forecasting.

In this article, we start by introducing the dataset at hands.
We then show that our mobility data from mobile networks are correlated with other well-known socio-economic indices that capture spatial dynamics of the population. 
Furthermore, we show that models using mobility data outperform the state-of-the-art in electricity demand forecasting by 10\% with respect to usual metrics.
To better understand this result, we characterise electricity savings during the sobriety period in France. 
Finally, we show that 
the {\it work} index we have defined (see Section \ref{sec:mob_dataset}) has a distinctive effect on electricity demand, and is able to explain observed drops in electricity demand during holidays. 
Other human spatial dynamics indices such as tourism at the national level did not prove to have a significant effect on national electricity demand.

The code to replicate the electricity dataset and implement the different models is available at \href{https://github.com/NathanDoumeche/Mobility_data_assimilation}{github.com/NathanDoumeche/Mobility\_data\_assimilation}. The corresponding dataset is available at \href{https://zenodo.org/records/10041368}{zenodo.org/records/10041368}. 
Hence, the change point results shown in Figure~\ref{fig:obst_right}, as well as the dataset and the benchmarks without mobility data of Table~\ref{table_score_target_agg2_orange}, are directly reproducible for future research, and can easily be updated to work for new time periods of interest.
Note, however, that the mobility indices are not publicly available.

\section{Using mobility data to forecast electricity demand} 
The goal of this section is to show how using mobility data leads to better performance in forecasting the French electricity demand during the energy crisis.
\subsection*{Datasets}
 
The reference dataset runs from 08/01/2013 to 28/02/2023. It consists of calendar data (dates and holidays), meteorological data (temperature), and historical data (electricity power load at different time scales). 
In this article, we consider this data to be a reference, because these features are commonly used to build state-of-the-art models in electricity load forecasting \cite{hammad2020methods, hong2020energy, nti2020review}, in particular for the French electricity load \cite{ obst2021adaptative,vilmarest2022state, Vilamarest2024adaptive}.
All these data are public and distributed under the Etalab open source licence. 
The calendar data are extracted from the French open source database \cite{dataGouvJourFeries, dataGouvVacances}. 
This regroups holiday periods according to France's three holiday timetables---in France holidays depend on the region you live in---as well as the French national holidays.
This calendar dataset has no missing values. 
The meteorological data are extracted  from the SYNOP Météo-France database \cite{meteoFrance}.
Météo-France is the French public agency responsible for the national weather and climate service. 
The dataset consists of 3-hourly temperature measurements from 62 meteorological stations located throughout French territory.
This dataset has many missing values, which we have imputed as follows. First, if a station has a missing value at time $t$ and the station's measurements are available 3 hours before and 3 hours after $t$, the missing value is imputed as the mean of these two measurements. If no such values are available, the missing temperature is imputed as the temperature of the nearest station. If however all stations in a region have missing values, the temperature of each station is imputed by taking the mean of the temperature at the same hour from the day before and the day after. 
Finally, the historical electricity load dataset is extracted from the RTE's public releases \cite{rteData}.
RTE (Réseau de Transport d'Electricité) is France's transmission system operator.
It provides high quality data on regional electricity consumption in France with a frequency of 30 minutes. The national electricity load has no missing values, which is valuable since this is the final target throughout this article.

\label{sec:mob_dataset}
In this work, the reference dataset is  complemented by mobility indices. 
These mobile phone data were provided by Orange's Flux Vision  business service  \cite{fluxVision}, in the form of daily presence data reports. 
These include the number of visitors in the 101 geographical areas of mainland France, which correspond to the second level of national administrative divisions. 
For each location and each day, the data are stratified by the type of visitor (resident, usually present, tourist, excursionist, recurrent excursionist) and origin (foreign, local, non-local). 
The mobile phone data were anonymised in compliance with strict privacy requirements and audited by the French data protection authority (Commission Nationale de l'Informatique et des Libertés). 
Computation of the presence data reports is based on the on-the-fly processing of signalling messages exchanged between mobile phones and the mobile network, usually collected by mobile network operators to monitor and optimise mobile network activity. 
Such messages contain information about the identifiers of the mobile subscriber and of the antenna handling the communication, the timestamp, and the type of event (e.g., voice call, SMS, handover, data connection, location update). 
Knowing the location of antennas makes it possible to reconstruct the approximate position of a communication device.
All these data were then used to compute the total number of individuals in a given area, without saving any residual information that could be traced back to the individual users. 
More specifically, at any given day, each individual was characterised based on their pattern of movement and their origin as follows. 
\begin{itemize}
\item Resident: person whose spends much of their time in the study area, and spent at least 22 nights (not necessarily consecutive) there over the past eight weeks.
\item Usually present: person who is not a resident of the study area but has been seen in the study area repeatedly: more than four times in different weeks during the previous eight weeks.
\item Tourist: person who spends the night in the study area who is neither resident nor usually present.
\item Excursionist: person not staying overnight the night before and the night of the study day, and present less than 5 times during the day in the last 15 days. 
\item Recurrent excursionist: person who has not spent the night before and the current night in the study area and who has been present more than five times during the day in the previous 15 days.
\end{itemize}
The night corresponding to a given day is the period between 8 p.m. of that day and 8 a.m. of the following day. Moreover, origin is categorised as follows:
\begin{itemize}
\item Foreign: person with a foreign SIM card.
\item Local: person with a billing address in the study area.
\item Non-local: person with a billing address outside the study area.
\end{itemize}
This data was then corrected by Orange Flux Vision to account for spatial and temporal biases, and so to be representative of the general population. 
To this end, they use spatially-stratified market share data, socio-economic data from the national statistics institute (Insee), mobile phone ownership data also from Insee, and
customer socio-demographic information provided upon subscription.
From these data, we constructed three indices. The \textit{work} index corresponds to the number of recurrent excursionists, the \textit{tourism} index to the number of foreign plus non-local tourists, and the \textit{resident} index to the number  of  residents plus ``usually presents''.
In this article, the mobility dataset covered the periods from 01/07/2019  to 01/03/2020, from 01/07/2020 to 01/03/2021, from 01/07/2021 to 01/03/2022, and from 01/07/2022 to 01/03/2023.

Relying on such mobility indices 
is an original new way of tracking human mobility to better characterise electricity demand.
Indeed, mobility is a complex signal, since the vibrancy of places varies over years, but also over the course of a day. 
For example, it is known that individuals circulate through a number of places throughout the day---on average between 2.5 and 4 per person in French metropolitan areas \cite{mtect}---month, and year, whether for housing, work, education, personal relationships, and leisure.
Determining the appropriate index and its level (i.e., the geographical scale at which it is aggregated) to measure a given phenomenon  are both major difficulties in analysing mobility dynamics \cite{pora2023telework}.
Our high quality dataset was designed to precisely quantify human presence over France at a very high frequency with respect to census or survey data, and has already been studied as such to account for residential behaviour \cite{Levy2023who}.
Its advantage is that it allows one not only to quantify with a high degree of accuracy the population present at a given time and place, but also to characterise the way they inhabit that place (i.e., residing, working, or exploring).

\begin{figure}
    \centering
    \includegraphics[width = 0.5\textwidth]{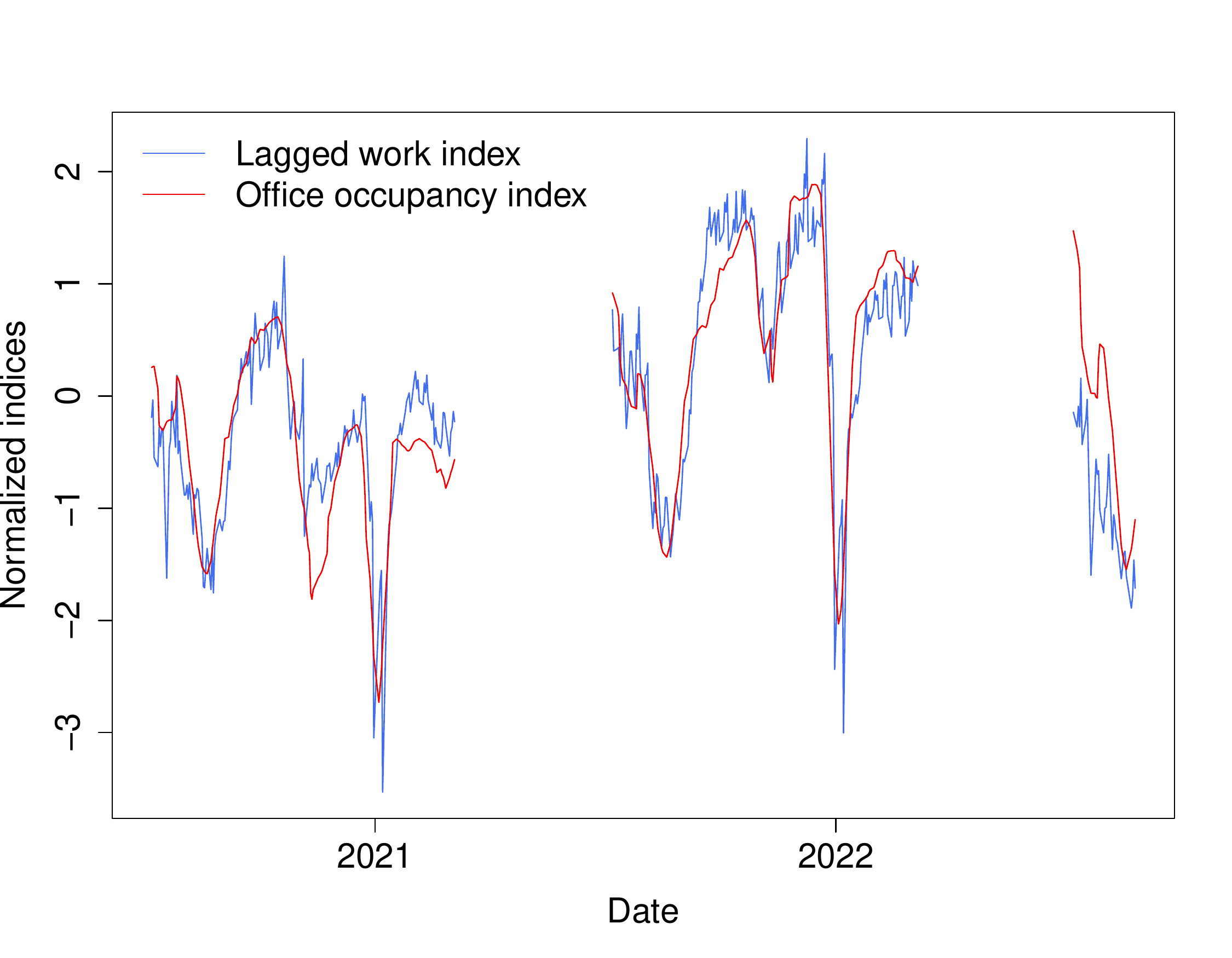}
    \caption{Comparison of work indices. Mobile network-based work index and the normalcy index's office occupancy one. The work index in blue is lagged by 7 days. Weekends and bank holidays are excluded. Both indices have been standardised, i.e.,  the empirical mean has been subtracted and the result divided by the empirical standard deviation. The mobile network dataset only covers the period from July to March each year.}
    \label{fig:normalcy}
\end{figure}

To investigate the ability of our dataset to characterise work behaviour, we compare it to the \textit{office occupancy} index from The Economist's normalcy index \cite{theeconomist2022theglobal}. 
This index was developed during the COVID-19 pandemic to evaluate the impact of 
government policies on human behaviour.
It tracks eight variables (sports attendance, time at home, traffic congestion, retail footfall, office occupancy, flights, film box office, and public transport) at the national level; this open data can be found here:
\url{https://github.com/TheEconomist/normalcy-index-data}. 
The \textit{office occupancy} index is derived from  Google's COVID-19 community mobility reports, which are no longer being updated as of mid-October 2022 \cite{googleCovid}.
Similar to our \textit{work} index, the \textit{office occupancy} index measures the tendency of workers to work on-site rather than remotely.
As the \textit{office occupancy} index is only available from February 2020 to October 2022, and mobile-network dataset only covers the periods form July to February of each year, Figure~\ref{fig:normalcy} has missing values.
As illustrated in Figure~\ref{fig:normalcy}, the office occupancy variable is highly correlated  (87\%) with the 7-day lagged \textit{work} index when excluding weekends and bank holidays. 
Moreover, our \textit{work} index is more suitable than the \textit{office occupancy} index for operational use, because it is seven days ahead of the \textit{office occupancy} index, meaning that it captures the variations in office occupancy earlier.
As detailed in Section~I.C. of the Supplementary material, the \textit{work} index is also more informative than the \textit{office occupancy} index, because it captures the reduction in office occupancy during weekends and holidays. 
 This is very valuable because holidays are known to  have a significant impact on  electricity demand, while their effect is difficult to evaluate. 
This often leads to having to analyse regular days and holidays separately \cite{Krstonijevic2022adaptive}.
In addition, in Section~I.B. of the Supplementary material, we demonstrate how tourism trends are related to another index from the same dataset.

\subsection*{Mobility data and  electricity demand forecasting}

\label{sec:benchmark}
We run a benchmark of state-of-the-art models to measure the benefits of incorporating mobility data into 
load forecasting techniques (see Section~II of the Supplementary material for a more complete description of the models). 
In this field, the state-of-the-art is generally divided into three classes of forecasts \cite{Singh2012Load, wang2023benchmarks}: statistical models that approximate electricity demand by simple relationships between explanatory variables, data assimilation techniques that update a model using recent observations, and data-driven machine learning methods whose results may be more difficult to explain but are more expressive. 
 
Here, we focus on the state-of-the-art in French load forecasting, during the energy crisis, thereafter referred to as the sobriety period.
Section~\ref{sec:energy_savings} explains how this period was precisely determined. 
To evaluate the benefits of using mobility data to forecast France's national electricity load, we thus run a benchmark on this sobriety period, i.e., from 01/09/2022 to 28/02/2023. The training period spanned 08/01/2013 to 01/09/2022. Results are presented in Table~\ref{table_score_target_agg2_orange} in terms of root mean square error (RMSE) and mean absolute percentage error (MAPE). 
Bold values highlight the best forecasts in each category of models. 

Indeed, models were evaluated according to the following test errors. Let $T_{test}$ be the test period, $(y_t)_{t \in T_{test}}$ the target, and  $(\hat y_t)_{t \in T_{test}}$ an estimator of $y$. The root mean square error is defined by
$\mathrm{RMSE}(y, \hat y) = (\frac{1}{T_{test}}\sum_{t= \in T_{test}} (y_t-\hat y_t)^2)^{1/2}$
and the mean absolute percentage error is defined by
$\mathrm{MAPE}(y, \hat y) = \frac{1}{T_{test}}\sum_{t \in T_{test}} \frac{|y_t-\hat y_t|}{|y_t|}$.
Both  errors are useful for operational uses. Since the sampling of time series are dependent, confidence intervals were obtained by time series bootstrapping \cite{lahiri2023resampling} using the \texttt{tseries} package \cite{trapletti2023tseries} forecasting.
All the models, as well as their weights and their optimization, are direct reproductions from state-of-the-art benchmarks.
Indeed, the GAM model was extracted from \cite{obst2021adaptative}, 
the static and dynamic Kalman filters were adapted from \cite{vilmarest2022state}, 
the Viking algorithm comes from \cite{Vilamarest2024adaptive}, the
 GAM boosting is from \cite{bentaieb2014a}, and
the random forest and random forest with bootstrap  were taken from \cite{gohery2023random}.
A full description of these models can be found in Section~II of the Supplementary material.
Note that we have not included neural networks in this benchmark because they have not shown state-of-the-art performance in forecasting the French electricity load \cite{Vilamarest2024adaptive, campagne2024machine}.

Overall, Table~\ref{table_score_target_agg2_orange} shows that incorporating mobility data improves the performance of all models.
In particular, the best forecast using the mobility data (aggregation of experts) has a lower error than the best forecast without mobility data, with a performance gain of about 15\% in RMSE and 10\% in MAPE. 
The NA values in the table are due to the fact that neither persistence nor SARIMA include exogenous data, making it impossible to include the mobility data in these models.
These gains are statistically significant, because they leave the confidence intervals obtained by bootstrapping.
Furthermore, the ranking of the models is consistent with past studies \cite{obst2021adaptative, vilmarest2022state}. 
We remark that the time series bootstrap in the \textit{Random forests + bootstrap} improves the performance of the random forest algorithm without mobility data---confirming the results of \cite{gohery2023random}---but is not the case when adding mobility data.
\begin{table*}[!t]
\caption{Benchmark with and without mobility data. The numerical performance is measured in RMSE (GW) and MAPE (\%).}
\centering

\begin{tabular}{|l|c|c|}
\hline
& Without  mobility & With  mobility  \\
  \hline
  \textit{Statistical model} &&\\
  Persistence (1 day) & 4.0$\pm$0.2 GW, 5.5$\pm$0.3 \%& N.A.,  N.A.\\
  SARIMA  &  2.4$\pm$0.2 GW,  \textbf{3.1}$\pm$0.2 \%& N.A., N.A. \\
  GAM & 2.3$\pm$0.1 GW, 3.5$\pm$0.2 \%  & \textbf{2.17}$\pm$0.08 GW, 3.3$\pm$0.1 \% \\
  \hline
    \textit{Data assimilation }&&\\
  Static Kalman & 2.1$\pm$0.1 GW, 3.1$\pm$0.2 \%  &  1.72$\pm$0.08 GW,  2.5 $\pm$0.1 \% \\
  Dynamic Kalman & 1.4$\pm$0.1 GW, 1.9$\pm$0.1  \% & 1.20$\pm$0.08 GW,  1.7 $\pm$0.1 \% \\
    Viking & 1.5$\pm$0.1 GW, 1.8$\pm$0.1 \% &  1.24$\pm$0.07 GW,  1.7 $\pm$0.1  \% \\
    Aggregation & 1.4$\pm$0.1 GW,  1.8$\pm$0.1 \% & \textbf{1.16}$\pm$0.07 GW,  \textbf{1.6} $\pm$0.1  \% \\
    \hline
    \textit{Machine learning}&&\\
    GAM boosting & 2.6$\pm$0.2 GW, 3.7$\pm$0.2 \%& 2.4$\pm$0.1 GW, 3.5$\pm$0.2 \% \\
    Random forests &  2.5$\pm$0.2 GW, 3.5$\pm$0.2 \%& \textbf{2.0}$\pm$0.1 GW, \textbf{2.7}$\pm$0.2 \%\\
    Random forests + bootstrap& 2.2$\pm$0.2 GW, 3.0$\pm$0.2 \% & \textbf{2.0}$\pm$0.1 GW, \textbf{2.7}$\pm$0.2 \%\\
   \hline
\end{tabular}
\label{table_score_target_agg2_orange}
\end{table*}

Moreover, holidays are known to behave differently from regular days \cite{Krstonijevic2022adaptive}. We therefore ran the same benchmark (see Table~I in the Supplementary material) when excluding holidays; the results suggest that incorporating mobility data still significantly improves forecasting performance.
These results all suggest that adding mobility data leads to RMSE and MAPE gains of around 10$\%$ when forecasting French electricity demand.

\section{Explainability of the models}
The goal of this section is to justify how we defined the sobriety period, and to better understand the impact of mobility on electricity demand.
\subsection*{Defining the sobriety period} 
\label{sec:energy_savings}
The energy crisis in France had a significant impact on electricity demand in terms of electricity savings.
To quantify the savings due to the crisis, and distinguish them from other unrelated changes, the effects of temperature and time seasonality must be removed from the French electricity demand data.
The expected load given temperature and time, which we denote $\mathrm{L}\widehat{\mathrm{oa}} \mathrm{d}$, is estimated using a generalised additive model (GAM).
\begin{figure*}[!t]
\centering
\subfloat[]{\includegraphics[width=0.49\textwidth]{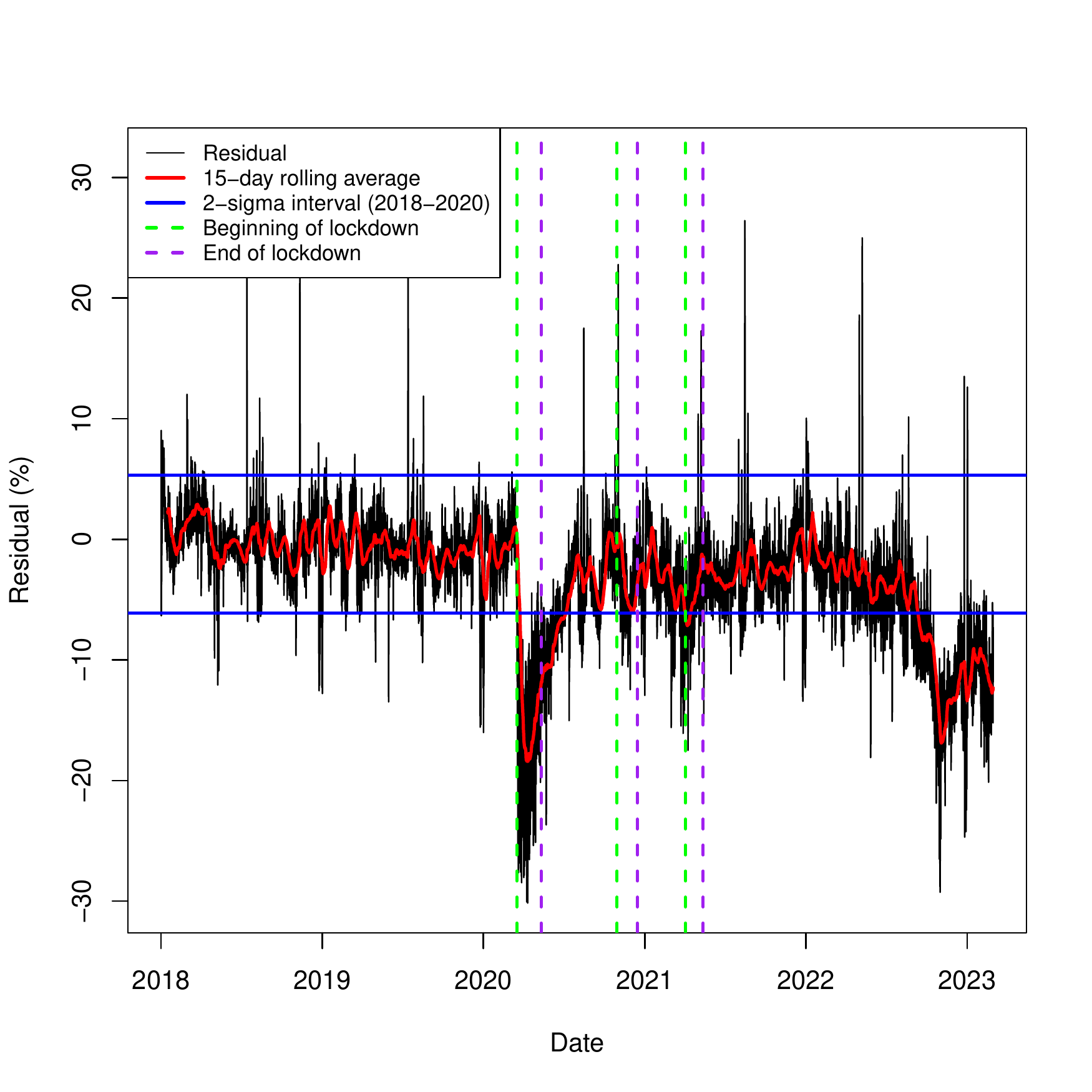}%
\label{fig:obst_left}}
\hfil
\subfloat[]{\includegraphics[width=0.49\textwidth]{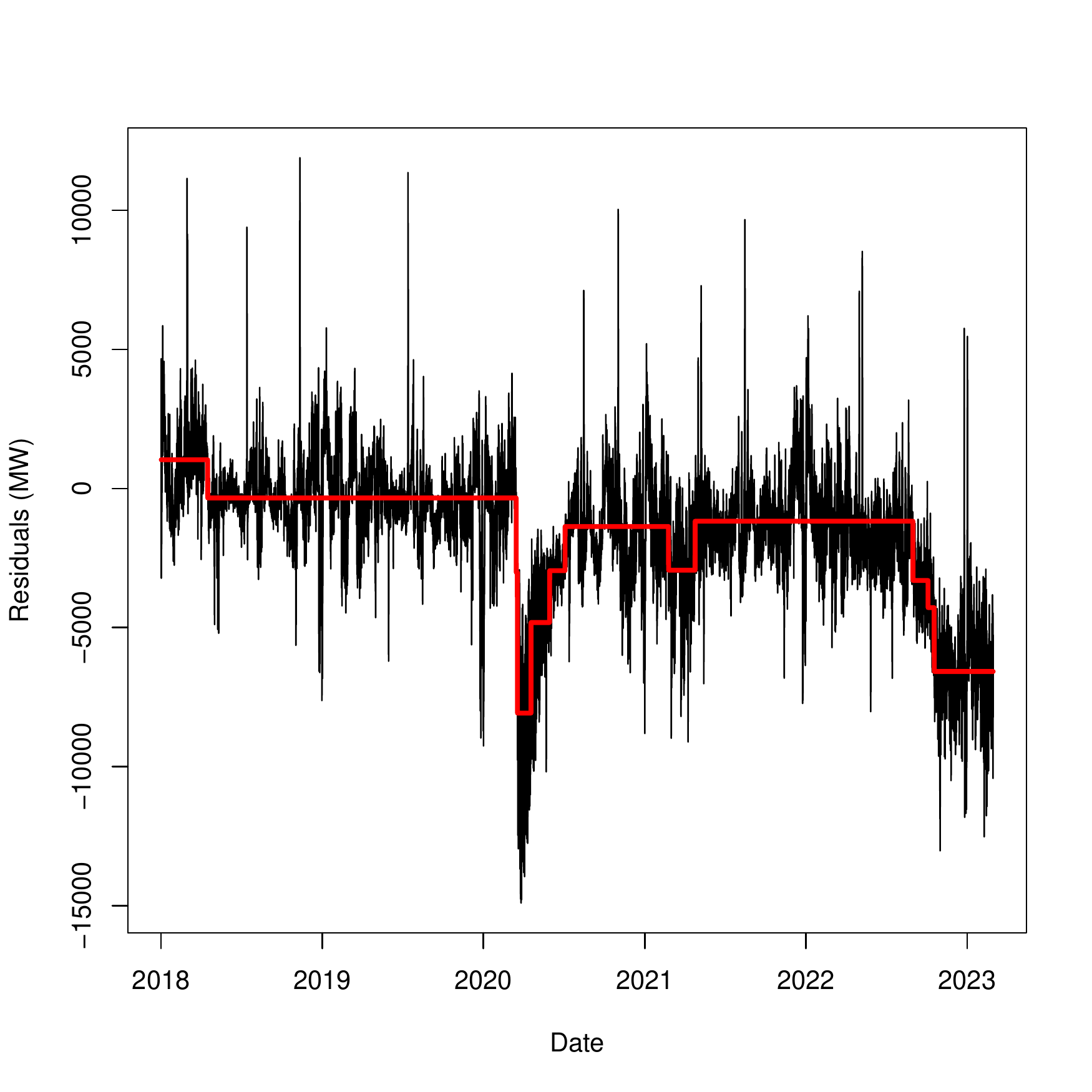}%
\label{fig:obst_right}}
\caption{Electricity demand corrected for the effects of temperature and annual seasonality. (a) Descriptive statistics of the residuals.  (b) The ten most important change points are represented by a change in the red line. The red line is the mean of the residuals between the change points.}
\end{figure*}
Figure~\ref{fig:obst_left} shows the residuals: $\mathrm{res} = \mathrm{Load} - \mathrm{L}\widehat{\mathrm{oa}} \mathrm{d}$, where Load is the actual value of the electricity demand.
This GAM was trained on the data from 01/01/2014 to 01/01/2018. The residuals were then evaluated from 01/01/2018 to 01/03/2023.
Therefore, residuals measure the gap between the electricity demand at a given time and the expected demand with respect to its time and temperature dependency between 2014 and 2018. Negative residuals correspond to electricity savings.  

In Figure~\ref{fig:obst_left}, the blue lines represent the 2$\sigma$ variations over the period spanning 01/01/2018 to 01/01/2020, and  correspond to typical variation in electricity demand around its expected value given the temperature and seasonal data.
The holidays deviate strongly from the expected trend and correspond to the peaks in the residuals.
Note that the 15-day rolling average in red only exits this confidence interval during the lockdowns and the 2022--2023 winter's sobriety period. 
This means that, during these events, the French electricity load is significantly lower than its expected values. 
As shown in Figure~\ref{fig:obst_right}, to detect these changes in the electricity demand, we ran a change point analysis \cite{killik2014changepoint, Aminikhanghahi2017a}  
using binary segmentation, which detects and orders the changes in the mean of the residuals.
The two most important change points of the 2018-2023 period were at the beginning of the sobriety period (04/10/2022) and the beginning of the first COVID-19 lockdown (15/03/2020).  During the sobriety period from 04/10/2022 to 01/03/2023, the residuals had a mean of -10.6$\%$. This result was close to the assessment made by the French transmission system operator (RTE) whose estimate was of a 9$\%$ decrease in consumption during the winter of 2022-2023 \cite{rte2023}.
Figure \ref{fig:obst_right} shows the ten most important change points in the residuals over the 2018-2023 period along with mean of the residuals between the change points.
These results confirm that there was indeed a significant drop of around 11$\%$  in French electricity demand during the sobriety period of 04/10/2022 to 01/03/2023. As this drop is visible in the electricity load adjusted for temperature and time, this means that temperature and seasonal data are not sufficient to accurately explain these energy savings.

\subsection*{Explaining the impact of mobility on electricity demand}
In this section, we use variable selection to offer insights into the performance of forecasts that use mobility data. 
We also investigate the link between electricity demand and the \textit{work} index---which emerges as the second most explanatory variable in our variable ranking (see below). 

Combining the calendar, meteorological, electricity, and mobile network datasets resulted in 38 features. 
Some of these features are highly correlated, e.g., see Section~I.A of the Supplementary material for details on the correlation between the \textit{temperature} and the \textit{school holidays} features.
Thus, to create highly explainable and robust forecasts, it is necessary to select a smaller number of highly explanatory features, in order to better understand how they relate to electricity demand.
Nevertheless, typical variable selection methods based on cross-validation \cite{wasserman2009high, Huang2010variable, marra2011practical} are not directly applicable to time series, because the samples are not independent. 
To reduce  the dimensionality of the problem, one solution is to rank the features by order of importance and then select the most important ones \cite{genuer2010variable}. 
In this paper, we considered three such ranking methods: minimum redundancy maximum relevance (mRMR), Hoeffding D-statistics, and Shapley values.
For multivariate time series, feature selection can be performed using the mRMR algorithm, which consists in selecting variables that maximise mutual information with the target \cite{peng2005feature, han2015joint} (here the electricity load). 
We used the \textit{mRMRe} package \cite{De2013rRMRe} in \texttt{R} for this analysis; the most important variables---in decreasing order of importance---were \textit{temperature}, the \textit{work} index, and the \textit{time of year}. 
The Hoeffding D-statistic ranking and the Shapley value ranking results are detailed in 
Section~IV.A of the Supplementary material.
All three rankings gave the same results, implying that the \textit{work} index is the second most inportant feature, after \textit{temperature}, and is more important than the calendar data.  
We note also that these analyses suggest that the \textit{tourism}  and \textit{residents} indices do not appear to be of great importance with respect to France's electricity demand (See Section~IV.A of the Supplementary material).

\begin{figure*}[!t]
\centering
\subfloat[]{\includegraphics[width=0.49\textwidth]{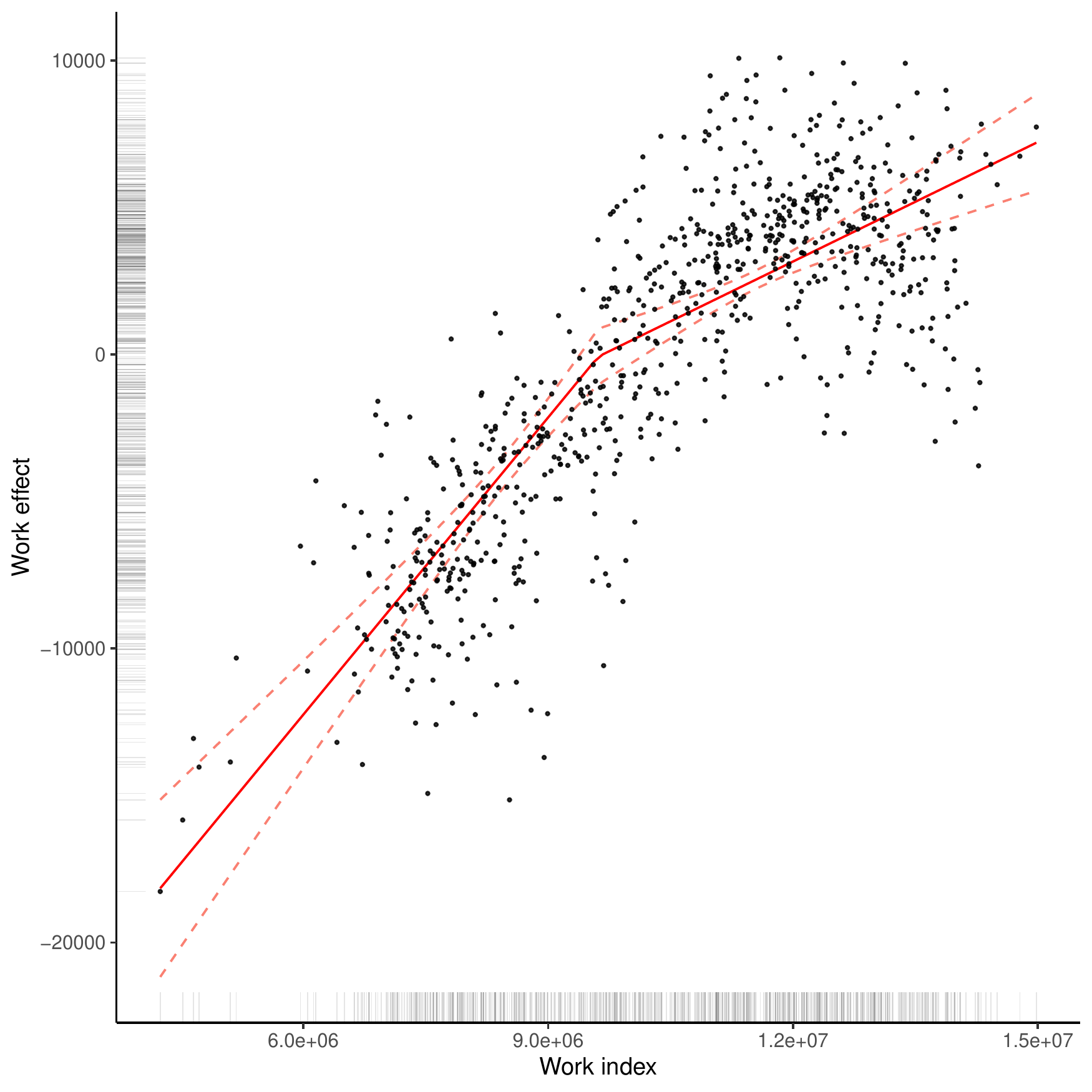}%
\label{fig:gam_effect_left}}
\hfil
\subfloat[]{\includegraphics[width=0.49\textwidth]{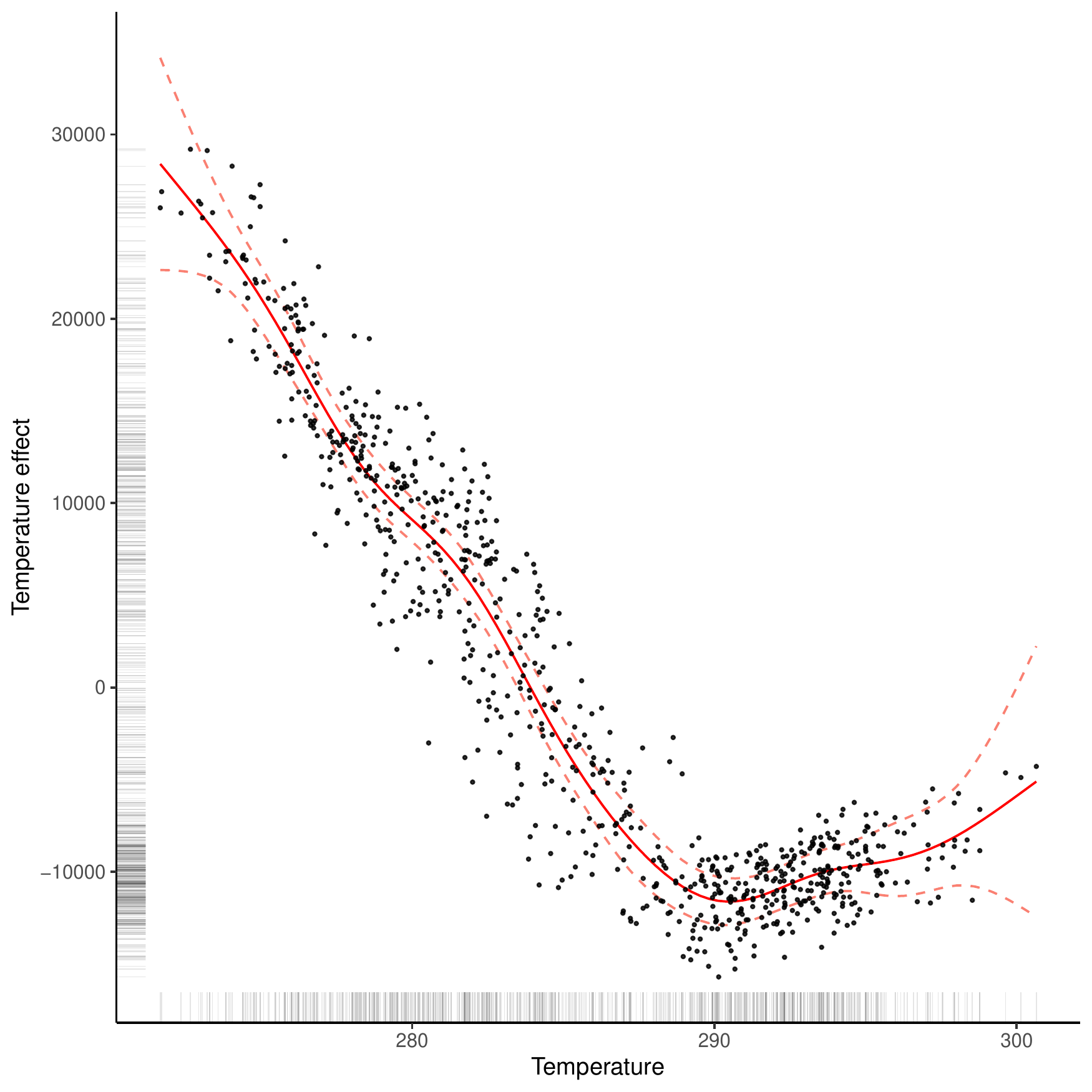}%
\label{fig:gam_effect_right}}
\caption{Effects of features on electricity demand. Each black point is an observation at 10 a.m. The effect given by the GAM regression is shown in red on both plots. Dotted red lines correspond to the 95$\%$ confidence interval of the effects.  (a) Temperature-corrected  electricity load as a function of the work index. (b) Work-index corrected electricity load as a function of temperature.  }
\end{figure*}

\label{sec:stat_ana}
Outperforming the state-of-the-art in Table~\ref{table_score_target_agg2_orange} indicates that mobility data had explanatory power when it came to electricity demand during the sobriety period. 
However, this result does not provide any formal insight into the future performance of the index. 
We therefore ran a statistical analysis of the predictive ability of this mobility data.
Moreover, state-of-the-art data assimilation techniques being difficult to analyse, we restrict ourselves to explainable models of the electricity demand. 
Since the effect of temperature is known to be nonlinear, we consider generalized additive models (GAMs) instead of standard linear regression.
As temperature and then the \textit{work} index were ranked as the two most important variables in our variable selection phase, we consider the electricity demand corrected for the effect of temperature.
Figures~\ref{fig:gam_effect_left} show that the electricity demand increases with the work index, i.e., the higher the number of people at work, the higher the electricity demand.
\begin{figure*}[!t]
\centering
\subfloat[]{\includegraphics[width=0.49\textwidth]{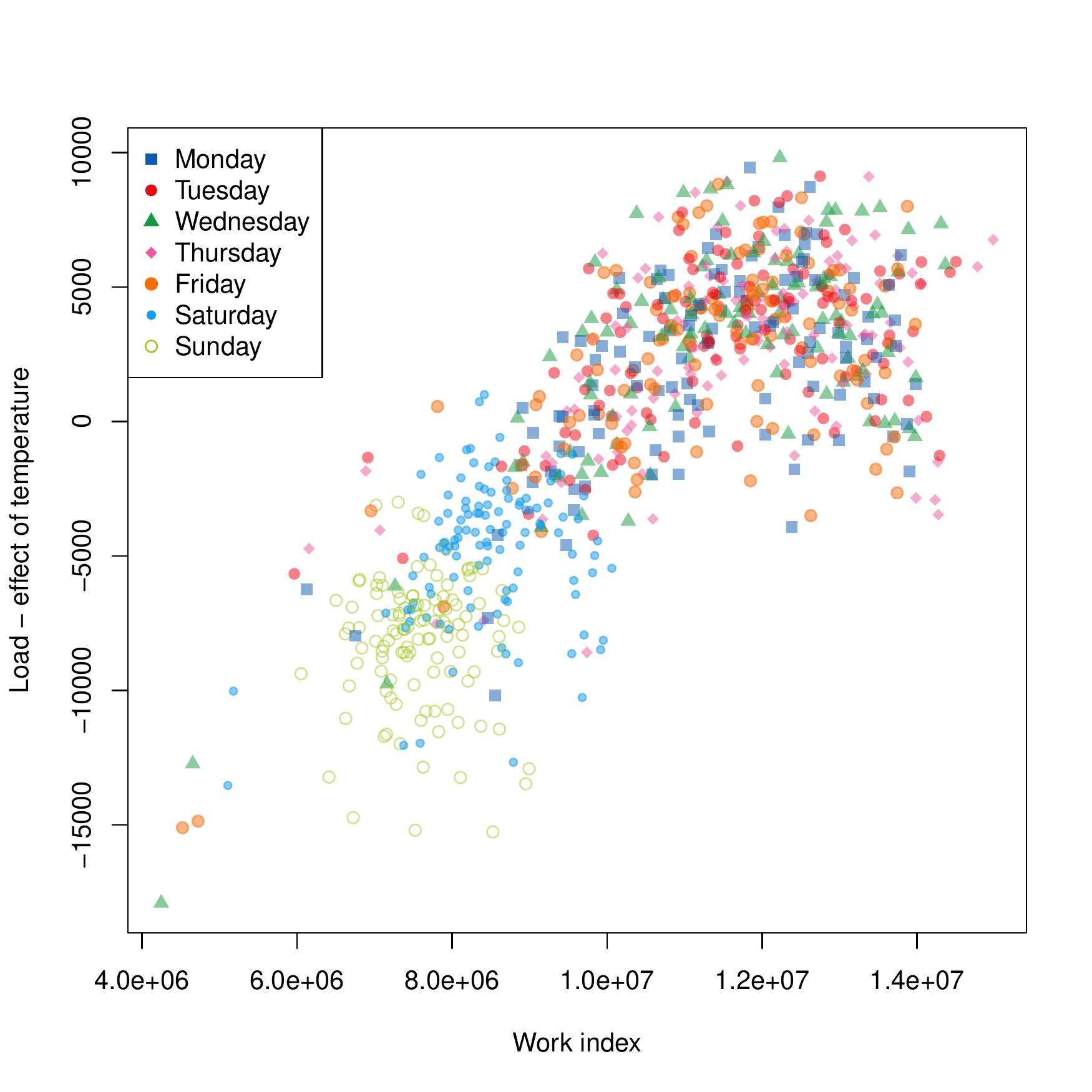}%
\label{fig:mob_left}}
\hfil
\subfloat[]{\includegraphics[width=0.49\textwidth]{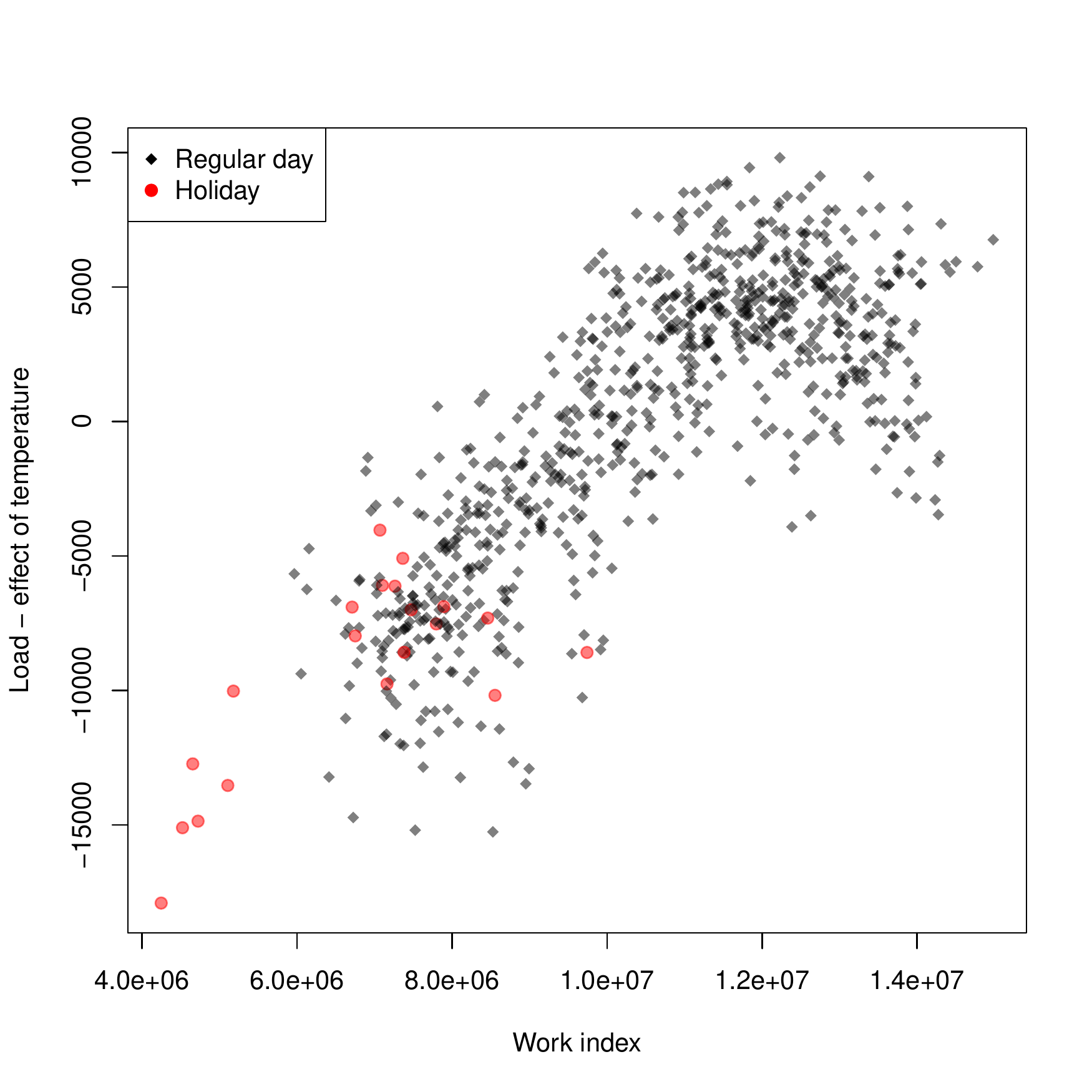}%
\label{fig:mob_right}}
\caption{Dynamics captured by the work index. Each point is an observation of the electricity load corrected for temperature as a function of the work index at 10 a.m. between July 2019 and March 2022.  (a) Dependence of the \textit{work} index on the day of the week. (b) The holiday pattern.}
\end{figure*}

Moreover, the \textit{work} index accounts for several consumption behaviours.
First, Figure~\ref{fig:mob_left} shows how the index accounts for the effect of weekends, thus capturing weekly seasonality related to typical work behaviour. 
Indeed, we see that weekdays, Saturdays (in purple), and Sundays (in yellow) correspond to specific clusters of points with a lower \textit{work} index. 
Figure~\ref{fig:mob_right} shows that the \textit{work} index is related to consumption differences during the holidays (in red) vs other days (in black). 
Note that they have the same relationship as on regular days. 
Therefore, the \textit{work} index summarises in a single feature both the effects of the day of the week and the holiday (7 features).

Moreover, the analysis of the impact of our \textit{work} index on electricity demand---when fixing the day of the week and excluding holidays, shows that lower \textit{work} index correspond to lower electricity demand (see Section~IV.B. in the Supplementary material).
This shows that lower work dynamics are associated with energy savings.
As expected, the effect is more pronounced during working hours.
As a result, the work index is more informative than calendar information alone. 
In fact, models using the \textit{work} index performed better on the atypical event of the sobriety period than models based on calendar data, which only capture seasonality in stationary signals (see Table~IV in the Supplementary material). 
This suggests that the \textit{work} index is explanatory of electricity demand.

\section{Conclusion}
In this work, we have shown that the period spanning September 2022 to March 2023 was atypical in terms of France's electricity demand.
During this so-called sobriety period, we observed a decrease in electricity demand similar to what happened during the first COVID-19 lockdown.
However, this period of significant electricity savings lasted for over six months, which is much longer than the 1-month COVID-related period.
These observations are consistent with those of French media and France's transmission system operators. 
These results suggest that additional phenomena to annual seasonality and temperature are responsible for the recent significant changes in the electricity consumption behaviour.

As evidenced by our benchmark in Table \ref{table_score_target_agg2_orange}, standard statistical models such as GAMs struggled during the sobriety period. Indeed, for the same state-of-the-art GAM, the RMSE and the MAPE when excluding holidays were respectively $55 \%$ and $87 \%$ higher than in the same test period two years earlier (September 2019 to March 2020) \cite{obst2021adaptative}.
Relying on a benchmark specific to France electricity load forecasting, we have shown that including mobile network mobility data in the analyses improves the state-of-the-art performance by  around 10\%.
Although evaluating the cost of load forecasting error is a difficult task, it has been estimated that a 1$\%$ reduction in load forecasting error would save an energy provider up to several hundred thousand USD per GW peak \cite{hong2016probabilistic}. 
In 2022, the average daily load peak in France was 58 GW.
According to this estimation, the gain of 0.2$\%$ in MAPE in Table~\ref{table_score_target_agg2_orange} resulting from exploiting mobility data would have amounted to tens of millions of USD per year at the national level.
In addition, we have shown that the \textit{work} index accounts for several consumption behaviours, including the impact of weekends and holidays on the electricity demand. 
Remark that these dynamics are not specific to the sobriety period, which suggests that the benefits of using mobility data would generalise to the post-crisis period. 
Overall, the higher the \textit{work} index, the higher the electricity demand. 

Future lines of research include studying the work index at a 1-hour frequency, over longer periods, and at the finer geographical scale of French administrative regions.
Indeed, as shown in Section~I.A of the Supplementary material, mobile network data effectively capture human spatial dynamics other than those related to work, such as residence and tourism. Although in this paper we have not found a significant effect of such behaviours on national electricity demand, it might become visible when working at the regional level. 
Although we have shown that a reduction in the \textit{work} index corresponds to a reduction in the electricity demand, further studies are needed to disentangle the effect of economic growth, employment rate, and remote working in this phenomenon.
Moreover, we have focused in this work on mean forecast performance, i.e., on the ability of the forecast to predict the expected value of the electricity demand.
Another interesting subject would be  to evaluate the variance in the electricity demand given the \textit{work} index, which would be helpful for practitioners when acting on the electricity market.
Finally, in practice, it currently requires several days to clean, aggregate, and adjust the indices. 
For operational use, further studies are therefore needed to quantify the impact of such a delay in the use of the \textit{work} index on the performance of benchmark forecasts, or conversely, to study the predictive capabilities of the \textit{work} index.

\setcounter{section}{0}
\renewcommand\thesection{\thechapter.\Alph{section}}

\section{Datasets and features}
\label{sec:data}
In this section, we provide further insight into our exploratory analysis of the mobility dataset. We show how the indices also capture holiday dynamics at the regional level by comparing the mobile network-based \textit{tourism} index with official tourism statistics from Insee, and by studying the temporal evolution of the \textit{work} index.

\subsection*{Regional human presence indices}
\label{sec:region}
\begin{figure*}[!t]
\centering
\subfloat[]{\includegraphics[width=0.49\textwidth]{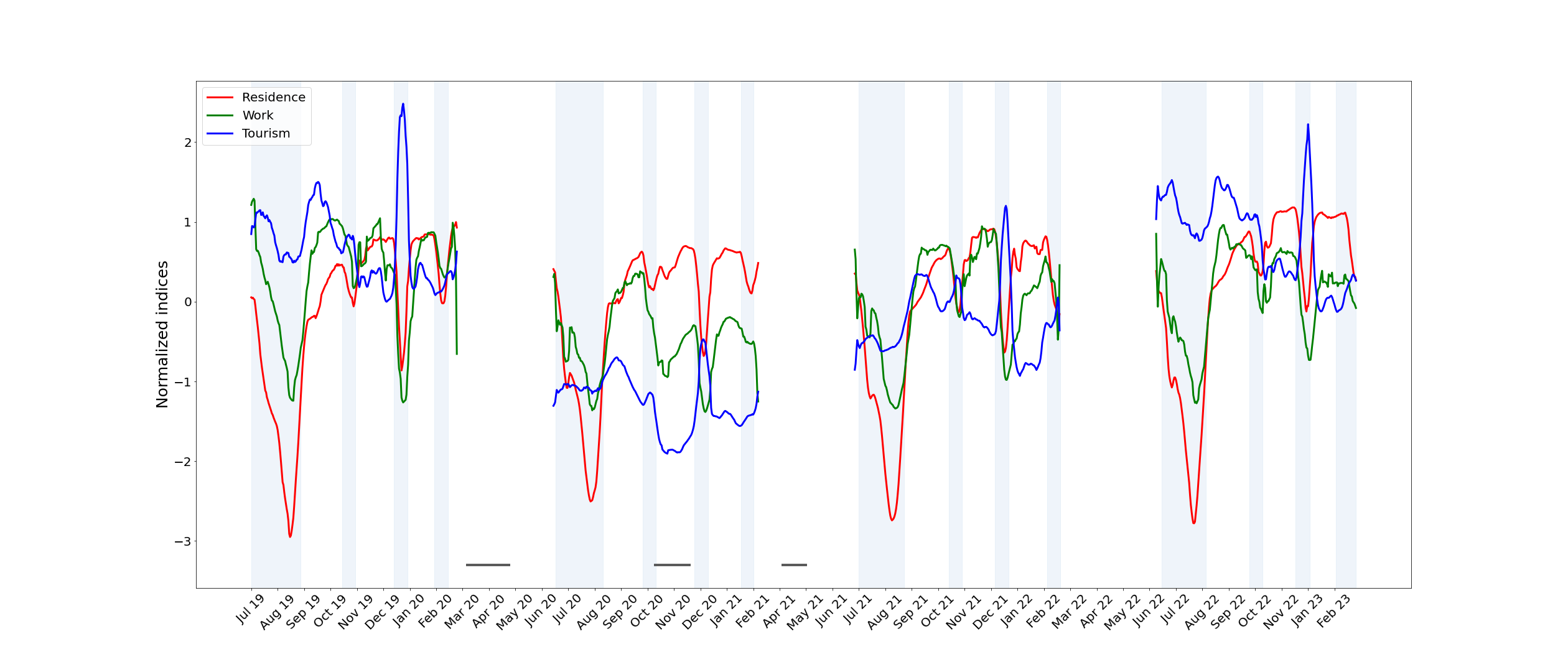}%
\label{fig:IDF_timeline_left}}
\hfil
\subfloat[]{\includegraphics[width=0.49\textwidth]{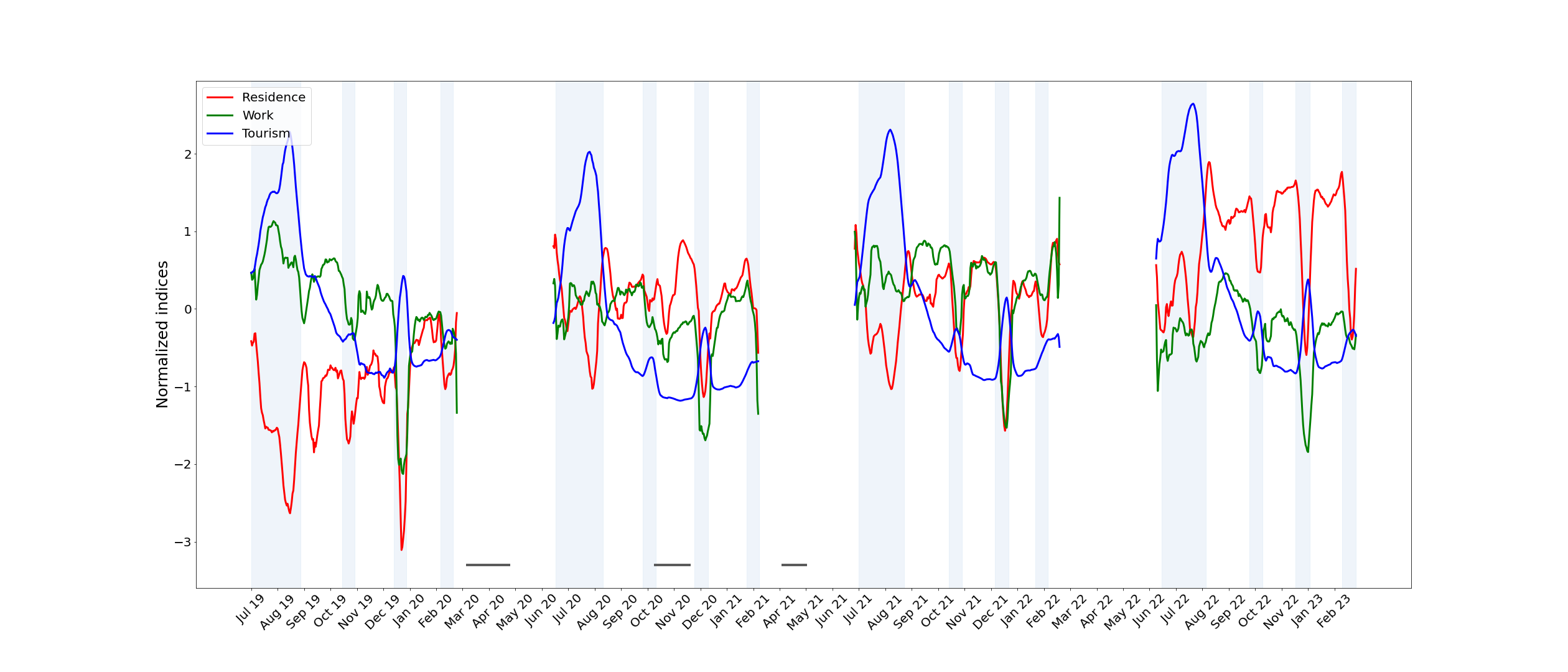}%
\label{fig:IDF_timeline_right}}
\caption{Regional indices. 7-day rolling average of mobility indices for the Île-de-France (a) and Provence-Alpes-Côte d'Azur (b) regions. Indices have been standardised, i.e., the empirical means have been subtracted and the result divided by the empirical standard deviations. The mobile network dataset only covers the period from July to March in each $12$-month period.
Shaded areas correspond to regional school holidays, and horizontal grey lines mark the three main COVID-19 lockdowns in France.}
\end{figure*}

Although for the purpose of national forecasts we have only relied on national-level indices, mobile network data were also available at the regional level, which helped to better understand the data at hand.
In order to obtain a preliminary understanding of the data, we computed Pearson's product-moment correlation coefficient $r$ between the human presence variables on the one hand, and the calendar and meteorological data on the other, for the regions of mainland France. 
This analysis confirmed that our indices matched several well-known human spatial dynamics.
In large urban regions such as Île-de-France (IDF) we observed a negative correlation between the \textit{residence} index and both of the calendar variables \textit{school holidays}  and \textit{summer holidays} ($r =-0.65$ and $r =- 0.84$, respectively), as well as \textit{temperature} ($r =-0.70$).
This captures how IDF residents leave their region during the holidays and then behave as tourists.
Consistently, in regions that are traditionally popular holiday destinations---such as the coastal region of Provence-Alpes-Côte d'Azur (PACA), the \textit{tourism index} variable was positively correlated with the calendar variables \textit{school holidays}  and \textit{summer holidays} ($r =0.58$ and $r =0.86$, respectively) and the meteorological variable \textit{temperature} ($r =0.82$). 
The \textit{work} index behaved similarly in both regions, with a negative correlation with the \textit{weekly holiday} calendar variable ($r = -0.54$ in IDF and $r = - 0.55$ in PACA). 
To better characterise the seasonal changes of the indices, we show the evolution of the daily \textit{tourism}, \textit{residence}, and \textit{work} indices in IDF (Figure \ref{fig:IDF_timeline_left}) and PACA (Figure \ref{fig:IDF_timeline_right}). In line with the Pearson correlation, in a region with a high level of economic activity such as IDF, the \textit{residence} and \textit{work} indices tended to increase during off-peak periods and to decrease during holidays. We observed the opposite for the \textit{tourism} index in PACA, which is a very tourist-oriented region. Moreover, unlike in IDF, the \textit{work} index in PACA did not decrease significantly during the summer holidays.  
This might be explained by the different make-up of the respective labour markets, with a high proportion of tourism workers in PACA.

We can also clearly see the effects of the COVID-19 health crisis. In IDF, for example, the \textit{tourism} and \textit{work} indices significantly dropped during the crisis. 
These then gradually increased during the post-COVID period, but without reaching pre-COVID levels.
This was especially pronounced for the \textit{work} index, probably because of changes in work organisation triggered by the health crisis, and also as an effect of the energy crisis. In PACA, on the other hand, we observed a lower impact on tourism, partly due to tourist origin (there are more local tourists, i.e., who do not cross regional borders) than in IDF. Of note, the \textit{residence} index seems to have gradually increased since the end of the COVID-19 crisis in PACA. This phenomenon of migration to certain regions of France has been documented by Insee in the report \cite{michailesco2023migrations}, but deserves a more in-depth analysis.

\subsection*{The tourism index from mobile-phone data}
\label{subsec:tourism}
\begin{figure}
    \centering
    \includegraphics[width = 0.49\textwidth]{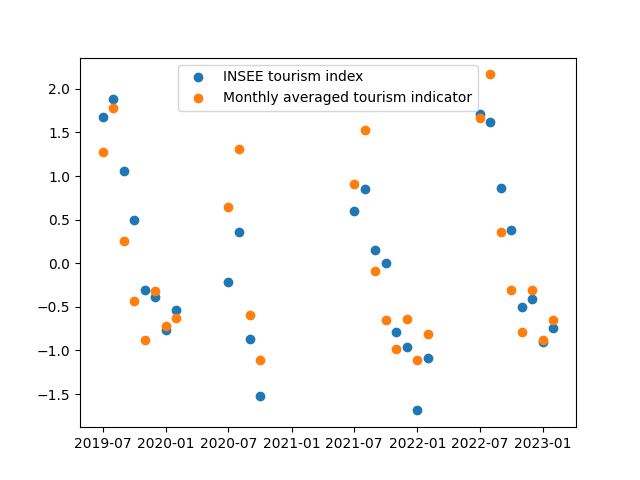}
    \caption{ Comparison of the Insee and the mobile network tourism indices.}
    \label{fig:tourism}
\end{figure}

The evaluated number of tourists and residents has been shown to be correlated with electricity demand in highly touristic areas \cite{Bakhat2011estimation, lai2011the}. 
For this reason, we created and studied a \textit{tourism} index at the national level.
Traditionally, most similar assessments have been carried out on a monthly or annual basis. 
One strength of our mobile phone-based \textit{tourism} index is that it can be calculated at finer temporal and geographical scales. 
To further assess the \textit{tourism} index's performance as a proxy for tourism activity, we compared its monthly
average with the Insee tourism index \cite{INSEEData}, as shown in Figure \ref{fig:tourism}.
We obtained an 87\% correlation between the two signals, showing that the tourism index efficiently captures tourism trends. However, our study found that tourism had no significant impact on French electricity demand (see Section \ref{sec:variable_selection}).

\subsection*{\textit{Work} index and calendar features} \label{subsec:temp_features}
\begin{figure}
    \centering
    \includegraphics[width = 0.49\textwidth]{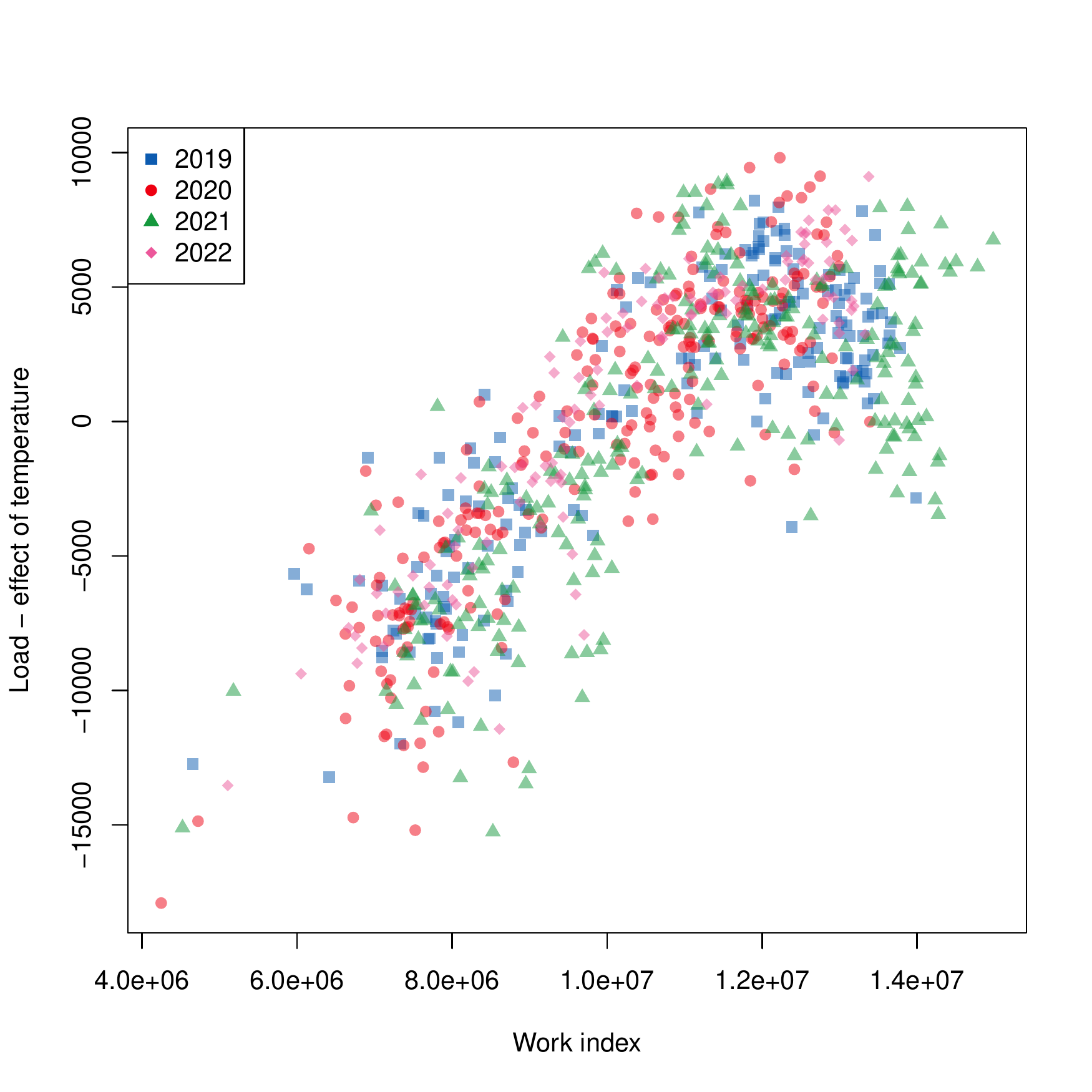}
    \caption{Residuals as a function of the \textit{work} index over the years 2019--2022. Each point is an observation between July 2019 and March 2022.}
    \label{fig:year}
\end{figure}

\begin{figure*}[!t]
\centering
\subfloat[]{\includegraphics[width=0.49\textwidth]{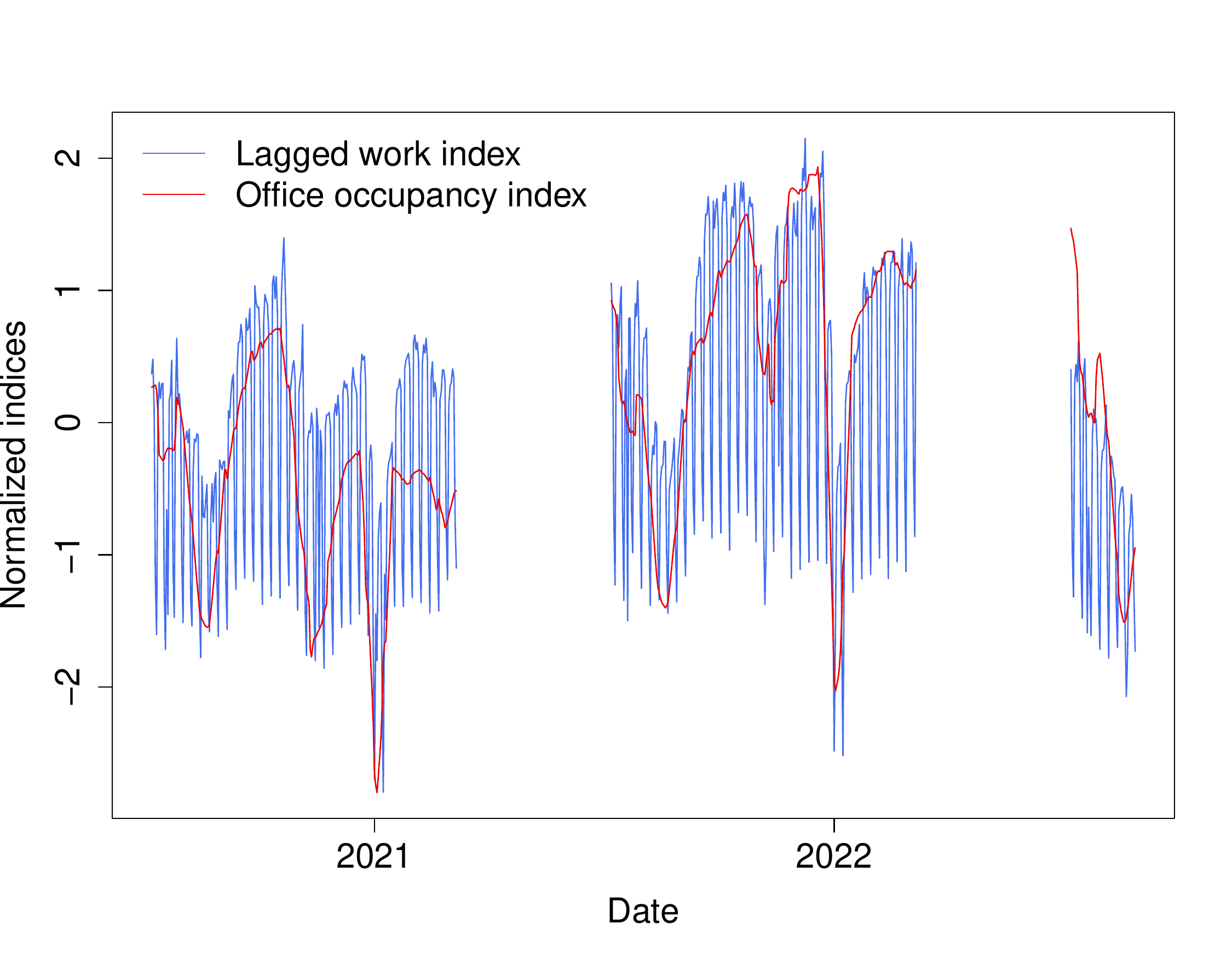}%
\label{fig:normalcy2_left}}
\hfil
\subfloat[]{\includegraphics[width=0.49\textwidth]{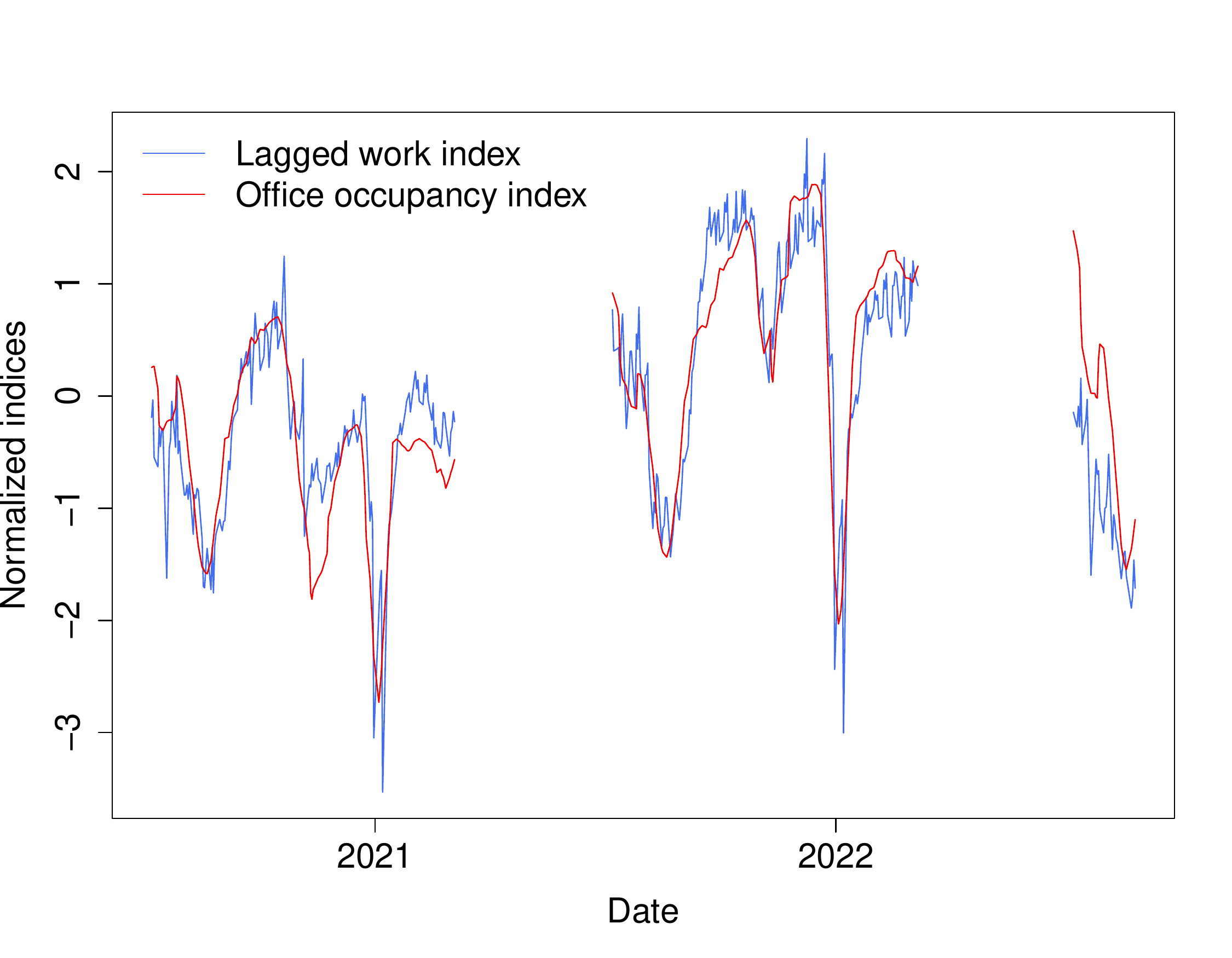}%
\label{fig:normalcy2_right}}
\caption{Comparison of work indices. Comparison of the 7-day lagged mobile phone based index and the normalcy office occupancy index on all days (a), and when excluding weekends and holidays (b).}
\end{figure*}

\begin{figure*}[!t]
\centering
\subfloat[]{\includegraphics[width=0.49\textwidth]{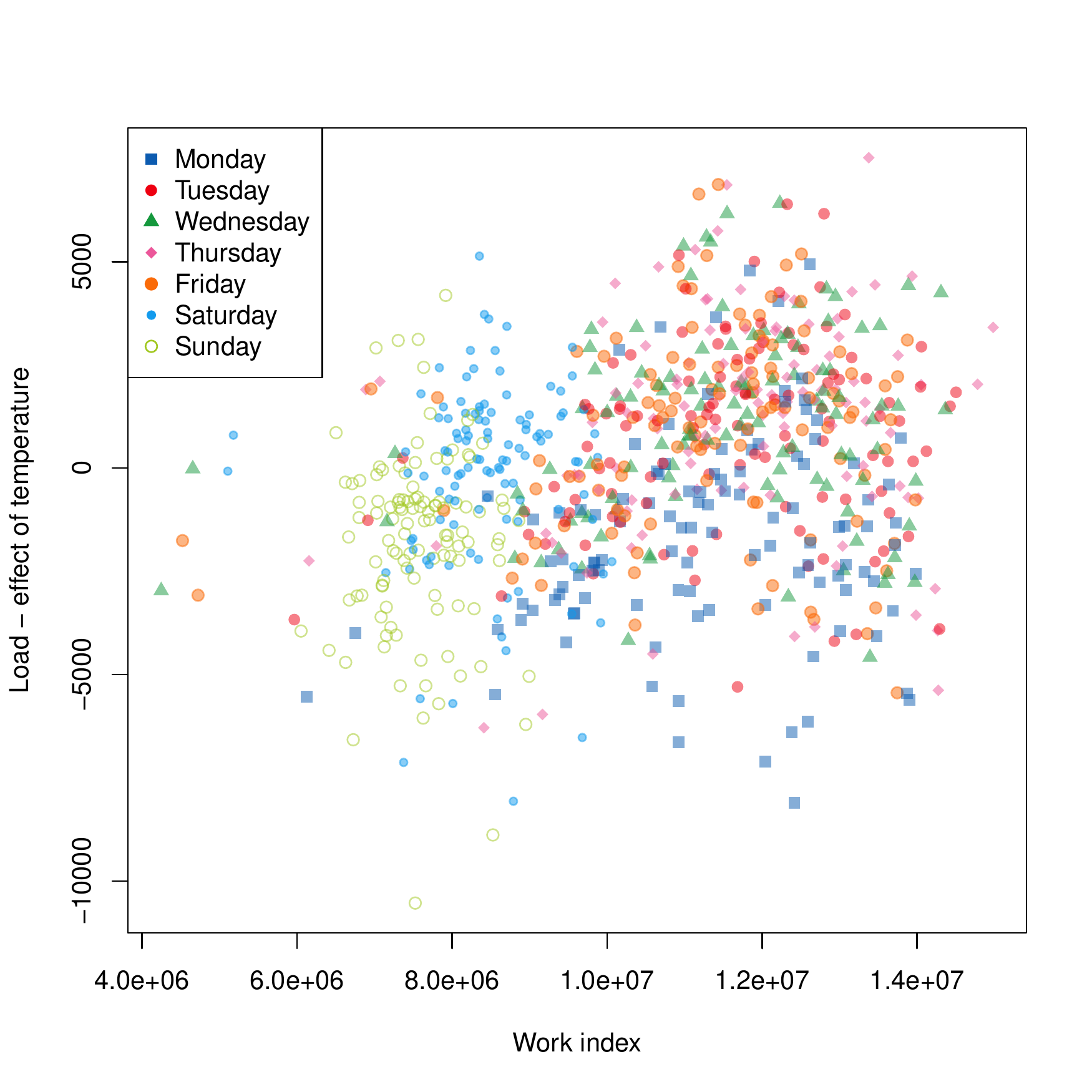}%
\label{fig:hours_comparison_left}}
\hfil
\subfloat[]{\includegraphics[width=0.49\textwidth]{figures/paper_5/day_dependency.pdf}%
\label{fig:hours_comparison_right}}
\caption{Electricity demand corrected for temperature as a function of the \textit{work} index for each day of the week. Each point is an observation between July 2019 and March 2022. (a) 2 a.m. (b) 10 a.m.}
\end{figure*}

As explained in the introduction, several phenomena occurred between 2020 and 2023 that significantly changed human behaviour and affected French electricity demand. 
To better understand the impact of the \textit{work} index on electricity demand, it was therefore important to see whether this dependence has changed over time.
Figure~\ref{fig:year} shows that the dependence of electricity demand in the \textit{work} index has been stationary over the years 2019--2022.
This shows that this relationship is robust to the aforementioned events, which is an argument to believe that the results of this article will generalise well to future periods of interest.
In addition, as shown in  Figure~\ref{fig:normalcy2_left} and \ref{fig:normalcy2_right}, unlike The Economist's office occupancy index, our \textit{work} index from mobile data captures the reduction in work activity due to weekends and holidays. 
Furthermore, as expected, Figures~\ref{fig:hours_comparison_left} and \ref{fig:hours_comparison_right} show that the \textit{work} index is only useful for electricity demand forecasting during typical work hours.
Indeed, the electricity demand corrected for the temperature effect had a significant dependence in the index at 10 a.m., but not at 2 a.m.
See Section \ref{sec:remote_wkg} for more details.

\section{Benchmark and models}
\label{app:Benchmark}
In this section, we detail the framework and the models in Table I of the main document.
\subsection*{Handling missing values in mobile network data}
\label{app:missing}
There are two types of missing data in our datasets. First, the datasets are regularly sampled time series but with different frequencies. Indeed, recall that the calendar and the electricity datasets have 30-minute frequencies, while the meteorological dataset has a 3-hour one, while the mobile phone dataset has a 1-day frequency.
A common method to deal with differences in sampling frequency is to impute the missing value by interpolation \cite{emmanuel2021a}.
The interpolation method for meteorological data is described in the Methods section of the main paper, while the Orange indices are considered constant throughout the day.

Second, the mobile network dataset only covers the periods ranging from 01/07/2019 to 01/03/2020,  from 01/07/2020 to 01/03/2021, from 01/07/2021 to 01/03/2022, and from 01/07/2022 to 01/03/2023. Though various techniques have been developed to tackle sampling irregularities in time series \cite{shukla2021survey}, dealing with large sets of consecutive missing values is still very challenging.
The three main approaches when studying time series with consecutive missing values are deletion, imputation, and imputation with masking \cite{emmanuel2021a}.
First, deletion consists of discarding any observation with at least one missing value.
Though this is the simplest way to deal with missing values, it can  introduce a bias if the missing data are not-at-random, i.e., if the missing data are actually informative with respect to the target \cite{little2019statistical}. 
Second, in regression, imputation techniques aim to ``fill in'' missing values.
The state-of-the-art in time series imputation is wide-ranging and an active field of research \cite{ma2020transfer}.
Note that imputation that maximises a regression model's performance is not necessarily that which reconstructs missing values most accurately  \cite{zhang2021missing, ayme2023naive}.
This makes it more difficult to understand and explain the true effect of imputed features on a target variable.
Third, imputation with masking consists of imputing the missing values and keeping track of which observations have been imputed by adding a new feature equal to 1 if the observation comes from an actual measurement, and 0 if it was imputed.
In this paper, the pattern of missing data is regular, spanning each year from March to July, and does not depend on the  explanatory variables (temperature, \textit{work} index, etc.).
Thus, to simplify the analysis, we chose the deletion framework and have not tried to impute the missing values of the mobile network indices.

Furthermore, Table~I in the main text not only shows that mobile phone indices help to improve the performance of state-of-the-art forecasting algorithms, but also attests that this is still true even when comparing the complete open dataset with the incomplete mobile phone dataset.
Indeed, on the one hand, models ``without mobility data'' were trained on the complete calendar, weather, and electricity datasets, spanning from 08/01/2013 to 01/09/2022.
On the other hand, models ``with mobility data'' were created in a two-step process using the transfer learning framework presented in \cite{antoniadis2021hierarchical}.
First, a model trained without mobility data from 08/01/2013 to 01/09/2022 provided an estimate $\hat{Load}$ of the electricity demand $Load$.
Then, another model was trained in the deletion framework to forecast the error $err = Load - \hat{Load}$, also known as the residual, using the mobile phone dataset.
This second forecast is denoted by $\hat{err}$.
The final forecast was therefore the sum of the two forecasts $\hat{Load} + \hat{err}$.
Notice that this framework gives an advantage to the reference forecast ``without mobility data''.
In fact, the gains from using mobile phone data are much higher if the training periods of all models are restricted to the period for which mobile phone data are available (although we have not included these results in the paper for the sake of simplicity).
However, the framework we chose allowed us to assess the interest of using mobile phone data from an operational point of view.
It ensured that the best models trained using the mobile phone dataset outperformed the best models trained on the full, open datasets. 
Therefore, the gains of $10 \%$ we obtained are likely to be much higher if we had access to a more complete mobile phone dataset.
We chose this residuals method to account for the mobile phone data because it gave better results than directly training models ``with mobility data'' on all datasets restricted  to the period for which the mobile phone data ere available (once again, we have not included these results in the paper for simplicity).

\subsection*{Statistical models}

\paragraph{Time series models}
Persistence models are the simplest type of model for time series. They consist of estimating the target with its own lags and are common baselines in time series benchmarks because of their simplicity, ability to capture trends, explainability, and robustness to sudden changes in data distributions. In Table~I of the main article, the persistence estimator corresponds to a 24-hour lag in electricity demand.

Seasonal autoregressive integrated moving average (SARIMA) models \cite{box2015time} are also commonly used in time series analysis. Here, we trained one model for each of the 48 half-hours in a day to capture daily seasonality in the data. 
Each model was then fitted with weekly seasonality by running the \texttt{auto.arima} method of the \texttt{forecast} package in \texttt{R}.

\paragraph{Generalized additive models}
\label{subs:variables}
Generalized additive models (GAMs) are a generalisation of linear regression. 
Instead of learning linear coefficients linking certain features $\boldsymbol{x} = (x_{1}, \hdots,x_{d})$ to a target $y$, a GAM learns the nodes and coefficients of the regression of the features onto the target with respect to a spline basis. 
More precisely, given a target time series $y = (y_t)_{t \in T}$ indexed by $T$, and some explanatory variables $\boldsymbol{x} = (x_{t,1},\hdots,x_{t,d})_{t \in T}$, the response variable $y$ is written in the form:
\begin{equation*}
    y_t = \beta_0 + \sum_{j=1}^d f_j(x_{t,j}) + \varepsilon_t \,,
\end{equation*}
where $\varepsilon = (\varepsilon_t)_{t\in T}$ are independent and identically distributed (i.i.d.) random noise.
Though the target $y_t$ at time $t$ is in fact a real number, each potentially explanatory time series $x_{k} = (x_{t,k})_{t \in T}$ has a dimension $d_k \geq 1$; i.e., at time $t$, $x_{t,k} \in \mathbb{R}^{d_k}$. 
Therefore, nonlinear effects of multiple variables are allowed, such as, for instance, $y_t = \beta_0 + f_1(x_{t, 1}) + \varepsilon_t$ with $x_{t,1} \in \mathbb{R}^2$.
The goal of GAM optimisation is to find the best nonlinear functions $f_1, \hdots, f_d$ to fit $y$.
Thus, each nonlinear effect $f_j$ is decomposed over a spline basis $(B_{j,k})_{1\leq j \leq d,\; k \in \mathbb{N}}$, with coefficients $\boldsymbol\beta_j$, such that
\begin{equation*}
    f_j(x) = \sum\limits_{k=1}^{m_j} \beta_{j,k} B_{j,k}(x) \,,
\end{equation*}
where $m_j$ corresponds to the dimension of the spline basis. 
The offset $\beta_0$ is chosen so that the functions $f_j$ are centred. The coefficients $\beta_0, \boldsymbol{\beta}_1, \ldots, \boldsymbol{\beta}_{d}$ are then obtained by penalised least squares. The penalty term involves the second derivatives of the functions $f_j$, forcing the effects to be smooth (see \cite{wood2017generalized}).

The GAM model used in the experiments presented in Table   \ref{table_score_target_agg}  was taken from \cite{obst2021adaptative}.  
As is usual in load forecasting with GAMs, we considered one model per half-hour of the day, with the 48 half-hour time series treated independently.
Therefore, 48 models were fitted, one for each half-hour of each day.
Given a half-hour $h$, our model was therefore:
\begin{align}\label{eq:modele_GAM_FR}
    \text{Load}_{h,t} =\ & \sum_{i=1}^7 \sum_{j=0}^1 \alpha_{h,i,j}\;\boldsymbol{1}_{\text{DayType}_t=i}\;\boldsymbol{1}_{\text{DLS}_t=j}  + \sum_{i=1}^7 \beta_{h,i}\; \text{Load1D}_t \;\boldsymbol{1}_{\text{DayType}_t=i}   \\
    & + \gamma\; \text{Load1W}_t + f_{h,1}(t) + f_{h,2}(\text{ToY}_t) + f_{h,3}(t,\; \text{Temp}_{h,t}) + f_{h,4}(\text{Temp95}_{h,t}) 
       \nonumber\\
   & + f_{h,5}(\text{Temp99}_{h,t}) +f_{h,6}(\text{TempMin99}_{h,t},\; \text{TempMax99}_{h,t})  + \varepsilon_{h,t} \nonumber\,,
\end{align}
where the timestamp $t$ corresponds to the day, and: 
\begin{itemize}
    \item
    $\text{Load}_{h,t}$ is the electricity load on day $t$ at time $h$.
    \item
    $\text{DayType}_t$ is a categorical variable indicating the type of day of the week.
    \item
    $\text{DLS}_t$ is a binary variable indicating whether $t$  is daylight saving time or standard time.
    \item
    $\text{Load1D}$ and $\text{Load1W}$ are the loads of the previous day and previous week, respectively.
    \item
    $\text{ToY}_t$ is the time of year, growing linearly from 0 at midnight when January 1 begins, to 1 on  December 31 at 23:30 pm.
    \item
    $\text{Temp}_{h,t}$ is the national average temperature at time $h$ on day $t$.
    \item
    $\text{Temp95}_{h,t}$ and $\text{Temp99}_{h,t}$ are exponentially smoothed temperatures with respective factors $\alpha=0.95$ and $0.99$. For example,  $\alpha=0.95$ corresponds to
    \[\text{Temp95}_{h,t}=\alpha \text{Temp95}_{h-1,t} + (1-\alpha) \text{Temp}_{h,t}.\]
    \item
    $\text{TempMin99}_{h,t}$ and $\text{TempMax99}_{h,t}$ are respectively the minimal and maximal values of $\text{Temp99}$ on day $t$ over all time instants $i$ such that $i \leq h$.
\end{itemize}
We ran these models  in \texttt{R} using the \texttt{mgcv} library \cite{wood2015package}. 
We used the default thin-plate spline basis to represent the $f_j$'s, except for the time of year effect $f_2$ for which we chose cyclic cubic splines (see \cite{wood2017generalized} for a full description of the spline basis). 
To replicate the GAM of \cite{vilmarest2022state}, the dimensions of the bases were taken equal to $5$, except for $f_2$ which had a basis of dimension $20$. 

\subsection*{Data assimilation techniques}

\paragraph{State space models}
State space models are efficient in capturing time-varying structures (as opposed to seasonality) in time series \cite{hyndman2008forecasting}.
In particular, the Kalman filter is a powerful mathematical and algorithmic tool introduced by \cite{Kalman1960} for state space model estimation. 
In electricity load forecasting, Kalman filters have been used to update the output of a GAM using recent observations of electricity demand \cite{vilmarest2022state}. 

Following the notation of \eqref{eq:modele_GAM_FR}, let $f(\boldsymbol{x}_t) = (1,f_1(x_{t,1}),\hdots,f_d(x_{t,d}))^\top$. 
Our goal is to estimate a time-varying vector $\boldsymbol\theta_t \in \mathbb{R}^{d+1}$ such that $\mathbb{E}[y_t\mid \boldsymbol{x}_t] = \boldsymbol{\theta}_t^\top f(\boldsymbol{x}_t)$. 
This corresponds to adjusting the relative importance of each nonlinear effect, while preserving their shapes. 
This is achieved by considering the state space model
\begin{align*}
    & \theta_t - \theta_{t-1} \;\sim \mathcal{N}(0,\;Q_t), \\
	& y_t - \theta_t^\top x_t \sim \mathcal{N}(0,\;\sigma_t^2),
\end{align*}
where $\mathcal{N}(\mu, \sigma^2)$ is the multidimensional normal distribution with mean $\mu$ and variance matrix $\sigma^2$, $\theta_t$ is the latent state, $Q_t$ the process noise covariance matrix, and $\sigma_t^2$  the observation variance. Applying the recursive Kalman filter equations as described in section A of \cite{vilmarest2022state} provides us with both $ \theta_t$ and the conditional expectation $\mathbb{E}[y_t\mid \boldsymbol{x}_t]$, which is known to be the best forecast, i.e., minimizes the mean square error conditional on past observations and exogenous covariates $x_t$. 
As in \cite{vilmarest2022state}, we ran the three variants \textit{static}, \textit{dynamic}, and \textit{Viking} of the Kalman filter. The \textit{static} version is a degenerate case where $Q_t$ is null, which leads to low adaptation. The \textit{dynamic} variant supposes that $Q_t=Q$ and $\sigma_t=\sigma$ are constants  obtained by grid search optimisation on past observations. Finally, the \textit{Viking} version assumes that $Q_t$ and $\sigma_t$ are updated online (see \cite{vilmarest2022state} for more details).
In Table~I of the main article, the GAM model used in the state space models is that from \cite{obst2021adaptative}, while the
\textit{static} Kalman filter, \textit{dynamic} Kalman filter, and \textit{Viking} method are from  \cite{Vilamarest2024adaptive}.

\paragraph{Online aggregation of experts}
Online robust aggregation of experts \cite{cesa2006prediction} is a model agnostic technique for time series forecasting. This approach combines various forecasts (called experts) based on their past performance, in a streaming manner. 
This method allows for adaptation to changes in distributions by tracking the best experts.
Sequential expert aggregation assumes that the data are observed sequentially.
The target variable $Y$ (here electricity demand) is supposed a bounded sequence, i.e., $Y_1,\dots,Y_T \in [0,B]$, where $B>0$. 
Our goal is to forecast this variable step by step for each given time $t$. At each time $t$, $N$ experts offer forecasts of $Y_t$, denoted by $\left(\hat{Y}_{t}^{1},\dots,\hat{Y}_{t}^N\right) \in [0,B]^N$. 
These experts can be the result of any process, such as a statistical model, a physical model, or human-based expertise.
Then, the aggregation algorithm generates a forecast of $Y_t$ by the weighted average of the $N$ forecasts: 
\[\hat Y_t = \sum_{j=1}^N \hat{p}_{j, t} \;\hat{Y}_{t}^j,\] where the weight $\hat{p}_{j, t} \in \mathbb{R}$ depends on the performance of $\hat Y_t^j$ over the period $\{1, \hdots, t-1\}$. 
Then, $Y_t$ is observed and the next instance starts.

In our study, we ran the ML-Poly algorithm, first proposed by \cite{gaillard2014second} and subsequently implemented in \texttt{R} in the \texttt{opera}  package \cite{gaillard2016opera}. This algorithm identifies the best expert aggregation by giving more weight to  experts producing the lowest forecasting error, making it noteworthy due to the absence of parameter tuning required. 
In Table~I of the main article, all of the estimators related to data assimilation techniques are combined, i.e., the GAM, the static Kalman filter, the dynamic Kalman filter, and the Viking estimator. 

\subsection*{Machine learning}

\paragraph{Random forests}

Among the most robust machine learning techniques are random forests \cite{breiman2001random}. 
They consist of averaging a given number of decision trees generated by applying classification and regression trees \cite{breiman1984cart} to different subsets of the data obtained by bagging and random sampling of covariates.
Each decision tree estimates the target by a series of logical comparisons on the feature variables. 
An example of a decision tree of depth 3 is ``if \textit{temperature}  $>$ 30°C, if it is 10 a.m., and if it is a Wednesday, then \textit{electricity demand} = 6 GW''. 
Random forests require very little prior knowledge about a problem, which makes them good for benchmarking in applied machine learning problems.
In Table~I of the main article, the random forests all had 1000 trees of depth 6 (the square root of the number of features).
Random forests are usually trained on random subsets of the training sample.
To take advantage of the dependence of samples in time series, the random subsets can be drawn from a given number of consecutive measures. 
This is what occurs in the \textit{random forest + bootstrap} architecture \cite{gohery2023random}.

\paragraph{Gradient boosting}

Gradient boosting \cite{breiman1997arcing,friedman2001greedy} consists of successively fitting the errors of simple models---called weak learners---and then aggregating them. 
This is an ensemble technique, like random forests. 
Gradient boosting usually outperforms random forests \cite{grinsztajn2022tree}, at the cost of more parameters to calibrate. 
It has previously shown excellent performance on regression problems \cite{grinsztajn2022tree} and in forecasting challenges \cite{makridakis2022m5}.
In tree-based gradient boosting algorithms, weak learners are decision trees, whereas in GAM boosting algorithms \cite{buhlmann2007boosting}, weak learners are spline regression models.

\subsection*{Models with mobile phone data}
As explained in Section \ref{app:missing}, the forecasts trained on the dataset ``with mobility data'' actually consisted of two models. 
The first model was trained on the entire dataset ``without mobility data''. 
The second model estimated the error of the first model on the dataset ``with mobility data''. 
The GAM ``with mobility data'' is the sum of the GAM ``without mobility data'' and of the following GAM:
\begin{align*}
    err_{h,t} =\ & \sum_{i=1}^7 \sum_{j=0}^1 \tilde \alpha_{h,i,j}\;\boldsymbol{1}_{\text{DayType}_t=i}  + f_{h,7}(\text{ToY}_t) +f_{h,8}(\text{Work}_t) +f_{h,9}(\text{Residence}_t)  + \varepsilon_{h,t}, 
\end{align*}
with these abbreviations  defined in Section~\ref{subs:variables}.
The static Kalman filter, dynamic Kalman filter, and Viking estimators ``with mobility data'' were then computed by summing the effects of the two GAMs.
The GAM boosting ``with mobility data'' was the sum of the boosted GAM ``without mobility data'' and of a boosted GAM with all variables (calendar, meteorological, electricity, and mobile phone).
The random forest ``with mobility data'' was the sum of the \textit{random forest + bootstrap} model ``without mobility data'' and of a random forest with all variables.
The  \textit{random forest + bootstrap} ``with mobility data'' was the sum of the \textit{random forest + bootstrap} model ``without mobility data'' and of a  \textit{random forest + bootstrap} with all variables.

\subsection*{Excluding holidays}
As mentioned in Section~II.B of the main paper, holidays are known to behave differently from regular days \cite{Krstonijevic2022adaptive}. 
Therefore, we ran the same benchmark here, but excluding holidays, as well as the days directly before and after holidays, from both training and testing.
Table~\ref{table_score_target_agg} shows that, when excluding holidays, incorporating mobility data improved the best performance (aggregation of experts) by 8\% in RMSE and 6\% in MAPE.
Once again, the global performance improvement across all models was around 10$\%$.
Note that these gains are significant, because they leave the confidence interval obtained by bootstrapping (see Methods).
\begin{table*}[!t]
\centering
\caption{Benchmark excluding holidays. The numerical performance is measured in  RMSE (GW) and MAPE (\%).} 
\begin{tabular}{|l|c|c|}
  \hline
 &  Without mobility   & With mobility   \\
 \hline 
  \textit{Model} &&\\
  Persistence (1 day) & 4.0$\pm$0.2 GW,  5.0$\pm$0.3 \%  & N.A., N.A.\\
    SARIMA  & 2.0$\pm$0.2  GW, 2.6$\pm$0.2 \%   & N.A., N.A. \\
  GAM & 1.70$\pm$0.06 GW, 2.6$\pm$0.1 \% & \textbf{1.55}$\pm$0.05 GW, \textbf{2.43}$\pm$0.08 \%\\
  \hline
  \textit{Data assimilation } &&\\
  Static Kalman  & 1.43$\pm$0.05 GW, 2.20$\pm$0.08 \% & 1.07$\pm$0.04 GW, 1.63$\pm$0.06 \% \\
  Dynamic Kalman  & 1.10$\pm$0.04 GW, 1.58$\pm$0.05  \% & 0.96$\pm$0.03 GW, 1.39$\pm$0.04 \%\\
    Viking & 0.98$\pm$0.04 GW, 1.33$\pm$0.04 \% & 0.98$\pm$0.03 GW, 1.41$\pm$0.05 \%\\
    Aggregation  & 0.96$\pm$0.04 GW, 1.36$\pm$0.04 \% & \textbf{0.88}$\pm$0.03  GW, \textbf{1.28}$\pm$0.04 \%\\
    \hline
    \textit{Machine learning} &&\\
    GAM boosting & 2.3$\pm$0.1 GW, 3.3$\pm$0.2 \%& 2.2$\pm$0.1 GW, 3.1$\pm$0.2 \%\\
    Random forests & 2.1$\pm$0.1 GW, 3.0$\pm$0.1 \% & \textbf{1.8}$\pm$0.1 GW, \textbf{2.4}$\pm$0.1 \%\\
    Random forests + bootstrap & 1.9$\pm$0.1 GW, 2.6$\pm$0.1 \%&\textbf{1.8}$\pm$0.1 GW, \textbf{2.4}$\pm$0.1 \%\\
   \hline
\end{tabular}
\label{table_score_target_agg}
\end{table*}

\section{Change point detection}
In this section, we detail and justify the use of the model used in Section~III.A of the main article to assess energy savings, as well as the change point detection algorithm subsequently applied to it.
\subsection*{The seasonality model}
The model used in Section~III.A of the main article to capture the dependence of calendar and meteorological data on electricity demand is the following direct adaptation of the GAM from \citet{obst2021adaptative}:
\begin{align*}
    \mathrm{Load}_{h,t} =\ & \sum_{i=1}^7 \sum_{j=0}^1 \alpha_{h,i,j}\;\boldsymbol{1}_{\text{DayType}_t=i}\;\boldsymbol{1}_{\text{DLS}_t=j} + f_{h,1}(\text{ToY}_t) + f_{h,2}(\text{Temp95}_{h,t}) \\
    &+ f_{h,3}(\text{Temp99}_{h,t})  + f_{h,4}(\text{TempMin99}_{h,t},\; \text{TempMax99}_{h,t})  + \varepsilon_{h,t}. 
\end{align*}
We note that this corresponds to removing the dependence on the timestamp $t$ and on the lags Load1D and Load1W from equation \eqref{eq:modele_GAM_FR}.
On the one hand, these features were removed because they only captured the trend of the signal without explaining the phenomena at stake, which interfered with the interpretability of the model.
On the other hand, the remaining features account for well-known repeated phenomena such as the effects of weekends  (in DayType), holidays (in ToY), and heating and cooling (in the smoothed temperatures),
thus helping to explain seasonality in the signal.
This GAM was trained on data from 01/01/2014 to 01/01/2018. 
The residuals $\text{res} = \text{Load} - \hat{\text{Load}}$ were then evaluated from 01/01/2018 to 01/03/2023.
Between 01/01/2018 and 01/01/2020, this GAM had an average MAPE of $2.1\%$ and an average RMSE of $1.6$ GW.
This is comparable to the performance of the GAM in \cite{obst2021adaptative}, which had an average MAPE of $1.6\%$ and an average RMSE of $1.2$ GW.
At the cost of slightly lower performance, our GAM is more interpretable because it only takes seasonal phenomena into account.
We therefore consider it to be a good model for forecasting what the electricity demand should be over a multi-year time horizon, assuming that electricity consumption behaviour remains unchanged. 

\subsection*{Descriptive analysis of residuals}
\begin{figure*}[!t]
\centering
\subfloat[]{\includegraphics[width=0.49\textwidth]{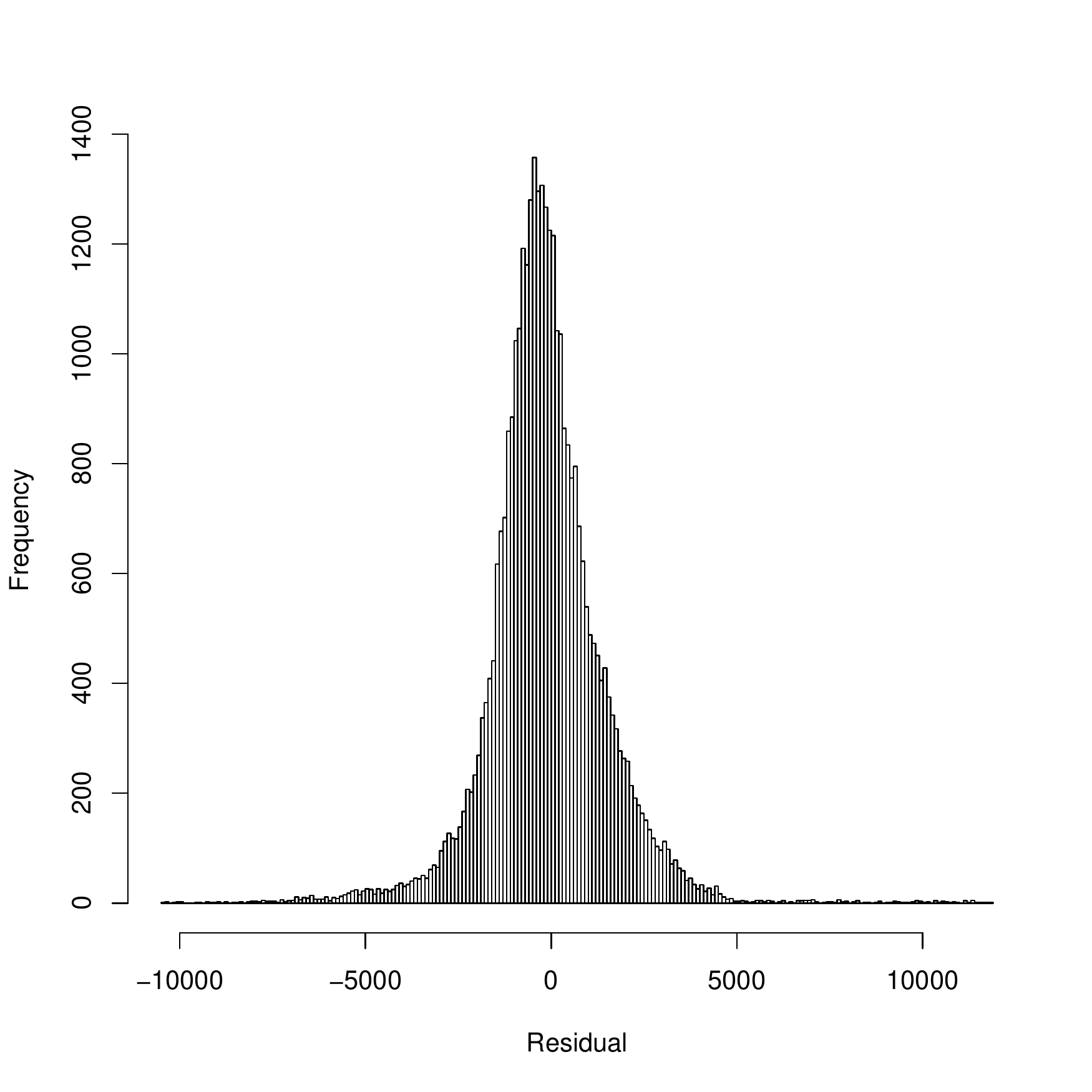}%
\label{fig:description_residuals_left}}
\hfil
\subfloat[]{\includegraphics[width=0.49\textwidth]{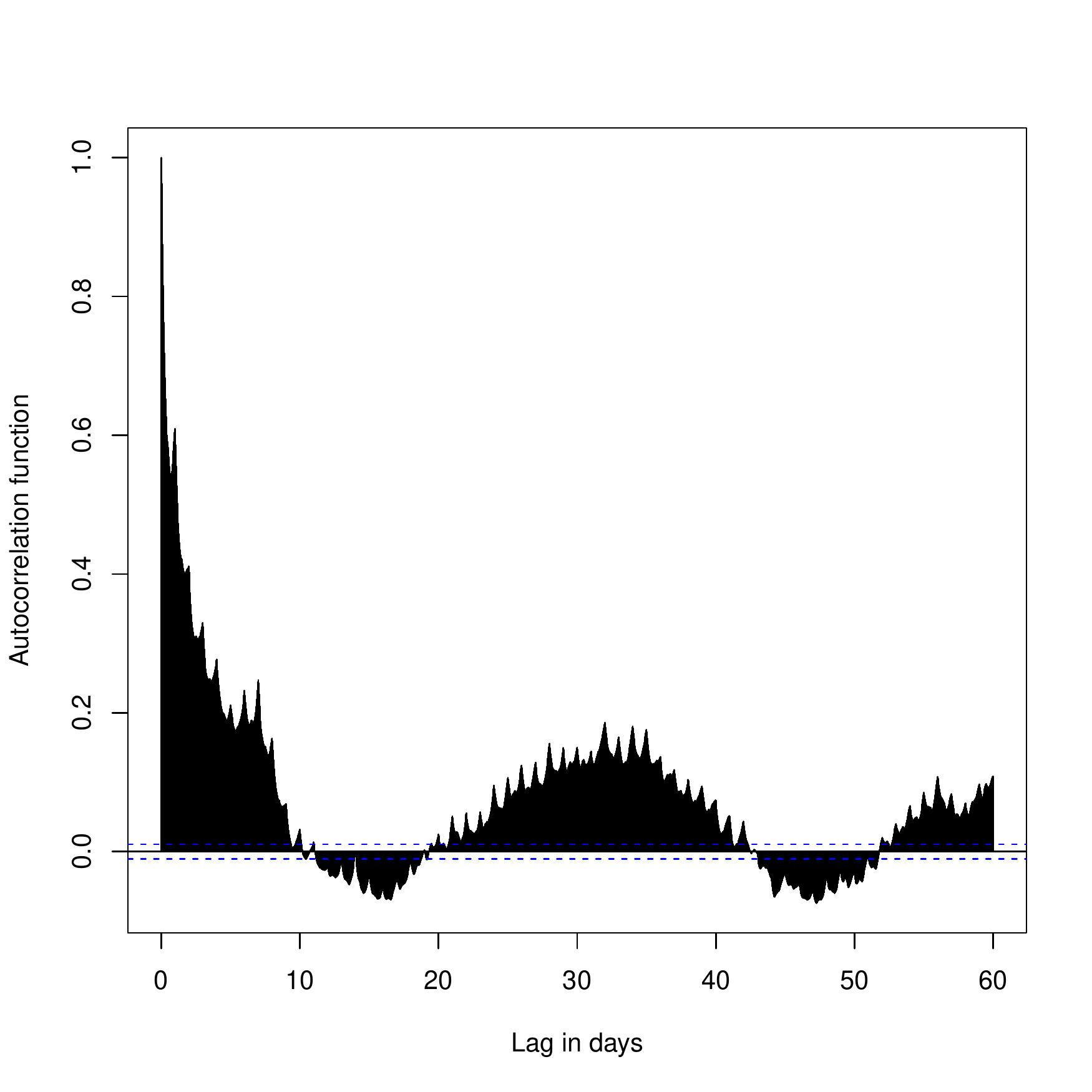}%
\label{fig:description_residuals_right}}
\caption{ Descriptive statistics of the residuals. (a) Histogram of the residuals between 01/01/2018 and 01/01/2020. (b) Auto\-cor\-re\-lation function of the residuals between 01/01/2018 and 01/01/2020. The dotted lines correspond to a confidence interval pertaining to the precision of the auto-correlation estimators.}
\end{figure*}

In this paragraph, we focus on the period spanning from 01/01/2018 to 01/01/2020.
As shown in Figure \ref{fig:description_residuals_left}, the residuals histogram is bell-shaped.
Since we had $2 \times 365 \times 48 = 35040$ observations, we chose the number of breaks in the histogram to be $\lfloor \sqrt{35040} \rfloor = 187$, where  $\lfloor \cdot \rfloor$ is the floor function.
Student's $t$-test showed that the expectation of the residuals was significantly lower than zero ($p < 2.2 \times 10^{-16}$) and was contained in the interval $[-0.16\;  \text{GW},\; -0.12\; \text{GW}]$ with a probability of 95\%.
The empirical mean was $-0.14$ GW, while the empirical standard deviation was $1.6$ GW.
An Anderson-Darling normality test suggested that the residuals did not follow a normal distribution ($p < 2.2 \times 10^{-16}$).
Moreover, as shown in Figure \ref{fig:description_residuals_right}, the autocorrelations of the residuals decreased slowly  and were significantly greater than zero, suggesting that the residuals were not stationary.
Further evidence for this came from a Box-Ljung test with a 1--day window ($p < 2.2 \times 10^{-16}$).

Both the fact that the expectation of the residuals was significantly less than zero and the residuals were not stationary indicated that other phenomena than calendar seasonality and temperature are involved, though their impact appears to be moderate since the estimator performs well.
This suggests that, even in this period without major events or decisions such as COVID-19 or sobriety, other features can be useful for better understanding electricity demand.

\subsection*{Ranking changes in the data distribution}
The descriptive analysis shows that the residuals are not stationary. 
Therefore, from a statistical point of view, it is pointless to look for the change points observed in Figure~2 of the main article in absolute terms.
In fact, the more precise the technique for detecting change points becomes, the more change points will be detected everywhere.
This is why we need quantitative information about the importance of the change points in order to rank them and determine which are the most significant ones.
A number of metrics have been developed to measure  the importance of change points \cite{Aminikhanghahi2017a}.
To assess the significance of the number of change points here, we sequentially compared the standard deviation of the residuals with the amplitude of the change points.
This resulted in 10 change points being considered in the following analysis.
The principle behind offline change-in-mean techniques is to segment the signal in such a way that approximating the signal by its mean in each segment results in the lowest possible variance.
However, finding such an optimum would be computationally expensive for our time series of around 70 000 observations.
Therefore, we rely on faster algorithms which have been developed to find approximations to the optimal change points, such as binary segmentation, as was used in Figure~2 of the main article.

\section{Statistical analysis}
In this section, we provide furthers analyses related to the variable selection detailed in the Results section. 
to further justify the study of the \textit{work} index in the statistical analysis of Section~III.B of the main article.

\subsection*{Variable selection: Hoeffding D-stastics and Shapley values}
\label{sec:variable_selection}
This paragraph complements the mRMR variable ranking performed in the Results section. 
To examine the variable selection process more closely, we computed the Hoeffding D-statistic, as shown in Table \ref{table:hoeff}.
\begin{table}[ht]
\centering
\caption{Hoeffding D-statistic.} 
\begin{tabular}{|l|llllll|}
  \hline
 &    Temp95 &  Work & Res. & Tour. & Toy & Dow   \\
  \hline
  Load &   \textbf{0.3}& 0.04 & 0.09& 0.2 & 0.07 & 0.01\\
  Load$\backslash$Temp & 0.02& \textbf{0.2}& 0.01 & 0.03 & 0.02& 0.09 \\
  Load$\backslash$Temp,Work& 0.006 & 0.007 & 0.01 & 0.01 & \textbf{0.04} &  0.006 \\
   \hline 
\end{tabular}
\vspace{1em}
\flushleft
{\small The statistic was computed on all available days from  2019 to March 2022.  \textit{Load}$\backslash$\textit{Features} stands for the \textit{Load} corrected for the effect of the \textit{Features}. Here, \textit{Res.} stands for the \textit{Residence} index, and \textit{Tour.} for the \textit{Tourism} index.}
\label{table:hoeff}
\end{table}
This is a distribution-free measure of the dependence between variables \cite{hoeffding1948a}; 
the closer it is to 1, the greater the dependence.  
We then computed the Shapley values of the same variables using the \texttt{SHAFF} algorithm \cite{pmlr-v151-benard22a}; results are shown in Table \ref{table:shapley}.
We see that with the three ranking methods, the three most important variables, in order of importance, were the \textit{temperature}, the \textit{work} index, and the \textit{time of year}.
Of note, the effect of the \textit{work} index only became clear after filtering out the effect of the temperature on electricity demand.
We see that the importance of \textit{tourism} and \textit{time of year} decreases when correcting electricity demand for temperature, due to their high correlation with temperature. 
As a result of this analysis, the \textit{tourism}  and \textit{residents} indices did not seem to have a significant impact on French electricity demand.
\begin{table}[ht]
\centering
\caption{Shapley values.}
\begin{tabular}{|l|llllll|}
  \hline
 &    Temp95 &  Work & Res. & Tour. & Toy & Dow   \\
  \hline
  Load & \textbf{0.31}  & 0.05 & 0.06  & 0.14 & 0.21& 0.03  \\
  Load$\backslash$Temp & 0.041 & \textbf{0.26}& 0.035& 0.072& 0.11& 0.19\\
  Load$\backslash$Temp,Work& 0.11& 0.10& 0.06& 0.07& \textbf{0.28}& 0.04\\
   \hline 
\end{tabular}
\vspace{1em}
\flushleft
{\small Shapley values were computed on all available days from 2019 to March 2022.  \textit{Load}$\backslash$\textit{Features} stands for the \textit{Load} corrected for the effect of the \textit{Features}. Here, \textit{Res.} stands for the \textit{Residence} index, and \textit{Tour.} for the \textit{Tourism} index.}
\label{table:shapley}
\end{table}

\subsection*{\textit{work} index and calendar features}
\label{sec:remote_wkg}
 
The variable selection analysis in Section \ref{sec:variable_selection} showed that the \textit{work} indicator has a very strong effect on the electricity demand, being the second most explanatory variable.
To better understand this effect, we compared---in Table \ref{table_GAM}---the performance of GAMs where we progressively added the features in the order of importance suggested by the variable selection analysis.
The \textit{Temp} GAM corresponds to the model
\[\text{Load}_{h,t} = f_{h}(\text{Temp95}_{h,t}) + \varepsilon_{h,t}.\]
The \textit{Temp + Work} GAM corresponds to the model
\[\text{Load}_{h,t} = f_{h, 1}(\text{Temp95}_{h,t}) + f_{h, 2}(\text{Work}_{h,t})  + \varepsilon_{h,t}.\]
The \textit{Temp + Time} GAM corresponds to the model
\begin{align*}
    \text{Load}_{h,t} =\ & \sum_{i=1}^7 \sum_{j=0}^1 \alpha_{h,i,j}\;\boldsymbol{1}_{\text{DayType}_t=i}\;\boldsymbol{1}_{\text{DLS}_t=j}  + \beta \boldsymbol{1}_{\text{Holidays}_t} + f_{h,1}(\text{ToY}_t) \\
    &+ f_{h,2}(\text{Temp95}_{h,t}) + \varepsilon_{h,t}.
\end{align*}
The \textit{Temp + Time + Work} GAM corresponds to the model
\begin{align*}
    \text{Load}_{h,t} =\ & \sum_{i=1}^7 \sum_{j=0}^1 \alpha_{h,i,j}\;\boldsymbol{1}_{\text{DayType}_t=i}\;\boldsymbol{1}_{\text{DLS}_t=j}  + \beta \boldsymbol{1}_{\text{Holidays}_t} + f_{h,1}(\text{ToY}_t)\\
    & + f_{h,2}(\text{Temp95}_{h,t}) +  f_{h, 3}(\text{Work}_{h,t})  + \varepsilon_{h,t}.
\end{align*}
The \textit{Temp +  Work + Lags} GAM corresponds to the model
\begin{align*}
    \text{Load}_{h,t} =\ & f_{h, 1}(\text{Temp95}_{h,t}) + f_{h, 2}(\text{Work}_{h,t}) +\sum_{i=1}^7  \alpha_{h,i,j}\;\boldsymbol{1}_{\text{DayType}_t=i}\;\text{Load1D}_{h,t} \nonumber\\
    & + \beta \text{Load1W}_{h,t}  + \varepsilon_{h,t}.
\end{align*}
The \textit{Temp +  Time + Lags} GAM corresponds to the model
\begin{align*}
    \text{Load}_{h,t} =\ & \sum_{i=1}^7 \sum_{j=0}^1 \alpha_{h,i,j}\;\boldsymbol{1}_{\text{DayType}_t=i}\;\boldsymbol{1}_{\text{DLS}_t=j} + \beta \boldsymbol{1}_{\text{Holidays}_t} + f_{h,1}(\text{ToY}_t)  
    + f_{h,2}(\text{Temp95}_{h,t})\\
    &+ \sum_{i=1}^7 \gamma_{h,i,j}\;\boldsymbol{1}_{\text{DayType}_t=i}\;\text{Load1D}_{h,t}+ \lambda \text{Load1W}_{h,t}  + \varepsilon_{h,t}.
\end{align*}
\begin{figure*}[!t]
\centering
\subfloat[]{\includegraphics[width=0.49\textwidth]{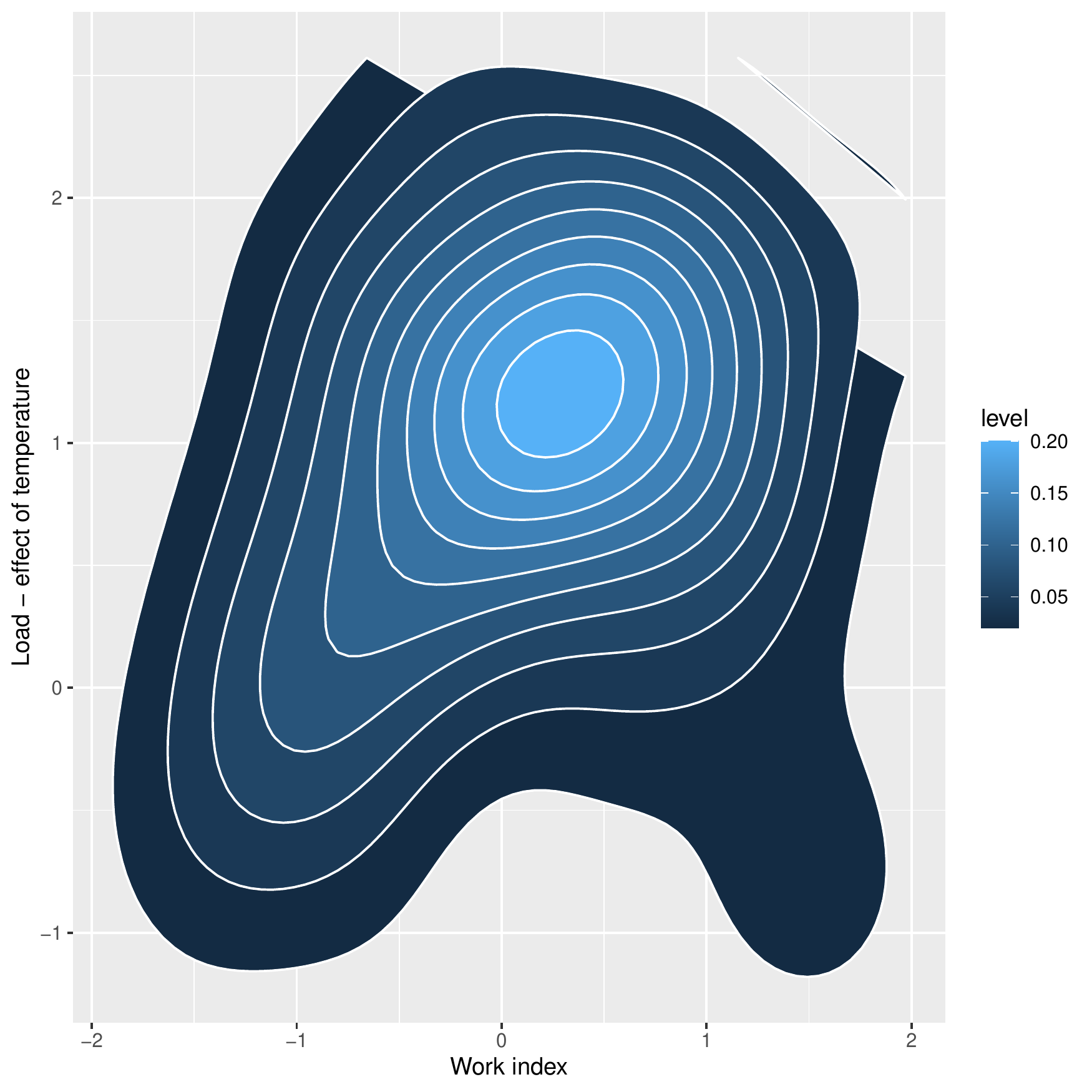}%
\label{fig:wednesday_left}}
\hfil
\subfloat[]{\includegraphics[width=0.49\textwidth]{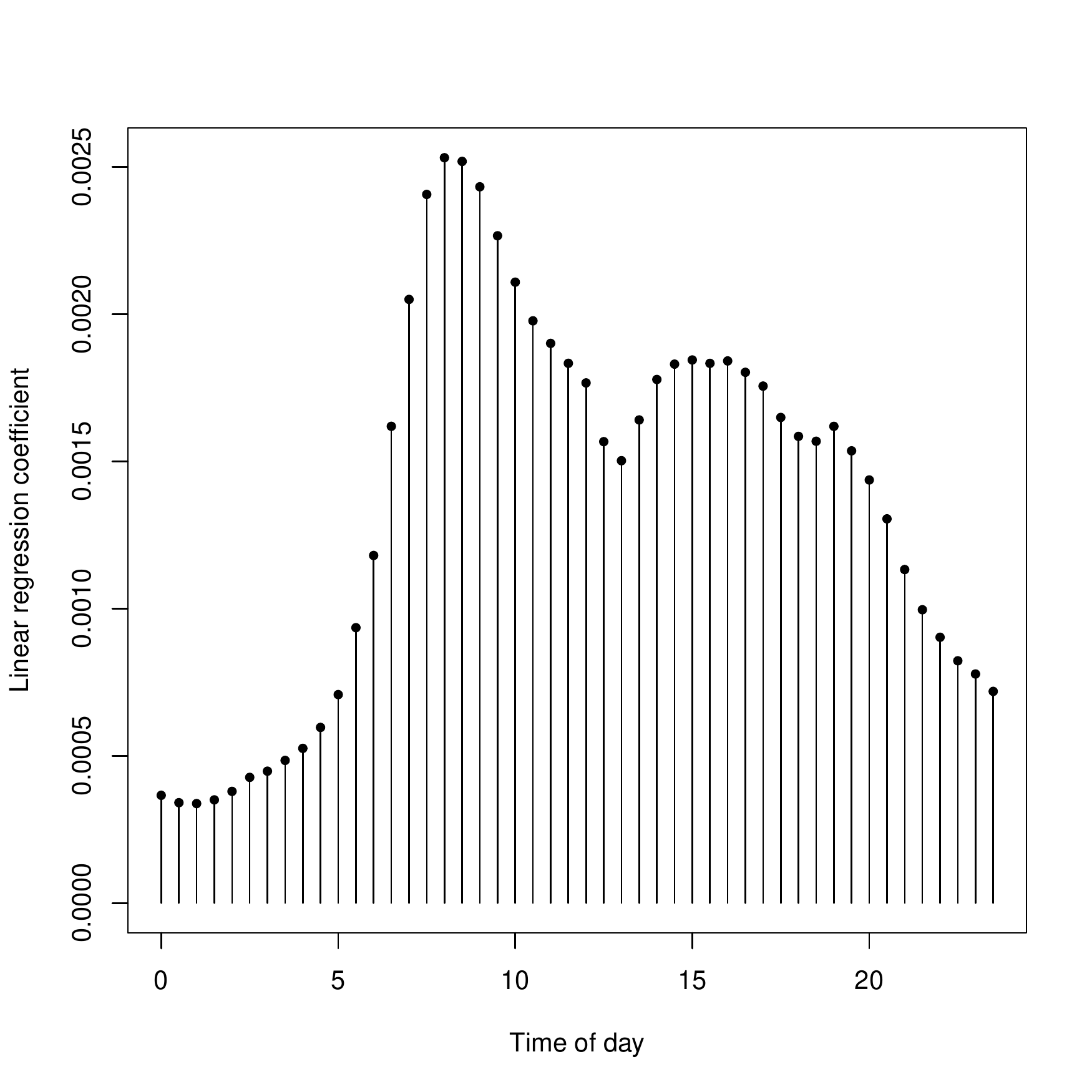}%
\label{fig:wednesday_right}}
\caption{Effect of the \textit{work} index on a given day at a given hour. (a) 2d density plots of residuals as function of the \textit{work} index at 10 a.m. on the Wednesdays between July 2019 and March 2022. (b) Regression coefficient of the \textit{work} index on electricity demand corrected for the effect of temperature on the training set spanning July 2019 to March 2022.}
\end{figure*}
The \textit{All variables} GAM corresponds to the model
\begin{align*}
    \text{Load}_{h,t} =\ &\sum_{i=1}^7 \sum_{j=0}^1 \alpha_{h,i,j}\;\boldsymbol{1}_{\text{DayType}_t=i}\;\boldsymbol{1}_{\text{DLS}_t=j} + \beta \boldsymbol{1}_{\text{Holidays}_t} + f_{h,1}(\text{ToY}_t)  + f_{h,2}(\text{Temp95}_{h,t})\\
    &+ \sum_{i=1}^7 \gamma_{h,i,j}\;\boldsymbol{1}_{\text{DayType}_t=i}\;\text{Load1D}_{h,t} + \lambda \text{Load1W}_{h,t} + f_{h,3}(\text{Work}_{h,t}) + \varepsilon_{h,t}.
\end{align*}
The p-values of the Fisher tests assessing the significance of the GAM effects were below 5$\%$ for all  GAMs.
We see in Table \ref{table_GAM} that replacing calendar data by the \textit{work} index was beneficial during atypical events whose behaviour differed from the past, i.e., the \textit{sobriety} period here.
Indeed, the time variables were only relevant during the \textit{normal period} spanning from July 2023 to September 2023, during which they still benefitted from the \textit{work} index. 
During the \textit{sobriety} period, the time variables---which only reconstruct past behaviour---were less explanatory than the \textit{work} index, which did not benefit from being coupled with them.
\begin{table}[ht]
\centering
\caption{Integration of mobility data in GAMs. This benchmark covers all days, including holidays. The performance is measured in RMSE (GW) and MAPE(\%).}
\begin{tabular}{lccccc}
  \hline
 & Normal period  & Sobriety \\
  \midrule
  \textit{Baseline} &&&\\
  Persistence (1 day) & 3.49  GW,  5.36 \% & 4.03 GW, 5.26 \%\\
  \midrule
    \textit{GAM} \\
  Temp & 3.53 GW, 6.78 \%& 6.13 GW, 9.60 \%\\
  Temp+Work & 1.62 GW, 3.10 \%& 4.98 GW, 8.11 \%\\
  Temp+Time & 1.33 GW, 2.46 \%& 5.60 GW, 9.62 \%\\
  Temp+Time+Work & 1.09 GW, 2.02 \%& 5.24 GW, 9.00 \%\\
  Temp+Work+Lags& 1.11 GW, 1.92 \%& \textbf{2.11} GW, \textbf{3.13} \%\\
  Temp+Time+Lags& 0.89 GW, 1.52 \%& 2.61 GW, 4.29 \%\\
  All variables & \textbf{0.80} GW, \textbf{1.38} \%& 2.60 GW, 4.32 \%\\
   \hline
\end{tabular}
\vspace{1em}
 
\label{table_GAM}
\end{table}

\subsection*{Work dynamics}
In Section~III.B of the main article, we explained how the \textit{work} index captures the effects of both the \textit{day of week} and  \textit{holidays} features. 
However, in both Section~II.B of the main paper and Section \ref{sec:remote_wkg}, we showed that the \textit{work} index improved the performance of the forecast, beyond the effect of the calendar features.
Let us briefly take a closer look at this effect.
To remove the effect of \textit{time of day} and \textit{holidays}, we worked on a specific day (here Wednesday) and removed holidays.
Figure \ref{fig:wednesday_left}  shows how the electricity demand on Wednesdays was still positively influenced by the \textit{work} index.
Furthermore, as expected, Figure \ref{fig:wednesday_right}  shows that the effect of the \textit{work} index was more important during working hours (from 6 a.m. to 8 p.m.).
These results confirm that on Wednesdays a high \textit{work} index corresponded to high electricity demand.
This effect could be due to economic growth (higher economic activity corresponding to both more people working which raises the \textit{work} index, and to higher electricity demand) and to energy saving due to remote working (a lower office occupancy corresponding both to a lower \textit{work} index and lower electricity demand).

\renewcommand\thesection{\thechapter.\arabic{section}}

\chapter{Forecasting time series with constraints}
\label{ch:contrib-3}

This chapter corresponds to the following paper: \citet{doumèche2025forecasting}.
\section{Introduction}

\paragraph{Time series forecasting.} Time series data are used extensively in many contemporary applications, such as forecasting supply and demand, pricing, macroeconomic indicators, weather, air quality, traffic, migration, and epidemic trends 
\citep{petropoulos2022forecasting}. 
However, regardless of the application domain, forecasting time series presents unique challenges due to inherent data characteristics such as observation correlations, non-stationarity, irregular sampling intervals, and missing values. These challenges limit the availability of relevant data and make it difficult for complex black-box or overparameterized learning architectures to perform effectively, even with rich historical data 
\citep{lim2021time}. 

\paragraph{Constraints in time series.} In this context, many modern frameworks incorporate physical constraints to improve the performance and interpretability of forecasting models. The strongest form of such constraints are typically derived from fundamental physical properties of the time series data and are represented by systems of differential equations. For example, weather forecasting often relies on solutions to the Navier-Stokes equations \citep[][]{schultz2021can}.
In addition to defining physical relationships, differential constraints can also serve as regularization mechanisms. For example, spatiotemporal regression on graphs can involve penalizing the spatial Laplacian of the regression function to enforce smoothness across spatial dimensions \citep[][]{jin2024spatio}.

However, time series rarely satisfy strict differential constraints, often adhering instead to more relaxed forms of constraints \citep[][]{coletta2023on}.
Perhaps the most successful example of such weak constraints are the generalized additive models \citep[GAMs,][]{hastie1986generalized}, which have been applied to time series forecasting in epidemiology \citep{wood2017generalized}, earth sciences \citep{augusting2009modeling}, and energy forecasting \citep{fasiolo2021fast}. 
GAMs model the target time series (or some parameters of its distribution) as a sum of nonlinear effects of the features, thereby constraining the shape of the regression function.
Another example of weak constraint appears in the context of spatiotemporal time series with hierarchical forecasting. Here, the goal is to combine regional forecasts into a global forecast by enforcing that the global forecast must be equal to the sum of the regional forecasts \citep{Wickramasuriya2019optimal}.
Although this may seem like a simple constraint, hierarchical forecasting is challenging because of a trade-off: using more granular regional data increases the available information, but also introduces more noise as compared to the aggregated total. Another common and powerful constraint in time series forecasting arises when combining multiple forecasts \citep{gaillard2014second}. 
This is done by creating a final forecast by weighting each of the initial forecasts, with the constraint that the sum of the weights must equal one.

\paragraph{PIML and time series.} Although  weak constraints have been studied individually and applied to real-world data, a unified and efficient approach is still lacking.
It is important here to mention physics-informed machine learning (PIML), which offers a promising way to integrate constraints into machine learning models. 
Based on the foundational work of \citet{raissi2019PINN}, PIML exploits the idea that constraints can be applied with neural networks and optimized by backpropagation, leading to the development of physics-informed neural networks (PINNs). 
PINNs have been successfully used to predict time series governed by partial differential equations (PDEs) in areas such as weather modeling \citep{kashinath2021physics}, and stiff chemical reactions \citep{ji2021stiff}. 
Weak constraints on the shape of the regression function have also been modeled with PINNs \citep[][]{daw2022lake}. 
However, PINNs often suffer from optimization instabilities and overfitting  \citep{doumeche2023convergence}.  
As a result, alternative methods have been developed for certain differential constraints that offer improved optimization properties over PINNs. For example, data assimilation techniques in weather forecasting have been shown to be consistent with the Navier-Stokes equations \citep{nickl2024on}. 
Moreover, \citet{doumeche2024physicsinformed} showed that forecasting with linear differential constraints can be formulated as a kernel method, yielding closed-form solutions to compute the exact empirical risk minimum. An additional advantage of this kernel modeling is that the learning algorithm can be executed on GPUs, leading to significant speedups compared to the gradient-descent-based optimization of PINNs \citep{doumèche2024physicsinformedkernellearning}.

\paragraph{Contributions.}
In this paper, we present a principled approach to effectively integrate constraints into time series forecasting. Each constrained problem is reformulated as the minimization of an empirical risk consisting of two key components: a data-driven term and a regularization term that enforces the smoothness of the function and the desired physical constraints.
For nonlinear regression tasks, we rely on a Fourier expansion.
Our framework allows for efficient computation of the exact minimizer of the empirical risk, which is easily optimized on GPUs for scalability and performance.

In Section~\ref{sec:weak}, we introduce a unified mathematical framework that connects empirical risks constrained by various forms of physical information. Notably, we highlight the importance of distinguishing between two categories of constraints: shape constraints, which limit the set of admissible functions, and learning constraints, which introduce an initial bias during parameter optimization. In Section~\ref{sec:shape}, we explore shape constraints and illustrate their relevance using the example of electricity demand forecasting.
In Section~\ref{sec:weight}, we define learning constraints and show how they can be applied to tourism forecasting.
This common modeling framework for shape and learning constraints allows for efficient integration of multiple constraints, as illustrated by the WeaKL-T in Section~\ref{sec:weight}, which combines hierarchical forecasting with additive models and transfer learning. 
Each empirical risk can then be minimized on a GPU using linear algebra, ensuring scalability and computational efficiency. 
This direct computation guarantees that the proposed estimator exactly minimizes the empirical risk, preventing convergence to potential local minima---a common limitation of modern iterative and gradient descent methods used in PINNs.
Our method achieves significant performance improvements over state-of-the-art approaches. 
The code for the numerical experiments and implementation is publicly available at \url{https://github.com/NathanDoumeche/WeaKL}.

\section{Incorporating constraints in time series forecasting}
\label{sec:weak}
Throughout the paper, we assume that $n$ observations $(X_{t_1}, Y_{t_1}), \ldots, (X_{t_n}, Y_{t_n})$ are drawn on $\mathbb{R}^{d_1} \times \mathbb{R}^{d_2}$. 
The indices $t_1, \ldots, t_n \in T$ correspond to the times at which an unknown stochastic process $(X,Y):=(X_t, Y_t)_{t \in T}$ is sampled.
Note that, all along the paper, the time steps need not be regularly sampled on the index set $T \subseteq \mathbb R$. 
We focus on supervised learning tasks that aim to estimate an unknown function $f^\star : \mathbb{R}^{d_1} \to \mathbb{R}^{d_2}$, under the assumption that $Y_t = f^\star(X_t) + \varepsilon_t$, where $\varepsilon$ is a random noise term. Without loss of generality, upon rescaling, we assume that $X_t:= (X_{1,t}, \hdots, X_{d_1,t}) \in [-\pi, \pi]^{d_1}$ and $-\pi \leq t_1 \leq  \cdots \leq  t_{n+1}\leq \pi$. The goal is to construct an estimator $\hat{f}$ for $f^\star$. 

A simple example to to keep in mind is when $Y$ is a stationary, regularly sampled time series with $t_j = j/n$, and the lagged value $X_j = Y_{t_{j-1}}$ serves as the only feature. In this specific case, where $d_1 = d_2$, the model simplifies to
$Y_t = f^\star(Y_{t-1/n})+\varepsilon_t$. Thus, the regression setting reduces to an autoregressive model. Of course, we will consider more complex models that go beyond this simple case.

\paragraph{Model parameterization.} We consider parameterized models of the form 
\begin{equation}
    f_{\theta}(X_t) = (f^1_{\theta}(X_t), \hdots, f^{d_2}_{\theta}(X_t)) =  (\langle \phi_1(X_t), \theta_1\rangle, \hdots, \langle \phi_{d_2}(X_t), \theta_{d_2}\rangle),
    \label{eq:model_def}
\end{equation} 
where each component  $f^\ell_\theta(X_t)$ is computed as the inner product of a feature map $\phi_\ell(X_t) \in \mathbb{C}^{D_\ell}$, with $D_\ell \in \mathbb N^\star$, and a vector $\theta_\ell \in \mathbb{C}^{D_\ell}$. 
The parameter vector $\theta \in \mathbb C^{D_1 + \cdots + D_{d_2}}$ of the model is defined as the concatenation of $\theta_1$, $\dots$, $\theta_{d_2}$. 
Note that $f_\theta$ is uniquely determined by $\theta$ and the maps~$\phi_\ell$.
To simplify the notation, we write $\dim(\theta) = D_1 + \cdots + D_{d_2}$. 

Our goal is to learn a parameter $\hat \theta \in \mathbb C^{\dim(\theta)}$ such that $\hat Y_t  = f_{\hat{\theta}}(X_t)$ is an estimator of the target $Y_t$.
Equivalently, $f_{\hat{\theta}}$ is an estimator of the target function $f^\star$. 
To this end, the core principle of our approach is to consider $\hat \theta$ to be a minimizer over $\mathbb C^{\dim(\theta)}$ of an empirical risk of the form
\begin{equation}
    L(\theta) = \frac{1}{n}\sum_{j=1}^n \|\Lambda(f_\theta(X_{t_j})-Y_{t_j})\|_2^2  + \|M\theta\|_2^2,
    \label{eq:risk}
\end{equation}
where $\Lambda$ and $M$ are complex-valued matrices with problem-dependent dimensions, which are not necessarily square. The matrix $M$ encodes a regularization penalty, which may include hyperparameters to be tuned through validation, as we will see in several examples.

\paragraph{Explicit formula for the empirical risk minimizer: WeaKL.}The following proposition shows how to compute the exact minimizer of \eqref{eq:risk}. 
(Throughout the document, $\ast$ denotes the conjugate transpose operation.)
\begin{prop}[Empirical risk minimizer.] 
\label{prop:emp_risk_min}
Suppose both $M$ and $\Lambda$ are injective.
Then, there is a unique minimizer to \eqref{eq:risk}, which takes the form
\begin{equation}
    \hat \theta = \Big( \Big( \sum_{j=1}^n \mathbb \Phi_{t_j}^\ast \Lambda^\ast \Lambda\mathbb \Phi_{t_j}\Big) + n M^\ast M\Big)^{-1} \sum_{j=1}^n \mathbb \Phi_{t_j}^\ast \Lambda^\ast \Lambda Y_{t_j},
    \label{eq:weakl}
\end{equation}
where $\mathbb \Phi_t$ is the $d_2\times \dim(\theta)$ block-wise diagonal feature matrix at time $t$, defined by
\begin{equation}
\mathbb \Phi_t = \begin{pmatrix}
    \phi_1(X_{t})^\ast & 0& 0 \\
    0 & \ddots & 0 \\
    0 & 0 & \phi_{d_2}(X_{t})^\ast
\end{pmatrix}
\label{eq:feature_matrix}.
\end{equation}
\end{prop}
This result, proven in Appendix~\ref{proof:kernel}, generalizes well-known results on kernel ridge regression \citep[see, e.g.,][Equation 10.17]{mehri2012foundations}. 
In the rest of the paper, we refer to the estimator $\hat \theta$ as the weak kernel learner (WeaKL). The strength of WeaKL lies in its exact computation via~\eqref{eq:weakl}. Unlike current implementations of GAMs and PINNs, WeaKL is free from optimization errors. Furthermore, since WeaKL relies solely on linear algebra, it can take advantage of GPU programming to accelerate the learning process. 
This efficiency enables effective hyperparameter optimization, as demonstrated in Section~\ref{sec:energy_crisis} through applications to electricity demand forecasting.

\paragraph{Algorithmic complexity.} The formula \eqref{eq:weakl} used in this article to minimize the empirical risk \eqref{eq:risk} can be implemented with a  complexity of $ O(\dim(\theta)^3 +  \dim(\theta)^2 n)$. 
Note that the dimensions $d_1$ and $d_2$ of the problem only impact the complexity of WeaKL through $\dim(\theta) = D_1 + \cdots + D_{d_2}$. 
By construction, $\dim(\theta) \geq d_2$, but the influence of $d_1$ is more subtle and depends on the chosen dimension $D_\ell$ of the maps $\phi_j: [-\pi, \pi]^{d_1}\to \mathbb{C}^{D_j}$. 
In particular, if all the maps have the same dimension, i.e., $D_\ell = D$, then $\dim(\theta) = Dd_2$.

Notably, this implementation runs in less than ten seconds on a standard GPU (e.g., an NVIDIA $L4$ with $24$ GB of RAM) when $\dim(\theta) \leq 10^3$ and $n \leq 10^5$. 
We believe that this framework is particularly well suited for time series, where data sampling is often costly, thus limiting both $n$ and $d_2$. Moreover, in many cases, the distribution of the target time series changes significantly over time, making only the most recent observations relevant for forecasting. This further limits the size of $n$. For example, in the Monash time series forecasting archive \citep{godahewa2021monash}, $19$ out of $30$ time series have $d_2 \leq 10^3$ and $n \leq 10^5$. 
However, there are relevant time series where either the dimension $d_2$ or the number of data points $n$ is large. 
In such cases, finding an exact minimizer of the empirical risk \eqref{eq:risk} becomes very computationally expensive. 
Efficient techniques have been developed to approximate the minimizer of \eqref{eq:risk} in these regimes \citep[see, e.g.,][]{meanti2020kernel}, but a detailed discussion of these methods is beyond the scope of this paper.

\paragraph{Some important examples.}
Let us illustrate the mechanism with two fundamental examples. Of course, the case where $\phi_\ell(x) = x$ and where $\Lambda$ and $M$ are identity matrices corresponds to the well-known ridge linear regression. 
On the other hand, a powerful example of a nonparametric regression map is the Fourier map, defined as $\phi_\ell(x) = (\exp(i \langle x, k \rangle / 2))_{\|k\|_\infty \leq m}^\top = (\exp(i \langle x, k \rangle / 2))_{-m\leq k_1, \hdots, k_{d_1} \leq m}^\top$, where the Fourier frequencies are truncated at $m \geq 0$. 
This map leverages the expressiveness of the Fourier basis to capture complex patterns in the data. Thus, for the $\ell$-th component of $f_{\theta}$, we consider the Fourier decomposition
\[
f^\ell_{ \theta}(x) =  \sum_{\|k\|_\infty \leq m}  \theta_{\ell,k} \exp(-i \langle x, k\rangle/2),
\]
which can approximate any function in $L^2([-\pi, \pi]^{d_1}, \mathbb{R})$ as $m \to \infty$. In this example, we have $\theta_{\ell}=(\theta_{\ell,k})_{\|k\|_\infty \leq m}^\top \in \mathbb C^{(2m+1)^d}$. 
Next, for $s \in \mathbb N^\star$, 
let $M$ be the $(2m+1)^{d_1}\times (2m+1)^{d_1}$ positive diagonal matrix such that
\[
\|M \theta_\ell\|_2^2  = \lambda \sum_{\|k\|_{\infty} \leq m} \theta_{\ell,k}^2 (1+\|k\|_2^{2s}),
\]
where $\lambda > 0$ is an hyperparameter.
Then, $\|M \theta_\ell\|_2$
is a Sobolev norm on the derivatives up to order $s$ of $f_{\theta_\ell}$.
When $\lambda = 1$, we will denote this norm by $\|f_{\theta}^\ell\|_{H^s}$. 
This approach regularizes the smoothness of $f_{\hat{\theta}}^{\ell}$, encouraging the recovery of smooth solutions. 
Moreover, choosing $\Lambda$ as the identity matrix and $\lambda = n^{-2s/(2s+d_1)}$ achieves the Sobolev minimax rate $\mathbb E(\|f_{\hat \theta}^\ell(X) -Y_\ell\|_2^2) = O(n^{-2s/(2s+d_1)})$ \citep{blanchard2020kernel}. 
This result justifies why the Fourier decomposition serves as an effective nonparametric mapping. 

These fundamental examples illustrate the richness of the approach, making it possible to incorporate constraints into models of chosen complexity, from very light models like linear regression, up to nonparametric models such as Fourier maps.

\paragraph{Classification of the constraints.} In order to clarify our discussion as much as possible, we find it helpful, after a thorough analysis of the existing literature, to consider two families of constraints. This distinction arises from the need to address two fundamentally different aspects of the forecasting problem.
\begin{enumerate}
\item {\bf Shape constraints}, described in Section~\ref{sec:shape}, include additive models, online adaption after a break, and forecast combinations (detailed in Appendix~\ref{sec:combination}). In these models, prior information is incorporated by selecting custom maps $\phi_\ell$. The set of admissible models  $f_\theta$ is thus restricted by shaping the structure of the function space through this choice of maps. Here, the matrix $M$ serves only as a regularization term, while $\Lambda$ is the identity matrix.

\item {\bf Learning constraints}, described in Section~\ref{sec:weight}, include transfer learning, hierarchical forecasting, and differential constraints (detailed in Appendix~\ref{sec:diff}). In these models, prior information or constraints are incorporated through the matrices $M$ and $\Lambda$. The goal is to increase the efficiency of parameter learning by introducing additional regularization.
\end{enumerate}
It is worth noting, however, that certain specific shape constraints cannot be penalized by a kernel norm, such as those in isotonic regression. In the conclusion, we discuss possible extensions to account for such constraints.

\section{Shape constraints}
\label{sec:shape}
\subsection*{Mathematical formulation}
In this section, we introduce relevant feature maps $\phi$ that incorporate prior knowledge about the shape of the function  $f^\star:[-\pi,\pi]^{d_1}\to \mathbb{C}^{d_2}$. To simplify the notation, we focus on the one-dimensional case where $d_2 = 1$ and $\Lambda = 1$. 
This simplification comes without loss of generality, since the feature maps developed in this section can be applied directly to \eqref{eq:model_def}.

As a result, the model reduces to $f_{\theta}(X_t) = \langle \phi_1(X_t), \theta_1 \rangle$, and \eqref{eq:weakl} simplifies to
\begin{equation}
    \hat \theta = ( \mathbb \Phi^\ast \mathbb \Phi + n M^\ast M)^{-1}  \mathbb \Phi^\ast \mathbb Y,
    \label{eq:weakl2}
\end{equation}
where $\mathbb Y = (Y_{t_1}, \hdots, Y_{t_n})^\top \in \mathbb R^n$ and the $n\times \dim(\theta)$ matrix $\mathbb \Phi$ takes the form 
\[ \mathbb \Phi = (\phi_1(X_{t_1})\mid \cdots \mid \phi_1(X_{t_n}))^\ast.
\]
Note that $\mathbb \Phi$ is the classical feature matrix, and that it is related to the matrix $\mathbb \Phi_t$ of \eqref{eq:feature_matrix} by $\mathbb \Phi^\ast \mathbb \Phi = \sum_{j=1}^n\mathbb \Phi_{t_j}^\ast \mathbb \Phi_{t_j} = \sum_{j=1}^n \phi_1(X_{t_j}) \phi_1(X_{t_j})^\ast$.

\paragraph{Additive model: Additive WeaKL.} The additive model constraint assumes that $f^\star(x_1, \hdots, x_{d_1}) = \sum_{\ell=1}^{d_1} g_\ell^\star(x_\ell)$, where $g_\ell^\star: \mathbb{R} \to \mathbb{R}$ are univariate functions. This constraint is widely used in data science, both in classical statistical models \citep{hastie1986generalized} and in modern neural network architectures \citep{agarwal2021neural}. Indeed, additive models are interpretable because the effect of each feature $x_\ell$ is captured by its corresponding function $g_\ell^\star$. In addition, univariate effects are easier to estimate than multivariate effects \citep{Ravikumar2009sparse}. These properties allow the development of efficient variable selection methods \citep[see, for example,][]{marra2011practical}, similar to those used in linear regression.

In our framework, the additivity constraint directly translates into the model as
\[
f_{\theta}(X_t) =  \langle \phi_{1}(X_{t}), \theta_{1}\rangle = \langle \phi_{1,1}(X_{1,t}), \theta_{1,1}\rangle + \cdots + \langle \phi_{1,d_1}(X_{d_1,t}), \theta_{1,d_1}\rangle,
\]
where $\phi_1$ is the concatenation of the maps $\phi_{1,\ell}$, and $\theta_1$ is the concatenation of the vectors  $\theta_{1,\ell}$. 
Note that the maps $\phi_{1,\ell}$ and the vectors $\theta_{1, \ell}$ can be multidimensional, depending on the model.
In this formulation, the effect of each feature is modeled by the function $g_\ell(x_\ell) = \langle \phi_{1,\ell}(x_\ell), \theta_{1,\ell}\rangle$, which can be either linear or nonlinear in $x_\ell$.
The empirical risk then takes the form
\begin{equation}
    L(\theta) = \frac{1}{n} \sum_{j=1}^n |f_\theta(X_{t_j}) - Y_{t_j}|^2 + \sum_{\ell=1}^{d_1}\lambda_\ell\|M_\ell\theta_{1,\ell}\|_2^2, \label{eq:weaklGAM}
\end{equation}
where $\lambda_\ell >0$ are hyperparameters and $M_\ell$ are regularization matrices.
There are three types of effects that can be taken into account:
\begin{itemize}
    \item[$(i)$] A linear effect is obtained by setting $\phi_{1,\ell}(x_\ell) = x_\ell \in \mathbb R$. 
    To regularize the parameter $\theta_{1, \ell}$, we set $M_\ell = 1$. This corresponds to a ridge penalty.
    \item[$(ii)$] A nonlinear effect can be modeled using the Fourier map $\phi_{1,\ell}(x_\ell) = (\exp(i  k x_\ell  / 2))_{-m\leq k \leq m}^\top$. 
    To regularize the parameter $\theta_{1, \ell}$, we set $M_\ell$ to be the $(2m+1)\times (2m+1)$ diagonal matrix defined by $M_\ell =\mathrm{Diag}((\sqrt{1+k^{2s}})_{-m\leq k\leq m})$, penalizing the Sobolev norm. 
    A common choice for the smoothing parameter $s$, as used in GAMs, is $s = 2$ \citep[see, e.g.,][]{wood2017generalized}.
    \item[$(iii)$] If $x_\ell$ is a categorical feature, i.e., $x_\ell$ takes values in a finite set $E$, we can define a bijection $\psi: E \to \{1, \hdots, |E|\}$. The effect of $x_\ell$ can then be modeled as $g_\ell(x_\ell) = \langle \phi_{1,\ell}(x_\ell), \theta_1 \rangle$, where $\phi_\ell = \phi \circ \psi$ and $\phi$ is the Fourier map with $m = \lfloor |E|/2 \rfloor$. To regularize the parameter $\theta_{1, \ell}$, we set $M_\ell$ as the identity matrix, which corresponds to applying a ridge penalty.
\end{itemize}
Overall, similar to GAMs, WeaKL can be optimized to fit additive models with both linear and nonlinear effects. The parameter $\hat \theta$ of the WeaKL can then be computed using \eqref{eq:weakl2}
with 
\[M = \begin{pmatrix}
        \sqrt{\lambda_1} M_1& 0  & 0\\
        0 & \ddots&  0\\
        0 & 0& \sqrt{\lambda_{d_1}} M_{d_1}
    \end{pmatrix}.\]
To stress that this WeaKL results from the enforcement of additive constraints, we call it the \textit{additive WeaKL}.
Note that, contrary to GAMs where identifiability issues must be addressed \citep{wood2017generalized}, WeaKL does not require further regularization, since $\hat \theta$ is the unique minimizer of the empirical risk~$L$. 
Note that the hyperparameters $\lambda_\ell$, along with the number $m$ of Fourier modes and the choice of feature maps  $\phi_\ell$, can be determined by model selection, as described in Appendix~\ref{sec:tuning}.

\paragraph{Online adaption after a break: Online WeaKL.}
For many time series, the dependence of $Y$ on $X$ can vary over time. For example, the behavior of $Y$ may change rapidly following extreme events, resulting in structural breaks. A notable example is the shift in electricity demand during the COVID-19 lockdowns, as illustrated in use case $1$. To provide a clear mathematical framework, we assume that the distribution of $(X, Y)$ follows an additive model that evolves smoothly over time. Formally, considering $(t, X_t)$ as a feature vector, we assume that
\begin{equation}
    f^\star(t, x_1, \hdots, x_{d_1}) = h_0^\star(t)+ \sum_{\ell=1}^{d_1} (1+ h_\ell^\star(t))  g_\ell^\star(x_\ell),
    \label{eq:model}
\end{equation}
where $g_\ell^\star$ and $h_\ell^\star$ are univariate functions. This model forms the core of the Kalman-Viking algorithm \citep{vilmarest2024viking}, which has demonstrated state-of-the-art performance in forecasting electricity demand and renewable energy production \citep{obst2021adaptative, vilmarest2022state, Vilamarest2024adaptive}. 

We assume that we have at hand estimators $\hat g_\ell$ of $g_\ell^\star$ that we want to update over time. For example, these estimators can be obtained by fitting an additive WeaKL model, initially assuming $h_\ell^\star = 0$. The functions $h_\ell^\star$ are then estimated by minimizing the empirical risk
\begin{equation}
    L(\theta) = \frac{1}{n}\sum_{j=1}^n \Big|h_{\theta_0}(t_j) + \sum_{\ell=1}^{d_1} (1+h_{\theta_\ell}(t_j)) \hat g_\ell(X_{\ell,t_j})-Y_{t_j}\Big|^2 + \sum_{0\leq \ell \leq d_1} \lambda_\ell\|h_{\theta_\ell}\|_{H^s}^2,
    \label{eq:risk_online}
\end{equation}
where $\lambda_\ell > 0$ are hyperparameters regularizing the smoothness of the functions $h_{\theta_\ell}$. Here, $h_{\theta}(t) = \langle \phi(t), \theta\rangle$, and $\phi$ is the Fourier map $\phi(t) =(\exp(i k t/2))_{-m\leq k \leq m}^\top$. The prior $h_{\theta_\ell} \simeq 0$ reflects the idea that the best a priori estimate of $Y$'s behavior follows the stable additive model. Defining $W_t = Y_t - \sum_{\ell=1}^{d_1}\hat g_\ell(X_{\ell,t})$, the empirical risk can be reformulated as
\begin{equation*}
    L(\theta) = \frac{1}{n}\sum_{j=1}^n |\langle \phi_1(t_j, X_{t_j}), \theta\rangle - W_{t_j}|^2 + \|M \theta\|_2^2,
\end{equation*}
with
$\phi_1(t, X_t) = 
    ((\exp(ik t/2))_{- m\leq k \leq  m},
    (\hat g_\ell(X_{\ell,t})\exp(ik t/2))_{- m\leq k \leq  m})_{\ell=1}^{d_1})^\top \in \mathbb C^{(2m+1)(d_1+1)}$,
\[M = \begin{pmatrix}
        \sqrt{\lambda_0} D& 0  & 0\\
        0 & \ddots&  0\\
        0 & 0& \sqrt{\lambda_{d_1}} D
    \end{pmatrix},\]
and $D$ is the $(2m+1)\times (2m+1)$ diagonal matrix
$D =\mathrm{Diag}((\sqrt{1+k^{2s}})_{-m\leq k\leq m})$.
From \eqref{eq:weakl2}, we deduce that the unique minimizer of the empirical loss $L$ is
\begin{equation}
    \hat \theta  = ({\mathbb{\Phi}} ^\ast {\mathbb{\Phi}} + n  M^\ast M)^{-1}{\mathbb \Phi}^\ast  \mathbb W,
    \label{eq:online}
\end{equation}
where  $\mathbb W = (W_{t_1}, \hdots, W_{t_n})^\top \in \mathbb R^n$. 

This formulation allows to forecast the time series $Y$ at the next time step, $t_{n+1}$, using
\begin{align*}
\hat Y_{t_{n+1}} &= f_{\hat \theta}(t_{n+1}, X_{t_{n+1}}) = \langle \phi_1(t_{n+1}, X_{t_{n+1}}), \hat \theta\rangle \\
&=   h_{\hat \theta_0}(t_{n+1}) + \sum_{\ell=1}^{d_1} (1+h_{\hat \theta_\ell}(t_{n+1})) \hat g_\ell(X_{\ell,t_{n+1}}).
\end{align*}
Since the model is continuously updated over time, this corresponds to an online learning setting.
To emphasize that Equation~\eqref{eq:online} arises from an online adaptation process, we refer to this model as the \textit{online WeaKL}.
Unlike the Viking algorithm of \citet{vilmarest2024viking}, which approximates the minimizer of the empirical risk through an iterative process, online WeaKL offers a closed-form solution and exploits GPU parallelization for significant speedups.
As shown in Section~\ref{sec:energy_crisis}, our approach leads to improved performance in electricity demand forecasting.

\subsection*{Application to electricity load forecasting}
\label{sec:energy_crisis}
In this subsection, we apply shape constraints to two use cases in electricity demand forecasting and demonstrate the effectiveness of our approach.
In these electricity demand forecasting problems, the focus is on short-term forecasting, with particular emphasis on the recent non-stationarities caused by the COVID-19 lockdowns and by the energy crisis.

\paragraph{Electricity load forecasting and non-stationarity.} Accurate demand forecasting is critical due to the costly nature of electricity storage, coupled with the need for supply to continuously match demand.  
Short-term load forecasting, especially for 24-hour horizons, is particularly valuable for making operational decisions in both the power industry and electricity markets.
Although the cost of forecasting errors is difficult to quantify, a $1\%$ reduction in error is estimated to save utilities several hundred thousand USD per gigawatt of peak demand \citep{hong2016probabilistic}. 
Recent events such as the COVID-19 shutdown have significantly affected electricity demand, highlighting the need for updated forecasting models  \citep{zarbakhsh2022human}.

\paragraph{Use case 1: Load forecasting during COVID.} In this first use case, we test the performance of our WeaKL on the IEEE DataPort Competition on Day-Ahead Electricity
Load Forecasting \citep{Farrokhabadi2022day}.
Here, the goal is to forecast the electricity demand of an unknown country during the period following the Covid-19 lockdown.
The winning model of this competition was the Viking model of Team~4 \citep{vilmarest2022state}, with a mean absolute error (MAE) of $10.9$ gigawatts (GW). For comparison, a direct translation of their model into the online WeaKL framework---using the same features and maintaining the same additive effects---results in an MAE of $10.5$ GW. In parallel, we also apply the online WeaKL methodology without relying on the variables selected by \citet{vilmarest2022state}. Instead, we determine the optimal hyperparameters $\lambda_\ell$ and select the feature maps $\phi_\ell$ through a hyperparameter tuning process (see Appendix~\ref{sec:tuning}). This leads to a different selected model with a MAE of $9.9$ GW (see Appendix~\ref{sec:case_study1} for a complete description of the models). 
Thus, the online WeaKL given by \eqref{eq:online} outperforms the state-of-the-art by $9\%$. 
As done in the IEEE competition \citep{Farrokhabadi2022day}, we assess the significance of this result by evaluating the MAE skill score using a block bootstrap approach (see Appendix~\ref{sec:case_study1}). 
It shows that the online WeaKL outperforms the winning model proposed by \citet{vilmarest2022state} with a probability above $90\%$.
The updated results of the competition are presented in Table~\ref{tab:ieee}. 
Note that a great variety of models were benchmarked in this competition, like Kalman filters (Team~4), autoregressive models (Teams~4 and 7), random forests (Teams~4 and 6), gradient boosting (Teams~6 and 36),  deep residual networks (Team~19), and averaging (Team~13).

\begin{table}[h]
\centering
\caption{Performance of the online WeaKL and of the top $10$ participants of the IEEE competiton. A specific bootstrap test shows that the WeaKL significantly outperform the winning team.}
\begin{tabular}{lccccccccccc}
\toprule
Team & WeaKL &  4 &   14 &   7 &   36 &   19 &   23 &   9 &   25 &   13 &   26 \\
\midrule
MAE (GW)  & \textbf{9.9} & 10.9 & 11.8 & 11.9 & 12.3 & 12.3 & 13.9 & 14.2 & 14.3 & 14.6 & 15.4\\
\bottomrule
\end{tabular}

\label{tab:ieee}
\end{table}

\paragraph{Use case 2: Load forecasting during the energy crisis.}
In this second use case, we evaluate the performance of our WeaKL within the open source benchmark framework proposed by \citet{doumeche2023human}. This benchmark provides a comprehensive evaluation of electricity demand forecasting models, incorporating the GAM boosting model of \cite{bentaieb2014a}, the GAM of \cite{obst2021adaptative}, the Kalman models of \cite{vilmarest2022state}, the time series random forests of \cite{gohery2023random}, and the Viking model of \cite{Vilamarest2024adaptive}. 
The goal here is to forecast the French electricity demand during the energy crisis in the winter of 2022-2023. Following the war in Ukraine and maintenance problems at nuclear power plants, electricity prices reached an all-time high at the end of the summer of 2022. In this context, French electricity demand decreased by $10\%$ compared to its historical trends \citep{doumeche2023human}. 
This significant shift in electricity demand can be interpreted as a structural break, which justifies the application of the online WeaKL given by \eqref{eq:online}.

In this benchmark, the models are trained from 8 January 2013 to 1 September 2022, and then evaluated from 1 September 2022 to 28 February 2023.
The dataset consists of temperature data from the French meteorological administration \citet{meteoFrance}, and electricity demand data from the French  transmission system operator \citet{rteData}, sampled with a half-hour resolution.  
This translates into the feature variable 
\[X =(\mathrm{Load}_1, \mathrm{Load}_7, \mathrm{Temp}, \mathrm{Temp}_{950},  \mathrm{Temp}_{\mathrm{max 950}}, \mathrm{Temp}_{\mathrm{min 950}}, \mathrm{ToY},  \mathrm{DoW}, \mathrm{Holiday},t),
\]
where $\mathrm{Load}_1$ and $\mathrm{Load}_7$ are the electricity demand lagged by one day and seven days, $\mathrm{Temp}$ is the temperature, and $\mathrm{Temp}_{950}$,  $\mathrm{Temp}_{\mathrm{max 950}}$, and $\mathrm{Temp}_{\mathrm{min 950}}$ are smoothed versions of $\mathrm{Temp}$. The time of year $\mathrm{ToY} \in \{1, \hdots, 365\}$ encodes the position within the year. 
The day of the week $\mathrm{DoW} \in \{1, \hdots, 7\}$ encodes the position within the week. 
In addition, $\mathrm{Holiday}$ is a boolean variable set to one during holidays, and $t$ is the timestamp. 
Here, the target $Y = \mathrm{Load}$ is the electricity demand, so $d_1 = 10$ and $d_2 = 1$.

We compare the performance of two of our WeaKLs against this benchmark.
First, our additive WeaKL is a direct translation of the GAM formula proposed by  \cite{obst2021adaptative} into the additive WeaKL framework given by \eqref{eq:weaklGAM}. Thus, $f_\theta(x) = \sum_{\ell=1}^{10} g_\ell(x_\ell)$, where:
\begin{itemize}
    \item the effects $g_1$, $g_2$, and $g_{10}$ of $\mathrm{Load}_1$, $\mathrm{Load}_7$, and $t$ are linear,
    \item the effects $g_3,\dots, g_7$ of $\mathrm{Temp}$, $\mathrm{Temp}_{950}$,  $\mathrm{Temp}_{\mathrm{max 950}}$, $ \mathrm{Temp}_{\mathrm{min 950}}$, and $\mathrm{ToY}$ are nonlinear with $m=10$,
    \item the effects $g_8$ and $g_9$ of   $\mathrm{DoW}$ and $\mathrm{Holiday}$ are categorical with $|E| = 7$ and $|E| = 2$.
\end{itemize}

\begin{wrapfigure}[\lenghtfig]{r}{0.4\textwidth}
    \centering 
    \vspace{-1em}
    \includegraphics[width=\linewidth, trim={0.4cm 0.3cm 0.2cm 0.9cm},clip]{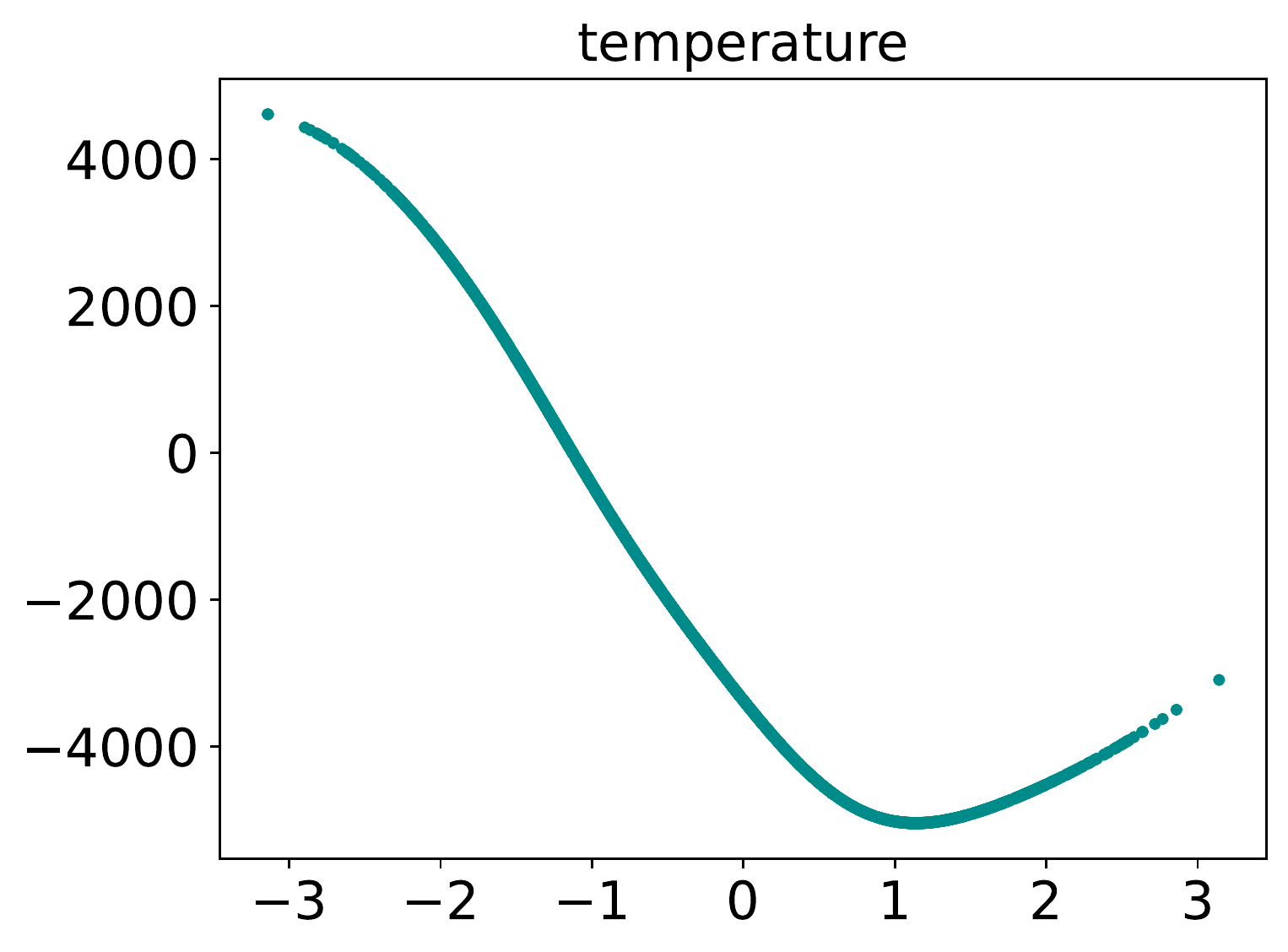}
    \vspace{-2em}
    \caption{Effect in MW of the temperature in the additive WeaKL.}
    \label{fig:WeaKL_effect}
 \end{wrapfigure}
\noindent The weights $\theta$ are learned using data from 2013 to 2021, while the optimal hyperparameters $\lambda_1, \dots, \lambda_{10}$ are tuned using a validation set covering the period from 2021 to 2022.
Once the additive WeaKL is learned, it becomes straightforward to interpret the impact of each feature on the model. For example, the effect $\hat g_3: \mathrm{Temp} \mapsto \langle\phi_{1,3}(\mathrm{Temp}), \hat \theta_{1, 3}\rangle $ of the rescaled temperature feature ($\mathrm{Temp} \in [-\pi, \pi]$) is illustrated in Figure~\ref{fig:WeaKL_effect}.

Second, our online WeaKL is the online adaptation of $f_\theta$  in response to a structural break, as described by \eqref{eq:online}.
The hyperparameters $\lambda_0, \dots, \lambda_{10}$ in \eqref{eq:risk_online} are chosen to minimize the error over a validation period from $1$ April $2020$ to $1$ June $2020$, corresponding to the first COVID-19 lockdown. Note that this validation period does not immediately precede the test period, which is uncommon in time series analysis. However, this choice ensures that the validation period contains a structural break, making it as similar as possible to the test period. Next, the functions $h_0, \dots, h_{10}$ in \eqref{eq:model} are trained on a period starting from $1$ July $2020$, and updated online. 

The results are summarized in Table~\ref{table_score_target_agg2bid}. The errors and their standard deviations are assessed by stationary block bootstrap (see Appendix~\ref{sec:block-bootstrap}). Since holidays are notoriously difficult to predict, performance is evaluated over the entire period (referred to as \textit{Including holidays}), and separately excluding holidays and the days immediately before and after (referred to as \textit{Excluding holidays}). 
Over both test periods, the additive WeaKL  significantly outperforms the GAM, while the online WeaKL outperforms the state-of-the-art by more than $10\%$ across all metrics.

\begin{figure}
    \centering
    \includegraphics[width=1\linewidth, trim={2.2cm 1.2cm 3cm 2.1cm},clip]{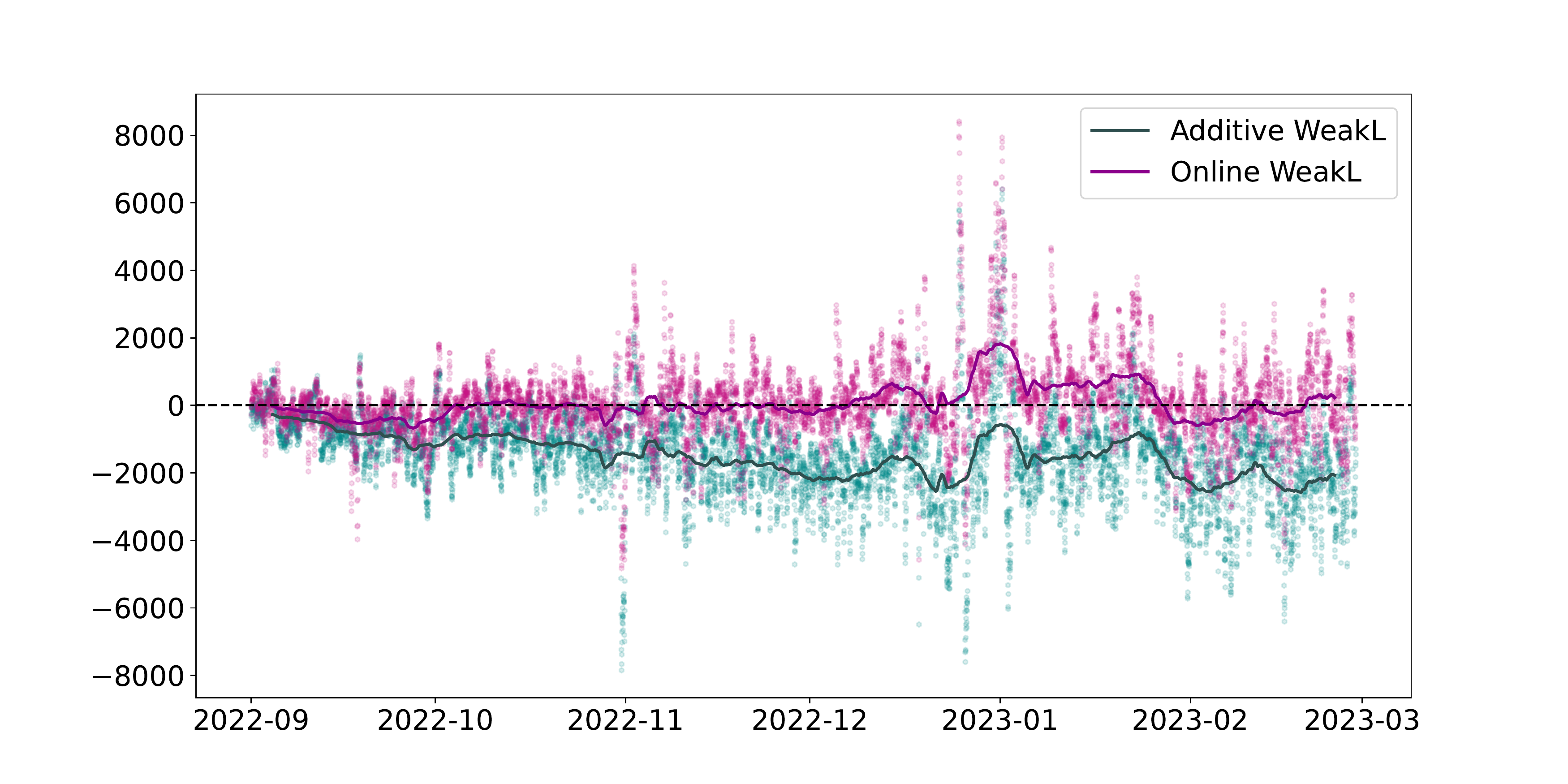}
    \caption{Error $Y_t - \hat Y_t$ in MW of the WeaKLs on the test period including holidays. Dots represent individual observations, while the bold curves indicate the one-week moving averages.}
    \label{fig:err_time_weaKL}
\end{figure}

Figure~\ref{fig:err_time_weaKL} shows the errors of the WeaKLs as a function of time during the test period, which includes holidays. 
During the sobriety period, electricity demand decreased, causing the additive WeaKL to overestimate demand, resulting in a negative bias. Interestingly, this bias is effectively corrected by the online WeaKL, which explains its strong performance. This shows that the online update of the effects effectively corrects biases caused by shifts in the data distribution.

Then, we compare the running time of the algorithms. 
Note that, during hyperparameter tuning, the GPU implementation of WeaKL makes it possible to train $1.6\times 10^5$ additive WeaKL over a period of eight years in less than five minutes on a single standard GPU (NVIDIA $L4$). 
As for the online WeaKL, the training is more computationally intensive because the model must be updated in an online fashion.
However, training $9.2 \times 10^3$ online WeaKLs over a period of two years takes less than two minutes.
This approach is faster than the Viking algorithm, which takes over $45$ minutes to evaluate the same number of parameter sets on the same dataset, even when using $10$ CPUs in parallel. A detailed comparison of the running times for all algorithms is provided in Appendix~\ref{sec:sobriety}.
\begin{table}[H]
\centering
\caption{Benchmark for load forecasting during the energy crisis}
\begin{tabular*}{\textwidth}{@{\extracolsep\fill}lcccc}
  \toprule
  & \multicolumn{2}{@{}c@{}}{ Including holidays} & \multicolumn{2}{@{}c@{}}{Excluding holidays} \\\cmidrule{2-3}\cmidrule{4-5}%
 & RMSE (GW)& MAPE (\%) &  RMSE (GW)& MAPE (\%)\\
  \midrule
  \textit{Statistical model} &&&&\\
  Persistence (1 day) & 4.0$\pm$0.2 & 5.5$\pm$0.3 & 4.0$\pm$0.2  & 5.0$\pm$0.3\\
  SARIMA  &  2.4$\pm$0.2   & 3.1$\pm$0.2 & 2.0$\pm$0.2  & 2.6$\pm$0.2\\
  GAM & 2.3$\pm$0.1 & 3.5$\pm$0.2   & 1.70$\pm$0.06 & 2.6$\pm$0.1 \\
  \midrule
    \textit{Data assimilation }\\
  Static Kalman & 2.1$\pm$0.1 & 3.1$\pm$0.2   &  1.43$\pm$0.05 & 2.20$\pm$0.08 \\
  Dynamic Kalman & 1.4$\pm$0.1 & 1.9$\pm$0.1   & 1.10$\pm$0.04 & 1.58$\pm$0.05  \\
    Viking & 1.5$\pm$0.1 & 1.8$\pm$0.1 &  0.98$\pm$0.04 & 1.33$\pm$0.04\\
    Aggregation & 1.4$\pm$0.1 & 1.8$\pm$0.1 & 0.96$\pm$0.04 & 1.36$\pm$0.04\\
    \midrule
    \textit{Machine learning}\\
    GAM boosting & 2.6$\pm$0.2 & 3.7$\pm$0.2 & 2.3$\pm$0.1 & 3.3$\pm$0.2 \\
    Random forests &  2.5$\pm   $0.2& 3.5$\pm$0.2& 2.1$\pm$0.1 & 3.0$\pm$0.1\\
    Random forests + bootstrap & 2.2$\pm$0.2 & 3.0$\pm$0.2 & 1.9$\pm$0.1 & 2.6$\pm$0.1\\
    \midrule
    \textit{WeaKLs}\\
    Additive WeaKL & 1.95$\pm$0.08 & 3.0 $\pm$0.1& 1.55$\pm$0.06 & 2.32$\pm$0.09  \\
    Online WeaKL &  \textbf{1.14$\pm$0.09}& \textbf{1.5$\pm$0.1}&  \textbf{0.87$\pm$0.04 }& \textbf{1.17$\pm$0.05} \\
   \bottomrule
\end{tabular*}
\label{table_score_target_agg2bid}
\end{table}
Both use cases demonstrate that WeaKL models are very powerful. Not only are they highly interpretable---thanks to their ability to fit into a common framework and produce simple formulas---but they are also competitive with state-of-the-art techniques in terms of both optimization efficiency (they can run on GPUs) and performance (measured by MAPE and RMSE).
\section{Learning constraints}
\label{sec:weight}
\subsection*{Mathematical formulation}
Section~\ref{sec:shape} focused on imposing constraints on the shape of the regression function $f^\star$. In contrast, the goal of the present section is to impose constraints on the parameter $\theta$. 
We begin with a ge\-ne\-ral method to enforce linear constraints on $\theta$, and subsequently apply this framework to transfer learning, hierarchical forecasting, and differential constraints.

\paragraph{Linear constraints.}
Here, we assume that $f^\star$ satisfies a linear constraint.
By construction of $f_\theta$ in \eqref{eq:model_def}, such a linear constraint directly translates into a constraint on $\theta$.
For example, the linear constraint $f^{\star\,1}(X_t) = 2f^{\star\,2}(X_t)$ can be implemented by enforcing $\theta_1 = 2\theta_2$.
Thus, in the following, we assume a prior on $\theta$ in the form of a linear constraint. Formally, we want to enforce that $\theta \in \mathcal{S}$, where $\mathcal{S}$ is a known linear subspace of $\mathbb{C}^{\dim(\theta)}$.
Given an injective $\dim(\theta) \times \dim(\mathcal{S})$ matrix $P$ such that $\mathrm{Im}(P) = \mathcal{S}$, then, as shown in Lemma~\ref{lem:ortho}, $\|C\theta\|_2^2$ is the square of the Euclidean distance between $\theta$ and $\mathcal S$, where $C=\mathrm{I}_{\dim(\theta)} - P(P^\ast P)^{-1}P^\ast$.  In particular, $\|C\theta\|_2^2 = 0 $ is equivalent to $\theta \in \mathcal S$, and $\|C\theta\|_2^2 = \|\theta\|_2^2$ if $\theta \in \mathcal S^\perp$. From this observation, there are two ways to enforce $\theta \in \mathcal S$ in the empirical risk \eqref{eq:risk}.

On the one hand, suppose that $f^\star$ exactly satisfies the linear constraint.
This happens in particular when the constraint results from a physical law. 
For example, to build upon the use cases of Section~\ref{sec:energy_crisis}, assume that we want to forecast the electricity load of different regions of France, i.e., the target $Y \in \mathbb R^3$ is such that $Y_1$ is the load of southern France, $Y_2$ is the load of northern France, and $Y_3 = Y_1+Y_2$ is the national load. 
This prototypical example of hierarchical forecasting is presented in Section~\ref{sec:toy-example}, where we show how incorporating even a simple constraint can significantly improve the model's performance. In this example, we know that $f^\star$ satisfies the constraint $f^{\star\,3} = f^{\star\,1} + f^{\star\,2}$.
When dealing with such exact priors, a sound approach is to consider only parameters $\theta$ such that $C\theta = 0$, or equivalently, $\theta = P\theta'$. Letting $\Pi_\ell$ be the $D_\ell\times \dim(\theta)$ projection matrix such that $\theta_\ell = \Pi_\ell \theta$, we have $\langle \phi_\ell (X_t), \theta_\ell\rangle = \langle \phi_\ell (X_t), \Pi_\ell \theta\rangle  = \langle P^\ast \Pi_\ell^\ast \phi_\ell (X_t),  \theta'\rangle$. 
Thus, minimizing the empirical risk \eqref{eq:risk} over $\theta' \in \mathbb C^{\dim(\mathcal S)}$ simply requires changing $\phi_\ell$ to $P^\ast \Pi_\ell^\ast \phi_\ell$, which is equivalent to replacing $\mathbb{\Phi}_t$ with $\mathbb{\Phi}_t P$ in \eqref{eq:weakl}. 

On the other hand, suppose that the linear constraint serves as a good but inexact prior. 
For example, building on the last example, let $X_t$ be the average temperature in France at time $t$. We expect the loads $Y_1$ in southern France and $Y_2$ in northern France to behave similarly. 
In both regions, lower temperatures lead to increased heating usage (and thus higher loads), while higher temperatures result in increased cooling usage (also leading to higher loads). 
Therefore, $f^{\star\,1}$ and $f^{\star\,2}$ share the same shape, resulting in the prior $f^{\star\,1} \simeq f^{\star\,2}$. 
This prototypical example of transfer learning is explored in the following paragraphs. Such inexact constraints can be enforced by adding a penalty $\lambda \|C\theta\|_2^2$ in the empirical risk \eqref{eq:risk}, where $\lambda > 0$ is an hyperparameter. (Equivalently, this only consists in replacing $M$ with $(\sqrt{\lambda} C^\top \mid  M^\top)^\top$ in \eqref{eq:risk}.)
This ensures that $\|C\hat \theta\|_2^2$ is small, while allowing the model to learn functions that do not exactly satisfy the constraint. 

These approaches are statistically sound, since under the assumption that $Y_t = f_{\theta^\star}(X_t)+ \varepsilon_t$, where $\theta^\star \in \mathcal{S}$, both estimators have lower errors compared to unconstrained regression. This is true in the sense that, almost surely,
\[\frac{1}{n}\sum_{j=1}^n\| f_{\theta^\star}(X_{t_j}) - f_{\hat \theta_C}(X_{t_j})\|_2^2 + \|M(\theta^\star- \hat \theta_C)\|_2^2\leq \frac{1}{n}\sum_{j=1}^n\| f_{\theta^\star}(X_{t_j}) - f_{\hat \theta}(X_{t_j})\|_2^2 + \|M(\theta^\star- \hat \theta)\|_2^2,\]
where $\hat \theta$ is the unconstrained WeaKL and $\hat \theta_C$ is a WeaKL integrating the constraint $C\theta^\star \simeq 0$ (see Proposition~\ref{prop:prop_lin} and Remark~\ref{rem:comment_prop_lin}).

\paragraph{Transfer learning.} Transfer learning is a framework designed to exploit similarities between different prediction tasks when $d_2 >1$. The simplest case involves predicting multiple targets $Y_1, \hdots, Y_{d_2}$ with similar features $X_1, \hdots, X_{d_2}$.
For example, suppose we want to forecast the electricity demand of $d_2$ cities. Here, $Y_\ell$ is the electricity demand of the city $\ell$, while $X_\ell$ is the average temperature in city $\ell$.
The general function $f^\star$ estimating $(Y_1, \hdots, Y_{d_2})$ can be expressed as $f^\star(X) = f^\star(X_1, \hdots, X_{d_2}) = (f^{\star\,1}(X_1), \hdots, f^{\star\,d_2}(X_{d_2}))$. The transfer learning assumption is $f^{\star\,1} \simeq \cdots \simeq f^{\star\,d_2}$. 
Equivalently, this corresponds to the linear constraint $\theta \in \mathrm{Im}(P)$, where $P = (\mathrm{I}_{2m+1} \mid \cdots \mid \mathrm{I}_{2m+1})^\top$ is a $(2m+1)d_1\times (2m+1)$ matrix. 
Thus, one can apply the framework of the last paragraph on linear constraints as inexact prior using $P$.

\paragraph{Hierarchical forecasting.} Hierarchical forecasting involves predicting multiple time series that are linked by summation constraints. This approach was introduced by \citet{athanasopoulos2009hierarchical} to forecast Australian domestic tourism. Tourism can be analyzed at various geographic scales. 
For example, at time $t$, one could consider the total number $Y_{A,t}$ of tourists in Australia, and the number $Y_{S_i,t}$ of tourists in each of the seven Australian states $S_1,\hdots, S_7$. By definition, $Y_{A,t}$ is the sum of the $Y_{S_i, t}$, which leads to the exact summation constraint $Y_{A,t} = \sum_{i=1}^7 Y_{S_i, t}$. Furthermore, since each state $S_i$ is composed of $z_i$ zones $Z_{i,1}$, $\dots$, $Z_{i, z_i}$, an additional hierarchical level can be introduced. 
Note that the number of zones depends on the state, for a total of 27 zones.
This results in another set of summation constraints
$Y_{S_i, t} =  Y_{Z_{i,1}, t}+\cdots + Y_{Z_{i,z_i}, t}$. 
\begin{figure}
    \centering
    \includegraphics[width=0.7\linewidth, trim={0 3.5cm 0 0},clip]{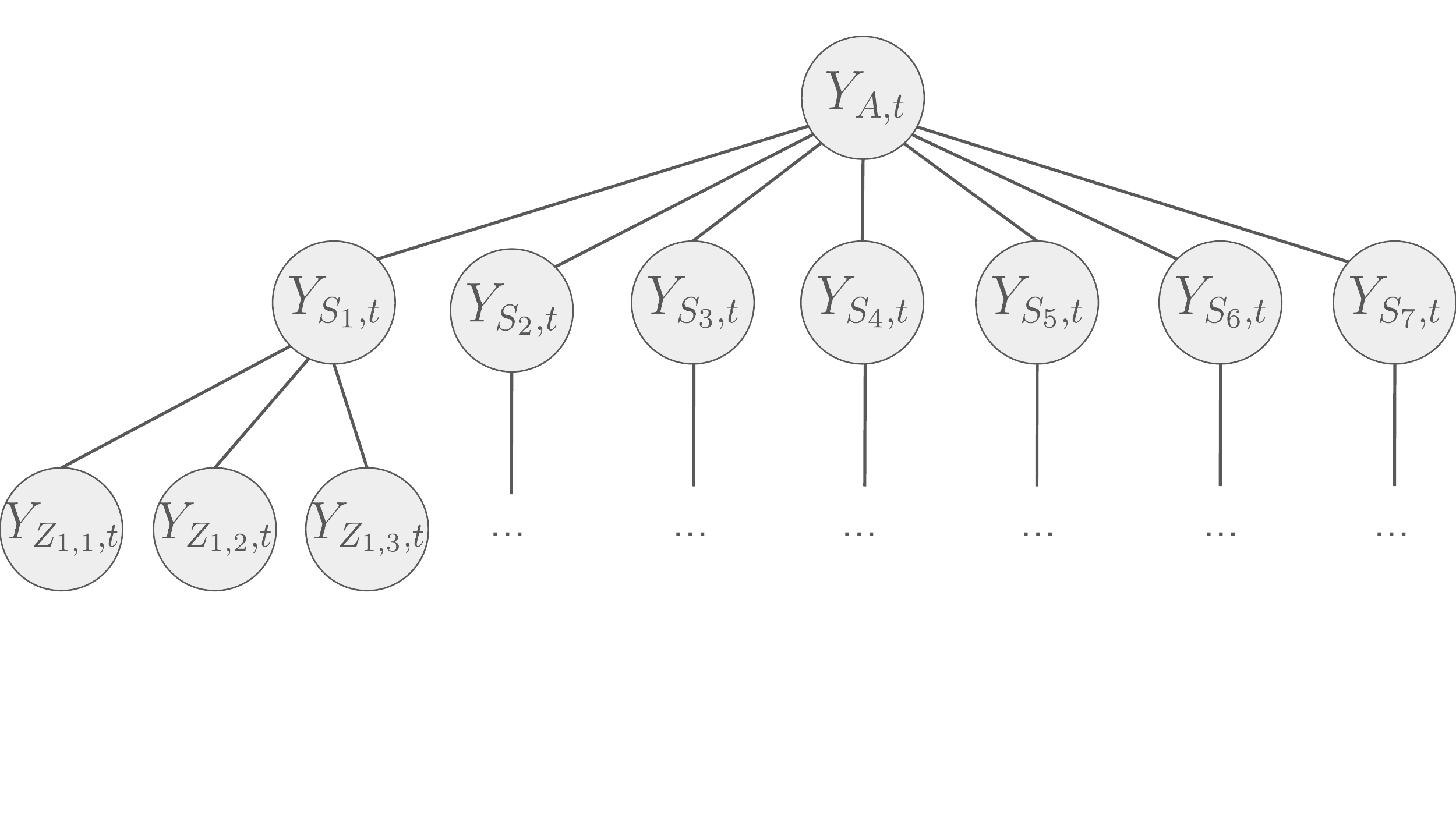}
    \caption{Graph representing the hierarchy of Australian domestic tourism.}
    \label{fig:DAC}
\end{figure}
Overall, the complete set of summation constraints can be represented by a directed acyclic graph, as shown in Figure~\ref{fig:DAC}. 
Alternatively, these constraints can be expressed by a $35 \times 27$ summation matrix $S$ that connects the bottom-level series $Y_b= ( Y_{Z_{1,1}},\hdots, Y_{Z_{7,z_7}})^\top \in \mathbb R^{27}$ to all hierarchical nodes $Y = (Y_{Z_{1,1}},\dots, Y_{Z_{7,z_7}}, Y_{S_1}, \hdots, Y_{S_7}, Y_A)^\top \in \mathbb R^{35}$ through the relation $Y = S Y_b$.
Thus, by letting $\mathbb 1 = (1, \hdots, 1)^\top \in \mathbb{R}^{27}$, and defining $\mathbb 1^{(j)}\in \mathbb R^{27}$ by $\mathbb 1^{(j)}_i = \left\{\begin{array}{cc}
     1 &   \hbox{ if } \sum_{k=1}^{j-1} z_k \leq i \leq \sum_{k=1}^j z_k\\
     0 &   \hbox{ otherwise}
\end{array}\right.$,  we have that $S = (\mathrm{I}_{27} \mid \mathbb 1^{(1)} \mid \cdots \mid \mathbb 1^{(7)} \mid  \mathbb 1)^\top$. 
The goal of hierarchical forecasting is to take advantage of the summation constraints defined by $S$ to improve the predictions of the vector $Y$ representing all hierarchical nodes.

This context can be easily generalized to many time series forecasting tasks. Spatial summation constraints, which divide a geographic space into different subspaces, have been applied in areas such as electricity demand forecasting \citep{bregere2022online}, electric vehicle charging demand forecasting \citep{amara-ouali2024forecasting}, and tourism forecasting \citep{Wickramasuriya2019optimal}.
Summation constraints also arise in multi-horizon forecasting, where, for example, an annual forecast must equal the sum of the corresponding monthly forecasts \citep{kourentzes2019cross}.
Finally, they also appear when goods are categorized into different groups \citep{pennings2017integrated}.

There are two main approaches to hierarchical forecasting. The first, known as forecast reconciliation, attempts to improve an existing estimator $\hat{Y}$ of the hierarchical nodes $Y$ by multiplying $\hat{Y}$ by a so-called reconciliation matrix $P$, so that the new estimator $P \hat Y$ satisfies the summation constraints. 
Formally, it is required that $\mathrm{Im}(P) \subseteq \mathrm{Im}(S)$, where $S$ is the summation matrix. 
The goal is for $P\hat{Y}$ to have less error than $\hat{Y}$. 
The strengths of this approach are its low computational cost and its ability to seamlessly integrate with pre-existing forecasts. 
Various reconciliation matrices, such as the orthogonal projection $P = S(S^\top S)^{-1}S$ on $\mathrm{Im}(S)$ (see the paragraph above on linear constraints), have been shown to reduce forecasting errors and to even be optimal under certain assumptions 
\citep{Wickramasuriya2019optimal}. Another complementary approach is to incorporate the hierarchical structure of the problem directly into the training of the initial estimator $\hat{Y}$ 
\citep{rangapuram21end}.
While this method is more computationally intensive, it provides a more comprehensive solution than reconciliation methods because it uses the hierarchy not only to shape the regression function, but also to inform the learning of its parameters. In this paper, we build on this approach to design three new estimators, all of which are implemented in Section~\ref{sec:tourism}.

As for now, we denote by $\ell_1$ the total number of nodes and $\ell_2 \leq \ell_1$ the number of bottom nodes. Thus, $Y=(Y_{\ell})_{1\leq \ell \leq \ell_1}^\top$ represents the global vector of all nodes, while $Y_b=(Y_{\ell})_{1\leq \ell \leq \ell_2}^\top$ represents the vector of the bottom nodes.
The $\ell_1 \times \ell_2$ summation matrix $S$ is defined so that, for all time index~$t$, the summation identity $Y_t = S Y_{b,t}$ is satisfied. 

\paragraph{Estimator 1. Bottom-up approach: WeaKL-BU.} In the bottom-up approach, models are fitted only for the bottom-level series $Y_b$, resulting in a vector of estimators $\hat{Y}_b$. The remaining levels are then estimated by $\hat{Y} = S \hat{Y}_b$, where $S$ is the summation matrix. 

To achieve this, forecasts for each bottom node $1 \leq \ell \leq \ell_2$ are constructed using a set of explanatory variables $X_\ell \in \mathbb R^{d_\ell}$ specific to that node. Together, these explanatory variables $X_1, \hdots, X_{\ell_2}$ form the feature $X\in \mathbb R^{d_1+\dots +d_{\ell_2}}$. A straightforward choice of features are the lags of the target variable, i.e., $X_{\ell, t} = Y_{\ell, t-1}$, though many other choices are possible. Next, for each bottom node $1 \leq \ell \leq \ell_2$, we fit a parametric model $f_{\theta_\ell}(X_{\ell, t})$ to predict the series $Y_{\ell, t}$. 
Each function $f_{\theta_\ell}$ is parameterized by a mapping $\phi_\ell$ (e.g., a Fourier map or an additive model) and a coefficient vector $\theta_\ell$, such that
$f_{\theta_\ell}(X_{\ell,t}) = \langle \phi_\ell(X_{\ell,t}), \theta_\ell \rangle$.
Therefore, the model for the lower nodes $Y_{b, t}$ can be expressed as $\mathbb \Phi_t \theta$, where  $\theta = (\theta_1, \hdots, \theta_{\ell_2})^\top$ is the vector of all coefficients, and $\mathbb{\Phi}_t$ is the feature matrix at time $t$ defined in \eqref{eq:feature_matrix}. 
Overall, the model for all levels $Y_t = S Y_{b, t}$ is $S \mathbb{\Phi}_t \theta$, and the empirical risk corresponding to this problem is given by
\[L(\theta) = \frac{1}{n}\sum_{j=1}^n\|\Lambda( S\mathbb \Phi_{t_j} \theta - Y_{t_j})\|_2^2 + \|M\theta\|_2^2,\]
where $\Lambda$ is a $\ell_1\times \ell_1$ diagonal matrix with positive coefficients,  and $M$ is a penalty matrix that depends on the $\phi_\ell$ mappings, as in Section~\ref{sec:shape}. 

Since $\Lambda$ scales the relative importance of each node in the learning process, the choice of its coefficients plays a critical role in the performance of the estimator. 
In the experimental Section~\ref{sec:tourism}, $\Lambda$ will be learned through hyperparameter tuning. Typically, $\Lambda_{\ell, \ell}$ should be large when $\mathrm{Var}(Y_\ell | X_\ell)$ is low---that is, the more reliable $Y_\ell$ is as a target \citep{Wickramasuriya2019optimal}. From~\eqref{eq:feature_matrix}, we deduce that the minimizer $\hat \theta$ of the empirical risk is 
    \begin{equation}
    \label{eq:pikl-bu}
        \hat \theta = \Big(\Big(\sum_{j=1}^n \mathbb \Phi_{t_j}^\ast S^\ast\Lambda^\ast \Lambda S\mathbb \Phi_{t_j}\Big) + n M^\ast M\Big)^{-1} \sum_{j=1}^n \mathbb \Phi_{t_j}^\ast \Lambda^\ast \Lambda Y_{t_j}.
    \end{equation}
We call $\hat{\theta}$ the {\it WeaKL-BU}. Setting $\Lambda = \mathrm{I}_{\ell_1}$, i.e., the identity matrix, results in treating all hierarchical levels equally, which is the setup of \citet{rangapuram21end}. On the other hand, setting $\Lambda_{\ell, \ell} = 0$ for all $\ell \geq \ell_2$ leads to learning each bottom node independently, without using any information from the hierarchy. 
This is the traditional bottom-up approach.

\paragraph{Estimator 2. Global hierarchy-informed approach: WeaKL-G.} 
The context is similar to the bottom-up approach, but here models are fitted for all nodes $1 \leq \ell \leq \ell_1$, using local explanatory variables $X_\ell \in \mathbb{R}^{d_\ell}$, where $d_\ell \geq 1$. Thus, the model for $Y_{t}$ is given by $\mathbb \Phi_t \theta$, where $\theta = (\theta_1, \hdots, \theta_{\ell_1})^\top$ is the vector of coefficients and $\mathbb{\Phi}_t$ is the feature matrix at time $t$ defined in \eqref{eq:feature_matrix}.
To ensure that the hierarchy is respected, we introduce a penalty term:
\[
\|\Gamma(S\Pi_b\mathbb \Phi_t\theta-\mathbb \Phi_t\theta)\|_2^2  = \|\Gamma(S\Pi_b-\mathrm{I}_{\ell_1})\mathbb \Phi_t\theta\|_2^2,
\]
where $\Gamma$ is a positive diagonal matrix and $\Pi_b$ is the projection operator on the bottom level, defined as $\Pi_b \theta = (\theta_1, \hdots, \theta_{\ell_2})^\top$. As in the bottom-up case, $\Gamma$ encodes the level of trust assigned to each node. 
In Section~\ref{sec:tourism}, we learn $\Gamma$ through hyperparameter tuning. This results in the empirical risk
\[L(\theta) = \frac{1}{n}\sum_{j=1}^n\|\mathbb \Phi_{t_j} \theta - Y_{t_j}\|_2^2 + \frac{1}{n}\sum_{j=1}^n\|\Gamma(S\Pi_b-\mathrm{I}_{\ell_1})\mathbb \Phi_{t_j}\theta\|_2^2 + \|M\theta\|_2^2.\]
where $M$ is a penalty matrix that depends on the $\phi_\ell$ mappings, as in Section~\ref{sec:shape}.
This empirical risk is similar to the one proposed by \citet{Zheng2023coherent}, where a penalty term is used to enforce hierarchical coherence during the learning process.
From \eqref{eq:feature_matrix}, we deduce that the  minimizer is given by 
    \begin{equation}
    \label{eq:pikl-G}
        \hat \theta = \Big(\sum_{j=1}^n (\mathbb \Phi_{t_j}^\ast\mathbb \Phi_{t_j}+ \mathbb \Phi_{t_j}^\ast(\Pi_b^\ast S^\ast-\mathrm{I}_{\ell_1})\Gamma^\ast \Gamma(S\Pi_b-\mathrm{I}_{\ell_1})\mathbb \Phi_{t_j})+nM^\ast M\Big)^{-1} \sum_{j=1}^n \mathbb \Phi_{t_j}^\ast Y_t.
    \end{equation}
We refer to $\hat{\theta}$  as the {\it WeaKL-G}.
The fundamental difference between \eqref{eq:pikl-bu} and \eqref{eq:pikl-G} is that the WeaKL-BU estimator only learns parameters for the $\ell_2$ bottom nodes, whereas the WeaKL-G estimators learns parameters for all nodes. We emphasize that WeaKL-BU and WeaKL-G follow different approaches. While WeaKL-BU adjusts the lower-level nodes and then uses the summation matrix $S$ to estimate the higher levels, WeaKL-G relies directly on global information, which is subsequently penalized by $S$. In the next paragraph, we complement the WeaKL-BU estimator by adding transfer learning constraints.

\paragraph{Estimator 3. Hierarchy-informed transfer learning: WeaKL-T.} In many hierarchical forecasting applications, the targets $Y_{\ell}$ are of the same nature throughout the hierarchy. Consequently, we often expect them to be explained by similar explanatory variables $X_{\ell}$ and to have similar regression functions estimators $f_{\hat{\theta}_\ell}$ \citep[e.g.,][]{leprince2023hierarchical}. For this reason, we propose an algorithm that combines WeaKL-BU with transfer learning.

Therefore, we assume that there is a subset $J \subseteq \{1,
\hdots, \ell_2\}$ of similar nodes and weights $(\alpha_i)_{i\in J}$ such that we expect $\alpha_i f_{\hat \theta_{i}}(X_{i,t}) \simeq \alpha_j f_{\hat \theta_{j}}(X_{j,t})$ for $i, j \in J$. 
In particular, there is an integer $D$ such that $\theta_j \in \mathbb{C}^{D}$ for all $j\in J$.
Therefore, denoting by $\Pi_J$ the projection on $J$ such that $\Pi_J\theta = (\theta_j)_{j\in J}\in \mathbb C^{D|J|}$, this translates into the constraint that $\Pi_J \theta \in \mathrm{Im}(M_J)$ where $M_J = (\alpha_1 \mathrm{I}_{D}, \hdots, \alpha_{|J|} \mathrm{I}_{D})^\top$.
As explained in the paragraph on linear constraints, we enforce this inexact constraint by penalizing the empirical risk with the addition of the term $\|(\mathrm{I}_{D|J|}-P_J)\Pi_J\theta\|_2^2$, where $P_J = M_J(M_J^\ast M_J)^{-1}M_J^\ast$ is the orthogonal projection onto the image of $M_J $.
This leads to the empirical risk
\[L(\theta) = \frac{1}{n}\sum_{j=1}^n\|\Lambda( S\mathbb \Phi_{t_j} \theta - Y_{t_j})\|_2^2+\lambda \|(\mathrm{I}_{D|J|}-P_{J})\Pi_{J}\theta\|_2^2 + \|M\theta\|_2^2,\]
where $M$ is a penalty matrix that depends on the $\phi_\ell$ mappings, as in Section~\ref{sec:shape}.
We call {\it WeaKL-T} the minimizer $\hat \theta$ of $L$. It is given by
    \begin{equation}
    \label{eq:pikl-T}
        \hat \theta = \Big(\Big(\sum_{j=1}^n \mathbb \Phi_{t_j}^\ast S^\ast\Lambda^\ast \Lambda S\mathbb \Phi_{t_j}\Big)+  n\lambda \Pi_{J}^\ast(\mathrm{I}_{D|J|}-P_{J})\Pi_{J}+nM^\ast M\Big)^{-1} \sum_{j=1}^n \mathbb \Phi_{t_j}^\ast\Lambda^\ast \Lambda Y_{t_j}.
    \end{equation}

\subsection*{Application to tourism forecasting}
\label{sec:tourism}
\paragraph{Hierarchical forecasting and tourism.}

In this experiment, we aim to forecast Australian domestic tourism using the dataset from \citet{Wickramasuriya2019optimal}. The dataset includes monthly measures of Australian domestic tourism from January 1998 to December 2016, resulting in $n = 216$ data points. 
Each month, domestic tourism is measured at four spatial levels and one categorical level, forming a five-level hierarchy. At the top level, tourism is measured for Australia as a whole. 
It is then broken down spatially into $7$ states, $27$ zones, and $76$ regions. 
Then, for each of the $76$ regions, four categories of tourism are distinguished  according to the purpose of travel: holiday, visiting friends and
relatives (VFR), business, and other. This gives a total of five levels (Australia, states, zones, regions, and categories), with $\ell_2 = 76 \times 4 = 304$ bottom nodes, and $\ell_1 = 1 + 7 + 27 + 76 + \ell_2 = 415$ total nodes.

\paragraph{Benchmark.} The goal is to forecast Australian domestic tourism one month in advance. Models are trained on the first $80\%$ of the dataset and evaluated on the last $20\%$. 
Similar to \citet{Wickramasuriya2019optimal}, we only consider autoregressive models with lags from one month to two years.
This setting is particularly interesting because, although each time series can be reasonably fitted using the $216$ data points, the total number of targets $\ell_1$ exceeds $n$. Consequently, the higher levels cannot be naively learned from the lags of the bottom level time series through linear regression.

The bottom-up (BU) model involves running $304$ linear regressions $\hat Y^{\mathrm{BU}}_{\ell,t} = \sum_{j=1}^{24}a_{\ell, j}Y_{\ell,t-j}$ for $1\leq \ell \leq \ell_2$, where $Y_{\ell,t-j}$ is the lag of $Y_{\ell, t}$ by $j$ months. 
The final forecast is then computed as $\hat Y^{\mathrm{BU}}_{t} = S\hat Y^{\mathrm{BU}}_{\ell,t}$,  where $S$ is the summation matrix.  
The Independent (Indep) model involves running separate linear regressions for each target time series using its own lags. This results in $415$ linear regressions of the form $\hat Y^{\mathrm{Indep}}_{\ell,t} = \sum_{j=1}^{24}a_{\ell, j}Y_{\ell,t-j}$ for $1\leq \ell \leq \ell_1$. 
Rec-OLS is the estimator resulting from OLS adjustment of the Indep estimator, i.e., taking $P = S(S^\ast S)^{-1}S$ \citep{Wickramasuriya2019optimal}. 
MinT refers to the estimator derived from the minimum trace adjustment of the Indep estimator \citep[see MinT(shrinkage) in][]{Wickramasuriya2019optimal}.
PIKL-BU refers to the estimator \eqref{eq:pikl-bu}, where, for all $1\leq \ell \leq 304$, $X_{\ell,t} = (Y_{\ell, t-j})_{1\leq j \leq 24}$ and $\phi_\ell(x) = x$.
PIKL-G is the estimator \eqref{eq:pikl-G}, where, for all $1\leq \ell \leq 415$, $X_{\ell,t} = (Y_{\ell, t-j})_{1\leq j \leq 24}$ and $\phi_\ell(x) = x$. 
Finally, PIKL-T is the estimator \eqref{eq:pikl-T}, where $X_{\ell,t} = (Y_{\ell, t-j})_{1\leq j \leq 24}$ and $\phi_\ell(x) = x$. 
In the latter model, all the auto-regressive effects are penalized to enforce uniform weights, which means that $\alpha_\ell = 1$ and $J = \{1, \hdots, \ell_2\}$ in \eqref{eq:pikl-T}.
The hyperparameter tuning process to learn the matrix $\Lambda$ for the WeaKLs is detailed in Appendix~\ref{sec:hierarchical_details}.

\paragraph{Results.} Table~\ref{table_australia} shows the results of the experiment. The mean square errors (MSE) are computed for each hierarchical level and aggregated under {\it All levels}. 
Their standard deviations are estimated using block bootstrap with blocks of length $12$.
The models are categorized based on the features they utilize.
We observe that the WeaKL-type estimators consistently outperform all other competitors in every case. This highlights the advantage of incorporating constraints to enforce the hierarchical structure of the problem, leading to an improved learning process.

\begin{table}[H]
\centering
\caption{Benchmark in forecasting Australian domestic tourism} 
\begin{tabular}{lcccccc}
\toprule
 &  & &$\;$\hfill  MSE  & ($\times 10^6$) \hfill $\;$&   &  \\
 \cmidrule{2-7}%
 & Australia & States & Zones & Regions & Categories & All levels \\
\midrule
\textit{Bottom data} &&&&&&\\
BU & $5.3\!\pm\!0.5$ & $2.0\!\pm\!0.2$ & $1.37\!\pm\!0.05$ & $\mathbf{1.19\!\pm\!0.02}$ & $\mathbf{1.17\!\pm\!0.03}$ & $11.0\!\pm\!0.7$ \\
WeaKL-BU & $\mathbf{4.5\!\pm\!0.5}$ & $\mathbf{1.9\!\pm\!0.3}$ & $\mathbf{1.34\!\pm\!0.05}$ & $\mathbf{1.19\!\pm\!0.03}$ & $\mathbf{1.17\!\pm\!0.03}$ & $\mathbf{10.1\!\pm\!0.6}$ \\
\midrule
\textit{Own lags} &&&&&&\\
Indep & $3.6\!\pm\!0.6$ & $1.8\!\pm\!0.2$ & $1.42\!\pm\!0.05$ & $1.23\!\pm\!0.03$ & $1.17\!\pm\!0.03$ & $9.2\!\pm\!0.7$ \\
WeaKL-G & $\mathbf{3.6\!\pm\!0.5}$ & $\mathbf{1.8\!\pm\!0.2}$ & $\mathbf{1.37\!\pm\!0.05}$ & $\mathbf{1.18\!\pm\!0.03}$ & $\mathbf{1.15\!\pm\!0.03}$ & $\mathbf{9.0\!\pm\!0.7}$ \\
\midrule
\textit{All data} &&&&&&\\
Rec-OLS & $3.5\!\pm\!0.5$ & $1.8\!\pm\!0.2$ & $1.35\!\pm\!0.05$ & $1.18\!\pm\!0.02$ & $1.17\!\pm\!0.03$ & $8.9\!\pm\!0.7$ \\
MinT & $3.6\!\pm\!0.4$ & $1.7\!\pm\!0.1$ & $1.29\!\pm\!0.05$ & $\mathbf{1.15\!\pm\!0.03}$ & $1.17\!\pm\!0.03$ & $8.9\!\pm\!0.5$ \\
WeaKL-T & $\mathbf{3.1\!\pm\!0.3}$ & $\mathbf{1.7\!\pm\!0.1}$ & $\mathbf{1.27\!\pm\!0.05}$ & $\mathbf{1.15\!\pm\!0.02}$ & $\mathbf{1.12\!\pm\!0.03}$ & $\mathbf{8.3\!\pm\!0.4}$ \\
\bottomrule
\end{tabular}
\label{table_australia}
\end{table}

\section{Conclusion}

In this paper, we have shown how to design empirical risk functions that integrate common linear constraints in time series forecasting. For modeling purposes, we distinguish between shape constraints (such as additive models, online adaptation after a break, and forecast combinations) and learning constraints (including transfer learning, hierarchical forecasting, and differential constraints). 
These empirical risks can be efficiently minimized on a GPU, leading to the development of an optimized algorithm, which we call WeaKL. 
We have applied WeaKL to three real-world use cases---two in electricity demand forecasting and one in tourism forecasting---where it consistently outperforms current state-of-the-art methods, demonstrating its effectiveness in structured forecasting problems.

Future research could explore the integration of additional constraints into the WeaKL framework. For example, the current approach does not allow for forcing the regression function $f_\theta$ to be non-decreasing or convex. However, since any risk function $L$ of the form \eqref{eq:risk} is convex in $\theta$, the problem can be formulated as a linearly constrained quadratic program. While this generally increases the complexity of the optimization, it can also lead to efficient algorithms for certain constraints. In particular, when $d=1$, imposing a non-decreasing constraint on $f_\theta$ reduces the problem to isotonic regression, which has a computational complexity of $O(n)$ \citep{wright1980isotonic}.

\setcounter{section}{0}
\renewcommand\thesection{\thechapter.\Alph{section}}

\section{Proofs}
The purpose of this appendix is to provide detailed proofs of the theoretical results presented in the main article. Appendix~\ref{proof:kernel} elaborates on the formula that characterizes the unique minimizer of the WeaKL empirical risks, while Appendix~\ref{sec:constraints} discusses the integration of linear constraints into the empirical risk framework.

\subsection*{A useful lemma}
\begin{lem}[Full rank]
    The matrix \[\tilde M = \frac{1}{n}\Big( \sum_{j=1}^n \mathbb \Phi_{t_j}^\ast \Lambda^\ast \Lambda\mathbb \Phi_{t_j}\Big) +  M^\ast M\] is invertible. Moreover, for all $\theta\in\mathbb C^{\dim \theta}$, $\theta^\star \tilde M \theta \geq \lambda_{\min}(\tilde M)\|\theta\|_2^2$, where $\lambda_{\min}(\tilde M)$ is the minimum eigenvalue of $\tilde M$.
    \label{lemma:full}
\end{lem}
\begin{proof}
    First, we note that $\tilde M$ is a positive Hermitian square matrix. Hence, the spectral theorem guarantees that $\tilde M$ is diagonalizable in an orthogonal basis of $\mathbb C^{\dim(\theta)}$ with real eigenvalues. In particular, it admits a positive square root, and the min-max theorem states that $\theta^\ast \tilde M \theta = \|\tilde M^{1/2} \theta\|_2^2 \geq \lambda_{\min}(\tilde M^{1/2})^2\|\theta\|_2^2 = \lambda_{\min}(\tilde M)\|\theta\|_2^2$. This shows the second statement of the lemma.
    
    Next, for all $\theta \in \mathbb C^{\dim \theta}$, $\theta^\ast \tilde M \theta \geq \theta^\ast  M^\ast M \theta$.
    Since $M$ is full rank,  $\mathrm{rank}(M) = \dim(\theta)$. Therefore, $\tilde M \theta = 0 \Rightarrow \theta^\ast \tilde M \theta = 0 \Rightarrow \theta^\ast M^\ast M \theta = 0  \Rightarrow \|M\theta\|_2^2 = 0 \Rightarrow M\theta = 0 \Rightarrow \theta = 0$. Thus, $\tilde M$ is injective and, in turn, invertible.
\end{proof}

\subsection*{Proof of Proposition~\ref{prop:emp_risk_min}}
\label{proof:kernel}
The function $L: \mathbb C^{\dim(\theta)} \to \mathbb R^+$ can be written as
\[L(\theta) = \frac{1}{n}\Big( \sum_{j=1}^n (\mathbb \Phi_{t_j}\theta- Y_{t_j})^\ast \Lambda^\ast \Lambda(\mathbb \Phi_{t_j}\theta- Y_{t_j})\Big)  + \theta^\ast M^\ast M\theta.\]
Recall that the matrices $\Lambda$ and $M$ are assumed to be injective. 
Observe that $L$ can be expanded as
\[L(\theta + \delta \theta) = L(\theta) + 2 \mathrm{Re}(\langle \tilde M\theta-\tilde Y, \delta \theta\rangle) + o(\|\delta \theta\|_2^2),\]
where $\tilde Y = \frac{1}{n} \sum_{j=1}^n \mathbb \Phi_{t_j}^\ast \Lambda^\ast \Lambda Y_{t_j}.$
This shows that $L$ is differentiable and that its differential at $\theta$ is the function $dL_\theta: \delta \theta \mapsto 2 \mathrm{Re}(\langle \tilde M\theta - \tilde Y, \delta \theta\rangle)$.
Thus, the critical points $\theta$ such that $dL_{\theta} = 0$ satisfy 
\[\forall\; \delta \theta \in \mathbb C^{\dim(\theta)}, \; \mathrm{Re}(\langle \tilde M\theta- \tilde Y, \delta \theta\rangle) = 0.\]
Taking $\delta \theta = \tilde M \theta- \tilde Y$ shows that $\|\tilde M\theta- \tilde Y\|_2^2 = 0$, i.e., $\tilde M\theta = \tilde Y$. From Lemma~\ref{lemma:full}, we deduce that $\theta = \tilde M^{-1} \tilde Y$, which is exactly the $\hat \theta_n$ in \eqref{eq:weakl}.

From Lemma~\ref{lemma:full}, we also deduce that, for all $\theta$ such that $\|\theta\|_2$ is large enough, one has $L(\theta) \geq \lambda_{\min}(\tilde M)\|\theta\|_2^2/2$. 
Since $L$ is continuous, it has at least one global minimum. Since the unique critical point of $L$ is $\hat \theta_n$, we conclude that $\hat \theta_n$ is the unique minimizer of $L$. 
\subsection*{Orthogonal projection and linear constraints}
\label{sec:constraints}
\begin{lem}[Orthogonal projection]
    \label{lem:ortho}
    Let $\ell_1, \ell_2 \in \mathbb N^\star$. Let $P$ be an injective $\ell_1 \times \ell_2$ matrix with coefficients in $\mathbb C$. Then
    $C = \mathrm{I}_{\ell_1} - P(P^\ast P)^{-1}P^\ast$ is the orthogonal projection on $\mathrm{Im}(P)^\perp$, where $\mathrm{Im}(P)$ is the image  of $P$ and $\mathrm{I}_{\ell_1}$ is the $\ell_1\times \ell_1$ identity matrix.
\end{lem}
\begin{proof}
First, we show that $P^\ast P$ is an $\ell_2 \times \ell_2$ matrix of full rank. Indeed, for all $x\in \mathbb C^{\ell_2}$, one has $P^\ast P x = 0 \Rightarrow x^\ast P^\ast P x = 0 \Rightarrow \|Px\|_2^2 = 0$. Since $P$ is injective, we deduce that $\|Px\|_2^2 = 0 \Rightarrow x = 0$. This means that $\ker P^\ast P = \{0\}$, and so that $P^\ast P$ is full rank. Therefore, $(P^\ast P)^{-1}$ is well defined.

Next, let $C_1 = P(P^\ast P)^{-1}P^\ast$. Clearly, $C_1^2 = C_1$, i.e., $C_1$ is a projector. Since $C_1^\ast = C_1$, we deduce that $C_1$ is an orthogonal projector.
 In addition, since $C_1 = P\times ((P^\ast P)^{-1}P^\ast)$, $\mathrm{Im}(C_1) \subseteq \mathrm{Im}(P)$. Moreover, if $x \in \mathrm{Im}(P)$, there exists a vector $ z$ such that $x = Pz$, and $C_1x = P(P^\ast P)^{-1}P^\ast Pz = Pz =x$. Thus, $x \in \mathrm{Im}(C_1)$. This shows that $\mathrm{Im}(C_1) = \mathrm{Im}(P)$. We conclude that $C_1$ is the orthogonal projection on $\mathrm{Im}(P)$ and, in turn, that $C = \mathrm{I}_{\ell_1} - C_1$ is the orthogonal projection on $\mathrm{Im}(P)^\perp$.
\end{proof}

The following proposition shows that, given the exact prior knowledge $C\theta^\star = 0$, enforcing the linear constraint  $C\theta = 0$ almost surely improves the performance of WeaKL.
\begin{prop}[Constrained estimators perform better.]
\label{prop:prop_lin}
    Assume that $Y_t = f_{\theta^\star}(X_t) + \varepsilon_t$ and that $\theta^\star$ satisfies the constraint $C\theta^\star = 0$, for some matrix $C$. (Note that we make no assumptions about the distribution of the time series $(X, \varepsilon)$.)
    Let $\Lambda$ and $M$ be injective matrices, and let $\lambda \geq 0$ be a hyperparameter. 
    Let $\hat \theta$ be the WeaKL given by \eqref{eq:weakl} and let $\hat \theta_C$ be the WeaKL obtained by replacing $M$ with $(\sqrt{\lambda}C^\top \mid M^\top)^\top$ in \eqref{eq:weakl}.
    Then, almost surely,
    \[\frac{1}{n}\sum_{j=1}^n\| f_{\theta^\star}(X_{t_j}) - f_{\hat \theta_C}(X_{t_j})\|_2^2 + \|M(\theta^\star- \hat \theta_C)\|_2^2\leq \frac{1}{n}\sum_{j=1}^n\| f_{\theta^\star}(X_{t_j}) - f_{\hat \theta}(X_{t_j})\|_2^2 + \|M(\theta^\star- \hat \theta)\|_2^2.\]
\end{prop}
\begin{proof} Recall from \eqref{eq:weakl} that
\[
    \hat \theta = P^{-1} \sum_{j=1}^n \mathbb \Phi_{t_j}^\ast \Lambda^\ast \Lambda Y_{t_j} \quad \mbox{and} \quad 
     \hat \theta_C = \big(P+ \lambda n C^\ast C\big)^{-1} \sum_{j=1}^n \mathbb \Phi_{t_j}^\ast \Lambda^\ast \Lambda Y_{t_j},
\]
where $P = ( \sum_{j=1}^n \mathbb \Phi_{t_j}^\ast \Lambda^\ast \Lambda\mathbb \Phi_{t_j}) + n M^\ast M$.
Since $C\theta^\star = 0$, we see that 
\begin{equation}
    \theta^\star = \big(P+ \lambda n C^\ast C\big)^{-1}P\theta^\star
    \label{eq:theta_star}.
\end{equation}
Subtracting \eqref{eq:theta_star} to, respectively, $\hat \theta$ and $\hat \theta_C$, we obtain
\[
     \theta^\star- \hat \theta = P^{-1/2} \Delta \quad \mbox{and}\quad 
    \theta^\star - \hat \theta_C = \big( P +\lambda n C^\ast C\big)^{-1} P^{1/2} \Delta,
\]
where \[\Delta = P^{-1/2}\Big(P\theta^\star- \sum_{j=1}^n \mathbb \Phi_{t_j}^\ast \Lambda^\ast \Lambda Y_{t_j}\Big).\] 
Moreover, according to the Loewner order \citep[see, e.g.,][Chapter~7.7]{Horn2012matrix}, we have that
$P^{-1/2}C^\ast C P^{-1/2} \geq 0$ and $(P^{-1/2}C^\ast C P^{-1/2})^2 \geq 0$. 
(Indeed, since $P$ is Hermitian, so is $P^{-1/2}C^\ast C P^{-1/2}$.)
Therefore, $(\mathrm{I} +\lambda n  P^{-1/2}C^\ast C P^{-1/2})^2 \geq \mathrm{I}$ and $( \mathrm{I} +\lambda n  P^{-1/2}C^\ast C P^{-1/2})^{-2} \leq \mathrm{I}$ \citep[see, e.g.,][Corollary~7.7.4]{Horn2012matrix}.
Consequently,
\[\|P^{1/2}(\theta^\star - \hat \theta_C)\|_2^2 = \Delta^\ast \big( \mathrm{I} +\lambda n  P^{-1/2}C^\ast C P^{-1/2}\big)^{-2} \Delta \leq \|\Delta\|_2^2 = \|P^{1/2}(\theta^\star - \hat \theta)\|_2^2.\]
Observing that $\|P^{1/2}(  \theta^\star-\hat \theta_C)\|_2^2 = \frac{1}{n}\sum_{j=1}^n\| f_{\theta^\star}(X_{t_j}) - f_{\hat \theta_C}(X_{t_j})\|_2^2 + \|M(\theta^\star- \hat \theta_C)\|_2^2$ and $\|P^{1/2}( \theta^\star- \hat \theta)\|_2^2 = \frac{1}{n}\sum_{j=1}^n\| f_{\theta^\star}(X_{t_j}) - f_{\hat \theta}(X_{t_j})\|_2^2 + \|M(\theta^\star- \hat \theta)\|_2^2$ concludes the proof.
\end{proof}
\begin{remark}
\label{rem:comment_prop_lin}
    Taking the limit $\lambda \to \infty$ in Proposition~\ref{prop:prop_lin} does not affect the result and corresponds to restricting the parameter space to $\ker(C)$, meaning that, in this case, $C \hat \theta_C = 0$.
    
Note also that the proposition is extremely general, as it holds almost surely without requiring any assumptions on either $X$ or $\varepsilon$.
Here, the error of $\hat \theta$ is measured by 
\[\frac{1}{n}\sum_{j=1}^n\| f_{\theta^\star}(X_{t_j}) - f_{ \hat \theta}(X_{t_j})\|_2^2 + \|M(\theta^\star- \hat \theta)\|_2^2,\]
which quantifies both the error of $\hat \theta$ at the points $X_{t_j}$ and in the $M$ norm.
Under additional assumptions on $X$ and $\varepsilon$, this discretized risk can be shown to converge to the $L^2$ error, $\mathbb E\| f_{\theta^\star}(X) - f_{ \hat \theta}(X)\|_2^2$, using Dudley’s theorem \citep[see, e.g., Theorem~5.2 in the Supplementary Material of][]{doumeche2023convergence}.

However, the rate of this convergence of $\hat \theta$ to $\theta^\star$ depends on the properties of $C$ and $M$, as well as the growth of $\dim(\theta)$ with $n$.
For instance, when the penalty matrix $M$ encodes a PDE prior, the analysis becomes particularly challenging and remains an open question in physics-informed machine learning.
Therefore, we leave the study of this convergence outside the scope of this article.
\end{remark}

\section{More WeaKL models}
\subsection*{Forecast combinations}
\label{sec:combination}
To forecast a time series $Y$, different models can be used, each using different implementations and sets of explanatory variables. Let $p$ be the number of models and let $\hat{Y}^1_t, \ldots, \hat{Y}^p_t$ be the respective estimators of $Y_t$.
The goal is to determine the optimal weighting of these forecasts, based on their performance evaluated over the time points $t_1 \leq  \cdots \leq t_n$. 
Therefore, in this setting, $X_t = (t, \hat{Y}^1_t, \ldots, \hat{Y}^p_t)$, and the goal is to find the optimal function linking $X_t$ to $Y_t$.
Note that, to avoid overfitting, we assume that the forecasts $\hat{Y}^1_t, \ldots, \hat{Y}^p_t$ were trained on time steps before~$t_1$.
This approach is sometimes referred to as the online aggregation of experts \citep{Remlinger2023expert, Antoniadis2024Aggregation}. Such forecast combinations are widely recognized to significantly improve the performance of the final forecast \citep{timmermann2006handbook, vilmarest2022state,petropoulos2022forecasting, amara-ouali2024forecasting}, as they leverage the strengths of the individual predictors. 

Formally, this results in the model
\[f_\theta(X_t) = \sum_{\ell=1}^p (p^{-1}+ h_{\theta_\ell}(t) )\hat Y^\ell_{t},\]
where $h_{\theta_\ell}(t) = \langle \phi(t), \theta_\ell\rangle$, $\phi$ is the Fourier map $\phi(t) =(\exp(i k t/2))_{-m\leq k \leq m}^\top$, and $\theta_\ell \in \mathbb{C}^{2m+1}$.
The $p^{-1}$ term introduces a bias, ensuring that $h_{\theta_\ell} = 0$ corresponds to a uniform weighting of the forecasts $\hat Y^\ell$.
The function $f^\star$ is thus estimated by minimizing the loss
    \[L(\theta) = \frac{1}{n}\sum_{j=1}^n \Big|\Big(\sum_{\ell=1}^p (p^{-1}+ h_{\theta_\ell}(t_j) )\hat Y^\ell_{t_j}\Big) - Y_{t_j}\Big|^2  + \sum_{\ell=1}^{p} \lambda_\ell \|h_{\theta_\ell}\|_{H^s}^2,\]
    where $\lambda_\ell > 0$ are hyperparameters.
Again, a common choice for the smoothing parameter is to set $s = 2$. 
Let $\phi_1(X_t) = 
    (
    (\hat Y^\ell_{t}\exp(ik t/2))_{- m\leq k \leq  m})_{\ell=1}^p)^\top \in \mathbb C^{(2m+1)p}$.
The Fourier coefficients that minimize the empirical risk are given by
\[
\hat \theta  = ({\mathbb{\Phi}} ^\ast {\mathbb{\Phi}} + n M^\ast  M)^{-1}{\mathbb \Phi}^\ast   \mathbb W,
\]
where $\mathbb W = (W_{t_1}, \hdots, W_{t_n})^\top$ is such that $W_t = Y_t - p^{-1}\sum_{\ell=1}^p \hat Y^\ell_t$,
\[M = \begin{pmatrix}
        \sqrt{\lambda_1} D& 0  & 0\\
        0 & \ddots&  0\\
        0 & 0& \sqrt{\lambda_{d_1}} D
    \end{pmatrix},\]
and $D$ is the $(2m+1)\times (2m+1)$ diagonal matrix
$D =\mathrm{Diag}((\sqrt{1+k^{2s}})_{-m\leq k\leq m})$. 

\subsection*{Differential constraints}
\label{sec:diff}
As discussed in the introduction, some time series obey physical laws and can be expressed as solutions of PDEs. Physics-informed kernel learning (PIKL) is a kernel-based method developed by \citet{doumèche2024physicsinformedkernellearning} to incorporate such PDEs as constraints. It can be regarded as a specific instance of the WeaKL framework proposed in this paper. In effect, given a bounded Lipschitz domain $\Omega$ and a linear differential operator $\mathscr D$, using the model $f_{ \theta}(x) = \langle \phi(x), \theta\rangle$, where $\phi(x) = (\exp(i  \langle x, k \rangle / 2) )_{\|k\|_\infty \leq m}$ is the Fourier map and $\theta$ represents the Fourier coefficients, the PIKL approach shows how to construct a matrix $M$ such that
\[
\int_\Omega \mathscr{D}(f_\theta, u)^2 \, du = \|M \theta\|_2^2.
\]
Thus, to incorporate the physical prior $\forall x \in \Omega,\; \mathscr D(f^\star, x) = 0$ into the learning process, the empirical risk takes the form
\[
L(\theta) = \frac{1}{n}\sum_{i=1}^n |f_\theta(X_{t_i}) - Y_{t_i}|^2 + \lambda \int_\Omega \mathscr{D}(f_\theta, u)^2 \, du =  \frac{1}{n}\sum_{i=1}^n |f_\theta(X_{t_i}) - Y_{t_i}|^2 + \|\sqrt{\lambda}M\theta\|_2^2,
\]
where $\lambda > 0 $ is a hyperparameter.
From \eqref{eq:weakl2} it follows that the minimizer of the empirical risk is
$\hat \theta = (\mathbb \Phi^\ast \mathbb \Phi+nM)^{-1} \mathbb\Phi^\ast \mathbb Y$. It is shown in \citet{doumeche2024physicsinformed} that, as $n \to \infty$, $f_{\hat{\theta}}$ converges to $f^\star$ under appropriate assumptions. Moreover, incorporating the differential constraint improves the learning process; in particular, $f_{\hat{\theta}}$ converges to $f^\star$ faster when $\lambda > 0$.

\section{A toy-example of hierarchical forecasting}
\label{sec:toy-example}
\paragraph{Setting.} We evaluate the performance of WeaKL on a simple but illustrative hierarchical forecasting task. In this simplified setting, we want to forecast two random variables, $Y_1$ and $Y_2$, defined as follows:
\[Y_1 = \langle X_1, \theta_1 \rangle + \varepsilon_1, \quad Y_2 = \langle X_2, \theta_2 \rangle - \varepsilon_1 + \varepsilon_2,
\]
where $X_1$, $X_2$, $\varepsilon_1$, and $\varepsilon_2$ are independent. The feature vectors are $X_1 \sim \mathcal{N}(0, \mathrm{I}_d)$ and $X_2 \sim \mathcal{N}(0, \mathrm{I}_d)$, with $d \in \mathbb N^\star$. The noise terms follow Gaussian distributions $\varepsilon_1 \sim \mathcal{N}(0, \sigma_1^2)$ and $\varepsilon_2 \sim \mathcal{N}(0, \sigma_2^2)$, with $\sigma_1, \sigma_2 > 0$.
Note that the independence assumption aims at simplifying the analysis in this toy-example by putting the emphasis on the impact of the hierarchical constraints rather than on the autocorrelation of the signal, though in practice this assumption is unrealistic for most time series. 
This is why we will develop a use case of hierarchical forecasting with real-world time series in Section~\ref{sec:tourism}.

What distinguishes this hierarchical prediction setting is the assumption that $\sigma_1 \geq \sigma_2$. 
Consequently, conditional on $X_1$ and $X_2$, the sum $Y_1 + Y_2= \langle  X_1, \theta_1 \rangle + \langle  X_2, \theta_2 \rangle + \varepsilon_2$ has a lower variance than either $Y_1$ or $Y_2$. 
We assume access to $n$ i.i.d.~copies $(X_{1,i}, X_{2,i}, Y_{1,i}, Y_{2,i})_{i=1}^n$ of the random variables $(X_1, X_2, Y_1, Y_2)$. 
The goal is to construct three estimators $\hat{Y}_1$, $\hat{Y}_2$, and $\hat{Y}_3$ of $Y_1$, $Y_2$, and $Y_3:=Y_1+Y_2$.

\paragraph{Benchmark.} We compare four techniques. The \textit{bottom-up (BU)} approach involves running two separate ordinary least squares (OLS) regressions that independently estimate $Y_1$ and $Y_2$ without using information about $Y_1 + Y_2$. Specifically,
\[
\hat{Y}_1^{\mathrm{BU}} = \langle X_1, \hat{\theta}_1^{\mathrm{BU}} \rangle, \quad \hat{Y}_2^{\mathrm{BU}} = \langle X_2, \hat{\theta}_2^{\mathrm{BU}} \rangle,
\]
where the OLS estimators are
\[
\hat{\theta}_1^{\mathrm{BU}} = (\mathbb{X}_1^\top \mathbb{X}_1)^{-1} \mathbb{X}_1^\top \mathbb{Y}_1, \quad \hat{\theta}_2^{\mathrm{BU}} = (\mathbb{X}_2^\top \mathbb{X}_2)^{-1} \mathbb{X}_2^\top \mathbb{Y}_2.
\]
Here, $\mathbb X_1 = (X_{1,1} \mid  \cdots \mid 
    X_{1,n})^\top$ and  $\mathbb X_2 = (X_{2,1}\mid \cdots \mid
    X_{2,n})^\top$ are $n \times d$ matrices, while  $\mathbb Y_1 = (Y_{1,1}, \hdots ,
    Y_{1,n})^\top$ and  $\mathbb Y_2 = (Y_{2,1}, \hdots ,
    Y_{2,n})^\top$ are vectors of $\mathbb R^n$.
To estimate $Y_3$, we simply set $\hat Y_3^{\mathrm{BU}}  = \hat Y_1^{\mathrm{BU}}  + \hat Y_2^{\mathrm{BU}} $.

The \textit{Reconciliation (Rec)} approach involves running three independent forecasts of $Y_1$, $Y_2$, and $Y_3$, followed by using the constraint that the updated estimator $\hat Y_3^{\mathrm{Rec}}$ should be the sum of $\hat Y_1^{\mathrm{Rec}}$ and $\hat Y_2^{\mathrm{Rec}}$ \citep{Wickramasuriya2019optimal}. To estimate $Y_3$, we run an OLS regression with 
$\mathbb X = (\mathbb X_1\mid  \mathbb X_2)$  and $\mathbb Y = \mathbb Y_1 +  \mathbb Y_2$. In this approach, 
\[
\begin{pmatrix}
    \hat Y_{3,t}^{\mathrm{Rec}}\\
    \hat Y_{1,t}^{\mathrm{Rec}}\\
    \hat Y_{2,t}^{\mathrm{Rec}}
\end{pmatrix} = S (S^T S)^{-1} S^T\begin{pmatrix}
    &\langle X_{t}, (\mathbb X^\top \mathbb X)^{-1}\mathbb X^\top \mathbb Y\rangle\\
    &\langle X_{1,t}, \hat \theta_1^{\mathrm{BU}}\rangle\\
    &\langle X_{2,t}, \hat \theta_2^{\mathrm{BU}}\rangle
\end{pmatrix},
\]
with $S = \begin{pmatrix}
    1 & 1 \\
    1 & 0 \\
    0 & 1
\end{pmatrix}$ and $X_t = (X_{1,t} \mid X_{2,t})$.

The \textit{Minimum Trace (MinT)} approach is an alternative update method that replaces the update matrix $S (S^\top S)^{-1} S^T$ with
$S (J-JWU(U^\top W U)^{-1}U^\top)$,
$J = \begin{pmatrix}
    0 & 1 & 0\\
    0 & 0 & 1 
\end{pmatrix}$,
$W$ the $3 \times 3$ covariance matrix of the prediction errors on the training data, and
$U = \begin{pmatrix}
    -1 &1 & 1
\end{pmatrix}^\top$ \citep{Wickramasuriya2019optimal}.
This approach extends the linear projection onto $\mathrm{Im}(S)$ and better accounts for correlations in the noise of the time series. 
Finally, we apply the WeaKL-BU estimator \eqref{eq:pikl-bu} with $M=0$, $\phi_1(x) = x$, $\phi_2(x) = x$, and $\Lambda = \mathrm{Diag}(1,1, \lambda)$, where $\lambda > 0$ is a hyperparameter that controls the penalty on the joint prediction $Y_{1} + Y_{2}$. 
It minimizes the empirical loss
\[
L(\theta_1, \theta_2) = \frac{1}{n}\sum_{i=1}^n |\langle X_{1,i}, \theta_1\rangle- Y_{1,i}|^2 + |\langle X_{2,i}, \theta_2\rangle-Y_{2,i}|^2 + \lambda |\langle X_{1,i}, \theta_1\rangle + \langle X_{1,i}, \theta_2\rangle- Y_{1,i}-Y_{2,i}|^2,
\]
In the experiments, we set $\lambda = \sigma_2^{-2}$ for simplicity, although it can be easily learned by cross-validation.
\begin{figure}
    \centering
    \includegraphics[width=0.4\linewidth]{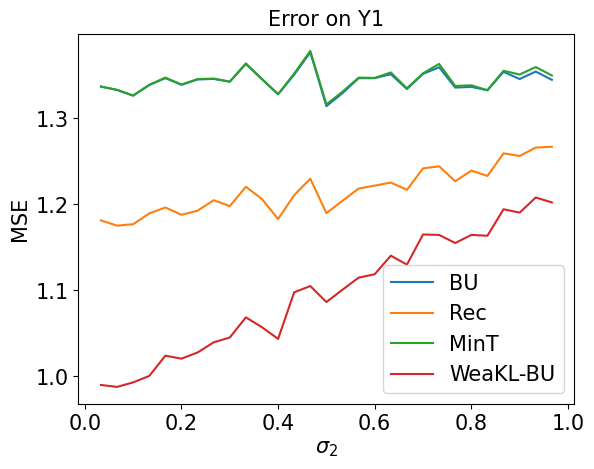}
    \includegraphics[width=0.4\linewidth]{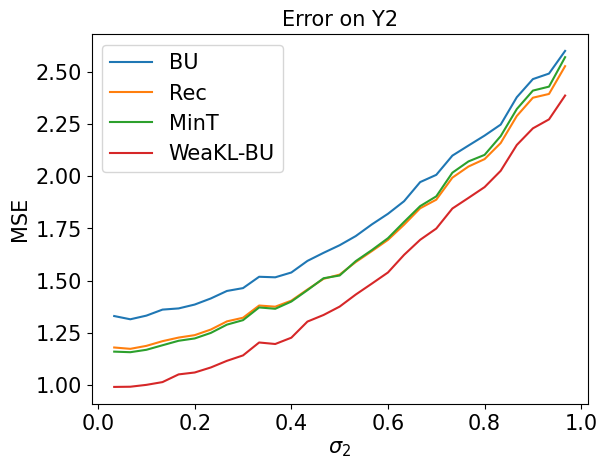}
    \includegraphics[width=0.4\linewidth]{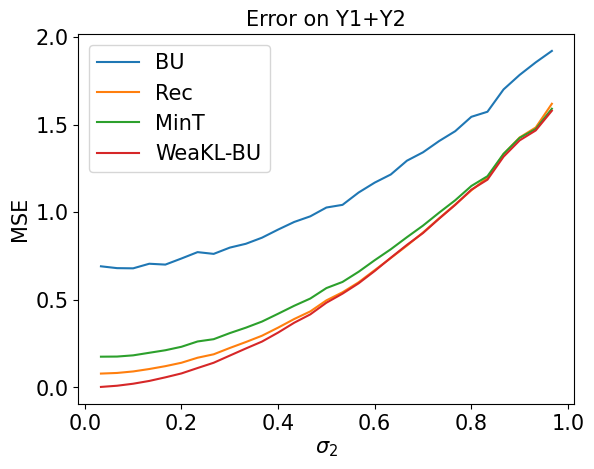}
    \includegraphics[width=0.4\linewidth]{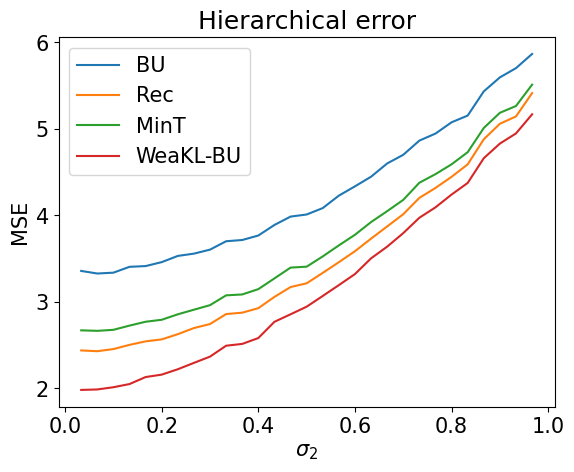}
    \caption{Hierarchical forecasting performance with $2d/n = 0.5$.}
    \label{fig:hier}
\end{figure}

\paragraph{Monte Carlo experiment.} To compare the performance of the different methods, we perform a Monte Carlo experiment. Since linear regression is invariant under multiplication by a constant, we set $\sigma_1=1$ without loss of generality. 
Since $\sigma_2 \leq \sigma_1$, we allow $\sigma_2$ to vary from $0$ to $1$. For each value of $\sigma_2$, we run $1000$ Monte Carlo simulations, where each simulation uses $n = 80$ training samples and $\ell = 20$ test samples. In each Monte Carlo run, we independently draw $\theta_1 \sim \mathcal N(0, I_d)$, $\theta_2 \sim \mathcal N(0, I_d)$, $X_{1,i} \sim \mathcal N(0, I_{ d})$, $X_{2,i} \sim \mathcal N(0, I_{d})$, $\varepsilon_{1,i} \sim \mathcal N(0, 1)$, and $\varepsilon_{2,i} \sim \mathcal N(0, \sigma_2^2)$, where $1 \leq i \leq n$. 
Note that, on the one hand, the $L^2$ error of an OLS regression on $Y_1 + Y_2$ is $\sigma_2^2 (1 + 2d/n)$, while on the other hand, the minimum possible $L^2$ error when fitting  $Y_1 + Y_2$ is $\sigma_2^2$.
Thus,  a large $2d/n$ is necessary to observe the benefits of hierarchical prediction. To achieve this, we set $d = 20$, resulting in $2d/n = 0.5$.

The models are trained on the $n$ training data points, and their performance is evaluated on the $\ell$ test data points using the mean squared error (MSE). Given any estimator $(\hat{Y}_1$, $\hat{Y}_2$, $\hat{Y}_3)$ of $(Y_1, Y_2, Y_1+Y_2)$, we compute the error $\ell^{-1}\sum_{j=1}^\ell| Y_{1, n+j}- \hat Y_{1, n+j}|^2$ on $Y_1$, the error $\ell^{-1}\sum_{j=1}^\ell| Y_{2, n+j}- \hat Y_{2, n+j}|^2$ on $Y_2$, and the error $\ell^{-1}\sum_{j=1}^\ell| Y_{1, n+j} + Y_{2, n+j}- \hat Y_{3, n+j}|^2$ on $Y_1 + Y_2$. 
The hierarchical error is defined as the sum of these three MSEs, which are visualized in Figure \ref{fig:hier}. 

\paragraph{Results.} Figure \ref{fig:hier} clearly shows that all hierarchical models (Rec, MinT, and WeaKL) outperform the naive bottom-up model for all four MSE metrics. Among them, our WeaKL consistently emerges as the best performing model, achieving superior results for all values of $\sigma_2$. Our WeaKL delivers gains ranging from $10\%$ to $50\%$ over the bottom-up model, solidifying its effectiveness in the benchmark.

The strong performance of WeaKL can be attributed to its approach, which goes beyond simply computing the best linear combination of linear experts to minimize the hierarchical loss, as reconciliation methods typically do. Instead, WeaKL directly optimizes the weights $\theta_1$ and $\theta_2$ to minimize the hierarchical loss.
Another way to interpret this is that when the initial forecasts are suboptimal, reconciliation methods aim to find a better combination of those forecasts, but do so without adjusting their underlying weights. In contrast, the WeaKL approach explicitly recalibrates these weights, resulting in a more accurate and adaptive hierarchical forecast.

\paragraph{Extension to the over-parameterized limit.} Another advantage of WeakL is that it also works for $d$ such that $2n \geq 2d \geq n$. 
In this context, the Rec and MinT algorithms cannot be computed because the OLS regression of $\mathbb Y$ on $\mathbb X$ is overparameterized ($2d$ features but only $n$ data points). 
To study the performance of the benchmark in the $n \simeq d$ limit, we repeated the same Monte Carlo experiment, but with $d = 38$, resulting in $d/n = 0.95$.
\begin{figure}
    \centering
    \includegraphics[width=0.4\linewidth]{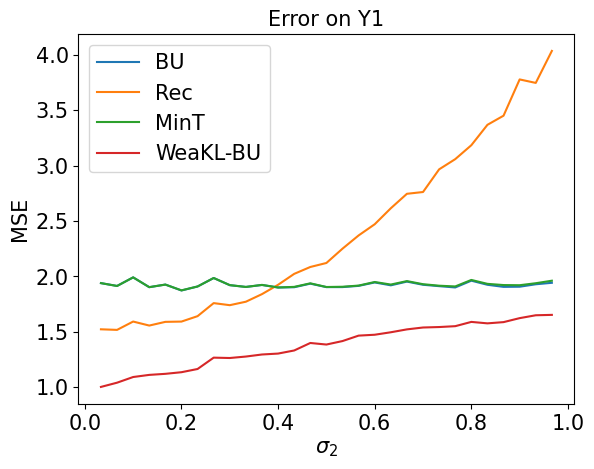}
    \includegraphics[width=0.4\linewidth]{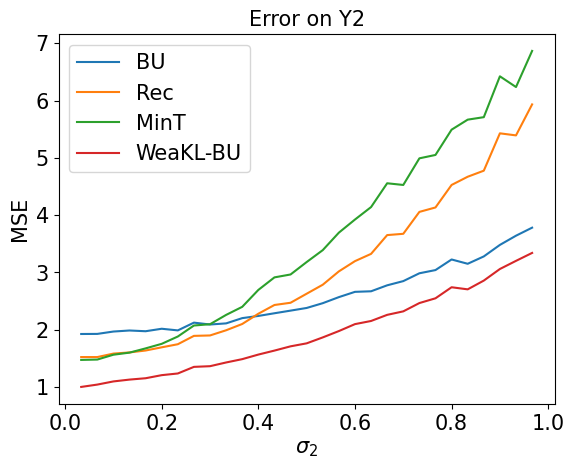}
    \includegraphics[width=0.4\linewidth]{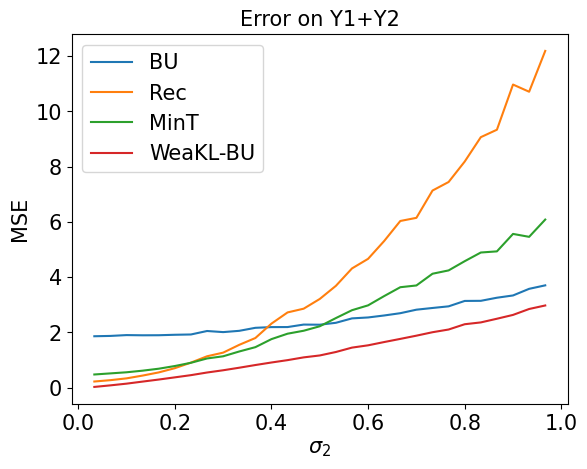}
    \includegraphics[width=0.4\linewidth]{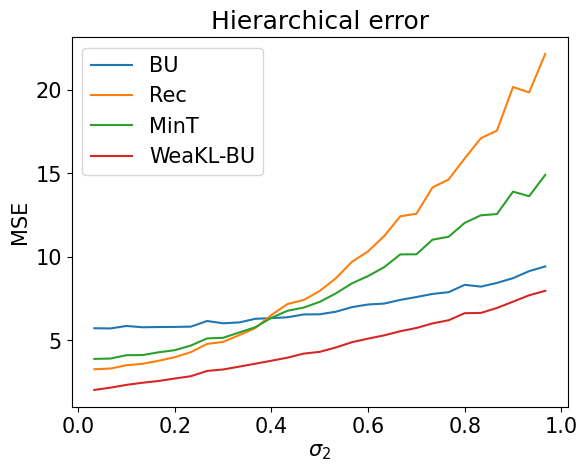}
    \caption{Hierarchical forecasting performance with $2d/n = 0.95$.}
    \label{fig:hier_2}
\end{figure}
The MSEs of the methods are shown in Figure~\ref{fig:hier_2}. These results further confirm the superiority of the WeaKL approach in the overparameterized regime. 
Note that such overparameterized situations are common in hierarchical forecasting.  
For example, forecasting an aggregate index-such as electricity demand, tourism, or food consumption-at the national level using city-level data across $d \gg 1$ cities (e.g., local temperatures) often leads to an overparameterized model.

\paragraph{Extension to non-linear regressions.} For simplicity, our experiments have focused on linear regressions. However, it is important to note that the hierarchical WeaKL can be applied to nonlinear regressions using exactly the same formulas. Specifically, in cases where $Y_1 = f_1(X_1) + \varepsilon_1$ and $Y_2 = f_2(X_2) - \varepsilon_1 + \varepsilon_2$, where $f_1$ and $f_2$ represent nonlinear functions, the WeaKL approach remains valid. This is because the connection to the linear case is straightforward: the WeaKL essentially performs a linear regression on the Fourier coefficients of $X_1$ and $X_2$, seamlessly extending its applicability to nonlinear settings.

\section{Experiments}
This appendix provides comprehensive details on the use cases discussed in the main text.
Appendix~\ref{sec:tuning} describes our hyperparameter tuning technique.
Appendix~\ref{sec:block-bootstrap} explains how we evaluate uncertainties.
Appendix~\ref{sec:half-hour} outlines our approach to handling sampling frequency in electricity demand forecasting applications.
Appendix~\ref{sec:case_study1} details the models used in Use case 1, while Appendix~\ref{sec:sobriety} focuses on Use case $2$, and Appendix~\ref{sec:hierarchical_details} covers the tourism demand forecasting use case.

\subsection*{Hyperparameter tuning}
\label{sec:tuning}
\paragraph{Hyperparameter tuning of the additive WeaKL.} Consider a WeaKL additive model
\begin{align*}
    f_{\theta}(X_t) &=  \langle \phi_{1,1}(X_{1,t}), \theta_{1,1}\rangle + \cdots + \langle \phi_{1,d_1}(X_{d_1,t}), \theta_{1,d_1}\rangle,
\end{align*}
where the type (linear, nonlinear, or categorical) of the effects are specified. Thus, as detailed in Section~\ref{sec:shape},
\begin{itemize}
    \item[$(i)$] If the effect $\langle \phi_{1,\ell}(X_{\ell,t}), \theta_{1,j}\rangle$ is assumed to be linear, then $\phi_{1,j}(X_{\ell,t}) = X_{\ell,t}$,
    \item[$(ii)$] If the effect $\langle \phi_{1,\ell}(X_{\ell,t}), \theta_{1,\ell}\rangle$ is assumed to be  nonlinear, then $\phi_{1,\ell}$ is a Fourier map with $2m_\ell +1$ Fourier modes,
    \item[$(iii)$] If the effect $\langle \phi_{1,\ell}(X_{\ell,t}), \theta_{1,\ell}\rangle$ is assumed to be categorical with values in $E$, then $\phi_{1,\ell}$ is a Fourier map with $2\lfloor |E|/2\rfloor+1$ Fourier modes.
\end{itemize}
We let $\mathbf{m} = \{m_\ell\mid \hbox{the effect $\langle \phi_{1,\ell}(X_{\ell,t}), \theta_{1,\ell}\rangle$ is nonlinear}\}$ be the concatenation of the numbers of Fourier modes of the nonlinear effects.
The goal of hyperparameter tuning is to find the best set of hyperparameters $\lambda = (\lambda_1, \hdots, \lambda_{d_1})$ and 
$\mathbf{m}$ for the empirical risk \eqref{eq:weaklGAM} of the additive WeaKL.

To do so, we split the data into three sets: a training set, then a validation set, and finally a test set.
These three sets must be disjoint to avoid overfitting, and the test set is the dataset on which the final performance of the method will be evaluated.
The sets should be chosen so that the distribution of $(X,Y)$ on the validation set resembles as much as possible the distribution of $(X,Y)$ on the test set.

We consider a list of potential candidates for the optimal set of hyperparameters $(\lambda ,\mathbf{m})_{\mathrm{opt}}$. 
Since we have no prior knowledge about $(\lambda, \mathbf{m})$, we chose this list to be a grid of parameters. For each element $(\lambda, \mathbf{m})$ in the grid, we compute the minimizer $\hat \theta(\lambda, \mathbf{m})$ of the loss \eqref{eq:weaklGAM} over the training period. Then, given $\hat \theta(\lambda, \mathbf{m})$, we compute the mean squared error (MSE) of $f_{\hat \theta(\lambda, \mathbf{m})}$ over the validation period. This procedure is commonly referred to as grid search. The resulting estimate of the optimal hyperparameters $(\lambda, \mathbf{m})_{\mathrm{opt}}$ corresponds to the values of  $(\lambda, \mathbf{m})$ that minimize the MSE of  $f_{\hat \theta(\lambda, \mathbf{m})}$ over the validation period. The performance of the additive WeaKL is then assessed based on the performance of $f_{\hat \theta(\lambda, \mathbf{m})_{\mathrm{opt}}}$ on the test set.
\paragraph{Hyperparameter tuning of the online WeaKL.} Consider an online WeaKL
\begin{equation*}
    f_\theta(t, x_1, \hdots, x_{d_1}) = h_{\theta_0}(t)+ \sum_{\ell=1}^{d_1} (1+ h_{\theta_\ell}(t))  \hat g_\ell(x_\ell),
\end{equation*}
where the effects $\hat g_\ell$ are known, and the updates $h_{\theta_\ell}(t) = \langle \phi(t), \theta_\ell\rangle$ are such that $\phi$ is the Fourier map $\phi(t) =(\exp(i k t/2))_{-m_j\leq k \leq m_j}^\top$, with $m_j \in \mathbb N^\star$.
We let $\mathbf{m} = \{m_j\mid 0\leq j \leq d_1\}$ be the concatenation of the numbers of Fourier modes.
The goal of hyperparameter tuning is to find the best set of hyperparameters $\lambda = (\lambda_0, \hdots, \lambda_{d_1})$ and 
$\mathbf{m}$ for the empirical risk \eqref{eq:risk_online} of the online WeaKL.

To do so, we split the data into three sets: a training set, then a validation set, and finally a test set.
This three sets must be disjoint to avoid overfitting.
Moreover, the training set and the validation set must be disjoint from the data used to learn the effects $\hat g_\ell$.
The test set must be the set on which the final performance of the method will be evaluated. 
The sets should be chosen so that the distribution of $(X,Y)$ on the validation set resembles as much as possible the distribution of $(X,Y)$ on the test set.
Similarly to the hyperparameter tuning of the additive WeaKL, we then consider a list of potential candidates for the optimal hyperparameter $(\lambda ,\mathbf{m})_{\mathrm{opt}}$, which can be a grid.
Then, we compute the minimizer $\hat \theta(\lambda, \mathbf{m})$ of the loss \eqref{eq:risk_online} on the training period, and the resulting estimation of $(\lambda, \mathbf{m})_{\mathrm{opt}}$ is the set of hyperparameters $(\lambda, \mathbf{m})$ such that the MSE of $f_{\hat \theta(\lambda, \mathbf{m})}$ on the validation period is minimal.
The performance of the online WeaKL is thus measured by the performance of $f_{\hat \theta(\lambda, \mathbf{m})_{\mathrm{opt}}}$ on the test set.

\subsection*{Block bootstrap methods}
\label{sec:block-bootstrap}
\paragraph{Evaluating uncertainties with block bootstrap.}
The purpose of this paragraph is to provide theoretical tools for evaluating the performance of time series estimators.
Formally, given a test period $\{t_1, \hdots, t_n\}$, a  target time series $(Y_{t_j})_{1\leq j \leq n}$, and an estimator $(\hat Y_{t_j})_{1\leq j \leq n}$ of $Y$, the goal is to construct confidence intervals that quantify how far $\mathrm{RMSE}_{n} = (n^{-1}\sum_{j=1}^n |\hat Y_{t_j} - Y_{t_j}|^2)^{1/2}$ deviates from its expectation 
$\mathrm{RMSE} = (\mathbb{E} |\hat Y_{t_1} - Y_{t_1}|^2)^{1/2}$, and  how far  $\mathrm{MAPE}_n = n^{-1}\sum_{j=1}^n |\hat Y_{t_j} - Y_{t_j}| |Y_{t_j}|^{-1}$ deviates from its expectation $\mathrm{MAPE} = \mathbb{E}( |\hat Y_{t_1} - Y_{t_1}| |Y_{t_1}|^{-1})$.
Here, we assume that $Y$ and $\hat{Y}$ are strongly stationary, meaning their distributions remain constant over time.
Constructing such confidence intervals is non-trivial because the observations $Y_{t_j}$ in the time series $Y$ are correlated, preventing the direct application of the central limit theorem.
The block bootstrap algorithm is specifically designed to address this challenge and is defined as follows.

Consider a sequence $Z_{t_1},Z_{t_2},\hdots,Z_{t_n}$ such that the quantity of interest can be expressed as $g(\mathbb{E}(Z_{t_1}))$, for some function $g$. 
This quantity is estimated by $g(\bar Z_n)$, where $\bar Z_n = n^{-1}\sum_{j=1}^n Z_{t_j}$ is the empirical mean of the sequence.
For example, $\mathrm{RMSE} = g(\mathbb E(Z_{t_1}))$ and $\mathrm{RMSE}_n = g(\bar Z_n)$ for $g(x) = x^{1/2}$ and $Z_{t_j}=(Y_{t_j}-\hat{Y_{t_j}})^2$, while $\mathrm{MAPE} = g(\mathbb E(Z_{t_1}))$ and  $\mathrm{MAPE}_n = g(\bar Z_n)$  for $g(x) = x$ and $Z_{t_j}=|\hat Y_{t_j} - Y_{t_j}| |Y_{t_j}|^{-1}$.
The goal of the block bootstrap algorithm is to estimate the distribution of $g(\bar Z_n)$.

Given a length $\ell \in \mathbb N^\star$ and a starting time $t_j$, we say that $(Z_{t_j}, \hdots, Z_{t_{j+\ell-1}})\in \mathbb R^\ell$ is a block of length $\ell$ starting at $t_j$. We draw  $b = \lfloor n/\ell\rfloor +1$ blocks of length $\ell$ uniformly from the sequence  $(Z_{t_1}, Z_{t_2}, \dots, Z_{t_n})$ and then concatenate these blocks to form the sequence $Z^\ast = (Z_1^\ast, Z_2^\ast, \dots, Z_{b\ell}^\ast)$.
Thus, $Z^\ast$ is a resampled version of $Z$  obtained with replacement.

For convenience, we consider only the first $n$ values of $Z^\ast$ and compute the bootstrap version of the empirical mean: $\bar{Z}^\ast_n=\frac{1}{n}\sum_{j=1}^nZ^\ast_j$. By repeatedly resampling the $b$ blocks and generating multiple instances of $\bar{Z}^\ast_n$, the resulting distribution of $\bar{Z}^\ast_n$  provides a reliable estimate of the distribution of $\bar{Z}_n$.
In particular, under general assumptions about the decay of the autocovariance function of $Z$, choosing $\ell = \lfloor n^{1/4} \rfloor$ leads to 
\[\sup_{x\in\mathbb{R}}|\mathbb{P}(T^{\ast}_n\leq x\mid Z_{t_1},\hdots,Z_{t_n})-\mathbb{P}(T_n\leq x)| = O_{n\to \infty}(n^{-3/4}),\]  where $T^{\ast}_n=\sqrt n(\bar{Z}^\ast_n-\mathbb{E}(\bar{Z}^\ast_n\mid Z_{t_1},\hdots,Z_{t_n}))$ and $T_n=\sqrt n(\bar{Z}_n-\mathbb{E}(Z_{t_1}))$ \citep[see, e.g.][Theorem 6.7]{lahiri2023resampling}.
Note that this convergence rate of $n^{-3/4}$ is actually quite fast, as even if the $Z_{t_j}$  were i.i.d., the empirical mean $\bar{Z}_n$  would only converge to a normal distribution at a rate of $n^{-1/2}$
(by the Berry-Esseen theorem). This implies that the block bootstrap method estimates the distribution of $\bar{Z}_n$ faster than $\bar{Z}_n$  itself converges to its Gaussian limit.

The choice of $\ell$ plays a crucial role in this method.
For instance, setting $\ell = 1$ leads to an underestimation of the variance of $\bar{Z}_n$  when the $Z_{t_j}$ are correlated \citep[see, e.g.][Corollary 2.1]{lahiri2023resampling}.
In addition, block resampling introduces a bias, as $Z_{t_n}$ belongs to only a single block and is therefore less likely to be resampled than  $Z_{t_{\lfloor n/2\rfloor}}$.
This explains why $\mathbb{E}(\bar{Z}^\ast_n \mid Z_{t_1}, \dots, Z_{t_n}) \neq \bar{Z}_n$.
To address both problems, \citet{politis1994stationary} introduced the stationary bootstrap, where the block length $\ell$ varies and follows a geometric distribution.

\paragraph{Comparing estimators with block bootstrap.}
Given two stationary estimators $\hat Y^1$ and $\hat Y^2$ of $Y$,  the goal is to develop a test of level $\alpha \in [0,1]$ for the hypothesis $H_0: \mathbb E|\hat Y^1_t-Y_t| = \mathbb E|\hat Y^2_t-Y_t|$. Using the previous paragraph, such a test could be implemented by estimating two confidence intervals $I_1$ and $I_2$ for $\mathbb E|\hat Y^1_t-Y_t|$ and $\mathbb E|\hat Y^2_t-Y_t|$ at level $\alpha/2$ using block bootstrap, and then rejecting $H_0$  if  $I_1 \cap I_2 = \emptyset$. However, this approach tends to be conservative, potentially reducing the power of the test when assessing whether one estimator is significantly better than the other.  

To create a more powerful test, \citet{Messner2020evaluation} and \citet{Farrokhabadi2022day} suggest relying on the MAE skill score, which is defined by \[\mathrm{Skill}=1-\frac{\mathrm{MAE_{1}}}{\mathrm{MAE_{2}}},\]
where $\mathrm{MAE_{1}}$ and $\mathrm{MAE_{2}}$ are the mean average errors of $\hat{Y}^{1}$ and $\hat{Y}^2$, respectively. 
Note that $\mathrm{Skill}= (\mathrm{MAE}_{2} - 
\mathrm{MAE_{1}})/\mathrm{MAE_{2}}$
is the relative distance between the two $\mathrm{MAE}$s. Thus, $\hat{Y}^{1}$ is significantly better than $\hat{Y}^2$ if $\mathrm{Skill}$ is significantly positive.
A confidence interval for $\mathrm{Skill}$ can be obtained  by block bootstrap.
Indeed, consider the time series $Z$ defined as $Z_{t_j}=(|\hat{Y}^{1}_{t_j}-Y^1_{t_j}|,|\hat{Y}^{2}_{t_j}-Y^2_{t_j}|)$, and let $g(x,y)=1-x/y$. We use the block bootstrap method over this sequence to estimate $g(\mathbb E(Z))$ by generating different samples of $\mathrm{MAE_{1}}$ and $\mathrm{MAE_{2}}$. In particular, in Appendix~\ref{sec:case_study1}, $\hat{Y}^{1}$ corresponds to WeakL, while $\hat{Y}^{2}$ is the estimator of the winning team of the IEEE competition.

\subsection*{Half-hour frequency}
\label{sec:half-hour}

Short-term electricity demand forecasts are often estimated with a half-hour frequency, meaning that the objective is to predict electricity demand every $30$ minutes during the test period. This applies to both Use case 1 and Use case 2.
There are two common approaches to handling this frequency in forecasting models. One approach is to include the half-hour of the day as a feature in the models. The alternative, which yields better performance, is to train a separate model for each half-hour, resulting in $48$ distinct models. This superiority arises because the relationship between electricity demand and conventional features (such as temperature and calendar effects) varies significantly across different times of the day. For instance, electricity demand remains stable at night but fluctuates considerably during the day. This variability justifies treating the forecasting problem at each half-hour as an independent learning task, leading to $48$ separate models. Consequently, in both use cases, all models discussed in this paper---including WeaKL, as well as those from \citet{vilmarest2022state} and \citet{doumeche2023human}---are trained separately for each of the $48$ half-hours, using identical formulas and architectures. This results in $48$ distinct sets of model weights. 
For simplicity, and since the only consequence of this preprocessing step is to split the learning data into $48$ independent groups, this distinction is omitted from the equations.

\subsection*{Precisions on the Use case 1 on the IEEE DataPort Competition on Day-Ahead Electricity Load Forecasting}
\label{sec:case_study1}
In this appendix, we provide additional details on the two WeaKLs used in the benchmark for Use case 1 of the IEEE DataPort Competition on Day-Ahead Electricity Load Forecasting. The first model is a direct adaptation of the GAM-Viking model from \citet{vilmarest2022state} into the WeaKL framework. The second model is a WeaKL where the effects are learned through hyperparameter tuning.
\paragraph{Direct translation of the GAM-Viking model into the WeaKL framework.}

To build their model, \citet{vilmarest2022state} consider four primary models: an autoregressive model (AR), a linear regression model, a GAM, and a multi-layer perceptron (MLP). 
These models are initially trained on data from $18$ March $2017$ to $1$ January $2020$. 
Their weights are then updated using the Viking algorithm starting from $1$ March $2020$ \citep[][Table~3]{vilmarest2022state}. 
The parameters of the Viking algorithm were manually selected by the authors based on performance monitoring over the $2020$–$2021$ period \citep[][Figure7]{vilmarest2022state}. 
To further refine the forecasts, the model errors are corrected using an autoregressive model, which they called the intraday correction and implemented as a static Kalman filter. The final forecast is obtained by online aggregation of all models, meaning that the predictions from different models are combined in a linear combination that evolves over time. The weights of this aggregation are learned using the ML-Poly algorithm from the \texttt{opera} package \citep{gaillard2016opera}, trained over the period $1$ July $2020$ to $18$ January $2021$. 
The test period spans from $18$ January $2021$ to $17$ February $2021$. 
During this period, the aggregated model achieves a MAE of $10.9$~GW, while the Viking-updated GAM model alone yields an MAE of $12.7$~GW.

Here, to ensure a fair comparison between our WeaKL framework and the GAM-Viking model of \citet{vilmarest2022state}, we replace their GAM-Viking with our online WeaKL in their aggregation. Our additive WeaKL model is therefore a direct translation of their offline GAM formulation into the WeaKL framework. Specifically, we consider the additive WeaKL based on the features $X = (\mathrm{DoW}, \mathrm{FTemps95_{corr1}}, \mathrm{Load_1}, \mathrm{Load_7}, \mathrm{ToY}, t)$ corresponding to
\begin{equation*}
\begin{split} 
Y_t =& g_1^\star(\mathrm{DoW}_t)
   +g_2^\star(\mathrm{FTemps95_{corr1}}_t)
   + g_3^\star(\mathrm{Load_1}_t) +g_4^\star(\mathrm{Load_7}_t)+g_5^\star(\mathrm{ToY}_t)
      +g_6^\star(t) +\varepsilon_t,
\end{split}
\end{equation*}
where $g_1^\star$ is categorical with 7 values, $g_2^\star$ and $g_6^\star$  are linear,  $g_3^\star$, $g_4^\star$, and $g_5^\star$ are nonlinear.

$\mathrm{FTemps95_{corr1}}$ is a smoothed version of the temperature, while the other features remain the same as those used in Use case 2. The weights of the additive WeaKL model are determined using the hyperparameter selection technique described in Appendix~\ref{sec:tuning}. 
The training period spans from $18$ March $2017$ to $1$ November $2019$, while the validation period extends from $1$ November $2019$ to $1$ January $2020$. 
During this grid search, the performance of $250,047$ sets of hyperparameters $(\lambda, \mathbf m)\in \mathbb R^7\times \mathbb R^3$ is evaluated in less than a minute using a standard GPU (Nvidia L4 GPU, $24$~GB RAM, $30.3$ teraFLOPs for Float32). 
Notably, this optimization period exactly matches the training period of the primary models in \citet{vilmarest2022state}, ensuring a fair comparison between the two approaches.

Then, we run an online WeaKL, where the effects $\hat g_\ell$, $1\leq \ell \leq 7$, are inherited directly from the previously trained additive WeaKL. 
The weights of this online WeaKL are determined using the hyperparameter selection technique described in Appendix~\ref{sec:tuning}. 
The training period extends from $1$ February $2020$ to $18$ November $2020$, while the validation period extends from $18$ November $2020$ to $18$ January $2021$, immediately preceding the final test period to ensure optimal adaptation. During this grid search, we evaluate $625$ sets of hyperparameters $(\lambda, \mathbf m)\in \mathbb R^6\times \mathbb R^6$ in less than a minute using a standard GPU. 
Since $t$ is already included as a feature, the function  $h_0^\ast$ in Equation~\eqref{eq:model} is not required in this setting.

Finally, we evaluate the performance of our additive WeaKL (denoted as $\hbox{WeaKL}_{\mathrm{+}}$), our additive WeaKL followed by intraday correction ($\hbox{WeaKL}_{+,\mathrm{intra}}$), our online WeaKL ($\hbox{WeaKL}_{\mathrm{on}}$), our online WeaKL with intraday correction ($\hbox{WeaKL}_{\mathrm{on, intra}}$), and an aggregated model based on \citet{vilmarest2022state}, where the GAM and GAM-Viking models are replaced by our additive and online WeaKL models ($\hbox{WeaKL}_{\mathrm{agg}}$). The test period remains consistent with \citet{vilmarest2022state}, spanning from $18$ January $2021$ to $17$ February $2021$. Their performance results are presented in Table~\ref{tab:gam-vik} and compared to their corresponding translations within the GAM-Viking framework.
\begin{table}
    \centering
    \caption{Comparing GAM-Viking with its direct translation in the WeaKL framework on the final test period}
    \begin{tabular}{|c|ccccc|}
        \hline
        Model GAM &  $\hbox{GAM}_{\mathrm{+}}$ & $\hbox{GAM}_{+,\mathrm{intra}}$ & $\hbox{GAM}_{\mathrm{on}}$ & $\hbox{GAM}_{\mathrm{on, intra}}$ & $\hbox{GAM}_{\mathrm{agg}}$\\
        \hline
         MAE (GW) & 48.3 & 22.7 & 13.2 & 12.7 & 10.9\\
         \hline
         \hline
        Model WeaKL &  $\hbox{WeaKL}_{\mathrm{+}}$ & $\hbox{WeaKL}_{+,\mathrm{intra}}$ & $\hbox{WeaKL}_{\mathrm{on}}$ & $\hbox{WeaKL}_{\mathrm{on, intra}}$ & $\hbox{WeaKL}_{\mathrm{agg}}$\\
        \hline
         MAE (GW) & 58.0 & 23.4 & 11.2 & 11.3 & 10.5\\
         \hline
    \end{tabular}
    \label{tab:gam-vik}
\end{table}
Thus, $\hbox{GAM}_{\mathrm{+}}$ refers to the offline GAM, while $\hbox{GAM}_{+,\mathrm{intra}}$ corresponds to the offline GAM with an intraday correction. Similarly, $\hbox{GAM}_{\mathrm{on}}$ represents the GAM-Viking model, and $\hbox{GAM}_{\mathrm{on, intra}}$ denotes the GAM-Viking model with an intraday correction. Finally, $\hbox{GAM}_{\mathrm{agg}}$ corresponds to the final model proposed by \citet{vilmarest2022state}.

The performance $\hbox{GAM}_{\mathrm{+}}$, $\hbox{GAM}_{\mathrm{+, intra}}$, $\hbox{WeaKL}_{\mathrm{+}}$, and $\hbox{WeaKL}_{\mathrm{+, intra}}$ in Table~\ref{tab:gam-vik} alone is not very meaningful because the distribution of electricity demand differs between the training and test periods. To address this, Table~\ref{tab:gam-vik-norm} presents a comparison of the same algorithms, trained on the same period but evaluated on a test period spanning from $1$ January $2020$ to $1$ March $2020$. In this stationary period, WeaKL outperforms the GAMs.
\begin{table}
    \centering
    \caption{Comparing GAM with its direct translation in the WeaKL framework on a stationary test period.}
    \begin{tabular}{|c|cccc|}
        \hline
        Model &  $\hbox{GAM}_{\mathrm{+}}$ &  $\hbox{WeaKL}_{\mathrm{+}}$ & $\hbox{GAM}_{+,\mathrm{intra}}$ & $\hbox{WeaKL}_{+,\mathrm{intra}}$\\
        \hline
         MAE (GW) & 20.7 & 19.1 & 19.3 & 19.2\\
         \hline
    \end{tabular}
    \label{tab:gam-vik-norm}
\end{table}

Moreover, in Table~\ref{tab:gam-vik}, the online WeaKLs clearly outperform the GAM-Viking models, achieving a reduction in MAE of more than $10\%$.
As a result, replacing the GAM-Viking models in the aggregation leads to improved overall performance. Notably, the WeaKLs are direct translations of the GAM-Viking models, meaning that the performance gains are due solely to model optimization and not to any structural changes.

\paragraph{Pure WeaKL.}
In addition, we trained an additive WeaKL using a different set of variables than those in the GAM model, aiming to identify an optimal configuration. Specifically, we consider the additive WeaKL with 
\begin{align*}
    X= (&\mathrm{FcloudCover\_corr1},\mathrm{Load1D},\mathrm{Load1W},\mathrm{DayType},\mathrm{FTemperature\_corr1},\\  &\mathrm{FWindDirection}, \mathrm{FTemps95\_corr1},\mathrm{Toy},\mathrm{t}),
\end{align*} where 
\begin{itemize}
    \item[$(i)$] the effects of $\mathrm{FclouCover\_corr1}$, $\mathrm{Load1D}$, and $\mathrm{Load1W}$ are nonlinear,
    \item[$(ii)$] the effect of $\mathrm{DayType}$ is categorical with 7 values,
    \item[$(iii)$] the remaining effects are linear.
\end{itemize}
This model is trained using the hyperparameter tuning process detailed in Appendix~\ref{sec:tuning}, with the training period spanning from $18$ March $2017$ to $1$ January $2020$, and validation starting from $1$ October $2019$.
Next, we fit an online WeaKL model, with hyperparameters tuned using a training period from $1$ March $2020$ to $18$ November $2020$ and a validation period extending until $18$ January $2021$.

To verify that our pure WeaKL model achieves a significantly lower error than the best model from the IEEE competition, we estimate the $\mathrm{MAE}$ skill score by comparing our pure WeaKL to the model proposed by \citet{vilmarest2022state}. To achieve this, we follow the procedure detailed in Appendix~\ref{sec:block-bootstrap}, using block bootstrap with a block length of $\ell = 24$ and $3000$ resamples to estimate the distribution of the $\mathrm{MAE}$ skill score, $\mathrm{Skill}$. Here, $\hat{Y}^1$ represents the WeaKL, while $\hat{Y}^2$ corresponds to the estimator from \citet{vilmarest2022state}. To evaluate the performance difference, we estimate the standard deviation $\sigma_n$ of $\mathrm{Skill}_n$ and construct an asymptotic one-sided confidence interval for $\mathrm{Skill}$. Specifically, we define
$\mathrm{Skill}_n = 1 - (\sum_{j=1}^n |\hat Y^1_{t_j} - Y_{t_j}| )/(\sum_{j=1}^n |\hat Y^2_{t_j} - Y_{t_j}|)$
and consider the confidence interval $[\mathrm{Skill}_n - 1.28 \sigma_n, +\infty[$, which corresponds to a confidence level of $\alpha = 0.1$. The resulting interval, $[0.007, +\infty[$, indicates that the $\mathrm{Skill}$ score is positive with at least $90\%$ probability. Consequently, with at least $90\%$ probability, the WeaKL  chieves a lower $\mathrm{MAE}$ than the best model from the IEEE competition.

\subsection*{Precision on the use Use case 2 on forecasting the French electricity load during the energy crisis}
\label{sec:sobriety}
This appendix provides detailed information on the additive WeaKL and the online WeaKL used in Use case 2, which focuses on forecasting the French electricity load during the energy crisis.
\paragraph{Additive WeaKL.} As detailed in the main text, the additive WeaKL is built using the following features:
\[X =(\mathrm{Load}_1, \mathrm{Load}_7, \mathrm{Temp}, \mathrm{Temp}_{950},  \mathrm{Temp}_{\mathrm{max 950}}, \mathrm{Temp}_{\mathrm{min 950}}, \mathrm{ToY},  \mathrm{DoW}, \mathrm{Holiday},t).
\]

The effects of $\mathrm{Load}_1$, $\mathrm{Load}_7$, and $t$ are modeled as linear. The effects of $\mathrm{Temp}$, $\mathrm{Temp}_{950}$,  $\mathrm{Temp}_{\mathrm{max 950}}$, $ \mathrm{Temp}_{\mathrm{min 950}}$, and $\mathrm{ToY}$ are modeled as nonlinear with $m=10$. The effects of $\mathrm{DoW}$ and $\mathrm{Holiday}$ are treated as categorical, with $|E| = 7$ and $|E| = 2$, respectively. The model weights are selected through hyperparameter tuning, as detailed in Appendix~\ref{sec:tuning}. The training period spans from $8$ January $2013$ to $1$ September $2021$, while the validation period covers $1$ September $2021$ to $1$ September $2022$. Notably, this is the exact same period used by \citet{doumeche2023human} to train the GAM. The objective of the hyperparameter tuning process is to determine the optimal values for $\lambda = (\lambda_1, \hdots, \lambda_{10}) \in (\mathbb R^+)^{10}$ and $\mathbf{m} = (m_3, m_4, m_5, m_6, m_7) \in (\mathbb N^\star)^5$ in \eqref{eq:weaklGAM}. As a result, the additive WeaKL model presented in Use case 2 is the outcome of this hyperparameter tuning process.

\paragraph{Online WeaKL.} Next, we train an online WeaKL to update the effects of the additive WeaKL. To achieve this, we apply the hyperparameter selection technique detailed in Appendix~\ref{sec:tuning}. The training period spans from $1$ February $2018$ to $1$ April $2020$, while the validation period extends from $1$ April $2020$ to $1$ June $2020$. These periods, although not directly contiguous to the test period, were specifically chosen because they overlap with the COVID-19 outbreaks. This is crucial, as it allows the model to learn from a nonstationary period. Moreover, since online models require daily updates, the online WeaKL is computationally more expensive than the additive WeaKL. The training period is set to two years and two months, striking a balance between computational efficiency and GPU memory usage. Using the parameters $(\lambda, \mathbf m)$ obtained from hyperparameter tuning, we then retrain the model in an online manner with data starting from $1$ July $2020$, ensuring that the rolling training period remains at two years and two months.

\paragraph{Error quantification.} Following the approach of \citet{doumeche2023human}, the standard deviations of the errors are estimated using stationary block bootstrap with a block length of $\ell = 48$ and 1000 resamples.

\paragraph{Model running times.} Below, we present the running times of various models in the experiment that includes holidays:
\begin{itemize}
\item  GAM: $20.3$ seconds. 
\item Static Kalman adaption: $1.7$ seconds.
\item Dynamic Kalman adaption: $48$ minutes, for an hyperparameter tuning of $10^4$ sets of hyperparameters \citep[see][II.A.2]{obst2021adaptative}.
\item Viking algorithm: $215$ seconds (in addition to training the Dynamic Kalman model).
\item Aggregation: $0.8$ seconds.
\item GAM boosting model: $6.6$ seconds.
\item Random forest model: $196$ seconds.
\item Random forest + bootstrap model: $34$ seconds.
\item Additive WeaKL: grid search of $1.6\times 10^5$ hyperparameters: $257$ seconds; training a single model: $2$ seconds.
\item Online WeaKL: grid search of $9.2\times 10^3$ hyperparameters: $114$ seconds; training a single model: $52$ seconds.
\end{itemize}
\subsection*{Precisions on the use case on hierarchical forecasting of Australian domestic tourism with transfer learning}
\label{sec:hierarchical_details} The matrices $\Lambda$ for the WeaKL-BU, WeaKL-G, and WeaKL-T estimators are selected through hyperparameter tuning. Following the procedure detailed in Appendix~\ref{sec:tuning}, the dataset is divided into three subsets: training, validation, and test. The training set comprises the first $60\%$ of the data, the validation set the next $20\%$, and the test set the last $20\%$. 
The optimal matrix, $\Lambda_{\mathrm{opt}}$, is chosen from a set of candidates by identifying the estimator trained on the training set that achieves the lowest MSE on the validation set. The model is then retrained using both the training and validation sets with $\Lambda = \Lambda_{\mathrm{opt}}$, and its performance is evaluated on the test set. Given that $d_1 = 415 \times 24 = 19,920$, WeaKL involves matrices of size $d_1^2 \simeq 4\times 10^8$, requiring several gigabytes of RAM. Consequently, the grid search process is computationally expensive. For instance, in this experiment, the grid search over $1024$ hyperparameter sets for WeaKL-T takes approximately $45$ minutes.

\renewcommand\thesection{\thechapter.\arabic{section}}

\part{Conclusion}
\label{part:conclusion}
%



\paragraph{Main contributions.}
In this thesis, we have proven several results on the statistical properties of physics-informed machine learning.
Chapter~\ref{ch:my-domain} focuses on the theoretical properties of physics-informed neural networks (PINNs), showing their risk consistency for linear and nonlinear PDE systems, and their strong convergence for linear PDE systems.
In Chapter~\ref{ch:requirements}, we prove that PDE solving and hybrid modeling  tasks with linear PDEs can be reframed as kernel methods, which makes it possible to characterize the impact of the physics on the convergence rate.
In Chapter~\ref{ch:related-work}, we introduce an algorithm to efficiently implement the kernel method of Chapter~\ref{ch:related-work} on GPUs.
We show that this kernel method outperforms PINNs on two PDE solving tasks.
Then, we have focused on forecasting energy signals during atypical periods.
In Chapter~\ref{ch:contrib-1}, we present the results of the Smarter Mobility Data Challenge on forecasting the occupancy of electric vehicle charging stations.
In Chapter~\ref{ch:contrib-2}, we study the integration of mobility data to forecast the French electricity demand during COVID.
Finally, in Chapter~\ref{ch:contrib-3}, we adapt the kernel framework of Chapter~\ref{ch:related-work} to common linear constraints in time series forecasting, and show that these new kernel methods outperform the state-of-the-art in electricity demand and tourism forecasting.
Many questions remain open regarding the topics developed in this thesis.

\paragraph{Convergence of PINNs.} First, the convergence of physics-informed machine learning algorithms, and in particular of PINNs, is not well-established for nonlinear PDEs.
In \citet{doumeche2023convergence}, we proved the risk-consistency of the PINNs estimator, but we warned that this does not necessarily mean that the resulting PINN indeed satisfy the penalized PDE.
This is due to the fact that the theoretical risk function is not weakly continuous with respect to the Sobolev norm.
Some results have been proven in PDE solving, for example regarding the Navier-Stokes equation, but they mainly focus on modifications of the PINNs algorithm.
For instance, \citet{deryck2023error} consider neural networks which weights are bounded during the gradient descent, and then let the bound on the weights grow, which disregards the overfitting scenarios that we pointed out in Chapter~\ref{ch:my-domain}.
Moreover, most theoretical papers consider to have at hand an exact minimizer of the empirical risk, while it is known that the gradient descent of PINNs behaves badly, especially when the penalized PDE is nonlinear \citep{wang2022when,bonfanti2024challengesnonlinearregimephysicsinformed}.

\paragraph{Curse of the dimension.} Regarding the theory of PINNs, the impact of the dimension $d_1$ on the theoretical performance of PINNs remains unknown.
Indeed, many machine learning algorithms are known to become extremely computationally expensive as the dimension $d_1$ grows, while becoming less accurate.
For example, our Fourier kernel method has a complexity of $(2m+1)^{2d_1}n+ (2m+1)^{3d_1}$, which scales exponentially in $d_1$, while the Sobolev minimax rate $n^{-2s/(2s+d_1)}$ worsens exponentially in $d_1$.
This is known as the curse of the dimension.
Meanwhile, neural networks are known to be able to overcome this issue when the data $X$ lies in a submanifold of $\mathbb R^{d_1}$ of lower dimension \citep[see, e.g.,][Section 2.6]{bach2024learning}.
While some experiments suggest that PINNs also alleviate the curse of the dimension \citep{hu2024tackling}, it has not been proven formally.

\paragraph{Minimax convergence rates of hybrid modeling tasks.} 
In Chapter~\ref{ch:my-domain}, we show that the impact of linear PDEs on the convergence rate in hybrid modeling tasks can be quantified by the impact on the effective dimension of the kernel methods.
In simple cases, we managed to bound the effective dimension to show that the physical prior speeds-up the learning process.
In Chapter~\ref{ch:requirements}, we show how to approximate the effective dimension for more complex linear PDEs. 
Similarly, for second-order linear elliptic PDEs in dimension $d=2$, and for domains with $C^2$ boundary $\partial \Omega$, \citet{azzimonti2015blood} and \citet{arnone2022spatialRegression} showed that the hybrid modeling task outperform the Sobolev minimax rate.
However, further studies should be made to determine the impact of both the PDE and the smoothness of the domain $\Omega$ on the convergence rate, in particular for classical linear and nonlinear PDEs.

\paragraph{Adapting our kernel methods to more contexts.} 
The kernel methods we developed are tailored to PDE solving and hybrid modeling tasks.
Their main advantage is that they admit closed-form formulas for the minimizer of the empirical risk, which can be efficiently implemented on GPUs.
However, to tackle high dimensional scenario, the complexity of the optimization should be reduced, and many usual techniques like gradient descent, conjugate gradient \citep{blanchard2010optimal}, or the Nyström method \citep{yang2012nystrom}.
Moreover, adapting our PIKL algorithm to handle nonlinear PDEs, and developing similar kernel methods for PDE learning and operator learning are all promising avenues of research.  

%
{%
\setstretch{1.1}
\renewcommand{\bibfont}{\normalfont\small}
\setlength{\biblabelsep}{0pt}
\setlength{\bibitemsep}{0.5\baselineskip plus 0.5\baselineskip}
\printbibliography[nottype=online]
}
\cleardoublepage







\newpage 

\end{document}